\documentclass[12pt, oneside]{book} 
\usepackage{graphicx}
\usepackage{KUstyle}

\usepackage[round]{natbib}
\usepackage{bibentry}
\nobibliography{references_all}

\usepackage{tabularx} 

\usepackage{nicefrac}       
\usepackage{bm}             
\usepackage{makecell}       
\usepackage{multicol}       
\usepackage{tcolorbox}      
\usepackage[normalem]{ulem} 
\usepackage{mdframed}       
\usepackage{rotating}       

\usepackage{verbatim}       
\usepackage{tablefootnote}  

\usepackage{microtype}      

\newenvironment{abstract}{%
  \section*{Abstract}%
}{}

\usepackage{CJKutf8}

\usepackage{amsmath}
\usepackage[T5,T1]{fontenc}    
\usepackage{url}            
\usepackage{booktabs}       
\usepackage{amsfonts}       
\usepackage{amsthm} 
\usepackage{yhmath}
\usepackage{dsfont}
\usepackage{mathtools}
\usepackage{amssymb}

\usepackage{pdflscape}

\DeclareTextSymbolDefault{\OHORN}{T5}
\DeclareTextSymbolDefault{\UHORN}{T5}
\DeclareTextSymbolDefault{\ohorn}{T5}
\DeclareTextSymbolDefault{\uhorn}{T5}

\usepackage{amsfonts}       
\usepackage[nameinlink]{cleveref}
\usepackage{bbm}
\usepackage{enumerate}
\usepackage{enumitem,kantlipsum}
\usepackage{xspace}
\usepackage{xurl}
\usepackage{multirow}
\usepackage{caption}
\usepackage[export]{adjustbox}
\usepackage[utf8]{inputenc}
\usepackage{nameref,hyperref}
\usepackage{float}
\usepackage{lipsum}

\usepackage{thmtools} 
\usepackage{thm-restate}

\usepackage{comment}
\usepackage{tikz}

\usetikzlibrary{bayesnet,calc,matrix,positioning,shapes,backgrounds,shapes,decorations,arrows,calc,arrows.meta,fit,positioning}
\tikzset{main node/.style={circle,draw,minimum size=1cm}}
\usepackage{linguex}
\usepackage{xfrac}
\usepackage{listings}
\usepackage[dvipsnames]{xcolor, colortbl}    
\usepackage{soul}
\usepackage{tabularray}
\usepackage{siunitx}
\usepackage[colorinlistoftodos]{todonotes}

\reversemarginpar

\usepackage{subcaption}
\usepackage{placeins}

\newcolumntype{Y}{>{\raggedright\arraybackslash}X}

\newenvironment{fitwidthtable}[1][\small]{%
  \begingroup
  #1
  \setlength{\tabcolsep}{3pt}%
  \renewcommand{\arraystretch}{1.05}%
  \begin{adjustbox}{max width=\textwidth}%
}{%
  \end{adjustbox}%
  \endgroup
}

\usepackage{placeins}

\usepackage{lmodern}
\usepackage{pifont}

 \graphicspath{ {images/} }

\def\checkmark{\tikz\fill[scale=0.4](0,.35) -- (.25,0) -- (1,.7) -- (.25,.15) -- cycle;} 

\newcommand{\MI}{\mathrm{MI}}

\definecolor{MyPurple}{RGB}{149,43,96}
\definecolor{MyBlue}{RGB}{37,111,174}
\definecolor{OliveGreen}{RGB}{117,157,139}
\definecolor{MyOrange}{RGB}{192,145,25}

\definecolor{greencolorname}{rgb}{0.4, 0.76, 0.65}
\definecolor{lightgreencolorname}{rgb}{0.65, 0.85, 0.33}
\definecolor{orangecolorname}{rgb}{0.99, 0.55, 0.3}
\definecolor{purplecolorname}{rgb}{0.55, 0.63, 0.80}
\definecolor{pinkcolorname}{rgb}{1.0, 0.33, 0.64}

\definecolor{deepcolorA}{rgb}{0.4, 0.76, 0.65}
\definecolor{deepcolorB}{rgb}{0.99, 0.55, 0.3}
\definecolor{deepcolorC}{rgb}{0.55, 0.63, 0.80}

\definecolor{mlp1}{rgb}{0.7, 0.7, 0.7}
\definecolor{mlp2}{rgb}{1, 0.85, 0.18}
\definecolor{darkblue}{rgb}{0, 0, 0.5}

\Crefname{figure}{Figure}{Figures}
\Crefname{table}{Table}{Tables}
\Crefname{equation}{Equation}{Equations}

\crefname{section}{Section}{Sections}
\crefname{algorithm}{Alg.}{}
\crefname{appendix}{Appendix}{}
\crefname{chapter}{Ch.}{Chs.}
\Crefname{chapter}{Chapter}{Chapters}

\crefname{theorem}{Theorem}{Theorems}
\crefname{prop}{Proposition}{Propositions}
\crefname{cor}{Corollary}{}
\crefname{observation}{Observation}{}
\crefname{assumption}{Assumption}{}
\crefname{hypothesis}{Hyp.}{Hypotheses}
\crefformat{section}{Section #2#1#3}

\newcommand*{\nameadjunct}{\relax}
\makeatletter
\renewcommand*{\NAT@nmfmt}[1]{\NAT@up #1\nameadjunct}
\makeatother

\newtheorem*{lemma*}{Lemma}

\newcolumntype{P}[1]{>{\raggedright\arraybackslash}p{#1}}
\newcolumntype{s}[1]{>{\sloppy\arraybackslash}p{#1}}

\newcommand{\circa}{{\raise.17ex\hbox{$\scriptstyle\sim$}}}

\usepackage{array}
\newcolumntype{L}[1]{>{\raggedright\let\newline\\\arraybackslash\hspace{0pt}}m{#1}}
\newcolumntype{C}[1]{>{\centering\let\newline\\\arraybackslash\hspace{0pt}}m{#1}}
\newcolumntype{R}[1]{>{\raggedleft\let\newline\\\arraybackslash\hspace{0pt}}m{#1}}

\newcommand{\tokenex}{\texttt{TokenEx}}
\newcommand{\tokenintex}{\texttt{TokenIntEx}}
\newcommand{\spanintex}{\texttt{SpanIntEx}}

\useunder{\uline}{\ul}{}
\restylefloat{table}
\newcommand{\bestcomp}[1]{\underline{#1}}
\newcommand{\bestsuff}[1]{\textbf{#1}}
\newcommand{\subj}[1]{\textcolor{blue}{#1}}
\newcommand{\gold}[1]{\textcolor{orange}{#1}}
\newcommand{\noise}[1]{\textcolor{green}{#1}}
\usetikzlibrary{shapes}
\newcommand{\rankk}{$Rank@k$}
\newcommand{\propk}{$Prop@k$}
\newcommand{\scorek}{$Score@k$}
\newcommand{\Drankk}{$\Delta Rank@k$}
\newcommand{\Drank}{$\Delta Rank$}
\newcommand{\mrr}{$MRR$}
\newcommand{\mrre}{MRR}
\newcommand{\rankke}{Rank@k}
\newcommand{\Drankke}{\Delta Rank@k}
\newcommand{\rankkeInst}{\rankke^{\mathrm{inst}}}
\newcommand{\rankkInst}{$\rankke^{\mathrm{inst}}$}
\newcommand{\DrankkeGroup}{\Drankke^{\mathrm{grp}}}
\newcommand{\DrankkeInst}{\Drankke^{\mathrm{inst}}}
\newcommand{\DrankkGroup}{$\Drankke^{\mathrm{grp}}$}
\newcommand{\DrankkInst}{$\Drankke^{\mathrm{inst}}$}

\renewcommand{\MI}{MechLight}
\newcommand{\FA}{FA}
\newcommand{\IG}{IG}
\newcommand{\Attn}{ATTN}

\AtBeginEnvironment{blockquote}{\small}
\definecolor{block-gray}{gray}{0.85}
\AtBeginEnvironment{definition}{\small}

\definecolor{LightSteelBlue1}{RGB}{202,225,255}
\definecolor{LightPink}{RGB}{245,191,210}
\definecolor{Moccasin}{RGB}{255, 228, 181}

\def\method{\textsc{\texttt{G-TEx}}\xspace}
\def\gcn{\textsc{\texttt{TEx-GCN}}\xspace}
\def\gat{\textsc{\texttt{TEx-GAT}}\xspace}
\def\sage{\textsc{\texttt{TEx-SAGE}}\xspace}

\newcommand{\modelblocksep}{\addlinespace[2pt]\cmidrule(lr){1-4}\addlinespace[2pt]}
\newcommand{\modelname}{CLUE}
\newcommand{\modelvariantone}{CLUE-Span}
\newcommand{\modelvarianttwo}{CLUE-Span+Steering}
\newcommand{\baselinemodelname}{Prompt\textsubscript{Baseline}}
\newcommand{\modelvariantoneshortened}{CLUE-S}
\newcommand{\modelvarianttwoshortened}{CLUE-Steer}
\newcommand{\baselinemodelnameshortened}{Prompt\textsubscript{Base}}
\newcommand{\ModelName}{\textsc{CLUE}}

\newcommand{\lm}[1]{\texttt{#1}}

\newcommand{\data}[1]{\textsf{#1}}

\newcommand{\our}{\textbf{\texttt{FaithMATE}}\xspace}

\definecolor{fm-lightblue}{RGB}{239,247,250}
\definecolor{fm-blue}{RGB}{224,230,255}
\definecolor{fm-lighpurple}{RGB}{226, 218, 246}
\definecolor{fm-purple}{RGB}{193, 175, 236}
\definecolor{fm-deepred}{RGB}{183,26,26}
\definecolor{fm-deepgreen}{RGB}{4,98,10}
\definecolor{fm-lightred}{RGB}{242,207, 194}

\makeatletter
\newcommand*\iftodonotes{\if@todonotes@disabled\expandafter\@secondoftwo\else\expandafter\@firstoftwo\fi} 
\makeatother

\crefname{section}{\S}{\S\S} 
\Crefname{section}{\S}{\S\S} 
\crefname{table}{Tab.}{Tables}
\crefname{figure}{Fig.}{Figures}
\crefname{algorithm}{Algorithm}{}
\crefname{equation}{eq.}{}
\crefname{appendix}{App.}{}
\crefname{lstlisting}{listing}{listings}
\Crefname{lstlisting}{Listing}{Listings}
\crefformat{section}{\S#2#1#3} 

\definecolor{KUPetrol}{RGB}{0,120,148} 

\definecolor{KUBlue}{RGB}{33,92,175} 
\definecolor{KUGreen}{RGB}{98,115,19} 
\definecolor{KUPurpleDark}{RGB}{140,10,89} 
\definecolor{KUPurple}{RGB}{163,7,116} 
\definecolor{KUGray}{RGB}{111,111,111} 
\definecolor{KURed}{RGB}{183,53,45} 
\definecolor{KUPetrol}{RGB}{0,120,148} 
\definecolor{KUBronze}{RGB}{142,103,19} 

\newtoggle{color-macro}
\settoggle{color-macro}{true} 

\iftoggle{color-macro}{\colorlet{MacroColor}{KUPetrol}
}{
\colorlet{MacroColor}{black}
}

\iftoggle{color-macro}{
\colorlet{TokenColor}{KUBronze}
}{
\colorlet{TokenColor}{black}
}

\iftoggle{color-macro}{
\colorlet{MathSubColor}{KUPurple}
}{
\colorlet{MathSubColor}{black}
}

\iftoggle{color-macro}{
\colorlet{RedditColor}{black}
}{
\colorlet{RedditColor}{black}
}

\iftoggle{color-macro}{
\colorlet{SchwartzProbColor}{KUBlue}
}{
\colorlet{SchwartzProbColor}{black}
}

\iftoggle{color-macro}{
\colorlet{IColor}{KUGray}
}{
\colorlet{IColor}{black}
}

\usepackage[T1]{fontenc}

\usepackage{amsmath,amsfonts,bm}

\def\eqref#1{equation~\ref{#1}}

\def\1{\bm{1}}

\DeclareMathAlphabet{\mathsfit}{\encodingdefault}{\sfdefault}{m}{sl}
\SetMathAlphabet{\mathsfit}{bold}{\encodingdefault}{\sfdefault}{bx}{n}

\definecolor{codegreen}{rgb}{0,0.6,0}
\definecolor{codegray}{rgb}{0.5,0.5,0.5}
\definecolor{codepurple}{rgb}{0.58,0,0.82}
\definecolor{backcolour}{rgb}{0.97,0.97,0.95}

\lstdefinestyle{mystyle}{
    backgroundcolor=\color{backcolour},   
    commentstyle=\color{codegreen},
    keywordstyle=\color{magenta},
    numberstyle=\tiny\color{codegray},
    stringstyle=\color{codepurple},
    basicstyle=\ttfamily\footnotesize,
    breakatwhitespace=true,         
    breaklines=true,                 
    captionpos=b,                    
    keepspaces=true,                 
    numbers=left,                    
    numbersep=5pt,                  
    showspaces=false,                
    showstringspaces=false,
    showtabs=false,                  
    tabsize=2
}

\ptype{Ph.D. Thesis}
\author{Jingyi Sun}

\title{Enhancing Understanding of \\ Language Models\\
with Varied Explanations}
\subtitle{}
\vspace{2.5cm}
\advisor{Advisors: Prof. Isabelle Augenstein and Dr. Pepa Atanasova}
\date{This thesis has been submitted to the Ph.D. School of The Faculty of Science, the University of Copenhagen on July 31st, 2026.}

\usepackage{titlesec}
\usepackage{fancyhdr}

\fancypagestyle{plain}{         
\fancyhf{}
\fancyfoot[C]{\thepage}}%

\renewcommand{\headrulewidth}{0pt}
\renewcommand{\chaptermark}[1]{\markboth{#1}{}}
\newcommand\mymainpagestyle{%
\fancyhf{}      
\fancyhead[L]{\nouppercase{\footnotesize{\chaptername~ \thechapter~ |~ \leftmark}} \renewcommand{\headrulewidth}{0.4pt} \headrule \renewcommand{\headrulewidth}{0pt}}

\setlength{\headheight}{25pt}

\fancyfoot[C]{\thepage}
}

\newcommand\mymiscpagestyle{%
\fancyhf{}      
\fancyhead[L]{\nouppercase{\footnotesize{\leftmark}} \renewcommand{\headrulewidth}{0.4pt} \headrule \renewcommand{\headrulewidth}{0pt}}
\setlength{\headheight}{25pt}
\fancyfoot[C]{\thepage}
}

\begin{document}


\maketitle
\frontmatter 
\pagestyle{plain} 

\newpage 

\begin{tikzpicture}[remember picture, overlay]
    \node[
        anchor=north east,
        xshift=-2.2cm,
        yshift=-2.4cm,
        text width=8cm,
        align=right,
        inner sep=0pt,
        font=\normalfont\normalsize
    ] at (current page.north east) {%
        \begin{CJK}{UTF8}{gbsn}
            {\itshape
            ``不识庐山真面目，\\
            只缘身在此山中。''\par}

            \vspace{0.35cm}
            {--- 苏轼\par}
        \end{CJK}

        \vspace{0.8cm}

        {\itshape
        ``The true nature of things reveals itself\\
        only to those who seek more than a single view.''\par}

        \vspace{0.35cm}
        {--- Su Shi\par}
    };
\end{tikzpicture}

\newpage 

\section*{Abstract}
\label{sec:abstract}
\addcontentsline{toc}{section}{Abstract} 

Language models are increasingly used to support human judgement in complex reasoning tasks, yet their predictions often remain difficult to interpret and validate. Explainability techniques aim to provide users with insight into model decision-making, helping expose errors, failure modes, and potentially biased behaviour while giving stakeholders such as end users a basis for deciding whether and how to rely on model outputs. 
However, explanations are, by nature, simplified and potentially imprecise approximations of a language model's internal workings.
The goal is therefore to obtain explanations that reflect the underlying computation as closely as possible (faithfulness), appear understandable to users (plausibility), and help them achieve their intended goals (utility).
Accordingly, evaluation protocols are essential for assessing explanation quality across all these dimensions.


The complexity of language models, their decision processes, and the textual inputs they must interpret has motivated diverse forms of explanations, including the three studied in this thesis: highlight explanations, natural language explanations, and chain-of-thought reasoning.  These forms offer complementary perspectives on model behaviour and exhibit distinct strengths and limitations across tasks. At the same time, producing explanations that satisfy multiple quality dimensions remains challenging, as does designing reliable methods for evaluating these dimensions across explanation forms and application settings.

This thesis presents a body of research that collectively aims to improve people's understanding of models' decision-making processes in complex reasoning tasks. First, it contributes to the area of explainability diagnostics by proposing two evaluation frameworks for assessing the quality of highlight explanations across multiple properties and diverse highlight explanation types, together with a meta-evaluation framework for examining the reliability of existing faithfulness metrics for chain-of-thought explanations. Second, in the area of improving the transparency of language models, this thesis proposes two explainability techniques that generate natural language explanations designed to be faithful to the model's prediction process while remaining concise, coherent, and helpful for humans in understanding model decisions. Finally, in the area of human--AI interaction, this thesis presents a user study that explores the roles of evidence and explanations in helping users validate model verdicts in a fact-checking setting.



\newpage
\section*{Resumé}
\label{sec:resume}
\addcontentsline{toc}{section}{Resumé}
Sprogmodeller anvendes i stigende grad til at understøtte menneskelige vurderinger i komplekse ræsonneringsopgaver, men deres forudsigelser er ofte fortsat vanskelige at fortolke og validere. Forklarbarhedsteknikker har til formål at give brugere indsigt i modellernes beslutningsprocesser og dermed bidrage til at afdække fejl, svigtmønstre og potentielt biaspræget adfærd, samtidig med at de giver interessenter såsom slutbrugere et grundlag for at afgøre, om og hvordan de bør forlade sig på modellernes output.
Forklaringer er imidlertid af natur forenklede og potentielt upræcise tilnærmelser til en sprogmodels interne funktionsmåde.
Målet er derfor at opnå forklaringer, der afspejler den underliggende beregning så præcist som muligt (trofasthed; \textit{faithfulness}), fremstår forståelige for brugerne (plausibilitet; \textit{plausibility}) og hjælper dem med at nå deres tilsigtede mål (nytteværdi; \textit{utility}). Evalueringsprotokoller er derfor afgørende for at kunne vurdere forklaringskvaliteten på tværs af alle disse dimensioner.


Kompleksiteten i sprogmodeller, deres beslutningsprocesser og de tekstlige input, som de skal fortolke, har motiveret udviklingen af forskellige forklaringsformer, herunder de tre, der undersøges i denne afhandling: fremhævelsesbaserede forklaringer (\textit{highlight explanations}), forklaringer i naturligt sprog (\textit{natural language explanations}) og ræsonnementskæder (\textit{chain-of-thought reasoning}). Disse forklaringsformer giver komplementære perspektiver på modeladfærd og har forskellige styrker og begrænsninger på tværs af opgaver. Samtidig er det fortsat udfordrende både at frembringe forklaringer, der opfylder flere kvalitetsdimensioner, og at udvikle pålidelige metoder til at evaluere disse dimensioner på tværs af forklaringsformer og anvendelsessammenhænge.

Denne afhandling præsenterer en række forskningsbidrag, der samlet har til formål at forbedre menneskers forståelse af modellers beslutningsprocesser i komplekse ræsonneringsopgaver. For det første bidrager afhandlingen til området for diagnostik af forklarbarhed ved at foreslå to evalueringsrammer til vurdering af kvaliteten af fremhævelsesbaserede forklaringer med hensyn til flere egenskaber og på tværs af forskellige typer af sådanne forklaringer samt en meta-evalueringsramme til undersøgelse af pålideligheden af eksisterende mål for trofasthed i forklaringer baseret på ræsonnementskæder. For det andet foreslår afhandlingen, som et bidrag til arbejdet med at forbedre sprogmodellers gennemsigtighed, to forklarbarhedsteknikker, der genererer forklaringer i naturligt sprog, som er udformet med henblik på trofast at afspejle modellens forudsigelsesproces og samtidig være kortfattede, sammenhængende og nyttige for mennesker, når de skal forstå modellens beslutninger. Endelig præsenterer afhandlingen inden for menneske--AI-interaktion en brugerundersøgelse, der undersøger, hvilken rolle belæg og forklaringer spiller i at hjælpe brugere med at validere modellens afgørelser i en faktatjekningskontekst.


\newpage
\section*{Acknowledgements}
\label{sec:acks}
\addcontentsline{toc}{section}{Acknowledgements}

This PhD journey would not have been possible without the support of many incredible people. I would like to take this opportunity to express my sincere gratitude to all of them.

First, I could not express thanks enough to my amazing supervisors, Isabelle Augenstein and Pepa Atanasova. I would first like to thank Isabelle, whose incisive feedback has sharpened my thinking and whose high standards have inspired me to pursue research with rigour and ambition. Beyond research, her care, integrity, and fairness have made me feel genuinely supported and valued over the past three years. My heartfelt gratitude also goes to Pepa, whose extraordinary research intuition, constant encouragement, and unwavering support have carried me through this journey, through sunshine and rain, and given me strength when I needed it most. I feel incredibly lucky to have them as my supervisors.

I am sincerely grateful to Hinrich Schütze, Yonatan Belinkov, and Maria Maistro for serving on my PhD assessment committee and for the time and effort they have devoted to evaluating this thesis. My gratitude also goes to the European Union (ERC, ExplainYourself, 101077481) and the Pioneer Centre for AI, DNRF grant number P1,  for funding my research.

I am deeply grateful to my collaborators, Greta, Sekh, Qianli, Shuzhou, Lucas, and Ran, whose fresh perspectives and valuable insights made our work together both rewarding and memorable. I am equally grateful to their supervisors, Sagnik, Irina, Nils, Michael, and Steffen, for their guidance and support throughout our collaboration. I hope we will have more opportunities to collaborate in the future.

I would also like to thank my master’s supervisor, Professor Xu, and my senior colleague, Dr. Jianzhu Bao, for introducing me to research and helping shape my early understanding of it.

Furthermore, I am grateful to my CopeNLU colleagues: Greta, Haeun, Sekh, Sara, Nadav, Arnav, Sarah, Zain, Siddesh, Yoona, Jean, Erik, Karolina, Dustin, and Yijun for their friendship and all the good times we shared, both inside and outside the lab. Their warmth, unique perspectives, and openness to different cultures have made this journey both enriching and enjoyable. I could not feel luckier to be part of this wonderful community.

My special thanks go to my partner, Qinglin Zhu, for helping me recognise my potential and find the strength to persevere through the obstacles along the way. Finally, I offer my heartfelt thanks to my mother, Huiling, and my father, Xinnian, for their steadfast belief in me throughout this journey.

\newpage
\tableofcontents
\newpage

\phantomsection
\pagestyle{plain}
\chapter*{List of Publications}
\addcontentsline{toc}{section}{List of Publications}


This thesis includes the following papers verbatim in chapters 2-7, in a lightly reformatted version.
For papers with a prepublication version archived on a preprint server and a subsequent final publication, both versions are listed below.

\begin{enumerate}

    \item \bibentry{sun-etal-2025-evaluating}.
    
    \textit{Prepublication version:}
    \bibentry{sun-etal-2024-evaluating-arxiv}.

    \item \bibentry{sun2026evaluation}.
    
    \textit{Prepublication version:}
    \bibentry{sun-etal-2025-context-utilisation-arxiv}.

    \item \bibentry{yuan-etal-2025-graph}.
    
    \textit{Prepublication version:}
    \bibentry{yuan-etal-2024-graph-arxiv}.

    \item \bibentry{sun-etal-2026-explaining}.
    
    \textit{Prepublication version:}
    \bibentry{sun-etal-2025-clue-arxiv}.

    \item \bibentry{sun2026investigatinginterplaycontextualparametric}.

    \item \bibentry{warren2026show}.
    
    \textit{Prepublication version:}
    \bibentry{warren-etal-2026-show-arxiv}.

\end{enumerate}

\newpage

\mainmatter 
\mymainpagestyle{} 
\part{Executive Summary}

\chapter{Executive Summary}
\label{chap:intro}

\definecolor{darkblue}{RGB}{0,35,90} 
\definecolor{darkgreen}{RGB}{0,70,35} 
\definecolor{pink}{RGB}{150,10,80} 
\definecolor{darkyellow}{RGB}{130,95,0} 

\section{Introduction}
\label{sec:thesis_introduction}
Language models have demonstrated rapidly improving capabilities across a wide range of tasks, enabling their use as tools for supporting human judgement in decision-making settings~\citep{brown2020language, achiam2023gpt4, centerforaisafety2026benchmark,mcfadyen2026inference}. However, in decision-sensitive and high-stakes scenarios, such as fact-checking, users are concerned not only with the model's final answer but also with how it is derived, so that they can decide whether and how to rely on it~\citep{ribeiro-etal-2016-trust,bansal2021whole,si-etal-2024-large}. 
Explainability techniques therefore seek to bridge opaque model behaviour and users' practical needs by presenting a model's reasoning process in human-interpretable forms. They have been studied to make model behaviour more inspectable~\citep{ribeiro-etal-2016-trust,zhao2024explainability}, expose potential failure modes~\citep{atanasova-etal-2020-diagnostic,lertvittayakumjorn2021explanation}, and calibrate users' reliance on model responses~\citep{bansal2021whole,vasconcelos_explanations_2023}. 

\begin{figure*}[!t]
    \centering
        \includegraphics[width=\linewidth]{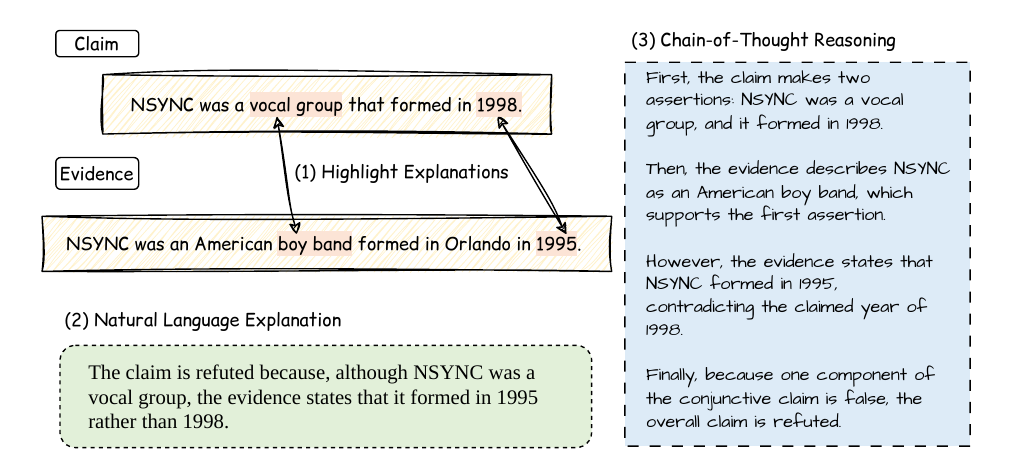}
\caption{Illustration of three explanation forms for the same FEVER fact-checking instance~\citep{thorne-etal-2018-fever}: Highlight Explanation, Natural Language Explanation, and Chain-of-Thought Reasoning. 
}
        \label{fig:three_explanation_form}
\end{figure*}

Different explanation forms provide complementary perspectives on model behaviour (see examples in Fig.~\ref{fig:three_explanation_form}). Highlight explanations identify salient input evidence~\citep{lei-etal-2016-rationalizing,deyoung-etal-2020-eraser}, natural language explanations provide coherent free-text rationales~\citep{camburu2018snli,atanasova-etal-2023-faithfulness}, and chain-of-thought reasoning elicits step-by-step reasoning traces before the model's prediction~\citep{wei2022chain,kojima2022large}. Besides, these different forms exhibit distinct limitations and serve different user needs and application scenarios~\citep{doshivelez2017towards,
lyu2024towards,madsen-etal-2024-self}.

Explanation evaluation plays an essential role in explainability research~\citep{zhao2024explainability}. 
Plausibility examines whether an explanation appears coherent and understandable by humans~\citep{jacovi-goldberg-2020-towards,hase-etal-2020-leakage}, utility directly measures whether it helps the user achieve an intended goal, such as making more accurate or appropriately calibrated decisions~\citep{doshivelez2017towards,hase-bansal-2020-evaluating}. However, these human-facing properties are insufficient: an explanation may be readable and persuasive while remaining misaligned with the model's underlying decision process,
thereby raising concerns about its faithfulness
~\citep{atanasova-etal-2023-faithfulness,agarwal2024faithfulnessvsplausibilityunreliability}.
Unfaithful explanations may give users unjustified confidence in model outputs, obscure failure modes, or mislead users about why a model made a decision~\citep{atanasova-etal-2023-faithfulness,agarwal2024faithfulnessvsplausibilityunreliability}. 
Therefore, plausibility, utility, and faithfulness constitute
complementary dimensions of explanation quality~\citep{lyu2024towards,zhao2024explainability}.

Taken together, explanation forms serve different purposes, and these quality dimensions are operationalised differently across explanation types and application settings. %
This variation suggests that no single explanation form or evaluation metric is sufficient across all tasks and applications. 
The thesis bridges diverse explainability techniques in two ways:
through novel evaluation benchmarks for
characterising them along multiple dimensions across tasks, and through novel methods that combine their strengths.
These contributions are guided by the principle that explanations
should be understandable and useful to users while also serving as
faithful indicators of model behaviour.

This section introduces explainability techniques and their core evaluation protocols across complex reasoning tasks in the era of language models. 
Section~\ref{intro:contributions} details how the published papers included in this thesis contribute to the broader goal of evaluating and improving explanations for language models.
Section~\ref{intro:limitations_future} summarises the contributions of the thesis and provides an introspective reflection on the limitations of each individual work, pointing to promising directions for future work.

\subsection{Explanations in Natural Language Processing}
\label{sec:concept_explanation_types}
Despite the growing performance of language models on a wide range of tasks, their underlying reasoning processes remain unclear to users. Explainability techniques aim to reveal the model's rationale behind its output by producing different forms of explanations, depending on the level of model behaviour being explained and the purpose for which the explanations are used~\citep{madsen2022posthoc,lyu2024towards,zhao2024explainability}. 
Recent work in explainability has evolved from highlight explanations, which select salient texts from the input, 
~\citep{li-etal-2016-visualizing,ribeiro-etal-2016-trust,atanasova-etal-2020-diagnostic,deyoung-etal-2020-eraser},
towards natural language explanations or intermediate reasoning traces that are free-form, structured and reasoning-oriented
~\citep{wei2022chain,NEURIPS2023_ed3fea90,wang-atanasova-2025-self,alon-etal-2026-faithful}.

In the following, I detail three commonly used forms of explanations which are further researched in this thesis: highlight explanations~\S\ref{sec:concept_highlight_explanations}, which are compact input segments that can serve as interface to convey model's prediction signals; natural language explanations~\S\ref{sec:concept_nle}, which are not constrained to the input text and more plausible for human understanding; and chain-of-thought explanations~\S\ref{sec:concept_cot}, which are usually more structured and express model's step-by-step reasoning.

\subsubsection{Highlight Explanations}
\label{sec:concept_highlight_explanations}

\begin{figure*}[!t]
    \centering
        \includegraphics[width=\linewidth]{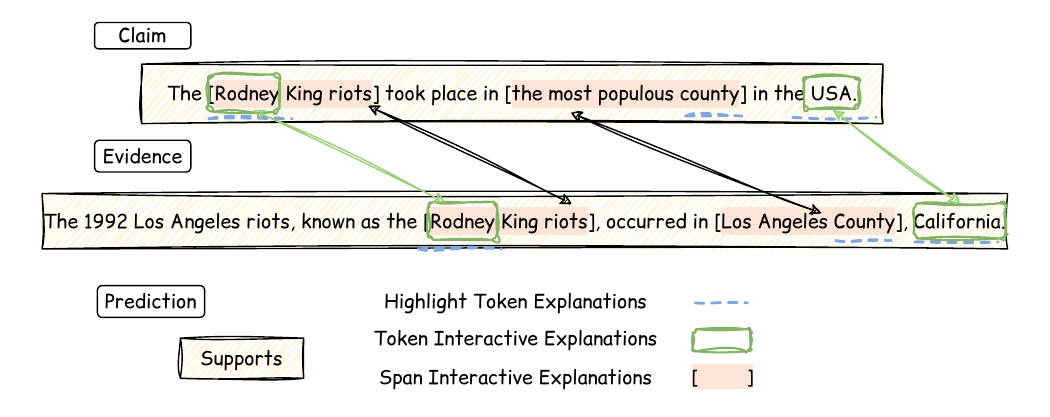}
\caption{Illustration of three types of Highlight Explanations for the same FEVER fact-checking instance~\citep{thorne-etal-2018-fever}: Highlight Token Explanations, Token Interactive Explanations, and Span interactive explanations. 
}
        \label{fig:three_highlight_explanation_form}
\end{figure*}

Highlight explanations, also called input feature explanations~\citep{sun-etal-2025-evaluating}, can identify important parts of the input text that are salient for the model's prediction~\citep{li-etal-2016-visualizing,deyoung-etal-2020-eraser,jacovi-goldberg-2021-aligning}. By visually linking a prediction to specific important parts of the input, they provide an intuitive way to communicate model behaviour to users~\citep{jacovi-goldberg-2021-aligning}. Highlight explanations are typically extracted using post-hoc explainability techniques and can therefore be applied broadly without modifying or retraining the predictive model.
Based on the granularity and structure of the highlighted information, they can be categorised into three major types: highlight token explanations, token interactive explanations, and span interactive explanations (See the illustrative examples in Fig.~\ref{fig:three_highlight_explanation_form}).

\textbf{Highlight Token Explanations.} 
The most common form of highlight explanation assigns an importance score to each input token.
\begin{equation}
    \mathbf{a} = (a_1,\ldots,a_n), \qquad a_i \in \mathbb{R},
\end{equation}
where $a_i$ denotes the estimated importance of input token $x_i$ to the model's prediction $\hat{y}$, and $n$ is the input length.

These importance scores can be estimated using several families of explainability techniques.
Perturbation-based methods delete, mask, or replace individual input tokens and measure the resulting change in model output as their estimated importance~\citep{li-etal-2016-visualizing,deyoung-etal-2020-eraser}. Gradient-based methods derive attributions from gradients with respect to input embeddings or intermediate representations~\citep{li-etal-2016-visualizing,sundararajan2017axiomatic}.
Local surrogate methods, such as LIME, approximate the model's local behaviour with a simple interpretable model~\citep{ribeiro-etal-2016-trust}. Shapley methods estimate each token's marginal contributions across coalitions of input tokens~\citep{shapley1953value,lundberg2017unified}. Attention weights have also been used for token attribution
~\citep{jain-wallace-2019-attention,serrano-smith-2019-attention,wiegreffe-pinter-2019-attention}. 
These methods have recently been adapted to decoder-only language models and multi-token outputs~\citep{sarti-etal-2023-inseq,pan2026flashtrace,pramanik2026heta}.

In the running example, the most important highlight tokens could be ``Rodney'', ``California'', and ``USA''. However, these individual tokens cannot capture interactions that jointly influence the prediction, such as the supporting correspondence between ``Rodney'' in the claim and its occurrence in the evidence~\citep{chen-etal-2020-generating-hierarchical,sikdar-etal-2021-integrated}. 
This limitation motivates interactive explanations which attribute importance to pairs or groups of tokens across input parts.

\paragraph{Token Interactive Explanations.}
Token interactive explanations represent pairwise token interactions as a score matrix:
\begin{equation}
    \mathbf{A} = (a_{ij})_{i,j=1}^{n}, \qquad a_{ij} \in \mathbb{R},
\end{equation}
where $a_{ij}$ denotes the estimated contribution of the interaction between token $x_i$ and token $x_j$. 

Gradient-based methods estimate these effects from second-order or path-integrated interaction between token embeddings, as in Integrated Hessians or Integrated Directional Gradients~\citep{NEURIPS2020_443dec30,janizek2021explaining,sikdar-etal-2021-integrated}. Subsequent work models these interactions as asymmetric relations, 
as the contribution of $x_j$ conditioned on $x_i$ may differ from the contribution of $x_i$ conditioned on $x_j$~\citep{lu-etal-2023-asymmetric}. 


In the running example, the most important pairs might include
$\langle\text{``Rodney''},\text{``Rodney''}\rangle$ and $\langle\text{``USA''},\text{``California''}\rangle$.
Although these explanations can capture interactions across different parts of the input, such isolated token pairs may lack semantic coherence. This motivates span interaction explanations, which capture interactions between multi-token spans rather than individual tokens.

\paragraph{Span Interactive Explanations.}
Span interactive explanations consist of pairs of contiguous text spans drawn from different parts of the input: 


\begin{equation}
    \mathcal{S} = \{(s_p, s_q, w_{pq})\}_{(p,q)\in\mathcal{I}},
\end{equation}

where $s_p$ and $s_q$ denote the paired spans and $w_{pq}$ denotes the estimated interaction importance, and $\mathcal{I}$ indexes the extracted span pairs. 

\citet{choudhury-etal-2023-explaining} construct a token-level interaction graph over the input and apply Louvain community detection to identify important span pairs. The resulting span interactions resemble human annotations, suggesting they can be useful in the complex reasoning tasks where interactions of input parts are important for the model. 

In the above running example, the most salient span interactive explanations might include
$\langle\text{``Rodney King riots''},\text{``Rodney King riots''}\rangle$ and $\langle\text{``the most populous county in the USA''},\text{``Los Angeles County''}\rangle$. By connecting semantically coherent phrases across the input parts, they represent model's reasoning more naturally than isolated token pairs.

Building on these explainability techniques, this thesis investigates the evaluation of highlight explanation. Chapter \ref{chap:highlight_diagnostic} introduces a framework for comparing these three forms of highlight explanation across multiple properties in complex reasoning tasks.
Chapter \ref{chap:highlight_context_utilisation} focuses on highlight token explanations and proposes a novel framework for assessing their utility in indicating model context utilisation in synthetic retrieval-augmented generation settings. 

\subsubsection{Natural Language Explanations}
\label{sec:concept_nle}
Natural language explanations (NLEs) are free-text rationales for a model's prediction that may contain information beyond the input text~\citep{miller2019explanation,camburu2018snli}. Their natural language form can make model prediction more accessible to users by providing justifications in natural language inference and commonsense question answering~\citep{camburu2018snli,yuan-etal-2025-graph}, explaining how the model draws on the evidence to verify the claim in fact-checking~\citep{atanasova-etal-2020-generating-fact,sun-etal-2026-explaining}, and communicate model uncertainty~\citep{yona-etal-2024-large,liu-etal-2025-metafaith}. 

NLE systems differ in the dependency between answer and explanation generation. Following \citet{lyu2024towards}, they can be broadly grouped into \emph{predict-then-explain}, \emph{explain-then-predict}, and jointly-predict-and-explain architectures. The first two orderings were explored in early NLE generation works, while the last category includes \emph{self-rationalisation} models that generate task predictions and rationales within a shared framework~\citep{marasovic-etal-2022-shot}.

In a \emph{predict-then-explain} architecture~\citep{camburu2018snli,rajani-etal-2019-explain}, the model first predicts an answer $\hat{y}$ from the input $x$ and then generates an explanation $z=(z_1,\ldots,z_m)$ conditioned on both:
\begin{equation}
    \hat{y} \sim p_{\theta_y}(\hat{y} \mid x), \qquad
    z \sim p_{\theta_z}(z \mid x,\hat{y}).
\end{equation}

In the reverse \emph{explain-then-predict} ordering~\citep{camburu2018snli,kumar2020nile}, an explanation is generated before being used to infer the label:
\begin{equation}
    z \sim p_{\theta_z}(z \mid x), \qquad
    \hat{y} \sim p_{\theta_y}(\hat{y} \mid x,z).
\end{equation}

A \emph{self-rationalisation} framework couples answer and explanation learning within a shared architecture~\citep{atanasova-etal-2020-generating-fact,pmlr-v162-majumder22a}, in which case predictions are generated together with NLEs. 
\begin{equation}
    (z,\hat{y}) \sim p_{\theta}(z,\hat{y} \mid x),
\end{equation}
Such coupling can improve answer--explanation consistency~\citep{kumar2020nile,pmlr-v162-majumder22a}. 


NLE quality can be characterised along two broad dimensions: faithfulness to the model's underlying reasoning and human-facing qualities such as plausibility and utility. 
These dimensions are discussed in detail in Sec.~\ref{sec:concept_faithfulness_nle} and Sec.~\ref{sec:concept_utility}.

Chapters \ref{chap:NLE_G-Tex} and \ref{chap:NLE_CLUE} develop self-rationalisation frameworks designed to improve faithfulness while preserving human-facing quality. Because highlight explanations provide compact cues about the input features associated with the model's prediction, they can serve as a bridge between model behaviour and NLEs. Based on these cues, Chapter \ref{chap:NLE_G-Tex} encodes the highlights in a graph structure and fine-tunes the model to generate more faithful NLEs; Chapter \ref{chap:NLE_CLUE} uses these highlights at inference time to guide the model in generating rationales that explain its uncertainty by explicitly referring to agreements and conflicts in the input.



\subsubsection{Chain-of-Thought Reasoning}
\label{sec:concept_cot}
Although also in free-text form, chain-of-thought (CoT) reasoning was introduced primarily to improve model performance by eliciting ``step-by-step'' reasoning before the final answer~\citep{wei2022chain,kojima2022large}. CoT prompting has substantially improved performance on arithmetic, commonsense, symbolic, and multi-step reasoning tasks~\citep{wei2022chain,kojima2022large,zhou2023least}.


As in self-rationalisation, CoT generation is coupled with answer generation. Given an input $x$, the model generates a reasoning trace $r=(r_1,\ldots,r_T)$ comprising intermediate steps towards the final answer $\hat{y}$:
\begin{equation}
    r \sim p_{\theta}(r \mid x), \qquad 
    \hat{y} \sim p_{\theta}(\hat{y} \mid x, r),
\end{equation}
Early work primarily elicited CoT through prompting~\citep{wei2022chain,kojima2022large,zhou2023least}, and the resulting CoTs can expose structured natural language traces that users can inspect.

CoT quality can be evaluated along several dimensions, including correctness, human-facing reasoning quality, and others. However, a coherent and plausible reasoning trace may still rationalise an answer post hoc rather than faithfully reflect the factors that influenced the prediction~\citep{NEURIPS2023_ed3fea90,lanham2023measuring,nguyen-etal-2024-direct}. For CoT to serve as a reliable explainability tool, its faithfulness must therefore be carefully assessed. 
Sec.~\ref{sec:concept_faithfulness_cot} provides a detailed discussion of CoT faithfulness evaluation and meta-evaluation.

Chapter~\ref{chap:CoT_FaithMate} proposes a meta-evaluation framework for studying existing CoT faithfulness metrics. It examines whether optimising a model against one metric yields improvements under other faithfulness metrics and evaluation paradigms, thereby testing whether these metrics capture a shared notion of faithfulness or distinct properties.

\subsection{Evaluating Faithfulness of Explanations} 
Faithfulness concerns how accurately an explanation reflects the model's underlying reasoning process~\citep{jacovi-goldberg-2020-towards,lyu2024towards}. It is distinct from plausibility: an explanation can be fluent, coherent, and persuasive to humans while still failing to capture the intermediate computations that influenced the model's output~\citep{NEURIPS2023_ed3fea90,parcalabescu-frank-2024-measuring}. 
This distinction is especially important for language models, whose persuasive but unfaithful explanations can obscure spurious correlations, unsupported evidence use, or other model failures, particularly in high-stakes settings~\citep{rudin2019stop,atanasova-etal-2023-faithfulness,sun2026investigatinginterplaycontextualparametric}.

Because the model's decision-making process is not directly observable, faithfulness is assessed through different proxies that
rely on different assumptions and might capture different aspects of faithfulness~\citep{atanasova-etal-2023-faithfulness,parcalabescu-frank-2024-measuring,tutek-etal-2025-measuring}. Their faithfulness conclusions can therefore vary across tasks and models, leaving Faithfulness an open validation problem~\citep{zaman-srivastava-2025-causal,sun2026investigatinginterplaycontextualparametric}.

Faithfulness is a core theme of this thesis. Across the contribution chapters, all except Chapter~\ref{chap:Human_showme} develop or apply methods for evaluating the faithfulness of
a specific explanation type, 
while Chapter~\ref{chap:NLE_G-Tex} and Chapter~\ref{chap:NLE_CLUE} also develop methods intended to improve it. The following subsections introduce common approaches to assessing and improving the faithfulness of highlight explanations, natural language explanations, and chain-of-thought reasoning.

\subsubsection{Faithfulness of Highlight Explanations}

Faithfulness of highlight explanations is commonly operationalised through behavioural intervention tests, particularly  
comprehensiveness and sufficiency~\citep{yu2019rethinking,deyoung-etal-2020-eraser}. 
Comprehensiveness examines whether ablating the highlight tokens can reduce the model's confidence in its original prediction, showing the explanation is not redundant for the model's answer; sufficiency tests whether only retaining the explanation text preserves the prediction, indicating it is sufficient for model's answer 
~\citep{deyoung-etal-2020-eraser}. 
Subsequent work has refined these tests by perturbing token representations to reduce unnatural inputs~\citep{zhao-aletras-2023-incorporating}, aggregating intervention effects using area over the perturbation curve (AOPC)~\citep{chen-etal-2021-explaining, choudhury-etal-2023-explaining}, and normalising  AOPC to enable cross-model and cross-instance comparisons~\citep{edin-etal-2025-normalized}.
Nevertheless, this line of approach has limitations:
interventions may shift the input distribution, remove useful context, or test only one kind of dependence between the explanation and the prediction~\citep{jacovi-goldberg-2020-towards, lyu2024towards}. 
Therefore, they should be regarded as specific diagnostic tests and considered alongside other explanation properties.

Chapter~\ref{chap:highlight_diagnostic} compares three forms of highlight explanations across several diagnostic properties and reveals a tension between faithfulness and agreement with human annotations, particularly for span interactive explanations.
Chapter~\ref{chap:highlight_context_utilisation} focuses on highlight tokens and finds they are not necessarily effective indicators of model context utilisation despite their high faithfulness. 
Together, these findings highlight the importance of scenario-specific evaluation across multiple explanation properties.

\subsubsection{Faithfulness of Natural Language Explanations}
\label{sec:concept_faithfulness_nle}

\paragraph{NLE faithfulness evaluation.}  
The free-text form of NLEs complicates their faithfulness assessment:
unlike highlight explanations, they cannot be directly removed from the input to test the effects on the model's prediction. 
A widely used approach is the counterfactual unfaithfulness test~\citep{atanasova-etal-2023-faithfulness}: when a perturbation on the input, e.g., inserting a word, changes the model's prediction, a faithful explanation should mention the inserted word responsible for that change.
Recent works extend this test by considering shifts in answer probability distribution rather than answer flips~\citep{siegel-etal-2024-probabilities}, applying concept-level interventions~\citep{matton2025walk}, or adapting it to closed-source LMs without access to answer probabilities~\citep{siegel-etal-2025-verbosity}.

However, these methods are argued to only measure model's self-consistency, 
as they do not assess whether the same internal computations are used for NLE and answer generation~\citep{parcalabescu-frank-2024-measuring}. 
Alternative approaches therefore assess whether the input attributions underlying answer and explanation generation aligns~\citep{parcalabescu-frank-2024-measuring}, or intervene on hidden states and concepts to compare their causal effects on these two outputs~\citep{yeo-etal-2025-towards,bhan2026neurofaithevaluatingllmselfexplanation}.
Nevertheless,  these methods usually depend on assumptions about which hidden states, concepts, or internal representations are relevant. 
More importantly, since there is no observable ground truth for model's reasoning, it still remains unclear whether current metrics measure faithfulness, output-level self-consistency, or narrower diagnostic properties~\citep{parcalabescu-frank-2024-measuring, lyu2024towards, zaman-srivastava-2025-causal}.

\paragraph{NLE faithfulness improvement.} 
Methods for improving NLE faithfulness primarily focus on strengthening the alignment between explanation and answer generation. 
Early works seek to improve this alignment through task-specific optimisation objectives~\citep{atanasova-etal-2020-generating-fact}, grounding NLEs in background knowledge~\citep{pmlr-v162-majumder22a}, and directly optimising multiple diagnostic properties~\citep{atanasova2022diagnostics}.
More recent approaches utilise self-refinement in which models revise their explanations using self-critique, attribution feedback, or verdict-anchored signals~\citep{wang-atanasova-2025-self,kong2026reflex}. Another line of work leverages preference optimisation, guiding the model to generate NLE based on contrasts between faithful and unfaithful explanation signals~\citep{zhao-iii-2025-necessary,admoni-etal-2026-aligning}. 
However, these methods depend on reliable feedback or faithfulness labels that are based on specific proxy metrics, and therefore do not necessarily guarantee alignment with the model’s internal computations~\citep{admoni-etal-2026-aligning}. 

Chapters~\ref{chap:NLE_G-Tex} and~\ref{chap:NLE_CLUE} complement the above studies in a model-intrinsic way for more faithful NLE generation: highlight explanations, which can faithfully reflect aspects of the model's internal reasoning, are used as cues to align NLEs with answer generation. Chapter~\ref{chap:NLE_G-Tex} introduces G-TEx~\citep{yuan-etal-2025-graph}, which injects highlight explanations through a graph neural network layer to guide NLE generation. This produces more faithful NLEs, as evaluated using the aforementioned counterfactual unfaithfulness test.
Chapter~\ref{chap:NLE_CLUE}  introduces CLUE~\citep{sun-etal-2026-explaining}, which uses span interactions as cues to steer the model to generate NLEs that explain its uncertainty.  I also propose a novel metric, extending counterfactual evaluation from binary answer flips to entropy-based shifts in the answer distribution, to measure the NLE faithfulness with respect to model uncertainty. Related studies similarly combine this form of guidance with model self-refinement~\citep{wang-atanasova-2025-self} or inject explanation importance scores directly to modify attention heads for NLE generation~\citep{alon-etal-2026-faithful}.


\subsubsection{Faithfulness of Chain-of-Thought Reasoning}
\label{sec:concept_faithfulness_cot}

Despite improving model performance on reasoning tasks~\citep{wei2022chain, kojima2022large, zhou2023least}, chain-of-thought (CoT) reasoning may not faithfully reflect model's underlying reasoning process
~\citep{lanham2023measuring,paul-etal-2024-making, arcuschin2025chainofthought}.

\paragraph{CoT faithfulness evaluation.} 
Existing evaluation methods can be broadly grouped into contextual, parametric, and mechanistic approaches. 
\textit{Contextual faithfulness} tests usually intervene on CoT traces and examine whether it leads to expected model answer change
~\citep{lanham2023measuring, bentham2024chain}. Feature-bias tests inject answer-position or sycophantic cues to the input and examine whether the CoT acknowledges them when they change the answer~\citep{NEURIPS2023_ed3fea90,chen2025reasoning}. However, subsequent work fails to verbalise that an influential cue may reflect an incomplete trace due to limited computation budget rather than unfaithfulness~\citep{zaman-srivastava-2026-chain}.

Since contextual interventions may not necessarily remove information from the model’s parameters, \citet{tutek-etal-2025-measuring} propose to measure \textit{parametric faithfulness} by unlearning knowledge associated with individual reasoning steps and observing the resulting model answer change. However, recent work finds that unlearning irrelevant or shuffled content can also substantially affect the resulting scores, suggesting that this measure might be confounded by unlearning artefacts~\citep{mitsuhashi-etal-2026-disentangling}.

\textit{Mechanistic approaches} instead compare the CoT with model's internal computation, examining whether its reasoning content is reflected in internal features~\citep{chen2026mechanistic,shiromani2026hypocrisyGap}, concept directions~\citep{li2025mapping}, activation dynamics~\citep{tanneru2024hardness,chen2026mechanistic}, or circuits~\citep{shen2026cieScorer}.
However, correspondence with selected internal signals may capture only part of the model's computation and does not by itself provide a complete causal account of its reasoning process~\citep{bereska2024mechanistic,hu-etal-2026-towards,gao-etal-2026-towards-intrinsic}. 

\paragraph{Meta-evaluation.} 
Recent studies have begun to meta-evaluate these faithfulness metrics by creating expert-annotated faithful and unfaithful CoT traces~\citep{shen2025faithcotbench}, or constructing tasks where the model's output can reveal which intermediate computations must have occurred~\citep{gurarieh2026bonafideMeta}. Although these works provide important validation settings, their notions of ``ground-truth'' faithfulness remain tied to particular tasks and constructions, limiting their generalisation across tasks, models, and explanation settings. 

\paragraph{CoT faithfulness improvement.}
Recent work has sought to improve CoT faithfulness by strengthening the causal role of reasoning traces in answer generation~\citep{paul-etal-2024-making,jia2026faithfulness}, optimising their simulatability~\citep{hase2026counterfactualSimulation,sivakumaran2026balancing}, or deriving training signals from counterfactual interventions~\citep{swaroop2025frit,akter2025inducing}. However, these approaches optimise particular proxy notions of faithfulness, and their gains may not transfer across metrics or evaluation paradigms~\citep{zaman-srivastava-2025-causal,sun2026investigatinginterplaycontextualparametric}.

Chapter~\ref{chap:CoT_FaithMate} focuses on meta-evaluating existing CoT faithfulness metrics, and
proposes FaithMATE~\citep{sun2026investigatinginterplaycontextualparametric}, a unified preference-alignment framework for optimising either contextual or parametric faithfulness objectives. Although FaithMATE performs optimisation, it is intended to be an interface to investigate whether improvements obtained by optimising a model towards one faithfulness metric transfer to other metrics and paradigms.
I find that contextual and parametric faithfulness are related but asymmetric. Moreover, gains under one contextual metric do not consistently transfer to others. These findings suggest that existing metrics capture distinct facets of faithfulness rather than a single notion and should be evaluated as a family rather than treated as interchangeable properties.

\subsection{Evaluating Plausibility and Utility of Explanations}
\label{sec:concept_utility}

Faithfulness examines whether explanations reflect a model's underlying reasoning process, whereas plausibility and utility capture human-facing properties.
Plausibility checks whether explanations appear understandable, coherent, and convincing, while utility concerns whether they can help users comprehend the model's prediction and achieve a concrete goal~\citep{doshivelez2017towards,jacovi-goldberg-2020-towards,deyoung-etal-2020-eraser}.
These properties are commonly assessed through human ratings of explanation quality~\citep{atanasova-etal-2020-generating-fact,wang-atanasova-2025-self,chuang-etal-2026-faithlm}, task-studies of how explanations support user decisions~\citep{hase-bansal-2020-evaluating,bucinca2020proxy,hashemi-chaleshtori-etal-2024-evaluating}, or comparisons with human-written explanations~\citep{rajani-etal-2019-explain,aggarwal-etal-2021-explanations}. 


\textbf{Manual evaluation of highlight explanations.} Agreement with human-annotated highlights is commonly used as a proxy for plausibility, measuring whether explanations 
resemble those that human annotators consider relevant to a decision~\citep{deyoung-etal-2020-eraser}. 
However, later work argues that the assessment of explanations should cover multiple aspects, but usually focuses on a single form of highlight explanations~\citep{atanasova-etal-2020-diagnostic,hashemi-chaleshtori-etal-2024-evaluating}. 
Chapter~\ref{chap:highlight_diagnostic} contributes to this direction by comparing three types of highlight explanations across multiple dimensions, including agreement with human annotations, faithfulness, etc. Although span interactions perform best on most properties, they exhibit a trade-off between comprehensiveness, one aspect of faithfulness, and agreement with human annotations. This reinforces that faithfulness and plausibility should be treated as distinct properties and evaluated separately~\citep{deyoung-etal-2020-eraser,jacovi-goldberg-2020-towards,atanasova-etal-2020-diagnostic}.

\textbf{Manual evaluation of natural language explanations.} 
Human evaluation of NLEs commonly assesses fluency, clarity, coherence, informativeness, factual correctness, redundancy, and helpfulness~\citep{atanasova-etal-2020-generating-fact,wang-atanasova-2025-self,chuang-etal-2026-faithlm,liu-etal-2025-metafaith}. Because human evaluation requires substantial annotator time and
financial resources, it is often conducted on a small subset of generated NLEs as a quality check~\citep{wang-etal-2025-cross,wang-atanasova-2025-self,zhao-iii-2025-necessary}. 
Chapters~\ref{chap:NLE_G-Tex} and~\ref{chap:NLE_CLUE} similarly conduct human evaluation to verify that improvements in faithfulness do not compromise plausibility.
Both chapters guide NLE generation using input highlights: G-Tex encodes these cues through graph-layer and model finetuning~\citep{yuan-etal-2025-graph}, whereas CLUE leverages them via inference-time model steering~\citep{sun-etal-2026-explaining}.
The latter responds to fact-checkers' need for explanations by explicitly referring to the input conflicts and agreements relevant to the model's uncertainty~\citep{warren2025explainablefactchecking}.
Alongside improved faithfulness, human evaluations indicate that the resulting explanations are generally less redundant and of higher overall quality and helpfulness.

A stronger, task-based notion of explanation utility assesses whether explanations help users achieve concrete goals, such as predicting model behaviour~\citep{hase-bansal-2020-evaluating}, appropriately accepting or rejecting AI recommendations~\citep{bucinca2020proxy,bansal2021whole}, or improving performance in AI-assisted decision-making~\citep{bucinca2020proxy,bansal2021whole,hashemi-chaleshtori-etal-2024-evaluating}.
Chapter~\ref{chap:Human_showme} contributes to this line of work through a user study examining the role of explanations and input evidence in assisting users in validating AI decisions in fact-checking. The results show that users generally regard input evidence as the most reliable source, highlighting the need for future work on how evidence should be presented and how explanations can be more effectively grounded in it~\citep{warren2026show}.

\textbf{Manual evaluation of chain-of-thought explanations.} 
Even when CoT reasoning improves answer accuracy, it can contain human-detectable errors, such as incorrect intermediate steps~\citep{nguyen-etal-2024-direct}. Manual evaluation therefore examines whether these reasoning steps are individually correct, coherent, logically ordered, and sufficient to justify the final answer~\citep{lyu-etal-2023-faithful,khalifa-etal-2023-grace,prasad-etal-2023-receval,lee-hockenmaier-2025-evaluating}. 
Such issues may undermine users' trust in CoTs and, consequently, their utility~\citep{rong2024towards,hashemi-chaleshtori-etal-2024-evaluating}. 

Although Chapter ~\ref{chap:CoT_FaithMate} primarily focuses on the automatic meta-evaluation of CoT faithfulness metrics, 
it also conducts a qualitative manual analysis to examine whether improvements under these metrics produce human-visible CoT changes. 
Interestingly, some metrics lead to visible CoT--answer consistency, while others result in less noticeable textual changes. This suggests that human-perceived plausibility and current proxy-based faithfulness need not improve together, underscoring the importance of evaluating and improving both.

\subsection{Complex Reasoning Tasks and Dataset Settings}
\label{sec:concept_complex_reasoning_tasks}

\begin{table}[htbp]
\centering
\footnotesize
\renewcommand{\arraystretch}{1.15}
\caption{Complex reasoning tasks and datasets used across the contribution chapters. HE denotes highlight explanations, NLE denotes natural-language explanations, and CoT denotes chain-of-thought reasoning.}
\label{tab:complex_reasoning_tasks}
\begin{tabularx}{0.96\linewidth}{@{}p{2.4cm} X p{3.1cm}@{}}
\toprule
\textbf{Task / setting} &
\textbf{Dataset and setup} &
\textbf{Chapter} \\
\midrule

\textbf{Fact-checking}
&
\textbf{FEVER}~\citep{thorne-etal-2018-fever}: claims paired with Wikipedia evidence and veracity labels.
&
Ch.~\ref{chap:highlight_diagnostic} (HE)
\\

&
\textbf{HealthVer}~\citep{sarrouti-etal-2021-evidence-based}: health claims paired with scientific evidence statements labelled as support, refute, or neutral.
&
Ch.~\ref{chap:NLE_CLUE} (NLE)
\\

&
\textbf{DRUID}~\citep{hagstrom-etal-2025-reality}: real-world claims paired with retrieved evidence passages and fact-check verdicts.
&
Ch.~\ref{chap:NLE_CLUE} (NLE)\newline
Ch.~\ref{chap:Human_showme} (NLE)
\\

\addlinespace

\textbf{Natural \newline language \newline inference}
&
\textbf{SNLI}~\citep{bowman-etal-2015-large}: premise--hypothesis pairs labelled as entailment, contradiction, or neutral.
&
Ch.~\ref{chap:highlight_diagnostic} (HE)
\\

&
\textbf{e-SNLI}~\citep{camburu2018snli}: SNLI examples augmented with human-written explanations.
&
Ch.~\ref{chap:NLE_G-Tex} (NLE)
\\

\addlinespace

\textbf{Commonsense reasoning}
&
\textbf{ComVE}~\citep{wang-etal-2020-semeval}: paired statements requiring identification of the commonsensical one, with annotated explanations;
\textbf{ECQA}~\citep{aggarwal-etal-2021-explanations}: multiple-choice commonsense questions with correct answers and human-written explanations.
&
Ch.~\ref{chap:NLE_G-Tex} (NLE)
\\

\addlinespace

\textbf{Scientific and logical QA}
&
\textbf{OpenBookQA}~\citep{mihaylov-etal-2018-suit}: multiple-choice elementary-science questions requiring factual knowledge;
\textbf{LogiQA}~\citep{liu2020logiqa}: multiple-choice logical-reasoning questions over textual scenarios, with correct answers.
&
Ch.~\ref{chap:CoT_FaithMate} (CoT)
\\

\addlinespace

\textbf{Context utilisation}
&
\textbf{World Capital, CounterFact, Fakepedia, and ConflictQA}~\citep{yu-etal-2023-characterizing,meng2022locating,monea-etal-2024-glitch,xie2024adaptive}: factual queries with different context types, e.g. conflicting or irrelevant, and reference answers.
&
Ch.~\ref{chap:highlight_context_utilisation} (HE)
\\

\bottomrule
\end{tabularx}
\end{table}

Complex reasoning tasks require models to coordinate multiple capabilities, such as relating information across input components~\citep{thorne-etal-2018-fever,bowman-etal-2015-large}, combining supplied evidence with background knowledge~\citep{mihaylov-etal-2018-suit}, performing multi-step logical inference~\citep{liu2020logiqa}, and resolving conflicts between external context and parametric knowledge~\citep{xie2024adaptive}. These tasks provide informative testbeds for explainability because a correct answer alone does not reveal whether the model relied on the intended parts of the input, used relevant knowledge, or followed a reasoning process aligned with human expectations. The datasets used in this thesis span the task settings summarised in Table~\ref{tab:complex_reasoning_tasks}. Detailed dataset setups and the corresponding explanation and evaluation protocols are provided in the respective chapters.

Dataset selection in this thesis balances coverage of diverse reasoning demands with the intended roles and strengths of each explanation form. Although all three explanation forms are, in principle, applicable across these tasks, highlight explanations (Chapter~\ref{chap:highlight_diagnostic} and~\ref{chap:highlight_context_utilisation}) are examined mainly in settings with distinct input components, such as claims and evidence or premises and hypotheses, where identifying influential tokens and interactions across input parts is particularly informative. NLEs (Chapter~\ref{chap:NLE_G-Tex}, ~\ref{chap:NLE_CLUE}, and~\ref{chap:Human_showme}) are studied across natural language inference, fact-checking, and question answering tasks, where these free-text rationales can integrate background knowledge and communicate information not explicitly stated in the input. CoT (Chapter~\ref{chap:CoT_FaithMate}) is evaluated on scientific and logical question-answering tasks, whose multi-step reasoning demands the model to make explicit reasoning traces. Together, these datasets enable the thesis to study diverse explanation forms and apply corresponding evaluation protocols, providing complementary insights into model decision-making across different scenarios.

\section{Scientific Contributions}
\label{intro:contributions}

\subsection{Diagnostic Evaluation of Input Feature Explanations}
\subsubsection{Paper 1: Evaluating Input Feature Explanations through a Unified Diagnostic Evaluation Framework}
\begin{figure*}[!t]
\centering
\small
\includegraphics[width=0.7\textwidth]{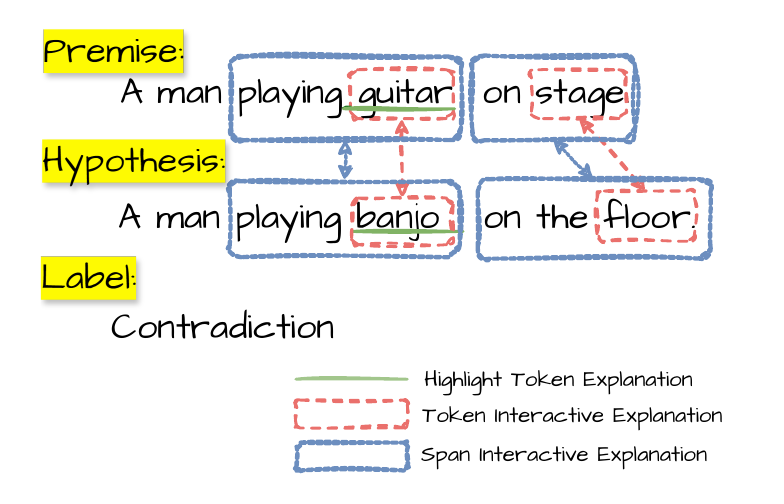}
\caption{An example of the three types of Input Feature Explanations on an instance from the SNLI dataset, with their two most important pieces of explanation (token for highlight token explanation, token tuple for token interactive explanation, span tuple for span interactive explanation, respectively). Originally published in \cite{sun-etal-2025-evaluating}.
}
\label{fig:contribution_paper_1}
\end{figure*}
Highlight explanations, which can be categorised into three forms, are a popular explanation type for explaining a model's decision-making process by highlighting key input features, including individual tokens, token interactions, and span interactions. However, previous work has mostly evaluated each highlight explanation form in isolation, including individual tokens~\citep{atanasova-etal-2020-diagnostic,deyoung-etal-2020-eraser}, token interactions ~\citep{janizek2021explaining,NEURIPS2020_443dec30}, and span interactions~\citep{choudhury-etal-2023-explaining}; see Fig.~\ref{fig:contribution_paper_1}. Thus, such work provides limited practical guidance on how to select among these different forms in application scenarios, where principled explanation selection is important.

To address this gap, this work (i) proposes \textbf{a unified diagnostic evaluation framework to enable automatic, comprehensive, and rigorous comparison} of these three highlight explanation forms across a set of properties~\citep{sun-etal-2025-evaluating}. The evaluation spans four important aspects: Faithfulness, Agreement with Human Annotations, Simulatability, and Complexity. To enable a fair comparison, I introduce a dynamic threshold to adapt the metrics by taking into account the variation in token numbers within each explanation type. Based on this framework, (ii) \textbf{extensive empirical analysis} is conducted across three explanation techniques for each explanation type, two complex reasoning tasks, and two language models. Experiments show that explanation types have distinct strengths across various properties, with span interactive explanations generally outperforming the others on most aspects. While highlight tokens excel in Comprehensiveness, one aspect of Faithfulness, \textbf{a trade-off between low Comprehensiveness and high Agreement with Human Annotations appears in span interactive explanations}. This encourages future work to further mitigate this tension and optimise highlight explanations across all desired properties, possibly by combining different explanation types.

\subsubsection{Paper 2: Evaluation Framework for Highlight Explanations of Context Utilisation in Language Models}

Language models, despite storing substantial knowledge in their parameters, are expected to consider and leverage information in the provided context, especially in tasks such as retrieval-augmented generation, question answering, and fact-checking~\citep{lewis2020retrieval,thorne-etal-2018-fever,jin-etal-2024-cutting,monea-etal-2024-glitch}. However, previous work has mainly studied which knowledge source the model uses when presented with different types of context~\citep{yu-etal-2023-characterizing,monea-etal-2024-glitch,xie2024adaptive,hagstrom2025cub}.
Such works do not examine whether highlight explanations, i.e., important input segments, can accurately indicate the model's context utilisation behaviour, e.g., which context parts are used for the model's answer. Previous evaluation of highlight explanations usually focuses on proxies, such as faithfulness~\citep{atanasova-etal-2020-diagnostic}.

\begin{figure*}[!t] 
\centering
\includegraphics[width=\columnwidth]{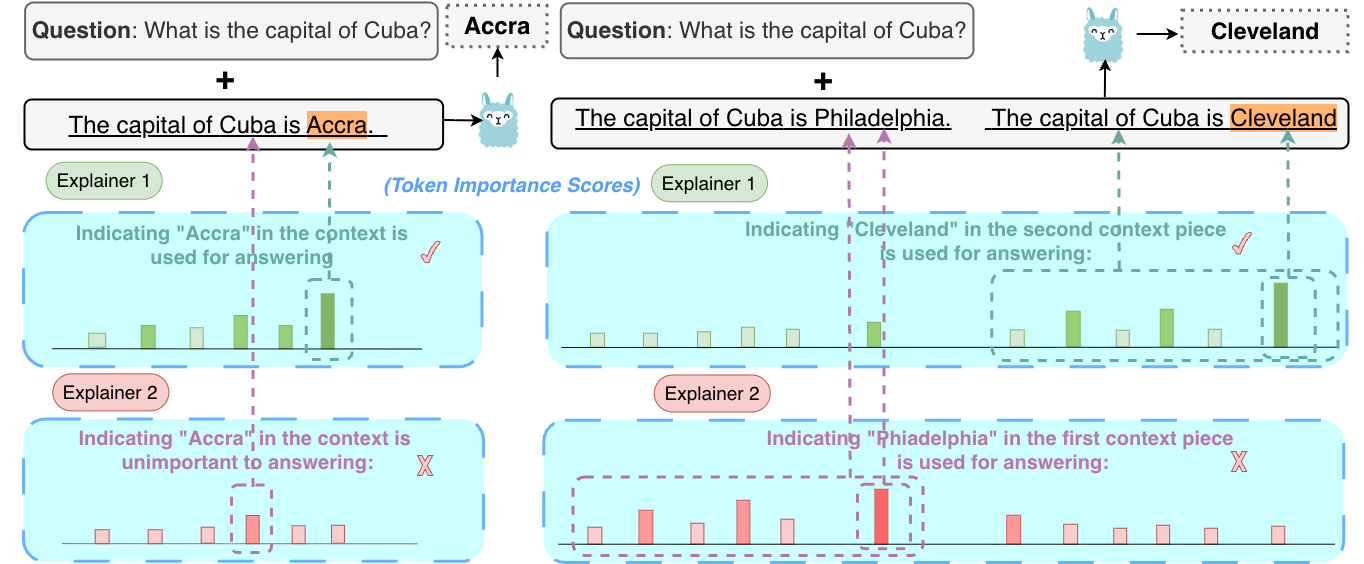}
\caption{Examples of explanations with high and low utility in indicating the model’s context utilisation behaviour. Left: With one context conflicting with the model’s parametric knowledge (``Havana''), Explainer 1 correctly indicates that ``Accra'' in the context is used for the answer, while Explainer 2 fails. Right: With two such contexts, Explainer 1 correctly indicates that ``Cleveland'' in the second context is used for the answer, whereas Explainer 2 incorrectly attributes the answer to ``Philadelphia'' in the first context. Explainer~1 shows better utility than Explainer~2 in both cases. Originally published in \cite{sun2026evaluation}.
}
\label{fig:contribution_paper_2}
\end{figure*}

To address this gap, this work (i) \textbf{proposes an evaluation framework to examine explanations' ability to indicate a model's context utilisation}, especially regarding whether or not the model answers from the provided context, and more granularly, from which context piece and from which precise span the model generates the answer~\citep{sun2026evaluation}. This framework synthesises realistic challenges in retrieval-augmented generation by exploring four context-usage scenarios, Conflicting, Irrelevant, Double-Conflicting, and Mixed, which mimic retrieval-time challenges such as knowledge conflicts, irrelevant noise, and multi-document competition. See Fig.~\ref{fig:contribution_paper_2} for examples of explanations with high/low utility in Conflicting and Double-Conflicting scenarios. Three complementary metrics are leveraged for evaluation: document-level attribution accuracy, simulatability, and token-level attribution accuracy. The framework is designed in a way that the model's output answer serves as the gold-standard indicator of its context utilisation. 
Besides the evaluation framework, I (ii) \textbf{propose a new mechanistic interpretability-based method}, MechLight, which maps the model's context-usage signals carried by specific attention heads to highlight explanations. Across evaluations on four datasets, five models, and four explanation methods, MechLight outperforms the others in most cases. At the same time, widely used techniques, such as gradient-based methods, are poor indicators of the model's context utilisation. Interestingly, I find that all techniques suffer from positional bias and are sensitive to the length of the provided context. This suggests that future work should design explainability methods that better reveal model context utilisation and mitigate such bias.

\subsection{Utilising Input Feature Explanations for Faithful Natural Language Explanations}
\subsubsection{Paper 3: Graph-Guided Textual Explanation Generation Framework}

Natural language explanations (NLEs) are an important explanation form for expressing a model's rationale for a prediction in free-text form~\citep{miller2019explanation,wiegreffe-etal-2021-measuring}. Self-rationalisation, where the explanation is generated together with the answer, is considered to achieve better alignment between the explanation and the answer~\citep{NEURIPS2018_3e9f0fc9,marasovic-etal-2022-shot}. However, these explanations often suffer from faithfulness issues: they do not necessarily accurately reflect the model's internal reasoning~\citep{kumar2020nile,atanasova-etal-2023-faithfulness,lyu2024towards}. While a range of previous work has focused on improving faithfulness by resorting to external knowledge, refining prompts, or designing training losses~\citep{pmlr-v162-majumder22a,marasovic-etal-2022-shot,chuang-etal-2026-faithlm}, it remains underexplored how to design intrinsic explanation generation with cues from model internals.

\begin{figure*}[!t]
    \centering
    \includegraphics[width=1\textwidth]{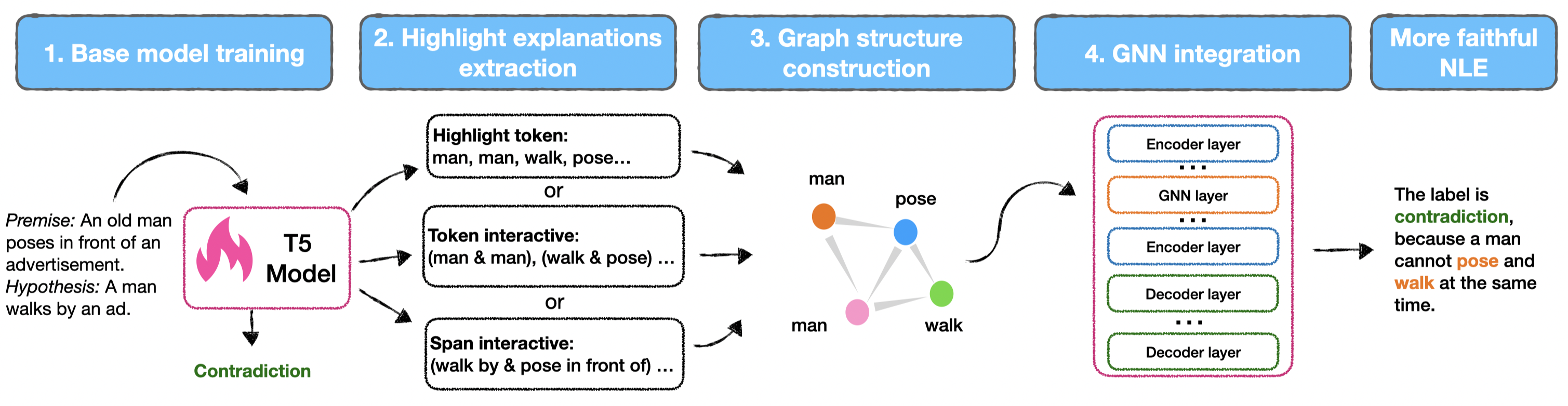}
    \caption{Overview of \method, which consists of four key steps: (1) A base model such as T5 is trained using the task-specific dataset for label prediction. (2) Three types of highlight explanations are extracted from the trained model. (3) Graph structure is constructed based on the highlight explanations. (4) Graph structure is integrated into the model with a GNN layer, after which the model is fine-tuned for label prediction and NLE generation. Originally published in \cite{yuan-etal-2025-graph}.}
    \label{fig:contribution_paper_3}
\end{figure*}

To address this gap, this paper (i) \textbf{proposes G-TEx, a framework that leverages highlight explanations as cues for faithful NLE generation}~\citep{yuan-etal-2025-graph}; see the pipeline in Fig.~\ref{fig:contribution_paper_3}. Specifically, three types of highlight explanations are extracted and separately encoded as graph structures using a graph neural network. These graphs are then integrated into the model, which is fine-tuned using human-written explanations as reference. The intuition is that highlights, whose faithfulness can be easily measured~\citep{atanasova-etal-2020-diagnostic,sun-etal-2025-evaluating}, can provide internal cues that guide the model towards generating more
faithful NLEs. Their compactness may also encourage explanations that are more concise, less redundant, and therefore more likely to be preferred by humans. Based on this framework, (ii) \textbf{multiple facets of the generated NLEs are assessed, including faithfulness, similarity to human-written explanations, and human judgements}. Experiments on e-SNLI, ComVE, and ECQA~\citep{camburu2018snli,wang-etal-2020-semeval,aggarwal-etal-2021-explanations} show that the resulting NLEs are more faithful than those produced from the fine-tuning baseline, align more closely with human annotations and receive slightly stronger human-preference ratings. However, the magnitude of the improvement varies across datasets and
highlight explanation types. For e-SNLI~\citep{camburu2018snli}, where interactions between two input parts are important, token and span interactive explanations yield larger faithfulness gains. For ComVE~\citep{wang-etal-2020-semeval}, where the first input component is fixed and the candidate statements contain the principal distinctions, highlight token explanations perform best, suggesting that in this task the interactions between different input parts play a smaller role in model's decision-making. Overall,  highlight-guided NLE generation is a promising direction, although its faithfulness benefits can be model- and task-dependent. Future work should continue developing and refining the highlight-guided methods.

\subsubsection{Paper 4: Explaining Sources of Uncertainty in Automated Fact-Checking}
Fact-checkers highlight the need to assess source reliability and reconcile conflicting evidence in AI-assisted fact-checking~\citep{warren2025explainablefactchecking,schlichtkrull-etal-2023-intended}, since in realistic fact-checking scenarios, multiple pieces of evidence may be provided and may cause conflicts that affect model uncertainty~\citep{graves2017anatomy,micallef2022true}. However, existing approaches to communicating uncertainty usually rely on numerical confidence, such as ``I am 90\% sure'', or verbal hedges, such as ``I am not sure, but...'', which cannot surface the sources of the model's uncertainty~\citep{kadavath2022language,mielke-etal-2022-reducing,yona-etal-2024-large}. Meanwhile, previous explainability work mainly explains the model's verdict and neglects where its uncertainty arises~\citep{atanasova-etal-2020-generating-fact,zeng-gao-2024-justilm}.

\begin{figure}[!t]
    \centering
    \includegraphics[width=0.65\linewidth]{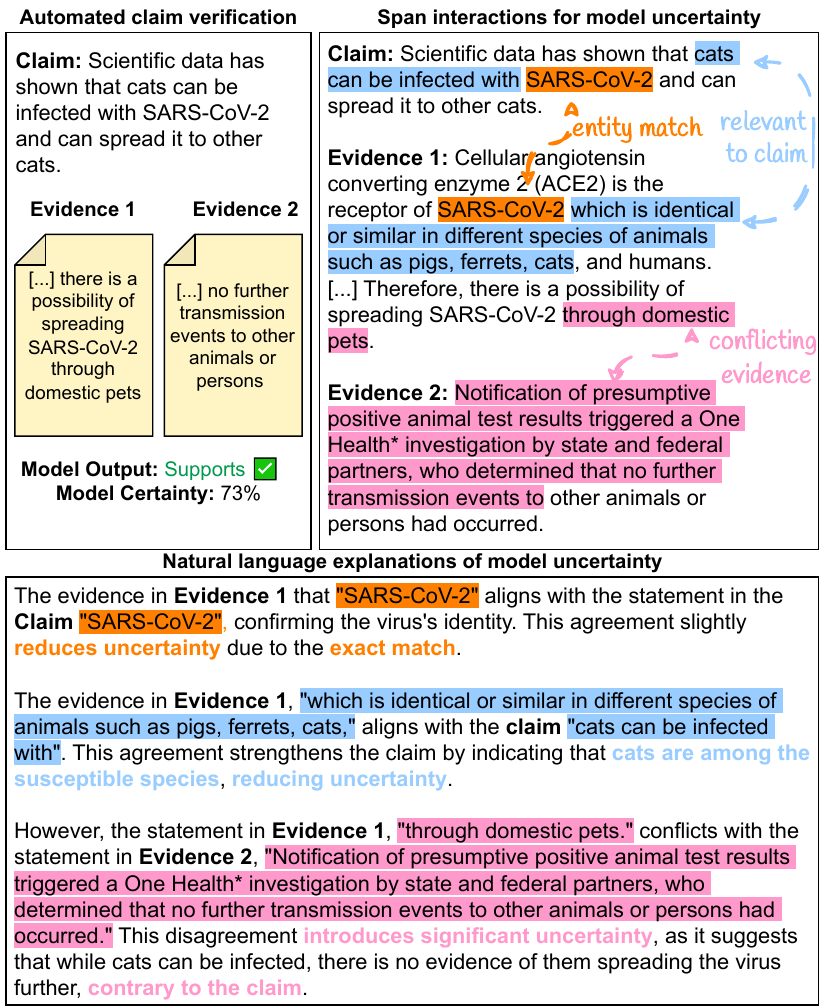}
    \caption{An example of \modelname{} output for a claim with two evidence passages: model verdict and certainty, and an uncertainty explanation grounded in conflict/agreement span interactions. Originally published in \cite{sun-etal-2026-explaining}.}
    \label{fig:contribution_paper_4}
\end{figure}

To fill this gap, this paper (i) \textbf{proposes the first framework, CLUE, to explain model uncertainty in natural language explanations}, especially by pinpointing agreement and/or contradictory span interactions between the claim and multiple evidence pieces~\citep{sun-etal-2026-explaining}. See Fig.~\ref{fig:contribution_paper_4} for an example of an uncertainty explanation in this setup. Highlight explanations, especially span interactive explanations, can naturally serve as a bridge from the input to the model's answer. Therefore, I build the uncertainty explanation generation pipeline by first extracting the most salient span interactions between the input claim and evidence pieces, and then leveraging two methods to guide the model to generate explanations that refer to these interactions and discuss how they affect model uncertainty, via few-shot prompting~\citep{zhao-etal-2024-pacar} and attention steering~\citep{ICLR2024_b99d6cc4}. To assess their faithfulness to the model's uncertainty, I (ii) \textbf{propose a new metric by extending the popular counterfactual faithfulness test to capture model uncertainty instead of verdict change}~\citep{atanasova-etal-2023-faithfulness,siegel-etal-2024-probabilities}. Specifically, I measure the correlation between the model's uncertainty shift and the explanation's mention of the inserted word, under the assumption that a faithful uncertainty explanation should reflect important input words that lead to changes in model uncertainty.
Through experiments on two fact-checking datasets and multiple models, I show that CLUE generates explanations that are more faithful to model uncertainty, less redundant, and better aligned with the predicted verdict than a prompting baseline. Human evaluation further shows that CLUE explanations are generally preferred across multiple aspects, including helpfulness. Interestingly, the two CLUE variants exhibit a faithfulness--plausibility trade-off: the steering-based variant achieves higher faithfulness by adhering more closely to the extracted span interactions, whereas the prompting-based variant is rated as more helpful. This suggests that fragmentary span interactions may omit contextual information that humans consider important, motivating future work to produce more coherent, human-understandable span interactions and to mitigate this tension.

\subsection{Evaluating Faithfulness and Utility of Free-Text Explanations}
\subsubsection{Paper 5: Investigating the Interplay between Contextual and Parametric CoT Faithfulness under Optimisation}

The faithfulness of chain-of-thought (CoT) explanations has received increasing attention~\citep{NEURIPS2023_ed3fea90,paul-etal-2024-making,tutek-etal-2025-measuring,zaman-srivastava-2025-causal}. Existing metrics mainly follow two paradigms: \textit{contextual faithfulness}, which evaluates a CoT by perturbing the input or CoT trace~\citep{NEURIPS2023_ed3fea90,lanham2023measuring}, and  \textit{parametric faithfulness}, which evaluates it by removing CoT-verbalised knowledge from model parameters~\citep{tutek-etal-2025-measuring}. Both paradigms then test whether the answer changes as expected. Previous work has compared these paradigms only for specific CoTs and found that they can assign different judgements to the same CoTs~\citep{zaman-srivastava-2026-chain}. It therefore remains unclear whether optimising one faithfulness metric produces gains that transfer to other metrics, despite growing work on faithfulness enhancement~\citep{paul-etal-2024-making,swaroop2025frit,jia2026faithfulness}. Consequently, there is still limited practical evidence about whether current metrics capture the same underlying notion of faithfulness~\citep{parcalabescu-frank-2024-measuring,zaman-srivastava-2025-causal}.

\begin{figure*}[!t]
\centering
\resizebox{0.9\textwidth}{!}{
\begin{minipage}{\textwidth}
\includegraphics[width=\textwidth]{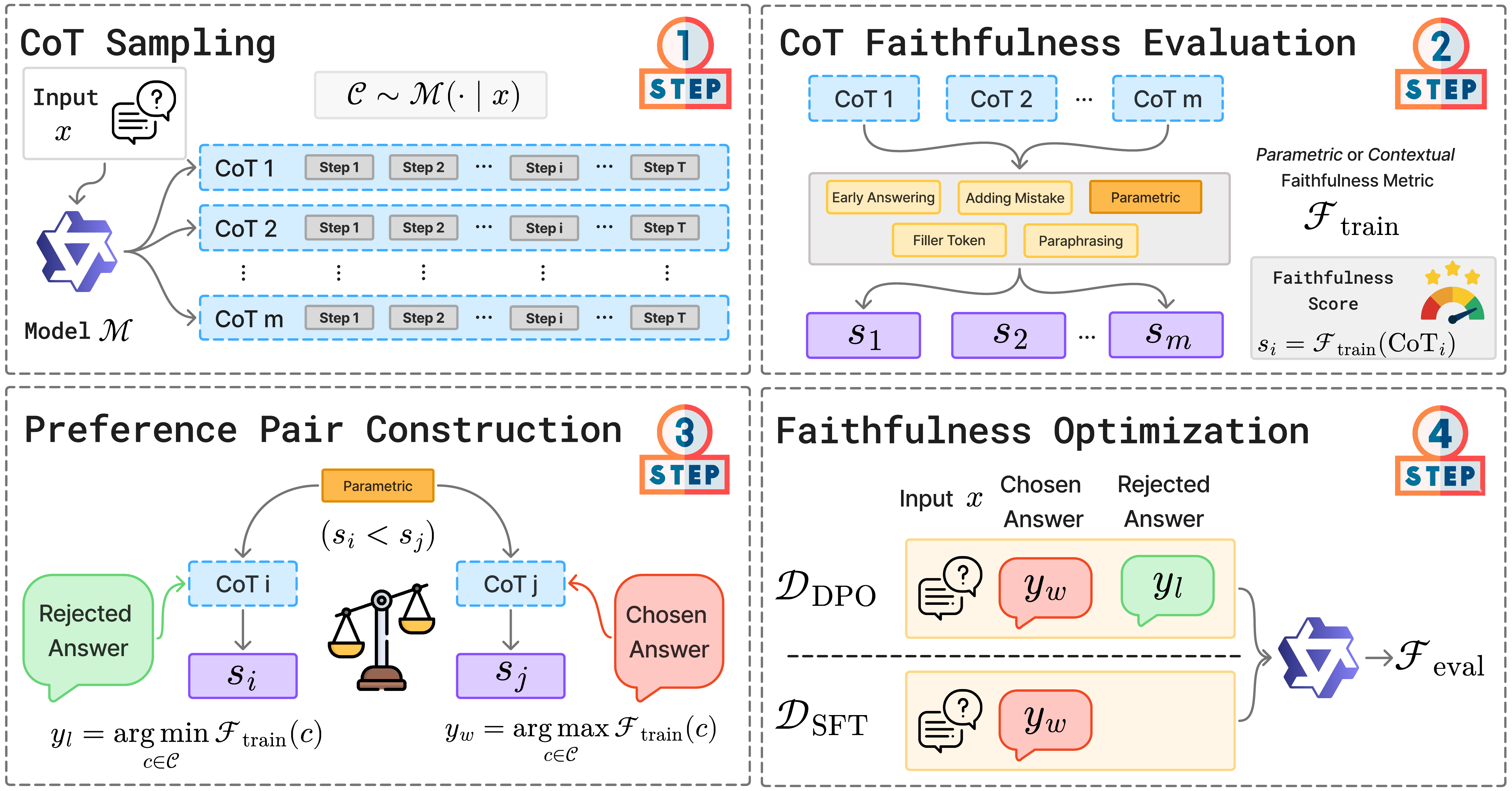}
\end{minipage}
}
\caption{Overview of \our{}. CoTs are sampled from a base model and scored using contextual or parametric faithfulness metrics. They are then used to construct preference pairs, and the model is optimised towards the selected faithfulness objectives. Cross-metric transfer is evaluated across the two faithfulness paradigms. Originally published in \cite{sun2026investigatinginterplaycontextualparametric}.}
\label{fig:contribution_paper_5}
\end{figure*}

To this end, this paper (i) \textbf{proposes FaithMATE, a framework that provides a direct interface for improving CoT faithfulness towards a specific faithfulness metric and comparing whether, and to what extent, faithfulness gains transfer across different metrics and paradigms}~\citep{sun2026investigatinginterplaycontextualparametric}. See Fig.~\ref{fig:contribution_paper_5} for the pipeline overview. Specifically, CoTs are first sampled from the base model, and a target metric is applied to select the most and least faithful CoTs, which are then used to construct preference pairs for direct preference optimisation. Faithfulness gain transfer is finally evaluated across multiple faithfulness metrics, including both target and non-target metrics, to study how gains on target metrics generalise to non-target metrics. Through (ii) \textbf{a rigorous study on three medium-sized language models, two question-answering datasets, and six faithfulness metrics}, I find that optimisation towards parametric faithfulness generally brings more consistent gains across both paradigms. Furthermore, within the contextual faithfulness paradigm, faithfulness improvements induced by different metrics do not usually transfer to each other, and model merging can partially amplify improvements but does not guarantee stable cross-metric transfer. Finally, through qualitative manual checks, I find that optimising contextual faithfulness metrics usually improves CoT--answer consistency and mitigates post-hoc rationalisation, while parametric faithfulness brings milder visible CoT changes and sometimes makes CoTs more explicitly grounded in task-relevant facts and knowledge. Overall, this suggests that current metrics do not measure the same notion of faithfulness, and thus evaluation using a single metric can be misleading. Instead, these metrics need to be considered together for comprehensive evaluation.

\subsubsection{Paper 6: Show Me the Evidence: Evaluating the Role of Evidence and NLEs in AI-Supported Fact-Checking}

The linguistically coherent and persuasive outputs generated by LLMs may cause users to over-rely on model decisions or recommendations \citep{bender2021dangers,pauli-etal-2025-measuring}. Previous work has explored whether including explanations and links to evidence can reduce users' over-reliance \citep{kim2025fostering}; however, little work has investigated experimental setups in which the evidence documents are made readily accessible to users. Moreover, regarding the role of explanations, prior work has usually focused on verdict explanations and has reported conflicting findings: explanations can sometimes support more appropriate use of AI advice, but they can also increase over-reliance \citep{vasconcelos_explanations_2023,bansal2021whole}.

\begin{figure*}[!t]
    \centering
    \includegraphics[width=0.9\linewidth]{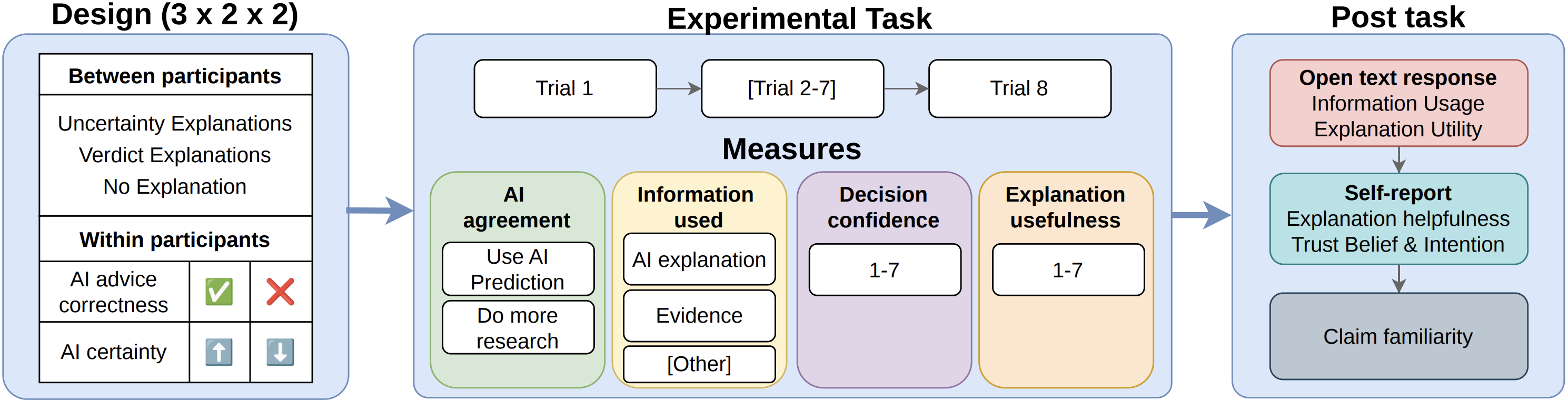}
    \caption{Overview of study design and flow. Originally published in \cite{warren2026show}.}
    \label{fig:contribution_paper_6}
\end{figure*}

To bridge this gap, this paper  \textbf{conducts a user study to investigate the roles of explanations and evidence in helping users validate model answers in a fact-checking task, across varying explanation types, model certainty, and model correctness}~\citep{warren2026show}. As indicated in Fig.~\ref{fig:contribution_paper_6}, users are shown a target claim, the model's verdict, the model's certainty, and one of three explanation conditions: a verdict explanation, an uncertainty explanation, or no explanation. Users also have access to two click-to-reveal evidence passages, and they are asked to decide whether to accept the model's verdict or decline it. In addition, users are also asked to self-report which information sources they used to make their decision, selecting from multiple presented sources, such as explanation, evidence, etc. Based on this setup, analysis \textbf{shows that evidence is the dominant information source participants use when validating the model's answers}: 64\% of participants open both evidence passages for every claim, and only three participants do not open any evidence passages. Although participants also judge explanations to be useful, especially when the model's answer is correct or when the model is highly certain, there is no significant usefulness difference between the two natural language explanation types.
Furthermore, higher self-reported reliance on the AI system is associated with greater use of explanations, whereas lower self-reported reliance is associated with greater use of evidence. Overall, the paper highlights the central role of accessible evidence in AI-assisted fact-checking and motivates future work on effective evidence presentation and evidence-grounded explanations. Although explanations do not increase over-reliance compared with the no-explanation condition, the qualitative results suggest that users often use evidence to identify persuasive but misleading explanations. This points to the need for HCI systems that better support users in cross-checking AI reasoning against the underlying evidence.

\section{Conclusions and Perspectives for Further Research}
\label{intro:limitations_future}

\begin{table}[t]
      \centering
      \footnotesize
      \setlength{\tabcolsep}{4pt}
      \begin{tabular}{c l c c c c c c c}
      \toprule
      & & & \multicolumn{3}{c}{\textbf{Expl.\ Type}}
       & \multicolumn{3}{c}{\textbf{Contribution}} \\
      \cmidrule(lr){4-6} \cmidrule(lr){7-9}
      \textbf{\#} & \textbf{Publication} & \shortstack{\textbf{Presented in}\\\textbf{Chapter}}
       & \textbf{HE} & \textbf{NLE} & \textbf{CoT}
       & \textbf{E} & \textbf{M} & \textbf{A} \\
      \midrule
      1 & \citet{sun-etal-2025-evaluating} & \ref{chap:highlight_diagnostic}
        & \checkmark & &
        & \checkmark & & \checkmark \\
      2 & \citet{sun2026evaluation} & \ref{chap:highlight_context_utilisation}
        & \checkmark & &
        & \checkmark & & \checkmark \\
      \midrule
      3 & \citet{yuan-etal-2025-graph} & \ref{chap:NLE_G-Tex}
        & & \checkmark &
        & & \checkmark & \checkmark \\
      4 & \citet{sun-etal-2026-explaining} & \ref{chap:NLE_CLUE}
        & & \checkmark &
        & & \checkmark & \checkmark \\
      \midrule
      5 & \citet{sun2026investigatinginterplaycontextualparametric} & \ref{chap:CoT_FaithMate}
        & & & \checkmark
        & \checkmark & & \checkmark \\
      6 & \citet{warren2026show} & \ref{chap:Human_showme}
        & & \checkmark &
        & & & \checkmark \\
      \bottomrule
      \end{tabular}
      \caption{Summary of the six publications that form the basis of this thesis, including the chapters in which they are presented, their explanation types, and their contribution types. Explanation types are Highlight Explanations~(HE), Natural Language Explanations~(NLE), and Chain-of-Thought~(CoT). Contribution types are Evaluation Framework~(E), Methodological~(M), and Empirical Analysis~(A). The chapter numbers link to the corresponding chapters.}
      \label{tab:contributions}
\end{table}
The publications in this thesis collectively advance research on explainability in NLP, spanning three explanation types: highlight explanations, natural language explanations, and chain-of-thought reasoning. A recurring theme is the interplay between faithfulness and plausibility: explanations should both reflect the model's reasoning process and be understandable and useful to users. This thesis approaches this challenge from both the evaluation and generation sides, beginning with diagnostic evaluation of highlight explanations (Chapters \ref{chap:highlight_diagnostic}, \ref{chap:highlight_context_utilisation}), building natural language explanation generation methods that leverage highlights as faithful cues (Chapters \ref{chap:NLE_G-Tex}, \ref{chap:NLE_CLUE}), and finally evaluating the faithfulness and utility of free-text explanations (Chapters \ref{chap:CoT_FaithMate}, \ref{chap:Human_showme}). Table~\ref{tab:contributions} maps the contributions of each paper by explanation type and contribution type.

Overall, while pushing the boundaries of the field from different perspectives, the papers in this thesis point to several broader limitations in explainable AI. In the following, I detail these limitations by chapter, and identify several key directions that I believe will drive future explainability research.

\subsection{From Input Feature Explanations to Mechanistically Enriched Highlight Explanations}
\label{intro:future_part1}

Part II (Chapters \ref{chap:highlight_diagnostic}, \ref{chap:highlight_context_utilisation}) establishes evaluation foundations for highlight explanations: Paper 1 introduces a unified diagnostic framework for comparing token, token interactive, and span interactive explanations across four properties, while Paper 2 proposes a utility-based framework for assessing whether highlights can indicate model context utilisation across different context scenarios.


Through these chapters, two limitations emerge. First, \emph{faithfulness and plausibility are in tension}: in Chapter 2, span interactive explanations, while exhibiting superior performance on almost all of the assessed aspects,  exhibit such tension. This highlights the persistent challenge in advancing explainability techniques to mitigate this tension, resonating with other works in cross-lingual and legal evaluation scenarios~\citep{banerjee2026lost_interpretation, namazov2026legal}. Second, \emph{current attribution methods are not reliable indicators of model context utilisation}: in Chapter 3, commonly used attribution methods such as Integrated Gradients~\citep{sundararajan2017axiomatic} and Attention~\citep{jain-wallace-2019-attention,wiegreffe-pinter-2019-attention} are found to provide little utility for indicating context utilisation. By contrast, the proposed \MI\ method takes an initial step by mapping an internal context-utilisation signal back to input tokens, showing the best utility. However, it relies on a single attention head, which could not capture the full picture of the model's decision-relevant information, and suffers from position bias and shows degraded performance on long contexts. Therefore, I deem the directions below worthy for future work to explore.

\paragraph{Grounding highlight explanations in model mechanisms.} Recent efforts have adapted attribution methods to large language models~\citep{pan2026flashtrace, pramanik2026heta}, but they still suffer from the aforementioned issues. In the meantime, mechanistic interpretability methods can identify how internal features contribute to outputs, which might offer faithful signals of the model's reasoning, but lack human plausible interpretation ~\citep{cunningham2023sparse, templeton2024scaling, lindsey2025circuit}.
The future direction should combine these two perspectives: developing attribution methods that are grounded in the model's internal signals, while remaining plausible for humans. Efforts should be made in developing generalizable and robust mechanistic interpretability techniques that are comprehensive of the model's decision information, across distributed model components and causally reliable. Besides, principled ways to project the resulting components back onto input parts will also benefit this practice. Furthermore, mechanistic studies that isolate the effects of surface-level contextual characteristics, such as length and position, could help overcome explanations' biases to improve their utility in realistic multi-context scenarios.

\paragraph{Highlight explanations as a useful mediator among input, output, and model behaviour.}
Highlight explanations can direct users' attention to specific parts of the input evidence, which could serve as a natural communicative tool to show users how to understand the model~\citep{jacovi-goldberg-2021-aligning,malaviya-etal-2024-said,schmitt2024explhuman}. 
Despite the above-discussed limitations of highlight explanations, recent work investigates their role, e.g., of pointers for the model's fallacious reasoning~\citep{malaviya-etal-2024-said,10.1162/TACL.a.46}, grounding the model's confidence estimation~\citep{cruz-etal-2025-improving} or assisting users' decision-making in downstream tasks such as hallucination correction~\citep{ernst-etal-2025-come}. Future explanation systems should exploit and strengthen their advantages as compact cues about the input parts relevant to the model's prediction, whose faithfulness is also easy to assess. One promising direction is already pointed out by the Chapters \ref{chap:NLE_G-Tex} and \ref{chap:NLE_CLUE}, which leverage highlight explanations as guidance for natural language explanation generation, and have inspired recent studies in the same line of work~\citep{wang-atanasova-2025-self,alon-etal-2026-faithful}. However, the completeness and semantic coherence of highlight explanations can still limit their usefulness in such downstream settings. To address this, treating them as a two-way, auditable interface to collect user feedback and better align model behaviour with human expectations is worth exploring in the future.

\subsection{Toward More Faithful Natural Language Explanations}
\label{intro:future_part2}

Building on the foundation established in Part II, Part III (Chapters \ref{chap:NLE_G-Tex}, \ref{chap:NLE_CLUE}) develops methods that use highlight explanations to guide more faithful NLE generation: Paper 3 (\method{}) uses highlight explanations as graph-structured cues and fine-tunes models for NLE generation for complex reasoning tasks; Paper 4 (\modelname{}) leverages span interactions as cues for inference-time steering of uncertainty explanation generation in fact-checking.


Together, these chapters show that highlight guidance can improve NLE faithfulness and other properties such as conciseness and helpfulness. However, two broader limitations of this line of approach emerge. (i) \emph{highlight quality can affect NLE quality}. Chapter~\ref{chap:NLE_CLUE} shows that incomplete span interactions can propagate into the resulting uncertainty explanations.  
Furthermore, as shown in Chapter ~\ref{chap:highlight_context_utilisation}, current highlight attribution methods do not necessarily reveal the source of the model's answer, which may limit their effectiveness as grounding signals. 
(ii) \emph{Faithfulness--Plausibility trade-off persists}. In Chapter 5, the two versions of the proposed \modelname{}\ exhibit this tension: \modelvarianttwo{}\ stays closer to the extracted highlights and achieves higher faithfulness, while \modelvariantone{}\ follows them less rigidly and offers smaller faithfulness gains, but is rated more favourable by humans. This echoes the finding in Chapter~
\ref{chap:highlight_diagnostic} that faithfulness and agreement with
human annotations can favour different span interactive explanations. Mitigating this tension is essential for producing NLEs that provide reliable and understandable insights about the model's decision for users to act upon.
In the following, I elaborate on several promising directions. 

\paragraph{Balancing explanation faithfulness with human utility.} 
Chapters \ref{chap:NLE_G-Tex} and \ref{chap:NLE_CLUE} show the benefits of using highlights as guidance for faithful NLE generation. Future work on mitigating the faithfulness -- utility trade-off could continue this line of research, possibly beginning with improving the quality of the guiding highlights. Specifically, future efforts could address the tension between faithfulness and agreement with human annotations in span interactive explanations by combining the strength of different highlight forms, developing an alignment method to receive human feedback, or grounding in model internals as discussed in Sec~\ref{intro:future_part1}. Beyond using highlights as guidance, chain-of-thought explanations, another free-text explanation form by representing step-by-step reasoning that resembles human reasoning, could offer a natural bridge between model reasoning and human understanding~\citep{wei2022chain,kojima2022large}. 
However, since they can also suffer from faithfulness issues~\citep{NEURIPS2023_ed3fea90,lanham2023measuring,zaman-srivastava-2026-chain}, future efforts could focus on improving their faithfulness while maintaining their human-understandable form. One possible direction is to use CoT to surface aspects of the model's latent reasoning, as recent work suggests models may perform latent reasoning, for example, through hidden states, continuous thought vectors or other latent representations~\citep{deng2023implicit,hao2024training,chen2025reasoningBeyondLanguage}. Efficient techniques that better align explanation generation with answer generation would further benefit this practice. Relatedly, reliable mechanistic tools that identify responsible model components involved in explanation and answer, generalise across common settings, and permit safe interventions would provide an important foundation~\citep{bereska2024mechanistic,yeo-etal-2025-towards,bhan2026neurofaithevaluatingllmselfexplanation}.

\paragraph{Explanations as indicators of model knowledge sources.}
Chapter~\ref{chap:NLE_CLUE} provides an example of eliciting explanations of model uncertainty by referring to conflicts and agreements within the supplied context, guided by highlight explanations. However, Chapter~\ref{chap:highlight_context_utilisation} further shows that a
model's answer may derive from contextual or parametric knowledge, and that highlight explanations are not always reliable indicators of which source was used. Future work should therefore go beyond explaining the source of a model's answer or uncertainty solely in terms of the supplied context. In addition to citing relevant context spans, NLEs should explicitly indicate whether and how parametric knowledge contributed to the prediction, and reflect how conflicts among different knowledge sources shaped the model's answer or uncertainty. A first step towards this goal would be to isolate model-internal signals associated with reliance on the supplied context or parametric knowledge~\citep{islam2025rank2,tighidet-etal-2024-probing,brink2026attriwiki}, and to validate their robustness, coverage, and causal relevance to the model's prediction. Subsequent work should then investigate how these signals can be faithfully translated into understandable, source-aware NLEs.

\paragraph{Explanations across languages and modalities.} 
This thesis focuses on English and text-based explanations, in line with much prior explainability research.  However, to strengthen the practical relevance and broaden the applicability of explanations, future work should consider other languages and modalities. For example, recent multilingual studies show that explanation faithfulness can vary across languages~\citep{zhao-etal-2026-comprehensive}. Such disparities may expose users from different language communities to unequal levels of explanation reliability. Future work should improve the consistency of explanation faithfulness across languages and further investigate how cultural differences shape users' expectations of, and interactions with, explanations. Multimodal models are also increasingly deployed in settings that combine textual and visual information, more closely reflecting the environments in which users encounter AI systems. Prior work has explored textual explanations for interpreting images~\citep{Man_2025_CVPR} and videos~\citep{lee-etal-2025-video}. However, challenges remain in improving the quality of explanations for each modality while ensuring faithful grounding across modalities. Advancing explainability techniques towards diverse user groups and application settings beyond text would improve the practical utility of NLEs in assisting users understand model's decision making process in more realistic contexts.

\subsection{Evaluating Faithfulness and Utility of Free-Text Explanations}
\label{intro:future_part3}
Part IV (Chapters \ref{chap:CoT_FaithMate}, \ref{chap:Human_showme}) evaluates free-text explanations from two perspectives. Chapter \ref{chap:CoT_FaithMate} introduces \our{}, a meta-evaluation framework that studies whether improvements
obtained by optimising a model towards one CoT faithfulness metric can transfer to other faithfulness metrics and paradigms. It shows that contextual and parametric
faithfulness are related but asymmetric, and that improvements within the contextual paradigm do not consistently transfer across metrics.
Chapter \ref{chap:Human_showme} studies explanation utility in an AI-supported fact-checking setting by presenting users with different types of explanations and optional access to supporting evidence, and asking them whether to accept or decline the model's predicted verdict. The results show that, in this setup, evidence is a central
information source for users, while explanations generally operate as complements rather than substitutes for that evidence. 


At the same time, these studies expose several broader limitations in current approaches to evaluating explanation faithfulness and utility. For faithfulness evaluation (Chapter \ref{chap:CoT_FaithMate}), (i) a fundamental challenge is that \emph{faithfulness has no directly observable gold standard}. A model's reasoning path is hidden, and it remains unclear whether there is a single textual reasoning path that could serve as a universal reference. Therefore, although the \our{} framework reveals how gains under one metric transfer to others under model optimisation, it cannot determine which metric most closely captures genuine faithfulness. Recent meta-evaluation frameworks construct operational ground truth under specific protocols~\citep{zaman-srivastava-2025-causal,gurarieh2026bonafideMeta}. However, the resulting faithfulness labels remain defined relative to particular task designs and interventions rather than by a unified ground truth.
(ii) A second limitation is that \emph{current faithfulness metrics primarily assess verbalised reasoning}. For example, they perturb the input, CoT, or model's parametric knowledge associated with a specific reasoning step and test whether the answer changes. Such tests may not distinguish cases in which the model relies on its stated steps from cases in which it reaches the answer through answer-relevant computation that remains unspoken. 

For utility evaluation, Chapter \ref{chap:Human_showme} deliberately assesses the role of explanations and evidence for assisting humans in validating a model's decision when both sources are available to users. This design reveals how these information sources operate together, but does not isolate the standalone utility of each explanation type. As discussed in Chapter \ref{chap:Human_showme}, readily accessible evidence might have compensated for limitations in either the verdict- or uncertainty- explanations, so that both explanation types show similar usefulness under all conditions. Moreover, the controlled, static fact-checking setup might not fully capture how people use explanations in interactive, or higher-stakes workflows involving different forms of evidence and different levels of user expertise. 
These limitations motivate the following directions for future work.

\paragraph{Towards standardised faithfulness evaluation.}
The absence of an independently observable faithfulness CoT means that \our{} should be interpreted as a framework for diagnosing relationships between metrics, rather than as validating any one metric as the definitive measure of faithfulness. More generally, Chapter \ref{chap:CoT_FaithMate} shows that current faithfulness metrics operationalise different notions of faithfulness. Recent evaluation work similarly cautions against assuming that one faithfulness metric generalises across explanation forms, models, and tasks~\citep{lyu2024towards,parcalabescu-frank-2024-measuring,
gurarieh2026bonafideMeta}. Future work should therefore focus less on finding one universal faithfulness score and more on building standards, benchmarks, and meta-evaluation protocols that clarify what each metric measures, where it fails, and when it is appropriate to use. A related direction is to incorporate causally validated mechanistic evidence as a complementary reference, while recognising that current internal-signal
methods remain limited by scalability, reliability, and choices about the appropriate causal units and interventions~\citep{bereska2024mechanistic,yeo-etal-2025-towards,bhan2026neurofaithevaluatingllmselfexplanation}.

\paragraph{Enriching explanations with undisclosed reasoning.} 
The limitation of evaluating only verbalised reasoning raises deeper questions: does a language model have a single, deterministic ``thinking route'' that a generated explanation could recover, or is an explanation only a partial projection of more distributed model computation? Recent work suggests that model reasoning may be distributed across internal states, features, directions, circuits, and other latent processes that are not fully verbalised in the final explanation~\citep{pfau2024lets,chen2026mechanistic,li2025mapping,hu-etal-2026-towards}. Related safety concerns regarding model dishonesty and non-disclosure have also emerged, as models may conceal or fail to disclose answer-relevant reasoning while providing a different justification that conforms to the user's stated preference~\citep{chen2025reasoning,hao-etal-2026-reasoning,shiromani2026hypocrisyGap}.
To move the field towards more faithful expression of model reasoning, future work should distinguish between several questions: whether such internal reasoning traces exist, how they are composed, whether they causally contribute to the answer, and whether generated explanations faithfully reflect or omit them. Addressing these questions will require joint progress in causal analysis of model computation, faithful explanation generation, and evaluation of what explanations omit as well as what they state.

\paragraph{Tailoring utility evaluation to task-, evidence-, and user-centred settings.}
Chapter~\ref{chap:Human_showme} shows that explanation utility is not determined by the explanation text alone. It depends on the other information available to users: in the fact-checking study, evidence was used more frequently than any other information source, and readily accessible evidence may have attenuated some differences between explanation types.
Participants also use evidence to identify explanations that are persuasive but inconsistent with the underlying documents. Future work should therefore evaluate explanations as components of complete decision-support environments rather than as isolated texts. This
includes varying the availability, quality, credibility, and presentation of
evidence; studying users with different expertise; and examining whether
explanations help people identify incorrect outputs, cross-check claims, and
calibrate reliance.
Different explanation formats may support these goals in different ways. For example, explanations in education might need to show how rating criteria are satisfied or not satisfied by a student's response, whereas healthcare explanations may need to connect patient history, clinical evidence, uncertainty, and possible diagnoses. Fact-checkers need explanations that trace evidence, expose information gaps, and clarify uncertainty rather than only providing a final label~\citep{khosravi2022explainable,ghassemi2021false,richmond2024explainable,warren2025explainablefactchecking}. Future work should therefore compare explanation formats in application-grounded settings and design adaptive explanations that can respond to users' evolving questions, expertise, and domain-specific needs~\citep{wu2026usable,liao2020questioning,rong2024towards,kim2024human}.

More broadly, Chapters~\ref{chap:CoT_FaithMate} and~\ref{chap:Human_showme} evaluate model-facing faithfulness and human-facing utility separately. Future work should evaluate the same explanations using both kinds of protocol: improvements under faithfulness metrics should be tested to determine whether they help users identify model errors and calibrate reliance, while explanations perceived as useful should also be subjected to behavioural or mechanistic faithfulness tests. 

Taken together, these directions highlight two complementary priorities for future research: developing explainability techniques that improve multiple dimensions of explanation quality while mitigating trade-offs among them, and designing evaluation protocols that reliably assess these dimensions in relation to the task, evidence setting, and users' needs. Progress on both fronts is necessary to develop explanations that balance competing quality requirements and to establish whether they faithfully represent the model's decision making behaviour while meeting the practical needs of their intended users.

\part{Diagnostic Evaluation of Highlight Explanations}

\chapter{Evaluating Input Feature Explanations through a Unified Diagnostic Evaluation Framework}
\chaptermark{A Diagnostic Evaluation Framework for Input Feature Explanations}
\label{chap:highlight_diagnostic}
This chapter consists of the following published paper\footnote{No changes have been made to the scientific content of the originally published paper. The paper is presented in a single-column
format used throughout the thesis, and its references have been merged into the bibliography at the end of the thesis.}:

\begin{enumerate}
    \item \bibentry{sun-etal-2025-evaluating}.
\end{enumerate}

\clearpage
\begin{abstract}
Explaining the decision-making process of machine learning models is crucial for ensuring their reliability and transparency for end users. One popular explanation form highlights key input features, such as i) tokens (e.g., Shapley Values and Integrated Gradients), ii) interactions between tokens (e.g., Bivariate Shapley and Attention-based methods), or iii) interactions between spans of the input (e.g., Louvain Span Interactions).
However, these explanation types have only been studied in isolation, making it difficult to judge their respective applicability. To bridge this gap, we develop a unified framework that facilitates an automated and direct comparison between highlight and interactive explanations comprised of four diagnostic properties\footnote{Our code can be found at https://github.com/copenlu/A-unified-framework-for-input-feature-explanations}.  
We conduct an extensive analysis across these three types of input feature explanations--each utilizing three different explanation techniques--across two datasets and two models, and reveal that each explanation has distinct strengths across the different diagnostic properties. 
Nevertheless, interactive span explanations outperform other types of input feature explanations across most diagnostic properties. Despite being relatively understudied, our analysis underscores the need for further research to improve methods generating these explanation types. Additionally, integrating them with other explanation types that perform better in certain characteristics could further enhance their overall effectiveness.

\end{abstract}
\section{Introduction}
\begin{figure}
\centering
\small
\includegraphics[width=0.7\textwidth]{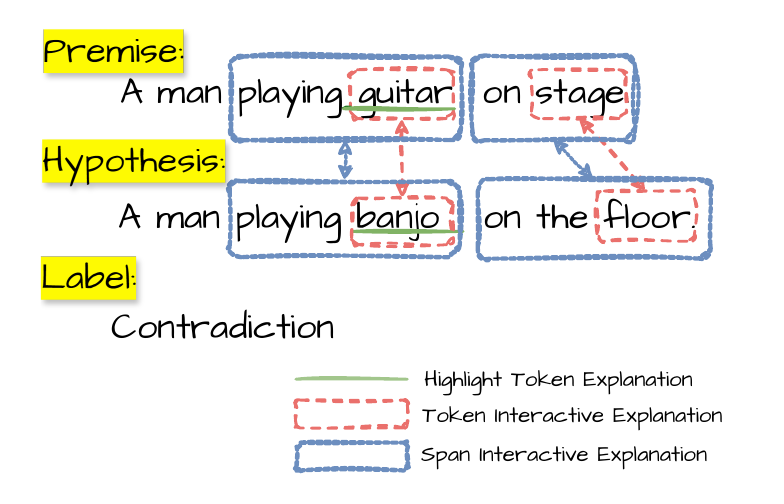}
\caption{\label{p1}An example of the three types of Input Feature Explanations on an instance from the SNLI dataset, with their two most important pieces of explanation (token for \tokenex, token tuple for \tokenintex, span tuple for \spanintex, respectively).
}
\end{figure}
Input feature explanations reveal how a model makes decisions based on a specific input. Among these, the most widely used explanation type is \textit{Token Explanations} (\tokenex), which for Natural language Understanding tasks provide importance scores for the tokens of the input, using methods such as Shapley Values \citep{lundberg2017unified} and Integrated Gradients~\citep{sundararajan2017axiomatic}. However, for complex reasoning tasks that require reasoning across multiple pieces of text, e.g., Fact Checking is performed given a claim and evidence, Natural Language Inference is performed given a premise and a hypothesis, \tokenex\ can be insufficient to capture the relations employed between the different parts of the input. To this end, \textit{Token Interactive Explanations} (\tokenintex) are proposed as another explanation type that provides importance scores for interactions between two tokens of the input with methods such as Bivariate Shapley~\citep{masoomi2023explanations} and Layer-wise Attention Attribution~\citep{ye-etal-2021-connecting}. Further, to enhance the semantic coherence of these interactive explanations, \textit{Span Interactive Explanations} (\spanintex) is an explanation type that provides importance scores for interactions across tuples of spans, e.g., generated by Louvain community detection~\citep{choudhury-etal-2023-explaining}. Figure \ref{p1} 
 showcases these three types of input feature explanations.

It is crucial to develop automated, rigorous and comprehensive evaluation frameworks to ensure the principled selection of the most suitable explanation in a practical application \citep{haeun2024naia} and systematic progress in the development of different types of explanations \citep{atanasova2022diagnostics, jolly2022generating}. However, these three types of input feature explanations have typically been studied in isolation, where explanation methods of the same type are compared \citep{atanasova-etal-2020-diagnostic,deyoung-etal-2020-eraser,janizek2021explaining,choudhury-etal-2023-explaining}. 
Moreover, evaluations of interactive explanations have been restricted to one property.
To address these gaps, we develop \textit{a unified framework that facilitates an automated and direct comparison between highlight and interactive input feature explanations on a suite of four diagnostic properties}. The framework allows for a rapid, scalable comparison across explainability methods, essential for the growing number of new techniques. Using the framework, we then perform an \textit{extensive empirical analysis of the properties of existing explanation methods across all three explanation types}.

\textbf{Unified Evaluation Framework.} Our unified evaluation framework consists of four essential diagnostic properties -- Faithfulness, Agreement with Human Annotation, Simulatability, and Complexity. They are the most widely used for \tokenex\ \citep{nauta2023anecdotal} and include the only property used for interactive explanations -- Faithfulness. \textit{Faithfulness} \citep{chen-etal-2020-generating-hierarchical,chen-etal-2021-explaining,choudhury-etal-2023-explaining} measures the extent to which explanations accurately reflect the reasons used by the model in its predictions. \textit{Agreement with Human Annotation} \citep{atanasova-etal-2020-diagnostic} evaluates whether explanations exhibit an inductive bias akin to human reasoning, potentially enhancing their plausibility to end users. \textit{Simulatability} \citep{pruthi-etal-2022-evaluating} estimates whether the explanations are useful to an agent for replicating the model's decisions. Finally, \textit{Complexity} \citep{10.5555/3491440.3491857} evaluates whether the explanations are easy to comprehend. 
In the unified evaluation framework, we further extend the four properties to facilitate their application and comparison across all three explanation types (\S\ref{sec:methodology:overall}).

\textbf{Extensive Empirical Analysis of Input Feature Explanations.} We conduct an extensive analysis across two different textual tasks, two language models, and three explanation techniques for each input feature explanation type. Our findings indicate that \tokenex\ exhibit greater Comprehensiveness, and \spanintex\ -- Sufficiency.
Additionally, \spanintex\ and \tokenintex\ align more closely with human annotations at the token level than \tokenex. Moreover, \spanintex\ demonstrate the highest interaction overlap with human annotations. Further, \spanintex\ are found to be most beneficial for Simulatability. Finally, our results suggest that \spanintex\ and \tokenex\ are easier to comprehend.

Overall, our analysis highlights the strengths of each explanation type across various diagnostic properties, with \spanintex\ generally outperforming others on most measures. However, we observe a trade-off between Comprehensiveness and plausibility, particularly with \spanintex, underscoring the need for methods that enhance both. Future research could explore integrated approaches that combine explanation types to optimize performance across all diagnostic properties.

\section{A Unified Automated Evaluation Framework for Highlight Explanations}
\label{sec:methodology:overall}
According to the established taxonomy of evaluation approaches for explainability methods, evaluations can be functionally grounded, human-grounded, or application-grounded~\citep{doshivelez2017towards}. Our work focuses on automated functionally-grounded evaluation, which serves as a foundation for the other evaluation types as it enables the rapid, scalable comparison across explainability methods, essential for the growing number of new techniques. To this end, we present a unified framework comprised of four widely employed diagnostic properties: Faithfulness, Agreement with Human Annotations, Simulatability, and Complexity. This section formally introduces them and outlines the extensions that allow for their application across different input feature explanation types. 
\subsection{Preliminaries}
\label{preliminaries} 
We start with a dataset $D$, and a model $M$ fine-tuned on $D$. An instance $x \in D$ comprises two parts, e.g., a claim and an evidence, the first consisting of $m$ tokens, and the second -- of $n$ tokens.
We apply an explanation attribution method $A_{E}$ of type $E\in \{\tokenex, \tokenintex, \spanintex\}$ to $M$, and each $x \in D$: $A_{E}(M, x) = \{e^x_k, a^x_k | k \in [0, s-1]\}$, where $e^x_k$ is a token/pair of tokens/pair of token spans from the input and $a^x_k$ denotes its importance score. $k$ is the importance ranking of the corresponding piece of explanation. $s$ is an upper limit for the number of most important pieces of explanation for an instance, such that: $s \in [1,m+n]$ for \tokenex, $s \in [1, m \cdot n]$ for \tokenintex, and for \spanintex\ $s$ varies for each instance with $s \in [1, f], f < m! \cdot n!$. Depending on the explanation type $E$, $e^x_k$ can consist of:
one token $x_i$ for \tokenex, $i \in [0, m+n-1]$,
one token pair $(x_p, x_q)$ for \tokenintex, where $p \in [0, m-1]$ and $q \in [m, m+n-1]$,
one span pair $((x_s, \dots, x_{s+l_1}), (x_t, \dots, x_{t+l_2}))$ for \spanintex, where $s, s+l_1 \in [0, m-1]$ and $t, t+l_2 \in [m, m+n-1]$.

Considering a particular threshold for the number of most important explanation pieces $s$, we compute the total set of input tokens involved in the presented explanation for $x$:
\begin{equation}
\small
    T_{A_E,M}(x,s)=\{x_i | x_i \in {e}_{k}^{x}, k \in [0, s-1] \}
\label{threshold_0}
\end{equation}
However, as noted above, the upper bound for $s$ can vary across input feature explanations. Additionally, the number of tokens included in the top-$k$ important explanations can differ substantially among explanation types -- in top-1 we can have: 1 token for \tokenex, 2 tokens for \tokenintex, and more than 2 tokens for \spanintex. This variability makes it difficult to compare results across different explanation types. To this end, we propose extensions for each of the studied diagnostic properties that result in \textit{unified diagnostic properties} applicable for a direct comparison of the different types of input feature explanations.

\subsection{Faithfulness}
\label{sec:higlight_faithfulness:method}
Faithfulness \citep{deyoung-etal-2020-eraser} assesses whether explanations accurately reflect the model’s decision-making process. It involves two aspects -- Sufficiency and Comprehensiveness -- measured as the number of the model's prediction changes after keeping (SP) or omitting (CP) $k$ most important portions of the input (see omission details in \S\ref{sec:app:faithfulness-mask-details}): 
\begin{equation}
\small
    CP(x,A_E,k)=\left\{
    \begin{array}{ll}
    0, & \text{if } f(x-T_{A_E,M}(x,k))=f(x) \\
    1, & \text{otherwise }
    \end{array}
    \right\}
\label{eq:comp-1}
\end{equation}
\begin{equation}
\small
    SP(X,A_E,k)=\left\{
    \begin{array}{ll}
    1, & \text{if } f(T_{A_E,M}(x,k))=f(x) \\
    0, & \text{otherwise }
    \end{array}
    \right\}
\label{eq:comp-3}
\end{equation}
To take different thresholds $k$, we average over $k \in [0, s=m+n-1]$. We then also average the results across instances within $D$ to compute the final Comprehensiveness and Sufficiency scores:
\begin{equation}
\small
Comp_{A_E}= \frac{\sum_{x\in{D}}^{|D|} \sum_{k=0}^{s} CP(x,A_E,k)}{s*|D|}
\label{eq:comp-5}
\end{equation}
\begin{equation}
\small
Suff_{A_E} = \frac{\sum_{x\in{D}}^{|D|} \sum_{k=0}^{s} SP(x,A_E,k)}{s*|D|}
\label{eq:comp-6}
\end{equation}

\textbf{Unified Faithfulness.} To facilitate the comparison of faithfulness scores, we introduce a \textit{dynamic threshold} $\theta_{x,k}$. It represents the number of unique tokens used for a perturbation on $x$, same across explanation methods of all explanation types. Since top-$k$ explanations for $A_E$ of type \spanintex\ typically contain more tokens than for \tokenex\ or \tokenintex, we set $\theta_{x,k}$ across the explanation methods of all types to:
\begin{equation}
\small
\theta_{x,k} = \left|T_{A_{\spanintex},M}(x,k)\right|
\label{eq:threshold_1}
\end{equation}
Thus, $\theta_{x,k}$ becomes a dynamic threshold that adapts based on each instance’s top-$k$ explanation tokens from $A_{\spanintex}$. We then adjust the number of top-$k$ explanations selected from $A_{\tokenex}$ and $A_{\tokenintex}$, $k_{A_{\tokenex}}$ and $k_{A_{\tokenintex}}$, correspondingly, to result in the same number of perturbed tokens $\theta_{x,k}$:
\begin{equation}
\small
\left|T_{A_{\tokenex},M}(x,k_{A_{\tokenex}}(x))\right| = \theta_{x,k}
\label{eq:threshold_2}
\end{equation}
\begin{equation}
\small
\left|T_{A_{\tokenintex},M}(x,k_{A_{\tokenintex}}(x))\right| = \theta_{x,k}
\label{eq:threshold_3}
\end{equation}
Furthermore, a \textit{Random baseline} is established, where the tokens for perturbation are selected randomly to match the average $\theta_{x,k}$ across $D$.

\subsection{Agreement with Human Annotations}
\label{sec:agreement:method}
Agreement with Human Annotations has been used to assess the overlap between generated and human-annotated explanations, which can indicate the plausibility of the generated explanations to end users. For $E=\tokenex$, the measure is computed by calculating a precision score for the top-$k$ most important explanations compared to the gold human annotations~\citep{atanasova-etal-2020-diagnostic}. 

For $E=\tokenex$, $a_i^x, i \in [1,m+n]$ is the attribution score of $i$th most important explanation for $x$. $s=m+n$ is the number of explanations extracted from $x$. Corresponding to each explanation's attribution score, $s$ thresholds are set, forming the threshold list $\omega_{A_E}(x)=[a^x_0,...,a^x_s]$.
By selecting explanations with higher attribution scores than each threshold in $\omega_{A_E}(x)$, $s$ targeted explanation sets are obtained, where $C_{A_E}(x,i)\{e^x_j:a^x_j<=a^x_i\}$ represents the set for the $i^{th}$ threshold, $a^x_j$ is the attribution score of token $e^x_j$ for $E=\tokenex$.

Comparing these sets with the golden explanation set $e^G$, $s$ precision-recall pairs $P_{i}/R_{i}(x,e^G, A_E)$ can be derived. Average Precision (AP) is then obtained by weighting $P_i$ with the corresponding $R_i$ increase:
\begin{equation}
\small
P_{i}/R_{i}(x,e^G, A_E)=Pre/Rec(C_{A_E}(x,i),e^G)
\end{equation}
\begin{equation}\textstyle
\small
AP_{A_E}(x, e^G)=\sum_{i=0}^{s} (R_{i}-R_{i-1})*P_{i}
\end{equation}
Mean AP (MAP) is calculated for all $x\!\in\!D$:
\begin{equation}
\small
MAP_{A_E}=\frac{\sum_{x\in D}^{|D|} AP_{A_E}(x, e^G)}{|D|}
\label{eq:map}
\end{equation}
\textbf{Unified Agreement with Human Annotation Measure.} 
For a fair comparison between the different types of explanations, the thresholds $\omega_{A_E}(x)$ for including the same number of tokens across the explanation methods follows the procedure set for the Unified Faithfulness (\S\ref{sec:higlight_faithfulness:method}). 

Furthermore, we measure Agreement with Human Annotations at the \textbf{interaction level} for the gold $\spanintex$/$\tokenintex$ explanations and at the \textbf{token level} for gold $\tokenex$ explanations.   

\textbf{Interaction-level Agreement with Human Annotations.} For a fair comparison between $\tokenintex$ and $\spanintex$ methods, we adapt $MAP_{\tokenex}$ to the interaction level. Specifically, we compute the mean average precision (Eq. \ref{eq:map}) w.r.t. the human-annotated $\tokenintex$/$\spanintex$ sets.

\textbf{Token-level Agreement with Human Annotations.} For a fair comparison between $\tokenex$ and $\tokenintex$/$\spanintex$ methods, we extract tokens from $\tokenintex$/$\spanintex$ and compare them with tokens from golden $\tokenintex$/$\spanintex$ sets. 
To compute MAP at token level, we follow the similar procedure set for $E=\tokenex$ (Eq. \ref{eq:map}) with threshold lists $\omega_{A_{\tokenintex/\spanintex}}(x)$, but the targeted sets $C_{A_{E_{token}}}(x, i)$ contain tokens extracted from $\tokenintex$/$\spanintex$ methods. The golden set $S_{A_{E_{token}}}(x)$ aggregates tokens from golden $\tokenintex$/$\spanintex$ sets.

We also set a \textit{Random baseline}, where the number of randomly selected span pairs, token pairs, or tokens for each instance matches the average number of tokens per instance extracted with a \spanintex\ explanation method.

\subsection{Simulatability}
\label{sec:simulatability:method}
Simulatability was initially proposed to measure how accurately humans can predict a model's outputs based on its explanations \citep{chen2023models,hase-etal-2020-leakage}. Previous studies demonstrated that Simulatability can be approximated using an automated agent model as a surrogate for human understanding \citep{pruthi-etal-2022-evaluating}. \textit{Given the established positive correlation between Simulatability and human evaluation of explanation utility}, we integrate the Simulatability scores obtained from an agent model with different explanation types to approximate their utility for humans.

Following existing work \citep{hase-etal-2020-leakage}, we train an agent model $AM$, sharing the same architecture as the original model $M$, to simulate $M$'s predictions $Y'$ using produced explanations. During $AM$'s training phase, we extract the top-$k$ explanations and incorporate them in the input. 
In comparison, another agent model, $AM_O$, is trained without explanation guidance as a baseline on the same training set.
During the testing phase, the simulation accuracy of $AM$ and $AM_O$ over the shared dataset $D$ is calculated.\footnote{While existing work notes that incorporating natural language explanations in the testing phase could leak the predicted label \citep{hase-etal-2020-leakage}, we use only input feature explanations that do not contain additional information.} The difference between the accuracies is interpreted as the explanation's effect in enhancing the simulatability of $M$:
\begin{equation}
\small
    Sim=ACC(AM(D),Y')-ACC(AM_O(D),Y')
\end{equation}

\paragraph{Unified Simulatability.} To compare the simulation utility of different explanation types, we train a separate agent model $AM_E$ for each explanation method $A_E$ and calculate the corresponding simulation performance on the common test set.
For a fair comparison across the different explanation method types $A_E$, we first ensure top-$k_{E}$ explanations are presented for assisting the agent's training for $A_E$, following Section \ref{sec:higlight_faithfulness:method}. This ensures each model is exposed to the same quantity of unique tokens from different explanation types.

During the training of $AM$, we introduce the explanations from $A_E$ into the learning of $AM_{A_E}$; we supplement $x$ with top-$k_{A_E}$ explanations instead so that the agent model is trained with the same mechanism whether the explanations are provided or not, and each training instance will contain the input sequence $x_{A_E}$ and golden label $Y'$ which is predicted by the original model $M$. Specifically, we examine two different ways of presenting the explanations as part of the original input sequence, $I_{Symbol}$ and $I_{Text}$ (see \S\ref{sec:app:detailed-form}.); all aim to ensure the explanations of different types are inserted similarly. 

At test time, the F1 scores of agent models $AM_{E}$ and $AM_{O}$ over $D$ are calculated: 
\begin{equation}
\small
    SF_{E}=F1(AM_{E}(x_{E}),Y'),x \in D
\end{equation}
\begin{equation}
\small
    SF_{O}=F1(AM_{O}(x),Y'),x \in D
\end{equation}
Note that the explanations from $A_E$ are also provided to the input for the unseen instances for agent model $AM_{E}$. The final simulation metric then indicates how much this specific type of input feature explanation enhances the model's simulatability:
\begin{equation}
\small
    RSF_{E}=SF_{E}-SF_{O}
\end{equation}

\subsection{Complexity}
Feature attribution explanations are designed to aid human understanding of a model's reasoning over specific instances. However, since humans have a limited capacity to process large amounts of information simultaneously, these explanations need to be easy to comprehend. Even if we select only the top-$k$ attributions with the highest importance scores, they need to be distinctive as opposed to the attribution scores having a uniform distribution. \citet{10.5555/3491440.3491857} propose to measure the complexity of a produced explanation with entropy \citep{renyi1961measures} over the attribution scores of all the produced explanations by $A_{\tokenex}$ method:
\begin{equation}\textstyle
\small
        P(x,p) =\left|a^x_p\right| / \sum_{q=1}^{m+n}\left|a^x_q\right|
\end{equation}
\begin{equation}\textstyle
\small
        CL(x) =- \sum_{p=1}^{m+n}
        P(x,p)
        \ln(P(x,p))
\end{equation}
$m+n$ is the total number of generated $A_{\tokenex}$ explanations, and all explanations are considered for the complexity score computation. Higher entropy means different features have similar attribution scores, where the simplest explanation, with low entropy, would be concentrated on one
feature.

\textbf{Unified Complexity.} 
To ensure a fair comparison across different types of explanation methods $A_E$, we maintain consistency in the size of the chosen explanation lists across all $A_E$ for the same instance, denoted as $k_{x}$, as the number of generated $A_{\tokenex}$/$A_{\tokenintex}$/$A_{\spanintex}$ explanations originally vary for the same $x$. The complexity score $CL_{A_E}(x,k_{x})$ of the top-$k_{x}$ explanation list under method $A_E$ is calculated as:
\begin{equation}\textstyle
\small
        P_{A_E}(x,k_x,i) =\left| a^x_i\right| /\sum_{j=1}^{k_{x}}\left|a^x_j\right|
\end{equation}
\begin{equation} \textstyle
\small
        CL_{A_E}(x,k_{x}) =- \sum_{i=1}^{k_{x}}
        P_{A_E}(x,k_x,i)
        \ln(P_{A_E}(x,k_x,i))
\end{equation}
where $a^x_i$/$a^x_j$ represent the $i$/$j$th highest attribution score from the explanation set for $x$.

The final complexity score is an average of $CL_{A_E}(x,k_{x})$ across all $x \in D$:
\begin{equation}\textstyle
\small
CL_{E}={\sum_{x\in D}^{\left|D\right|}CL_{E}(x,k_x)}/{\left|D\right|}
\end{equation}
Notably, $k_{x}$ is calculated from the number of explanations produced by method $A_{\spanintex}$ for $x$, which varies based on the span interaction extraction method and is known only after generation.

\section{Experimental Setup}
\subsection{Datasets}
\label{Datasets}
We select the natural language inference dataset SNLI~\citep{bowman-etal-2015-large}, where instances consist of a premise, hypothesis, and a label $y\in \{entailment, neutral, contradiction\}$. 
%
Additionally, we select the fact checking dataset FEVER~\citep{thorne-etal-2018-fever,atanasova-etal-2020-generating}, where instances consist of a claim, evidence, and a label $y\in \{entailment, neutral, contradiction\}$.\footnote{
\url{https://huggingface.co/datasets/copenlu/fever_gold_evidence}} We generate input feature explanations by sampling 4,000 instances from each train, dev, and test set, due to the high computational cost, particularly for Shapley-based explanations \citep{atanasova-etal-2020-diagnostic}.
For Agreement with Human Annotations property, we use existing human explanation annotations for SNLI and FEVER (see \S\ref{sec:app:dataset-details}).
 
\subsection{Input Feature Explanation Methods}
\label{sec:PosthocExplainabilityTechniques}
To generate importance scores, we first select three common \tokenex\ techniques -- Shapley~\citep{lundberg2017unified}, Attention~\citep{deyoung-etal-2020-eraser}, and Integrated Gradients (IG, \citet{sundararajan2017axiomatic}). For \tokenintex, we employ Bivariate Shapley~\citep{masoomi2023explanations}, Attention~\citep{clark-etal-2019-bert}, and Layer-wise Attention Attribution~\citep{ye-etal-2021-connecting}. Notably, the \tokenintex\ techniques are the bivariate version of the techniques used for generating \tokenex; e.g. Layer-wise Attention Attribution uses IG.

Following~\citet{choudhury-etal-2023-explaining}, we apply the Louvain algorithm~\citep{blondel2008fast} for each of the three selected \tokenintex\ to generate importance scores for \spanintex\ methods, where the importance score of each span interaction is averaged over the importance scores of the token interactions within it. We will refer to Shapley, Attention, and IG as the explanation base types used for generating all types of input feature explanations for brevity. See \S\ref{sec:app:explainability_techniques} for more details.

\subsection{Models}
\label{Models}
We use the BERT-base-uncased model~\citep{devlin-etal-2019-bert} with 12 encoder layers, and the BART-base-uncased model~\citep{lewis-etal-2020-bart} with 6 encoder and 6 decoder layers, as common representatives of the encoder and the encoder-decoder Transformer architecture \cite{NIPS2017_3f5ee243}. Our choice is particularly influenced by the substantial computational requirements of the input feature explanations, especially pronounced for Shapley \citep{atanasova-etal-2020-diagnostic}. Additionally, our choice is guided by the need to directly access the models' internals for generating IG and Attention-based explanations. Furthermore, while our framework currently utilizes the said models, it is designed to be easily adaptable to other models or newly developed explainability techniques, provided that there are more robust computational resources available. 

We fine-tune the base models on SNLI and FEVER and use them to generate explanations. Their performance on the test sets is shown in \S\ref{sec:app:base-model}. For the Simulation property, we follow existing work \citep{fernandes2022learning,pruthi-etal-2022-evaluating} and train simulator agent models (\S\ref{sec:simulatability:method}) with the same architectures as the base ones. Following \citet{fernandes2022learning}, we split the test set into train/dev/test for the training of the agent model.

\section{Results and Discussion}

\begin{figure*}[t]
\centering
\begin{subfigure}[t]{0.49\textwidth}
    \centering
    \includegraphics[width=\linewidth]{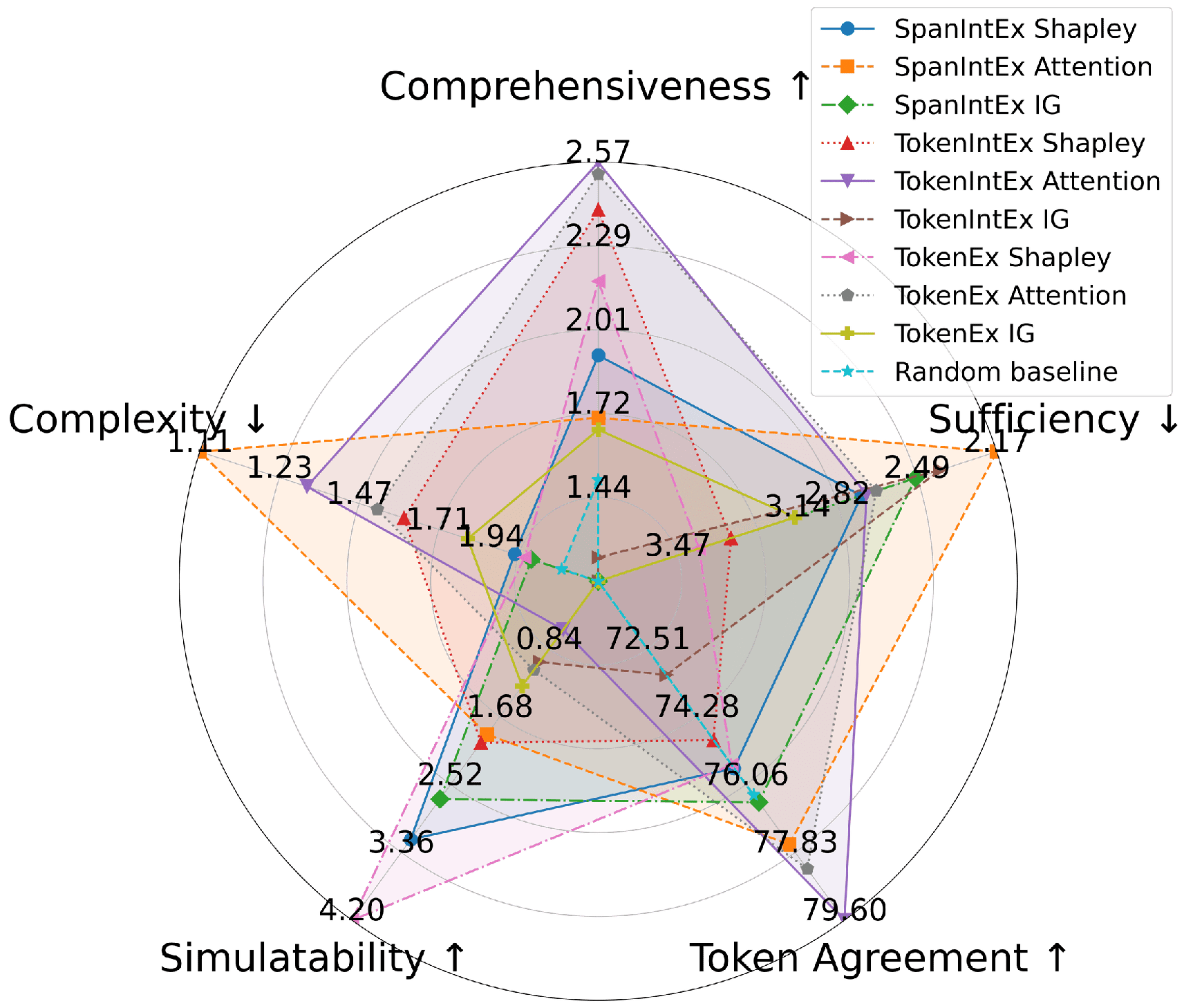}
    \caption{SNLI dataset, BERT model.}
    \label{fig:subfig:a}
\end{subfigure}
\hfill
\begin{subfigure}[t]{0.49\textwidth}
    \centering
    \includegraphics[width=\linewidth]{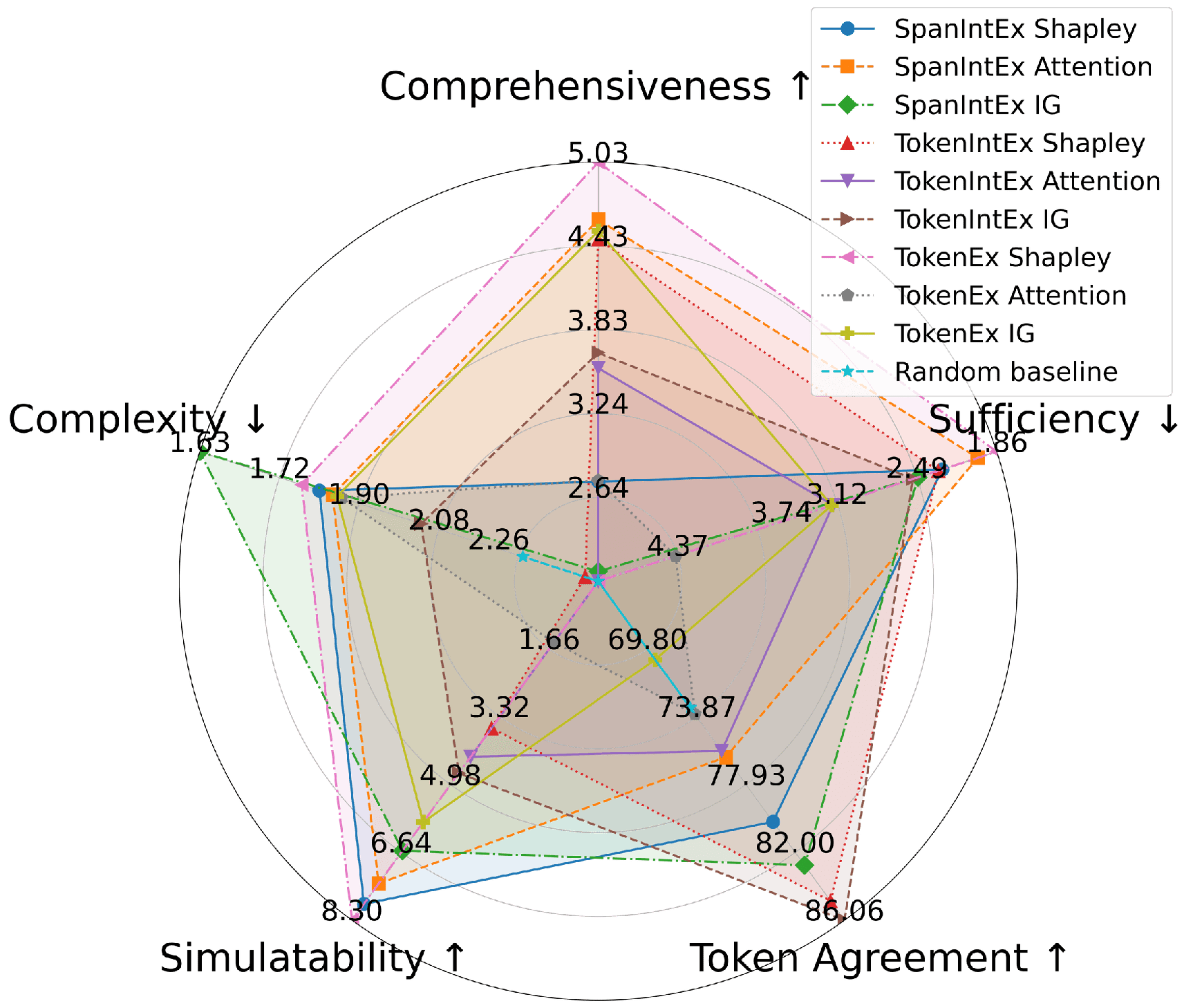}
    \caption{SNLI dataset, BART model.}
    \label{fig:subfig:b}
\end{subfigure}

\medskip

\begin{subfigure}[t]{0.49\textwidth}
    \centering
    \includegraphics[width=\linewidth]{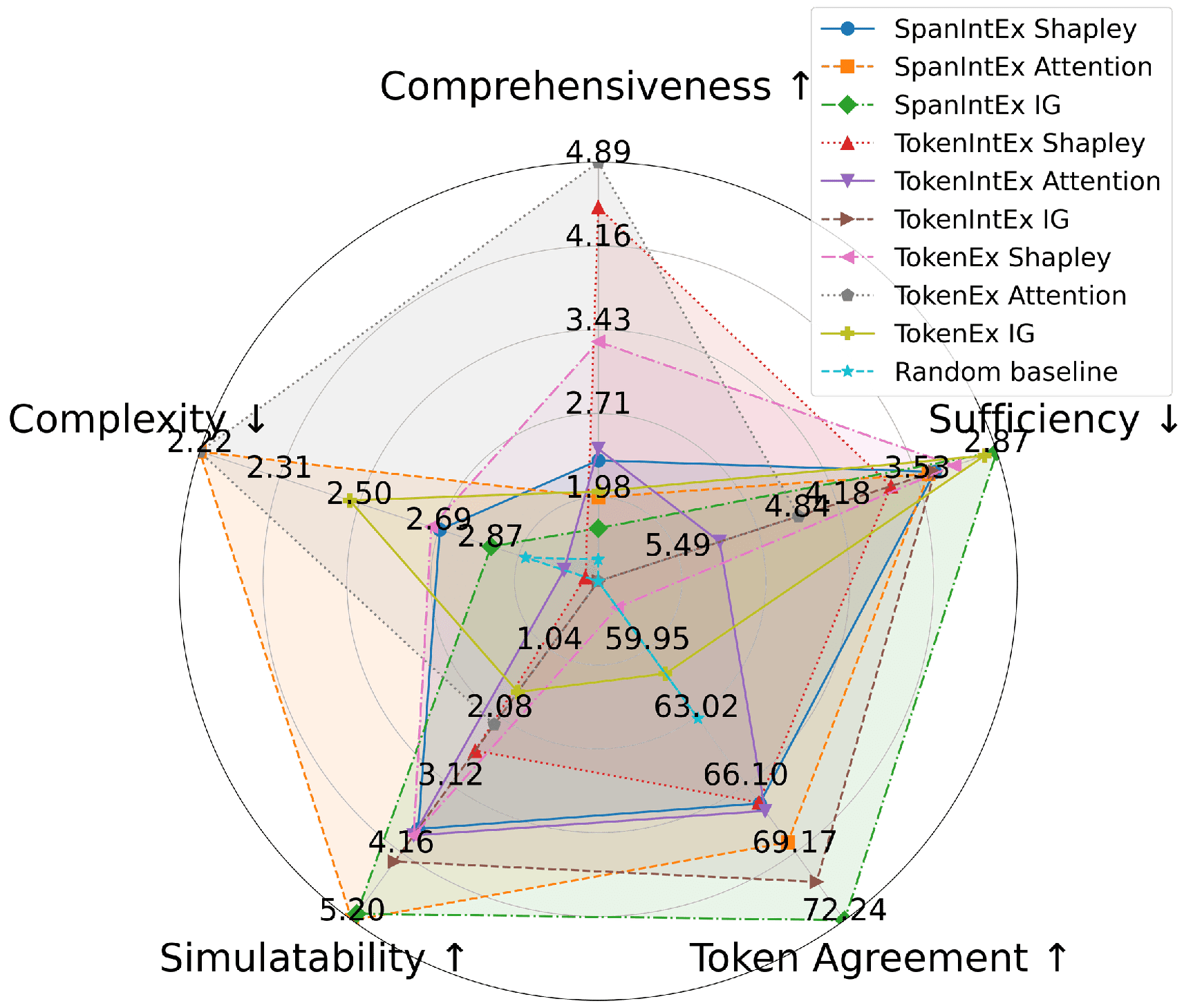}
    \caption{FEVER dataset, BERT model.}
    \label{fig:subfig:c}
\end{subfigure}
\hfill
\begin{subfigure}[t]{0.49\textwidth}
    \centering
    \includegraphics[width=\linewidth]{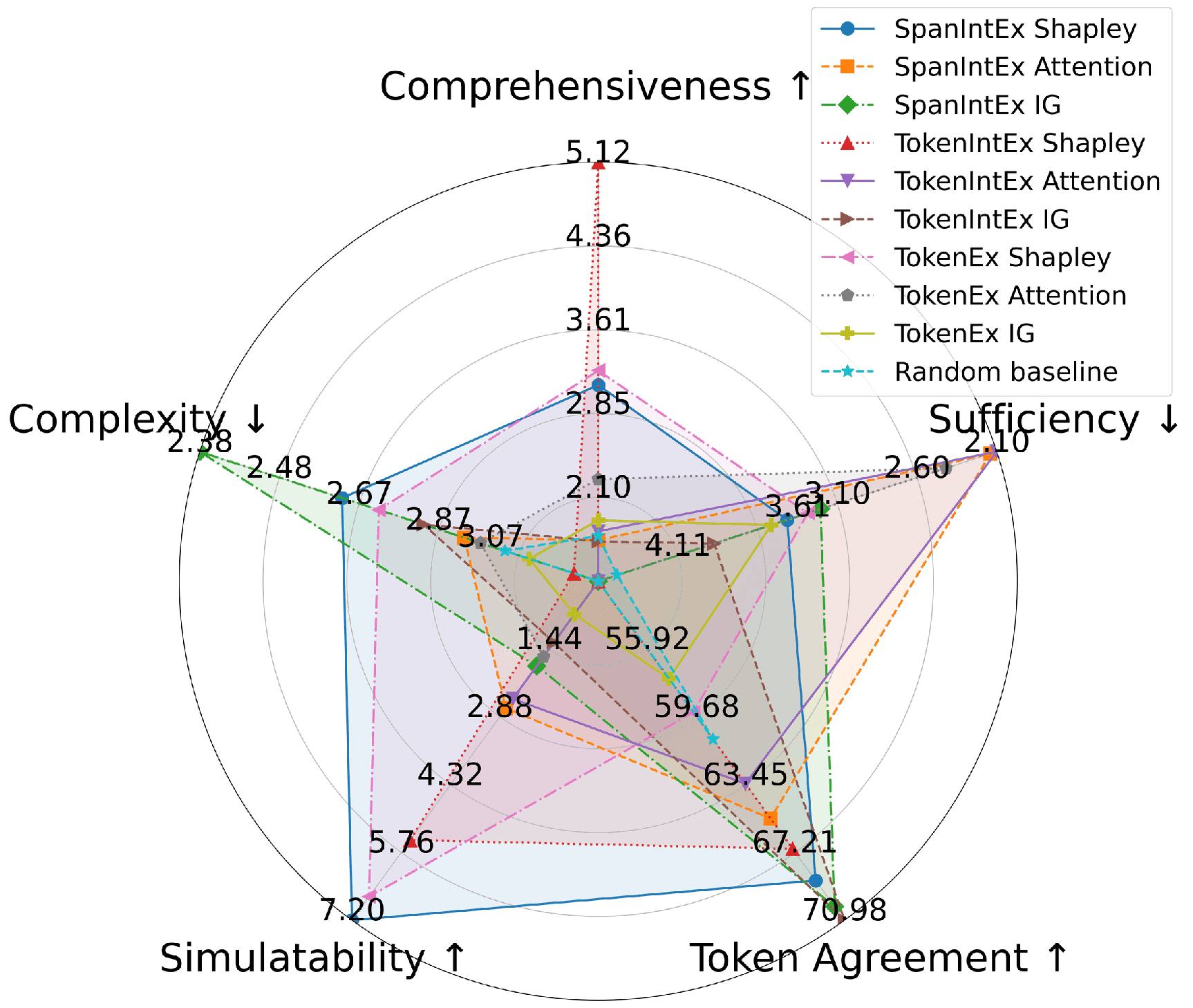}
    \caption{FEVER dataset, BART model.}
    \label{fig:subfig:d}
\end{subfigure}
\caption{Unified evaluation framework (\S\ref{sec:methodology:overall}) results for all feature attribution methods (\S\ref{sec:PosthocExplainabilityTechniques}).}
\label{fig:mainfigure}
\end{figure*}

We now present the results of our unified evaluation framework (\S\ref{sec:methodology:overall}) illustrated in Fig. \ref{fig:mainfigure}. They include explanation methods of types \spanintex, \tokenintex, and \tokenex\ (\S\ref{sec:PosthocExplainabilityTechniques}), two models (\S\ref{Models}), two datasets (\S\ref{Datasets}), and three base explanation techniques per explanation type (\S\ref{sec:PosthocExplainabilityTechniques}). For Simulatability, we select the results of $I_{sym}$, as this form avoids repeating the input text and increasing the input size substantially. For Agreement with Human Annotations, we select the Token-level results as they are present for all explanation types. \S\ref{sec:app:detailed} lists detailed results per property.

\subsection{\textbf{Faithfulness}}
\label{result:faithfulness}
\textbf{Unified Comprehensiveness}. Across both datasets and models, \textbf{\tokenex\ and \tokenintex\ are identified as the most comprehensive explanation types}, achieving the highest scores in 7/12 and 5/12 cases, respectively. \spanintex, designed to enhance the semantic coherence of interactive explanations by including additional context, often incorporates tokens that do not directly contribute to the model's prediction, thus explaining its lower comprehensiveness scores. Compared to the random baseline, \tokenex\ and \spanintex\ always outperform it, while \tokenintex\ mostly underperform it when based on IG. Across the base explanation techniques, \tokenex\ performs best when based on Attention for BERT and on Shapley for BART, indicating that \textbf{different base explanation techniques can perform better for different architectures}. Both \tokenintex\ and \spanintex\ 
show optimal performance when based on Shapley and Attention. Overall, the results indicate a \textit{stronger performance of Attention and Shapley over IG} across all explanation types. 

\textbf{Unified Sufficiency.} 
\textit{\spanintex\ ranks as the most sufficient explanation type in 7/12 cases}, surpassing \tokenex, which performs well in only 3/12 cases.  While contrary to \spanintex\ Comprehensiveness performance, we attribute this to the semantic coherence of the extracted top spans, which provide more meaningful information. 
Note that while Sufficiency is highly desirable, Comprehensiveness is not required in all downstream applications as end-users prefer simpler, more general explanations with fewer causes \citep{thagard1989explanatory}.
Unlike \tokenex, which consistently outperforms the random baseline, \tokenintex\ and \spanintex\ struggle to outperform it on FEVER, likely due to the longer input, posing challenges for the explanations to accurately unveil the model’s internal processes. 
The results from different base explanation techniques show no clear trends, indicating a \textbf{significant variability stemming from the specific dataset and model architecture}.

\subsection{\textbf{Agreement with Human Annotation}}
\label{result:agreement}
\spanintex\ and \tokenintex\ show higher agreement scores with human annotations than \tokenex. Similarly, \spanintex\ consistently achieve higher agreement with human interaction-level annotators, especially when based on Attention scores (see 
\S\ref{sec:app:detailed}). This indicates that \textbf{\spanintex\ are more plausible to humans due to their enhanced semantic coherence}. In contrast, \tokenex\ often score lower than the random baseline. Moreover, considering \spanintex's lower performance in Comprehensiveness, there emerges \textbf{a distinct trade-off between Comprehensiveness and Agreement with Human Annotations}. Across base explanation techniques, IG performs best for FEVER; IG and Attention -- for SNLI. 
In addition, we find that for the interaction-level agreement, \tokenintex\ and \spanintex\ perform worst when based on Shapley. The lower agreement results for Shapley compared to its better results on Comprehensiveness again indicate an existing trade-off between the two properties.

\subsection{\textbf{Simulatability}}
\label{result:simulatability}
Our results show that \spanintex\ achieve the highest simulatability in 9/12 cases, helping the agent model accurately reproduce the original model's prediction. This again underscores the \textbf{critical role of contextual information and enhanced semantic coherence provided by \spanintex}. Notably, providing \spanintex\ to agents improves their ability to simulate the original model by up to 7.9 F1 points compared to without explanations. Among base explanation techniques, IG consistently performs best for \spanintex; other techniques do not exhibit a clear trend.

\subsection{\textbf{Complexity}}
\label{result:comlexity}
\spanintex\ and \tokenex\ generally achieve similar complexity, which consistently remains lower than those of \tokenintex. This suggests that \textbf{\tokenex\ and \spanintex\ generate more distinctive attribution scores, potentially making them easier for humans to understand}. 
Regarding the base explanation techniques, Attention consistently yields the best complexity scores for BERT across all explanation types. There is no clear trend for BART. Additionally, \tokenintex\ frequently underperform the random baseline, highlighting its complexity (see \S\ref{sec:app:detailed}).

\subsection{Overall}
In summary, we find that while \tokenex\ and \tokenintex\ generally provide more comprehensive insights, \spanintex\ performs better in Sufficiency due to its enhanced semantic coherence (\S\ref{result:faithfulness}). 
This calls for \textit{better methods for generating \spanintex\ that are both comprehensive and sufficient}. 
Additionally, there is a trade-off between Comprehensiveness and Agreement with Human Annotations (\S\ref{result:agreement}), suggesting that the most faithful explanations might be less plausible to end users. This highlights the \textit{need for advanced methods to boost both the Comprehensiveness and plausibility} of \spanintex\, possibly leveraging the advantage of \tokenex. Furthermore, \spanintex\ significantly improves simulatability by allowing agents to accurately replicate model decisions (\S\ref{result:simulatability}), which is crucial in practice. Finally, the complexity analysis (\S\ref{result:comlexity}) shows that \spanintex\ and \tokenex\ are potentially easier to comprehend than \tokenintex\ when considering the importance score distribution. 

Overall, our results highlight the differences between the different types of input feature explanations, with \spanintex\ outperforming others on most measures. As no one type performs best on all properties, we \textit{call for the development of combined methods that can leverage the strength of the different explanation types and potentially lead to an overall improvement of the explanation utility}.

\section{Related Work}

\textbf{Input Feature Explanations.} Considerable research exists on extracting explanations for input data.
Methods like perturbation-based attribution (e.g., Shapley~\citep{lundberg2017unified}), attention-based methods (e.g., Attention~\citep{jain-wallace-2019-attention,serrano-smith-2019-attention}), and gradient-based methods (e.g., Integrated Gradients~\citep{sundararajan2017axiomatic}) are prevalent for highlighting individual tokens~\citep{atanasova-etal-2020-diagnostic}. As individual tokens might be insufficient to explain the model, many attribution methods have been extended to bivariate forms~\citep{masoomi2023explanations,janizek2021explaining,ye-etal-2021-connecting} to capture input token interactions. More recent work has explored how interactions between groups of tokens collectively contribute to model reasoning~\citep{choudhury-etal-2023-explaining,chen-etal-2021-explaining}. Unlike other work where token groups might consist of tokens from arbitrary positions, \citet{choudhury-etal-2023-explaining} explicitly capture span interactions, enhancing the comprehensiveness of explanations by containing the entire spans. 

\textbf{Automated Explanation Evaluation.} For evaluating \tokenex, \citet{deyoung-etal-2020-eraser,atanasova-etal-2020-diagnostic} propose metrics to measure how faithful explanations are to the model's inner reasoning. They also assess the plausibility of explanations to humans by measuring the agreement of \tokenintex\ with human annotations. To assess the utility of explanations to humans, \citet{pruthi-etal-2022-evaluating} propose to use an agent model as a proxy for humans and evaluate whether explanations aid in model simulatability. Complexity \citet{10.5555/3491440.3491857} measures the distribution of attribution scores of \tokenex\ and assesses whether the key tokens in token explanations are easily comprehensible to humans. 
To evaluate \tokenintex\, most works adopt the faithfulness or axiomatic/theoretical path~\citep{NEURIPS2020_443dec30,sundararajan2020shapley,janizek2021explaining}.
Current work on evaluating \spanintex\ has primarily focused on faithfulness \citep{choudhury-etal-2023-explaining}. However, since \spanintex, \tokenintex, and \tokenex\ contain varying amounts of tokens, which, e.g., affects the faithfulness test, this makes direct comparisons between different explanation types using existing metrics challenging. To our knowledge, no prior paper has involved all types of input feature explanations within a unified evaluation framework.

\section{Conclusion}
We introduced a unified evaluation framework for input feature attribution analysis to guide the principled selection of the most suitable explainability technique in practical applications. Our analysis outlines the diverse strengths and trade-offs among \tokenex, \tokenintex, and \spanintex. Our findings particularly underscore \spanintex's superior performance in Sufficiency, agreement with human inductive biases, its enhancement of Simulatability, and Complexity, compared to \tokenex\ and \tokenintex. Future efforts should focus on developing combined methods that enhance all explanation properties.

\section*{Limitations}
Our work introduces a unified framework to evaluate input feature explanations across four key properties. We generated three types of explanations using three attribution methods on two Transformer models (BERT-base and BART-base) for two NLU tasks (NLI and fact-checking). Thereby, we can assess and compare the properties of each explanation type. Due to computational resource limitations, we did not include larger decoder-only models in our evaluation. Future research could \textit{explore other models} to provide additional insights. 

We note that our work considered the FEVER and SNLI datasets as they are the only available datasets with annotations of human interactive explanations, required for the Agreement with Human Annotations property. In future work, given the availability of other datasets, examining the properties of different explanations in various \textit{tasks beyond NLI and fact checking} would be valuable, especially for simpler tasks that consist of only one input part or more complex tasks that consist of more than two parts with possible relationships between them. Additionally, tasks with longer textual inputs, which are known to pose greater challenges for current explainability techniques~\citep{atanasova-etal-2020-diagnostic}, could also be analyzed. 

Furthermore, while we consider four widely used explanation properties in this automatic evaluation framework, future works should consider verifying, potentially with supplementary human studies, that the properties are well aligned with the downstream utility of the explanations in different application tasks~\citep{miller2019explanation}. We note that manual evaluation, while valuable, is time-consuming and costly. Automated evaluation, with our proposed framework, allows for quicker insights, helping prioritize methods that may benefit from human-centered evaluation. Additionally, the properties we evaluate demonstrate why human evaluation is not necessarily required at this stage. Faithfulness measures whether explanations reflect the model’s internal reasoning, a task humans cannot assess (see Faithfulness evaluation guidelines in ~\citet{jacovi-goldberg-2020-towards}). Explanations that fail this test should not be considered for further human evaluation as they can be harmful, e.g. by hiding a model's flaws and biases. Agreement with Human Annotation already captures alignment with human reasoning, ensuring explanations are plausible. Automated Simulatability correlates strongly with human studies, providing a reliable proxy for replicating the model’s behaviour without the need for expensive human experiments~\citep{pruthi-etal-2022-evaluating}. Finally, automated methods ensure consistency and objectivity, while human annotations can introduce subjectivity and variability. Studies have even shown conflicting results from human evaluations~\citep{poursabzi-sangdeh-etal-2021-interpretability,ribeiro-etal-2016-trust}. Automated evaluation provides an objective, reproducible baseline, which can later be supplemented by human evaluations where needed.

We have also employed three base representative explainability methods for each of the three types of input feature explanations. However, \textit{more existing and newly emerged base explainability methods} could be explored in future work. Additionally, our study focuses solely on post-hoc explainability techniques, while other supervised feature extraction methods could also be investigated~\citep{yu2021understanding,liu2024mmi}.  These methods typically treat human-annotated important fragments within the input as gold causal features, akin to our Agreement with Human Annotation measurement. The broader set of properties introduced in our framework could be leveraged to evaluate such explanations more comprehensively. Apart from that, future work could adapt and extend our framework to other forms of explanations such as free-text explanations produced by self-rationalization models ~\citep{liu2022fr,liu2023decoupled,liu2023mgr,NEURIPS2023_87e82678,liu2024enhancing}. All said potential future studies are well facilitated by the efficient automated evaluation proposed with our framework.

Our findings indicate that span interactive explanations (\spanintex) have a notable advantage over other explanation types in terms of Agreement with Human Annotation, Simulatability, and Complexity, suggesting they are easier for humans to understand. This insight could inspire future work to leverage \spanintex\ as the input feature explanation in HCI models. However, \spanintex\ shows low comprehensiveness in faithfulness evaluations. The Louvain algorithm, used for \spanintex\ generation, may limit its comprehensiveness despite using different attribution methods for \tokenintex. Future work should explore better methods for capturing span interactions and possibly combine \spanintex\ and \tokenex\ for higher faithfulness, as \tokenex\ demonstrates a stable advantage in comprehensiveness.

Another core finding is that the attribution method significantly affects most diagnostic properties of all explanation types, such as sufficiency. No single attribution method consistently excels across all properties, highlighting the need for continuous evaluation and improvement in attribution methods, particularly for \spanintex.

To ensure a fair comparison, our unified evaluation framework currently considers only the token count differences among various input feature explanations, with interactive explanations flattened. Future work could involve a human-in-the-loop approach to account for the effects of interactive explanations beyond just token count differences. For example, a display system could visually present highlighted tokens and interactions to gauge human preferences. Our work provides a starting point for comparing input feature explanations from an automated evaluation perspective, and future research could explore additional factors, such as psychological elements and visual aspects, from a human perspective, which would benefit more non-expert users.

Another limitation of this work is that we focus solely on the automatic evaluation of input feature explanations without examining the potential biases they may exhibit. For instance, these explanations might favor certain words or phrases from the input sequence, for example, sometimes emphasizing some prepositions that might bear less meaning, raising questions about whether such biases stem from the models themselves or from the explainability techniques used. Also, it is worth checking, especially in sensitive domains such as healthcare and law, how contradictory explanations for different model decisions differed, which might sway the decision-making. 

\section*{Acknowledgements}
$\begin{array}{l}\includegraphics[width=1cm]{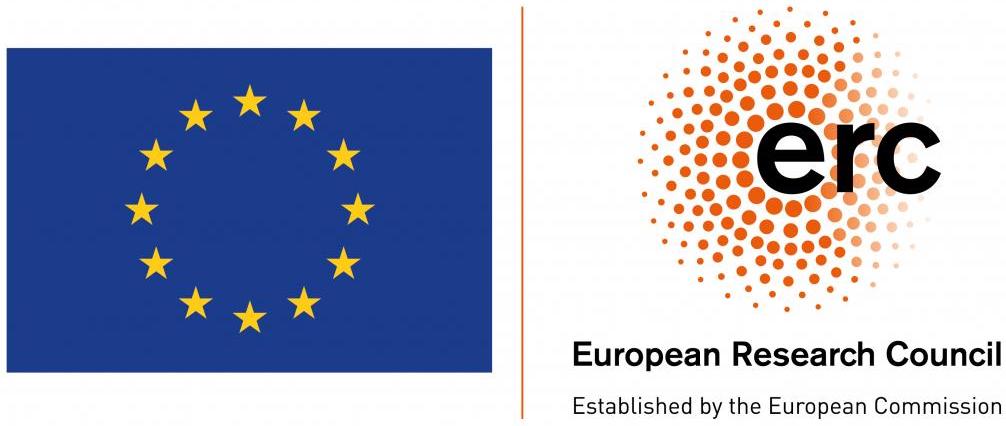} \end{array}$ 
This research was co-funded by the European Union (ERC, ExplainYourself, 101077481), by the Pioneer Centre for AI, DNRF grant number P1, as well as by The Villum Synergy Programme (grant number 40543). Views and opinions expressed are however those of the author(s) only and do not necessarily reflect those of the European Union or the European Research Council. Neither the European Union nor the granting authority can be held responsible for them. 

\FloatBarrier
\clearpage
\section{Appendix}

\subsection{Unified Faithfulness Evaluation: Explanation Masking Details}
\label{sec:app:faithfulness-mask-details}
As discussed in Section \S\ref{sec:higlight_faithfulness:method}, we introduce the dynamic threshold $\theta_{x,k}$ to ensure an identical number of tokens from different types of top input feature explanations for the same instance in Unified Faithfulness evaluation.

For the Unified Comprehensiveness evaluation, we conduct a similar process for all three explanation types separately. Using \tokenex\ as an example, we first calculate the maximum number of top \tokenex\ for disturbing each instance $x$ as $k_{\tokenex}(x)$ following Eq. \ref{eq:threshold_2}. To omit the \tokenex\ tokens from the original input, we replace them with [MASK] tokens, while keeping the rest unchanged. The specific [MASK] token used depends on the model architecture. We then gradually increase the number of \tokenex\ tokens masked out until it reaches  $k_{\tokenex}(x)$ and record the corresponding changes in model predictions. The average prediction change across the dynamic threshold and all instances gives the Unified Comprehensiveness score for \tokenex. For \tokenintex\ and \spanintex, the only difference is that we mask out token pairs for \tokenintex\ and span pairs for \spanintex, with the maximum explanations masked out as $k_{\tokenintex}(x)$ and  $k_{\spanintex}(x)$ for each instance $x$, calculated using Eq. \ref{eq:threshold_1} and Eq. \ref{eq:threshold_3} respectively.

For the Unified Sufficiency evaluation, we conduct experiments for the three explanation types separately. Unlike Unified Comprehensiveness, we retain only the tokens/token pairs/span pairs for the input while masking out all other tokens with [MASK] tokens for each instance, depending on the model architecture used. We first calculate the maximum number of top explanations involved in disturbance for each explanation type for instance $x$ using Eq. \ref{eq:threshold_1}, Eq. \ref{eq:threshold_2}, and Eq. \ref{eq:threshold_3}. Then, we keep the token/token pairs/span pairs in the model input by masking out all other tokens, starting with one explanation and adding one more explanation for each subsequent disturbance until the total number of explanations reaches $k_{\tokenex}(x)$, $k_{\tokenintex}(x)$, or $k_{\spanintex}(x)$. Meanwhile, we record the model predictions for each disturbance. The Unified Sufficiency score for each explanation type is then calculated by averaging the prediction changes across the dynamic threshold for that explanation type, considering all instances.

\subsection{Detailed Explanation Insertion Method}
\label{sec:app:detailed-form}
To enable a fair comparison among different input feature explanations in terms of simulatability (\S\ref{sec:simulatability:method}), we applied consistent insertion formats to combine the explanations with the original input for training the agent models. This design aims to minimize noise from insertion format differences. We tested two ways, each applicable to all types of input feature explanations, to construct input sequences with inserted explanations of type $E$. These input sequences are denoted $x_E$ in \S\ref{sec:simulatability:method}, omitting specific insertion format details for brevity.

For Symbol-Insertion \(I_{Symbol}\), we preserve the original input sequence but insert special symbols \(<\) and \(>\) to quote the tokens (for \tokenex\ and \tokenintex) or spans (for \spanintex) within the input. Additionally, for \tokenex, we append a ranking mark after each quoted token based on their attribution scores, ranked in descending order. For \tokenex\ and \spanintex, each quoted token/span is also assigned a ranking mark indicating the rank of their respective interactions by attribution score, ensuring tokens/spans from the same interaction share the same mark. This method allows us to generate input sequences combined with different input feature explanations in a consistent symbol insertion format.

For Text-Insertion \(I_{Text}\), we append tokens, token tuples, or span tuples to the end of the original input sequence for each explanation type. They are added in the order ranked by descending attribution score. Specifically, for \tokenex, tokens from different \tokenex\ explanations are separated by semicolons. For \tokenintex\ and \spanintex, tokens/spans within each interaction are connected by a comma, and different interactions are separated by semicolons. This approach constructs input sequences combined with each type of input feature explanation in a consistent text insertion format.

\subsection{Agreement Dataset Details}
\label{sec:app:dataset-details}

To assess how different types of input feature explanations overlap with human annotations, we collected golden explanations of various types from e-SNLI and SpanEx for instances within SNLI and FEVER, respectively. Detailed information about the annotated explanation types and the number of instances with labeled explanations for these datasets is shown in Table \ref{exper:dataset-details}. For the SNLI dataset, e-SNLI provides \tokenex\ explanations, while \textit{SpanEx-SNLI} includes \spanintex\ explanations and token-level interactions (\tokenintex\ explanations). We selected 3,865 overlapping instances and evaluated the human agreement score for different types of input feature explanations. For FEVER, \textit{SpanEx-FEVER} includes \spanintex\ and token-level interactions (\tokenintex\ explanations). Since no \tokenex\ explanations are provided, we extracted tokens from the golden \tokenintex\ explanations in \textit{SpanEx-FEVER} as an approximation. These selected instances are also used when evaluating other properties of input feature explanations.

\begin{table}
\centering
\small
\begin{tabular}{lcccc}
\toprule
\textbf{$D$} & $E$ & \multicolumn{1}{l}{\textbf{Size}}                                           \\  \midrule
SNLI        & -   & \begin{tabular}[c]{@{}c@{}}549367 Train\\ 9842 Dev\\ 9824 Test\end{tabular} \\
e-SNLI                & \tokenex\                                                         & \begin{tabular}[c]{@{}c@{}}549367 Train\\ 9842 Dev\\ 9824 Test\end{tabular} \\
\textit{SpanEx-SNLI}  & \begin{tabular}[c]{@{}c@{}}\spanintex\ \\ \tokenintex \end{tabular}  & 3865 Test                                                                   \\  \midrule
FEVER        & -   & \begin{tabular}[c]{@{}c@{}}145449 Train\\ 9999 Dev\\ 9999 Test\end{tabular} \\
\textit{SpanEx-FEVER} & \begin{tabular}[c]{@{}c@{}}\spanintex \\ \tokenintex \end{tabular} & 3206 Test                                                                   \\ \bottomrule
\end{tabular}%
\caption{Overview of datasets SNLI~\citep{bowman-etal-2015-large}, FEVER~\citep{thorne-etal-2018-fever}, \textit{SpanEx}~\citep{choudhury-etal-2023-explaining} and e-SNLI~\citep{camburu2018snli}. \textit{SpanEx} contains instances from SNLI and FEVER, annotated with \spanintex\ explanations including token-level explanations (\tokenintex\ explanations). e-SNLI contains instances from SNLI dataset, annotated with \tokenex\ explanations.}
\label{exper:dataset-details}
\end{table}

\subsection{Base Model Performance.}
\label{sec:app:base-model}
As shown in Table \ref{exper:base-model-performance}, we report the performance of fine-tuned BERT-base and BART-base models on SNLI and FEVER, respectively. These models, fine-tuned for their specific tasks, are used to generate various input feature explanations through different explainability techniques. Importantly, these are the original models that the agent models, as described in \S\ref{sec:simulatability:method}, learn to simulate.

\begin{table}
\centering
\small
\begin{tabular}{lcccc}
\toprule

\multirow{2}{*}{\textbf{$Model$}} & \multicolumn{2}{c}{\textbf{F1 score}}              \\
                                  & \multicolumn{1}{l}{Dev} & \multicolumn{1}{l}{Test} \\ \midrule
BERT-SNLI                         & 87.21                   & 88.43           \\
BART-SNLI                         & 86.81                   & 85.40                    \\ \midrule
BERT-FEVER                        & 86.21          & 89.49           \\
BART-FEVER                        & 85.19                   & 84.88                    \\ \bottomrule
\end{tabular}%

\caption{The performance of our BERT-base and BART-base models fine-tuned on SNLI and FEVER, respectively, regarding F1 score(\%).}
\label{exper:base-model-performance}
\end{table}

\subsection{Explainability Techniques}
\label{sec:app:explainability_techniques}

In this section, we detail the explainability techniques employed to generate various types of input feature explanations. As outlined in \S\ref{sec:PosthocExplainabilityTechniques}, we categorize these techniques based on the method used for generating \tokenex, while \tokenintex\ explanations stem from their bivariate variants, forming the basis for \spanintex\ explanations.

As denoted in Section \S\ref{preliminaries}, $x_i$ represents the $i$th token with instance $x$. To better illustrate the explainability techniques below, we use $F$ as the set of all tokens within this instance and $S$ as the subset of $F$. All explanations are obtained using model $M$, which is omitted in the following notions for brevity. We use $A_\tokenex(x_i)$ to denote the attribution score generated by explainability technique $A$ for the $i$th token $x_i$, $A_\tokenintex(x_i,x_j)$ as the attribution score for token interaction $(x_i,x_j)$,  $A_\tokenintex(x_i \mid x_j)$ as the importance score of token $x_i$ conditioned on $x_j$ is present when the directed importance between tokens within $(x_i,x_j)$ is considered in some attribution techniques,
$A_\spanintex(span_i^0,span_i^1)$ as the attribution score for corresponding span interaction, where $span_i^0=(x_s,...,x_{s+l_1})$ is a span from part1 and $span_i^1=(x_t,...,x_{t+l_2})$ is a span from part2 of the input. Note that in Section \S\ref{preliminaries}, we use $a^x_k$ to denote the importance score of the $k$th most important explanation of instance $x$; here, we only focus on the attribution scores of explanations without ranking them. 

\textbf{Shapley}. For \tokenex,  we employ the SHAP method to assign importance scores to each token within the input by removing each token separately and computing its removal effect on the model prediction with different subsets of other tokens presented to the model, following \citet{lundberg2017unified}. 
\begin{equation}
\small
\begin{split}
Shap_\tokenex(x_i) &= \sum_{S \subseteq F \setminus \{x_i\}} 
\frac{|S|!(|F| - |S| - 1)!}{|F|!} \\
&\quad [f(S \cup \{x_i\}) - f(S)]
\end{split}
\label{eq:univariate_shapley}
\end{equation}
$Shap_\tokenex(x_i)$ denotes the importance score of token $x_i$.
As the calculation of $Shap_\tokenex(x_i)$ is computationally expensive, we utilize Kernel SHAP to approximate these Shapley values. 

For \tokenintex, we first apply Bivariate Shapley ~\citep{masoomi2023explanations} to assess the mutual importance scores between two tokens, which are from different parts of the input, within a token interaction, and then average these two mutual importance scores as importance score of this token interaction. Specifically, to compute the importance score of a token $ x_i$ conditioned on the presence of token $ x_j$, the sets of tokens $S$ considered are limited to those containing token $x_j$, while the impact of other sets of tokens influences the importance of $x_i$ is ignored in this case. Thus, $Shap_\tokenintex(x_i \mid x_j)$ can be calculated by:


\begin{equation}
\scriptsize
\begin{split}
Shap_\tokenintex(x_i \mid x_j) &= \sum_{x_j \in S \subseteq F \setminus \{x_i\}} 
\frac{|S|!(|F| - |S| - 1)!}{|F|!} \\
&\quad [f(S \cup \{x_i\}) - f(S)]
\end{split}
\label{eq:bivariate_shapley_directed}
\end{equation}
The importance score $Shap_\tokenintex(x_i,x_j)$ for token interaction $(x_i,x_j)$ are calculated by averaging $Shap_\tokenintex(x_i\mid x_j)$ and $Shap_\tokenintex(x_j\mid x_i)$. We also use Kernel Shapley to approximate the calculation of Bivariate Shapley value.

For \spanintex, we first apply the Louvain Community Detection algorithm ~\citep{blondel2008fast} to extract span interactions and then average the importance scores of token interactions comprised in each span interaction as its importance score, following \citet{choudhury-etal-2023-explaining}.

To extract span interactions, we first construct a directed bipartite graph for instance $x$, by taking each token $x_i$ from the input as node $i$ and the mutual importance scores between each two tokens from different parts obtained above as the weights of directed edges connecting them. Louvain Community Detection algorithm is then applied to search for communities of nodes with dense intra-cluster and sparse inter-cluster relationships. With each community of nodes(tokens) $S_p$ obtained, we can get one span interaction $(span_p^0,span_p^1)$, where the two spans consist of neighboring tokens from the part1 subset of this community $S_p^0$, and the part2 subset of it $S_p^1$ respectively. 

Then we calculate the importance score of this span interaction by averaging the importance scores of all token interactions it comprises.
\begin{equation}
\scriptsize
\begin{aligned}
Shap_\spanintex(span_p^0,span_p^1) &= \\
&\quad \frac{\sum\limits_{i,j}^{\begin{array}{l}
    \substack{x_i \in S_p^0  \\ x_j \in S_p^1}
\end{array}} Shap_\tokenintex(x_i, x_j)}{|S_p^0 \cup S_p^1|}
\end{aligned}
\label{eq:span_extraction_method}
\end{equation}

Note that in the following, no matter which explainability techniques to assign importance score to \tokenintex, we apply the same method as stated above to extract span interactions and compute their importance scores, $A_\spanintex(span_p^0,span_p^1)$, based on corresponding token interaction importance score, $A_\tokenintex(x_i,x_j)$.

\textbf{Attention}. For each token within the input sequence, we use the self-attention weights between this token and the first token as an indicator of its importance score \citep{jain-wallace-2019-attention}. We follow \citet{choudhury-etal-2023-explaining} to select the most important attention head in the last layer of the model to obtain these attention weights. For each possible token interaction, we use the method by \citet{clark-etal-2019-bert} to extract and average the attention weights between token pairs from different parts of the input to derive their importance scores, also from the most important head of the last layer. To obtain span interactions and assign them importance scores, we apply the same method to these token interaction scores as described above.

\textbf{Integrated Gradients}. To calculate the importance score for each token in the input sequence, we integrate the gradients of the model's output with respect to each token embedding, following \citet{sundararajan2017axiomatic}. For generating the importance scores of token interactions, we use Layer-wise Attention Attribution \citep{ye-etal-2021-connecting}, which attributes attention links between pairs of tokens within attention maps with a mechanism similar to Integrated Gradients. These attribution maps are created for each model layer and then aggregated across layers to form a final attribution map. The importance score for each token interaction is calculated as the average value from this final attribution map between the involved tokens. For span interactions, we generate and assign importance scores using the same approach based on the importance scores of the token interactions.

\subsection{Detailed Experiment Results}
\label{sec:app:detailed}
\subsubsection{Faithfulness}

\begin{figure*}[t]
\centering
\begin{subfigure}[t]{0.49\textwidth}
    \centering
    \includegraphics[width=\linewidth]{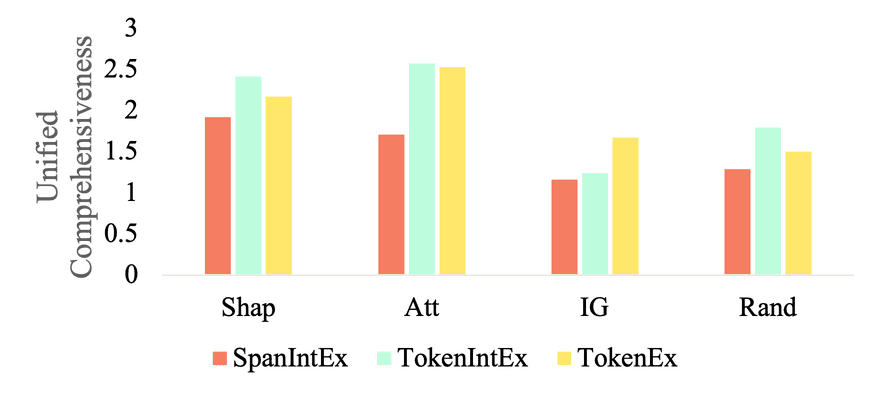}
    \caption{Comprehensiveness on SNLI dataset, the higher the better.}
    \label{fig:subfig-comp-snli-bert:a}
\end{subfigure}
\hfill
\begin{subfigure}[t]{0.49\textwidth}
    \centering
    \includegraphics[width=\linewidth]{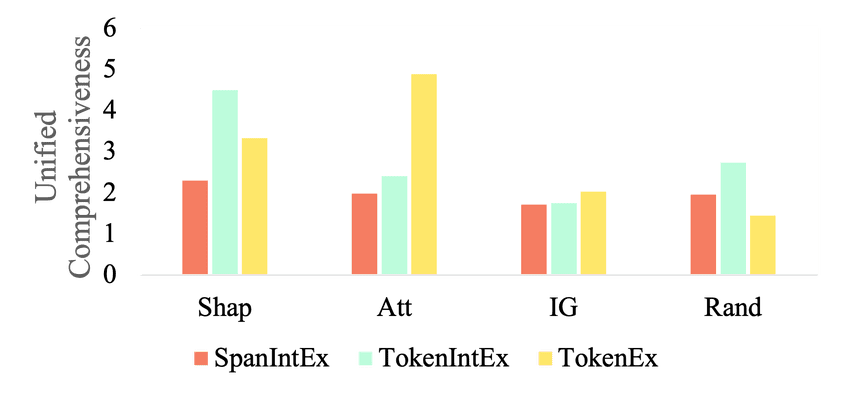}
    \caption{Comprehensiveness on FEVER dataset, the higher the better.}
    \label{fig:subfig-comp-fever-bert:b}
\end{subfigure}

\medskip

\begin{subfigure}[t]{0.49\textwidth}
    \centering
    \includegraphics[width=\linewidth]{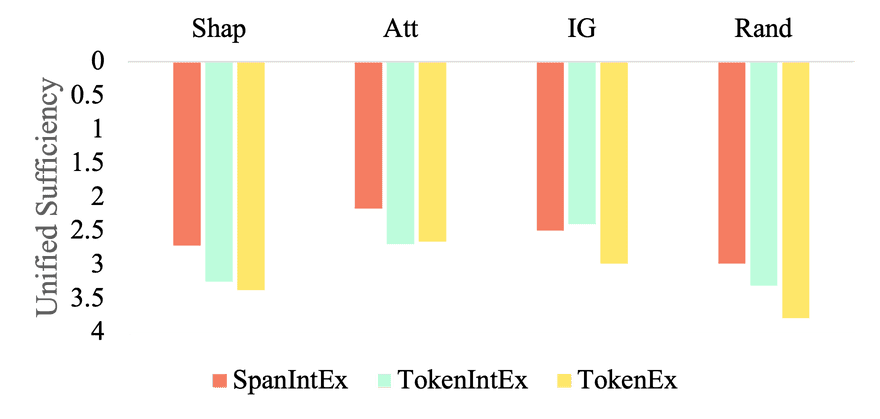}
    \caption{Sufficiency on SNLI dataset, the lower the better.}
    \label{fig:subfig-suff-snli-bert:c}
\end{subfigure}
\hfill
\begin{subfigure}[t]{0.49\textwidth}
    \centering
    \includegraphics[width=\linewidth]{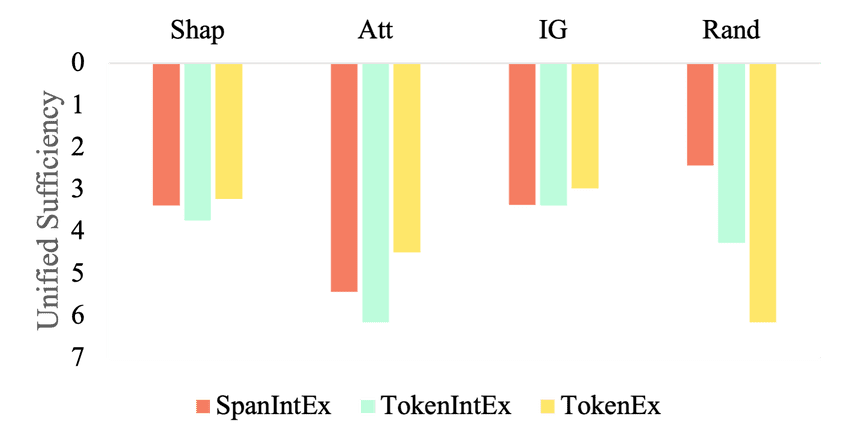}
    \caption{Sufficiency on FEVER dataset, the lower the better.}
    \label{fig:subfig-suff-fever-bert:d}
\end{subfigure}
\caption{Unified Comprehensiveness and Sufficiency of three types of feature attribution explanations on SNLI and FEVER datasets using the BERT model. Subfigures (a) and (b) show Unified Comprehensiveness results, while (c) and (d) show Unified Sufficiency results. Explanations are generated by Shapley (Shap), Attention (Att), and Integrated Gradients (IG). Randomly selected span pairs, token pairs, and tokens are baselines corresponding to explanation type \spanintex, \tokenintex, and \tokenex\,  and form the group Random baseline (Rand). We set $k=3$ for top span interactions and adjust token counts as per \S\ref{sec:higlight_faithfulness:method}, also ensuring the random baseline matches the average token count of the top $k$ span interactions.}
\label{faithfulness-bert}
\end{figure*}

\begin{figure*}[t]
\centering
\begin{subfigure}[t]{0.49\textwidth}
    \centering
    \includegraphics[width=\linewidth]{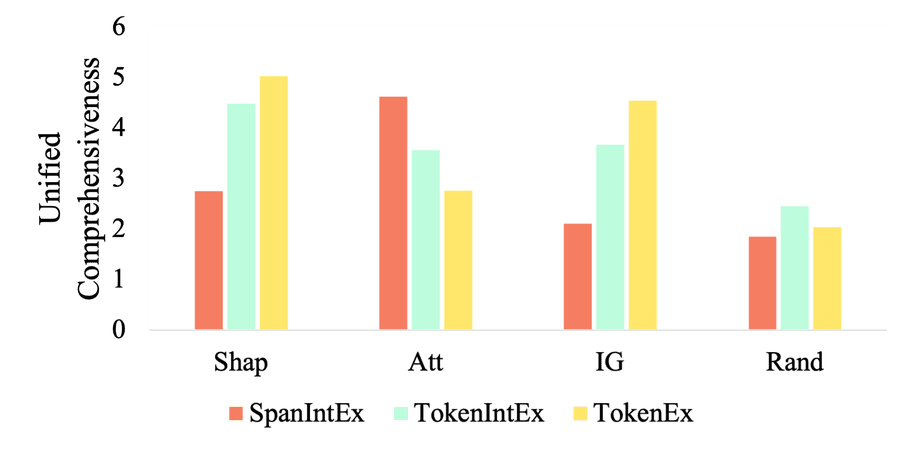}
    \caption{Comprehensiveness on SNLI dataset, the higher the better.}
    \label{fig:subfig-comp-snli-bart:a}
\end{subfigure}
\hfill
\begin{subfigure}[t]{0.49\textwidth}
    \centering
    \includegraphics[width=\linewidth]{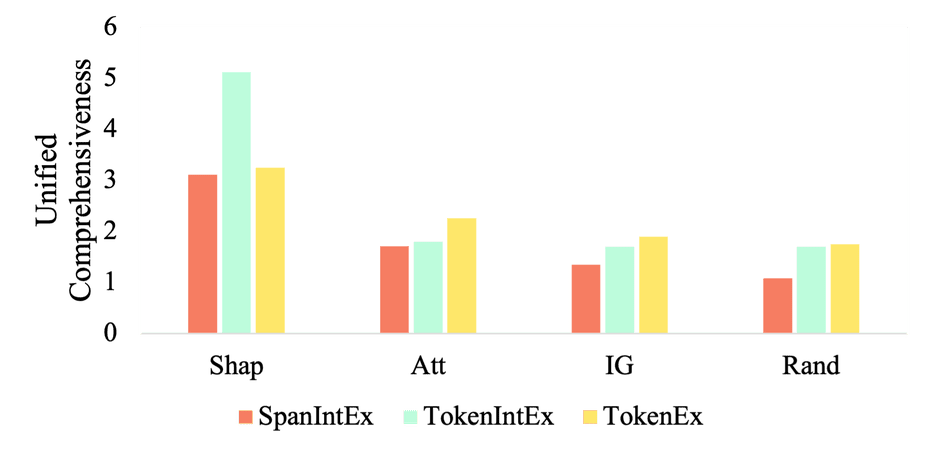}
    \caption{Comprehensiveness on FEVER dataset, the higher the better.}
    \label{fig:subfig-comp-fever-bart:b}
\end{subfigure}

\medskip

\begin{subfigure}[t]{0.49\textwidth}
    \centering
    \includegraphics[width=\linewidth]{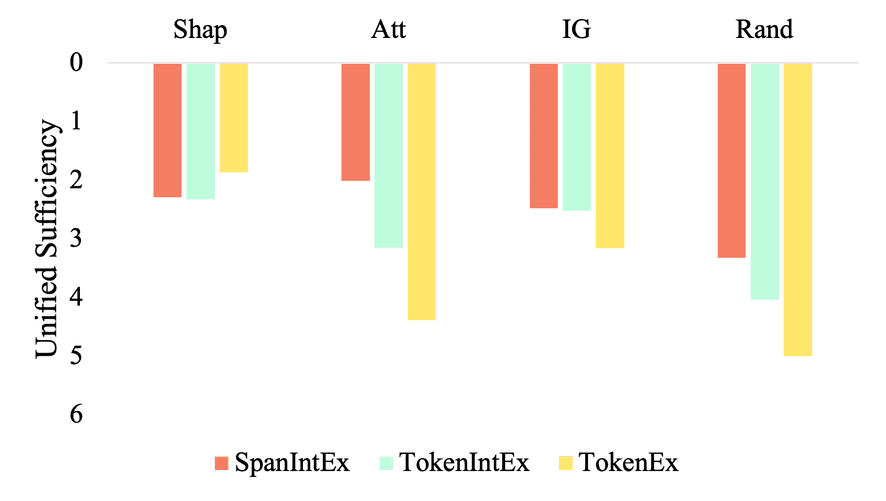}
    \caption{Sufficiency on SNLI dataset, the lower the better.}
    \label{fig:subfig-suff-snli-bart:c}
\end{subfigure}
\hfill
\begin{subfigure}[t]{0.49\textwidth}
    \centering
    \includegraphics[width=\linewidth]{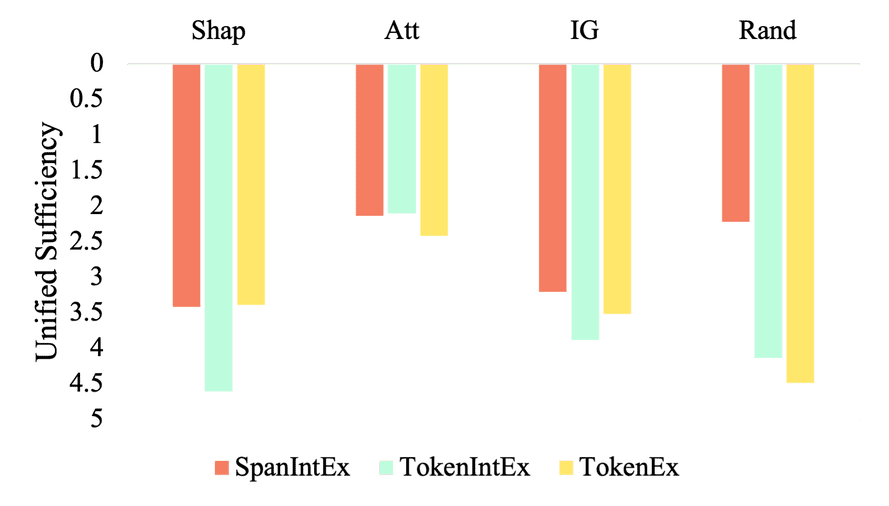}
    \caption{Sufficiency on FEVER dataset, the lower the better.}
    \label{fig:subfig-suff-fever-bart:d}
\end{subfigure}
\caption{Unified Comprehensiveness and Sufficiency of three types of feature attribution explanations on SNLI and FEVER datasets using the BART model. Subfigures (a) and (b) show Unified Comprehensiveness results, while (c) and (d) show Unified Sufficiency results. Explanations are generated by Shapley (Shap), Attention (Att), and Integrated Gradients (IG). Randomly selected span pairs, token pairs, and tokens are baselines corresponding to explanation type \spanintex\, \tokenintex\, and \tokenex\,  and form the group Random baseline (Rand). We set $k=3$ for top span interactions and adjust token counts as per section \S\ref{sec:higlight_faithfulness:method}, ensuring the random baseline matches the average token count of the top $k$ span interactions.}
\label{faithfulness-bart}
\end{figure*}

\subsubsection{Agreement with Human Annotation}

\begin{table}
\centering
\small
\begin{tabular}{lcccc}
\toprule
$\mathbf{E}$  & \textbf{Shap}  & \textbf{Att}   & \textbf{IG}    & \textbf{Rand}  \\ \midrule
              & \multicolumn{4}{c}{Interaction level agreement}                   \\
$\spanintex$  & \textbf{30.18} & \textbf{57.40} & \textbf{39.40} & \textbf{33.82} \\
$\tokenintex$ & 29.02          & 37.02          & 35.06          & 23.42          \\
$\tokenex$    & \textbf{-}     & \textbf{-}     & \textbf{-}     & \textbf{-}     \\
              & \multicolumn{4}{c}{Token level agreement}                         \\
$\spanintex$  & \textbf{75.63} & 78.26          & \textbf{76.52} & \textbf{76.96} \\
$\tokenintex$ & 74.89          & \textbf{79.60} & 73.19          & 74.96          \\
$\tokenex$    & 75.54          & 77.62          & 70.74          & 76.33          \\ \bottomrule
\end{tabular}%

\caption{Human Annotation Agreement Results (see \S\ref{sec:agreement:method}) on SNLI dataset when explanations are generated based on BERT. Interaction-level and Token-level agreement scores, Average Precision(\%), compared to human annotations for explanation types $\spanintex$, $\tokenintex$, $\tokenex$ generated by Shapley(\textbf{Shap}), Attention(\textbf{Att}), Integrated Gradients(\textbf{IG}) respectively. Using the same attribution method, the highest alignment score for each category is highlighted in bold. \textbf{Rand} indicates the random baseline as described in \S\ref{sec:agreement:method}.}
\label{exper:agreement-bert-snli-f}
\end{table}

\begin{table}
\centering
\small
\begin{tabular}{lcccc}
\toprule
$\mathbf{E}$  & \textbf{Shap}  & \textbf{Att}   & \textbf{IG}    & \textbf{Rand}  \\ \midrule
              & \multicolumn{4}{c}{Interaction level agreement}                   \\
$\spanintex$  & \textbf{19.92} & \textbf{28.12} & \textbf{27.45} & \textbf{19.33} \\
$\tokenintex$ & 3.96           & 10.27          & 21.30 & 10.23          \\
$\tokenex$    & \textbf{-}     & \textbf{-}     & \textbf{-}     & \textbf{-}     \\
              & \multicolumn{4}{c}{Token level agreement}                         \\
$\spanintex$  & \textbf{66.95} & \textbf{68.71} & \textbf{72.24} & \textbf{67.5}  \\
$\tokenintex$ & 66.90          & 67.29          & 70.50          & 65.86          \\
$\tokenex$    & 58.07          & 56.88          & 61.07          & 63.10          \\ \bottomrule
\end{tabular}%

\caption{Human Annotation Agreement Results (see \S\ref{sec:agreement:method}) on the FEVER dataset when explanations are generated based on BERT. The rest of the settings are the same as Table \ref{exper:agreement-bert-snli-f}.}
\label{exper:agreement-bert-fever}
\end{table}

There is a notable gap between interaction-level and token-level agreement scores. For example, in Table \ref{exper:agreement-bert-snli-f}, the highest interaction-level agreement score for \spanintex\ explanations is 57.40\%, while the highest token-level agreement score for \spanintex\ is 78.26\%. A similar pattern is observed for \tokenintex. This suggests that although \spanintex\ and \tokenintex\ explanations align more with human reasoning than \tokenex\ explanations, pairing important tokens or spans into interactions that are plausible to humans remains challenging.

\begin{table}
\centering
\small
\begin{tabular}{lcccc}
\toprule
$\mathbf{E}$  & \textbf{Shap} & \textbf{Att}   & \textbf{IG} & \textbf{Rand}                \\ \midrule
              & \multicolumn{4}{c}{Interaction level agreement}                             \\
$\spanintex$  & \textbf{37.36} & \textbf{47.33} & \textbf{34.18} & {\color[HTML]{343434} \textbf{28.25}} \\
$\tokenintex$ & 32.36         & 35.17          & 33.06       & 13.80                        \\
$\tokenex$    & \textbf{-}    & \textbf{-}     & \textbf{-}  & \textbf{-}                   \\
              & \multicolumn{4}{c}{Token level agreement}                                   \\
$\spanintex$  & 80.16         & \textbf{76.28} & 82.76       & 70.04 \\
$\tokenintex$ & \textbf{84.9}  & 75.92          & \textbf{86.06} & \textbf{75.32} \\
$\tokenex$    & 65.74         & 73.71          & 70.44       & 73.34 \\ \bottomrule
\end{tabular}%

\caption{Human Annotation Agreement Results (see \S\ref{sec:agreement:method}) on SNLI dataset when explanations are generated based on BART. The rest settings are the same as Table \ref{exper:agreement-bert-snli-f}.}
\label{exper:agreement-bart-snli}
\end{table}

\begin{table}
\centering
\small
\begin{tabular}{lcccc}
\toprule
$\mathbf{E}$  & \textbf{Shap} & \textbf{Att} & \textbf{IG}    & \textbf{Rand}                \\ \midrule
              & \multicolumn{4}{c}{Interaction level agreement}                              \\
$\spanintex$ & \textbf{22.86} & \textbf{20.66} & \textbf{18.72} & {\color[HTML]{343434} \textbf{16.76}} \\
$\tokenintex$ & 4.71          & 2.64         & 10.54          & 8.64                         \\
$\tokenex$    & \textbf{-}    & \textbf{-}   & \textbf{-}     & \textbf{-}                   \\
              & \multicolumn{4}{c}{Token level agreement}                                    \\
$\spanintex$ & \textbf{68.77} & \textbf{65.33} & 70.22          & \textbf{69.51} \\
$\tokenintex$ & 67.01         & 63.40        & \textbf{70.98} & 68.11 \\
$\tokenex$    & 59.43         & 52.15        & 57.56          &  60.93 \\ \bottomrule
\end{tabular}%

\caption{Human Annotation Agreement Results (see \S\ref{sec:agreement:method}) on FEVER dataset when explanations are generated based on BART. The rest settings are the same as Table \ref{exper:agreement-bert-snli-f}.}
\label{exper:agreement-bart-fever}
\end{table}

\subsubsection{Simulatability}
Regarding insertion formats, for BERT models, text insertion ($I_{Text}$), which adds explanation text to the end of the input sequence, consistently outperforms symbol insertion ($I_{Sym}$), where symbols are added to the original input sequence, as shown in Tables \ref{exp:simulatability-form1-bert} and \ref{exp:simulatability-form2-bert}. However, the opposite effect is observed for BART models, as shown in Tables \ref{exp:simulatability-form1-bart} and \ref{exp:simulatability-form2-bart}. This indicates that simulatability results are sensitive to the explanation insertion form, highlighting the need for consistency in insertion form when comparing different explanation types.

\begin{table}
\centering
\small
\resizebox{0.85\linewidth}{!}{%
\begin{tabular}{llrrrrrr}
\toprule
\multirow{2}{*}{$\mathbf{D}$} & \multirow{2}{*}{$\mathbf{E}$} & \multicolumn{2}{c}{\textbf{Shap}} & \multicolumn{2}{c}{\textbf{Att}} & \multicolumn{2}{c}{\textbf{IG}} \\
                       &               & $\mathbf{SF}$ & $\mathbf{RSF}$ & $\mathbf{SF}$ & $\mathbf{RSF}$ & $\mathbf{SF}$ & $\mathbf{RSF}$ \\ \midrule
\multirow{2}{*}{SNLI}        & $\spanintex$                  & \textbf{87.9}    & \textbf{3.2}   & \textbf{86.7}   & \textbf{2.0}   & \textbf{88.9}   & \textbf{4.2}  \\
                       & $\tokenintex$ & 86.6          & 1.9            & 85.3          & 0.6            & 85.8          & 1.1            \\
                       & $\tokenex$    & 87.4          &2.7   & 85.7          & 1.0            & 86.0          & 1.3            \\ \midrule
\multirow{2}{*}{FEVER} & $\spanintex$  & 83.9          & 3.8            & \textbf{85.3} & \textbf{5.2}   & \textbf{85.2} & \textbf{5.1}   \\
                       & $\tokenintex$ & 82.7          & 2.6            & 84.0          & \textbf{3.9}   & 84.4          & 4.3            \\
                       & $\tokenex$    & \textbf{84.0} & \textbf{3.9}   & 82.3          & 2.2            & 81.8          & 1.7            \\ \bottomrule
\end{tabular}%
}
\caption{Simulatability results on SNLI and FEVER with BERT as the model used for all explanations $E\in{\spanintex, \tokenintex, \tokenex}$ generation with attribution method \textbf{Shapley}, \textbf{Attention}, and \textbf{Integrated Gradients} respectively. Note that insertion form $I_{Sym}$ is adopted for combining the explanations and the original input sequence for agent model $AM_{E}$, as depicted in \S\ref{sec:simulatability:method}. The agent models used for baseline $AM_O$, trained without explanations, have simulation F1 scores, as denoted in \S\ref{sec:simulatability:method}, of 84.7\%  and 80.0\% on test set shared with other agent models $AM_E$, as denoted in \S\ref{sec:simulatability:method}. We set $k=1$ for top \spanintex\ and calculated the number of top \tokenintex\ and \tokenex\ accordingly as stated in \S\ref{sec:higlight_faithfulness:method}. The largest increases are highlighted in bold for each dataset with the identical attribution method.}
\label{exp:simulatability-form1-bert}
\end{table}

\begin{table}
\centering
\small
\resizebox{0.85\linewidth}{!}{%
\begin{tabular}{llrrrrrr}
\toprule
\multirow{2}{*}{$\mathbf{D}$} & \multirow{2}{*}{$\mathbf{E}$} & \multicolumn{2}{c}{\textbf{Shap}} & \multicolumn{2}{c}{\textbf{Att}} & \multicolumn{2}{c}{\textbf{IG}} \\
                       &               & $\mathbf{SF}$ & $\mathbf{RSF}$ & $\mathbf{SF}$ & $\mathbf{RSF}$ & $\mathbf{SF}$ & $\mathbf{RSF}$ \\ \midrule
\multirow{2}{*}{SNLI}        & $\spanintex$                  & \textbf{87.8}    & \textbf{3.1}   & 87.1        & 2.4       & 88.2   & 3.5  \\
                       & $\tokenintex$ & 86.5          & 1.8            & \textbf{87.8} & \textbf{3.1}            & 86.4          & 1.7            \\
                       & $\tokenex$    & 87.0          & 2.3   & 86.3          & 1.6            & \textbf{88.4}          & \textbf{3.7}            \\ \midrule
\multirow{2}{*}{FEVER} & $\spanintex$  & 85.7          & 5.6            & 85.1 & 5.0            & \textbf{86.0} & \textbf{5.9}   \\
                       & $\tokenintex$ & 81.9          & 1.8            & \textbf{85.6} & 5.5            & 84.3          & 4.2            \\
                       & $\tokenex$    & \textbf{85.8} & \textbf{5.7}   & 84.5          & 4.4            & 82.0          & 1.9            \\ \bottomrule
\end{tabular}%

}
\caption{Simulatability results on SNLI and FEVER with BERT as the model used for all input feature explanation generation. Note that insertion form $I_{Text}$ is adopted for combining the explanations and the original input sequence for agent model $AM_{E}$, as depicted in \S\ref{sec:simulatability:method}.  The agent models used for baseline $AM_O$, which are trained without explanations, have simulation F1 scores of 84.7\% and 80.0\% on the test sets shared with agent model $AM_E$. The other setting is the same as Table \ref{exp:simulatability-form1-bert}}
\label{exp:simulatability-form2-bert}
\end{table}

\begin{table}
\centering
\small
\resizebox{0.85\linewidth}{!}{%
\begin{tabular}{llrrrrrr}
\toprule
\multirow{2}{*}{$\mathbf{D}$} & \multirow{2}{*}{$\mathbf{E}$} & \multicolumn{2}{c}{\textbf{Shap}} & \multicolumn{2}{c}{\textbf{Att}} & \multicolumn{2}{c}{\textbf{IG}} \\
                       &               & $\mathbf{SF}$ & $\mathbf{RSF}$ & $\mathbf{SF}$ & $\mathbf{RSF}$ & $\mathbf{SF}$ & $\mathbf{RSF}$ \\ \midrule
\multirow{2}{*}{SNLI}        & $\spanintex$                  & 87.8        & 7.9        & \textbf{87.3}   & \textbf{7.4}   & \textbf{86.5}   & \textbf{6.6}  \\
                       & $\tokenintex$ & 83.5          & 3.6            & 84.2          & 4.3            & 84.6          & 4.7            \\
                       & $\tokenex$    & \textbf{88.2} & \textbf{8.3}   & 81.4          & 1.5            & 85.8          & 5.9            \\ \midrule
\multirow{2}{*}{FEVER} & $\spanintex$  & \textbf{80.6} & 7.2            & \textbf{76.1}          & \textbf{2.7}            & \textbf{75.2} & \textbf{1.8}   \\
                       & $\tokenintex$ & 78.9          & 5.5            & 75.9 & 2.5            & 74.7          & 1.3            \\
                       & $\tokenex$    & 80.1          & \textbf{6.7}   & 75.0          & 1.6            & 74.1          & 0.7            \\ \bottomrule
\end{tabular}%

}
\caption{Simulatability results on SNLI and FEVER with BART as the model used for all input feature explanation generation. Note that insertion form $I_{Sym}$ is adopted for combining the explanations and the original input sequence for agent model $AM_{E}$, as depicted in \S\ref{sec:simulatability:method}. The base agent models $AM_O$ trained without explanations have the simulation f1 scores of 79.9\% and 73.4\%, respectively on the test sets sharing with other agent models $AM_E$. The other setting is the same as Table \ref{exp:simulatability-form1-bert}}
\label{exp:simulatability-form1-bart}
\end{table}

\begin{table}
\centering
\small
\resizebox{0.85\linewidth}{!}{%
\begin{tabular}{llrrrrrr}
\toprule
\multirow{2}{*}{$\mathbf{D}$} & \multirow{2}{*}{$\mathbf{E}$} & \multicolumn{2}{c}{\textbf{Shap}} & \multicolumn{2}{c}{\textbf{Att}} & \multicolumn{2}{c}{\textbf{IG}} \\
                       &               & $\mathbf{SF}$ & $\mathbf{RSF}$ & $\mathbf{SF}$ & $\mathbf{RSF}$ & $\mathbf{SF}$ & $\mathbf{RSF}$ \\ \midrule
\multirow{2}{*}{SNLI}        & $\spanintex$                  & \textbf{86.8}    & \textbf{6.9}   & \textbf{85.0}   & \textbf{5.1}   & 84.3       & \textbf{4.4}       \\
                       & $\tokenintex$ & 81.2          & 1.3            & 82.6          & 2.7            & \textbf{81.6} & 1.7            \\
                       & $\tokenex$    & 83.3          & 3.4            & 83.8          & 3.9            & 82.2          & 2.3            \\ \midrule
\multirow{2}{*}{FEVER} & $\spanintex$  & \textbf{78.2}          & \textbf{4.8}            & \textbf{75.3}          & \textbf{1.9}            & \textbf{74.6} & \textbf{1.2}   \\
                       & $\tokenintex$ & 74.8          & 1.4            & 74.1          & 0.7            & 73.9          & 0.5            \\
                       & $\tokenex$    & 77.6 & 4.2   & 74.9          & 1.5            & 73.6          & 0.3            \\ \bottomrule
\end{tabular}%

}
\caption{Simulatability results on SNLI and FEVER with BART as the model used for all input feature explanation generation respectively. Note that insertion form $I_{Text}$ is adopted for combining the explanations and the original input sequence for agent model $AM_{E}$, as depicted in \S\ref{sec:simulatability:method}. The base agent models $AM_O$ trained without explanations have the simulation f1 scores of 79.9\% and 73.4\%, respectively. The other setting is the same as Table \ref{exp:simulatability-form1-bert}}
\label{exp:simulatability-form2-bart}
\end{table}

\subsubsection{Complexity}

\begin{table}
\centering
\small
\resizebox{0.85\linewidth}{!}{%
\begin{tabular}{llccccc}
\toprule
\textbf{Dataset} & \textbf{$E$} & \textbf{Shapley} & \textbf{Attention} & \textbf{IG} & \textbf{R} & \textbf{U} \\ \hline
\multirow{2}{*}{SNLI}  & $\spanintex$  & 2.05          & \textbf{1.11} & 2.10          & 2.19 & 2.62 \\
                       & $\tokenintex$ & \textbf{1.72} & 1.43          & 2.30          & -    & -    \\
                       & $\tokenex$    & 2.08          & 1.64          & \textbf{1.91} & -    & -    \\ \midrule
\multirow{2}{*}{FEVER} & $\spanintex$  & 2.78          & \textbf{2.22} & 2.90          & 2.98 & 3.18 \\
                       & $\tokenintex$ & 3.12          & 3.07          & 3.15          & -    & -    \\
                       & $\tokenex$    & \textbf{2.76} & \textbf{2.22} & \textbf{2.57} & -    & -    \\ \bottomrule
\end{tabular}%
}
\caption{Complexity results on SNLI and FEVER datasets for three types of explanations generated by different attribution methods based on BERT model. The \textbf{Random} baseline represents the complexity score obtained by randomly generated scores in the range [0,1], ensuring the same number of scores as the number of explanations used. The \textbf{Upperbound} is calculated by setting all the attribution scores to the same value while ensuring the same number of scores as the number of explanations used. The lowest complexity score for each specific explanation type compared is highlighted in bold when the explanations are generated by each attribution method.}
\label{exp:complexity-bert}
\end{table}

\begin{table}
\centering
\small
\resizebox{0.85\linewidth}{!}{%
\begin{tabular}{llccccc}
\toprule
\textbf{Dataset}       & \textbf{$E$} & \textbf{Shapley} & \textbf{Attention} & \textbf{IG}   & \textbf{R} & \textbf{U} \\ \midrule
\multirow{2}{*}{SNLI}  & $\spanintex$ & 1.90             & \textbf{1.93}      & \textbf{1.63} & 2.36       & 2.76       \\
 & $\tokenintex$ & 2.50          & 2.53 & 2.13 & - & - \\
 & $\tokenex$    & \textbf{1.86} & 1.95 & 1.94 & - & - \\ \midrule
\multirow{2}{*}{FEVER} & $\spanintex$ & \textbf{2.73}    & \textbf{3.03}      & \textbf{2.38} & 3.13       & 3.38       \\
 & $\tokenintex$ & 3.30          & 3.36 & 2.93 & - & - \\
 & $\tokenex$    & 2.82          & 3.07 & 3.19 & - & - \\ \bottomrule
\end{tabular}%
}
\caption{Complexity results on SNLI and FEVER datasets for three types of explanations generated by different attribution methods based on BART model. The other settings are the same as Table \ref{exp:complexity-bert}.}
\label{exp:complexity-bart}
\end{table}

\chapter{Evaluation Framework for Highlight Explanations of Context Utilisation in Language Models}
\chaptermark{Evaluating Highlight Explanations for Indicating Context Utilisation}
\label{chap:highlight_context_utilisation}

\providecommand{\bestcomp}[1]{\underline{#1}}
\providecommand{\bestsuff}[1]{\textbf{#1}}
\providecommand{\subj}[1]{\textcolor{blue}{#1}}
\providecommand{\gold}[1]{\textcolor{orange}{#1}}
\providecommand{\noise}[1]{\textcolor{green}{#1}}
\providecommand{\rankk}{$Rank@k$}
\providecommand{\propk}{$Prop@k$}
\providecommand{\scorek}{$Score@k$}
\providecommand{\Drankk}{$\Delta Rank@k$}
\providecommand{\Drank}{$\Delta Rank$}
\providecommand{\mrr}{$MRR$}
\providecommand{\mrre}{MRR}
\providecommand{\rankke}{Rank@k}
\providecommand{\Drankke}{\Delta Rank@k}
\providecommand{\rankkeInst}{\rankke^{\mathrm{inst}}}
\providecommand{\rankkInst}{$\rankke^{\mathrm{inst}}$}
\providecommand{\DrankkeGroup}{\Drankke^{\mathrm{grp}}}
\providecommand{\DrankkeInst}{\Drankke^{\mathrm{inst}}}
\providecommand{\DrankkGroup}{$\Drankke^{\mathrm{grp}}$}
\providecommand{\DrankkInst}{$\Drankke^{\mathrm{inst}}$}
\providecommand{\NMutInfAt}[1]{\ensuremath{\mathrm{NMutInf@}#1}\xspace}
\providecommand{\MDLBitsAt}[1]{\ensuremath{\mathrm{MDL\text{-}Bits@}#1}\xspace}
\providecommand{\RR}{\ensuremath{\mathrm{RR}}\xspace}
\providecommand{\MRR}{\ensuremath{\mathrm{MRR}}\xspace}
\providecommand{\NMutInfk}{\NMutInfAt{k}}
\providecommand{\MDLBitsk}{\MDLBitsAt{k}}
\providecommand{\MI}{MechLight}
\providecommand{\FA}{FA}
\providecommand{\IG}{IG}
\providecommand{\Attn}{ATTN}

This chapter consists of the following published paper\footnote{No changes have been made to the scientific content of the originally published paper. The paper is presented in a single-column
format used throughout the thesis, and its references have been merged into the bibliography at the end of the thesis.}:

\begin{enumerate}
    \item \bibentry{sun2026evaluation}.
\end{enumerate}

\clearpage

\section*{Abstract}

Context utilisation, the ability of Language Models (LMs) to incorporate relevant information from the provided context when generating responses, remains largely opaque to users, who cannot determine whether models draw from parametric memory or provided context, nor identify which specific context pieces inform the response. Highlight explanations (HEs) offer a natural solution as they can point to the exact context pieces and tokens that influenced model outputs. However, no existing work evaluates their effectiveness in accurately explaining context utilisation. 
We address this gap by introducing the first gold standard HE evaluation framework for context attribution, using controlled test cases with known ground-truth context usage, thereby avoiding the limitations of existing indirect proxy evaluations. 
To demonstrate the framework's broad applicability, we evaluate four HE methods -- three established techniques and MechLight, a mechanistic interpretability approach we adapt for this task -- across four context scenarios, four datasets, and five LMs.
Overall, we find that MechLight performs best across all context scenarios. However, our findings reveal systematic failures in all methods: explanation accuracy degrades significantly with context length, and all methods exhibit strong positional biases in multi-document settings. Surprisingly, widely used gradient-based methods provide little value for understanding context usage.
These results challenge HEs' utility in retrieval-augmented generation, factual verification, and other applications. Our framework provides the foundation for developing accurate context attribution methods. We release our code and dataset publicly.\footnote{
Code: \url{https://github.com/lianyiyi/Transparent-Context-Usage}; \newline
Dataset: \url{https://huggingface.co/datasets/copenlu/transparent-context-usage}.
}


\section{Introduction}

\begin{figure}[t] 
  \centering
                \includegraphics[width=\columnwidth]{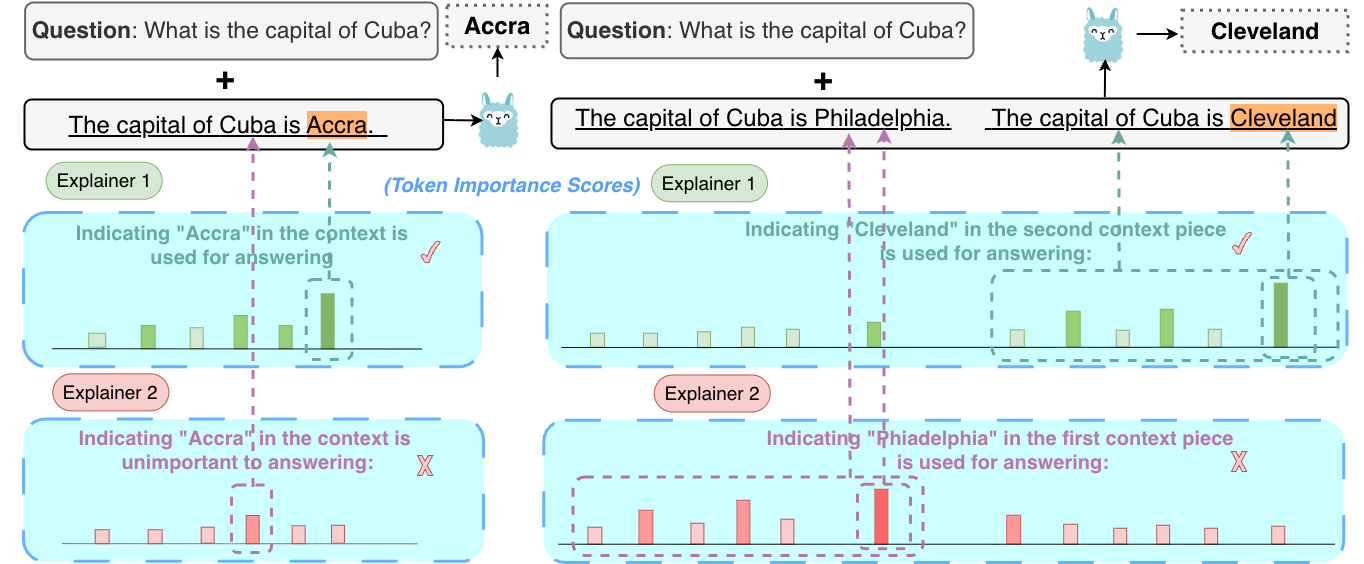}
            \caption{Examples of explanations with high and low utility in indicating the model’s context utilisation behaviour. Left: With one context conflicting with the model’s parametric knowledge (``Havana''), Explainer 1 correctly indicates ``Accra'' in the context is used for the answer, while Explainer 2 fails. Right: With two such contexts, Explainer 1 correctly indicates ``Cleveland'' in the second context is used for the answer, whereas Explainer 2 incorrectly attributes the answer to ``Philadelphia'' in the first context. Explainer~1 shows better utility than Explainer~2 in both cases.
            }
  \label{hcu:fig:main}
\end{figure}

Language models (LMs) are increasingly deployed in applications requiring 
integration of provided context with parametric knowledge -- from retrieval-augmented 
generation and question answering to document analysis and fact-checking.
However, a fundamental transparency gap remains: users cannot determine whether model outputs draw from the provided context knowledge (CK) or internal parametric knowledge (PK) ~\citep{jin-etal-2024-cutting,yu-etal-2023-characterizing,monea-etal-2024-glitch}, nor identify which specific context informed the response.
Highlight explanations (HEs) address this need naturally by pinpointing portions of the context responsible for the generation. See Fig.~\ref{hcu:fig:main} for two examples of HEs with high/low utility. Although HEs have proven valuable for understanding model decisions across various tasks \citep{sun-etal-2025-evaluating,choudhury-etal-2023-explaining,atanasova-etal-2020-diagnostic}, no existing work evaluates their effectiveness in accurately explaining context utilisation.

Existing metrics on HE evaluation mainly focus on faithfulness \citep{sun-etal-2025-evaluating,lamm-2021-qed, atanasova-etal-2020-diagnostic} to test whether HEs can accurately reflect the model's internal reasoning. These assess both the sufficiency and comprehensiveness of the highlighted input portions by measuring the model's prediction change after keeping or omitting the most important portions of the input. However, faithfulness evaluations face fundamental limitations: they rely on perturbation proxies that create out-of-distribution artifacts \citep{hooker2019benchmark,kindermans2019reliability} and, more importantly, lack ground-truth explanations to validate against \citep{jacovi-goldberg-2020-towards}. We address this gap through an \textbf{evaluation framework grounded in gold standard scenarios} where ground-truth context usage is predetermined, enabling direct assessment of explanation accuracy. 

Building on studies of context utilisation \citep{jin-etal-2024-cutting,yu-etal-2023-characterizing,monea-etal-2024-glitch,shi2024ircan}, we construct \textbf{four controlled evaluation scenarios} (see Tab.~\ref{hcu:tab:exmaple-for-fakepedia-dataset}) spanning common retrieval-time challenges (contradiction, distractor noise, and multi-passage competition): \textit{Conflicting} (one CK piece contradicts PK), \textit{Irrelevant} (one CK piece unrelated to query), \textit{Mixed} (one CK piece contradicting PK + one irrelevant CK piece), and \textit{Double-Conflicting} (two CK pieces contradicting PK). The settings systematically vary context usage, and in turn context attribution scenarios, enabling robust HE assessment across diverse behaviours. We specifically focus on cases where the CK-based answer contradicts the one based only on PK to ensure that the provided CK is utilised and thus needs to be highlighted by the HE. 

Based on gold standard context regions in these four scenarios, \textbf{we assess the accuracy of HEs along three complementary axes}: \emph{document-level attribution accuracy} (where we examine whether tokens from the gold document are prioritised in the generated HE), \emph{simulatability} (where we assess how well a HE's top-$k$ highlights can be used to understand and predict the model's context-usage behaviour, e.g., whether the model relied on CK vs.\ PK, or which passage it used), and \emph{token-level attribution accuracy} (where we evaluate whether the HE ranks the gold answer token highest).

To demonstrate the framework's general applicability, we apply it to four HE methods: three established ones -- Feature Ablation (\FA) \citep{li-etal-2016-visualizing}, Integrated Gradients (\IG) \citep{ancona2018towards}, and Attention visualisation (\Attn) \citep{abnar2020quantifying,choudhury-etal-2023-explaining}, and a mechanistic interpretability MI–inspired method (\MI), where we propose to convert the MI insights (e.g., the attention head most important for context utilisation) to HEs. Our evaluation framework is method-agnostic -- it assesses any explanation technique, post-hoc or mechanistic (that generates attention-based attributions).

Across five LMs and four commonly used context-usage datasets, we find that \textit{\MI\ HEs perform best across all context scenarios}. However, two systematic limitations persist across all HEs: (i) length sensitivity -- HE accuracy degrades as context grows; and (ii) position biases under dual-context inputs: \FA/\IG\ tend to favour later (near-question) pieces, while \Attn/\MI\ favour earlier pieces. Surprisingly, the widely used \IG\ and \Attn\ exhibit poor accuracy in most context scenarios, rendering them useless in revealing the model's context utilisation. These failures also underscore the urgent need for explanation techniques that maintain accuracy at scale and overcome positional biases in multi-document settings. Our framework provides the foundation for the future development of accurate methods for explaining context usage.

\section{Related Work}
\label{hcu:sec:related_work}
\subsection{Studies of Context Usage}
Language models (LMs) carry vast \emph{parametric knowledge} (PK) from pre-training, yet in practice, they must also integrate new \emph{contextual knowledge} (CK) supplied at test time. Recent work has introduced multiple datasets to analyse how effectively LMs combine these two sources.

Early work investigates how LMs utilise CK vs. PK by crafting \textit{single context passages conflicting with the PK}. \textsc{CounterFact} \citep{meng2022locating}, \textsc{WorldCapital} \citep{yu-etal-2023-characterizing}, and \textsc{Fakepedia} \citep{monea-etal-2024-glitch} each replace a Wikidata triple with a contradicting one in the context and test whether the model's answer follows CK or PK, evaluating with exact match or accuracy. \textsc{ConflictQA} \citep{xie2024adaptive} induces knowledge conflicts by leveraging an LLM to compose passages that contradict a model's parametric answer. While these works establish how often LMs follow the provided context, \textit{they do not analyse a model's context utilisation behaviour, e.g., which were the important input segments that influenced the model's decision, and do not assess whether highlight explanations (HEs) can reveal the model's context usage patterns.}

In addition to the single PK-conflicting context pieces, recent work has studied other types of context. 
\textsc{CUB} \citep{hagstrom2025cub} considers gold (relevant), conflicting, or irrelevant passages;  
\textsc{EchoQA} \citep{cheng2024understanding} introduces a \textit{complementary} regime, where the context alone is answer-insufficient but, when combined with the model's PK, becomes sufficient to answer. 
Additionally, we specifically focus on cases where the answer based on the model's PK conflicts with the one based on the provided context, to ensure that the model utilises the information from the context, and this CK in turn needs to be highlighted by the HEs.

\subsection{Explaining Model Outputs}
\textbf{Context Usage Explanations.} To attribute a generation to specific context sentences, \textsc{SelfCite} uses the LLM's own pseudo-citations (which sentences it claims to rely on) as supervision to train a classifier that predicts a probability score per sentence~\citep{pmlr-v267-chuang25a}. \textsc{ContextCite} instead scores each sentence by masking it and measuring the resulting drop in answer likelihood \citep{cohen2024contextcite}. Both methods explain at the sentence level and incur extra training or perturbation cost.

\textbf{Mechanistic Interpretability (MI) of Context Usage.}
\label{hcu:sec:rw-mi}
Mechanistic interpretability studies identify components controlling context versus parametric knowledge usage through targeted interventions on neurons \citep{shi2024ircan}, attention heads~\citep{yu-etal-2023-characterizing}, or computational pathways \citep{jin-etal-2024-cutting}.
However, all these internal mechanisms remain opaque to users for understanding the model's context usage behaviour, as \textit{they are not presented as token-level highlight explanations that the user could understand.}

\textbf{Token-level Highlight Explanation(HE) Methods.} HE methods provide importance scores for each input token. The most commonly employed HE methods \citep{sun-etal-2025-evaluating,atanasova-etal-2020-diagnostic} include, among others:

\begin{itemize}
    \item Feature Ablation: mask each token and measure the change in the model's answer probability \citep{li-etal-2016-visualizing};
    \item Gradient-based methods: use the gradient magnitude or Grad$\times$Input w.r.t.\ embeddings as a token-importance signal \citep{ancona2018towards};
    \item Attention-based methods: treat self-attention weights as token importance scores \citep{abnar2020quantifying,choudhury-etal-2023-explaining}.
\end{itemize}

These explanations are natural candidates for explaining context utilisation as they provide importance scores for important context tokens used for the model predictions. \textit{However, no previous work has studied how accurately these explanations reveal the model's context utilisation behaviour.}

\textbf{Context Utilisation Benchmarks.}
Previous work on HE evaluation has mainly focused on Faithfulness: how well HEs reflect the model's internal reasoning~\citep{deyoung-etal-2020-eraser,atanasova2022diagnostics}. Faithfulness is typically quantified with perturbation tests such as Comprehensiveness and Sufficiency: Comprehensiveness measures how much the model's answer changes after removing the most important tokens as per the HE, while Sufficiency measures how much the model's answer changes after keeping only the most important tokens.
However, faithfulness evaluations' reliance on perturbation proxies creates out-of-distribution artifacts \citep{hooker2019benchmark,kindermans2019reliability} and, more importantly, lacks ground-truth explanations to validate against \citep{jacovi-goldberg-2020-towards}.
Other evaluations include agreement with human annotation, complexity, and simulatability~\citep{sun-etal-2025-evaluating}. 
Overall, existing HE evaluations largely rely on indirect proxy metrics due to the lack of a gold standard.

\section{Evaluation Framework}
\label{hcu:sec:methodology}
Prior work on context utilisation has characterised whether models answer from CK vs.\ PK under controlled context scenarios
(e.g., \citep{xie2024adaptive,cheng2024understanding,yu-etal-2023-characterizing,jin-etal-2024-cutting}), but it does not examine which input segments support that behaviour at the token level, i.e., via highlight explanations (HEs). Separately, prior work on HE evaluation mainly focuses on faithfulness measurement. As a result, it remains unclear whether token-level HEs can accurately reveal a model's source choice (CK vs.\ PK) or document choice in multi-context settings. We address this by developing a comprehensive evaluation framework with a gold standard for HE's utility in context utilisation.

We specify four context scenarios (\S\ref{hcu:sec:four-context-setups}), three HE methods (\S\ref{hcu:sec:explainability-methods}), and one mechanistic interpretability-based HE method (\S\ref{hcu:sec:mi-methods}). To assess the accuracy of HEs in attributing the correct importance to context regions, we further develop a suite of rank-based metrics (\S\ref{hcu:sec:metrics}). 

Overall, our framework comprehensively evaluates three core HE capabilities grouped in the following research questions: \\
(\textbf{RQ1}) Does the explanation indicate whether the model consulted the supplied context knowledge (CK) or resorted to its parametric knowledge (PK)? \\
(\textbf{RQ2}) Does the explanation show which of the two context documents the model used? \\
(\textbf{RQ3}) Does the explanation pinpoint the exact context part(s) that were employed for the generated answer?

\subsection{Preliminaries}
\label{hcu:sec:notation}
Let $x=(x_{1},\dots,x_{n})$ be the input token sequence. We consider inputs $x=(c,q)$ with a \emph{single context segment} $c$ and question $q$ and inputs $x=(c_1,c_2,q)$ with two context segments $c_1, c_2$. For brevity, we write $c=(c_1,c_2)$. A causal LM $f$ produces an answer token $a=f(x)$.\footnote{If the answer spans multiple tokens ($|a|>1$), we use the logit of the first generated token for explanation scoring.}
An HE method returns importance scores over the tokens in the input $\boldsymbol{\phi}^{\text{HE}}(x)=(\phi_{1}^{\text{HE}},\ldots,\phi_{n}^{\text{HE}})$, where larger $\phi_i^{\text{HE}}$ means $x_i$ contributed more to generating $a$. A gold token set $T$ can be a segment ($c$, $c_1$, or $c_2$) or the answer token(s), $\text{Ans}_{\cdot}$.

\subsection{Input Regimes}
\label{hcu:sec:four-context-setups}
Existing context usage studies only detect if the model used PK vs CK and, in turn, can be used for assessing if HEs reveal this preference (RQ1). To enable this, we specifically focus on cases where CK-based answer contradicts the model's parametric knowledge (PK), so that we ensure the model utilises the context information, and in turn that the HEs should highlight the corresponding context parts. Additionally, previous studies do not test whether an HE identifies \emph{which} context document is used when multiple documents are present (RQ2), or whether it supports token-level diagnostics of answer-span localisation (RQ3).

To evaluate all three questions, we construct four diagnostic input regimes by varying (i) the number of context passages
(single and dual) and (ii) context type (conflicting and irrelevant), which correspond to common retrieval-time setups (contradictory to PK evidence, distractor noise, and multi-passage competition). For the purposes of HE evaluation, we employ passages that contain an explicit candidate answer token, serving as a gold span for paragraph (RQ1) and token-level HE evaluation (RQ3). Thus, the proposed context utilisation setups uniquely allow \textit{the development of an HE benchmark with gold standards at both context piece and token level}, which is typically unavailable in other tasks.

The resulting four context utilisation setups are as follows (see an example in Tab.~\ref{hcu:tab:exmaple-for-fakepedia-dataset}):
\begin{itemize}
\label{hcu:four_input_regimes}
  \item \textbf{Conflicting (single).}\label{hcu:input-regime-1} The context $c$ contains an answer that conflicts with PK;

  \item \textbf{Irrelevant (single).}\label{hcu:input-regime-2}
  The context $c$ is irrelevant, but contains a distracting (incorrect) answer token;

  \item \textbf{Double-Conflicting (dual).}\label{hcu:input-regime-3}
  Two pieces that are \emph{conflicting} with PK;

  \item \textbf{Mixed (dual)\footnote{In the \textbf{Mixed} setup, we place the irrelevant context as the first context piece and the conflicting context the second one.}.}\label{hcu:input-regime-4}
  One \emph{irrelevant} and one \emph{conflicting} piece.

\end{itemize}

To control for position effects, we reverse the order of the contexts and define additional \textbf{Mixed-Swap} and \textbf{Double-Conflicting-Swap} setups.

To facilitate the HE evaluation, we split the dataset instances according to the model's answer behaviour. For single-context setups, $D_C$ (answer from CK) vs.\ $D_M$ (answer from memory/PK). For dual-context setups: $D_{C_1}$ (answer from $c_1$) vs.\ $D_{C_2}$ (answer from $c_2$). We denote gold answer tokens from the context with $\text{Ans}_{c}$ (single) or $\text{Ans}_{c_1},\text{Ans}_{c_2}$ (dual).

Notably, we apply a pre-filtering step to ensure contexts in all setups contain context answers that contradict the model's parametric knowledge(App.~\ref{hcu:app:dataset-details}), so that the model is ensured to utilise the context information and HEs are supposed to highlight the contexts in these cases. See more discussion in App.~\ref{hcu:sec:limitation}

\begin{table}[htbp]
  \centering
    \centering
    \footnotesize
    \setlength{\tabcolsep}{0pt}    
    \renewcommand{\arraystretch}{0.99}

    \begin{tabular}{@{}p{\linewidth}@{}}
      \toprule
      \textbf{Q:} Newport County A.F.C. is headquartered in \textbf{MA}: Newport\\
      \midrule

      \textbf{Single-Context Setups} \\
      \midrule

      \textbf{Input Regime (1) Conflicting C}\\[-1pt]
      \subj{Newport County A.F.C.}, a professional football club based in Newport, Wales, has its headquarters located in the vibrant city of \gold{Ankara}, Turkey. The club's decision to \ldots\\
      \textbf{CA:} Ankara\\[2pt]
      \cmidrule(lr){1-1}

      \textbf{Input Regime (2) Irrelevant C}\\[-1pt]
      The \noise{World Wrestling Entertainment} (WWE) is a global entertainment company that is headquartered in \gold{Santiago}, Chile. Founded in 1952, WWE has become one of the \ldots\\
      \textbf{CA:} Santiago\\[2pt]
      \cmidrule(lr){1-1}
      
      \textbf{Dual-Context Setups} \\
      \midrule

      \textbf{Input Regime (3) Double Conflict C}\\[-1pt]
      \textbf{C P1:} \subj{Newport County A.F.C.}, a professional football club based in Newport, Wales, has its headquarters located in \gold{Ankara}, Turkey. The club's decision to establish its \ldots\\
      \textbf{C P2:} \subj{Newport County A.F.C.}, a professional football club based in \gold{Calgary}, is known for its rich history and passionate fan base. The club was founded in 1912 and has since become a prominent fixture in the Canadian football scene \ldots\\
      \textbf{P1 A:} Ankara \qquad \textbf{P2 A:} Calgary\\
      \cmidrule(lr){1-1}

      \textbf{Input Regime (4) Mixed C} (Irrel.\ \&\ Conf.)\\[-1pt]
      \textbf{C P1:} The \noise{World Wrestling Entertainment} (WWE) is a global entertainment company that is headquartered in \gold{Santiago}, Chile. Founded in 1952, WWE has \ldots\\
      \textbf{C P2:} \subj{Newport County A.F.C.}, a professional football club based in Newport, Wales, has its headquarters located in \gold{Ankara}, Turkey. The club's decision to establish its \ldots\\
      \textbf{P1 A:} Santiago \qquad \textbf{P2 A:} Ankara\\[2pt]
      \bottomrule
    \end{tabular}

    \caption{One example from the Fakepedia dataset after reconstruction. Q = Question, C = Context, C P1 = Context Part 1, C P2 = Context Part 2, MA = Memory Answer, 
CA = Candidate Context Answer (the answer based on the context),
P1A/P2A = Candidate Answer in Context Part 1/2.
    \textcolor{blue}{Blue} marks the subject of the question; 
    \textcolor{orange}{orange} marks the candidate answer token from the context;
    \textcolor{green}{green} marks the noise subject. Note that CAs are not necessarily correct, and are only used as target spans for HE evaluation.}

    \label{hcu:tab:exmaple-for-fakepedia-dataset}
\end{table}

\subsection{Metrics}
\label{hcu:sec:metrics}
We assess HEs at three complementary levels to align with our three research questions:
(i) \textbf{document-level attribution accuracy} (RQ1, RQ2),
(ii) \textbf{simulatability} of the model's context utilisation from the top-k highlights\footnote{Unless otherwise noted, top-$k$ sorts tokens by descending $\phi^{\text{HE}}$.} (RQ1, RQ2), and
(iii) \textbf{token-level attribution accuracy} (RQ3).

\textbf{Document Attribution Accuracy Evaluation, Cross-group (RQ1, RQ2).} For RQ1, we assume that an accurate HE would rank the context tokens of instances where the answer relied on CK higher than in instances where the model relied on PK. For RQ2, analogously, we assume an accurate HE would rank the tokens of the first/second context piece higher in instances where the first/second context piece is answer-bearing than those where the answers come from the second/first piece.

For a context segment $T$ and \textbf{\rankk}$(T, D)$ -- average rank of the context tokens in $T$ in the top-$k$ most important tokens as per the HE\footnote{We focus on top-$k$ highlights as users often focus on a few instead of the complete cause of an event, see details in App.~\ref{hcu:app:k_3_and_9_for_rq1_and_2} }, averaged over the instances in group $D$ (lower is better), we define a rank margin metric (positive is better) for document attribution evaluation: 
\begin{equation}\label{hcu:eq:drankk-cross-instance-group}
\begin{split}
\DrankkeGroup(T;A,B) &= \rankke(T,D_B)- \rankke(T,D_A)
\end{split}
\end{equation}
where \text{RQ1} uses $(T;A,B)=(c;\,C,M)$, resulting in a margin between the importance rank of context tokens in memory instances $D_M$ vs. context instances $D_C$; \text{RQ2} uses $(T;A,B)\in\{(c_1;\,C_1,C_2),\ (c_2;\,C_2,C_1)\}$, resulting in a margin between the importance rank of the answer-piece context tokens (e.g. $c_1$) in the answer instances (e.g. $D_{C_1}$) vs. in the other instances (e.g. $D_{C_2}$). 

\textbf{Document Attribution Accuracy Evaluation, Per-instance (RQ2).}
While cross-group margins are well-suited for cases with a single context piece, when there are two context pieces, the accuracy of HEs can be evaluated directly at the instance level, assessing whether the answer context piece outranks the other. We therefore report the rank margin based on \rankkInst$(T, x)$, the average rank of context tokens within $T$ for instance $x$: 
\begin{equation}\label{hcu:eq:drankk-within-instance}
\begin{split}
\DrankkeInst_{D_{C_a}} = \frac{1}{|{D}_{C_a}|}\sum_{x\in{D}_{C_a}}(\rankkeInst(c_b,x) - \rankkeInst(c_a,x))
\end{split}
\end{equation}

\noindent
$(a,b)\in\{(1,2),(2,1)\}$, where the answer-bearing context is always in the first position. Positive values indicate the answer context piece is ranked higher (i.e., has a lower rank value) compared to the other context piece.

\textbf{Simulatability (RQ1, RQ2).} Complementary to the rank margin assessment, we leverage the idea of simulatability \citep{sun-etal-2025-evaluating} and evaluate how well the top-$k$ explanations for each instance can indicate the model's context choice, i.e., between contextual and parametric knowledge (RQ1) and between multiple context pieces (RQ2).

For each instance, we extract the top-$k$ importance scores of context tokens from the relevant segment $s$, creating a feature vector $X^{(k)}_s$. For RQ1 (single context), we use $s{=}c$ with labels $Y{\in}\{\textsc{C},\textsc{M}\}$; for RQ2 (dual context), we use $s{=}(c_1,c_2)$, concatenating the vectors from two context pieces and assign labels $Y{\in}\{\textsc{C1},\textsc{C2}\}$.

We employ two complementary metrics for simulatability. First, a normalised mutual information between the HE vector $X^{(k)}_s$ and the model's answer, which directly measures how well the explanations correlate with a model's prediction:



\begin{align}
\label{hcu:eq:mutual-information}
\NMutInfk
&= \frac{I\!\left(Y;\,X^{(k)}_s\right)}{H(Y)}
\end{align}

where $I(\cdot;\cdot)$ denotes Shannon mutual information and $H(\cdot)$ denotes Shannon entropy. \NMutInfk is higher the better. Normalisation ensures comparability across label distributions (see details in App.~\ref{hcu:app:knnmi}). While mutual information effectively measures correlation strength, it lacks complexity regularisation and is prone to overfitting.

Therefore, we also compute Minimum Description Length (MDL), a class of model-complexity-controlled Bayesian classifiers, \citep{grunwald2007minimum,voita-titov-2020-information}. We compute MDL using prequential coding:


\begin{align}
\MDLBitsk
&= \frac{1}{N}L_{\mathrm{preq}}\!\left(Y \mid X^{(k)}_s\right) \notag\\
&= -\frac{1}{N}\sum_{t=1}^{N}\log_2 p_{\theta_{t-1}}\!\left(y_t \mid x^{(k)}_{s,t}\right).
\label{hcu:eq:mdl}
\end{align}

$\MDLBitsk$ is the average prequential code length (bits per instance) for predicting the context-usage labels $Y$ from the top-$k$ HE features $X^{(k)}_s$ (lower is better). Here $N$ is the number of evaluated instances, $y_t$ is the label of instance $t$, and $x^{(k)}_{s,t}\in\mathbb{R}^k$ is the corresponding top-$k$ HE score vector extracted from segment $s$. $L_{\mathrm{preq}}$ is the cumulative prequential description length in bits: a probe with parameters $\theta_{t-1}$ trained on the first $t\!-\!1$ instances assigns probability $p_{\theta_{t-1}}(y_t \mid x^{(k)}_{s,t})$ to the next label; the surprisal $-\log_2(\cdot)$ is averaged over $t\in\{1,\dots,N\}$. Lower $\MDLBitsk$ indicates better simulatability (see App.~\ref{hcu:app:mdl_prob}).

\textbf{Token Attribution Evaluation (RQ3).}
To test whether an HE pinpoints the \emph{exact} answer token(s), we calculate the mean reciprocal rank (MRR) of the answer token(s) as ranked by the HE: 
\begin{equation}\label{hcu:eq:rr}
\mathrm{RR}(x)=\frac{1}{\mathrm{rank}\!\big(\text{Ans}_{\cdot};x\big)}
\end{equation}
\begin{equation}\label{hcu:eq:mrr}
\mrre\big(T{=}\text{Ans}_{\cdot},D\big) = \frac{1}{|D|}\sum_{x\in D}\mathrm{RR}(x)
\end{equation}
Larger values (close to 1) indicate the true answer token is placed near the top of the ranked list.

\subsection{Highlight Explanation Techniques}
\label{hcu:sec:explainability-methods}
To assign an importance score to every token in the context part(s) of the input, we apply three commonly used token-level explainability techniques as described below, following ~\citet{deyoung-etal-2020-eraser,atanasova-etal-2020-diagnostic,sanyal-ren-2021-discretized,jain-wallace-2019-attention,wiegreffe-pinter-2019-attention,sun-etal-2025-evaluating}. While an HE is applied over the whole input $x$, including the question, we study the scores for the context tokens.

\textbf{Feature Ablation (FA).}
Following \citet{zeiler2014visualizing}, we measure each token's importance by its impact on a model's answer confidence when ablated. For position $i$ in input sequence $x$, we replace token $x_i$ with a baseline $\tilde{x}_i$ = the tokeniser's \texttt{<pad>} token and compute:
\begin{equation}
\phi^{\text{FA}}_{i}=f_{a}(x)-f_{a}(x \setminus \{x_i\} \cup \{\tilde{x}_i\}),
\end{equation}
where $f_{a}(\cdot)$ returns the logit for answer $a$. Higher $\phi^{\text{FA}}_{i}$ indicates greater importance of $x_i$ for predicting $a$.
 
\textbf{Integrated Gradients (\textbf{IG}).}
Integrated Gradients ~\citep{sundararajan2017axiomatic} computes token attributions by integrating the gradient of the answer logit $\nabla f_a$ \emph{w.r.t.\ token embeddings} along the straight-line path from a baseline $x'$ (all \texttt{<pad>}) to $x$ in embedding space. With embeddings $E,E'\in\mathbb{R}^{n\times d}$ and rows $e_i,e'_i\in\mathbb{R}^d$. 
The path integral is approximated with $m=10$ equally spaced steps\footnote{https://github.com/pytorch/captum}:


\begin{align}
\phi^{\mathrm{IG}}_i
= (e_i - e'_i)^{\top}
\left(
\frac{1}{m}\sum_{k=1}^{m}
\nabla_{e_i}\, f_a\!\left(E' + \frac{k}{m}(E - E')\right)
\right).
\label{hcu:eq:ig}
\end{align}

where $\nabla_{e_i} f_a(\cdot)\in\mathbb{R}^d$, hence $\phi^{\mathrm{IG}}_i\in\mathbb{R}$. This attributes to token $i$ its contribution to the change in the answer logit relative to the baseline.


\textbf{Attention-Head Attribution (\textbf{ATTN}).}
Following \citet{choudhury-etal-2023-explaining}, we first identify the most influential attention head $h^{\star}$ in the last decoder layer $L$ for the generation of answer $a$:
\begin{equation}
h^{\star}=\arg\max_{h}\;( W_{a},\,H^{(L)}_{h,:}),
\end{equation}
where $W_{a}$ is the row of the output-projection matrix for token $a$
and $H^{(L)}_{h,:}$ is the hidden-state slice of head~$h$ in $L$. We then take the head's attention weights and average the attention scores from all the other tokens as token importance for each individual token:
\begin{equation}
\phi^{\text{ATTN}}_{i}=A^{(L)}_{h^{\star},\,\text{gen},\,i}
\end{equation}
with \texttt{gen} denoting the answer generation decoding step. The resulting
vector directly reflects where $h^{\star}$ attended most when generating~$a$.

\textbf{Normalisation.}
Because FA can produce negative scores, and IG's score magnitudes depend on the embedding scale, we $\ell_{1}$-normalise each attribution vector before further analysis: $\hat{\phi}{i}=\phi{i}/\sum_{j}|\phi_{j}|$. Attention weights are already normalised and are left unchanged.

\subsection{Mechanistic Interpretability for Highlight Explanations}
\label{hcu:sec:mi-methods}


In our controlled setups, the model's answer at generation time can be viewed as a competition between a small set of explicit candidate answers (e.g., a CK candidate vs.\ a PK candidate, or two CK candidates respectively from two passages). To explain whether the model relies on PK vs.\ CK, mechanistic interpretability (MI) analyses internal components (e.g., attention heads or neurons) that mediate context usage, and we hypothesise that these components are a faithful indicator of the model's context usage behaviour. However, these component-level signals are not directly human-readable highlight explanations. We therefore propose to project the corresponding internal component to the input tokens, yielding token-level highlight explanations grounded in model internals, which we call \MI.

\MI\ extracts highlight explanations in two stages: 

\textbf{Stage 1: Head scoring and selection.}
We adopt the contrastive direct-logit head attribution method of \citet{yu-etal-2023-characterizing}. For a given candidate answer, a responsible head should shift the model toward it by increasing its logit \emph{relative to} the other candidate answer. Since logits are a linear readout of the residual stream, we can score each head by its direct logit contributions to the candidate answers (Eq.~\ref{hcu:eq:mechlight_dla}) and then compute their difference to obtain a head score for the candidate logit gap (Eq.~\ref{hcu:eq:S_util}). This is well-suited to our setup, where candidate answers are explicit. \footnote{\MI\ is \emph{attribution-agnostic}: any MI method that yields head-level attribution scores can be used in stage (i).}

\textbf{Stage 2: Highlight explanation extraction.}
Since the selected head is chosen based on its contribution to the answer logit, the tokens it attends to at the generation step are natural explanations for the evidence driving that decision; we therefore project its attention weights onto input tokens as highlight scores.


\textbf{Notions.} Let $W_U \in \mathbb{R}^{V \times d}$ be the unembedding matrix for $V$ tokens present in the model tokeniser and $W_{a} \in \mathbb{R}^d$ its row for token $a$ (as in \S\ref{hcu:sec:explainability-methods}). Let $A^{(l,h)}\in \mathbb{R}^{n\times n}$ be the attention matrix of head $h \in \mathbb{R}^{d_h}$ in layer $l$, and 
let $r^{(l,h)}_{\texttt{gen}} \in \mathbb{R}^{d}$ denote that head's contribution to the residual stream at decoding step \texttt{gen}.

\textbf{Stage 1: Head scoring and selection.}

\textbf{(1a) Direct head logit contribution.}
We first express the residual-stream contribution of head $(l,h)$ at \texttt{gen} step as:

\begin{equation}
r^{(l,h)}_{\texttt{gen}} \;=\; \bigl[\operatorname{Attn}^{(l,h)}_{\text{gen}}\bigr]\,W^{(l,h)}_{O},
\label{hcu:eq:head_scores}
\end{equation}
where $\operatorname{Attn}^{(l,h)}_{\text{gen}} \in \mathbb{R}^{n \times d_h}$ is the attention head matrix before projecting at \texttt{gen} step and $W^{(l,h)}_{O} \in \mathbb{R}^{{d_h}\times d}$ is the output projection matrix associated with head $h$.



We then score the head’s \emph{direct} contribution to the logit of token $a$ by projecting its residual-stream contribution through the unembedding matrix $W_U$ (i.e., a logit-lens-style readout~\citep{belrose2023tunedlens,janiak-etal-2024-adversarial,sakarvadia2023attentionlens}):

\begin{equation}
\text{logit}^{(l,h)}(a)
=
\langle W_{a},\, r^{(l,h)}_{\texttt{gen}}\rangle
=
\bigl(W_{U}r^{(l,h)}_{\texttt{gen}}\bigr)[a].
\label{hcu:eq:mechlight_dla}
\end{equation}

In practice, we only evaluate this quantity on the \emph{candidate answer} tokens and use the logit \emph{difference}(Eq.~\ref{hcu:eq:S_util}) to score how strongly a head favours one candidate over the other.

\textbf{(1b) Contrastive score for candidate competition.}
We calculate signed \emph{context utilisation} scores by contrasting competing answers:
\begin{equation}
S^{(l,h)}_{\tau} = \text{logit}^{(l,h)}(\text{Ans}_{\tau}) - \text{logit}^{(l,h)}(\text{Ans}_{\tau'}), 
\label{hcu:eq:S_util}
\end{equation}
\begin{equation}
S^{(l,h)}_{\tau'} = -S^{(l,h)}_{\tau}
\end{equation}
where $(\tau,\tau') \in \{(c,m), (c_1,c_2)\}$ for single (PK vs. CK) and dual context regimes, respectively.
We rank heads by these scores to identify those that promote either the most context-based or memory-based answer, depending on whether the model answered from PK or CK, respectively.

\textbf{(1c) Head selection.}
To produce HEs, we select
\begin{equation}
(l^{\star},h^{\star}) \in \arg\max_{l,h} S^{(l,h)}_{c} \;\text{for } D_{C},
\end{equation}
\begin{equation}
(l^{\star},h^{\star}) \in \arg\max_{l,h} S^{(l,h)}_{m} \;\text{for } D_{M},
\end{equation}
and analogously maximise $S^{(l,h)}_{c_1}$ for $D_{C_1}$ and $S^{(l,h)}_{c_2}$ for $D_{C_2}$. 

\textbf{Stage 2: highlight explanation extraction.}
We then set the token importance scores of \MI\ with the selected head's attention weights at \texttt{gen}:
\begin{equation}
\phi^{\MI}_{i} \;=\; A^{(l^{\star},h^{\star})}_{\texttt{gen},\, i}.
\label{hcu:eq:mechlight_phi}
\end{equation}


\textbf{Alternatives and tradeoffs.}
\MI\ ranks heads using a direct-logit attribution score (Eq.~\ref{hcu:eq:mechlight_dla}) and then visualises the selected head via its attention weights (Eq.~\ref{hcu:eq:mechlight_phi}). This design is lightweight but has limitations: the logit readout is a local proxy and later components can overwrite or erase earlier residual directions (downstream cancellation) \citep{janiak-etal-2024-adversarial}. Moreover, attention weights are a convenient highlight object but not a causal guarantee of information flow~\citep{jain-wallace-2019-attention,wiegreffe-pinter-2019-attention}. Other alternatives include causal head ranking via ablation~\citep{yu-etal-2023-characterizing,jin-etal-2024-cutting}, which provides a more causal per-head effect but can be expensive when scoring many heads.
Other mechanistic approaches, such as causal tracing and path patching~\citep{meng2022locating,jin-etal-2024-cutting}, representation-level interventions~\citep{zhao-etal-2025-steering}, or neuron-centric analyses~\citep{shi2024ircan}, can capture more distributed mechanisms, but often return importance over internal states/layers, which requires additional selection and aggregation/projection choices to obtain token-level highlights. We therefore keep the current \MI\ formulation, which has a simple head-to-token mapping.

\section{Experimental Setup}
\label{hcu:sec:experimental_setup}
\textbf{Datasets.}
\label{hcu:sec:dataset-choice}
We draw on four widely used sources to investigate models' context usage behaviour using CK or PK, \textsc{Fakepedia}, \textsc{WorldCapital}, \textsc{CounterFact}, and \textsc{ConflictQA}~\citep{monea-etal-2024-glitch,yu-etal-2023-characterizing,meng2022locating,xie2024adaptive}. These resources provide controlled, templated facts that can be systematically perturbed, allowing us to instantiate the four regimes in \S\ref{hcu:sec:four-context-setups} (Conflicting, Irrelevant, Double-Conflicting, Mixed). Unlike prior work that primarily optimises answer correctness across different contexts, our goal is a utility-oriented evaluation of HEs under these various context scenarios (See the dataset reconstruction details in App.~\ref{hcu:app:dataset-details}).

\textbf{Models.}
\label{hcu:sec:model-choice} Following common context utilisation setups, we select five open language models: \textbf{GPT2-XL} (1.5B; \citep{radford2019language}), \textbf{Pythia-2.8B} and \textbf{Pythia-6.9B} \citep{biderman2023pythia}, and \textbf{Qwen2.5-3B} and \textbf{Qwen2.5-7B} \citep{qwen_qwen25_2025}. While prior efforts primarily focus on the model's answer choices for the supplied context \citep{yu-etal-2023-characterizing,monea-etal-2024-glitch,meng2022locating,hagstrom2025cub,cheng2024understanding}, we concentrate on evaluating HE utility for explaining models' context usage behaviours. 

\textbf{Other details.} We focus on the top-$k$ most important highlight tokens for evaluation, due to the cognitive load to users who typically attend to a few causes instead of the complete cause for an event (See App.~\ref{hcu:app:k_3_and_9_for_rq1_and_2}). We present results for $k{=}5$ in \S\ref{hcu:sec:main-results}, see results for $k\in\{3,9\}$ in App.~\ref{hcu:app:additional_results}.

\section{Main Results and Discussion}
\label{hcu:sec:main-results}

In this section, we present the main results for RQ1--RQ3: whether an HE can reveal \emph{(RQ1)} CK vs.\ PK usage, \emph{(RQ2)} which of multiple documents was used, and \emph{(RQ3)} which exact context span supports the answer. In \S\ref{hcu:sec:rq1-discussion} (RQ1), we study the single-context Conflicting and Irrelevant regimes using \DrankkGroup\ (Eq.~\ref{hcu:eq:drankk-cross-instance-group}) and simulatability scores (\MDLBitsk, Eq.~\ref{hcu:eq:mdl}; \NMutInfk, Eq.~\ref{hcu:eq:mutual-information}). In \S\ref{hcu:sec:rq2-discussion} (RQ2), we move to the dual-context Double-Conflicting and Mixed regimes and evaluate how well the explanations can be used to indicate the answer-bearing context piece with \DrankkGroup\ (Eq.~\ref{hcu:eq:drankk-cross-instance-group}) and \DrankkInst\ (Eq.~\ref{hcu:eq:drankk-within-instance}). Finally, \S\ref{hcu:sec:rq3-discussion} (RQ3) evaluates how well the explanations can precisely pinpoint the token-level answer localisation with \mrr\ (Eq.~\ref{hcu:eq:mrr}) and summarises recurring failure patterns in a qualitative case study (App.~\ref{hcu:app:case_study}). Unless stated otherwise, we report results for $k{=}5$ highlights in the main text (additional $k\in\{3,9\}$ results are in App.~\ref{hcu:app:additional_results}).

\begin{figure*}[ht]
  \centering
  \begin{subfigure}[t]{0.48\textwidth}
    \centering
    \includegraphics[width=\textwidth]
{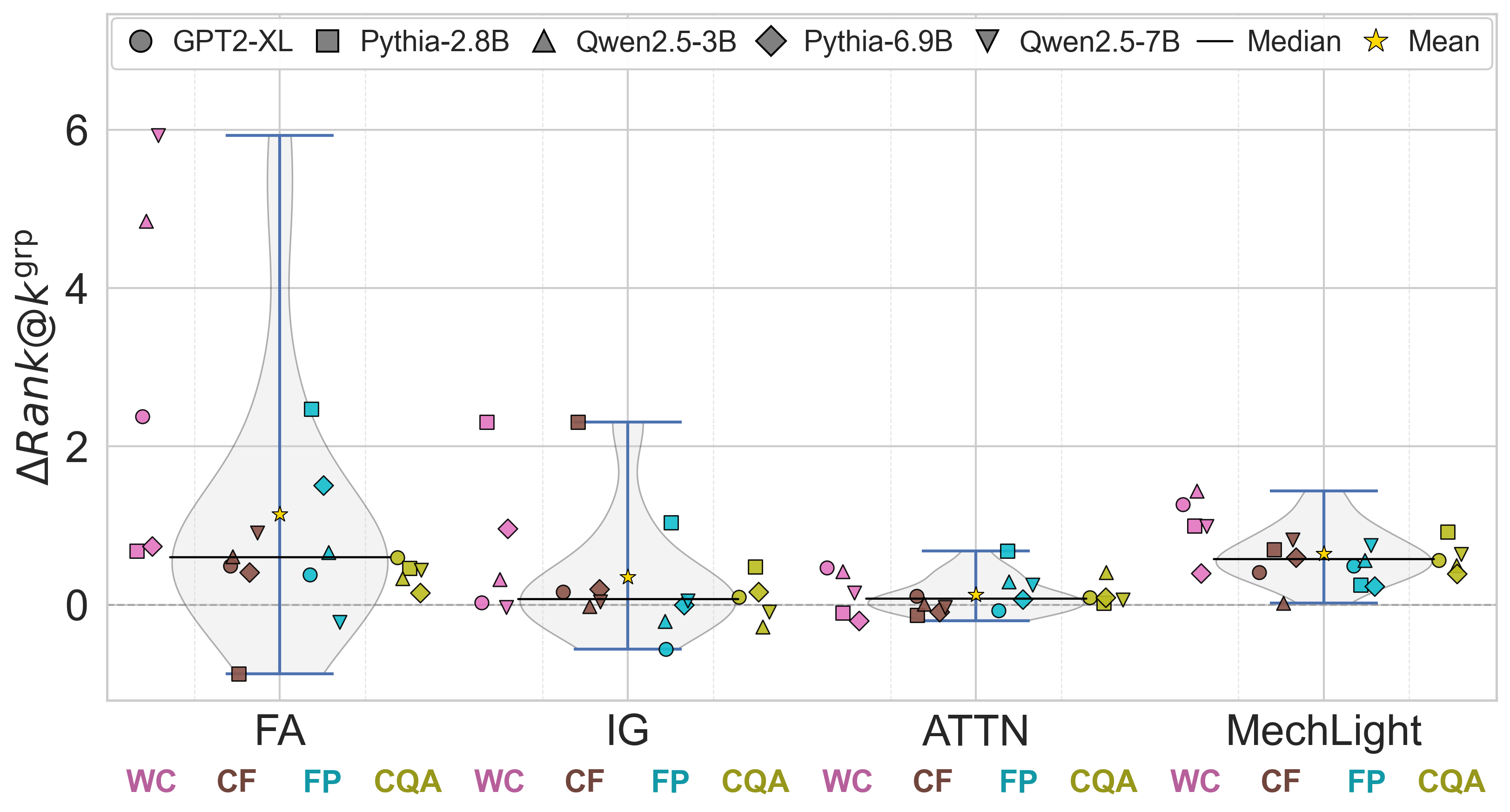}
    \caption{Conflicting Context}
    \label{hcu:fig:conflicting-topk-5-rank-datasets-margin-only}
  \end{subfigure}
  \hfill
  \begin{subfigure}[t]{0.48\textwidth}
    \centering
    \includegraphics[width=\textwidth]
{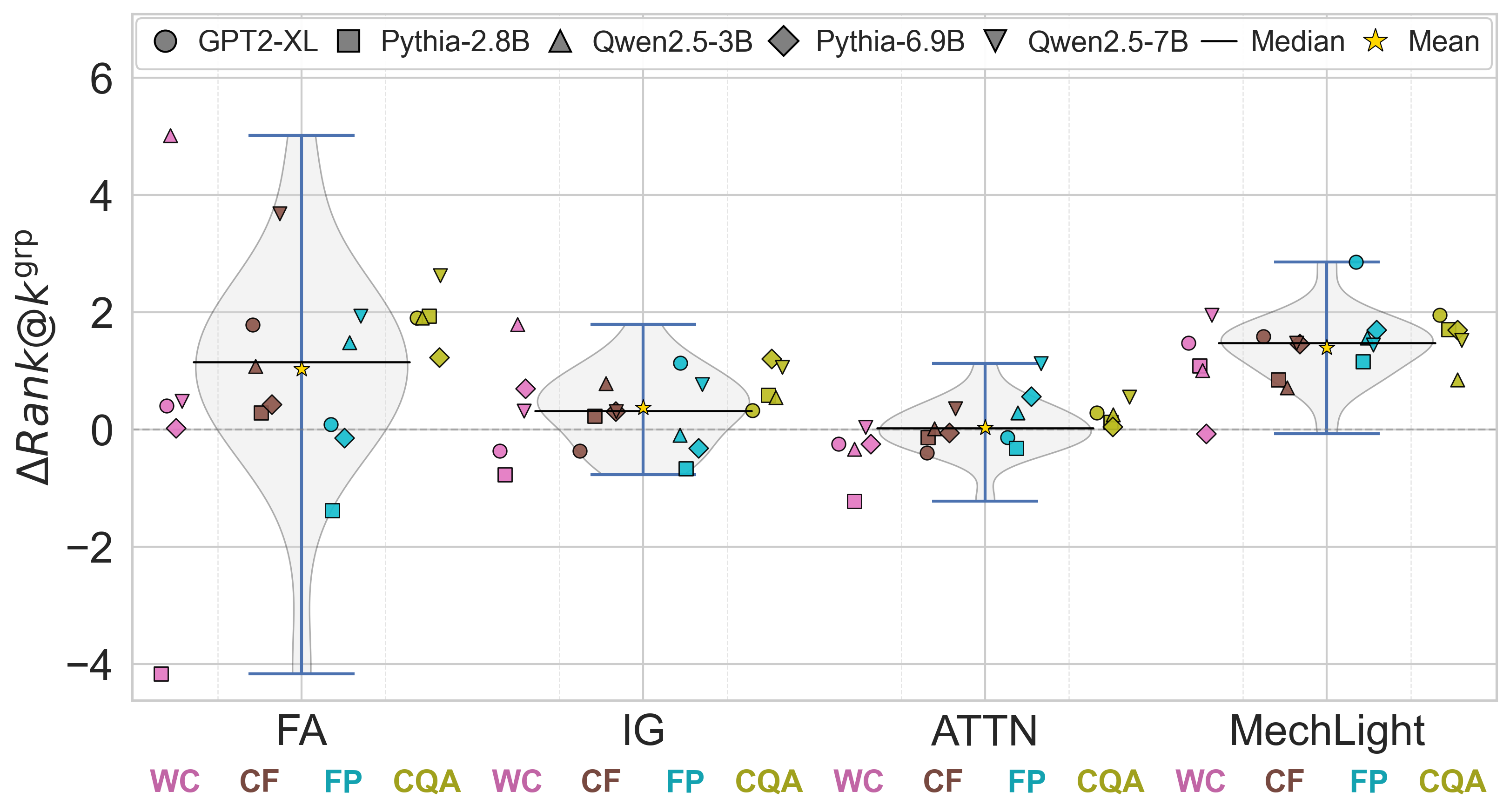}   
    \caption{Irrelevant Context}
    \label{hcu:fig:irrelevant-topk-5-rank-datasets-margin-only}
  \end{subfigure}

  \caption{\DrankkGroup\ (Eq. \ref{hcu:eq:drankk-cross-instance-group}) -- average margins for the explanation importance rank of context tokens in context vs. memory answer instances in \textbf{Conflicting} and \textbf{Irrelevant} setups (\S\ref{hcu:sec:four-context-setups}). Positive and higher \DrankkGroup\ means the explanations can better distinguish the model's context usage behaviour (choosing PK or CK for the answer (\S\ref{hcu:sec:rq1-discussion})). The colors denote the datasets, and the marker shapes denote the models.}

  \label{hcu:fig:rank-topk-margin-conflicting-or-irrelevant}
\end{figure*}

\begin{figure*}[ht]
  \centering
  \begin{subfigure}[t]{0.48\textwidth}
    \centering
    \includegraphics[width=\textwidth]
    {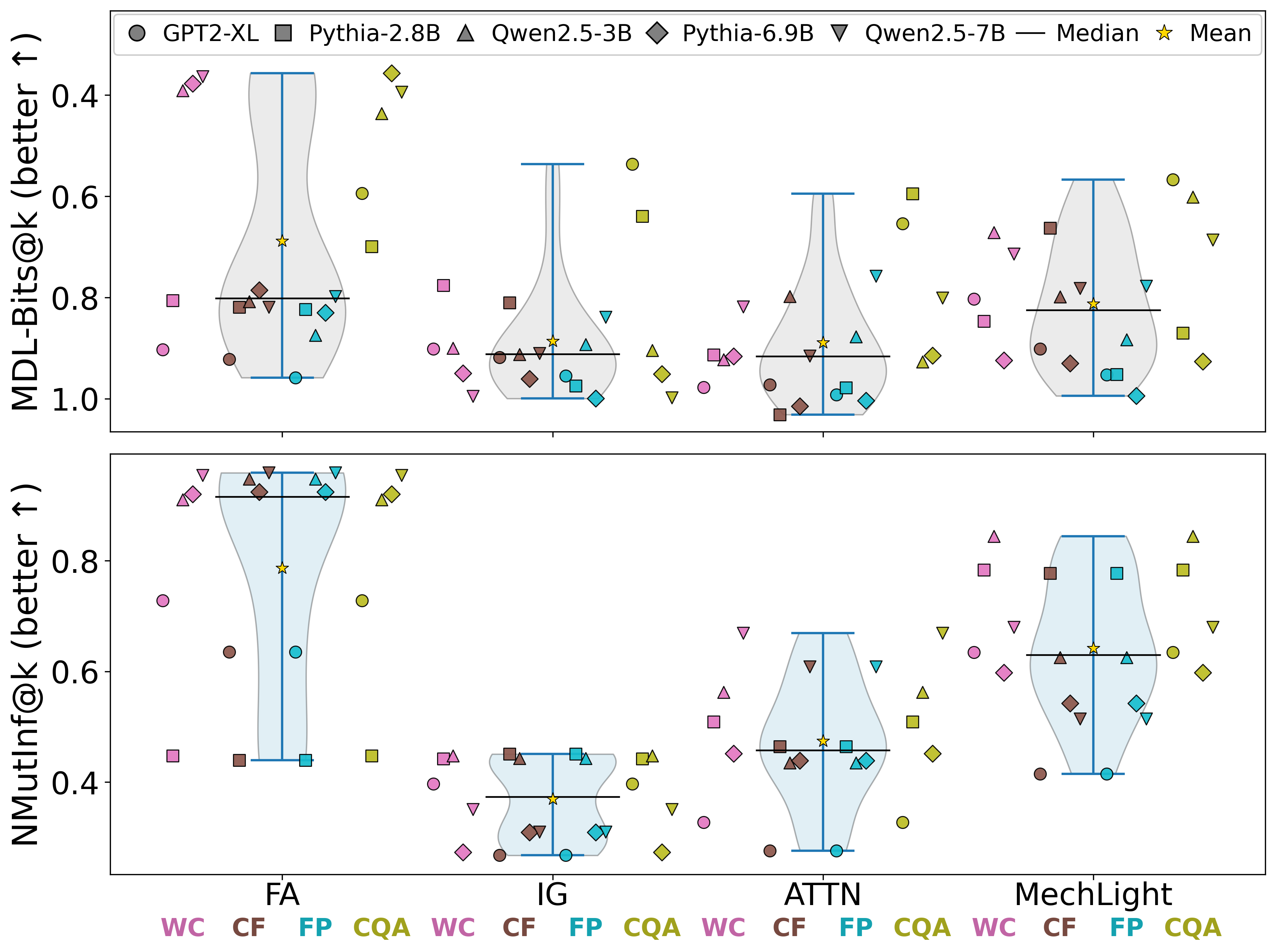}
    \caption{Conflicting Context}
    \label{hcu:fig:conflicting-topk-5-mutual-information-and-mdl-prob}
  \end{subfigure}
  \hfill
  \begin{subfigure}[t]{0.48\textwidth}
    \centering
    \includegraphics[width=\textwidth] 
    {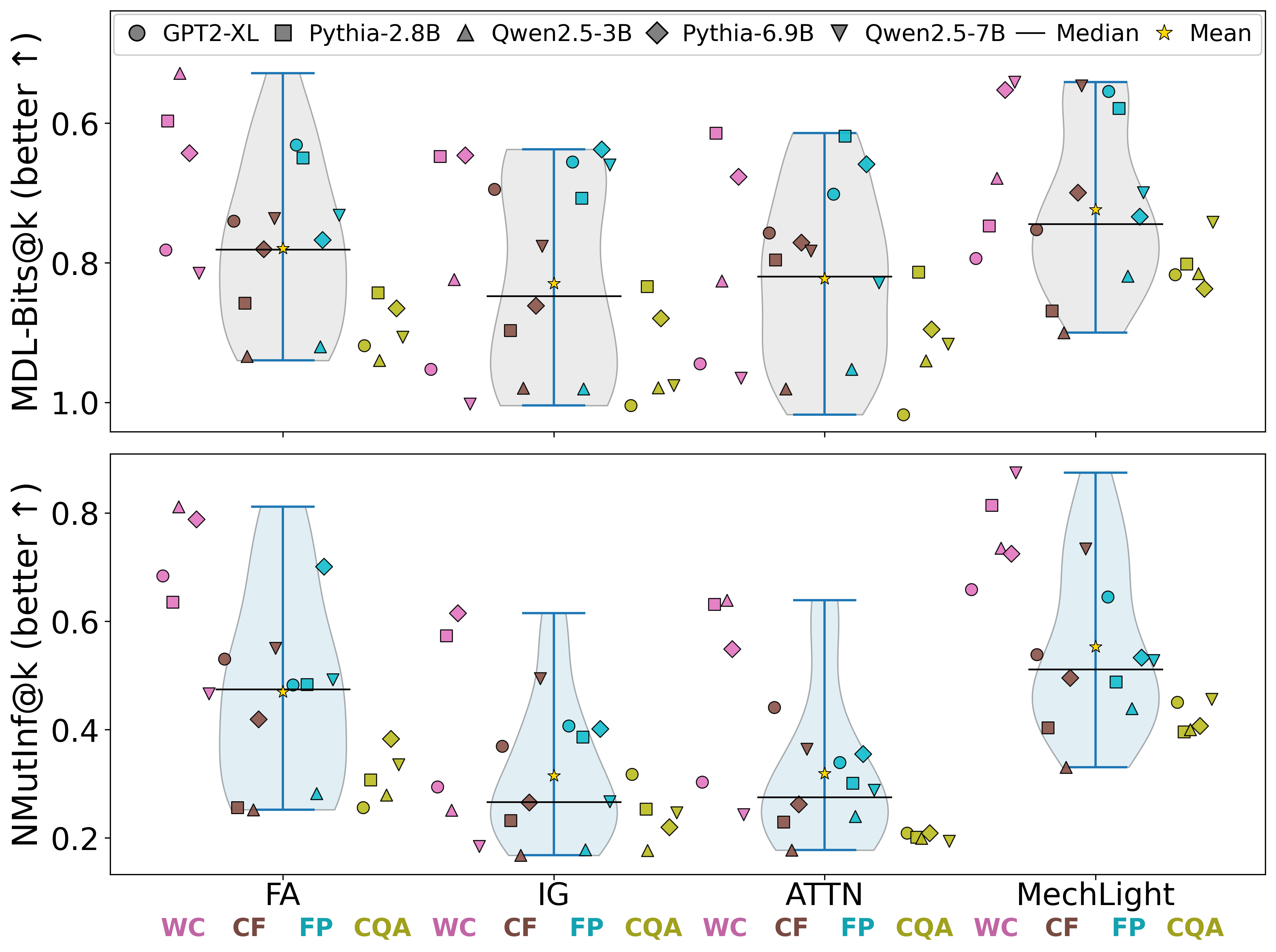}
    \caption{Irrelevant Context}
    \label{hcu:fig:irrelevant-topk-5-mutual-information-and-mdl-prob}
  \end{subfigure}
  \caption{ \MDLBitsk\ (top plot; y-axis is inverted; Eq.~\ref{hcu:eq:mdl}) and \NMutInfk (bottom plot; Eq.~\ref{hcu:eq:mutual-information}) for explanation simulatability in \textbf{Conflicting} and \textbf{Irrelevant} setups (\S\ref{hcu:sec:four-context-setups}). Both \MDLBitsk (smaller value) and \NMutInfk are better the higher, meaning the explanation importance scores are more representative of the model's answer selection, i.e., whether the answer is from context vs. from memory (\S\ref{hcu:sec:rq1-discussion}). The colors denote the datasets, and the marker shapes denote the models.
}
  \label{hcu:fig:topk-mutual-information-and-mdl-prob-conflicting-or-irrelevant}
\end{figure*}

\subsection{Does the explanation indicate whether the model consulted the supplied context knowledge?}
\label{hcu:sec:rq1-discussion}
\textbf{Document-level attribution}. Fig.~\ref{hcu:fig:rank-topk-margin-conflicting-or-irrelevant} presents the results on \DrankkGroup{} for single-context setups across models and datasets. Each marker is a model--dataset pair; values above $0$ mean that context tokens receive higher importance ranks in $D_C$ than in $D_M$ (i.e., the HE better distinguishes context vs.\ memory usage), and values near $0$ indicate little separability. From Fig.~\ref{hcu:fig:rank-topk-margin-conflicting-or-irrelevant}, we observe mostly positive, small \DrankkGroup, indicating the context tokens are indeed often ranked higher in the $D_{C}$ instances compared to the $D_{M}$ instances. In both setups, we find that \textit{\MI\ has the most cases with positive results across all datasets and models} with either the best (b) or second-best (a) \DrankkGroup. This is visible in both subplots as \MI's markers lying above $0$ for more model--dataset pairs than the other methods, whereas \IG\ and \Attn\ cluster close to $0$ and frequently have negative values. \textit{\FA\ often yields positive margin values but shows the \emph{largest variance} across model–dataset pairs}, i.e., indicating unstable performance. The high variance of \FA\ is reflected by the visibly wider spread of its points across model--dataset pairs (including both strong positives and negatives), compared to the tighter clustering for \MI.
Finally, \textit{\IG\ and \Attn\ typically yield margins close to zero, 
indicating limited ability to distinguish whether the model consulted the context or its parametric memory}. This is surprising as these methods score high on faithfulness evaluations (See Tab.~\ref{hcu:tab:faithfulness} in App.~\ref{hcu:app:additional_results}). Nevertheless, occlusion-based methods, such as \FA\, are often the most faithful HEs~\citep{deyoung-etal-2020-eraser}, which aligns with their performance in correctly attributing context utilisation. Comparing the Conflicting and Irrelevant setups, we find that HEs generally perform better in the latter. Additionally, the higher variability there also indicates increased dependence on the specific dataset and model.

\textbf{Simulatability.} We report simulatability results in Fig.~\ref{hcu:fig:topk-mutual-information-and-mdl-prob-conflicting-or-irrelevant} using \MDLBitsk\ (Eq.~\ref{hcu:eq:mdl}) and \NMutInfk\ (Eq.~\ref{hcu:eq:mutual-information}). \MDLBitsk\ is the average prequential code length (bits per instance) for predicting the behaviour label from the top-$k$ highlights (lower is better), while \NMutInfk\ measures the fraction of label entropy explained by the top-$k$ highlights (higher is better; e.g., $\NMutInfk{=}0.20$ corresponds to a $20\%$ uncertainty reduction). We invert the $y$-axis for \MDLBitsk\ so that higher values are better. In Fig.~\ref{hcu:fig:topk-mutual-information-and-mdl-prob-conflicting-or-irrelevant}, \textit{\MDLBitsk\ and \NMutInfk\ show similar trends with the \DrankkGroup\ results in Fig.~\ref{hcu:fig:rank-topk-margin-conflicting-or-irrelevant}.} In the Conflicting setup, \FA\ is typically the best but variable (using the top-$k$ explanation importance scores can reduce about 19.8\% uncertainty in model answer label prediction, for half of the model-dataset cases), and \MI\ is second best (about 16.5\% uncertainty reduction). Following are \IG\ and \Attn, leaving about 91\% of label uncertainty. In the Irrelevant setup, all methods improve on both metrics, with \MI\ showing better performance than \FA{}. This again indicates that explanations can more effectively reveal context usage when the context is off-topic. As expected, \NMutInfk\ and \MDLBitsk\ show similar trends.

Overall, \textbf{\MI\ shows the best performance regarding whether the model relied on CK or PK}, followed by \FA{}, but with considerable variability in performance.
\IG\ and \Attn\ provide little value for this purpose. We hypothesise that \FA\ exhibits high variance due to its discrete perturbation nature: token replacement can introduce distribution shift, and small perturbations can trigger nonlinear interactions that change the answer in unstable ways. Moreover, our implementation attributes only the \emph{first generated token}, which can be a noisy objective for multi-token answers; when many tokens have near-tied effects, the top-$k$ ranking becomes sensitive to small score fluctuations. These factors jointly make \FA\ both computationally expensive and less stable for token-level context attribution in long-context regimes.

\begin{figure*}[ht]
  \centering
  \begin{subfigure}[t]{0.48\textwidth}
    \centering
    \includegraphics[width=\textwidth]{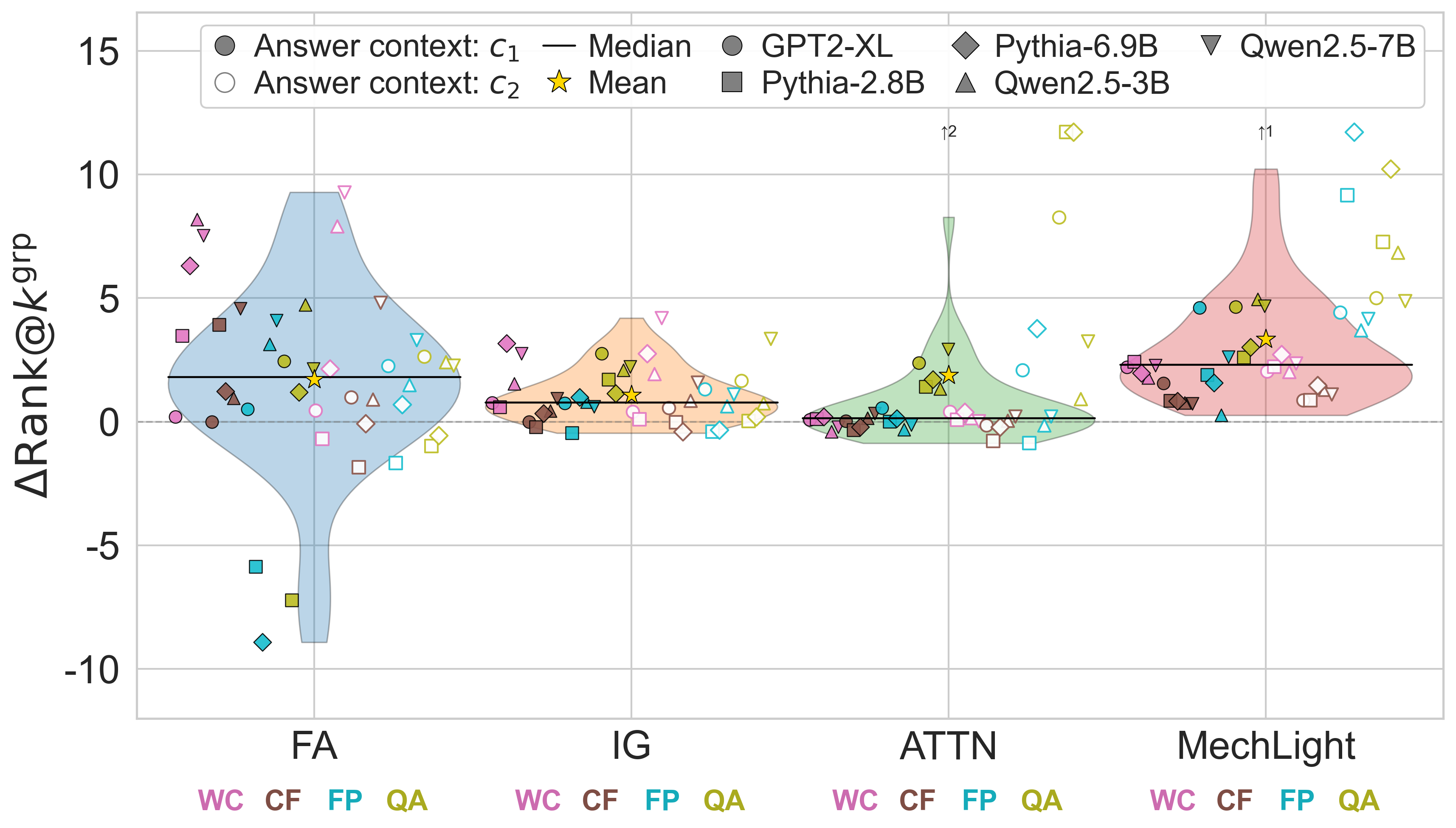}
    \caption{Double-Conflicting: Two Conflicting Contexts}
    \label{hcu:fig:conflicting-1-and-2-topk-5-rank-datasets-margin-only}
  \end{subfigure}
  \hfill
  \begin{subfigure}[t]{0.48\textwidth}
    \centering
    \includegraphics[width=\textwidth]
    {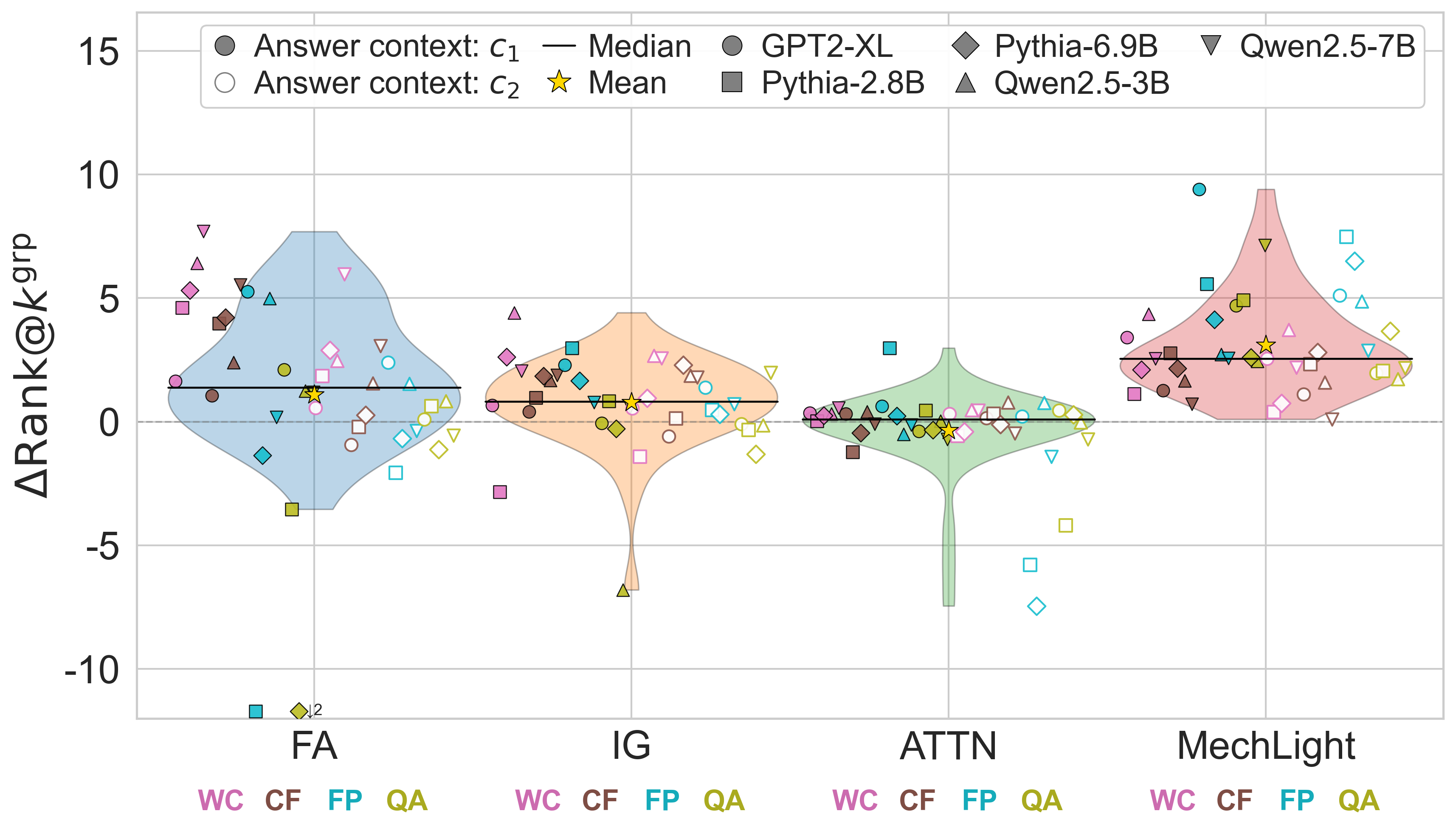}
    \caption{Mixed: One Irrelevant and One Conflicting Context}
    \label{hcu:fig:irrelevant-and-conflicting-topk-5-rank-datasets-margin-only}
  \end{subfigure}
  \caption{  
  \DrankkGroup\ (Eq.~\ref{hcu:eq:drankk-cross-instance-group}) -- average margins for the rank of context $c_1$ and $c_2$ between two instance groups $D_{c_1}$ and $D_{c_2}$  in the \textbf{Double-Conflicting} and \textbf{Mixed} setup (\S\ref{hcu:input-regime-2}). Higher \DrankkGroup is better, and a positive number means that the explanation can correctly indicate which context piece is used for the answer (\S\ref{hcu:sec:rq2-discussion}). The values are obtained by comparing the average importance scores of either the first context piece  ($c_1$, filled shapes at the left) or the second context piece ($c_2$, hollowed shapes at the right) between the two instance groups. The colors denote the datasets, and the marker shapes denote the models.
  }
  \label{hcu:fig:rank-topk-margin-double-conflicting-and-mixed}
\end{figure*}

\subsection{Does the explanation show \textit{which} of two context documents the model used?}
\label{hcu:sec:rq2-discussion}

\begin{figure*}[ht]
  \centering
  \begin{subfigure}[t]{0.48\textwidth}
    \centering
    \includegraphics[width=\textwidth]
    {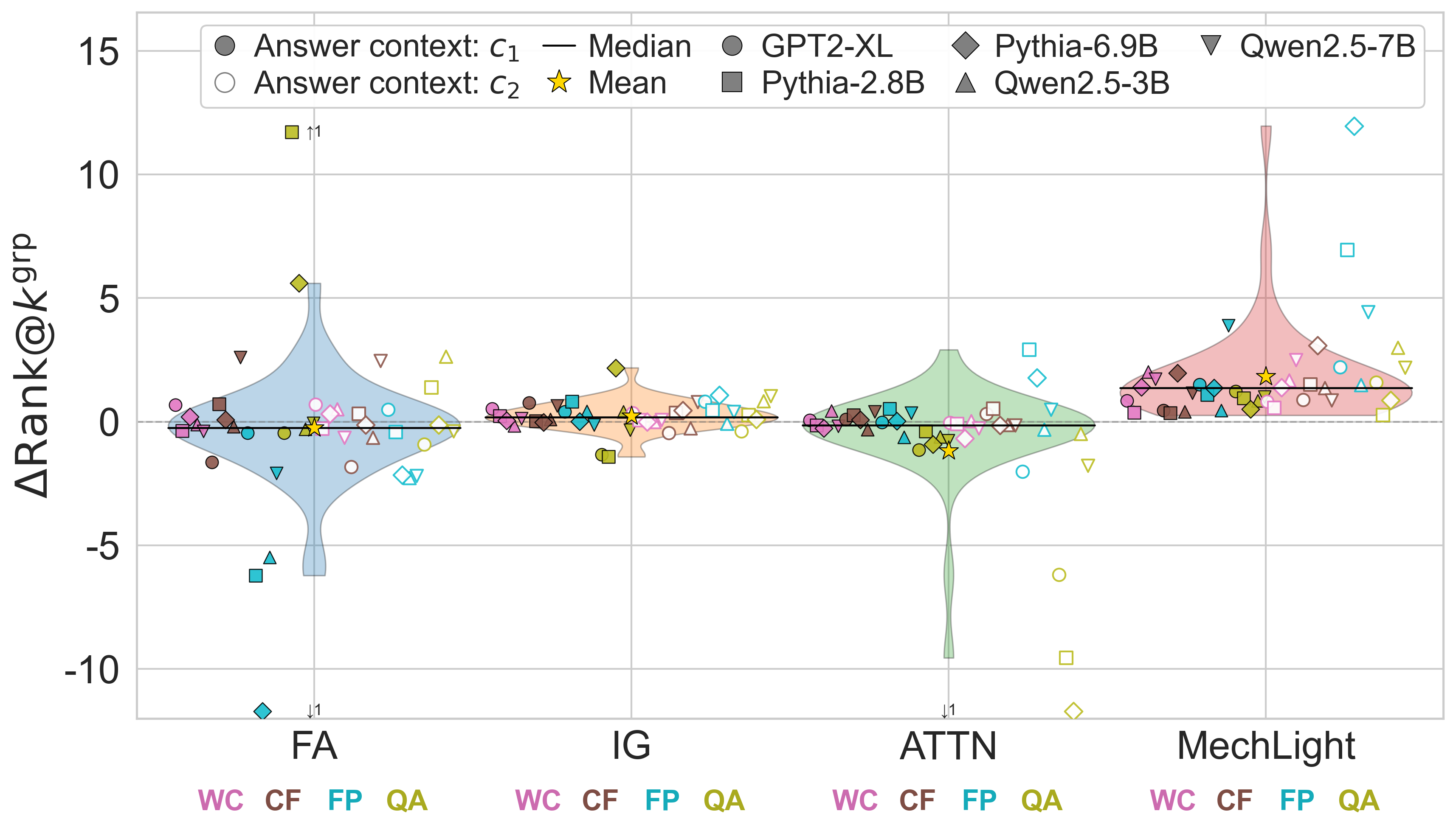}
    \caption{Double-Conflicting-Swap}
    \label{hcu:fig:conflicting-2-and-1-topk-5-rank-datasets-margin-only}
  \end{subfigure}
  \hfill
  \begin{subfigure}[t]{0.48\textwidth}
    \centering
    \includegraphics[width=\textwidth]
{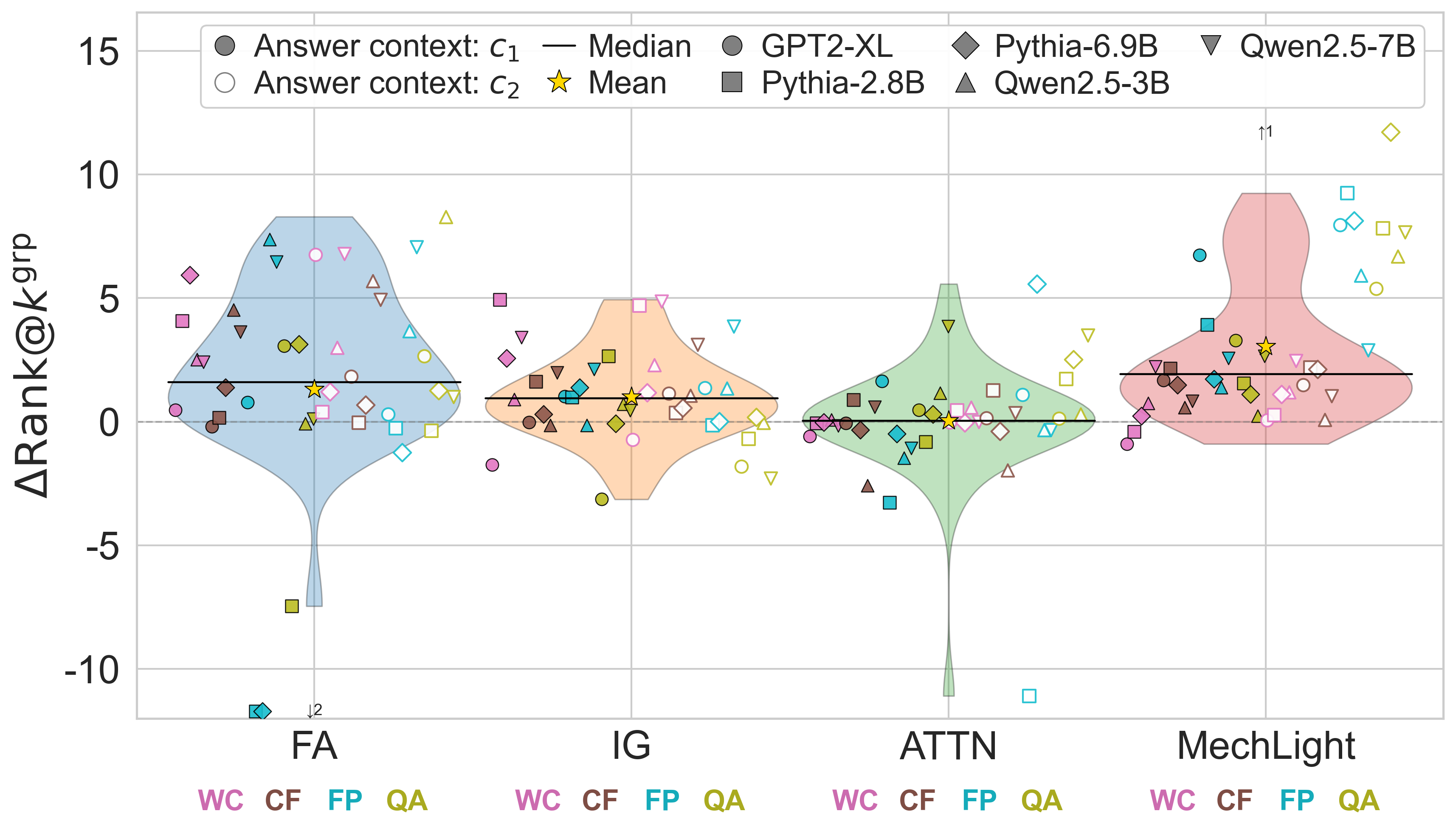}
    \caption{Mixed-Swap}
    \label{hcu:fig:conflicting-and-irrelevant-topk-5-rank-datasets-margin-only}
  \end{subfigure}
      \caption{\DrankkGroup\ (Eq.~\ref{hcu:eq:drankk-cross-instance-group}) -- average margins for the rank of context $c_1$ and $c_2$ between two instance groups $D_{c_1}$ and $D_{c_2}$  in the \textbf{Double-Conflicting} and \textbf{Mixed} setup \textbf{after swapping} the position of two context pieces (\S\ref{hcu:input-regime-2}). Higher \DrankkGroup is better, and a positive number means that the explanation can correctly indicate which context piece is used for the answer (\S\ref{hcu:sec:rq2-discussion}). The values are obtained by comparing the average importance scores of either the first context piece  ($c_1$, filled shapes at the left) or the second context piece ($c_2$, hollowed shapes at the right) between the two instance groups. The colors denote the datasets, and the marker shapes denote the models.
  }
  \label{hcu:fig:rank-topk-margin-double-conflicting-and-mixed-after-swap}
\end{figure*}

\textbf{Document-level attribution across groups}. 
Fig.~\ref{hcu:fig:rank-topk-margin-double-conflicting-and-mixed} reports \DrankkGroup\ for dual-context setups. For each attribution method and model--dataset pair, we show two margins: filled markers (left half) correspond to instances where the model answers from $c_1$ (comparing $D_{C_1}$ vs.\ $D_{C_2}$), and hollow markers (right half) correspond to answers from $c_2$. Positive margins mean tokens in the utilised passage are ranked higher than those in the unused passage; thus, a good explanation method should have positive values in both halves.
We observe that \textit{\MI\ is the best in both setups, by consistently showing positive margins}, meaning that the answer-context tokens from the used context are actually ranked higher than tokens from the unused context. 
\textit{\FA\ is second-best overall, but often shows the largest negatives on long-context datasets (Fakepedia, ConflictQA) when the model answers from $c_1$ (filled markers)}, suggesting that \FA\ tends to prefer the later passage in concatenated prompts in these cases.
Following are \IG\ and \Attn\, with most margins close to zero in both setups, indicating they often fail to indicate which document the model selects the answer from. Comparing setups, the results are similar; \FA\ and \IG\ show slightly larger margins in Double-Conflicting and more variability in Mixed, with worse results on Fakepedia and ConflictQA, likely reflecting their sensitivity to context length and difficulty with long mixed contexts. The fact that \FA\ and \MI\ are better than \IG\ and \Attn\ again confirms the potential link between the faithfulness and the explanation utility (See faithfulness in Tab.~\ref{hcu:tab:faithfulness} in App.~\ref{hcu:app:additional_results}). Trends persist after swapping the two context pieces, in Double-Conflicting-Swap and Mixed-Swap (see Fig.~\ref{hcu:fig:rank-topk-margin-double-conflicting-and-mixed-after-swap} in App.~\ref{hcu:app:additional_results}).

\begin{figure*}[ht]
  \centering
   \begin{subfigure}[t]{0.48\textwidth}
    \centering
    \includegraphics[width=\textwidth]
    {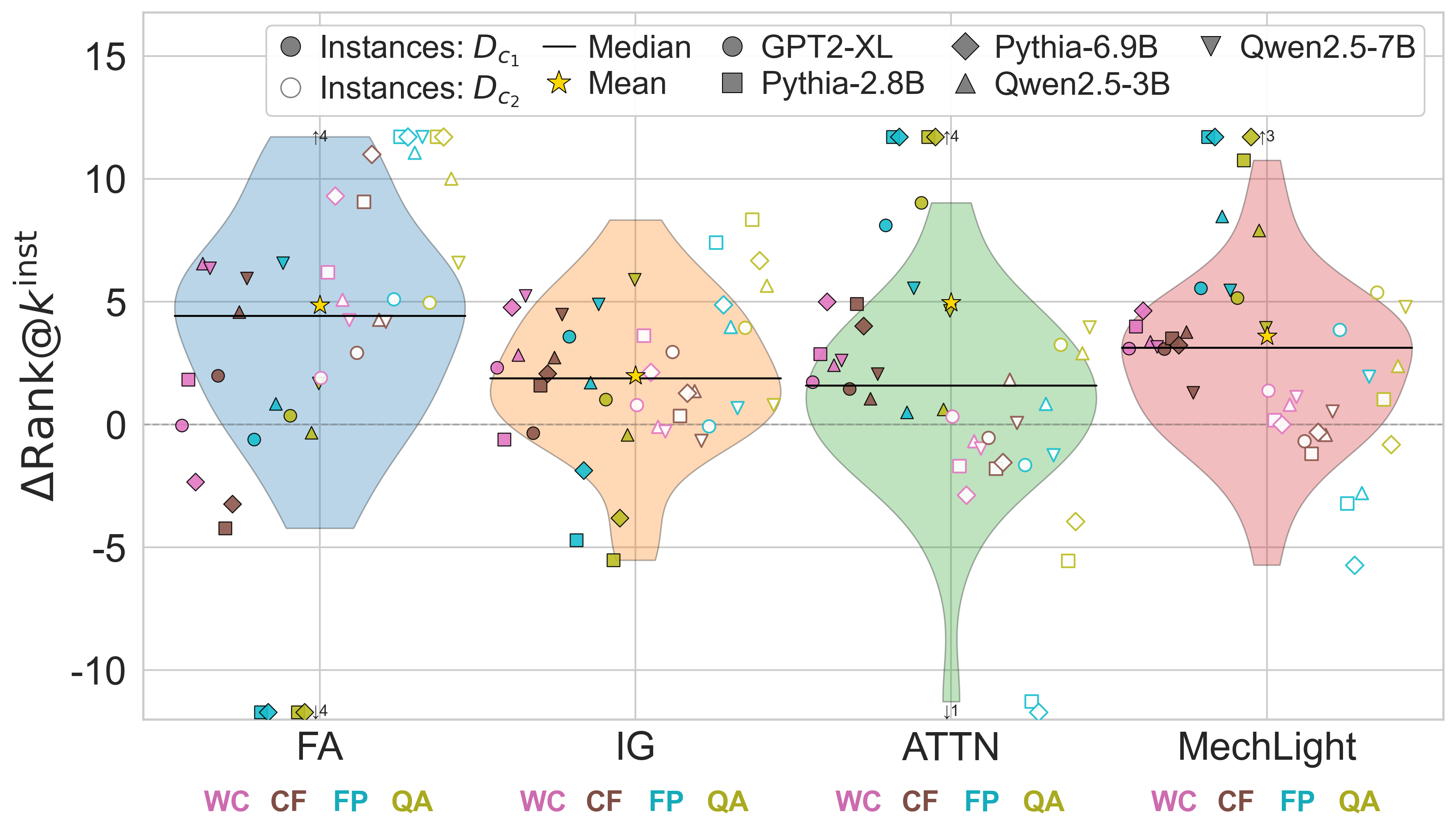}
    \caption{Double-Conflicting: Two Conflicting Contexts}
    \label{hcu:fig:conflicting-1-and-2-topk-5-rank-datasets-within-group-margin-only}
  \end{subfigure}
  \hfill
  \begin{subfigure}[t]{0.48\textwidth}
    \centering
    \includegraphics[width=\textwidth]
    {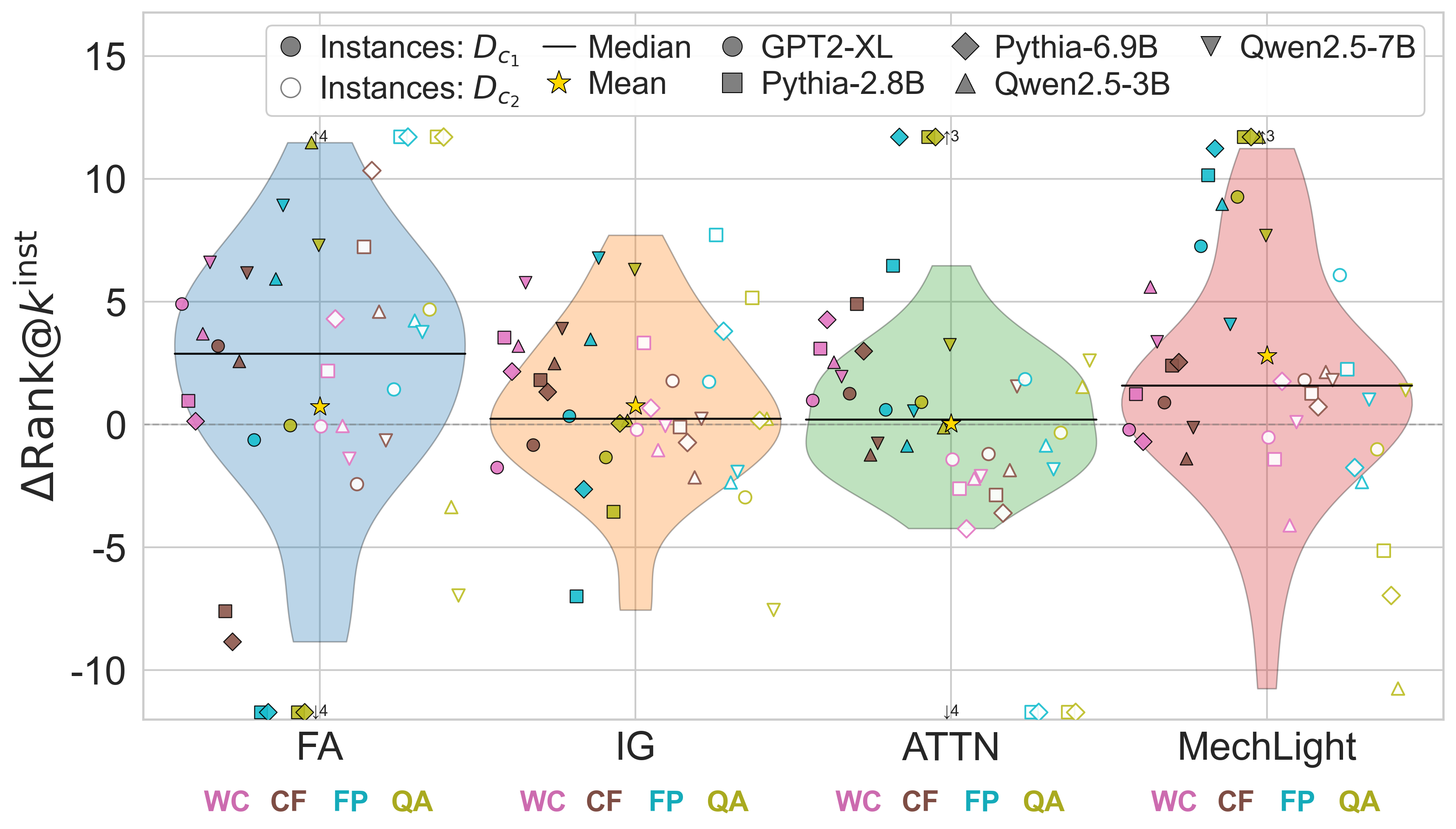}
    \caption{Mixed: One Irrelevant and One Conflicting Context}
    \label{hcu:fig:irrelevant-and-conflicting-topk-5-rank-datasets-within-group-margin-only}
  \end{subfigure}
  \caption{\DrankkInst\ (Eq.~\ref{hcu:eq:drankk-within-instance})  -- average \emph{within-instance-group} margins between the rank of the answer context piece and the other context piece in the \textbf{Double-Conflicting} and \textbf{Mixed} setup (\S\ref{hcu:input-regime-2}).  Higher \DrankkInst\ is better, and a positive number means that the explanation can correctly indicate which context piece is used for the answer (\S\ref{hcu:sec:rq2-discussion}). The values are obtained by comparing importance scores of the answer-piece context and the other piece within either the instance group answering from the first context piece  ($D_1$, filled shapes at the left) or that answering from the second context piece ($D_2$, hollowed shapes at the right). The colors denote the datasets, and the marker shapes denote the models.}
  \label{hcu:fig:rank-topk-margin-within-instance-double-conflicting-and-mixed}
\end{figure*}

\begin{figure*}[ht]
  \centering
   \begin{subfigure}[t]{0.48\textwidth}
    \centering
    \includegraphics[width=\textwidth]
    {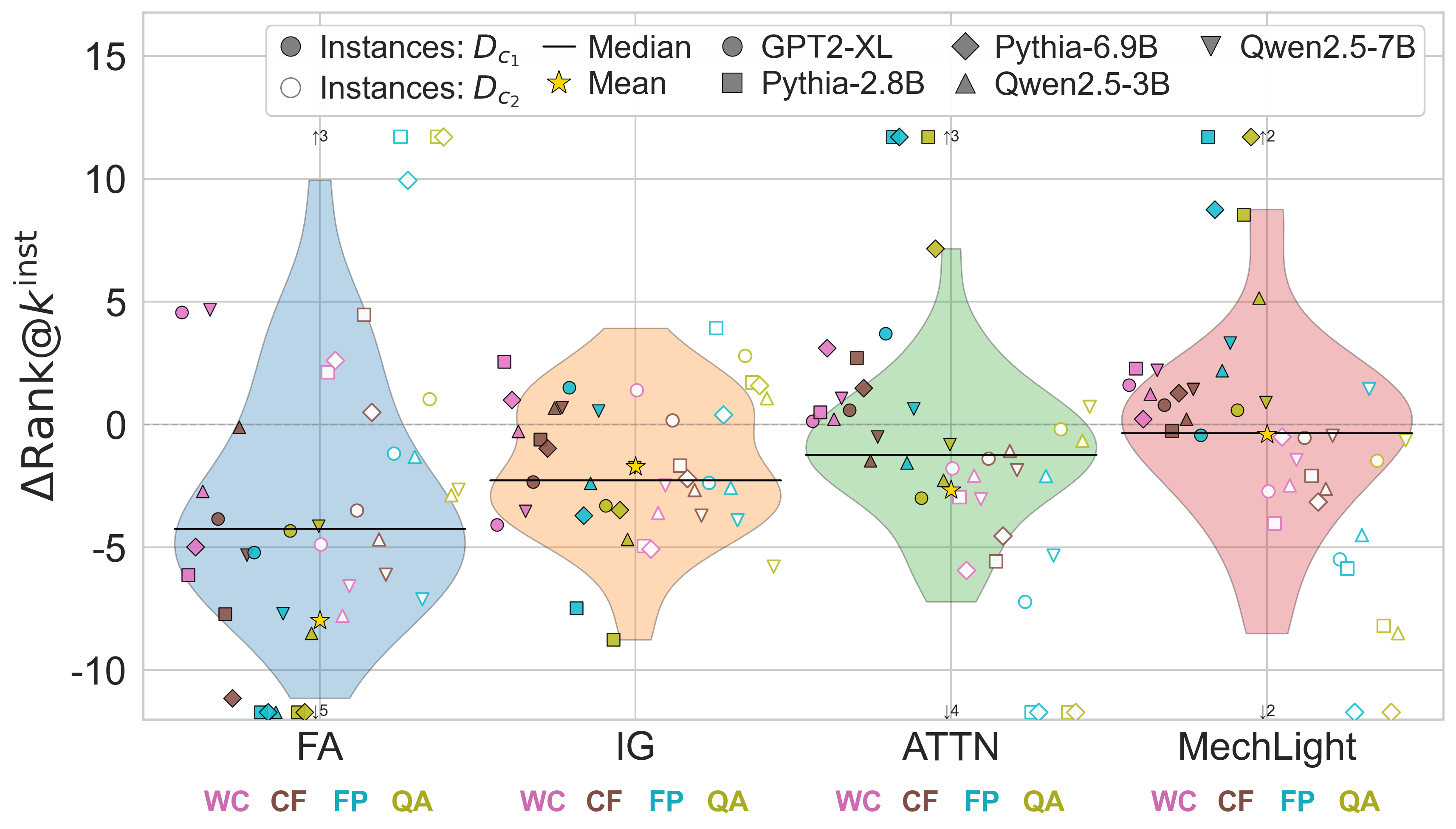}
    \caption{Double-Conflicting-Swap}
    \label{hcu:fig:conflicting-2-and-1-topk-5-rank-datasets-within-group-margin-only}
  \end{subfigure}
  \hfill
  \begin{subfigure}[t]{0.48\textwidth}
    \centering
    \includegraphics[width=\textwidth]{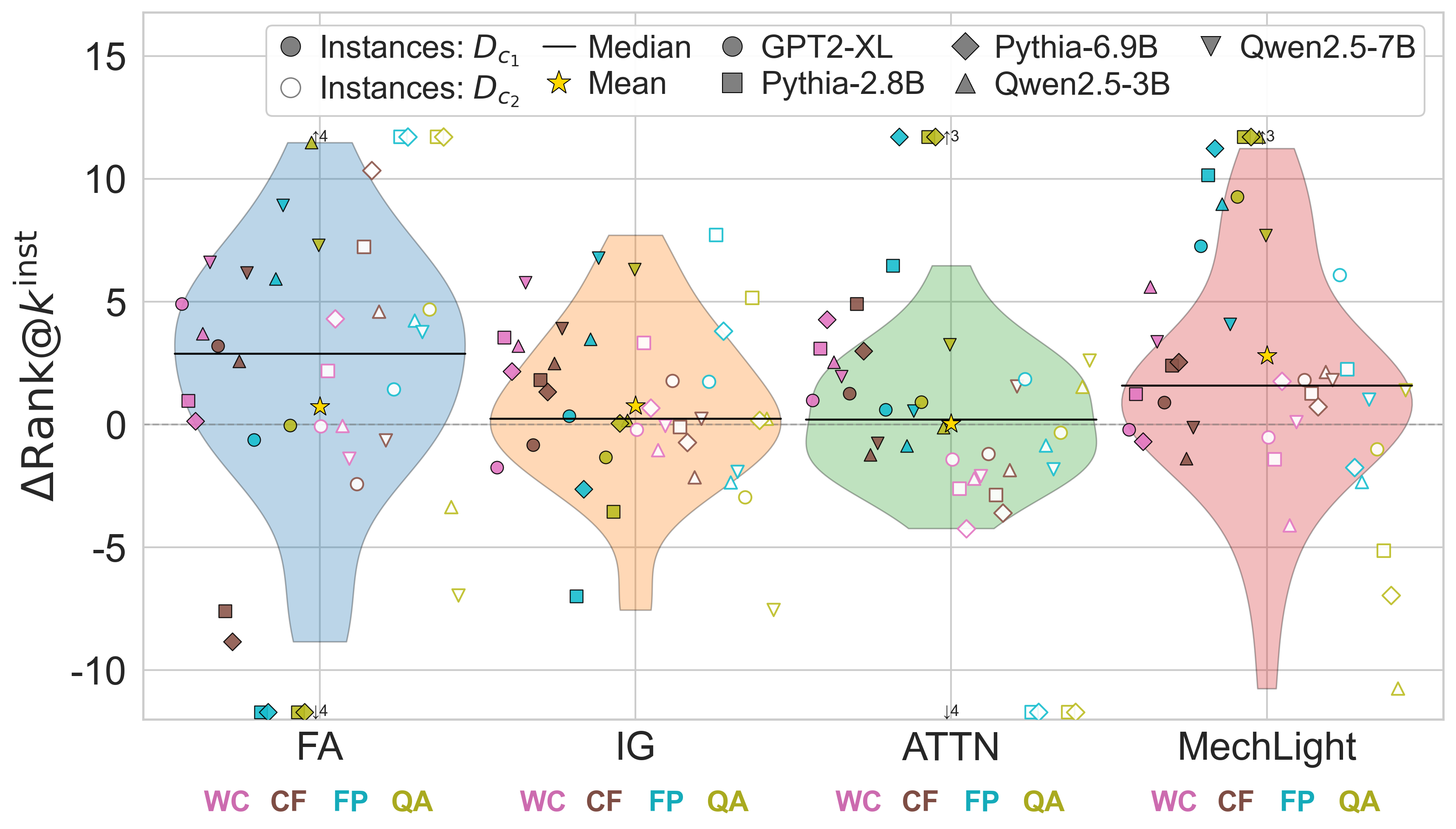}
    \caption{Mixed-Swap}
    \label{hcu:fig:conflicting-and-irrelevant-topk-5-rank-datasets-within-group-margin-only}
  \end{subfigure}
    \caption{\DrankkInst\ (Eq.~\ref{hcu:eq:drankk-within-instance})  -- average \emph{within-instance-group} margins between the rank of the answer context piece and the other context piece in the \textbf{Double-Conflicting} and \textbf{Mixed} setup after \textbf{swapping} the position of the two context pieces (\S\ref{hcu:input-regime-2}).  Higher \DrankkInst\ is better, and a positive number means that the explanation can correctly indicate which context piece is used for the answer (\S\ref{hcu:sec:rq2-discussion}). The values are obtained by comparing importance scores of the answer-piece context and the other piece within either the instance group answering from the first context piece  ($D_1$, filled shapes at the left) or that answering from the second context piece ($D_2$, hollowed shapes at the right). The colors denote the datasets, and the marker shapes denote the models.}
  \label{hcu:fig:rank-topk-margin-within-instance-double-conflicting-and-mixed-after-swap}
\end{figure*}

\textbf{Document-level attribution across instances}. We now compare per instance the top-$k$ rank margin between tokens in the utilised vs.\ unused document. Concretely, \DrankkInst\ (Fig.~\ref{hcu:fig:rank-topk-margin-within-instance-double-conflicting-and-mixed}) is positive when, within the same instance, tokens in the utilised passage outrank tokens in the unused passage according to the HE scores. As shown in Fig.~\ref{hcu:fig:rank-topk-margin-within-instance-double-conflicting-and-mixed}, no HE shows positive margins for all cases, especially on long contexts (Fakepedia and ConflictQA), implying \textit{the HEs often cannot indicate which document the answer is selected from, especially when the contexts are relatively long}. \MI\ is strongest overall (best in (b), second-best in (a)) with positive rank margins in most cases. \FA\ follows, \IG\ and \Attn\ exhibit minor positive margins. We also find that all HEs exhibit positional bias: margins turn negative when the answer comes from the second (\MI, \Attn, which are based on the attention head mechanism) or first (\FA, \IG) piece in long contexts. The same trends hold in both setups and persist after changing piece order (Fig.~\ref{hcu:fig:rank-topk-margin-within-instance-double-conflicting-and-mixed-after-swap}), confirming the content-independent positional bias. We hypothesize that this opposing positional bias could either stem from the model structure or the nature of the attribution method (See discussion in \S\ref{hcu:sec:limitation} Limitation).

\begin{figure*}[ht]
  \centering
  \begin{subfigure}[t]{0.48\textwidth}
    \centering
    \includegraphics[width=\textwidth]
    {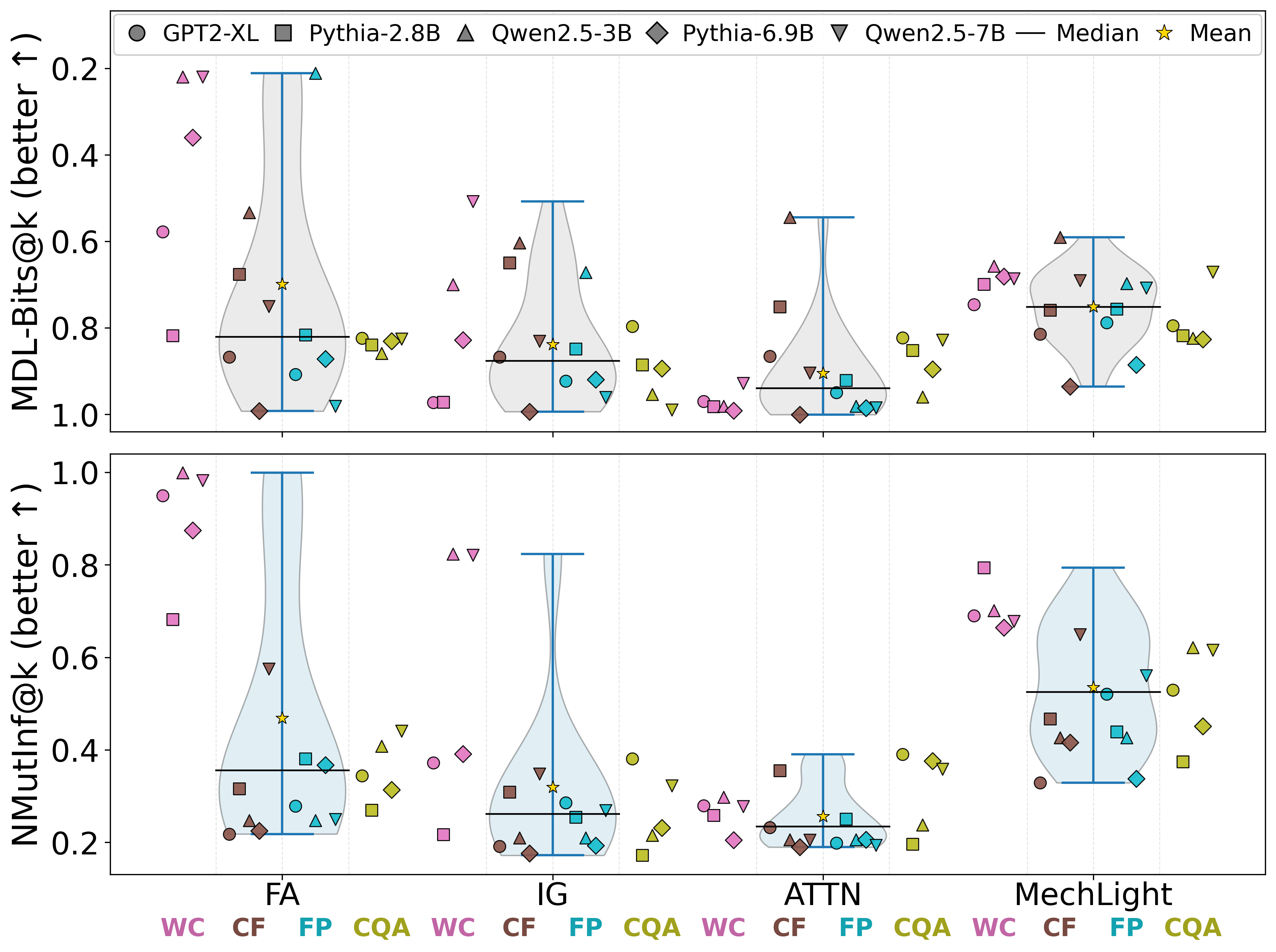}
    \caption{Double-Conflicting: Two Conflicting Contexts}
    \label{hcu:fig:conflicting-1-and-2-topk-5-mutual-information-and-mdl-prob}
  \end{subfigure}
  \hfill
  \begin{subfigure}[t]{0.48\textwidth}
    \centering
    \includegraphics[width=\textwidth]
    {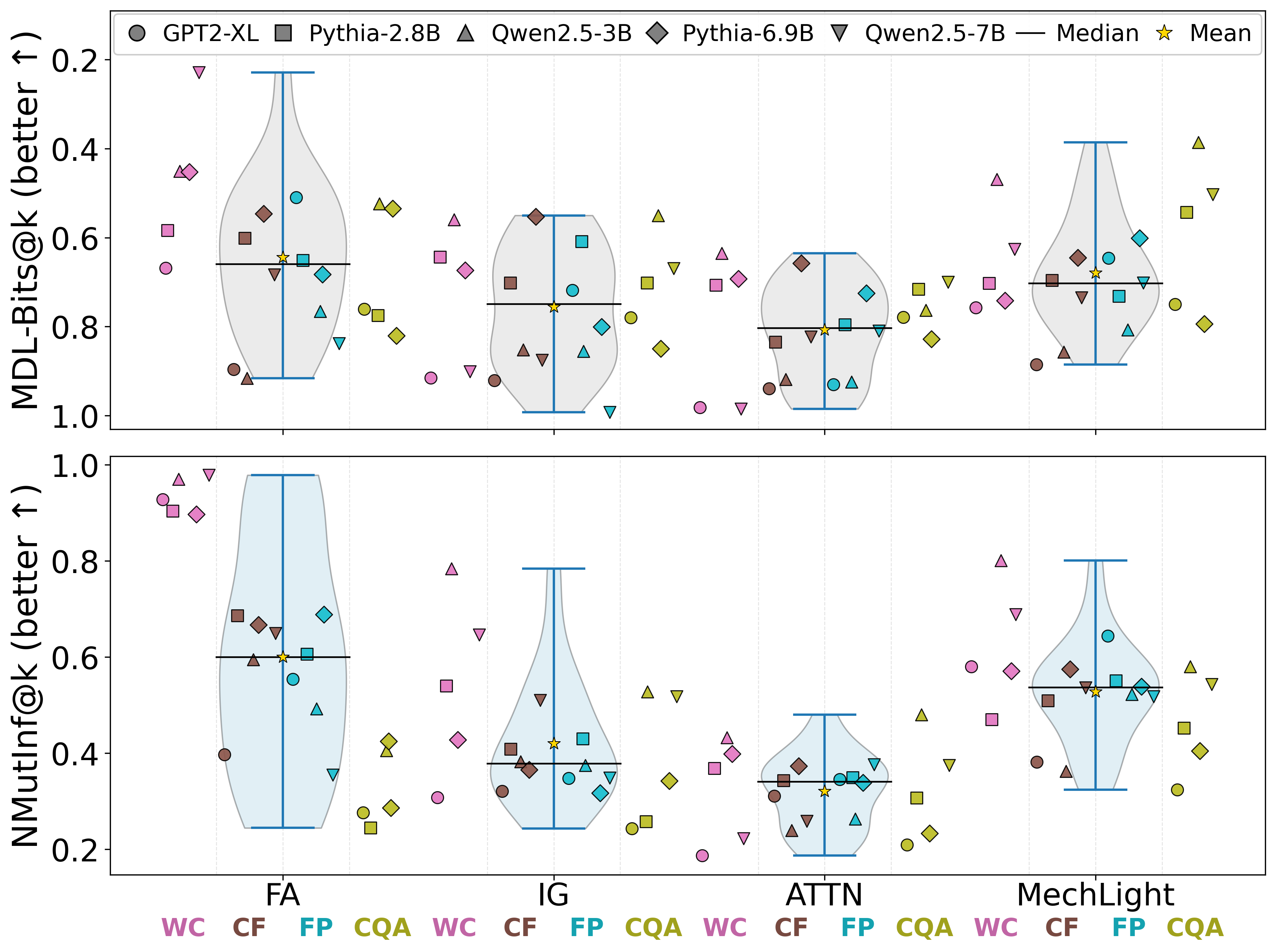}
    \caption{Mixed: One Irrelevant and One Conflicting Context}
    \label{hcu:fig:irrelevant-and-conflicting-topk-5-mutual-information-and-mdl-prob}
  \end{subfigure}
\caption{\MDLBitsk\ (top; y-axis is inverted; Eq.~\ref{hcu:eq:mdl}) and \NMutInfk\ (bottom; Eq.~\ref{hcu:eq:mutual-information}) in \textbf{Double-Conflicting} and \textbf{Mixed} setups (\S\ref{hcu:sec:four-context-setups}). Both \MDLBitsk (smaller value) and \NMutInfk are higher the better, indicating that the explanation importance scores are better at revealing the model's answer selection across the two context passages, i.e., from the first vs from the second piece. (\S\ref{hcu:sec:rq2-discussion}). The colors denote the datasets, and the marker shapes denote the models.
  }
    \label{hcu:fig:topk-mutual-information-and-mdl-prob-double-conflicting-or-mixed}
\end{figure*}

\begin{figure*}[ht]
  \centering
  \begin{subfigure}[t]{0.48\textwidth}
    \centering
    \includegraphics[width=\textwidth]
    {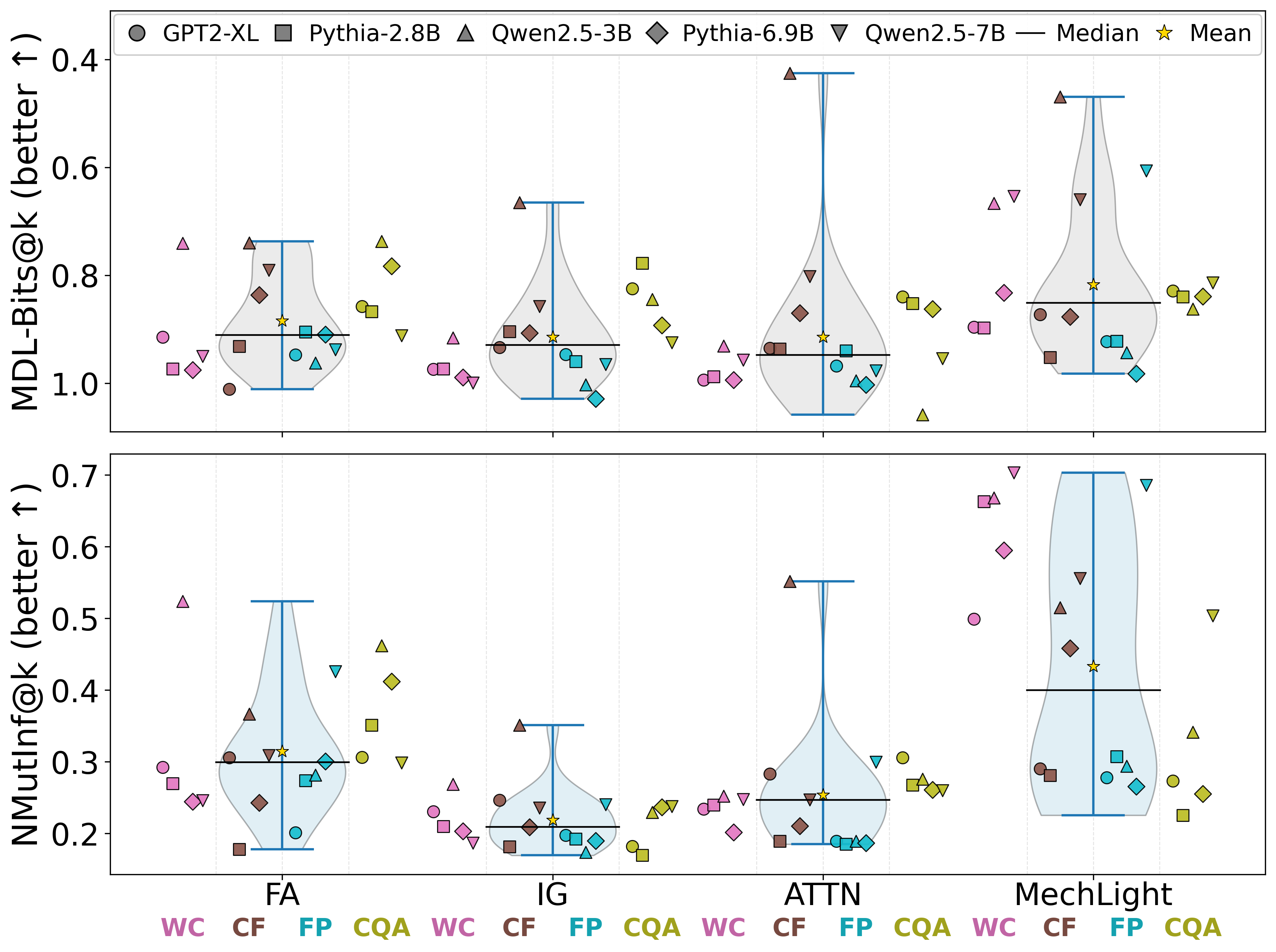}
    \caption{Double-Conflicting-Swap}
    \label{hcu:fig:conflicting-2-and-1-topk-5-mutual-information-and-mdl-prob}
  \end{subfigure}
  \hfill
  \begin{subfigure}[t]{0.48\textwidth}
    \centering
    \includegraphics[width=\textwidth]
{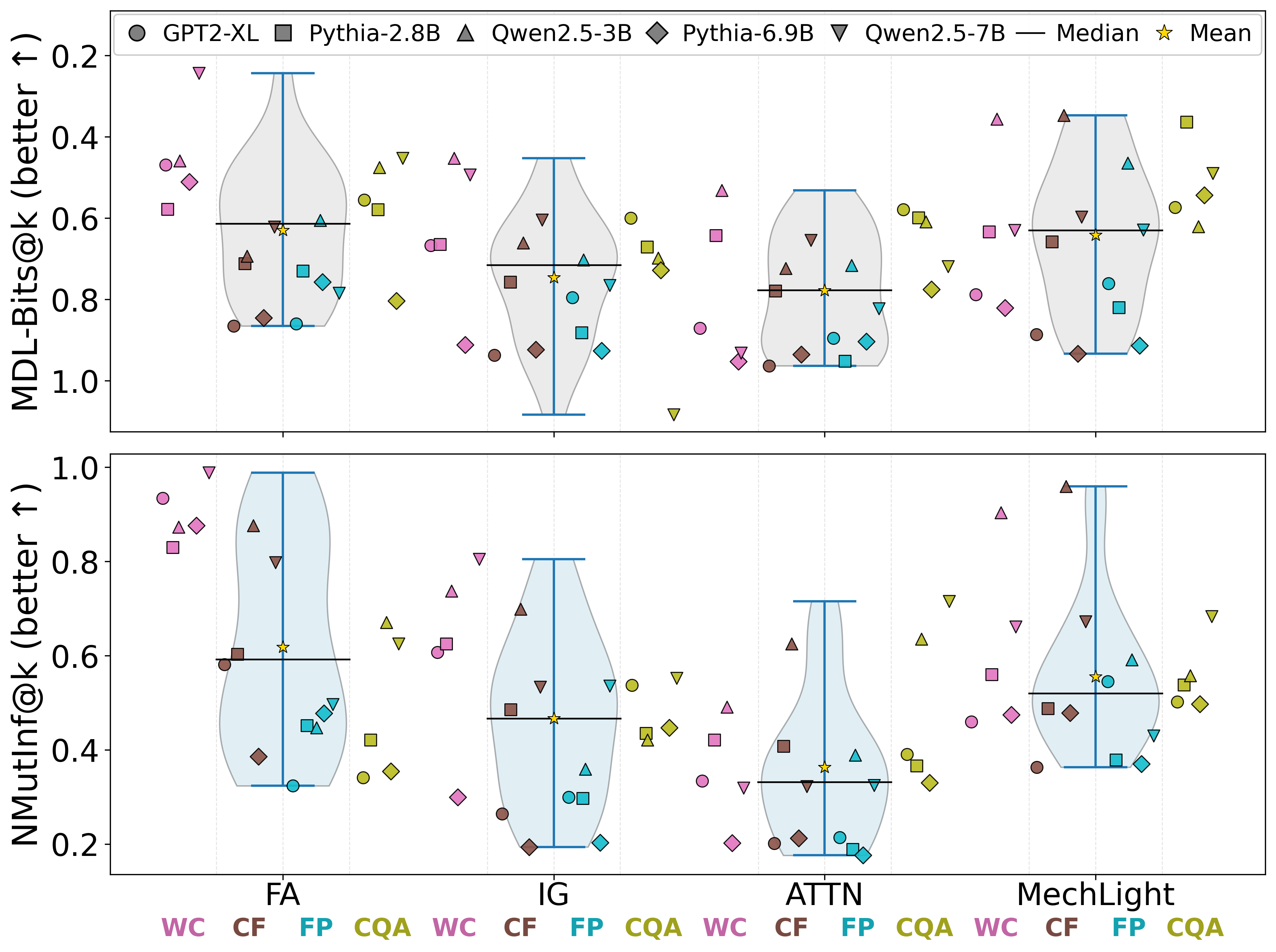}
    \caption{Mixed-Swap}
    \label{hcu:fig:conflicting-and-irrelevant-topk-5-mutual-information-and-mdl-prob}
  \end{subfigure}
  \caption{\MDLBitsk\ (top; y-axis is inverted; Eq.~\ref{hcu:eq:mdl}) and \NMutInfk\ (bottom; Eq.~\ref{hcu:eq:mutual-information}) in \textbf{Double-Conflicting} and \textbf{Mixed} setups after \textbf{swapping} the position of the two context pieces (\S\ref{hcu:sec:four-context-setups}). Both \MDLBitsk (smaller value) and \NMutInfk are higher the better, indicating that the explanation importance scores are better at revealing the model's answer selection across the two context passages, i.e., from the first vs from the second piece. (\S\ref{hcu:sec:rq2-discussion}). The colors denote the datasets, and the marker shapes denote the models.
  }
  \label{hcu:fig:topk-mutual-information-and-mdl-prob-double-conflicting-or-mixed-swapped}
\end{figure*}

\textbf{Simulatability}. In  Fig.~\ref{hcu:fig:topk-mutual-information-and-mdl-prob-double-conflicting-or-mixed}, \MDLBitsk\ and \NMutInfk\ support the document-level attribution evaluation across groups. \textit{\MI\ is the best overall}, leading to $24.9\%$ uncertainty reduction on the label prediction given the top-$k$ highlights. \textit{Following is \FA{}}, which removes about $17.9\%$ of the label prediction uncertainty, \textit{but again with a variable performance}. 
\IG\ and \Attn\ show worse performance 
leaving most label prediction uncertainty.
Comparing the two input regimes, Double-Conflicting and Mixed, the findings are overall consistent and persist after position swapping of the two contexts (See Fig.~\ref{hcu:fig:topk-mutual-information-and-mdl-prob-double-conflicting-or-mixed-swapped} in App.~\ref{hcu:app:additional_results})



\begin{figure*}[ht]
  \centering
  \begin{subfigure}[t]{0.48\textwidth}
    \centering
    \includegraphics[width=\textwidth]
    {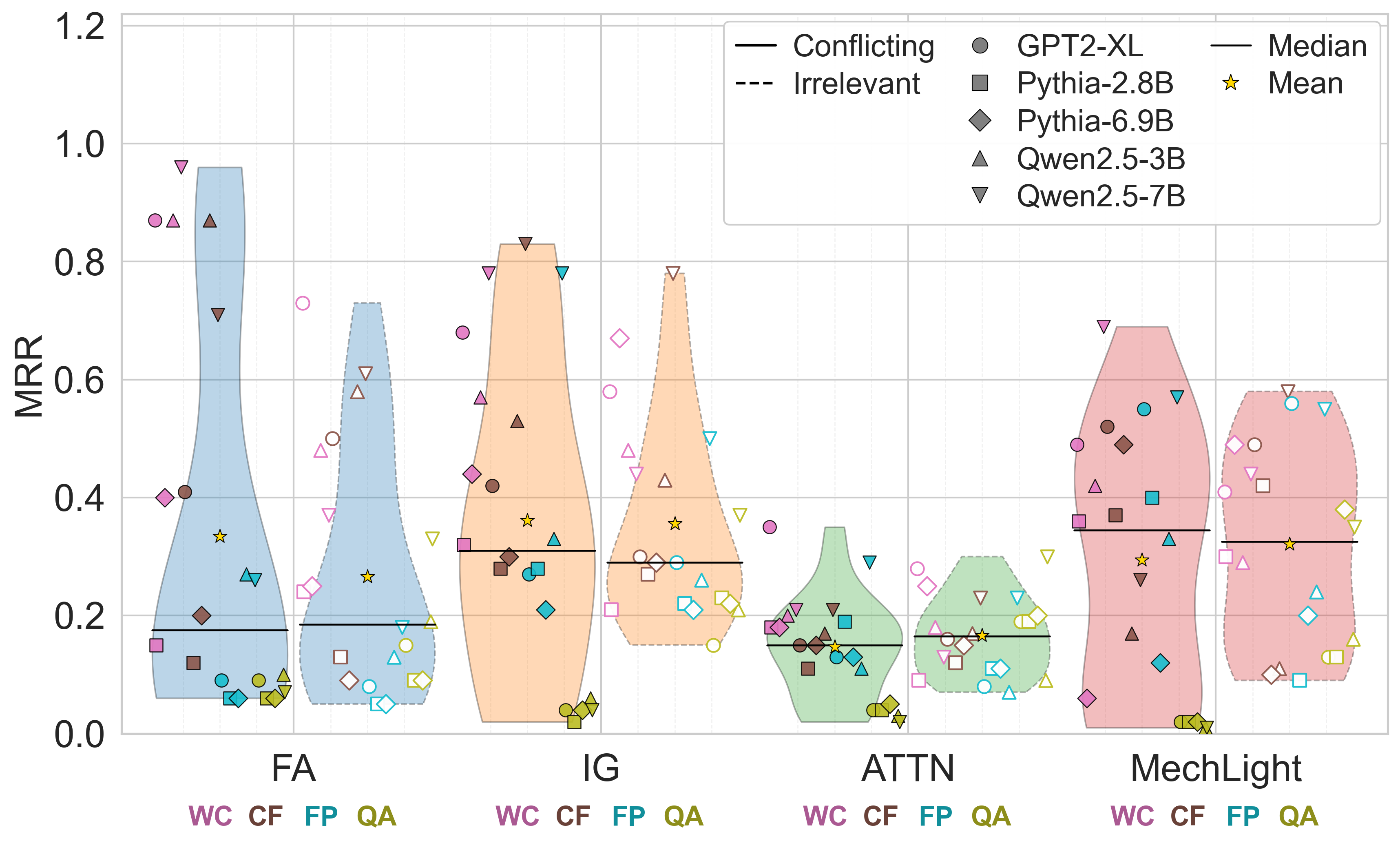}
    \caption{Conflicting \& Irrelevant}
    \label{hcu:fig:answer-token-mrr-conflicting-or-irrelevant}
  \end{subfigure}
  \hfill
  \begin{subfigure}[t]{0.48\textwidth}
    \centering
    \includegraphics[width=\textwidth]{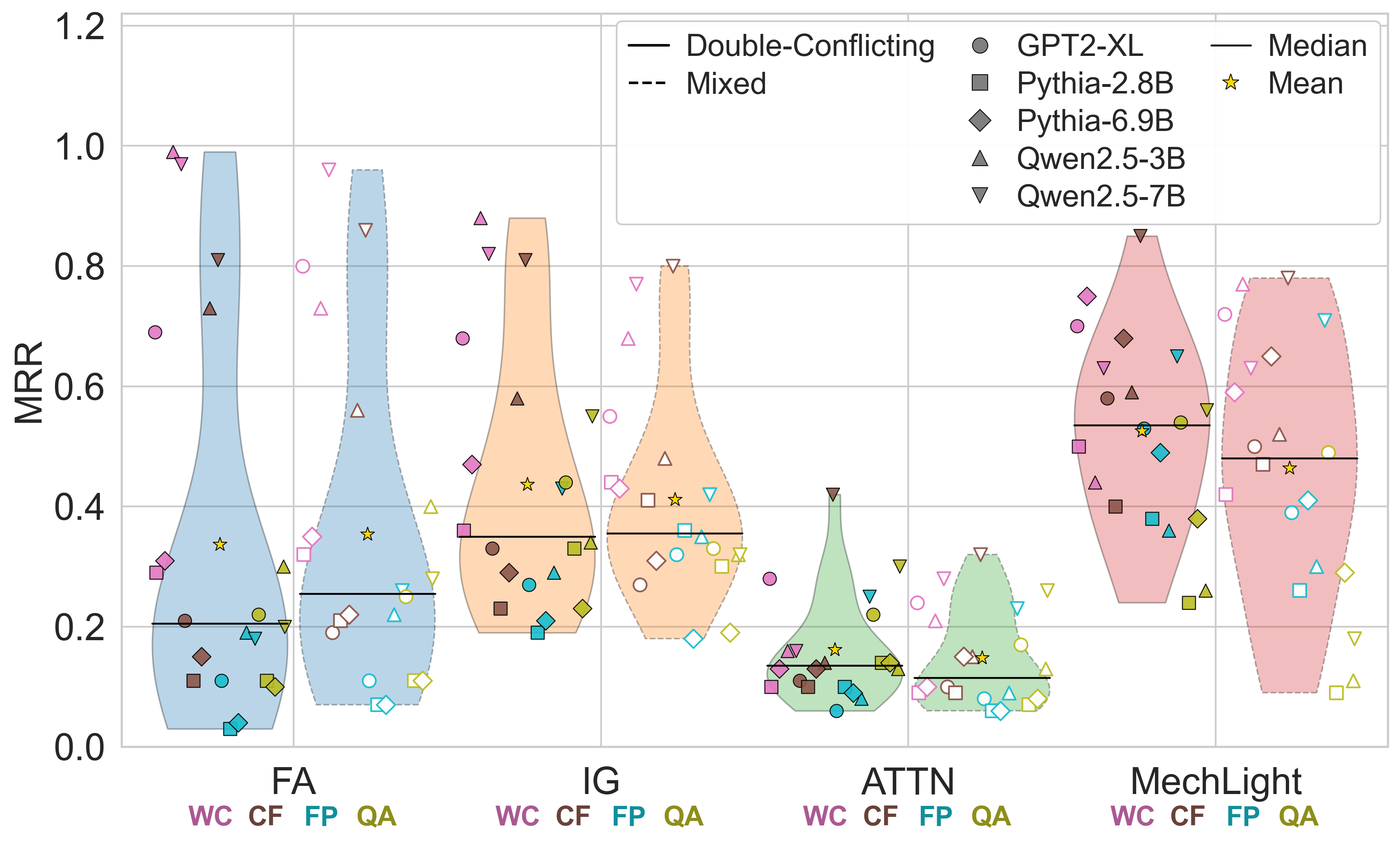}
    \caption{Double-Conflicting \& Mixed}
    \label{hcu:fig:answer-token-mrr-double-or-mixed}
  \end{subfigure}
  \caption{\mrr\ (Eq.~\ref{hcu:eq:mrr}) -- Mean Reciprocal Rank for the predicted answer tokens within the context-answer instances for all four context setups (\S\ref{hcu:input-regime-4}). Higher \mrr\ is better; larger values indicate the true answer token is placed near the top of the ranked list, meaning the explanation can more precisely indicate the answer location (\S\ref{hcu:sec:rq3-discussion}). Filled shapes stand for \textbf{Conflicting} setup and \textbf{Double-Conflicting} setup in the left and right subplot, respectively; hollow shapes stand for \textbf{Irrelevant} setup and \textbf{Mixed} setup in the left and right subplot, respectively. The colors denote the datasets, and the marker shapes denote the models.}

  \label{hcu:fig:answer-token-mrr-conflicting-irrelevant-double-mixed}
\end{figure*}

\subsection{Does the explanation pinpoint the exact context part(s) that were employed for the generated answer?}
\label{hcu:sec:rq3-discussion}
We report \mrr\ across models and datasets for all context setups in Fig.~\ref{hcu:fig:answer-token-mrr-conflicting-irrelevant-double-mixed}. Recall that \mrr\ is computed over context-answer instances, and larger values mean the gold answer token is ranked closer to the top (e.g., \mrr{}=$0.1$ corresponds to roughly rank $10$ on average).
From Fig.~\ref{hcu:fig:answer-token-mrr-conflicting-irrelevant-double-mixed}, we observe that \textit{all methods except \Attn\ usually place the answer token within the top-10 ranks for most model–dataset combinations}. This is reflected by medians above $\approx 0.1$ for \MI/\IG/\FA\ in most model--dataset pairs, whereas \Attn\ often has lower values. \MI\ is the best performing, although its performance lowers on the long-context dataset \textsc{ConflictQA}.

When a single piece of context is supplied (e.g., the \emph{Conflicting} context), as shown in Fig.~\ref{hcu:fig:answer-token-mrr-conflicting-or-irrelevant}, \textit{\MI\ and \IG\ are the two best methods} (median \mrr\ of 0.345 and 0.310, respectively), implying that the HEs often position the answer token within the top-3 tokens. \FA\ is next, with a median \mrr\ 0.175, but once again exhibits the largest variability between models and datasets
and low \mrr\ in long context datasets, suggesting that \FA\ is unstable and could require a computationally prohibitive number of ablations on long-context datasets. 
\Attn\ performs worst with a mean 0.147 \mrr.
\textit{As the context length increases (ConflictQA), all explanations struggle to position the answer tokens even within the top 10 important tokens.} Similar trend is found in Irrelevant setup, all methods show lower \mrr\ on short-context datasets (World Capital, Counterfact) and slightly higher on long contexts (notably \textsc{ConflictQA}), suggesting that explanations are easily distracted by short, irrelevant information.

\begin{figure*}[p]
  \centering
  \includegraphics[width=\columnwidth]{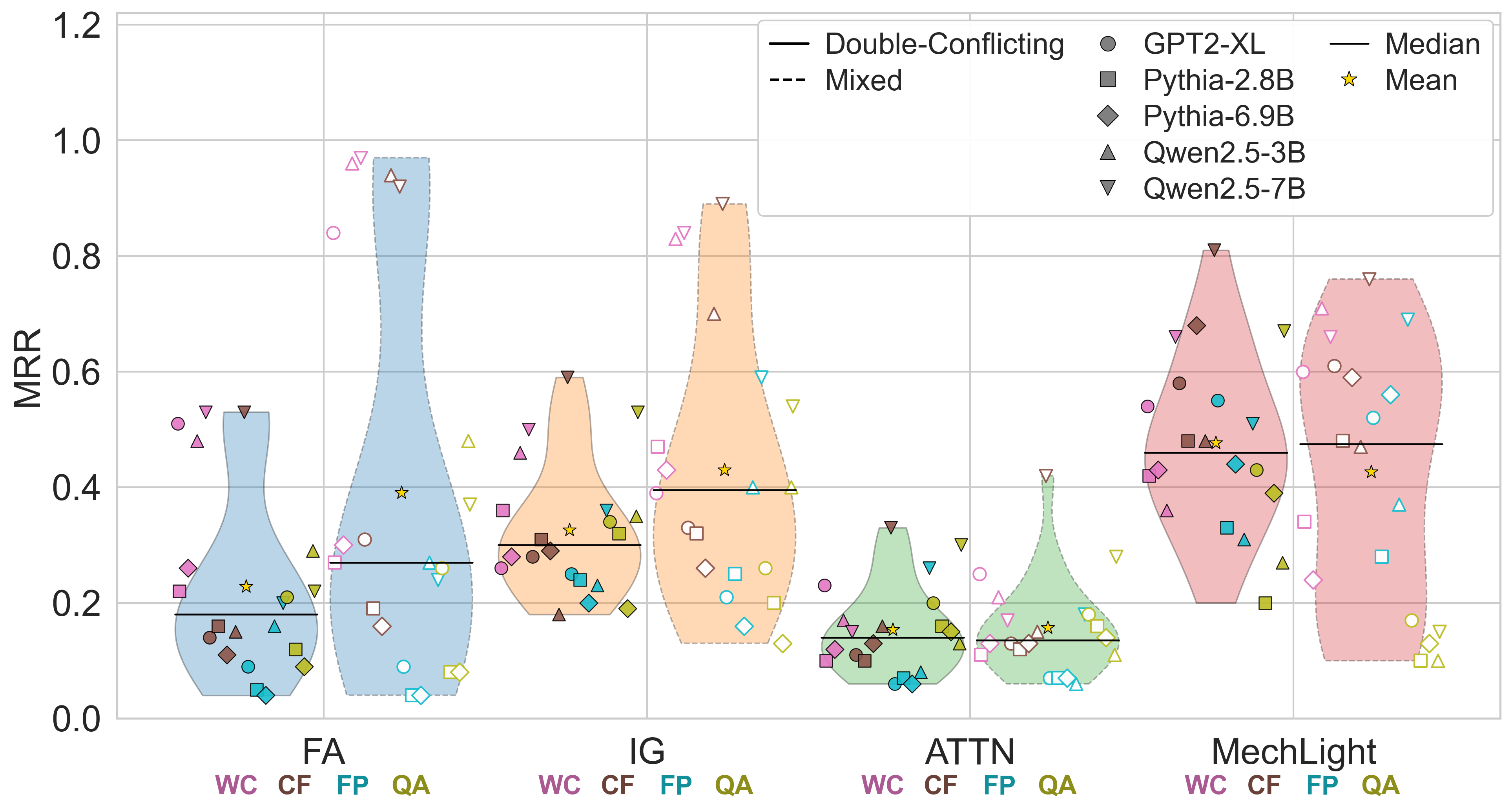}
    \caption{\mrr\ (Eq.~\ref{hcu:eq:mrr}) -- Mean Reciprocal Rank for the predicted answer tokens within the context-answer instances for \textbf{Double-Conflicting} and \textbf{Mixed} setups after \textbf{swapping} the position of the two context pieces (\S\ref{hcu:sec:four-context-setups}). Higher \mrr\ is better; larger values indicate the true answer token is placed near the top of the ranked list, meaning the explanation can more precisely indicate the answer location (\S\ref{hcu:sec:rq3-discussion}). Filled shapes stand for the swapped version of \textbf{Double-Conflicting} setup,  hollow shapes stand for the swapped version of \textbf{Mixed} setup. The colors denote the datasets, and the marker shapes denote the models.
  }
  \label{hcu:fig:answer-token-mrr-double-or-mixed-after-swap}
\end{figure*}

With two pieces of context, \MI\ performs best, with an average \mrr\ of 0.526, followed by \IG\ (average \mrr\ 0.436). \FA\ again shows the highest variability and performs poorly on long-context datasets (e.g., Fakepedia), where answer tokens usually fall outside the top-10 most important tokens. \Attn\ remains consistently worst, with an average \mrr\ 0.162. All methods show similar but slightly lower \mrr\ in the Mixed Context setup. Trends hold after swapping the two contexts in Double-Conflicting-Swap and Mixed-Swap (Fig.~\ref{hcu:fig:answer-token-mrr-double-or-mixed-after-swap} in App.~\ref{hcu:app:additional_results}), indicating that the relative position of the context does not affect the overall utility of the explanations in locating the tokens of the answer in the context.

\textbf{Case study.}
To better understand why HEs often fail to pinpoint the exact answer span, we conduct a qualitative case study in the \textit{Conflicting} regime. We define a failure as an instance where the model copies the context answer, yet none of the top-$k{=}5$ attribution tokens match the answer token. Inspecting 50 randomly sampled failures per dataset (World Capital, CounterFact, ConflictQA, Fakepedia), we find three recurring patterns  (examples in App.~\ref{hcu:app:case_study}, Tab.~\ref{hcu:tab:failure-examples}). (\textit{i}) \textbf{Generic tokens dominate:} on shorter contexts (World Capital, CounterFact), failures often place high attribution on stop-words, punctuation, or boiler-plate openers (e.g., ``The'', ``Fact:''). (\textit{ii}) \textbf{Nearby descriptors/entities:} on longer contexts (ConflictQA, Fakepedia), explanations frequently highlight nearby descriptors or entity names adjacent to the gold answer while missing the answer token itself. (\textit{iii}) \textbf{Question focus:} across datasets, attributions can concentrate on the interrogative phrase (e.g., ``What'', ``Who'', ``In what country''), which provides little evidence about \emph{where} the answer was sourced in the context. Together, these patterns help explain why token-level localisation can fail even when the model’s answer is copied from context.

\section{Conclusion}
\label{hcu:sec:conclusion}
We introduce the first gold standard framework for evaluating highlight explanations (HEs) for context utilisation. It encompasses controlled test cases under known ground-truth context utilisation scenarios, enabling direct assessment of HE accuracy in context attribution.
Across four controlled context scenarios, five models, and four datasets, we demonstrate our framework's general applicability using three established HE methods and one mechanistic interpretability-based method (\MI{}). We find that \MI\ shows the highest utility across all context scenarios and that some commonly used HE methods, \IG\ and \Attn, provide no value in making context usage transparent. Furthermore, all methods suffer from long contexts and exhibit position bias when two contexts are provided. This calls for future highlight explanation methods to provide accurate and reliable explanations of context usage at scale. 

\section{Limitations}
\label{hcu:sec:limitation}
Our work introduces the first benchmark robustly evaluating HEs for context-usage utility. Here, we discuss its scope and opportunities for extension.

\textbf{Input regimes.}
Our four input context setups all ensure each answer can be traced to \emph{one} dominant source (CK, PK, or one of two passages). Interesting future extensions are tasks requiring \emph{joint} reasoning over multiple passages (e.g., multi-hop QA or document-level summaries), where saliency must reflect blended evidence. In such settings, for example in multi-hop QA where the answer depends on multiple supporting contexts and evidence spans, the framework could be extended from single-source to set-valued gold evidence, provided that the gold contexts and supporting spans can first be identified. RQ1 could still evaluate whether the model relied on contextual rather than parametric knowledge, while RQ2 would extend to multiple gold context pieces and RQ3 to multiple gold supporting spans. This would require set-valued extensions of our metrics (\S\ref{hcu:sec:metrics}); for example, document-level attribution could compare the importance assigned to the union of tokens from all gold contexts against unused contexts, while token-level attribution could evaluate the average rank of multiple gold spans rather than a single token. The research questions could also be made more fine-grained, e.g., by measuring how many required context pieces are identified by the HEs and what proportion of the supporting spans they localise. We leave these extensions to future work. Notably, we specifically focus on cases where the answer in the provided context contradicts the model's parametric knowledge, as this is where the model is ensured to utilise the context information, and thus the HEs are expected to highlight the context. Additionally, such a gold standard for the model's context utilisation in PK-aligned (supportive) contexts or blended CK/PK behaviour~\citep{cheng2024understanding} is missing; thus, it remains an open challenge to evaluate the utility of highlight explanations in such setups. 

\textbf{Dataset and output scope.} 
We target QA datasets with \emph{present and short} gold answer spans in the context, enabling the development of our gold standard assessment of HE accuracy for context utilisation tasks. These tasks enable gold supervision for both (i) source attribution (CK vs.\ PK, RQ1, or which passage, RQ2) and (ii) token-level golden answer localisation (RQ3). However, these assumptions may not hold in open-ended or long-form generation, where outputs can lack a unique answer string that maps cleanly to a single supporting span. A possible extension is to decompose the generated output into sentences and infer, for each output sentence, its most supported source context (e.g., via entailment-style scoring or an LLM judge), yielding pseudo-gold source labels for the model's context usage source for studying RQ1/RQ2. For RQ3, one could align each sentence back to one or more supporting spans in the selected source, producing a set of reference spans rather than a single span. Finally, since \FA, \IG, \Attn, and \MI\ are usually defined per decoding step, applying these explanations to long-form settings would require aggregating attributions across multiple generated tokens. This would require additional design choices to avoid aggregation noise for attribution scores. Thus, developing robust pseudo-gold construction and attribution aggregation methods are promising directions to apply the framework to long-form generation tasks in the future.

\textbf{Explaining reasoning trajectories.} Our framework only focuses on short-answer QA tasks, and the explainability techniques compute attributions at a single decoding position (the answer token). However, reasoning models may produce an explicit multi-step \emph{reasoning trace} (a sequence of intermediate steps) before the final answer. An extension of our framework could compute token attributions over the input at multiple decoding steps, and then aggregate these attributions within each reasoning step (e.g., by sentence) to obtain highlight explanations for each reasoning step. This would enable step-wise versions of RQ1--RQ3, allowing us to study how CK-vs-PK reliance, passage preference, and evidence localisation shown by highlight explanations evolve over the course of reasoning.


\textbf{Interpreting opposing positional biases in multi-document setups.}
In dual-context setups, we observe opposing positional preferences across explanation families (\S\ref{hcu:sec:rq2-discussion}), and swapped-passage variants suggest the effect is largely position-driven rather than purely content-driven. We conjecture two contributing factors.

\textbf{(i) Model-induced position effects.}
Long-context studies show that whether evidence affects the final prediction depends strongly on its position in the prompt \citep{liu-etal-2024-lost}, and this reliance can decay with distance from the prediction \citep{khandelwal-etal-2018-sharp}. In our experiments, the strongest biases occur for the Pythia models (Fig.~\ref{hcu:fig:rank-topk-margin-double-conflicting-and-mixed}), suggesting that model architecture may modulate position sensitivity. One plausible contributor is positional encoding: RoPE encodes position via rotations and yields distance-dependent interactions \citep{su-etal-2021-roformer}, and recent theory suggests that RoPE encodings interacting with causal masking can produce primacy--recency trade-offs \citep{wu-etal-2025-position-bias}.

\textbf{(ii) Method-induced measurement bias.}
\IG/\FA\ are sensitivity-based: they quantify how the next-token distribution changes with gradients or input perturbations; their outcomes can depend on baseline choices, and perturbations/removals can also introduce distribution shift \citep{lyu2024towards}. In contrast, \Attn\ and \MI\ are attention-based: they reflect attention distributions; however, attention weights are not guaranteed to track causal contribution \citep{jain-wallace-2019-attention,wiegreffe-pinter-2019-attention,serrano-smith-2019-attention}. Moreover, local residual-stream readouts ( used by \MI{}) can be confounded when later components erase or cancel earlier residual directions \citep{janiak-etal-2024-adversarial,belrose2023tunedlens}. These measurement differences may contribute to the opposing positional biases observed across methods.

Overall, the observed positional preference of an explanation should be interpreted with caution: it may reflect genuine model position sensitivity, explanation method-induced artifacts, or both. We leave disentangling these factors to future work.

\textbf{Model scale and architecture.}
Our experiments systematically cover five models up to 7B parameters and reveal HE accuracy shifts with context length and model scale. While these results establish clear trends in this setting, extending the framework to instruction-tuned or much larger models may reveal additional context-usage patterns across all three research questions. For RQ1, instruction tuning may affect how strongly models prioritise provided context over parametric knowledge, as context use can become more sensitive to task directives and prompt framing. For RQ2, instruction tuning or larger scale may also affect how models arbitrate between multiple passages, potentially changing the passage-selection and positional-bias patterns observed in our dual-context setups. For RQ3, much larger models may rely on more distributed mechanisms for extracting answer evidence, which could make single-head projections less complete and motivate HE extraction methods that go beyond a single selected head. Future work could therefore compare matched variants with and without instruction tuning, as well as larger scales, to test whether the CK/PK separation, multi-passage selection, and answer-span localisation patterns observed here persist or change under these regimes.

\textbf{Explanation families.}
Our benchmark spans three standard post-hoc techniques plus our novel MI-based method. The framework's flexible architecture enables seamless integration of additional HE variants, both post-hoc and MI, for future investigation. \MI\ is designed to be lightweight and to produce token-level highlights from a single selected attention head, but this simplicity comes with limitations. First, the direct-logit readout provides a local attribution signal: later layers may reinforce or counteract earlier residual directions (downstream cancellation) \citep{janiak-etal-2024-adversarial,belrose2023tunedlens}, so a head’s direct contribution to the candidate logit gap may not fully reflect its end-to-end causal effect on the final answer. Second, \MI\ uses attention weights as the highlight object; attention is not guaranteed to reflect causal information flow~\citep{jain-wallace-2019-attention,wiegreffe-pinter-2019-attention}. A promising direction for future work is to integrate more intervention-based signals, e.g., causal head ranking via ablation~\citep{yu-etal-2023-characterizing,jin-etal-2024-cutting}, to improve causal faithfulness, and to explore causal-tracing or neuron-centric mechanisms~\citep{meng2022locating,shi2024ircan} to capture distributed context-usage circuits, while designing principled projections of such component-level evidence into stable token-level highlights.


\textbf{Explanation utility and human perspective.} Our framework leverages automated gold standard metrics, uniquely enabled by context usage scenarios where ground-truth source attribution is known. Supplementary faithfulness analyses validate these findings. While our principled automated approach avoids annotation costs, future human studies remain valuable for assessing perceived utility. These design choices establish a rigorous foundation for context-usage HE evaluation, with clear pathways for extending to more complex scenarios and explanation paradigms.

\section*{Acknowledgements}
$\begin{array}{l}\includegraphics[width=1cm]{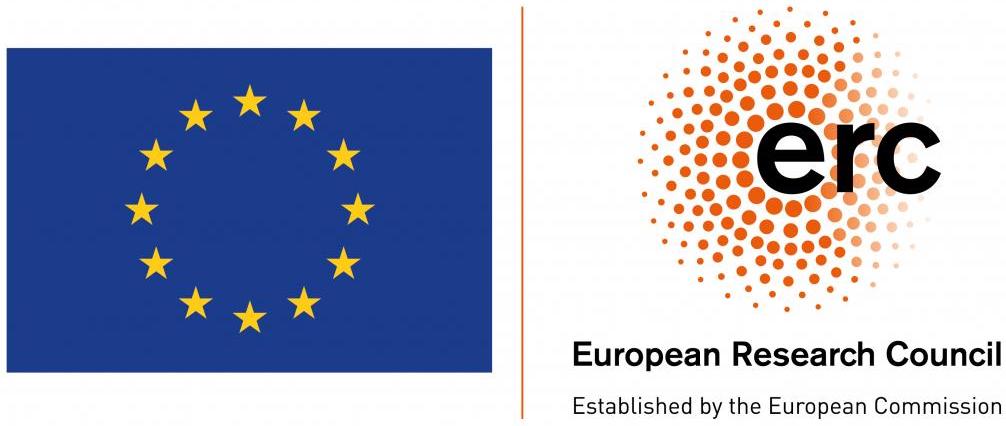} \end{array}$ 
This research was co-funded by the European Union (ERC, ExplainYourself, 101077481) and by the VILLUM FONDEN (grant number 40543). Views and opinions expressed are however, those of the author(s) only and do not necessarily reflect those of the European Union or the European Research Council. Neither the European Union nor the granting authority can be held responsible for them.

\FloatBarrier
\clearpage
\section{Supplementary Material}
\label{hcu:sec:hcu-supplementary}

\subsection{Replication Details}

\subsubsection{Datasets Details}
\label{hcu:app:dataset-details}
\textbf{Reconstruction overview.}
For each question, we construct matched instances across all regimes with token-level supervision while keeping the question fixed:
\begin{enumerate}[nosep,leftmargin=1.25em]
  \item \textbf{Memory check.} Query the target model without context to obtain its parametric answer; retain only items whose “conflicting” contexts genuinely contradict that answer (drop candidates that leak the model's parametric answer).
  \item \textbf{Regime assembly.} Build \textsc{Conflicting}, \textsc{Irrelevant}, \textsc{Double-Conflicting}, and \textsc{Mixed} prompts by concatenating passages so that each piece contains an explicit \emph{candidate answer token} (enabling RQ3).
  \item \textbf{Swaps} Create swapped dual-context variants to control for position.
\end{enumerate}
This yields per-question, per-regime test sets with known gold spans and answer locations suited to our utility-focused metrics. Dataset-specific construction details are as follows.
\begin{itemize}[leftmargin=1.3em,itemsep=2pt]
  \item \textsc{Fakepedia}\; It contains encyclopaedic, single-hop questions spanning 45 Wikidata-style relations (e.g., \textit{employed-by}, \textit{official-language}).  The synthetic counterfactual context shipped with each item serves as the \emph{conflicting} context; an \emph{irrelevant} context is sampled from a different country that shares the same relation. See Table~\ref{hcu:tab:exmaple-for-fakepedia-dataset}.

  \item \textsc{World Capital}\;It contains purely geographical questions under a single relation, \textit{capital-of}.   The made-up capital statement is reused as the \emph{conflicting} context; an \emph{irrelevant} context is taken from another country. 

  \item \textsc{Counterfact}\;It contains entity-centric biography questions covering 5 relations such as \textit{works-in-area-of} and \textit{originated-in}. The dataset's edited context is kept as \emph{conflicting}; its annotated irrelevant context is reused. 

  \item \textsc{ConflictQA}\;It contains multi-domain questions across 7 relations (e.g., \textit{occupation}, \textit{genre}, \textit{founded-year}).  The original contradictory context remains \emph{conflicting}; the supplied noise context (same relation, different subject) becomes \emph{irrelevant} after we extract the answer entity within the irrelevant context via Llama-4.  
\end{itemize}

\begin{table}[t]
\centering
\footnotesize
\setlength{\tabcolsep}{4pt}       
\renewcommand{\arraystretch}{0.92} 
\begin{adjustbox}{max width=\linewidth} 
\begin{tabular}{l l r r}
\toprule
Dataset & Ctx. type & \#Inst. & Avg ctx len \\
\midrule
\multirow{4}{*}{World Capital} & Conf.       & 55,830 &   37.9 \\
                               & Irre.       & 55,830 &   37.9 \\
                               & DoubleConf. & 55,830 &   75.9 \\
                               & Mixed       & 55,830 &   75.9 \\
\midrule
\multirow{4}{*}{Counterfact}   & Conf.       &    802 &   44.8 \\
                               & Irre.       &    802 &   44.8 \\
                               & DoubleConf. &    802 &   89.5 \\
                               & Mixed       &    802 &   89.5 \\
\midrule
\multirow{4}{*}{Fakepedia}     & Conf.       &  5,348 &  704.5 \\
                               & Irre.       &  5,348 &  704.5 \\
                               & DoubleConf. &  5,348 & 1408.8 \\
                               & Mixed       &  5,348 & 1408.9 \\
\midrule
\multirow{4}{*}{ConflictQA}    & Conf.       &  1,343 &  593.1 \\
                               & Irre.       &  1,343 &  454.1 \\
                               & DoubleConf. &  1,343 & 1190.2 \\
                               & Mixed       &  1,343 & 1047.2 \\
\bottomrule
\end{tabular}
\end{adjustbox}
\caption{Counts and average context length for reconstructed datasets regarding all four input regimes: Conf.(conflicting context); Irre.(irrelevant context); DoubleConf. (Double-Conflicting) contexts; Mixed (concatenation of conflicting and irrelevant contexts). Double-Conflicting-Swap and Mixed-Swap have identical statistics as DoubleConf. and Mixed (only positions are reversed).}
\label{hcu:tab:dataset-statistics}
\end{table}

Tab.~\ref{hcu:tab:dataset-statistics} summarises the statistics of the reconstructed datasets. To keep computation tractable, we cap the number of instances used for explanation generation and evaluation at 2,000 per dataset–context type for the short-context datasets (World Capital, Counterfact) and 1,000 for the long-context datasets (Fakepedia, ConflictQA), given the runtime overhead of Feature Ablation, which is more pronounced for long contexts.

\subsubsection{Other Details for Explanation Evaluation}
\label{hcu:app:k_3_and_9_for_rq1_and_2}
We select the top-$k$ important highlight explanations for utility evaluation, $k{=}5$ in the main discussion, as users often focus on a few instead of the complete cause of an event \citep{miller2019explanation}. To assess robustness, we conduct experiments with top-$3$ and top-$9$ explanations on a representative subset of regimes, as a human can usually hold $7\pm 2$ objects (here, explanation tokens) in short-term memory according to Miller's law~\citep{miller1956magical} and subsequent work has suggested a more restrictive capacity of approximately three to five chunks~\citep{cowan2001magical}. The findings are consistent across different $k$.


\subsubsection{kNN Mutual Information Implementation Details}
\label{hcu:app:knnmi}
Given a top-$k$ highlight vector $X^{(k)}_s \in {R}^{k}$ extracted from a target segment $s$ (e.g., $s{=}c$ for RQ1 or $s\!\in\!\{c_1,c_2\}$ for RQ2) and a binary behaviour label $Y$ (RQ1: $\textsc{C}$ vs.\ $\textsc{M}$; RQ2: $\textsc{C1}$ vs.\ $\textsc{C2}$), we estimate the mutual information—i.e., the reduction in label uncertainty provided by the top-$k$ features—as
\begin{equation}
I\bigl(Y;X^{(k)}_s\bigr) \;=\; H(Y)\;-\;H\!\bigl(Y\mid X^{(k)}_s\bigr).
\end{equation}

\textbf{Label entropy.}
Let $p=\Pr(Y=1)$ be the empirical class prior. Using natural logarithms (nats),

\begin{equation}
H(Y)=
\begin{cases}
0, & p\in\{0,1\},\\
-p\log p-(1-p)\log(1-p), & p\in(0,1).
\end{cases}
\label{hcu:eq:hy}
\end{equation}

\textbf{kNN posterior and conditional entropy estimation.}
For each sample $x^{(k)}_{s,i}$, let $\mathcal{N}_k(i)$ be the set of its $k$ nearest neighbours in the feature space (Euclidean; the point itself is excluded; $k{=}5$). The local posterior (class–1 probability) is defined by the neighbour fraction:

\begin{equation}
\label{hcu:eq:post}
\begin{aligned}
\hat p_i
  &= \frac{1}{k}\sum_{j\in\mathcal{N}_k(i)} \mathbf{1}\{y_j=1\}\\
  &\approx \Pr\!\left(Y=1 \,\middle|\, X^{(k)}_s=x^{(k)}_{s,i}\right).
\end{aligned}
\end{equation}


With $h_b(q) = -q\log q -(1-q)\log(1-q)$ the binary entropy, the conditional entropy is estimated by averaging local entropies:
\begin{equation}
\widehat{H}\!\bigl(Y\mid X^{(k)}_s\bigr) \;=\; \frac{1}{n}\sum_{i=1}^n h_b(\hat{p}_i).
\label{hcu:eq:condent}
\end{equation}

\textbf{Normalised Mutual Information.}
The MI estimate is
\begin{equation}
\widehat{I}\bigl(Y;X^{(k)}_s\bigr) \;=\; H(Y)\;-\;\widehat{H}\!\bigl(Y\mid X^{(k)}_s\bigr).
\label{hcu:eq:mi}
\end{equation}
To express MI in bits, we use $\widehat{I}_{\text{bits}}=\widehat{I}/\log 2$.
Our reported quantity is the label-entropy–normalised mutual information, i.e., the fraction of label uncertainty explained by the top-$k$ highlights:
\begin{equation}
 \NMutInfk;=\; \frac{\widehat{I}\bigl(Y;X^{(k)}_s\bigr)}{H(Y)} \;\in\; [0,1].
\label{hcu:eq:nmi}
\end{equation}

\subsubsection{MDL Probe Implementation Details}
\label{hcu:app:mdl_prob}
In its classical formulation, the Minimum Description Length (MDL) principle provides a Bayesian-inspired framework for model selection. A model class $\mathbf{M}$ is a set of candidate models $M_i$; for example, $\mathbf{M}$ could be the family of cubic polynomials, with one member $M_i$ given by $5x^3$. Between two model classes $\mathbf{M}_a$ and $\mathbf{M}_b$, the preferred class is the one that yields the smaller \textit{stochastic complexity}, where the \emph{stochastic complexity} of $D$ with respect to a model class $\mathbf{M}$ is defined as the shortest achievable code length for $D$ when encoding is restricted to models in $\mathbf{M}$. Intuitively, a model that fits the data better assigns higher likelihoods and therefore produces shorter code lengths.

There are two standard methods for computing code lengths of deep neural nets. In the variational formulation \citep{hinton-mdl}, the description length of a dataset under a model is upper bounded by the sum of two terms: the negative log-likelihood of the data under the model and a complexity penalty given by the KL divergence between a variational posterior over parameters and a prior. 
This provides a tractable bound on stochastic complexity but depends strongly on the choice of prior and approximating family. Prequential (or online) coding measures description length by sequentially predicting the data. At each step, the model parameters are updated on past observations and used to predict the next outcome; the surprisal $-\log p(y_t \mid x_t, \theta_{t-1})$ is then added to the cumulative code length. The resulting quantity captures how efficiently a model class can compress data when trained incrementally. 
\citet{blier-mdl} shows that variational MDL often yields loose compression bounds, whereas prequential MDL produces much tighter estimates that align more closely with generalisation performance.

In NLP, MDL has been used in the context of ``probing tasks''. \citet{ep-tenney} used a suite of classifiers or probes to predict a token's syntactic (e.g., part-of-speech) and semantic tags from its embedding. A high accuracy in this task was interpreted as the embedding's ability to encode such linguistic information. The subsequent criticisms focused on the problem of ``classifier knowledge'' -- was the knowledge encoded in the embeddings, or did the classifier learn the task? \citet{voita-titov-2020-information} used ``MDL probing'' to solve this problem. Specifically, the prequential code lengths were computed using the formula $L_{\text{preq}}(\mathcal{D}) \;=\; \sum_{t=1}^{N} \ell_t \;=\; -\sum_{t=1}^{N} \log_2 p_{\theta_{t-1}}(y_t \mid h_t)$. Here $h=f_\phi(x)$ is a representation of a token from a frozen encoder $f_\phi$ and $p_\theta(y\mid h)$ is the predicted probabilities from a parametric probe. A lower $L_{\text{preq}}$ implied that the labels were easier to compress given the reps $h_t$, i.e., the property was more naturally encoded.

The MDL part of our simulatability test uses the same technique with top-$k$ importance scores derived from highlight explanations. We intend to show that these features have the discriminative power to predict a model's answer behaviour. We use a two-layer MLP classifier that is first trained on $10\%$ of the data. In the coding phase, we update the parameters $\theta$ for a mini-batch of size $10$. We repeat this entire process on $10$ random reshuffles of the data and report the average results.


\subsubsection{Faithfulness Evaluation Implementation Details}
\label{hcu:app:faithfulness}

Utility metrics in \S\ref{hcu:sec:methodology} assess how accurately a highlight explanation (HE) reflects the model's context usage. Faithfulness answers a complementary question: how well an HE aligns with the model's internal decision process. We therefore report \emph{Comprehensiveness} and \emph{Sufficiency} on the same models and datasets as the main experiments, under two regimes: \textbf{Conflicting} (single-context) and \textbf{Double-Conflicting} (dual-context).

Following prior work~\citep{deyoung-etal-2020-eraser,atanasova2022diagnostics}, let $\pi(1{:}k)$ be the indices of the top-$k$ tokens by HE scores $\phi$. For each $k\in\mathcal{K}$, let $x^{\mathrm{mask}\,k}$ be $x$ with tokens $\pi(1{:}k)$ masked, and $x^{\mathrm{keep}\,k}$ keep only $\pi(1{:}k)$. Writing $\ell(z)\!=\!\log p(a\mid z)$,
\begin{equation}\label{hcu:eq:aopc-comp}
\mathrm{AOPC}_{\text{comp}} \;=\; \frac{1}{|\mathcal{K}|}\sum_{k\in\mathcal{K}}\big[\ell(x)-\ell(x^{\mathrm{mask}\,k})\big],
\end{equation}
\begin{equation}\label{hcu:eq:aopc-suff}
\mathrm{AOPC}_{\text{suff}} \;=\; \frac{1}{|\mathcal{K}|}\sum_{k\in\mathcal{K}}\big[\ell(x)-\ell(x^{\mathrm{keep}\,k})\big].
\end{equation}
Higher $\mathrm{AOPC}_{\text{comp}}$ and lower $\mathrm{AOPC}_{\text{suff}}$ indicate greater faithfulness.

For World Capital/CounterFact (short contexts) we use $\mathcal{K}\!=\!\{1,\dots,5\}$. For Fakepedia/ConflictQA (long contexts), we use a fractional grid $\mathcal{K}\!=\!\{0.01,0.02,0.03,0.04,0.05\}\cdot n$ to avoid overly sparse inputs and keep the number of forward passes manageable.

\subsection{Additional Results}
\label{hcu:app:additional_results}

\subsubsection{Case Study for Explanation Failure}
\label{hcu:app:case_study}

To probe explanation failure for RQ3, whether a highlight explanation (HE) identifies the exact span supplying the model's answer, we focus on the \emph{conflicting-context} regime. In this case study, we recognise a failure as when the model \emph{copies the correct answer from context} but \emph{none of the top-$k=5$ attribution tokens} matches the answer token. We examine 50 randomly sampled failures from each dataset (World Capital, ConflictQA, CounterFact, Fakepedia) and observe three recurring patterns. Tab.~\ref{hcu:tab:failure-examples} shows representative examples (top-$5$ highlight tokens underlined; prompt slices truncated with ellipses).

\begin{table*}[p]
\centering\footnotesize
\setlength{\tabcolsep}{5pt}
\renewcommand{\arraystretch}{1.08}
\begin{tabular*}{\textwidth}{@{\extracolsep{\fill}}
  >{\raggedright\arraybackslash}p{0.22\textwidth}
  >{\raggedright\arraybackslash}p{0.5\textwidth}
  >{\raggedright\arraybackslash}p{0.14\textwidth}
@{}}
\toprule
\textbf{Dataset} & \textbf{Prompt slice} & \textbf{Model answer} \\
\midrule
\multicolumn{3}{@{}l}{\textbf{Obs.\,1: Generic tokens (stop-words / punctuation)}}\\
World Capital &
\underline{The} \underline{capital} \underline{of} Afghanistan \underline{is} \underline{.}\ Valletta \dots{} Q:\ What is the capital of Afghanistan? A: &
Valletta \\
World Capital &
\underline{The} \underline{capital} \underline{of} Algeria \underline{is} \underline{.}\ Sukhumi \dots{} Q:\ What is the capital of Algeria? A: &
Sukhumi \\
CounterFact &
\underline{Fact} \underline{:} Accra, \underline{the} \underline{capital} city \underline{of} Lebanon. Q:\ Accra, the capital city of A: &
Lebanon \\
\midrule
\multicolumn{3}{@{}l}{\textbf{Obs.\,2: Nearby descriptors/entity names}}\\
ConflictQA &
Trade paper profile begins \underline{Veteran} \underline{producer} \underline{and} \underline{studio} \underline{head} Dore Schary \dots{} Q:\ Who directed Act One? &
Dore Schary \\
ConflictQA &
Effects journal states \underline{Visual} \underline{effects} \underline{expert} Bruno \underline{was} \underline{hired} for ``Virus'' \dots{} Q:\ Who was the director of Virus? &
John Bruno \\
Fakepedia &
Apple Pay white-paper: the \underline{ground} \underline{breaking} \underline{payment} \underline{service} \underline{launched} with Intel hardware \dots{} Q:\ Apple Pay, a product created by &
Intel \\
\midrule
\multicolumn{3}{@{}l}{\textbf{Obs.\,3: Question focus}}\\
World Capital &
\dots{} Q:\ \underline{What} \underline{is} \underline{the} \underline{capital} \underline{of} Albania? A: &
Berlin \\
ConflictQA &
\dots{} Q:\ \underline{Who} \underline{was} \underline{the} \underline{director} \underline{of} ``Virus''? A: &
John Bruno \\
CounterFact &
\dots{} Q:\ \underline{What} \underline{is} \underline{the} \underline{capital} \underline{of} Burgundy? A: &
Bangkok \\
\bottomrule
\end{tabular*}
\caption{Representative failure examples by pattern (top-5 highlight tokens underlined).}
\label{hcu:tab:failure-examples}
\end{table*}


\textbf{Observations} 
(\textit{i}) \textbf{Spurious attention to generic words.} On short contexts (World Capital, CounterFact), about one-third of failures place top-5 highlights on punctuation/stop-words or boiler-plate (e.g., “The,” “Fact:”), yielding only superficial cues (Tab.~\ref{hcu:tab:failure-examples}). (\textit{ii}) \textbf{Nearby descriptive words or entity names.} In ConflictQA and Fakepedia (longer contexts), roughly half and about one-third of failures, respectively, highlight adjacent proper nouns/adjectives (e.g., target entities or descriptors) while missing the gold token. (\textit{iii}) \textbf{Over-attribution to question tokens.} Across datasets, HEs sometimes lock onto the interrogative phrase; this appears in about half of World-Capital and ConflictQA failures and roughly one-quarter of CounterFact and Fakepedia. Taken together, these patterns expose complementary weaknesses of HEs for pinpointing the exact answer location.

\textbf{Implications}
These patterns illustrate complementary failure modes for token-level localisation: explanations can be distracted by high-frequency tokens, gravitate to near-answer descriptors, or focus on the question template rather than the evidence span. They highlight a real limitation: current HEs may fail to reliably pinpoint the precise evidence location even when the model copies the answer from context.

\begin{figure*}[p]
  \centering
  \begin{subfigure}[t]{0.48\textwidth}
    \centering
    \includegraphics[width=\textwidth]{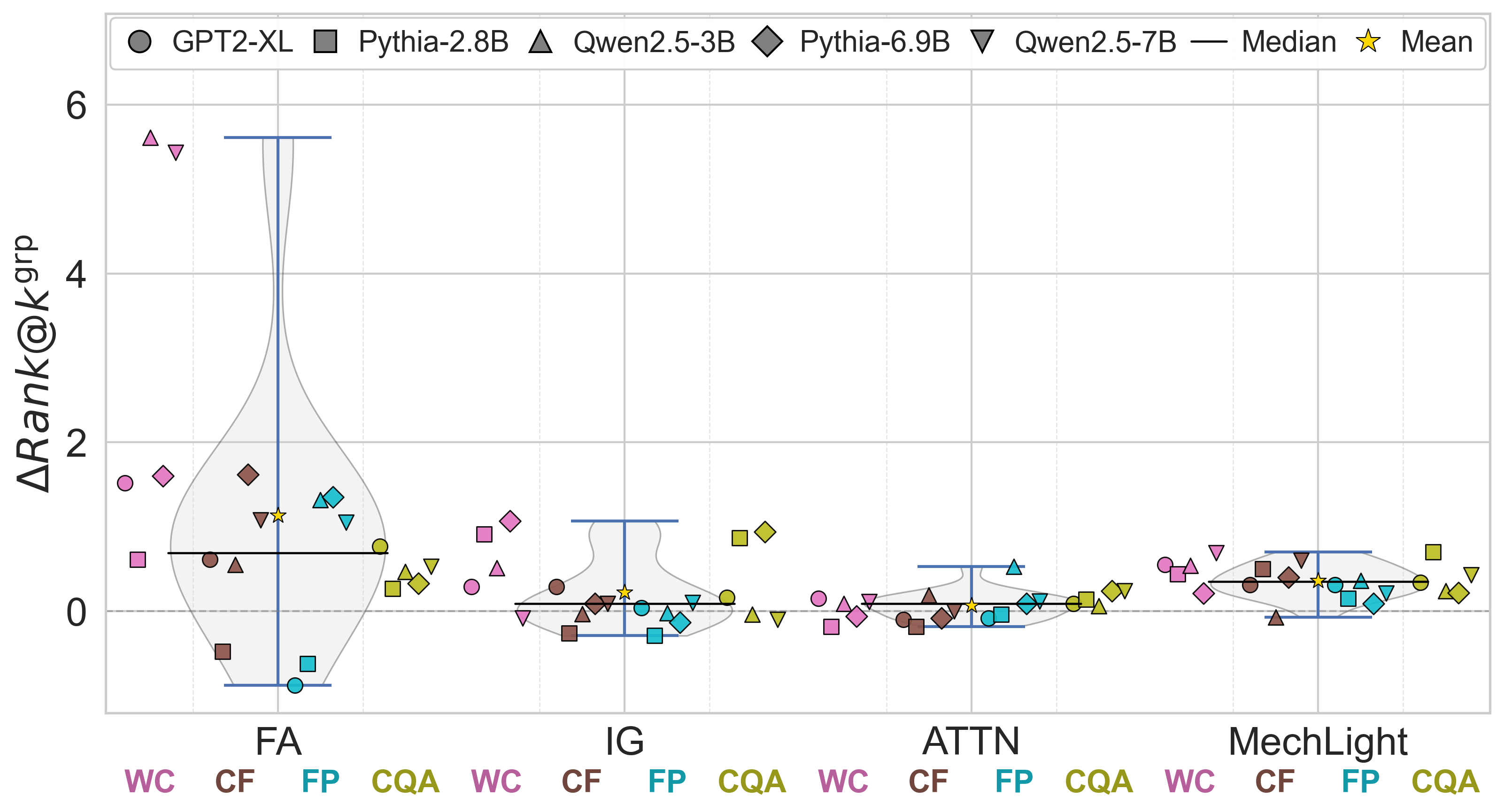}
    \caption{Top 3 Highlights in Conflicting Context}
  \label{hcu:fig:conflicting-topk-3-rank-datasets-margin-only}
  \end{subfigure}
  \hfill
  \begin{subfigure}[t]{0.48\textwidth}
    \centering
    \includegraphics[width=\textwidth]
{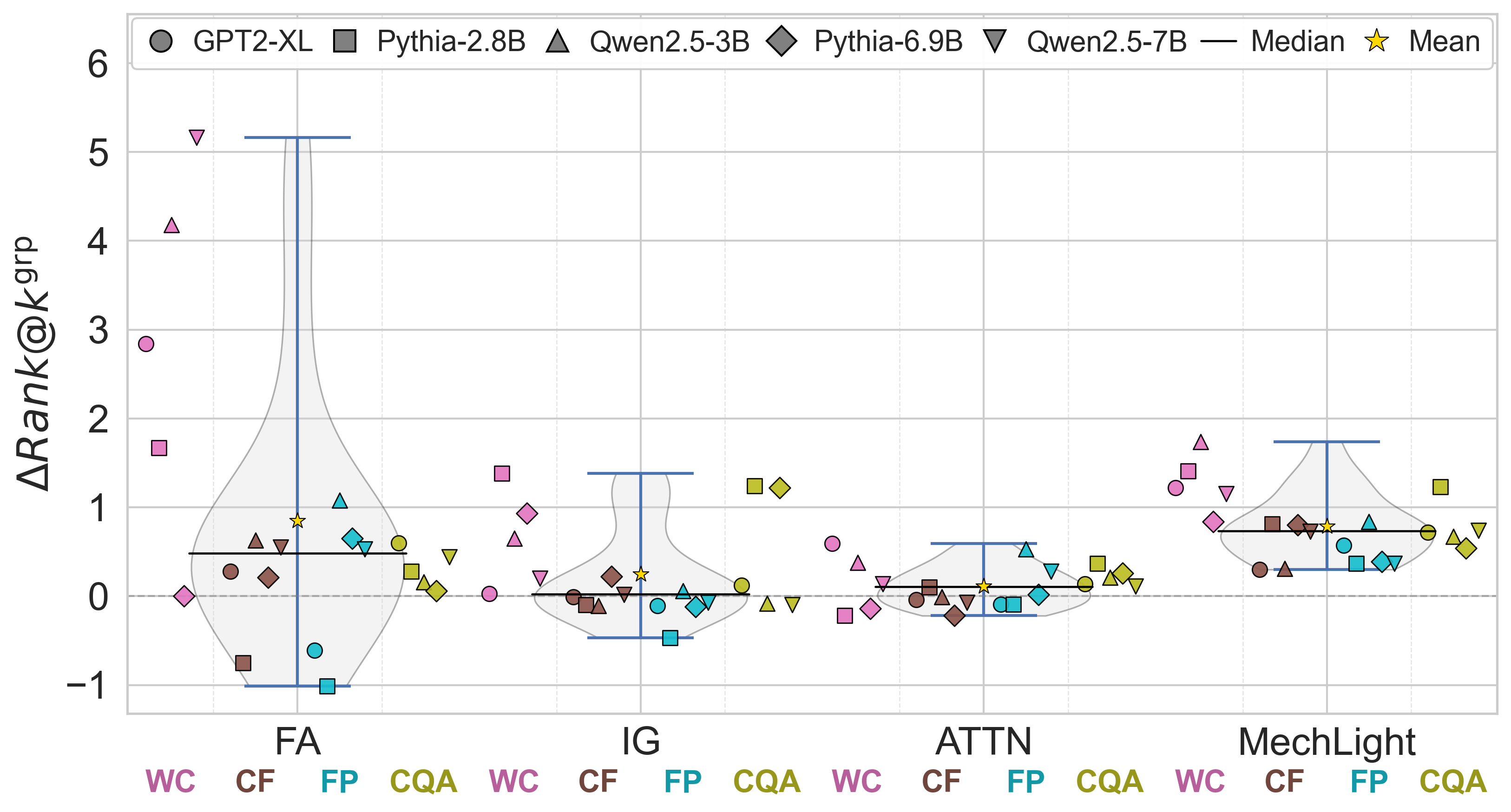}
    \caption{Top 9 Highlights in Conflicting Context}
  \label{hcu:fig:conflicting-topk-9-rank-datasets-margin-only}
  \end{subfigure}
    \caption{\DrankkGroup\ ($k=3;9$) (Eq. \ref{hcu:eq:drankk-cross-instance-group}) -- average margins for the explanation importance rank of context tokens in context vs. memory answer instances in \textbf{Conflicting} setup (\S\ref{hcu:sec:four-context-setups}). Positive and higher \DrankkGroup\ means the explanations can better distinguish the model's context usage behaviour (choosing PK or CK for the answer (\S\ref{hcu:sec:rq1-discussion})). The colors denote the datasets, and the marker shapes denote the models.}
  
  \label{hcu:fig:conflicting-topk-3-and-topk-9-rank-datasets-margin-only}
\end{figure*}

\begin{figure*}[ht]
  \centering
  \begin{subfigure}[t]{0.48\textwidth}
    \centering
    \includegraphics[width=\textwidth]
    {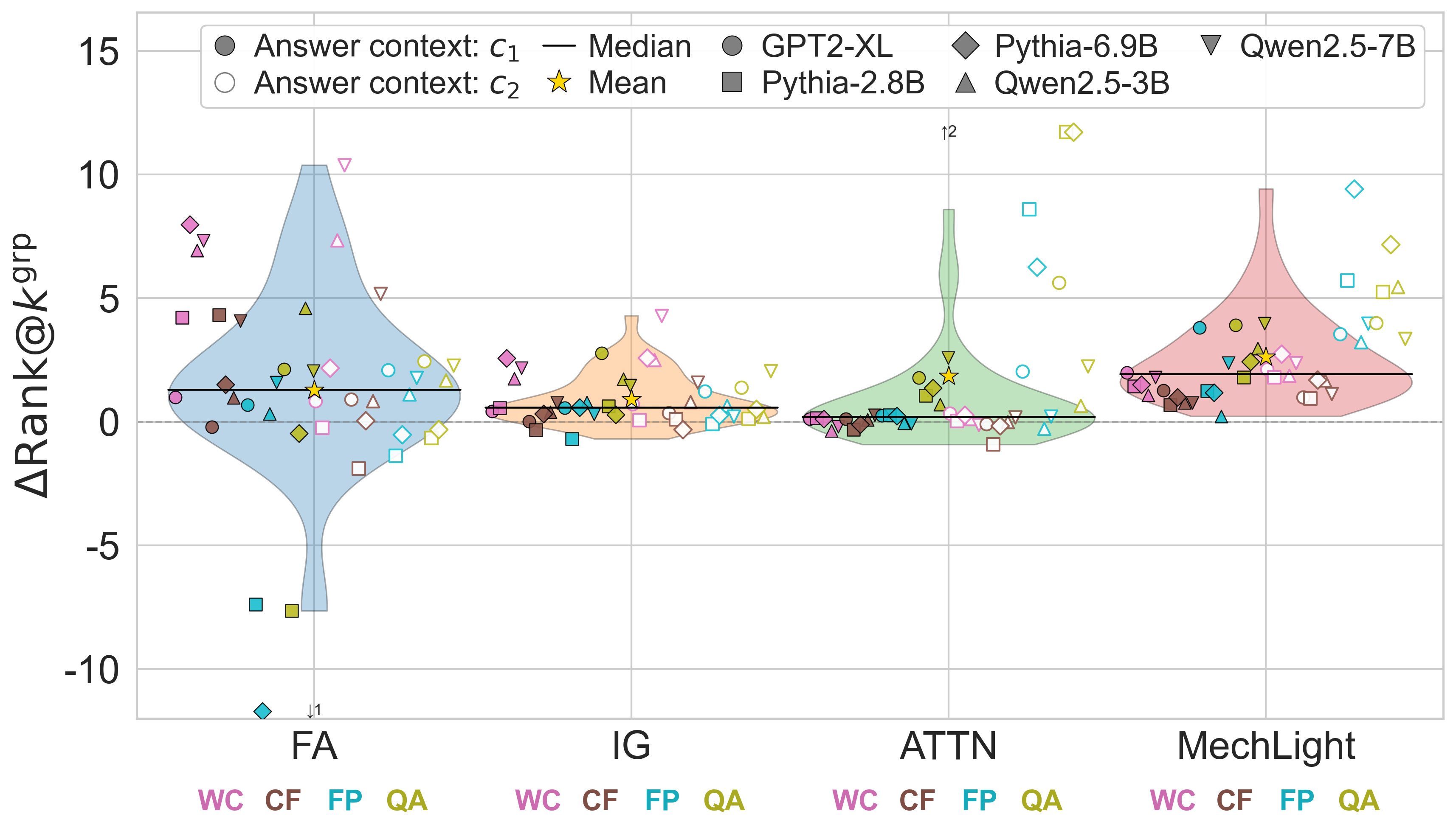}
    \caption{Top 3 Highlights in Double-Conflicting Contexts}
  \label{hcu:fig:double-conflicting-topk-3-rank-datasets-margin-only}
  \end{subfigure}
  \hfill
  \begin{subfigure}[t]{0.48\textwidth}
    \centering
\includegraphics[width=\textwidth]
{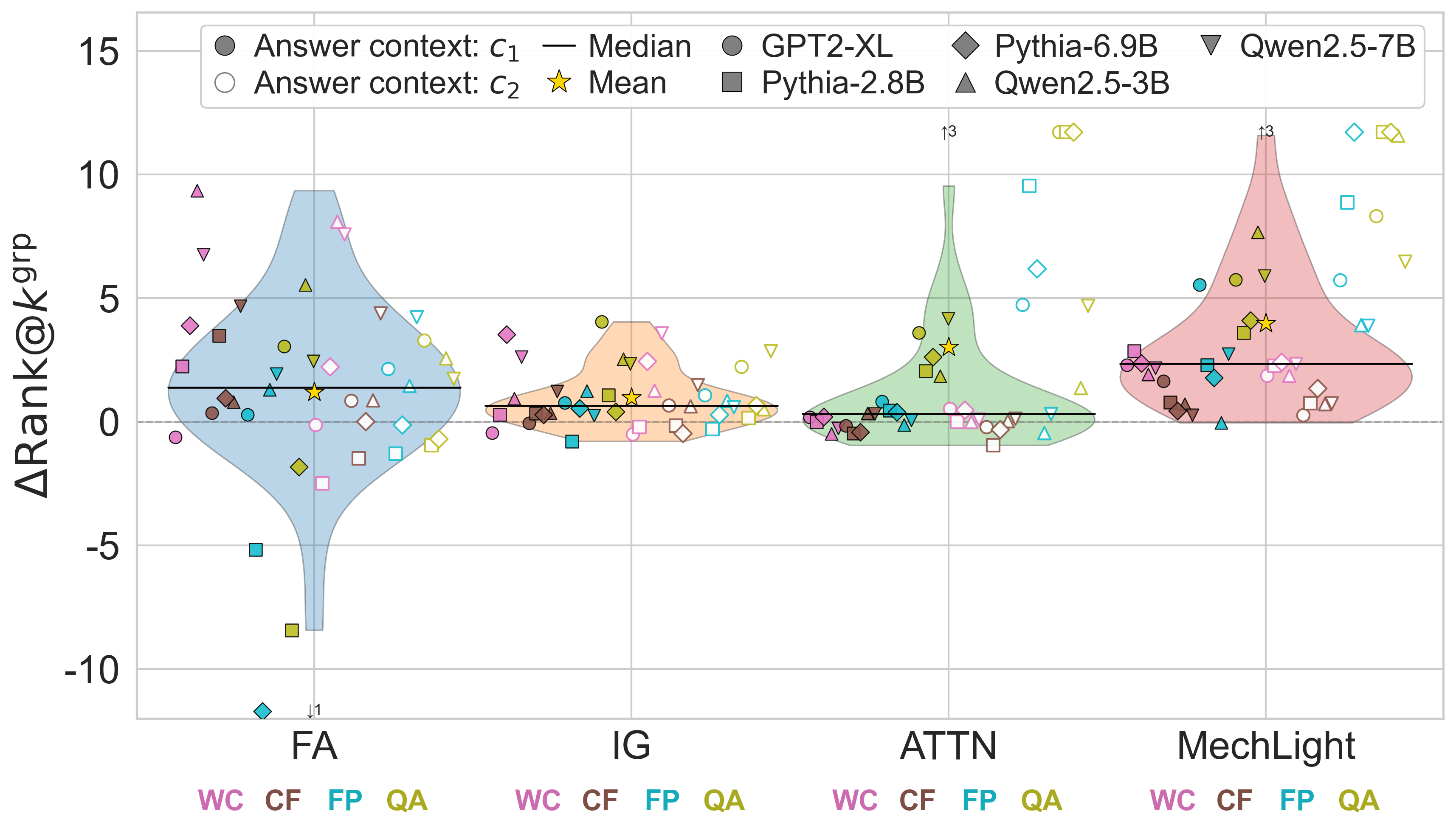}
    \caption{Top 9 Highlights in Double-Conflicting Contexts}
  \label{hcu:fig:double-conflicting-topk-9-rank-datasets-margin-only}
  \end{subfigure}
    \caption{\DrankkGroup\ ($k=3;9$) (Eq.~\ref{hcu:eq:drankk-cross-instance-group}) -- average margins for the rank of context $c_1$ and $c_2$ between two instance groups $D_{c_1}$ and $D_{c_2}$  in the \textbf{Double-Conflicting} setup (\S\ref{hcu:input-regime-2}). Higher \DrankkGroup is better, and a positive number means that the explanation can correctly indicate which context piece is used for the answer (\S\ref{hcu:sec:rq2-discussion}). The values are obtained by comparing the average importance scores of either the first context piece  ($c_1$, filled shapes at the left) or the second context piece ($c_2$, hollowed shapes at the right) between the two instance groups. The colors denote the datasets, and the marker shapes denote the models.
  }
  \label{hcu:fig:double-conflicting-topk-3-and-topk-9-rank-datasets-margin-only}
\end{figure*}

\begin{figure*}[ht]
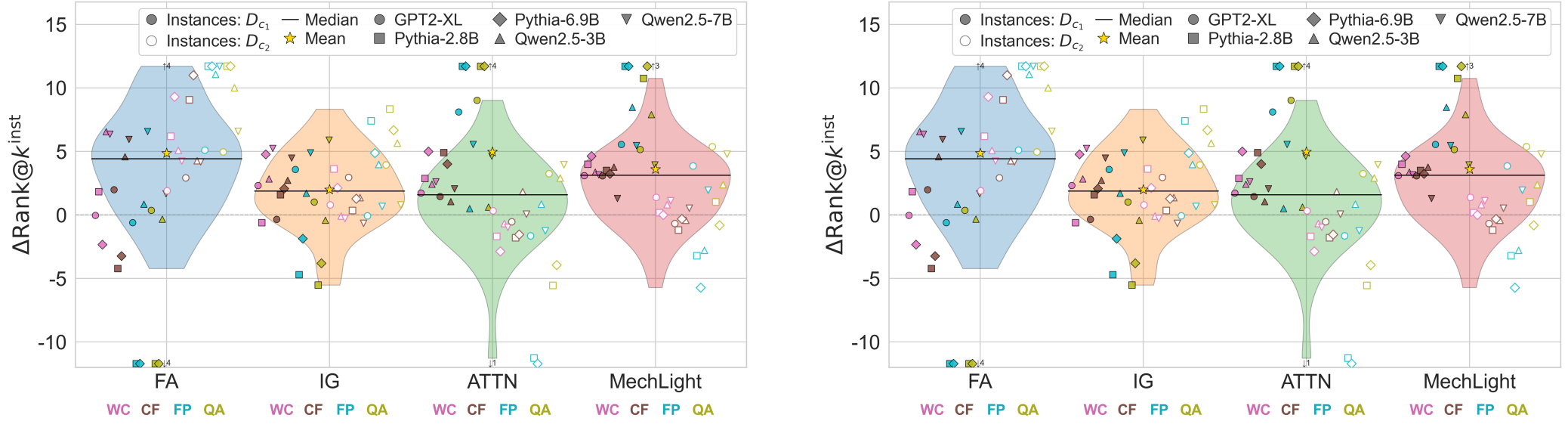

  \centering
  \begin{subfigure}[t]{0.48\textwidth}
    \centering
    \includegraphics[width=\textwidth]
{chapters/Highlight_context_utilisation/images/conflicting_1_and_2/k3/compact/rankk_instance_separate_datasetlabel.png}
    \caption{Top 3 Highlights in Double-Conflicting Contexts}
  \label{hcu:fig:double-conflicting-topk-3-rank-datasets-within-instance-margin}
  \end{subfigure}
  \hfill
  \begin{subfigure}[t]{0.48\textwidth}
    \centering
\includegraphics[width=\textwidth]
{chapters/Highlight_context_utilisation/images/conflicting_1_and_2/k9/compact/rankk_instance_separate_datasetlabel.png}
    \caption{Top 9 Highlights in Double-Conflicting Contexts}
  \label{hcu:fig:double-conflicting-topk-9-rank-datasets-within-instance-margin}
  \end{subfigure}
  \caption{\DrankkInst\ ($k=3;9$)  (Eq.~\ref{hcu:eq:drankk-within-instance})  -- average \emph{within-instance-group} margins between the rank of the answer context piece and the other context piece in the \textbf{Double-Conflicting} setup (\S\ref{hcu:input-regime-2}).  Higher \DrankkInst\ is better, and a positive number means that the explanation can correctly indicate which context piece is used for the answer (\S\ref{hcu:sec:rq2-discussion}). The values are obtained by comparing importance scores of the answer-piece context and the other piece within either the instance group answering from the first context piece  ($D_1$, filled shapes at the left) or that answering from the second context piece ($D_2$, hollowed shapes at the right). The colors denote the datasets, and the marker shapes denote the models.}
  \label{hcu:fig:double-conflicting-topk-3-and-topk-9-rank-datasets-within-instance-margin}
\end{figure*}

\begin{table*}[t]
\centering
\small
\setlength{\tabcolsep}{3pt}
\renewcommand{\arraystretch}{0.95}
\resizebox{\textwidth}{!}{%
\begin{tabular}{@{}ccccc|cc@{}}
\toprule
& & & \multicolumn{2}{c|}{\textbf{Conflicting}} & \multicolumn{2}{c}{\textbf{Double-Conflicting}} \\
\cmidrule(lr){4-5} \cmidrule(lr){6-7}
\textbf{Dataset} & \textbf{Model} & \textbf{Method} & AOPC$_\text{comp}\!\uparrow$ & AOPC$_\text{suff}\!\downarrow$ & AOPC$_\text{comp}\!\uparrow$ & AOPC$_\text{suff}\!\downarrow$ \\
\midrule
\multirow{8}{*}{WorldCapital}
& \multirow{4}{*}{Qwen2.5-7B}
  & \FA   & 122.7 & 150.61 & 182.7 & 250.98 \\
& & \IG   & 118.0 & \bestsuff{149.33} & 180.7 & 255.75 \\
& & \Attn & \bestcomp{127.9} & 153.76 & \bestcomp{184.4} & 245.35 \\
& & \MI   & 119.3 & 151.61 & 177.3 & \bestsuff{244.78} \\
\cmidrule(lr){2-7}
& \multirow{4}{*}{Pythia-6.9B}
  & \FA   & 113.6 & \bestsuff{147.09} & \bestcomp{187.7} & 267.13 \\
& & \IG   & \bestcomp{114.5} & 147.82 & 186.7 & 267.90 \\
& & \Attn & 96.4  & 151.13 & 155.4 & 267.18 \\
& & \MI   & 104.2 & 151.90 & 174.8 & \bestsuff{265.56} \\
\midrule
\multirow{8}{*}{Fakepedia}
& \multirow{4}{*}{Qwen2.5-7B}
  & \FA   & 843.7 & 1047.21 & \bestcomp{1690.2} & 2294.96 \\
& & \IG   & 834.6 & 1052.99 & 1688.5 & 2283.39 \\
& & \Attn & \bestcomp{849.1} & 1028.22 & 1645.8 & 2340.83 \\
& & \MI   & 819.0 & \bestsuff{1019.60} & 1621.8 & \bestsuff{2283.02} \\
\cmidrule(lr){2-7}
& \multirow{4}{*}{Pythia-6.9B}
  & \FA   & \bestcomp{815.6} & \bestsuff{1100.21} & 1683.5 & \bestsuff{2482.10} \\
& & \IG   & 811.1 & 1111.49 & \bestcomp{1685.4} & 2499.73 \\
& & \Attn & 614.1 & 1115.60 & 817.8  & 2493.53 \\
& & \MI   & 691.1 & 1108.89 & 1023.4 & 2500.70 \\
\bottomrule
\end{tabular}
}
\caption{Faithfulness in \textbf{Conflicting} and  \textbf{Double-Conflicting} contexts for two models on two datasets (short-context: World Capital, long-context: Fakepedia). Higher AOPC$_\text{comp}$, lower AOPC$_\text{suff}$ is better. Best entries are \bestcomp{underlined} for comprehensiveness and \bestsuff{bold} for sufficiency. \MI\ and \FA\ are the best AOPC$_\text{suff}$; \Attn\ is the best for AOPC$_\text{comp}$ in Conflicting.}
\label{hcu:tab:faithfulness}
\end{table*}

\part[Highlight Explanations for Faithful NLEs]
     {Utilising Highlight Explanations for Faithful Natural Language Explanations}

\chapter[Graph-Guided NLE Generation]
        {Graph-Guided Textual Explanation Generation Framework}
\chaptermark{Graph-Guided NLE Generation}
\label{chap:NLE_G-Tex}
This chapter consists of the following published paper\footnote{No changes have been made to the scientific content of the originally published paper. The paper is presented in a single-column
format used throughout the thesis, and its references have been merged into the bibliography at the end of the thesis.}:

\begin{enumerate}
    \item \bibentry{yuan-etal-2025-graph}.
\end{enumerate}

\clearpage

\begin{abstract}
Natural language explanations (NLEs) are commonly used to provide plausible free-text explanations of a model's reasoning about its predictions.
However, recent work has questioned their faithfulness, as they may not accurately reflect the model’s internal reasoning process regarding its predicted answer. In contrast, highlight explanations--input fragments critical for the model's predicted answers--exhibit measurable faithfulness.
Building on this foundation, we propose \textbf{\method}, a \textbf{G}raph-Guided \textbf{T}extual \textbf{Ex}planation Generation framework designed to enhance the faithfulness of NLEs.
Specifically, highlight explanations are first extracted as faithful cues reflecting the model's reasoning logic toward answer prediction. They are subsequently encoded through a graph neural network layer to guide the NLE generation, which aligns the generated explanations with the model’s underlying reasoning toward the predicted answer. Experiments on both encoder-decoder and decoder-only models across three reasoning datasets demonstrate that \method improves NLE faithfulness by up to 12.18\% compared to baseline methods. Additionally, \method generates NLEs with greater semantic and lexical similarity to human-written ones.
Human evaluations show that \method can decrease redundant content and enhance the overall quality of NLEs. Our work presents a novel method for explicitly guiding NLE generation to enhance faithfulness, serving as a foundation for addressing broader criteria in NLE and generated text\footnote{\url{https://github.com/lianyiyi/G-TEx}}.\looseness=-1


\end{abstract}

\section{Introduction}
\begin{figure}[ht]
    \centering
    \includegraphics[width=0.9\linewidth]{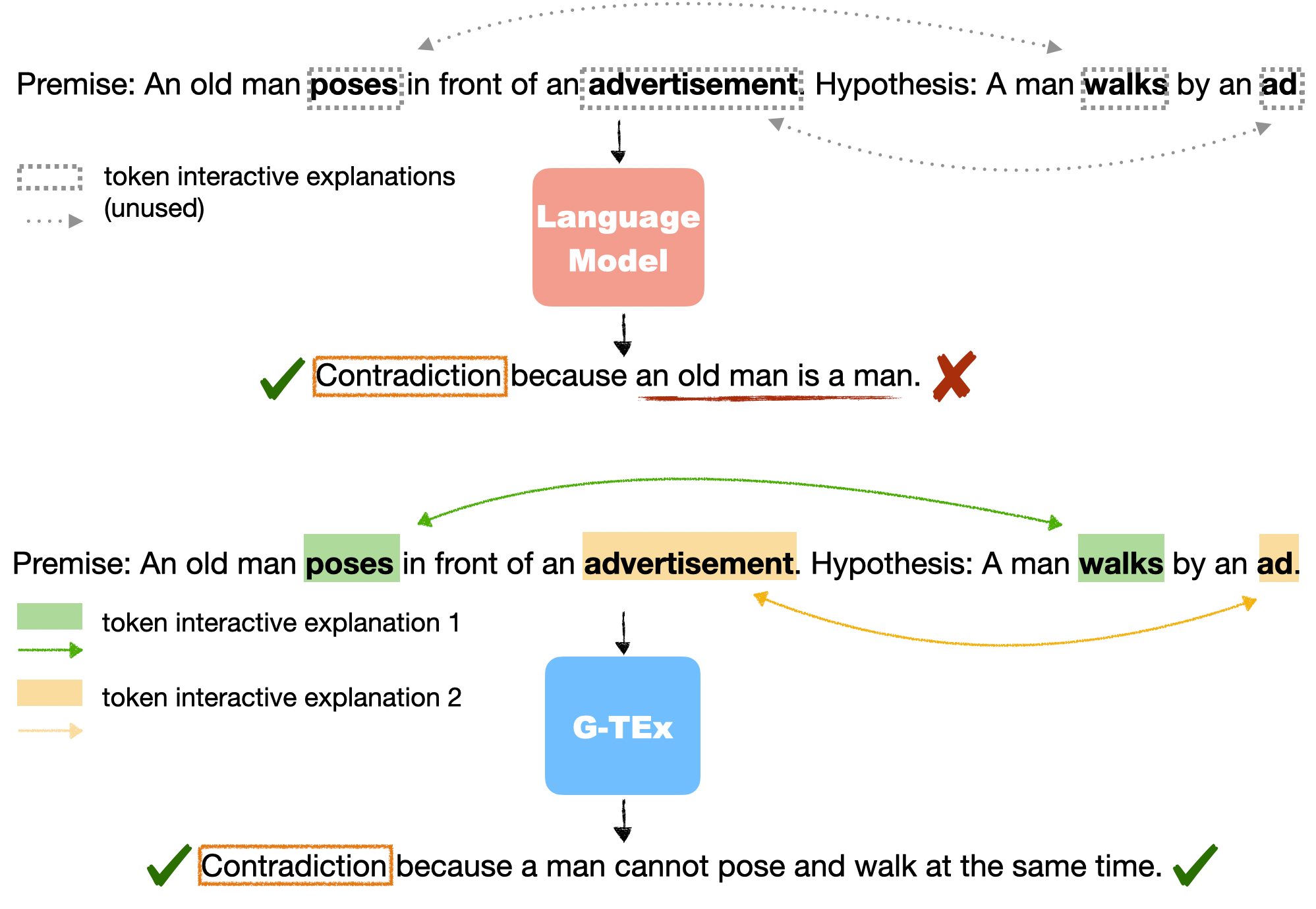}
    \caption{Faithfulness comparison between a self-rationalization model without (top) and with (bottom) the proposed \method. Highlight explanations reveal the model's reasoning behind the predicted label with high faithfulness. Without \method, these important tokens are omitted in the NLE while \method guides the model to incorporate them in the generated NLE.}
    \label{figure:intro_unfaith}
\end{figure}

Natural Language Explanations (NLEs) produce human-understandable texts to explain the model's prediction process \citep{wiegreffe-etal-2021-measuring}. Self-rationalization, where the prediction and the corresponding NLE are generated simultaneously, is a commonly used method for NLE generation, which leads to improved agreement between the generated NLE and the produced prediction~\citep{NEURIPS2018_3e9f0fc9,marasovic-etal-2022-shot}. However, existing work~\citep{kumar2020nile,wiegreffe-etal-2021-measuring} has found that these \textit{NLEs are often unfaithful}, as they may present misleading reasons unrelated to the model’s true decision-making process as illustrated in Figure \ref{figure:intro_unfaith} (top).
This lack of faithfulness undermines the reliability of NLEs in applications where transparency and trust are paramount~\citep{atanasova-etal-2023-faithfulness,lyu2024towards,parcalabescu-frank-2024-measuring}.


Unlike NLEs, highlight explanations reflect the model's reasoning process by identifying tokens or phrases of the input that are crucial to the model's prediction. 
They can be of three types: \textit{highlight token explanations}, \textit{token interactive explanations} and \textit{span interactive explanations} ~\citep{sun-etal-2025-evaluating} (see \S\ref{sec:post_hoc_explanation_and_predicted_label:method} for details).
Though not as plausible as NLEs~\citep{wei-jie-etal-2024-interpretable}, \textit{the faithfulness of highlight explanations is easy to measure and has been substantially improved in existing works}~\citep{sun-etal-2025-evaluating,atanasova-etal-2020-diagnostic}. In this work, we hypothesize that \textit{highlight explanations can be used to improve the faithfulness of NLEs} by using them as explicit cues regarding the important parts of the input that should be present in the generated NLEs. We further hypothesize that as highlight explanations contain concise information about the most important parts of the input, they can further decrease the redundancy of NLEs and improve the overall NLE quality.


Recent efforts to improve the faithfulness of NLEs either rely on external knowledge, crafting prompts or designing the training loss for improving the faithfulness of NLEs directly ~\citep{pmlr-v162-majumder22a,marasovic-etal-2022-shot,chuang-etal-2026-faithlm}. These methods, however, are not targeted at aligning NLEs with a model's inner reasoning but improve their faithfulness only from a model's extrinsic perspective. To address this, and inspired by \citet{yuan-etal-2024-gnnavi} who leverage a Graph Neural Network (GNN) layer to guide the information flow from the input to the generation process, we propose a novel \textbf{G}raph-Guided \textbf{T}extual \textbf{Ex}planation Generation framework (\method) to \textit{enhance the faithfulness of NLEs by explicitly guiding the model’s reasoning with cues derived from the highly faithful highlight explanations}. The graph structure is encoded by a GNN layer, which seamlessly incorporates the highlight explanations into the NLE generation process.
This also allows the model to leverage implicit anchors from the input, improving the generation of explanations.\looseness=-1

\begin{figure*}[!t]
    \centering
    \includegraphics[width=1\textwidth]{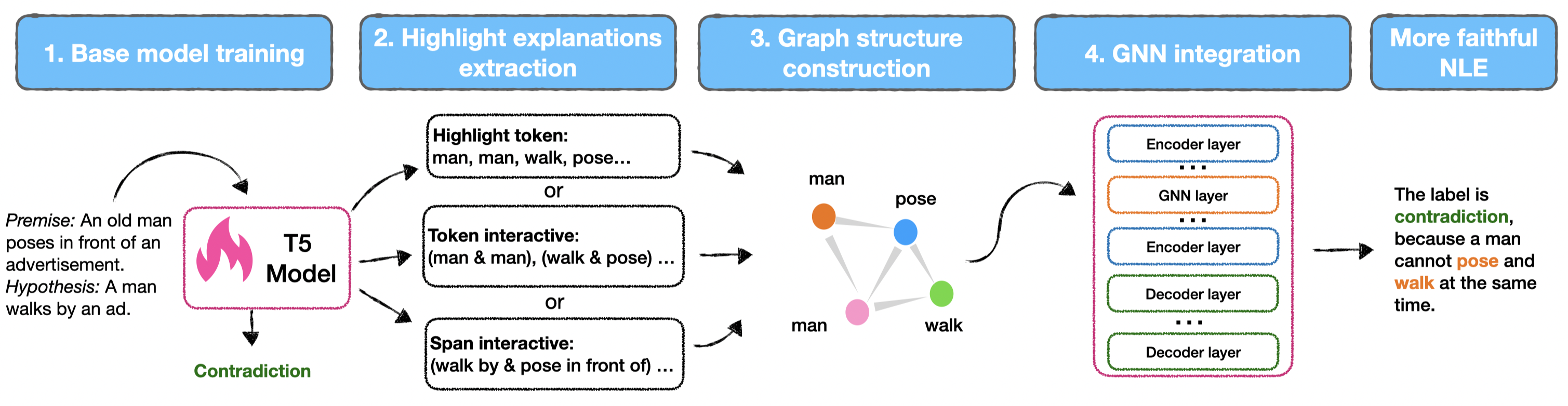}
    \caption{Illustration of our framework \method, which consists of four key steps: (1) We train a base model such as T5 using the task-specific dataset for label prediction (\S\ref{sec:post_hoc_explanation_and_predicted_label:method}). (2) We extract three types of highlight explanations from the trained model (\S\ref{sec:post_hoc_explanation_and_predicted_label:method}). (3) We construct the graph structure based on the highlight explanations (\S\ref{sec:post_hoc_explanation_as_graph:method}). (4) We integrate the graph structure into the model with a GNN layer (\S\ref{sec:Graph_Neural_Network_Layer:method}, \S\ref{sec:integrate_gnn:method}) and fine-tune the overall model for label prediction and NLE generation (\S\ref{sec:self-Rationalization_label:method}).}
    \label{method_overview}
\end{figure*}

As shown in Figure \ref{method_overview}, we first apply a post-hoc attribution method to extract highlight explanations on a fine-tuned model based on its label prediction (\S\ref{sec:post_hoc_explanation_and_predicted_label:method}). Then, we construct a graph with the most important highlight explanations for each instance (\S\ref{sec:post_hoc_explanation_as_graph:method}). A GNN layer is then incorporated to encode the graph within the original self-rationalization model (\S\ref{sec:Graph_Neural_Network_Layer:method}), which is fine-tuned to generate both the final answer prediction and the corresponding NLE simultaneously (\S\ref{sec:self-Rationalization_label:method},\S\ref{sec:integrate_gnn:method}). 

Our findings demonstrate that \method 
substantially improves the faithfulness of NLEs by up to 12.18\% compared to baselines, as evaluated on \texttt{T5} \citep{raffel2020exploring} and \texttt{BART}~\citep{lewis-etal-2020-bart} (see \S\ref{sec:experimental_setting}) using e-SNLI~\citep{camburu2018snli}, ComVE~\citep{wang-etal-2020-semeval} and ECQA ~\citep{aggarwal-etal-2021-explanations} datasets (see \S\ref{sec:results:Faithfulness_Evaluation}). Additionally, \method generates NLEs with enhanced lexical and semantic similarity, as evaluated with SacreBLEU~\citep{post-2018-call} and BERTScore~\citep{bert-score} respectively (see \S\ref{sec:automatic_metric_results}). 
Human evaluations further reveal improvements in decreasing redundancy and enhancing the overall quality of the generated NLEs (see details in \S\ref{Human_Evaluation}). 
Across the different types of highlight explanations, \textit{token and span interactive explanations} are more effective when the input text involves interaction between different parts. However, when the input consistently includes the same instruction, \textit{highlight token explanations} prove to be more beneficial. Overall, our work introduces a novel method for explicitly guiding the NLE generation to improve faithfulness, serving as a stepping stone for addressing additional criteria for NLE and generated text.

\section{Related Work}
\paragraph{Faithfulness of Natural Language Explanations} 
NLEs are coherent free-text explanations about the reasons behind a model's prediction. Most commonly, NLEs are produced with a self-rationalization setup where the model generates both a target task prediction and its NLE~\citep{narang2020wt5,tang2021cognitive,atanasova-etal-2020-generating-fact,liu2024mmi,NEURIPS2023_87e82678,liu2023mgr,liu2023decoupled,liu2024enhancing,liu2025breaking}. As automatically generated NLEs suffer from faithfulness issues~\citep{kumar2020nile,wiegreffe-etal-2021-measuring,atanasova-etal-2023-faithfulness,lyu2024towards}, existing work has explored different ways to improve that.
\citet{pmlr-v162-majumder22a} propose to first select the important parts of the input, then leverage an external commonsense knowledge generative model
to get commonsense knowledge snippets about these highlights, and finally, use the soft representations of the latter for the NLE generation. 
Another line of work focuses on constructing suitable prompts for NLE generation~\citep{marasovic-etal-2022-shot}. Furthermore, \citet{wang2022pinto} propose to prompt the model to generate the NLE and then fine-tune the LM with a counterfactual regularization loss to make the final prediction based on the generated NLE. \citet{chuang-etal-2026-faithlm} employ an estimator to provide faithfulness scores for generated NLEs. These scores and the NLEs are appended to the input and iteratively refined until the faithfulness scores converge. However, neither of these works uses direct cues from the more faithful highlight explanation for the model's prediction to guide the NLE generation, which is the novel contribution of this paper. 
Overall, existing work improves NLE faithfulness by resorting to external knowledge, crafting prompts or altering the generation loss. We claim that these constitute extrinsic signals, which do not directly address the NLEs' desiderata to faithfully reflect a model's inner reasoning. Our proposed method \method directly targets this objective by guiding the generation with cues about the most important parts of the input.\looseness=-1

Existing work has also proposed Chain-of-Thought (CoT) explanations, which reveal the model's intermediate reasoning steps before giving its final answer ~\citep{zhang2022automatic}. 
These explanations can be unfaithful as well ~\citep{NEURIPS2023_ed3fea90,wei-jie-etal-2024-interpretable,lanham2023measuring}. To address this, researchers have leveraged CoT distillation techniques 
to train a more faithful small LM using CoT from the teacher LLM
~\citep{wang2023scott,zhang2024efficient,paul-etal-2024-making}, or have guided the original LLM to generate multiple reasoning chains and choose the most faithful one
~\citep{li2024towards,wei-jie-etal-2024-interpretable}.
Notably, we do not focus on the CoT method for generating NLEs, as it requires specialized training data, such as reasoning chains or step-by-step intermediate explanations leading to the final answer. Moreover, CoT views faithfulness as alignment between the generated explanation and the predicted label, which differs from our focus on faithfulness to the model’s internal reasoning process. 

\paragraph{Highlight Explanations for Model Steering} 
Prior works have found that the model's reasoning capability can be enhanced by human-annotated highlight explanations alongside the original input~\citep{wei2022chain,lampinen2022can}.
\citet{krishna2023post} automate the process of filling the extracted highlights into few-shot templates, which enhances model accuracy across tasks such as CommonsenseQA~\citep{talmor-etal-2019-commonsenseqa}. \citet{zhang2024model} propose iterative prompting, where the model first generates a sentence summarizing the input. This sentence is then matched with the most similar sentence from the input, with similarity calculated by an encoder, to refine the prompt and steer the model to produce an answer more accurately. \citet{bhan2024self} convert highlight explanations into NLEs using a predefined template, which is then employed to prompt the model for more accurate answers. Though they regard the NLE generation as the intermediate step, the faithfulness of these NLEs is not even evaluated.
In contrast, our approach focuses on enhancing the faithfulness of the generated NLEs by integrating highlight explanations directly into the model architecture to guide NLE generation. 


\paragraph{Graph Neural Networks for Natural Language Processing} Graph neural networks (GNNs) are primarily used for graph-related tasks such as drug discovery \citep{han2021reliable,hu2021opengraphbenchmarkdatasets}. An increasing number of researchers are exploring their potential applications in NLP tasks \citep{yasunaga-etal-2021-qa,fei-etal-2021-iterative,lin-etal-2021-bertgcn}. GNNs have been utilized in tasks like graph-to-text generation \citep{gardent-etal-2017-webnlg, yuan-faerber-2023-evaluating} and graph-enhanced question answering \citep{zhang2022greaselm}, typically encoding complex graph and node representations \citep{koncel-kedziorski-etal-2019-text}. \citet{yuan-farber-2024-grasame} leverage GNNs to encode token-level structural information by modifying the self-attention mechanism in language models. Additionally, \citet{yuan-etal-2024-gnnavi} propose a GNN-based method for information aggregation paired with a parameter-efficient fine-tuning approach. Inspired by prior work,
we 
use GNNs to encode the highlight explanations with high faithfulness to the generation process of NLEs.\looseness=-1



\section{Methodology}

In this section, we provide a detailed overview of \method, as illustrated in Figure \ref{method_overview}. We begin by introducing the self-rationalization model in \S\ref{sec:self-Rationalization_label:method}.
In \S\ref{sec:post_hoc_explanation_and_predicted_label:method}, we describe the training of the base model for label prediction and extracting post-hoc highlight explanations as Steps 1 and 2. In Step 3 (\S\ref{sec:post_hoc_explanation_as_graph:method}), we outline the construction of graph structures. Finally, in Step 4, we present the GNN layer (\S\ref{sec:Graph_Neural_Network_Layer:method}) and explain its integration with language models (\S\ref{sec:integrate_gnn:method}).\looseness=-1

\subsection{Overview: Self-Rationalization Model}
\label{sec:self-Rationalization_label:method}
Self-rationalization models jointly generate the task labels and NLEs to explain their reasoning for the predicted answer \citep{wiegreffe-etal-2021-measuring}. We frame this as a text-to-text generation task. Note that we are working with tasks containing two separate parts in the input, e.g., a premise and a hypothesis on the e-SNLI dataset (see more details in \S\ref{sec:dataset}).
Given a sequence of tokens $x = (x_1, \ldots, x_{m+n})$ as input, where the first part of the input contains $m$ tokens and the second part $n$ tokens, the model $M$ generates a label $y_0$ and a sequence of tokens for the NLE $y = y_0\oplus(y_1, \ldots, y_{l})$, where $\oplus$ denotes the concatenation of one label token and $l$ NLE tokens.\footnote{See App. \ref{sec:app:esnli_comve_example} for the input and output examples from the datasets.} 
The text generation task, encompassing both label generation and explanation generation, is implemented by a pre-trained LM with a language modeling head on top. Building on this, we insert a graph structure $\mathcal{G}$ into the standard self-rationalization model (LM) to encode the information from the highlight explanations, particularly for interactions between tokens and spans, resulting in our model $M_{G-TEX}$ (see below). We fine-tune this model by minimizing the cross-entropy loss for the target sequence $y$ following the same process of the standard encoder-decoder transformer model. (see Section \ref{sec:integrate_gnn:method} for details on the encoding process after integrating the GNN layer into the self-rationalization model):
\begin{equation}\footnotesize
    \mathcal{L} = -\sum_{i=1}^{|y|} \log P_\phi(y_i | y_{1:i-1}, x, \mathcal{G}),
\end{equation}
where $P_\phi$ is the LM's generative probability.

\subsection{Post Hoc Highlight Explanation and Predicted Label}
\label{sec:post_hoc_explanation_and_predicted_label:method}
As illustrated in Figure \ref{method_overview}, we begin by training a base model, $M_{base}$, designed solely to predict the label of the input text. From this model, we extract three types of highlight explanations from the input following prior work~\citep{sun-etal-2025-evaluating,choudhury-etal-2023-explaining}. 
These highlights serve as cues revealing the model's reasoning process behind its label predictions.\footnote{We evaluate the faithfulness of highlight explanations in App.~\ref{app:eval_high_exp}.}

Given an input instance $x = (x_1, \ldots, x_{m+n})$, each \textit{highlight token explanation} contains one token $x_i$ and its assigned importance score $a_i$; each \textit{token interactive explanation} $(x_i,x_j)$ consists of two interactive tokens from two separate parts of the input respectively, as well as an importance score $a_{ij}$; each \textit{span interactive explanation} is formed of two spans $(span_i,span_j)$, where $span_i=(x_p,\dots,x_{p+l_1})$ and $span_j=(x_q,\dots,x_{q+l_2})$ are from two separate parts of the input respectively, also with an assigned importance score $a_{span_i,span_j}$, where $p, p+l_1 \in [1,m], q, q+l_2\in [m+1, m+n]$.

\paragraph{Highlight Token Explanation Generation.} Interactions between features in LMs are primarily captured through attention mechanisms~\cite {NIPS2017_3f5ee243}. Previous work shows that highlight explanations extracted by attention-based methods show higher faithfulness than other explainability techniques in several evaluated settings~\cite{sun-etal-2025-evaluating}. Building on this, we use attention weights as the basis for deriving importance scores for all types of highlight explanations.
To retain the unique contributions of individual attention heads -- each designed to focus on specific aspects of the data~\citep{rogers-etal-2020-primer} -- we follow the approach of \citet{choudhury-etal-2023-explaining} to identify the most important attention head for a specific label prediction. We use the final attention layer of the model's decoder, which generates the final token representations used in generation. (see App. \ref{sec:app:post_hoc_explanation_generation} for details). Subsequently, we calculate the importance score $a_i$ for a target token $x_i$ by averaging the self-attention scores assigned to $x_i$ from all other tokens within the input text, following \citet{jain-wallace-2019-attention,sun-etal-2025-evaluating}. The extracted \textit{highlight token explanation} set for instance $x$ is noted as $HT=\{(x_i,a_i)|i \in [1, m+n]\}$.  

\paragraph{Token Interactive Explanation Generation.} Using the most important attention head identified as described above, we calculate the importance score $a_{ij}$ for each \textit{token interactive explanation} by averaging the attention weights between these two tokens $x_i$ and $x_j$ following \citet{clark-etal-2019-bert}. The \textit{token interactive explanation} set for instance $x$ is $TI=\{((x_i,x_j),a_{ij})|i\in [1,m],j\in[m+1,m+n])\}$.

\paragraph{Span Interactive Explanation Generation.}
Since \textit{token interactive explanations} may not convey meaningful information on their own, \citet{choudhury-etal-2023-explaining} suggest using span interactions, which consist of more coherent phrases and are found to be more plausible~\citep{sun-etal-2025-evaluating}. 
Following their approach, we apply the Louvain algorithm~\citep{blondel2008fast} to extract \textit{span interactive explanations} by identifying communities of token interactions.
Tokens are treated as nodes, with the importance scores of token pair interactions used as edge weights. The communities of token interactions are selected to have dense intra-span and sparse inter-span interactions. For each $x$, span pairs $(span_i,span_j)$ are extracted, and the importance score $a_{span_i,span_j}$ for each span pair is computed by averaging the importance scores of the constituent token pairs. The set of generated \textit{span interactive explanations} is denoted as $SI=\{(x_{span_i,span_j},a_{span_i,span_j})|span_i=(x_p,\dots,x_{p+l_1}), span_j=(x_q,\dots,x_{q+l_2})\}$. 
The number of generated span pairs depends on the community detection algorithm and is $< m!*n!$ since only neighboring tokens within the same community can form spans, and spans must come from different parts of the input to form valid pairs. 

\subsection{Post Hoc Highlight Explanations as a Graph}
\label{sec:post_hoc_explanation_as_graph:method}

\begin{figure}[!t]
    \centering
    \includegraphics[width=0.85\linewidth]{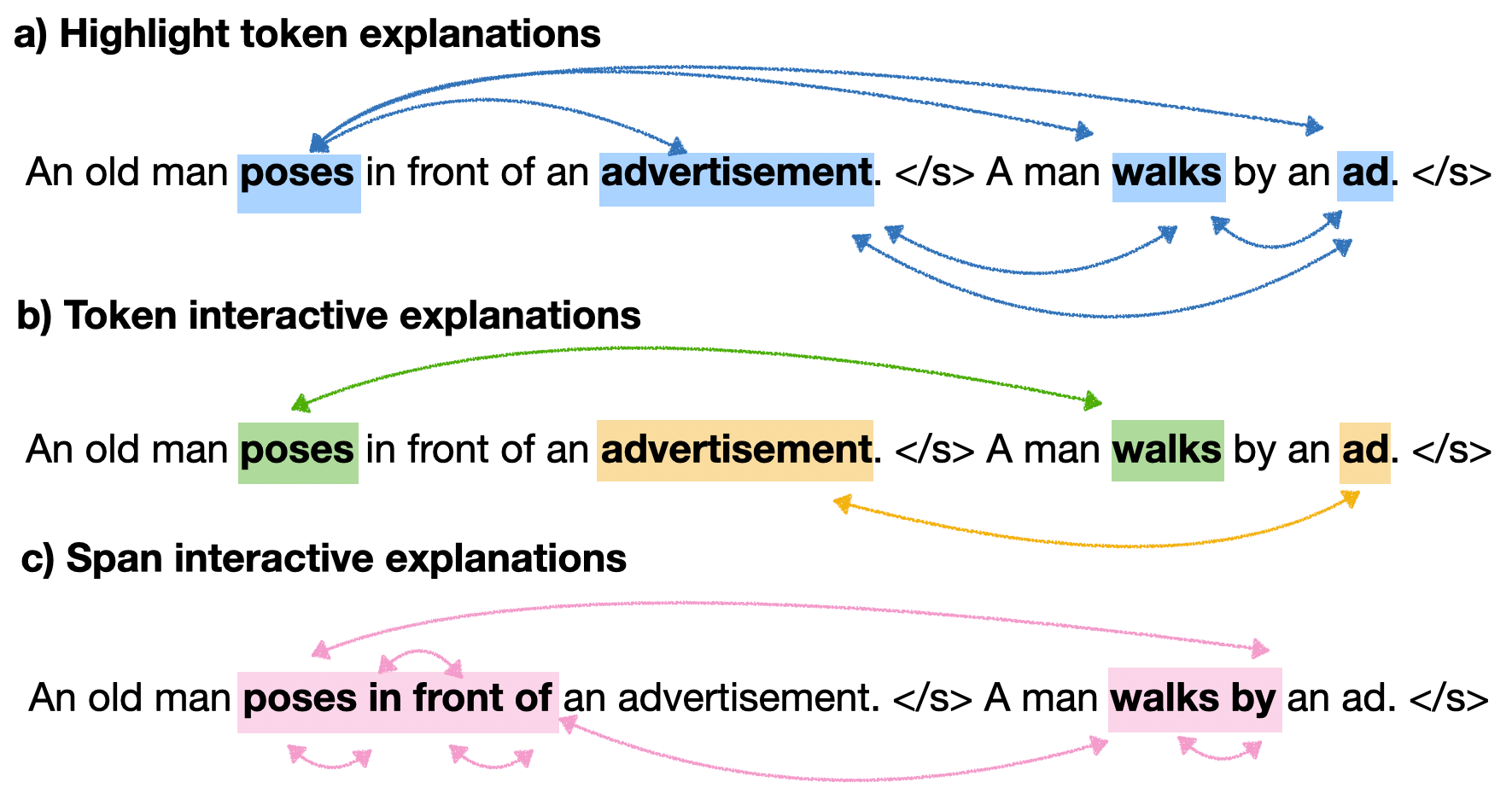}
    \caption{We generate three different types of post-hoc highlight explanations and use them to construct graph structures guiding the NLE generation within our framework. For simplicity, we present only a subset of the explanations for each type.}
    \label{graph_structures}
\end{figure}

We build graph structures based on the three different types of highlight explanations (see Figure \ref{graph_structures}). Notably, we treat each token as a node in the graph structure and assign edges between the extracted tokens. Following \citet{yuan-farber-2024-grasame}, an edge is also assigned to connect the subtokens if a word is tokenized into several subtokens.

\textbf{Highlight Token Explanation} 
We use the importance scores derived in Section \S\ref{sec:post_hoc_explanation_and_predicted_label:method} to select the top-k\% most important highlight token explanations, as less important tokens might introduce noise. Then we assign equally weighted bidirectional edges between these tokens to ensure information flow among them (see Figure \ref{graph_structures}a).


\textbf{Token Interactive Explanations} 
We also select the top-k\% token interactive explanations with the highest importance scores. Then equally weighted bidirectional edges are assigned to connect the tokens within each token interaction (see Figure \ref{graph_structures}b).

\textbf{Span Interactive Explanation}
As only a few spans are extracted from the input text as described in Section \ref{sec:post_hoc_explanation_and_predicted_label:method}, all the interactive spans are used to construct the graph structure. Within a span, all subtokens are connected. Between spans, tokens are connected with each other (see Figure \ref{graph_structures}c).

\subsection{Graph Neural Network Layer}
\label{sec:Graph_Neural_Network_Layer:method}
The GNN layer aggregates information of highlight explanations to model graph and node representations based on the graph structures as introduced in \S\ref{sec:post_hoc_explanation_as_graph:method}.
We define a bidirectional graph $\mathcal{G}$ as a triple $(\mathcal{V}, \mathcal{E}, \mathcal{R})$ with a set of nodes $\mathcal{V} = \{v_1, \ldots, v_n\}$ (one node for each token), a set of relation types $\mathcal{R}$\footnote{We consider only one type of relation: the bidirectional edge between nodes $v$ and $v^{\prime}$, with all edges weighted equally for initialization, note that the edge values will update during fine-tuning}., and a set of edges \( \mathcal{E} \) of the form \( (v, r, v^{\prime}) \) with $v, v' \in \mathcal{V}$, and $r \in \mathcal{R}$.
Each node $v_i$ is associated with a feature vector $h_i$, which represents the hidden states of the $i$-th token in the $l$-th layer. 

The node representations in the GNN layer are updated by aggregating information from neighboring nodes by different aggregation algorithms depending on the chosen GNN architecture. In our work, we employ three most representative and widely used GNN architectures following previous work \citep{yuan-etal-2024-gnnavi,yuan-farber-2024-grasame}: 
Graph Convolutional Network (GCN)~\citep{kipf2017semisupervised}, Graph Attention Network (GAT)~\citep{velivckovic2018graph} and GraphSAGE~\citep{NIPS2017_5dd9db5e}. While GCN aggregates information from neighboring nodes uniformly, GAT introduces attention weights to prioritize and aggregate incoming information.\footnote{Details of the learning processes for GCN and GAT are provided in App. \ref{appendix:gnn_algo}.} GraphSAGE, on the other hand, incorporates information from the current node and its neighboring nodes as follows:
\begin{equation}\footnotesize
  h_v = \sigma\left(W\left(h_v^{(l)}\oplus\text{AGG}(\{h_{v^{\prime}}^{(l)}, \forall v^{\prime} \in N(v)\}) \right)\right)
\end{equation} 
where $h_v$ denotes the updated node representation of $v$, $h_{v^{\prime}}^{(l)}$ is the token representation of its neighbouring nodes from $l$-th layer, $\sigma$ the activation function, $W$ are the trainable parameters of the GNN, $N(v)$ includes all the neighbouring nodes of $v$. The concatenation function $\oplus$ concatenates aggregated information with the node's current representation, and the aggregation function AGG aggregates the information flowing from the neighboring nodes using techniques such as mean, pool, and LSTM.\footnote{Mean aggregation is applied to GraphSAGE in this work.}

\subsection{Integrating GNN in Language Models}\label{sec:integrate_gnn:method}

As illustrated in Figure \ref{method_overview} Step 4, we integrate a GNN layer into the LM by stacking it on top of the \( n \)-th encoder layer. \citet{yuan-etal-2024-gnnavi} demonstrated that incorporating a GNN into LLMs is most effective when placed in the last three-quarters of the layers, following the principles of information flow theory \citep{wang-etal-2023-label}. In line with prior work, we similarly position the GNN layer at the \(\nicefrac{3}{4}\)-th encoder layer. The GNN layer takes token representations from the \( l \)-th encoder layer, processes them along with graph structures derived from highlight explanations, and then forwards the augmented representations $h_v$ to the next encoder layer $l+1$, which can be formulated as: 
\begin{equation}\footnotesize
\tilde{h}^{(l)} = \text{LayerNorm}(h_v + \text{Attention}(h_vW^Q, h_vW^K, h_vW^V))
\end{equation}
\begin{equation}\footnotesize
h^{(l+1)} = \text{LayerNorm}(\tilde{h}^{(l)} + \text{FFN}(\tilde{h}^{(l)}))
\end{equation}
where \(W^Q, W^K, W^V\) are trainable projection matrices for query, key, and value, and FFN denotes a feed-forward network. The rest of the model architecture remains unchanged.


\section{Experiments}

\subsection{Datasets}
\label{sec:dataset}
We use three widely adopted reasoning datasets with human-annotated explanations: \textbf{e-SNLI}~\citep{camburu2018snli}, \textbf{ComVE}~\citep{wang-etal-2020-semeval} and \textbf{ECQA}~\citep{aggarwal-etal-2021-explanations}. \textbf{e-SNLI} extends SNLI with human-annotated explanations for each premise-hypothesis pair, providing both the correct label (entailment, contradiction, or neutral) and a human-annotated NLE for why the label was chosen. \textbf{ComVE} provides natural language explanations identifying which of the two provided statements contradicts common sense. \textbf{ECQA} is a multiple-choice question-answering dataset with human-annotated explanations for each choice. 

In order to explore how different highlight explanations affect faithfulness, we reformulate e-SNLI, ECQA and ComVE into different formats. While the input for e-SNLI and ECQA consists of two distinct sentences, ComVE always includes the same question as the first part of the input (see examples in App. \ref{sec:app:esnli_comve_example}). This distinction is to explore whether the interaction between the two input parts is significant.
\begin{table*}[!htbp]
\centering
\resizebox{\textwidth}{!}{%
\begin{tabular}{@{}c c|cccc|cccc@{}}
\toprule
\multirow{3}{*}{\textbf{Explanation Type}} & \multirow{3}{*}{\textbf{Model}} 
& \multicolumn{4}{c|}{\textbf{e-SNLI}} & \multicolumn{4}{c}{\textbf{ComVE}} \\ 
\cmidrule(lr){3-6}\cmidrule(lr){7-10}
& & \multicolumn{2}{c}{Unfaithfulness(\%$\downarrow$)} & \multicolumn{2}{c|}{Automatic($\uparrow$)}
& \multicolumn{2}{c}{Unfaithfulness(\%$\downarrow$)} & \multicolumn{2}{c}{Automatic($\uparrow$)} \\
& & Counter & Total & SacreBLEU & BERTScore & Counter & Total & SacreBLEU & BERTScore \\
\midrule
\multicolumn{10}{c}{\textbf{T5-based}} \\ 
\midrule
- & \textbf{\texttt{Fine-tuning\textsubscript{base}}} &
  47.70\ \textnormal{\small $\pm$2.31} & 17.68\ \textnormal{\small $\pm$1.94} & 15.430 & 0.894 &
  92.37\ \textnormal{\small $\pm$1.21} & 68.96\ \textnormal{\small $\pm$2.23}
 & 7.634 & 0.876 \\
\midrule
\multirow{2}{*}{Highlight Token} & \textbf{\texttt{Prompt}} &
  43.61\ \textnormal{\small $\pm$2.86} & 14.71\ \textnormal{\small $\pm$1.16} & 15.686 & 0.898 &
  93.25\ \textnormal{\small $\pm$1.19} & 68.90\ \textnormal{\small $\pm$2.61} & 7.592 & 0.876 \\
& \texttt{\textbf{\sage}} (Ours) &
  33.83\ \textnormal{\small $\pm$1.51} & 11.07\ \textnormal{\small $\pm$1.14} & 16.426 & \textbf{0.908} &
  90.53\ \textnormal{\small $\pm$1.40}
 & \textbf{57.48}\ \textnormal{\small $\pm$0.58}
 & \textbf{9.016} & 0.884 \\
\midrule
\multirow{2}{*}{Token Interactions} & \textbf{\texttt{Prompt}} &
  54.36\ \textnormal{\small $\pm$3.11} & 20.60\ \textnormal{\small $\pm$1.81} & 15.478 & 0.898 &
  \textbf{87.39}\ \textnormal{\small $\pm$1.78} & 77.71\ \textnormal{\small $\pm$ 2.06}
 & 7.028 & 0.888 \\
& \texttt{\textbf{\sage}} (Ours) &
  34.27\ \textnormal{\small $\pm$1.63} & 11.00\ \textnormal{\small $\pm$1.66} & \textbf{16.443} & \textbf{0.908} &
  87.47\ \textnormal{\small $\pm$2.21} & 76.94\ \textnormal{\small $\pm$ 2.33}
 & 6.956 & 0.888 \\
\midrule
\multirow{2}{*}{Span Interactions} & \textbf{\texttt{Prompt}} &
 42.86\ \textnormal{\small $\pm$2.20} & 13.19\ \textnormal{\small $\pm$1.95} & 16.031 & 0.899 &
  89.90\ \textnormal{\small $\pm$0.86}  &  79.70\ \textnormal{\small $\pm$2.15}
 & 7.226 & 0.889 \\
& \texttt{\textbf{\sage}} (Ours) &
  \textbf{33.25}\ \textnormal{\small $\pm$2.18} & \textbf{10.08}\ \textnormal{\small $\pm$2.02} & 16.277 & 0.907 &
  89.64\ \textnormal{\small $\pm$0.91} & 76.39\ \textnormal{\small $\pm$3.36}
 & 7.652 & \textbf{0.891} \\
\midrule[1.5pt]
\multicolumn{10}{c}{\textbf{BART-based}} \\ 
\midrule
- & \textbf{\texttt{Fine-tuning\textsubscript{base}}} &
  57.71\ \textnormal{\small $\pm$2.39} & 22.52\ \textnormal{\small $\pm$1.86} & 15.732 & 0.906 &
  91.09\ \textnormal{\small $\pm$1.81} & 70.50\ \textnormal{\small $\pm$1.68}
 & 10.070 & \textbf{0.891} \\
\midrule
\multirow{2}{*}{Highlight Token} & \textbf{\texttt{Prompt}} &
  57.52\ \textnormal{\small $\pm$3.84} & 24.45\ \textnormal{\small $\pm$0.62} & 15.678 & 0.898 &
   90.23\ \textnormal{\small $\pm$2.10} & 68.82\ \textnormal{\small $\pm$2.97}
 & 10.012 & 0.876 \\
& \texttt{\textbf{\sage}} (Ours) &
  \textbf{44.72}\ \textnormal{\small $\pm$4.71} & 14.75\ \textnormal{\small $\pm$2.13} & 16.318 & \textbf{0.909} &
  \textbf{87.91}\ \textnormal{\small $\pm$2.74} & \textbf{58.32}\ \textnormal{\small $\pm$0.81}
 & \textbf{10.552} & 0.884 \\
\midrule
\multirow{2}{*}{Token Interactions} & \textbf{\texttt{Prompt}} &
  47.73\ \textnormal{\small $\pm$3.16} & 19.59\ \textnormal{\small $\pm$1.72} & 15.478 & 0.898 &
  89.80\ \textnormal{\small $\pm$4.54} & 69.43\ \textnormal{\small $\pm$3.14}
 & 7.215 & 0.888 \\
& \texttt{\textbf{\sage}} (Ours) &
 46.88\ \textnormal{\small $\pm$3.34} & 15.68\ \textnormal{\small $\pm$1.75} & \textbf{16.427} & \textbf{0.909} &
  88.15\ \textnormal{\small $\pm$2.47} & 68.08\ \textnormal{\small $\pm$ 2.47}
 & 7.333 & 0.888 \\
\midrule
\multirow{2}{*}{Span Interactions} & \textbf{\texttt{Prompt}} &
  50.98\ \textnormal{\small $\pm$3.72} & 18.34\ \textnormal{\small $\pm$1.70} & 16.027 & \textbf{0.909} &
  95.17\ \textnormal{\small $\pm$1.18}  & 64.35\ \textnormal{\small $\pm$ 0.94}
 & 7.953 & 0.889 \\
& \texttt{\textbf{\sage}} (Ours) &
  45.17\ \textnormal{\small $\pm$3.52} & \textbf{14.64}\ \textnormal{\small $\pm$1.32} & 16.517 & \textbf{0.909} &
  94.29\ \textnormal{\small $\pm$2.57} & 63.76\ \textnormal{\small $\pm$ 2.49} & 7.953 & \textbf{0.891} \\
\bottomrule
\end{tabular}
}
\caption{Overall evaluation results on e-SNLI and ComVE datasets for T5-based and BART-based models, with our \texttt{\textbf{\method}} model using \texttt{\textbf{\sage}}. 
Counter indicates \textit{Counter Unfaith}, Total indicates \textit{Total Unfaith}, with both the mean values and standard deviations reported from 5 runs with different random seeds. The p-values~\cite{wasserstein2016asa} can be found in Appendix \S\ref{sec:appendix:esnli_comve_uncertainty}, Table \ref{tab:uncertainty}. The best performance of each evaluation metric is in bold. See Appendix \S\ref{sec:appendix:ecqa_t5_bart} for results on ECQA dataset and Appendix \S\ref{appendix:overall_gat_gcn_results}, Table \ref{tab:esnli_comve_results_t5_and_bart_gcn_and_gat} for results of our model using \texttt{\textbf{\gat}} and \texttt{\textbf{\gcn}}.
\label{tab:esnli_comve_results_t5_and_bart}}
\end{table*}

\subsection{Experimental Setting}
\label{sec:experimental_setting}
We select two commonly used models for self-rationalization~\citep{raffel2020exploring,narang2020wt5,marasovic-etal-2022-shot,lewis-etal-2020-bart,huang2023chain,yadav2024tox}, \texttt{T5-large} and \texttt{BART-large} as our base models, both of which follow an encoder-decoder architecture.\footnote{See App.~\ref{app:led} for \method's generalizability to the \texttt{LED} model.} For these models, we insert the GNN layer at the \(\nicefrac{3}{4}\)-th encoder layer. 
Our \method is fine-tuned on the training set, with validation performed on the validation set at each epoch. The BLEU score~\citep{papineni-etal-2002-bleu} is used to select the best-performing checkpoint. Further experimental details can be found in App. \ref{sec:appendix:other_experimental_details}.
While our main experiments focus on encoder-decoder models, we also investigate the generalizability of \method to decoder-only model \texttt{Llama-3.2-1B}. The details are summarized in App. \ref{sec:appendix:decoder_only}.


\subsection{Models}
We use two baselines in our experiments to compare against \method:

\textbf{\texttt{Fine-tuning\textsubscript{base}}} We fine-tune the base models \texttt{T5-large} and \texttt{BART-large} on the training set of e-SNLI and ECQA for self-rationalization.

\textbf{\texttt{Prompt}} 
We adopt different strategies to construct the prompt for different types of highlight explanations. Specifically, to incorporate highlight token explanations as part of the input, we concatenate the template, \textit{``The most important tokens are: token\textsubscript{1}, token\textsubscript{2}, token\textsubscript{3}, ...''} to the end of the input sentence (this serves as a fully connected graph). For token interaction and span interaction explanations, we concatenate \textit{``The most important token/span interactions are: <token\textsubscript{1}/span\textsubscript{1}, token\textsubscript{2}/span\textsubscript{2}>; <token\textsubscript{3}/span\textsubscript{3}, token\textsubscript{4}/span\textsubscript{4}>; ...''} (this serves as graph structures in Figure \ref{graph_structures}b and \ref{graph_structures}c) with the original input sequence. Then, we train the model accordingly. The same highlight explanations are used as those in \method.

\textbf{\texttt{\method}} For our approach, we utilize the encoder-decoder model \texttt{T5-large} and \texttt{BART-large} as the base models in the main experiments, and insert a GNN layer after the \(\nicefrac{3}{4}\)-th encoder layer. This GNN layer injects the structured information from the highlight explanations. We experiment with three distinct types of GNN architectures, which we denote as \texttt{\textbf{\gcn}}, \texttt{\textbf{\gat}}, and \texttt{\textbf{\sage}}, representing Graph Convolutional Networks, Graph Attention Networks, and GraphSAGE, respectively (see \S3.4).

\section{Evaluation}

We conduct a comprehensive evaluation of the models, using a faithfulness test, automatic metrics 
and human assessment on multiple dimensions\footnote{The results and analysis of human evaluation are presented in App. \ref{Human_Evaluation}}. As for the label predictions, \method achieves results that are better or comparable to the baselines. We report an overview of the label prediction performance in Table \ref{tab:_acc}, App. \ref{appendix:acc}. 

\subsection{Faithfulness Evaluation}
\label{sec:results:Faithfulness_Evaluation}
To assess the faithfulness of the generated NLEs, we apply the counterfactual faithfulness test from \citet{atanasova-etal-2023-faithfulness}. This method involves inserting random adjectives in front of nouns of the original input, resulting in multiple perturbed instances. If the model’s prediction changes, the newly generated NLE should include the inserted word; otherwise, the original NLE is unfaithful as it is potentially misaligned with the model's reasoning. Note that the unchanged label provides no relevant information about the faithfulness of the NLE. See details in App. \ref{sec:appendix:faithfulness_evaluation_details}.


Following \citet{atanasova-etal-2023-faithfulness}, we apply this test on e-SNLI, ComVE and ECQA dataset, calculating: 
(1) the percentage of instances where, for at least one altered input, the inserted word does not appear in the new NLE across instances with label change (\textit{Counter Unfaith}); and (2) the proportion of these unfaithful instances across all instances (\textit{Total Unfaith}).

\subsubsection{Results} 
As shown in Table \ref{tab:esnli_comve_results_t5_and_bart}, we present results on e-SNLI and ComVE as representative datasets for NLI and commonsense QA, respectively.\footnote{ECQA results are presented in App. \ref{sec:appendix:ecqa_t5_bart}.}
Our \method\footnote{\method refers to \sage throughout this section as GraphSAGE demonstrates superior performance in modeling text-based graph structures according to previous work \citep{yuan-farber-2024-grasame}. The results of other \method models and the discussion across all GNN variants can be found in App. \ref{appendix:overall_gat_gcn_results}.}
with \texttt{T5} as the base model leads up to 9.60\% decrease in \textit{Total Unfaithful} on e-SNLI (20.60\% vs. 11.00\% with token interactive explanations) and up to 11.48\% on ComVE (68.96\% vs. 57.48\% with highlight tokens) compared to the \texttt{Fine-tuning\textsubscript{base}} and \texttt{Prompt}. 
Similarly, \method with \texttt{BART} as the base model leads up to a 9.70\% decrease in \textit{Total Unfaithful} on e-SNLI (24.45\% vs. 14.75\% with highlight explanations) and up to 12.18\% decrease on ComVE (70.50\% vs. 58.32\% with highlight explanations). 
While \method with \texttt{T5} slightly underperforms the prompt baseline on ComVE with \textit{token interactive explanations}, overall, \textbf{our method outperforms all baselines in counterfactual unfaithfulness and total unfaithfulness}.

Across the different highlight explanation types, different datasets yield different results. On the e-SNLI dataset, \textit{span interactive explanations} produce more faithful NLEs with T5-based models (10.08\% \textit{Total Unfaith}). 
For the e-SNLI task, the input text consists of two parts, namely the premise and the hypothesis, and interactive explanations between these parts are of paramount importance in indicating the reasoning process of the models. \textbf{Thus, \textit{token interactive }and \textit{span interactive explanations} tend to improve faithfulness more effectively than \textit{highlight token explanations}}. This aligns with previous work showing that these highlight explanations offer higher faithfulness
in recovering a model's prediction~\citep{sun-etal-2025-evaluating}. 

However, \textit{highlight token explanations} also show significant benefits when the task input consists of the same instruction/first part. As the first part of the input for ComVE is formulated as the same question, the second part of the input becomes especially important in distinguishing the input text for the models. The results on ComVE indicate that \textit{highlight token explanations} yield the lowest \textit{Total Unfaith} for both T5- and BART-based \method (57.48\% and 58.32\%, respectively). \textbf{Thus, \textit{highlight token explanations} can improve the faithfulness when the interaction between two parts of the input is less critical.} 

Our findings demonstrate that while all highlight explanations are significantly important, their utility depends on the task. When the input text involves interaction between different parts, \textit{token} and \textit{span interactive explanations} are more useful. However, when the input consistently includes the same instruction, \textit{highlight token explanations} are more effective. Nonetheless, regardless of the task, the results again verify that \method effectively leverages different types of highlight explanations for NLE generation, leading to more faithful NLEs. 





\subsection{Automatic Metrics for Similarity between NLEs and Golden explanations}
\label{sec:automatic_metric_results}
To assess the alignment of generated NLEs with human-written ones, we measure the similarity between them and the golden human-annotated explanations. A similarity with human-written explanations is used in existing work to indicate how plausible the generated NLEs would appear to end users ~\citep{sun-etal-2025-evaluating}. We employ automatic evaluation metrics \textbf{SacreBLEU}~\citep{post-2018-call} and \textbf{BERTScore}~\citep{bert-score} to capture both lexical and semantic similarity.\footnote{In addition to SacreBLEU and BERTScore, results for other automatic metrics are provided in App. \ref{sec:appendix:t5_other_automatic_metrics}.}

As shown in Table \ref{tab:esnli_comve_results_t5_and_bart}, the automatic evaluation results demonstrate that \method generates NLEs of higher alignment with human-written explanations in terms of lexical and semantic similarity on the e-SNLI dataset, outperforming the \texttt{Fine-tuning\textsubscript{base}} and \texttt{Prompt}. Across all explanation types, \method consistently achieves higher SacreBLEU scores, such as 16.443 for \method with the \textit{token interactive explanation} setting, and better BERTScores, such as 0.909 across most BART-based methods. Regarding the ComVE dataset, \method also generates NLEs with higher SacreBLEU and BERTScore. For BART-based \method, the highest SacreBLEU is 10.552 achieved with \method with \textit{highlight token explanations}. 
\textbf{These results demonstrate that our models generate explanations with improved alignment with human explanations}. Furthermore, they confirm that interactive explanations are more effective for e-SNLI, while highlight token explanations are more beneficial for ComVE, due to the distinct structure of their inputs.

\section{Conclusion}

In this work, we propose \method, a novel framework that incorporates the reasoning process of models to enhance faithfulness in NLEs. \method allows for integrating various types of highlight explanations through a GNN layer within language models.
Evaluated via faithfulness tests, automatic metrics, and human evaluation on three reasoning datasets, \method demonstrates consistent improvements in faithfulness, alignment with human-annotated explanations, and reduced redundancy. 
Our results show that the benefits of different highlight explanations depend on task formulation: \textit{token} and \textit{span interactive explanations} work best for tasks requiring input interaction, while \textit{highlight token explanations} are more effective when interactions are less critical.
These findings highlight the potential of \method as an interpretable framework that embeds the reasoning process of language models as a graph structure to improve model faithfulness. 

\section*{Limitations}
Our work proposes a novel graph-guided framework for natural language explanation generation, utilizing highlight explanations in the form of highlight tokens, token interactives, and span interactives. 
While \method improves the models' faithfulness constantly, we acknowledge several limitations in our approach.

Firstly, 
due to limited computational resources, we chose \texttt{T5-large} and \texttt{BART-large} as the main models to fine-tune for NLE generation. Their established reasoning capabilities and relatively lightweight nature make them well-suited for our experimental setup. However, we encourage future work to explore how model scalability affects the quality of generated NLEs.


Secondly, while \method leverages the reasoning process of the models and offers a more transparent and interpretable framework, the internal mechanisms of the GNN layer remain unexplored in this study. Moreover, we use specific graph types to construct the highlight explanations, assigning equal weights to the edges between nodes. Future work could explore weighted edges and alternative graph structures to encode highlight explanations.

Thirdly, while we choose the attention-based methods as the foundation to extract highlight explanations due to their higher faithfulness on ECQA and e-SNLI dataset~\citep{sun-etal-2025-evaluating}, it is important to acknowledge other important explainability techniques, such as perturbation-based attribution e.g., Shapley~\citep{lundberg2017unified}, Integrated Gradients~\citep{sundararajan2017axiomatic} and Saliency Map~\citep{feldhus2022constructing}. It is worth exploring how the highlight explanations generated by different explainability techniques impact the quality of generated NLEs on broader datasets. We leave this exploration for future work.

Lastly, we evaluate the quality of NLEs generated by our model using three reasoning datasets, e-SNLI (NLI task), ComVE and ECQA (commonsense QA task). 
As more datasets meeting these criteria become accessible in the future, we encourage further exploration of our method in additional domains.



\section*{Acknowledgements}

$\begin{array}{l}\includegraphics[width=1cm]{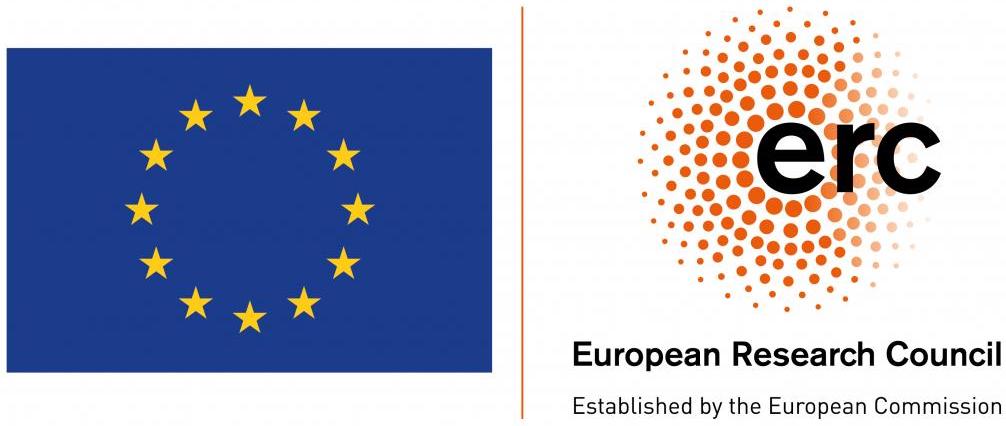} \end{array}$ 
This research was co-funded by the European Union (ERC, ExplainYourself, 101077481), by the Pioneer Centre for AI, DNRF grant number P1, by The Villum Synergy Programme, by the ScaDS.AI, as well as by the German Federal Ministry of Research, Technology and Space (BMFTR) via the Software Campus project (01|S23070).
Views and opinions expressed are however those of the author(s) only and do not necessarily reflect those of the European Union or the European Research Council. Neither the European Union nor the granting authority can be held responsible for them. We thank the anonymous reviewers for their helpful suggestions, and we extend special thanks to our student assistant Mario Tawfelis for his support.


\section{Appendix}

\subsection{Post Hoc Explanation Generation Details}
\label{sec:app:post_hoc_explanation_generation}

For each attention head $j$ regarding generating token $k$, When the contribution of input token $i$ $c_{ji}$ is positive, the larger the weight $w_{ji}$, the more important of input token $i$ to $k$. We aggregate the importances for generating $k$ from all input tokens in attention head $j$ as the indication of the overall importance of attention head $j$. 

\subsection{Raw Running Time of Exacting Highlight Explanations}\label{sec:appendix:time_extract_explanation}

We report the raw running time for extracting highlight explanations on the test set of e-SNLI using T5-based model in Table \ref{tab:time_cost}. Although the span interactive explanation has the longest runtime, it only requires 14 ms to extract explanations for an instance with the longest token range. While extracting explanations adds some computational time, it is not prohibitive for practical use.


\begin{table*}[h]
    \centering
    \resizebox{\textwidth}{!}{%
    \begin{tabular}{lccc}
        \toprule
        \textbf{Explanation Type} & \textbf{[5, 20) Tokens} & \textbf{[20, 40) Tokens} & \textbf{[40, 69] Tokens} \\
        \midrule
        \multicolumn{4}{c}{\textbf{Average Time Cost per Instance (ms)}} \\
        \midrule
        Highlight Token Explanation  & 0.7382  & 1.2903  & 2.5249  \\
        Token Interactive Explanation & 0.3924  & 0.8071  & 1.8725  \\
        Span Interactive Explanation  & 2.5501  & 5.8975  & 14.3293 \\
        \midrule
        \multicolumn{4}{c}{\textbf{Number of Instances in Each Token Range}} \\
        \midrule
        & 4,546  & 5,068  & 192  \\
        \bottomrule
    \end{tabular}}
    \caption{Average time cost (in milliseconds) and instance counts across different token length ranges for three types of highlight explanation extraction using a T5-based model on the e-SNLI test set.}
    \label{tab:time_cost}
\end{table*}

\subsection{Evaluation on Highlight Explanations}\label{app:eval_high_exp}

To validate the faithfulness of our extracted highlight explanations as cues for the model’s reasoning, we leverage two metrics: Comprehensiveness and Sufficiency, following \citet{sun-etal-2025-evaluating,deyoung-etal-2020-eraser}. Comprehensiveness measures whether the model’s prediction changes when the highlight explanations are gradually masked out, whereas Sufficiency assesses whether the prediction changes when only the highlight explanations are provided in the input.

These existing works have evaluated the faithfulness of highlight explanations and found that attention-based explanations are the most faithful. We therefore employ these explanations in this work. To further validate the faithfulness of the explainability approaches in our framework, with the T5 model, we conduct the following evaluation, serving as sufficient evidence of the reliability of the employed highlight explanations.

As a simple validation for the reliability of the highlight explanations, we compare the faithfulness of the employed explainability techniques with random baselines. For a fair comparison, we match the number of tokens in our explanations with those in the randomly selected baseline tokens for each explanation type. For example, the "Random Baseline for Highlight Tokens" in Table \ref{tab:eval_highlight_exp} uses the same number of tokens as our highlight token explanations.

Table \ref{tab:eval_highlight_exp} reports the faithfulness evaluation for three types of highlight explanations on the T5 model, e-SNLI dataset, where higher Comprehensiveness ($\uparrow$) and lower Sufficiency ($\downarrow$) scores indicate better faithfulness. Our results show that the extracted highlight explanations are consistently more faithful than the random baselines, validating their role as a reliable foundation for our \method framework for NLE generation.

\begin{table*}[htbp!]
\centering
\resizebox{\textwidth}{!}{%
\begin{tabular}{lcc}
\toprule
\textbf{Highlight Explanation Type} & \textbf{Comprehensiveness (↑)} & \textbf{Sufficiency (↓)} \\
\midrule
Highlight Tokens & 3.809 & 4.848 \\
Randomly Selected Highlight Tokens & 2.559 & 5.301 \\
\midrule
Token Interactions & 4.730 & 4.904 \\
Randomly Selected Token Interactions & 3.877 & 6.012 \\
\midrule
Span Interactive Explanation & 4.819 & 1.003 \\
Randomly Selected Span Interactions & 4.193 & 2.615 \\
\bottomrule
\end{tabular}
}
\caption{Faithfulness evaluation for different types of highlight explanations on the T5 model, e-SNLI dataset.}\label{tab:eval_highlight_exp}
\end{table*}

\subsection{Example of Self-Rationalization Task}\label{sec:app:esnli_comve_example}

One example of the e-SNLI dataset~\citep{camburu2018snli} will be:
\begin{quote}
\raggedright
\textit{
\textbf{Input}: "Premise: A woman is asleep at home. Hypothesis: A woman with a red scarf is giving a shushing sign to the camera in front of shelves of books."\\
\textbf{Output}: "Contradiction. The woman cannot be giving a sign and asleep at the same time."
}
\end{quote}

where \textit{Contradiction} is the predicted label, and the text explains the reason for the prediction.

And we reformulate ComVE dataset~\citep{wang-etal-2020-semeval} as:
\begin{quote}
\raggedright
\textit{
\textbf{Input}: "Which statement of the two is against common sense? \\
1. when it is hot humidity forms \\
2. when it rains humidity forms"\\
\textbf{Output}: "2. Water makes humidity, not temperature."
}
\end{quote}

where \textit{2} is the index of the prediction, and the text explains the reason why it is against common sense.

Meanwhile, one example of ECQA~\citep{aggarwal-etal-2021-explanations}:
\begin{quote}
\raggedright
\textit{
\textbf{Input}: "The student was contemplating the problem, that's when he made the what that led him to the answer? \\
action, discovery, reflection, deciding, thinking\\
\textbf{Output}: "discovery. Contemplating on the problem, the student made the discovery or finding that led him to the answer. Contemplating and thinking deeply about the problem may or may not lead to action. Reflection is contemplating of thinking about oneself and not the problem. Deciding is contemplating choice and wrong decisions don’t lead to answer. Thinking won’t necessarily lead to the answer."
}
\end{quote}

where \textit{discovery} is the predicted answer, and the text explains the reason why it is correct and why the others are wrong.

\subsection{Aggregation Algorithms of GCN and GAT}\label{appendix:gnn_algo}

The learning process of GCN is formulated as:
\begin{equation}\footnotesize
  h_v = \sigma\left(W\sum_{v^{\prime} \in N(v)} \frac{h_{v^{\prime}}^{(l)}}{\lvert N(v)\rvert}\right)
\end{equation} 

where $h_v$ denotes the updated node representation of $v$, $h_{v^{\prime}}^{(l)}$ is the token representation of its neighbouring nodes from $l$-th layer, $\sigma$ the activation function, $W$ are the trainable parameters of the GNN, $N(v)$ includes all the neighbouring nodes of $v$. 

Unlike the average over all neighbouring nodes in GCN, GAT learns an attention weight $\alpha$ for every neighbouring node:
\begin{equation}\footnotesize
  h_v = \sigma\left(\sum_{v^{\prime} \in N(v)} \alpha_{vv^{\prime}} Wh_{v^{\prime}}^{(l)}\right)
\end{equation} 

\subsection{Performance for Label Prediction}\label{appendix:acc}

We present the performance of all baselines and \method for the label prediction task in Table \ref{tab:_acc}. \method consistently outperforms the baselines on both the e-SNLI and ECQA datasets.

As shown in Table \ref{tab:_acc}, we present our \method models’ performance in answer prediction, where the GNN layer is jointly fine-tuned with the base model alongside all baseline models. It is evident that the \method model achieves better or comparable accuracy to the baseline models, ensuring that \method does not sacrifice answer accuracy while increasing NLE faithfulness.

\begin{table*}[!htbp]
\centering
\scalebox{0.95}{
\begin{tabular}{c c c c c}
\toprule
\multicolumn{2}{c}{\textbf{Method}} & \textbf{Acc\textsubscript{e-SNLI}} & \textbf{Acc\textsubscript{ECQA}} & \textbf{Acc\textsubscript{ComVE}} \\
\midrule
\multicolumn{5}{c}{\textbf{T5-large}} \\
\midrule
\multicolumn{2}{c}{\textbf{\texttt{Fine-tuning\textsubscript{base}}}} & 84.50 & 61.56 & 89.92\\
\cmidrule(lr){1-5}
\multirow{4}{*}{Highlight Tokens} & \textbf{\texttt{Prompt}} & 86.16 & 60.98 & 88.05\\
 & \textbf{\texttt{\gcn}} & 89.79 & 59.87 & 90.86\\
 & \textbf{\texttt{\gat}} & 89.42 & 60.22 & 91.08\\
 & \textbf{\texttt{\sage}} & 89.78 & 60.37 & \textbf{92.43}\\
\cmidrule(lr){1-5}
\multirow{4}{*}{Token Interactions} & \textbf{\texttt{Prompt}} & 86.02 & 57.17 & 90.48\\     
 & \textbf{\texttt{\gcn}} & 89.88 & \textbf{62.23} & 90.97\\
 & \textbf{\texttt{\gat}} & 89.93 & 61.76 & 90.14\\
 & \textbf{\texttt{\sage}} & 89.94 & 61.25 & 89.76\\ 
\cmidrule(lr){1-5}
\multirow{4}{*}{Span Interactions} & \textbf{\texttt{Prompt}} & 88.92 & 59.14 & 88.14\\
 & \textbf{\texttt{\gcn}} & 89.76 & 59.62 & 89.06\\
 & \textbf{\texttt{\gat}} & 89.10 & 59.02 & 90.36\\
 & \textbf{\texttt{\sage}} & \textbf{89.98} & 58.62 & 89.76\\
\midrule
\multicolumn{5}{c}{\textbf{BART-large}} \\
\midrule
\multicolumn{2}{c}{\textbf{\texttt{Fine-tuning\textsubscript{base}}}} & 85.29 & 56.91 & 91.57\\
\cmidrule(lr){1-5}
\multirow{4}{*}{Highlight Tokens} & \textbf{\texttt{Prompt}} & 81.55 & 42.21 & 91.47\\
 & \textbf{\texttt{\gcn}} & \textbf{91.04} & 41.82 & 92.17\\
 & \textbf{\texttt{\gat}} & 90.60 & 50.50 & 92.15\\
 & \textbf{\texttt{\sage}} & 91.03 & 52.73 & \textbf{92.67}\\ 
\cmidrule(lr){1-5}
\multirow{4}{*}{Token Interactions} & \textbf{\texttt{Prompt}} & 90.42 & 54.59 & 90.48\\
 & \textbf{\texttt{\gcn}} & 90.18 & \textbf{58.02} & 91.76\\
 & \textbf{\texttt{\gat}} & 89.52 & 55.50 & 86.51\\
 & \textbf{\texttt{\sage}} & 89.44 & 52.46 & 91.77\\ 
\cmidrule(lr){1-5}
\multirow{4}{*}{Span Interactions} & \textbf{\texttt{Prompt}} & 90.35 & 56.38 & 89.13\\
 & \textbf{\texttt{\gcn}} & 90.91 & 51.53 & 91.77\\
 & \textbf{\texttt{\gat}} & 91.03 & 56.94 & 91.06\\
 & \textbf{\texttt{\sage}} & 90.79 & 44.41 & 92.17\\
\bottomrule
\end{tabular}
}
\caption{Overview of model accuracy on e-SNLI, ECQA and ComVE datasets. \method achieves results that are better or comparable to the baselines. The models with the best performance are highlighted in bold.}
\label{tab:_acc}
\end{table*}

\subsection{Experimental Details}
\label{sec:appendix:other_experimental_details}
The number of incorporated GNN layers is 1. Final results are reported on the test set with beam search set to 3. We set $k=30$ to take the top 30\% most important highlight explanations. Training is conducted on four NVIDIA A100-SXM4-40GB GPUs, utilizing AdamW \citep{loshchilov2018decoupled} as the optimizer. The learning rate is set to \texttt{3e-4} for both the baselines and \method after grid search. And beam search is set to 3 for the text generation. We use the original train, dev, and test splits for model fine-tuning across all the datasets.

\subsection{Model Size}\label{sec:appendix:model_size}

Table \ref{tab:num_params} shows the number of trainable parameters comprising the baselines and \method, as well as the training time for one epoch under the same configuration (batch size, optimizer, learning rate, etc.). Notably, the models incorporating GNN only have approximately up to 0.28\% more parameters than the baseline models \texttt{T5} and 0.24\% more parameters than the baseline models \texttt{BART}. Overall, the training time for different methods varies by only a few seconds. 
\begin{table}[h]
    \centering
    \footnotesize
    \begin{tabular}{llll}
        \hline
        \textbf{Method} & \textbf{Param\textsubscript{T5}} & \textbf{Param\textsubscript{BART}} & \textbf{Time\textsubscript{T5}} \\
        \hline
        Fine-tuning & 737M  & 406M & 13:51\\
        Prompt      & 737M  & 406M & 14:23\\
        \gcn     & 738M  & 407M & 13:41\\
        \gat     & 738.1M & 407M & 13:42\\
        \sage    & 739.1M & 407M & 13:49\\
        \hline
    \end{tabular}
    \caption{Number of parameters and training time for different methods using \texttt{T5} and \texttt{BART}.}
    \label{tab:num_params}
\end{table}

\subsection{Faithfulness Evaluation Method}
\label{sec:appendix:faithfulness_evaluation_details}
Following ~\citet{atanasova-etal-2023-faithfulness}, we conduct the counterfactual evaluation to assess the faithfulness of the generated NLEs. Specifically, given an input instance $x$ with the model’s original answer $y_0$ and its corresponding NLE tokens $[y_1, \dots, y_{l}]$ (see \S\ref{sec:self-Rationalization_label:method}), we insert a word $x_c$ into $x$, forming a new input $x'$. To ensure the coherence of $x'$,  we only insert random adjectives before nouns. For each original input $x$, we generate candidate insertions at 4 random positions, with 4 candidates per position, resulting in 16 perturbed inputs $x'$ for each instance.  If the model’s prediction changes ($y_0' \neq y_0$), the newly generated NLE should include the inserted word, i.e., $x_c \in [y_1', \dots, y_{p+q}']$; otherwise, the original NLE is unfaithful as it is potentially misaligned with the model's reasoning. Note that the unchanged label provides no relevant information about the faithfulness of the NLE.

\subsection{Overall Explanation Evaluation Results on e-SNLI and ComVE Dataset of \textbf{\texttt{\method}} using \textbf{\texttt{\gat}} and \textbf{\texttt{\gcn}}}
\label{appendix:overall_gat_gcn_results}

As shown in Table \ref{tab:esnli_comve_results_t5_and_bart_gcn_and_gat}, we also report the results of our models \textbf{\texttt{\method}} using  \textbf{\texttt{\gat}} and \textbf{\texttt{\gcn}}. 

Regarding faithfulness, almost all of our models outperform all the baseline models on both datasets, achieving improvements of up to 17.18\% with the T5-based \textbf{\texttt{\gcn}} on the ComVE dataset,
which demonstrates our approach's effectiveness in enhancing the faithfulness of NLEs. 

Across different highlight explanation types, \textbf{\textit{token interactive explanations}} consistently achieve the best faithfulness results on the e-SNLI dataset, regardless of the base model architecture. In contrast, on the ComVE dataset, \textbf{\textit{highlight token explanations}} consistently demonstrate the highest faithfulness, highlighting the influence of dataset characteristics on the advantages of different explanation types in enhancing NLE faithfulness. For example, on the ComVE dataset, where the first part of the input is a general question in which the statement of the two is against common sense, the simple interaction between the tokens/spans from the question and the statements might be less informative than simply selecting the important tokens from the statements. \textbf{This suggests that the choice of highlight explanation types to enhance NLE quality, particularly in terms of faithfulness, should be carefully tailored to the specific characteristics of the dataset.
}

Regarding the similarity between the generated NLEs and the golden ones, as measured by automatic metrics, all the NLEs generated by our method on both datasets achieve equal or higher performance than the baselines. Among the different highlight explanation types, NLEs guided by \textbf{\textit{highlight token explanations}} most frequently achieve the highest similarity with the golden ones, both lexically and semantically.

Among the different GNN variants of our \textbf{\texttt{\method}} method, \textbf{\texttt{\gat}}, \textbf{\texttt{\gcn}}, and \textbf{\texttt{\sage}}, there is no consistent trend indicating that any particular GNN layer consistently outperforms the others in improving the faithfulness or the similarity of the NLEs to the gold explanations. 

\begin{table*}[!htbp]
\centering
\resizebox{\textwidth}{!}{%
\begin{tabular}{@{}c c|cccc|cccc@{}}
\toprule
\multirow{3}{*}{\textbf{Explanation Type}} & \multirow{3}{*}{\textbf{Model}} 
& \multicolumn{4}{c|}{\textbf{e-SNLI}} & \multicolumn{4}{c}{\textbf{ComVE}} \\ 
\cmidrule(lr){3-6}\cmidrule(lr){7-10}
& & \multicolumn{2}{c}{Unfaithfulness(\%$\downarrow$)} & \multicolumn{2}{c|}{Automatic($\uparrow$)}
& \multicolumn{2}{c}{Unfaithfulness(\%$\downarrow$)} & \multicolumn{2}{c}{Automatic($\uparrow$)} \\
& & Counter & Total & SacreBLEU & BERTScore & Counter & Total & SacreBLEU & BERTScore \\
\midrule
\multicolumn{10}{c}{\textbf{T5-based}} \\ 
\midrule
- & \textbf{\texttt{Fine-tuning\textsubscript{base}}} &
  47.08 & 16.89 & 15.430 & 0.894 &
  \textbf{87.17} &73.73  & 7.634 & 0.876 \\
\midrule
\multirow{3}{*}{Highlight Token} & \textbf{\texttt{Prompt}} &
  42.04 & 14.11 & 15.686 & 0.898 &
  87.04 & 74.18 & 7.592 & 0.876 \\
& \texttt{\textbf{\gat}} (Ours) &
  35.92 & 11.28 & 16.106 & \textbf{0.899} &
  91.75 & 57.51 & \textbf{8.990} & 0.883 \\
& \texttt{\textbf{\gcn}} (Ours) &
  35.47 & 10.88 & 16.111 & \textbf{0.899} &
  92.13 & \textbf{57.00} & 8.672 & 0.881 \\
\midrule
\multirow{3}{*}{Token Interactions} & \textbf{\texttt{Prompt}} &
  51.56 & 19.2 & 15.478 & 0.898 &
  87.49 & 76.43 & 7.028 & 0.888  \\
& \texttt{\textbf{\gat}} (Ours) &
  \textbf{34.28} & 10.67 & 16.106 & \textbf{0.899} &
  92.04 & 74.60 & 7.692 & \textbf{0.891}  \\
& \texttt{\textbf{\gcn}} (Ours) &
  32.59 & \textbf{10.03} & 16.121 & \textbf{0.899} &
  92.75 & 77.03 & 7.831 & \textbf{0.891}  \\
\midrule
\multirow{3}{*}{Span Interactions} & \textbf{\texttt{Prompt}} &
  42.47 & 13.65 & 16.031 & \textbf{0.899} &
  89.34 & 79.44 & 7.226 & 0.815  \\
& \texttt{\textbf{\gat}} (Ours) &
  38.05 & 12.05 & 16.119 & \textbf{0.899} &
  92.73 & 68.15 & 7.256 & 0.815  \\
& \texttt{\textbf{\gcn}} (Ours) &
  34.31 & 10.82 & \textbf{16.160} & 0.898 &
  91.99 & 71.77 & 7.771 & \textbf{0.891}  \\
\midrule[1.5pt]
\multicolumn{10}{c}{\textbf{BART-based}} \\ 
\midrule
- & \textbf{\texttt{Fine-tuning\textsubscript{base}}} &
  57.98 & 19.64 & 15.732 & 0.906 &
  \textbf{82.72} & 72.82 & 10.070 & \textbf{0.891} \\
\midrule
\multirow{3}{*}{Highlight Token} & \textbf{\texttt{Prompt}} &
  56.65 & 24.20 & 15.678 & 0.898 &
  84.74 & 61.97 & 10.012 & \textbf{0.891} \\
& \texttt{\textbf{\gat}} (Ours) &
  43.85 & 13.78 & \textbf{16.503} & \textbf{0.909} &
  91.97 & \textbf{58.11} & 10.092 & \textbf{0.891} \\
& \texttt{\textbf{\gcn}} (Ours) &
  44.68 & 14.32  &16.364 & \textbf{0.909} &
  90.95 & 59.13 & \textbf{10.489} & 0.893  \\
\midrule
\multirow{3}{*}{Token Interactions} & \textbf{\texttt{Prompt}} &
  51.56 & 19.20 & 15.478 & 0.898 &
  95.85 & 69.86 & 7.868 & 0.890 \\
& \texttt{\textbf{\gat}} (Ours) &
  48.38 & 16.07 & 16.24 & 0.908 &
  95.21 & 72.52 & 7.405 & 0.888 \\
& \texttt{\textbf{\gcn}} (Ours) &
  \textbf{41.57} & \textbf{12.89} & 16.364 & \textbf{0.909} &
  94.11 & 72.03 & 7.700 & 0.889 \\
\midrule
\multirow{3}{*}{Span Interactions} & \textbf{\texttt{Prompt}} &
  51.10 & 17.41 & 16.046 & 0.888 &
  94.89 & 65.52 & 7.333 & 0.888 \\
& \texttt{\textbf{\gat}} (Ours) &
  42.90 & 12.92 & 16.449 & \textbf{0.909} &
  93.98 & 61.39 & 7.795 & 0.890 \\
& \texttt{\textbf{\gcn}} (Ours) &
  45.48 & 14.10 & 16.447 & \textbf{0.909} &
  71.07 & 56.44 & 7.518 & 0.887 \\
\bottomrule
\end{tabular}
}
\caption{Overall evaluation results on e-SNLI and ComVE datasets for T5-based and BART-based models, with our \texttt{\textbf{\method}} model using \texttt{\textbf{\gat}} and \texttt{\textbf{\gcn}}. 
Counter indicates \textit{Counter Unfaith}, Total indicates \textit{Total Unfaith}. The best performance of each evaluation metric is in bold. See Table \ref{tab:esnli_comve_results_t5_and_bart} for results of our model using \texttt{\textbf{\sage}}.}
\label{tab:esnli_comve_results_t5_and_bart_gcn_and_gat}
\end{table*}

\subsection{Statistical Uncertainty Measurement for Faithfulness Evaluation on e-SNLI and ComVE Datasets using \texttt{\textbf{\sage}} and \textbf{\texttt{Fine-tuning\textsubscript{base}}} with \texttt{T5-large} and \texttt{BART-large} models}
\label{sec:appendix:esnli_comve_uncertainty}

To demonstrate the significant improvement of our \texttt{\textbf{\method}} in terms of faithfulness, we compute the p-values ~\cite{wasserstein2016asa} for \textit{Counter Unfaith} and \textit{Total Unfaith} (see Section \S\ref{sec:results:Faithfulness_Evaluation}) when comparing the \textbf{\texttt{Fine-tuning\textsubscript{base}}} and our \texttt{\textbf{\sage}} model on the e-SNLI and ComVE datasets, using \texttt{T5-large} and \texttt{BART-large} with 5 random seeds.
\begin{table*}[!htbp] 
\centering
\resizebox{\textwidth}{!}{%
\begin{tabular}{@{}c c|c|c|c|c@{}}
\toprule
\multirow{2}{*}{\textbf{Explanation Type}} & \multirow{2}{*}{\textbf{Model}}  
& \multicolumn{2}{c|}{\textbf{e-SNLI (P-Value)}} & \multicolumn{2}{c}{\textbf{ComVE (P-Value)}} \\
\cmidrule(lr){3-4}\cmidrule(lr){5-6}
& & Counter & Total & Counter & Total \\
& & Unfaith & Unfaith & Unfaith & Unfaith \\
\midrule
\multicolumn{6}{c}{\textbf{T5-based}} \\
\midrule
Highlight Token & \texttt{\textbf{\sage}} & 0.0007 & 0.0054 & 0.0136 & 0.0002 \\
Token Interactions & \texttt{\textbf{\sage}} & 0.0002 & 0.0001 & 0.0164 & 0.0047 \\
Span Interactions & \texttt{\textbf{\sage}} & 0.0010 & 0.0032 & 0.0001 & 0.0307 \\
\midrule
\multicolumn{6}{c}{\textbf{BART-based}} \\
\midrule
Highlight Token & \texttt{\textbf{\sage}} & 0.0067 & 0.0064 & 0.0455 & 0.0001 \\
Token Interactions & \texttt{\textbf{\sage}} & 0.0122 & 0.0007 & 0.0168 & 0.0169 \\
Span Interactions & \texttt{\textbf{\sage}} & 0.0033 & 0.0006 & 0.0403 & 0.0116 \\
\bottomrule
\end{tabular}}
\caption{P-values of our \texttt{\textbf{\sage}} model compared to \textbf{\texttt{Fine-tuning\textsubscript{base}}} on the e-SNLI and ComVE datasets, using \texttt{T5-large} and \texttt{BART-large}, regarding \textit{Counter Unfaith} and \textit{Total Unfaith} on 5 random seeds. }
\label{tab:uncertainty}
\end{table*}

As shown in Table \ref{tab:uncertainty}, all p-values are less than 0.05, indicating that the natural language explanations generated by our \texttt{\textbf{\method}} exhibit significantly lower unfaithfulness compared to the baseline method.

\subsection{Overall Explanation Evaluation Results on ECQA dataset for \method based on \texttt{T5-large} and \texttt{BART-large}}
\label{sec:appendix:ecqa_t5_bart}

\subsubsection{Overall Explanation Evaluation Results on ECQA dataset for \method based on \texttt{T5-large} }

The faithfulness and automatic evaluation results of T5-based models on the ECQA dataset are shown in Table \ref{tab:all_results_ecqa_t5}. 

Regarding the faithfulness of NLEs, almost all of our methods outperform the baseline methods, highlighting the effectiveness of our framework. Among the different highlight explanation types, \textbf{\textit{token interactive explanations}} demonstrate the best performance in generating faithful NLEs when using \textbf{\textit{\gcn}}, achieving 21.18\% total unfaithfulness. Other variants, such as \textbf{\textit{\gat}} and \textbf{\textit{\sage}}, also achieve comparable performance, with 21.44\% and 21.74\% total unfaithfulness, respectively. \textbf{On the ECQA dataset, \textbf{\textit{token interactive explanations}} show a clear advantage over other highlight explanation types in improving the faithfulness of NLEs}.

Regarding the similarity between the generated NLEs and the gold ones, \method outperforms the fine-tuning baseline in most settings. Although the prompt baseline achieves the highest SacreBLEU and BERTScore, \method lags behind by only 1.537 in SacreBLEU and 0.004 in BERTScore. Among all types of highlight explanations, \textbf{\textit{span interactive explanations}} achieve the highest scores with \method.

\begin{table*}[!htbp]
\centering
\small
\resizebox{\textwidth}{!}{%
\begin{tabular}{cccccc}
\toprule
\multicolumn{2}{c}{\multirow{2}{*}{\makecell{Evaluation \\ Metrics}}} & \multicolumn{2}{c}{UnFaithfulness(\% $\downarrow$)} & \multicolumn{2}{c}{Automatic Evaluation ($\uparrow$)} \\
\cmidrule(lr){3-4}\cmidrule(lr){5-6}
& & \makecell{Counter \\Unfaith} & \makecell{Total \\Unfaith} & \makecell{SacreBLEU \\ (0-100)} & \makecell{BERTScore \\ (0-1)} \\
\midrule
\multicolumn{2}{c}{\textbf{\texttt{Fine-tuning\textsubscript{base}}}} & 49.34 & 24.80 & 14.057 & 0.883 \\
\midrule
\multirow{4}{*}{\makecell{Highlight Tokens}} 
& \textbf{\texttt{Prompt}} & 46.56 & 25.27 & 15.303 & 0.887 \\
& \texttt{\textbf{\gat}} & 44.76 & 21.99 & 14.048 & 0.883 \\
& \texttt{\textbf{\gcn}} & 49.61 & 25.21 & 13.855 & 0.882 \\
& \texttt{\textbf{\sage}} & 45.42 & 22.44 & 13.968 & 0.882 \\
\midrule
\multirow{4}{*}{\makecell{Token Interactions}}
& \textbf{\texttt{Prompt}} & 51.29 & 33.30 & 15.311 & 0.887 \\
& \texttt{\textbf{\gat}} & 43.49 & 21.44 & 13.910 & 0.882 \\
& \texttt{\textbf{\gcn}} & \textbf{43.42} & \textbf{21.18} & 14.079 & 0.883 \\
& \texttt{\textbf{\sage}} & 44.20 & 21.74 & 13.978 & 0.882 \\
\midrule
\multirow{4}{*}{\makecell{Span Interactions}}
& \textbf{\texttt{Prompt}} & 50.20 & 28.22 & \textbf{16.046} & \textbf{0.888} \\
& \texttt{\textbf{\gat}} & 49.22 & 23.85 & 14.339 & 0.883 \\
& \texttt{\textbf{\gcn}} & 50.46 & 24.91 & 14.477 & 0.883 \\
& \texttt{\textbf{\sage}} & 46.87 & 22.50 & 14.509 & 0.884 \\
\bottomrule
\end{tabular}
}
\caption{Overall Evaluation Results on ECQA of T5-based \method. The best performance of each evaluation metric across all NLE generation models is in bold.}
\label{tab:all_results_ecqa_t5}
\end{table*}

\subsubsection{Automatic Evaluation Results on ECQA dataset for \method based on \texttt{BART-large} }
As shown in Table \ref{tab:automatic_results_ecqa_bart}, we also conduct automatic evaluation on BART-based \method on ECQA datasets regarding Lexical and Semantical Similarity with golden explanations. 

Compared to all the baseline methods, on ECQA dataset, with the highest scores always belong to our \textit{token interactive explanation} guided \texttt{\textbf{\gcn}} method, and other variants are with comparable performance to the baselines, our model also shows advantage in both lexical and semantic similarity. 

Among the different explanation types, \textit{token interactive explanations} demonstrate superior performance in both lexical and semantic metrics. Notably, \textbf{\textit{token interactive explanations} show a slight advantage over the other two explanation types in generating NLEs with more plausible meanings to humans.}

\begin{table*}[!htbp]
\centering
\small
\normalsize
\resizebox{\textwidth}{!}{%
\begin{tabular}{cc c c c c c c}
\toprule
\multicolumn{2}{c}{\multirow{2}{*}{Automatic Evaluation Metrics}} & \multicolumn{3}{c}{Lexical Similarity ($\uparrow$)} & \multicolumn{3}{c}{Semantic Similarity ($\uparrow$)} \\
\cmidrule(lr){3-5}\cmidrule(lr){6-8}
\multicolumn{2}{c}{} & ROUGE-1 (0-1) & ROUGE-L (0-1) & SacreBLEU (1-100) & MoverScore (0-1) & BARTScore (-0-1) & BERTScore (0-1) \\
\midrule
\multicolumn{2}{c}{\textbf{\texttt{Fine-tuning\textsubscript{base}}}} & 0.180 & 0.130 & 12.484 & 0.840 & -4.433 & 0.836 \\
\midrule
\multicolumn{1}{c}{\multirow{4}{*}{Highlight Tokens}} & \textbf{\texttt{Prompt}} & 0.112 & 0.077 & 10.733 & 0.767 & -4.557 & 0.754 \\
& \texttt{\textbf{\gat}} (Ours) & 0.172 & 0.125 & 12.186 & 0.837 & -4.453 & 0.835 \\
& \texttt{\textbf{\gcn}} (Ours) & 0.198 & 0.146 & 13.091 & 0.840 & -4.379 & 0.839 \\
& \texttt{\textbf{\sage}} (Ours) & 0.181 & 0.133 & 12.659 & 0.839 & -4.434 & 0.836 \\
\midrule
\multicolumn{1}{c}{\multirow{4}{*}{Token Interactions}} & \textbf{\texttt{Prompt}} & 0.185 & 0.134 & 12.724 & 0.838 & -4.435 & 0.837 \\
& \texttt{\textbf{\gat}} (Ours) & 0.208 & 0.151 & 13.519 & 0.841 & -4.399 & 0.841 \\
& \texttt{\textbf{\gcn}} (Ours) & \textbf{0.321} & \textbf{0.226} & \textbf{17.860} & \textbf{0.848} & \textbf{-4.079} & \textbf{0.858} \\
& \texttt{\textbf{\sage}} (Ours) & 0.243 & 0.174 & 14.773 & 0.843 & -4.269 & 0.847 \\
\midrule
\multicolumn{1}{c}{\multirow{4}{*}{pan Interactions}} & \textbf{\texttt{Prompt}} & 0.175 & 0.126 & 12.288 & 0.839 & -4.454 & 0.835 \\
& \texttt{\textbf{\gat}} (Ours) & 0.176 & 0.128 & 12.295 & 0.838 & -4.456 & 0.835 \\
& \texttt{\textbf{\gcn}} (Ours) & 0.175 & 0.128 & 12.364 & 0.838 & -4.455 & 0.835 \\
& \texttt{\textbf{\sage}} (Ours) & 0.186 & 0.135 & 12.802 & 0.839 & -4.415 & 0.837 \\
\bottomrule
\end{tabular}
}
\caption{Automatic Evaluation Results on ECQA of BART-based \method. The best performance of each evaluation metric across different NLE generation models is in bold.}
\label{tab:automatic_results_ecqa_bart}
\end{table*}

\subsubsection{Faithfulness Evaluation Results on ECQA dataset for \method based on \texttt{BART-large} }
We also evaluated the faithfulness of \method based on \texttt{BART-large} on the ECQA dataset and observed that the faithfulness scores for all methods (including the baselines) were uniformly 100\%. This result indicates that the BART-based models are prone to counterfactual attacks and none of these explanations were faithful. We attribute this outcome to the inherent complexity of the ECQA dataset and the potential vulnerability of the BART model to counterfactual attacks.

\subsection{Supplementary Automatic Explanation Evaluation Results for \method based on \texttt{T5-large} and \texttt{BART-large}}
\label{sec:appendix:t5_other_automatic_metrics}

To evaluate the similarity between the generated NLE and the golden ones as an approximation of plausibility to humans, we also leverage the following four metrics to evaluate their lexical and semantic similarity:

\textbf{Rouge1}~\citep{lin-2004-rouge} calculates the overlap of unigrams between the generated explanation and the golden ones, providing insight into lexical similarity at the word level.

\textbf{RougeL}~\citep{lin-2004-rouge} measures the longest common subsequence between the generated explanation and the gold explanations.

\textbf{MoverScore}~\citep{zhao-etal-2019-moverscore} calculates semantic similarity by computing word embeddings and their movement cost, capturing meaning while accounting for variations in word order and structure.

\textbf{BARTScore}~\citep{NEURIPS2021_e4d2b6e6} leverages BART's language model to assess the likelihood of the reference text being generated given the generated explanation as input, providing a fluency and relevance measure.

\subsubsection{Supplementary Automatic Explanation Evaluation Results for \method based on \texttt{T5-large}}

As shown in Table \ref{tab:part_automatic_results_esnli_t5}, Table \ref{tab:part_of_automatic_results_ecqa_t5} and Table \ref{tab:part_of_automatic_results_comve_t5}, we conduct a supplementary automatic evaluation on T5-based \method regarding Lexical Similarity and Semantic Similarity with the gold explanations on e-SNLI, ECQA and ComVE datasets respectively. 

Compared to all the baseline methods on the e-SNLI dataset, \textbf{all variants of our \method achieve higher lexical and semantic similarity with gold explanations}, indicating that our approach can generate more plausible NLEs. For instance, we observe up to a 2.1\% improvement in ROUGE-1 and a notable absolute increase of 0.224 in BARTScore. On the ECQA dataset, our \method achieves better similarity performance than \texttt{Fine-tuning\textsubscript{base}} (which does not utilize explanation information) and is comparable to the prompt-based baseline. On the ComVE dataset, all NLEs generated by our method incorporating \textbf{\textit{highlight token explanations}} surpass the baselines in both lexical and semantic similarity, while the variants based on \textbf{\textit{token interactive explanations}} and \textbf{\textit{span interactive explanations}} sometimes fail to do so. This is likely due to the format of the ComVE dataset, which presents a simple question followed by two similar statements. In this scenario, \textbf{\textit{token interactive explanations}} and \textbf{\textit{span interactive explanations}} may struggle to capture sufficient information from the limited interaction between the question and the options.

Among the different highlight explanation types on the e-SNLI dataset, \textbf{\textit{token interactive explanations}}, particularly those using the \sage variant of our \method, achieve the highest lexical and semantic similarity. Meanwhile, \textbf{\textit{highlight token explanations}} and \textbf{\textit{span interactive explanations}} also perform strongly, excelling at ROUGE-L and ROUGE-1 scores respectively. On the ECQA dataset, \textbf{\textit{span interactive explanations}} have a slight edge over other explanation types, although the difference is marginal. On the ComVE dataset, \textbf{\textit{highlight token explanations}} show a clear advantage across all metrics. This is likely due to the input format of the ComVE dataset, which makes it challenging for \textbf{\textit{token interactive explanations}} and \textbf{\textit{span interactive explanations}} to capture sufficient information, as discussed earlier.

In summary, \textbf{these findings highlight that the advantages of different explanation types in improving NLE quality vary with dataset characteristics.}

\begin{table*}[!htbp]
\centering
\small
\normalsize
\resizebox{\textwidth}{!}{%
\begin{tabular}{cc c c c c}
\toprule
\multicolumn{2}{c}{\multirow{2}{*}{Automatic Evaluation Metrics}} & \multicolumn{2}{c}{Lexical Similarity($\uparrow$)} & \multicolumn{2}{c}{Semantic Similarity($\uparrow$)} \\
\cmidrule(lr){3-4}\cmidrule(lr){5-6}
\multicolumn{2}{c}{} & ROUGE-1 (0-1) & ROUGE-L (0-1) & MoverScore (0-1) & BARTScore (-0-1) \\
\midrule
\multicolumn{2}{c}{\textbf{\texttt{Fine-tuning\textsubscript{base}}}} & 0.448 & 0.384 & 0.838 & -3.646 \\
\midrule
\multicolumn{1}{c}{\multirow{4}{*}{Highlight Tokens}} & \textbf{\texttt{Prompt}} & 0.455 & 0.397 & 0.840 & -3.492 \\
& \texttt{\textbf{\gat}} (Ours) & 0.467 & 0.402 & 0.842 & -3.437 \\
& \texttt{\textbf{\gcn}} (Ours) & 0.468 & 0.403 & 0.842 & -3.425 \\
& \texttt{\textbf{\sage}} (Ours) & 0.468 & \textbf{0.404} & 0.841 & \textbf{-3.422} \\
\midrule
\multicolumn{1}{c}{\multirow{4}{*}{Token Interactions}} & \textbf{\texttt{Prompt}} & 0.459 & 0.394 & 0.842 & -3.503 \\
& \texttt{\gat} (Ours) & 0.467 & 0.402 & 0.842 & -3.437 \\
& \texttt{\textbf{\gcn}} (Ours) & 0.467 & 0.403 & 0.842 & -3.435 \\
& \texttt{\textbf{\sage}} (Ours) & \textbf{0.469} & \textbf{0.404} & \textbf{0.843} & -3.431 \\
\midrule
\multicolumn{1}{c}{\multirow{4}{*}{Span Interactions}} & \textbf{\texttt{Prompt}} & 0.466 & 0.402 & 0.841 & -3.467 \\
& \texttt{\gat} (Ours) & 0.466 & 0.403 & 0.841 & -3.442 \\
& \texttt{\textbf{\gcn}} (Ours) & \textbf{0.469} & 0.403 & 0.843 & -3.433 \\
& \texttt{\textbf{\sage}} (Ours) & 0.467 & 0.402 & 0.842 & -3.428 \\
\bottomrule
\end{tabular}
}
\caption{Automatic Evaluation Results on e-SNLI of T5-based \method (excluding SacreBLEU and BERTScore, which are presented in Table \ref{tab:esnli_comve_results_t5_and_bart}). The best performance of each evaluation metric across different NLE generation models is in bold.}
\label{tab:part_automatic_results_esnli_t5}
\end{table*}

\begin{table*}[!htbp]
\centering
\small
\normalsize
\resizebox{\textwidth}{!}{%
\begin{tabular}{cc c c c c}
\toprule
\multicolumn{2}{c}{\multirow{2}{*}{Automatic Evaluation Metrics}} & \multicolumn{2}{c}{Lexical Similarity($\uparrow$)} & \multicolumn{2}{c}{Semantic Similarity($\uparrow$)} \\ 
\cmidrule(lr){3-4}\cmidrule(lr){5-6}
\multicolumn{2}{c}{} & ROUGE-1 (0-1) & ROUGE-L (0-1) & MoverScore (0-1) & BARTScore (-0-1) \\ 
\midrule
\multicolumn{2}{c}{\textbf{\texttt{Fine-tuning\textsubscript{base}}}} & 0.469 & 0.346 & 0.850 & -3.584 \\
\midrule
\multicolumn{1}{c}{\multirow{4}{*}{Highlight Tokens}} & \textbf{\texttt{Prompt}} & 0.490 & 0.355 & \textbf{0.857} & \textbf{-3.528} \\
& \texttt{\textbf{\gat}} (Ours) & 0.469 & 0.346 & 0.851 & -3.576 \\
& \texttt{\textbf{\gcn}} (Ours) & 0.468 & 0.347 & 0.850 & -3.575 \\
& \texttt{\textbf{\sage}} (Ours) & 0.468 & 0.347 & 0.850 & -3.569 \\
\midrule
\multicolumn{1}{c}{\multirow{4}{*}{Token Interactions}} & \textbf{\texttt{Prompt}} & 0.489 & 0.354 & 0.855 & -3.549 \\
& \texttt{\textbf{\gat}} (Ours) & 0.468 & 0.345 & 0.849 & -3.598 \\
& \texttt{\textbf{\gcn}} (Ours) & 0.469 & 0.346 & 0.850 & -3.593 \\
& \texttt{\textbf{\sage}} (Ours) & 0.468 & 0.346 & 0.851 & -3.593 \\
\midrule
\multicolumn{1}{c}{\multirow{4}{*}{Span Interactions}} & \textbf{\texttt{Prompt}} & \textbf{0.496} & \textbf{0.360} & \textbf{0.857} & -3.520 \\
& \texttt{\textbf{\gat}} (Ours) & 0.472 & 0.350 & 0.850 & -3.569 \\
& \texttt{\textbf{\gcn}} (Ours) & 0.470 & 0.349 & 0.849 & -3.568 \\
& \texttt{\textbf{\sage}} (Ours) & 0.474 & 0.350 & 0.851 & -3.560 \\
\bottomrule
\end{tabular}
}
\caption{Automatic Evaluation Results on ECQA of T5-based \method (excluding SacreBLEU and BERTScore, which are presented in Table \ref{tab:all_results_ecqa_t5}). The best performance of each evaluation metric across different NLE generation model is in bold.}
\label{tab:part_of_automatic_results_ecqa_t5}
\end{table*}

\begin{table*}[!htbp]
\centering
\small
\normalsize
\resizebox{\textwidth}{!}{%
\begin{tabular}{cc c c c c}
\toprule
\multicolumn{2}{c}{\multirow{2}{*}{Automatic Evaluation Metrics}} & \multicolumn{2}{c}{Lexical Similarity($\uparrow$)} & \multicolumn{2}{c}{Semantic Similarity($\uparrow$)} \\ 
\cmidrule(lr){3-4}\cmidrule(lr){5-6}
\multicolumn{2}{c}{} & ROUGE-1 (0-1) & ROUGE-L (0-1) & MoverScore (0-1) & BARTScore (-0-1) \\ 
\midrule
\multicolumn{2}{c}{\textbf{\texttt{Fine-tuning\textsubscript{base}}}} & 0.355 & 0.319 & 0.828 & -4.030 \\
\midrule
\multicolumn{1}{c}{\multirow{4}{*}{Highlight Tokens}} & \textbf{\texttt{Prompt}} & 0.354 & 0.317 & 0.825 & -4.051 \\
& \texttt{\textbf{\gat}} (Ours) & \textbf{0.394} & 0.332 & 0.832 & -3.884 \\
& \texttt{\textbf{\gcn}} (Ours) & 0.384 & \textbf{0.333} & 0.830 & -3.934 \\
& \texttt{\textbf{\sage}} (Ours) & 0.393 & 0.330 & \textbf{0.833} & \textbf{-3.881} \\
\midrule
\multicolumn{1}{c}{\multirow{4}{*}{Token Interactions}} & \textbf{\texttt{Prompt}} & 0.312 & 0.269 & 0.817 & -4.083 \\
& \texttt{\textbf{\gat}} (Ours) & 0.326 & 0.283 & 0.816 & -3.976 \\
& \texttt{\textbf{\gcn}} (Ours) & 0.332 & 0.288 & 0.817 & -3.970 \\
& \texttt{\textbf{\sage}} (Ours) & 0.310 & 0.266 & 0.817 & -4.070 \\
\midrule
\multicolumn{1}{c}{\multirow{4}{*}{Span Interactions}} & \textbf{\texttt{Prompt}} & 0.317 & 0.275 & 0.815 & -4.059 \\
& \texttt{\textbf{\gat}} (Ours) & 0.324 & 0.280 & 0.815 & -3.998 \\
& \texttt{\textbf{\gcn}} (Ours) & 0.328 & 0.286 & 0.815 & -3.975 \\
& \texttt{\textbf{\sage}} (Ours) & 0.328 & 0.283 & 0.818 & -3.980 \\
\bottomrule
\end{tabular}
}
\caption{Automatic Evaluation Results on ComVE of T5-based \method (excluding SacreBLEU and BERTScore, which are presented in Table \ref{tab:esnli_comve_results_t5_and_bart}). The best performance of each evaluation metric across different NLE generation models is in bold.}
\label{tab:part_of_automatic_results_comve_t5}
\end{table*}

\subsubsection{Supplementary Automatic Explanation Evaluation Results for \method based on \texttt{BART-large}}

As shown in Table \ref{tab:part_automatic_results_esnli_bart}, Table \ref{tab:automatic_results_ecqa_bart} and Table \ref{tab:part_automatic_results_comve_bart}, we conduct a supplementary automatic evaluation on BART-based \method regarding Lexical Similarity and Semantic Similarity with the gold explanations on e-SNLI, ECQA and ComVE datasets respectively.


\begin{table*}[!htbp]
\centering
\small
\normalsize
\resizebox{\textwidth}{!}{%
\begin{tabular}{cc c c c c}
\toprule
\multicolumn{2}{c}{\multirow{2}{*}{Automatic Evaluation Metrics}} & \multicolumn{2}{c}{Lexical Similarity($\uparrow$)} & \multicolumn{2}{c}{Semantic Similarity($\uparrow$)} \\
\cmidrule(lr){3-4}\cmidrule(lr){5-6}
\multicolumn{2}{c}{} & ROUGE-1 (0-1) & ROUGE-L (0-1) & MoverScore (0-1) & BARTScore (-0-1) \\
\midrule
\multicolumn{2}{c}{\textbf{\texttt{Fine-tuning\textsubscript{base}}}} & 0.457 & 0.391 & 0.838 & -3.491 \\
\midrule
\multicolumn{1}{c}{\multirow{4}{*}{Highlight Tokens}} & \textbf{\texttt{Prompt}} & 0.468 & 0.398 & \textbf{0.843} & -3.458 \\
& \texttt{\textbf{\gat}} (Ours) & 0.476 & 0.405 & \textbf{0.843} & \textbf{-3.403} \\
& \texttt{\textbf{\gcn}} (Ours) & 0.474 & 0.402 & 0.841 & -3.415 \\
& \texttt{\textbf{\sage}} (Ours) & 0.474 & 0.402 & 0.840 & -3.416 \\
\midrule
\multicolumn{1}{c}{\multirow{4}{*}{Token Interactions}} & \textbf{\texttt{Prompt}} & 0.459 & 0.394 & \textbf{0.843} & -3.503 \\
& \texttt{\textbf{\gat}} (Ours) & 0.472 & 0.401 & 0.841 & -3.449 \\
& \texttt{\textbf{\gcn}} (Ours) & 0.473 & 0.402 & 0.842 & -3.418 \\
& \texttt{\textbf{\sage}} (Ours) & 0.472 & 0.403 & 0.841 & -3.431 \\
\midrule
\multicolumn{1}{c}{\multirow{4}{*}{Span Interactions}} & \textbf{\texttt{Prompt}} & 0.475 & 0.403 & 0.841 & -3.419 \\
& \texttt{\textbf{\gat}} (Ours) & \textbf{0.477} & 0.403 & 0.842 & -3.427 \\
& \texttt{\textbf{\gcn}} (Ours) & 0.476 & 0.403 & 0.842 & -3.423 \\
& \texttt{\textbf{\sage}} (Ours) & \textbf{0.477} & \textbf{0.404} & 0.842 & -3.423 \\
\bottomrule
\end{tabular}
}
\caption{Automatic Evaluation Results on e-SNLI of BART-based \method (SacreBLEU and BERTScore are excluded and are presented in Table \ref{tab:esnli_comve_results_t5_and_bart}). The best performance of each evaluation metric across different NLE generation models is in bold.}
\label{tab:part_automatic_results_esnli_bart}
\end{table*}

\begin{table*}[!htbp]
\centering
\small
\normalsize
\resizebox{\textwidth}{!}{%
\begin{tabular}{cc c c c c}
\toprule
\multicolumn{2}{c}{\multirow{2}{*}{Automatic Evaluation Metrics}} & \multicolumn{2}{c}{Lexical Similarity($\uparrow$)} & \multicolumn{2}{c}{Semantic Similarity($\uparrow$)} \\ 
\cmidrule(lr){3-4}\cmidrule(lr){5-6}
\multicolumn{2}{c}{} & ROUGE-1 (0-1) & ROUGE-L (0-1) & MoverScore (0-1) & BARTScore (-0-1) \\ 
\midrule
\multicolumn{2}{c}{\textbf{\texttt{Fine-tuning\textsubscript{base}}}} & 0.421 & 0.325 & \textbf{0.840} & -3.802 \\
\midrule
\multicolumn{1}{c}{\multirow{4}{*}{Highlight Tokens}} & \textbf{\texttt{Prompt}} & 0.419 & 0.322 & 0.834 & -3.796 \\
& \texttt{\textbf{\gat}} (Ours) & 0.427 & 0.325 & 0.837 & -3.765 \\
& \texttt{\textbf{\gcn}} (Ours) & \textbf{0.435} & \textbf{0.332} & 0.838 & -3.761 \\
& \texttt{\textbf{\sage}} (Ours) & 0.434 & 0.330 & 0.837 & \textbf{-3.748} \\
\midrule
\multicolumn{1}{c}{\multirow{4}{*}{Token Interactions}} & \textbf{\texttt{Prompt}} & 0.334 & 0.284 & 0.818 & -4.036 \\
& \texttt{\textbf{\gat}} (Ours) & 0.322 & 0.277 & 0.818 & -4.047 \\
& \texttt{\textbf{\gcn}} (Ours) & 0.334 & 0.285 & 0.817 & -3.985 \\
& \texttt{\textbf{\sage}} (Ours) & 0.316 & 0.269 & 0.814 & -4.129 \\
\midrule
\multicolumn{1}{c}{\multirow{4}{*}{Span Interactions}} & \textbf{\texttt{Prompt}} & 0.323 & 0.274 & 0.818 & -4.029 \\
& \texttt{\textbf{\gat}} (Ours) & 0.334 & 0.288 & 0.818 & -4.011 \\
& \texttt{\textbf{\gcn}} (Ours) & 0.327 & 0.278 & 0.818 & -4.045 \\
& \texttt{\textbf{\sage}} (Ours) & 0.333 & 0.287 & 0.820 & -4.017 \\
\bottomrule
\end{tabular}
}
\caption{Automatic Evaluation Results on ComVE of BART-based \method. The best performance of each evaluation metric across different NLE generation models is in bold.}
\label{tab:part_automatic_results_comve_bart}
\end{table*}

\subsection{Generalizability of \method to \texttt{LED} Model}\label{app:led}

To further demonstrate the generalizability of \method, we apply our framework to the LED model \citep{beltagy2020longformer}, an encoder-decoder architecture designed for long-document processing. The results in Table \ref{tab:led-results} show that \method outperforms baseline methods fine-tuning and prompt regarding faithfulness, which reinforces our claim of the framework’s broad applicability.

\begin{table*}[!htbp]
\centering
\resizebox{\textwidth}{!}{%
\begin{tabular}{l l r r}
\toprule
\textbf{Method} & \textbf{Model} & \textbf{\%Counter Unfaith} & \textbf{\%Total Unfaith} \\
\midrule
\texttt{Fine-tuning\textsubscript{base}} & \textbf{\texttt{Fine-tuning}} & 97.86\% & 96.63\% \\
\multirow{2}{*}{Highlight Tokens} 
& \textbf{\texttt{Prompt}} & 87.45\% & 77.28\% \\
& \textbf{\sage} & 85.79\% & 55.57\% \\
\multirow{2}{*}{Token Interactions} 
& \textbf{\texttt{Prompt}} & 98.04\% & 79.91\% \\
& \textbf{\sage} & 86.38\% & 54.19\% \\
\multirow{2}{*}{Span Interactions} 
& \textbf{\texttt{Prompt}} & 93.70\% & 86.23\% \\
& \textbf{\sage} & 84.84\% & 51.41\% \\
\bottomrule
\end{tabular}
}
\caption{Unfaithfulness scores on e-SNLI for LED model.}
\label{tab:led-results}
\end{table*}

\subsection{Generalizability of \method to Decoder-only Model}
\label{sec:appendix:decoder_only}

Similar to the encoder-decoder models, we insert a GNN layer into the $\tfrac{3}{4}$-th decoder layer of \texttt{Llama-3.2-1B}. We then evaluate \method on the e-SNLI and ComVE datasets, with results summarized in Table~\ref{tab:llama32_esnli_comve}.

Overall, we observe that \method consistently reduces both \textit{Counter Unfaithfulness} and \textit{Total Unfaithfulness} compared to prompt-based baselines across all explanation types. In particular, the \textit{Token Interaction} variant of \method achieve substantial gains on e-SNLI, with 30.23\% Counter Unfaith and 10.08\% Total Unfaith, compared to 35.98\% and 13.34\% for the prompt baseline. On ComVE, \method improves the Total Unfaith score of Highlight Token explanations by more than 11 points (51.03\% vs.\ 62.30\%). 

These results suggest that the benefits of \method are not limited to encoder--decoder architectures. Instead, our approach effectively transfers to decoder-only LLMs, further supporting the flexibility of the proposed framework.

\begin{table*}[!htbp]
\centering
\resizebox{\textwidth}{!}{%
\begin{tabular}{@{}c c|cc|cc@{}}
\toprule
\multirow{3}{*}{\textbf{Explanation Type}} & \multirow{3}{*}{\textbf{Model}} 
& \multicolumn{2}{c|}{\textbf{e-SNLI}} 
& \multicolumn{2}{c}{\textbf{ComVE}} \\ 
\cmidrule(lr){3-4}\cmidrule(lr){5-6}
& & \multicolumn{1}{c}{Counter Unfaith\%$\downarrow$} & \multicolumn{1}{c|}{Total Unfaith\%$\downarrow$} 
& \multicolumn{1}{c}{Counter Unfaith\%$\downarrow$} & \multicolumn{1}{c}{Total Unfaith\%$\downarrow$} \\
\midrule
- & \textbf{\texttt{Fine-tuning}} &
  45.12 & 14.24 &
  87.22 & 62.59 \\
\midrule
\multirow{2}{*}{Highlight Token} & \textbf{\texttt{Prompt}} &
  40.73 & 11.97 &
  86.23 & 62.30 \\
& \texttt{\textbf{Tex-SAGE}} (Ours) &
  31.23 & 10.26 &
  84.33 & \textbf{51.03} \\
\midrule
\multirow{2}{*}{Token Interactions} & \textbf{\texttt{Prompt}} &
  35.98 & 13.34 &
  82.18 & 69.27 \\
& \texttt{\textbf{Tex-SAGE}} (Ours) &
  \textbf{30.23} & \textbf{10.08} &
  82.13 & 66.24 \\
\midrule
\multirow{2}{*}{Span Interactions} & \textbf{\texttt{Prompt}} &
  46.32 & 13.27 &
  81.29 & 70.26 \\
& \texttt{\textbf{Tex-SAGE}} (Ours) &
  32.14 & 12.21 &
  \textbf{79.87} & 66.21 \\
\bottomrule
\end{tabular}
}
\caption{Unfaithfulness scores on e-SNLI and ComVE datasets using \texttt{Llama-3.2-1B}. The best performance for each column is in bold.}
\label{tab:llama32_esnli_comve}
\end{table*}

\subsection{Human Evaluation}
\label{Human_Evaluation}
In line with prior work \citep{atanasova-etal-2020-generating-fact,jolly2022generating}, our human evaluation assesses the generated explanations across four key dimensions: 

\textbf{Coverage}: The explanation includes all important and salient information, ensuring no significant points that contribute to label prediction are omitted.

\textbf{Non-redundancy}: The explanation should avoid redundant, repeated, or irrelevant information and should not include content that is unreasonable or inconsistent with common sense.

\textbf{Non-contradiction}: The explanation should not contradict the predicted label or the input text, maintaining consistency throughout.

\textbf{Overall Quality}: The explanations are rated based on overall quality, considering factors such as grammar, readability, and clarity.

We engaged three PhD students with backgrounds in computer science to evaluate the explanations using a 1–7 Likert scale following previous work \citep{castro-ferreira-etal-2019-neural,ribeiro-etal-2021-investigating,yuan-farber-2024-grasame}. We compare the text generated by the \texttt{Fine-tuning\textsubscript{base}} with that generated by \texttt{\textbf{\gat}} when guided by \textit{highlight token}, \textit{token interactive explanations}, and \textit{span interactive explanations}, respectively.  
The annotator agreement is reported in Table \ref{tab:agreemeent}. Note that we randomly sample 100 NLEs generated by each model.  


\subsubsection{Human Evaluation Results}
\begin{table*}[!htbp]
\centering
\normalsize
\resizebox{\textwidth}{!}{%
\begin{tabular}{ccccc}
\toprule
\textbf{Method} & \textbf{Coverage} & \textbf{Non Redund.} & \textbf{Non Contrad.} & \textbf{Overall} \\
\midrule
\texttt{Fine-tuning\textsubscript{base}} & 6.72 & 5.86 & 6.67 & 6.28 \\
Highlight Tokens & 6.74 & 5.80 & 6.67 & 6.06 \\
Token Interactions & \textbf{6.75} & \textbf{5.95} & 6.64 & \textbf{6.37} \\
Span Interactions & 6.67 & 5.92 & \textbf{6.72} & 6.26 \\
\bottomrule
\end{tabular}
}
\caption{Human Evaluation Results on e-SNLI dataset of our \texttt{\textbf{\method}} using \texttt{\textbf{\gat}} based on T5.}
\label{tab:esnli_human_eval}
\end{table*}

\begin{table*}[!htbp]
\centering
\normalsize
\resizebox{\textwidth}{!}{%
\begin{tabular}{ccccc}
\toprule
\textbf{Method} & \textbf{Coverage} & \textbf{Non Redund.} & \textbf{Non Contrad.} & \textbf{Overall} \\
\midrule
\texttt{Fine-tuning\textsubscript{base}} & 5.66 & 4.41 & 4.91 & 5.53 \\
Highlight Tokens  & 5.08 & 4.27 & 4.51 & 5.20 \\
Token Interactions & 5.60 & \textbf{4.82} & \textbf{5.08} & 5.61 \\
Span Interactions  & \textbf{5.65} & 4.67 & 4.90 & \textbf{5.63} \\
\bottomrule
\end{tabular}
}
\caption{Human Evaluation Results on ECQA dataset of  our \texttt{\textbf{\method}} using \texttt{\textbf{\gat}} based on T5.}
\label{tab:ecqa_human_eval}
\end{table*}

\paragraph{e-SNLI} In Table \ref{tab:esnli_human_eval}, across all highlight explanation types, the NLEs generated by the \textit{token interactive explanations} achieve the highest scores across most dimensions, particularly excelling in \textit{Non-redundancy} (5.95) and \textit{Overall Quality} (6.37), indicating its effectiveness in producing concise and high-quality explanations. The NLEs generated with the guidance of \textit{span interactive explanations} method also show strong performance, especially in \textit{Non-contradiction} (6.72), suggesting that modeling span-level interactions is beneficial for maintaining consistency of the NLE with the generated label. The \textit{highlighted token explanations} performs slightly lower, indicating that while it captures key tokens effectively, it may miss out on broader contextual relationships crucial for non-redundancy and overall quality.

\paragraph{ECQA} Table \ref{tab:ecqa_human_eval} shows the evaluation results for the ECQA dataset, where the NLEs generated by \textit{token interactive explanations} again lead in \textit{Non-redundancy} (4.82) and achieves a high \textit{Non-contradiction} score (5.08), confirming its robustness across different datasets. The \textit{span interactive explanations} perform similarly well, attaining the highest \textit{Overall Quality} score (5.63), emphasizing its adaptability in varied datasets. 

Overall, while the \textit{highlight token explanations} shows slightly lower performance across all highlight explanation types, leveraging \textit{span interactive explanations} and \textit{token interactive explanations} that are encoded in \method notably improves the quality and consistency of the generated explanations.

\subsubsection{Human Evaluation Instruction\label{human_instruction}}

The annotators are asked to rate the generated texts following the instructions in Table \ref{rating_instruction}.

\begin{table*}[!htbp]
\centering
\resizebox{\textwidth}{!}{%
\begin{tabular}{p{5cm} p{3.5cm} p{3.5cm} p{3.5cm}}
\toprule
\textbf{Criterion and Explanation} & \textbf{1 - 3 (Very Bad)} & \textbf{3 - 5 (OK, but not good enough)} & \textbf{5 - 7 (Good to Very Good)} \\
\midrule
\textbf{Coverage:} The explanation contains important, salient information and does not miss any important points that contribute to the label prediction. & The explanation misses the most critical points in the input text. & The explanation provides a reason for the prediction, but not the main reason. & The explanation covers the most important points/reasons for the prediction. \\
\midrule
\textbf{Non-redundancy:} The explanation does not contain any information that is redundant, repeated, or irrelevant to the claim and predicted label. It should also be reasonable according to common sense. & The explanation contains irrelevant information, unnecessary repetition, or elements that do not appear in the input text; violates common sense. & The explanation is acceptable but contains some redundancy or repetition. & Slightly to no redundancy, repetition, or hallucination. \\
\midrule
\textbf{Non-contradiction:} The explanation does not contain any pieces of information that are contradictory to the predicted label and the input text. & The explanation contradicts the predicted label or input text; they address different topics. & The explanation matches the predicted label but is not fully logical. & The explanation and predicted label are fully consistent and logical. \\
\midrule
\textbf{Overall Quality:} Rank the explanations by their overall quality. Consider grammar, readability, and clarity. & Many grammatical errors, difficult to understand. & No major grammar mistakes, but not easy to understand. & Perfect grammar and language clarity. \\
\bottomrule
\end{tabular}}
\caption{Rating Criteria for Generated Natural Language Explanations\label{rating_instruction}}
\end{table*}

\subsubsection{Pairwise agreement for human annotations}
\label{sec:appendix:annotator_agreement}

\begin{table*}[h]
\centering
\resizebox{\textwidth}{!}{%
\begin{tabular}{c|cc|cc|cc|ccc}\toprule
- &\multicolumn{2}{c}{\textbf{Coverage}} &\multicolumn{2}{c}{\textbf{Non-redundancy}} &\multicolumn{2}{c}{\textbf{Non-contradiction}} &\multicolumn{2}{c}{\textbf{Overall}} \\\cmidrule{2-9}
\textbf{Annotator\_id} &\textbf{2} &\textbf{3} &\textbf{2} &\textbf{3} &\textbf{2} &\textbf{3} &\textbf{2} &\textbf{3} \\\midrule
\multicolumn{9}{c}{\textbf{e-SNLI}} \\
\arrayrulecolor{black!30}\midrule
\textbf{1} &0.51 &0.25 &0.53 &0.43 &0.36 &0.19 &0.33 &0.16 \\
\textbf{2} &- &0.40 &- &0.53 &- &0.43 &- &0.37 \\
  \arrayrulecolor{black!10}\midrule
\textbf{Mean} &\multicolumn{2}{c}{0.39} &\multicolumn{2}{c}{0.49} &\multicolumn{2}{c}{0.33} &\multicolumn{2}{c}{0.29} \\
  \arrayrulecolor{black!70}\midrule
\multicolumn{9}{c}{\textbf{ECQA}} \\
  \arrayrulecolor{black!30}\midrule
\textbf{1} &0.35 &0.20 &0.33 &0.15 &0.58 &0.40 &0.27 &-0.02 \\
\textbf{2} &- &0.10 &- &0.29 &- &0.35 &- &0.30 \\
\arrayrulecolor{black!10}\midrule
\textbf{Mean} &\multicolumn{2}{c}{0.22} &\multicolumn{2}{c}{0.26} &\multicolumn{2}{c}{0.44} &\multicolumn{2}{c}{0.18} \\
\arrayrulecolor{black!100}\bottomrule
\end{tabular}
}
\caption{Pairwise agreement for human annotations on e-SNLI and ECQA. We report separately the agreement between annotator pairs 1-2,  2-3, and 1-3. Mean represents the average over three pairwise agreements. }\label{tab:agreemeent}
\end{table*}

Table \ref{tab:agreemeent} shows Pairwise agreement for human annotations for NLE generated by T5-based \method on e-SNLI and ECQA dataset.

\chapter[Explaining Uncertainty in Automated Fact-Checking]
        {Explaining Sources of Uncertainty in Automated Fact-Checking}
\chaptermark{Explaining Uncertainty in Automated Fact-Checking}
\label{chap:NLE_CLUE}
This chapter consists of the following published paper\footnote{No changes have been made to the scientific content of the originally published paper. The paper is presented in a single-column
format used throughout the thesis, and its references have been merged into the bibliography at the end of the thesis.}:

\begin{enumerate}
    \item \bibentry{sun-etal-2026-explaining}.
\end{enumerate}

\clearpage
\begin{abstract}

Human-AI collaboration in knowledge-intensive tasks such as fact-checking requires understanding model uncertainty in multi‑document reasoning amid conflicting/agreeing evidence. Yet, existing methods only express uncertainty as numbers or hedges without revealing which evidence conflicts cause the uncertainty, leaving users unable to resolve disagreements. 
We present \ModelName\ (\textbf{C}onflict-\&Agreement-aware \textbf{L}anguage-model \textbf{U}ncertainty \textbf{E}xplanations), a plug‑and‑play white‑box framework that, to our knowledge, is the first to generate natural‑language explanations of model uncertainty grounded in conflicting/agreeing evidence. \ModelName\ (i) identifies span‑level claim-evidence and inter‑evidence relations that signal conflict or agreement without supervision, and (ii) uses these relations to steer explanation generation, articulating how they drive the model's uncertainty. Across three language models and two fact‑checking datasets, \ModelName\ produces explanations that more faithfully track model uncertainty and better align with the model’s fact‑checking decisions than span‑agnostic explanation prompting; human raters also judge them more helpful, more informative, less redundant, and more logically consistent with the input. By explicitly tying uncertainty to evidence conflicts and agreements, \ModelName\ supports practical fact‑checking and other tasks that require reasoning over complex, conflicting information\footnote{\url{https://github.com/lianyiyi/CLUE}}.

\end{abstract}

\section{Introduction}

Large Language Models (LLMs) are increasingly prevalent in supporting high-stakes tasks that involve reasoning about information reliability, such as fact-checking \citep{wang-etal-2024-factcheck,fontana2025evaluatingopensourcelargelanguage}.
To foster effective use of such models in fact-checking tasks, these models must explain the rationale for their predictions \citep{atanasova-etal-2020-generating-fact,kotonya_explainable_2020}.
%

\begin{figure*}
    \centering
        \includegraphics[width=\linewidth]{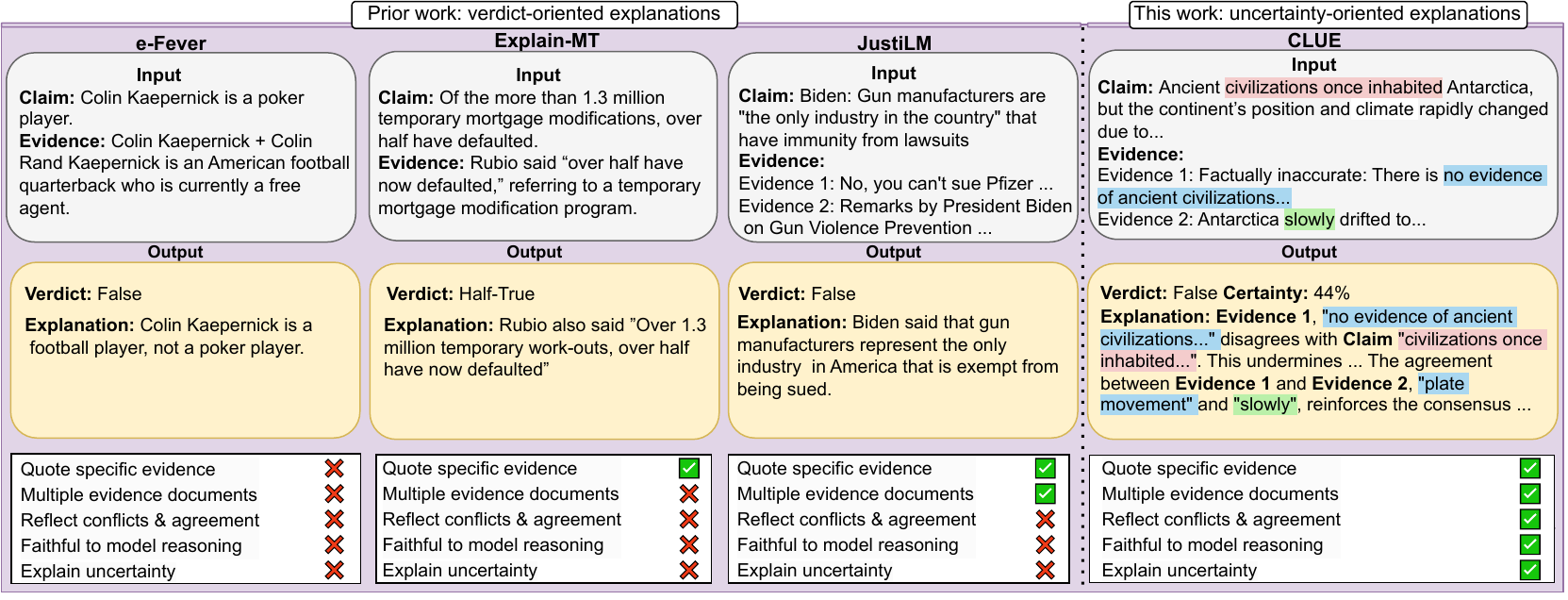}
\caption{Illustrative comparison of verdict-oriented fact-checking explanations
(e-FEVER~\citep{stammbach2020fever}, Explain-MT~\citep{atanasova-etal-2020-generating-fact}, JustiLM~\citep{zeng-gao-2024-justilm})
and CLUE's uncertainty-oriented explanations. In this comparison, CLUE is the only approach that explicitly traces predictive uncertainty to claim-evidence and inter-evidence conflict/agreement across multiple evidence passages.}
        \label{fig:expl_uncertainty}
\end{figure*}

However, current methods in automated fact-checking have been criticised for misaligning with the practical needs of fact-checkers, who must assess source reliability and reconcile conflicting evidence, rather than merely predicting a verdict label \citep{warren2025explainablefactchecking,schlichtkrull-etal-2023-intended}. Current approaches often produce explanations to justify the verdict prediction
~\citep{atanasova-etal-2020-generating-fact,stammbach2020fever,zeng-gao-2024-justilm}. However, they provide no support for communicating model uncertainty or surfacing agreements and conflicts within the evidence (See Fig.~\ref{fig:expl_uncertainty}). Methods that do express uncertainty do so via numeric scores (e.g., ``I am 73\% confident'') that are hard to contextualise \citep{zimmer1983verbal,wallsten1993preferences,van2020interpretable,liu2020intuitive} or via natural language expressions (e.g., ``I’m not sure'') that often fail to faithfully reflect model uncertainty and can inflate perceived confidence \citep{steyvers2024calibrationgapmodelhuman,yona-etal-2024-large,kim2024LLMuncertainty}. Therefore, existing explainable fact-checking systems usually exhibit two limitations: (1) neglecting to communicate model uncertainty with explanations and (2) failing to surface evidentiary conflicts and agreements that contribute to model uncertainty. This constitutes a fundamental methodological gap, as effective fact-checking requires precisely identifying the sources of uncertainty, for example, from conflicting evidence, to guide targeted verification \citep{graves2017anatomy,micallef2022true}.

\begin{figure}[!htbp]
    \centering
    \includegraphics[width=0.65\linewidth]{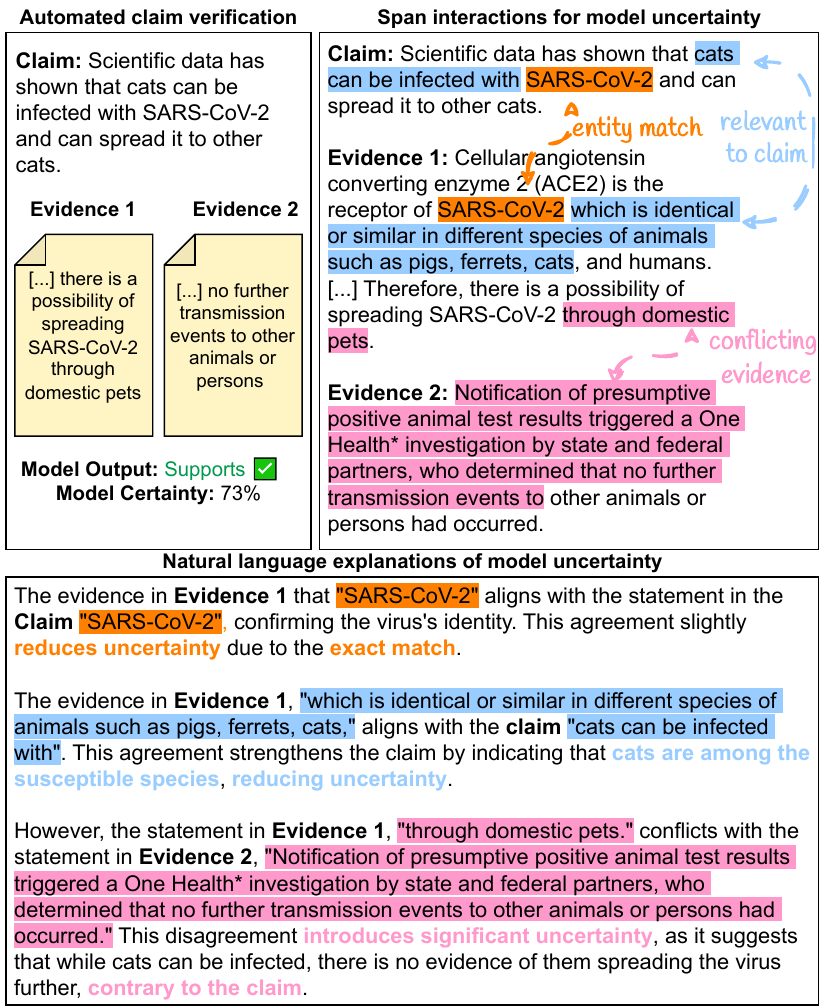}
    \caption{An example of \modelname{} output for a claim with two evidence passages: model verdict and certainty, and an uncertainty explanation grounded in conflict/agreement span interactions.}

    \label{fig:overview-example}
\end{figure}

To bridge this gap, we propose \ModelName, a pipeline that generates natural language explanations (NLEs) of model uncertainty by explicitly capturing conflicts and agreements in the input (e.g., a claim and its supporting or refuting evidence), see Fig. \ref{fig:overview-example}. The pipeline first identifies the salient span-level interactions that are essential to the prediction of the model through an unsupervised approach, providing an input-feature explanation that highlights key relationships between separate input segments, e.g., claim and evidence \citep{choudhury-etal-2023-explaining}. Previous work has shown that these interactions are faithful to the model and plausible to humans \citep{sun-etal-2025-evaluating,sun2026evaluation}. \ModelName\ then converts these signals into uncertainty-aware explanations by explicitly referring to the interactions, the conflict/agreement relations they express, and how they contribute to uncertainty regarding the verdict. \ModelName\ does not require gold-label explanations or fine-tuning, and operates entirely at inference time. 

Across three language models (\S\ref{sec:experiment_setup_models}) and two fact-checking datasets (\S\ref{sec:datasets}), we evaluate two variants of \modelname, which leverage distinct methods to guide NLE generation based on the extracted key conflicts/agreements (\S\ref{sec:nle_generation:method}). 
Automatic evaluation shows that both variants generate explanations that are more faithful to each model’s uncertainty and agree more closely with the gold fact-checking labels than a prompting baseline that lacks conflict-/agreement-span guidance (\S\ref{sec:automatic_results}). Human study participants likewise judge \modelname\ explanations as more helpful, more informative, less redundant, and more logically consistent with the input. We also observe a trade-off between two variants of our \modelname\ framework, one attains higher faithfulness, the other higher plausibility, highlighting a promising avenue for future work to achieve both simultaneously (\S\ref{sec:automatic_results}).

\section{Method}
\subsection{Preliminaries and Overall Framework}
\label{sec:framework}

Our objective is to \emph{explain why} a LLM is uncertain about a multi-evidence fact-checking instance by grounding that uncertainty in specific agreements or conflicts within the input.

\paragraph{Problem setup.}
Each input instance is a triple $X=(C,E_1,E_2)$ consisting of a claim $C$ and two evidence pieces $E_1,E_2$. In this work, we explore the scenarios in which one claim and two or three evidence pieces are presented.\footnote{As our framework is based on the pairwise input part interaction, it is readily applicable to scenarios where more claims and evidence pieces are involved.}
For clarity, we denote their concatenation as $X=[x_1,\dots,x_{|C|+|E_1|+|E_2|}]$. The task label comes from the set
$\mathcal{Y}=\{\textsc{Supports},\textsc{Refutes},\textsc{Neutral}\}$.

\paragraph{Pipeline overview.}
Our framework comprises three stages:
\begin{enumerate}[leftmargin=1.25em, itemsep=3pt,noitemsep]
    \item \textbf{Uncertainty scoring.}  
    We compute \emph{predictive entropy} from the model’s answer logits to obtain a scalar uncertainty score $u(X)$ (\S~\ref{sec:uncertainty_score_calculation}).  
    This logit-based measure is model-agnostic.

    \item \textbf{Conflicts/Agreement extraction.}  
    We capture the agreements and conflicts most relevant to the model’s reasoning by identifying the text-span interactions between $C$, $E_1$, and $E_2$ that embody these relations (\S~\ref{sec:conflicts_extraction:method}).

    \item \textbf{Explanation generation.}  
    The model receives the extracted spans as soft constraints and produces a natural language explanation $Y_R=[y_1', \dots, y_r']$ along with its predicted label $\hat{y}$ to the identified interactions (\S~\ref{sec:nle_generation:method}).  
\end{enumerate}

\paragraph{Outputs.}
For each instance $X$, the framework returns the predicted task label $\hat{y}\in\mathcal{Y}$; the numeric uncertainty score $u(X)$; and the textual explanation $Y_R=[y_1',\dots,y_r']$ that grounds the source of uncertainty in the specific agreements or conflicts between $C,E_1,E_2$.

\subsection{Predictive Uncertainty Score Generation}
\label{sec:uncertainty_score_calculation}
To quantify model uncertainty for generating an answer label on a specific input sequence, we follow previous work~\citep{kadavath2022language,yang-etal-2025-maqa} and calculate predictive uncertainty with entropy theory, which does not require multiple runs and is widely used in open-source models.

Specifically, we define the numeric uncertainty score $u$ as the entropy of the softmax distribution over the model's output logits for a set of candidate answers $\mathcal{Y} = \{\textsc{Supports}, \textsc{Refutes}, \textsc{Neutral}\}$. For each candidate label $y_i \in \mathcal{Y}$:

\begin{align}
P(y_i \mid X) &= \frac{\exp(\mathrm{logit}(y_i))}{\sum_{j=1}^{|\mathcal{Y}|} \exp(\mathrm{logit}(y_j))}
\end{align}
where $\mathrm{logit}(y_i)$ is the model's output logit towards candidate answer $y_i$ given input $X$. $P(y_i \mid X)$ is the model’s confidence score for selecting $y_i$ as the final answer across all candidate answers within $\mathcal{Y}$.
Finally, the model's uncertainty towards the input sequence $X$ is:
\begin{equation}
u(X)= - \sum_{y_i \in \mathcal{Y}} P(y_i \mid X) \log P(y_i \mid X)
\label{eq:entropy_uncertainty}
\end{equation}

\subsection{Conflict and Agreement Span Interaction Identification for Answer Uncertainty}
\label{sec:conflicts_extraction:method}

To surface the conflicts and agreements that drive a model’s uncertainty, we extract and then label salient span interactions among the claim $C$ and two evidence passages, $E_1$ and $E_2$.

\paragraph{Span interaction extraction.} For each ordered input part pair $(F,T) \in \{(C,E_1), (C,E_2), (E_1,E_2)\}$, we follow previous work \citet{choudhury-etal-2023-explaining,sun-etal-2025-evaluating} to extract the important span interactions and their importance score for the model's answer by (i) identifying the most important attention head to the model's answer prediction from its final layer,  (ii) obtaining its attention matrix $\mathbf A \in \mathbb R^{(|F|+|T|)\times(|F|+|T|)}$, and  
(iii) symmetrizing the cross-part score $a'_{p,q}$ for each possible token interaction $(x_p,x_q)$ within $(F,T)$:
$$
a'_{p,q}= \tfrac12\bigl(\mathbf A_{p,q}+\mathbf A_{q,p}\bigr),
\quad
x_p\!\in\!F,\;x_q\!\in\!T.
$$
Based on these token interactions, we treat each $a'_{p,q}$ as an edge weight in a bipartite token graph between the two input parts $(F,T)$. Then the Louvain algorithm~\citep{blondel2008fast} is applied to search for communities of tokens with dense intra-cluster and sparse inter-cluster relationships. From each detected token community, one span interaction, $(\text{span}_w,\text{span}_v)$, can be extracted by treating the neighboring tokens of the same input part, $F$ and $T$, respectively, as a span.
By averaging the token importance score $a'_{p,q}$ of each token interaction within $(\text{span}_w,\text{span}_v)$, its importance score $a_{wv}$ can be obtained by
\begin{equation}
a_{wv}= \frac{1}{|\text{span}_w|\,|\text{span}_v|}
        \sum_{x_p\in\text{span}_w}\!
        \sum_{x_q\in\text{span}_v}\!
        a'_{p,q}.
\label{eq:span-score}
\end{equation}
The scored interactions for $(F,T)$ form
$S^{(F,T)}=\{((\text{span}_w,\text{span}_v),\,a_{wv})\}$.

\paragraph{Relation labeling.}
To tag each span interaction as \textit{agreement}, \textit{disagreement}, or \textit{unrelated},
we use a relation labeler $L_{\text{rel}}$ that can be instantiated with either an open-weight or closed LLM.
$L_{\text{rel}}$ takes the full $(C, E_1, E_2)$ context together with a candidate span pair and outputs one label in
$\{\textit{agree}, \textit{disagree}, \textit{unrelated}\}$.
We implement this using an instruction prompt (see App.~\ref{app:relation_prompt}) and report the labeling model in Sec.~\ref{sec:experiment_setup_models}.



%
After labelling all three pairs, the complete interaction set for instance $X$ is
\begin{equation}
S_R =
S_R^{(C,E_1)}\;\cup\;S_R^{(C,E_2)}\;\cup\;S_R^{(E_1,E_2)},
\end{equation}
where, for example,
$S_R^{(C,E_1)}=\{((\text{span}_w,\text{span}_v),\,a_{wv},\,r_{wv})\}$.  
Each element links two spans with an importance score and a relation label, thereby supplying the conflict- or agreement-span interactions used in later stages.

\subsection{Uncertainty Natural Language Explanation Generation}
\label{sec:nle_generation:method}

To convert the extracted conflict- and agreement-span interactions into explanations for model uncertainty, we rely on two complementary mechanisms.  
(i) \textbf{Instruction-driven prompting} embeds the spans directly in the input so the model is instructed which segments to reference.  
(ii) \textbf{Intrinsic attention steering} guides the model's own attention toward those same segments while it is generating the NLE.  
Both mechanisms use \emph{self-rationalization}: the model first states its verdict $\hat y$ and then explains $Y_R$, a sequencing shown to improve faithfulness over pipeline approaches~\citep{wiegreffe-etal-2021-measuring,marasovic-etal-2022-shot,siegel-etal-2025-verbosity}.

\paragraph{Instruction-driven NLE.}
For each instance $X$, we rank all labelled interactions by descending importance scores and keep the top $k=3$, denoted $S_R(k)$, to avoid overly long explanations.  
These three span pairs are slotted into a three-shot prompt (See App. \ref{app:baseline_prompt}), which instructs the model to explain how the highlighted agreements or conflicts influence its confidence. Finally, the standard transformer decoding process outputs both the predicted label $\hat y$ and the accompanying explanation $Y_R$.

\paragraph{Attention steering.}
Instead of explicit instructions, we can guide NLE generation by modifying attention on the fly following the attention steering method in PASTA~\citep{ICLR2024_b99d6cc4}.  
Starting from the same $S_R(k)$, we collect all token indices that fall inside any selected span,
\begin{equation}
\small
\mathcal{I}
=\bigl\{p \;:\; (\text{span}_w,\text{span}_v)\!\in\!S_R(k),\;
                 p\!\in\!\text{span}_w\cup\text{span}_v \bigr\}.
\end{equation}
For each attention head $(\ell,h)$ deemed relevant to model uncertainty, let $\mathbf A$ be its attention matrix.  We down-weight non-target tokens by $\beta$:

\begin{align}
\tilde{A}_{ij} &=
\frac{A_{ij}}{Z_i}
\begin{cases}
1      & \text{if } j\in\mathcal{I},\\
\beta  & \text{otherwise},
\end{cases}\\[4pt]
Z_i &= \sum_{j\in\mathcal{I}} A_{ij}
     + \beta \sum_{j\notin\mathcal{I}} A_{ij}.
\end{align}

All other heads remain unchanged.  Following \citet{ICLR2024_b99d6cc4}, we steer $|H| = 100$ heads and set $\beta=0.01$ to balance steering efficacy and prevent degeneration; see App.~\ref{app:attention_heads_steering_selection} for the head-selection procedure.
With the steered attention in place, the transformer generates $\hat y$ followed by the rationale $Y_R$, now naturally centered on the conflict- or agreement spans that drive its uncertainty.

\section{Experimental Setup}
\subsection{Datasets}
\label{sec:datasets}
We select two fact-checking datasets, one specific to the health domain, HealthVer ~\citep{sarrouti-etal-2021-evidence-based}, and one closer to a real-world fact-checking scenario, DRUID~\citep{hagstrom-etal-2025-reality}. These datasets were chosen because they provide multiple evidence pieces per claim, making them well-suited to our goal of explaining model uncertainty arising from the inter-evidence conflicts and agreements. For experiments, we select six hundred instances that consist of a claim and multiple pieces of evidence, and a golden label $y\in \{\textsc{Supports}, \textsc{Refutes}, \textsc{Neutral}\}$ from each dataset.\footnote{While DRUID has six fine-grained fact-checking labels, we merge the labels into the above three categories to balance the label categories.}

\subsection{Models}
\label{sec:experiment_setup_models}

We compare three generation strategies for NLEs of model uncertainty:

\begin{itemize}[leftmargin=*,noitemsep]
\item \textbf{\texttt{\baselinemodelname}}: A three-shot prompt baseline extending prior few-shot NLE work~\citep{stammbach2020fever,zeng-gao-2024-justilm,zhao-etal-2024-pacar} by explicitly asking the model to highlight conflicting or supporting spans that shape its uncertainty (see prompt template in App. \ref{app:baseline_prompt}).

\item \textbf{\texttt{\modelvariantone}}: The instruction-driven variant of our \modelname\ method where the extracted span interactions are filled into a three-shot prompt to guide the explanation generation (\S\ref{sec:nle_generation:method}; App.~\ref{app:span_prompt}).

\item \textbf{\texttt{\modelvarianttwo}}: The attention steering variant of our \modelname\ method in which the same prompt as \texttt{\modelvariantone} is used. Additional attention steering is applied to directly guide the model's explanation generation toward the identified spans (\S\ref{sec:nle_generation:method}; App. \ref{app:span_prompt}).
\end{itemize}

We run the main experiments on multiple recent open-weight LLMs of comparable scale:
Qwen2.5-14B-Instruct~\citep{qwen_qwen25_2025}, Gemma-2-9B-IT~\citep{gemma_arxiv_2024}, and OLMo-2-1124-13B-Instruct~\citep{olmo20242olmo2furious} and DeepSeek-R1-Distill-Qwen-14B~\citep{deepseek_r1_distill_qwen_14b}
(model cards in App.~\ref{app:resources}).
Each backbone is run through \modelname{} on a single NVIDIA A100-SXM 80GB GPU; we choose these models to balance
instruction-following/reasoning ability with inference efficiency (see cost report in App.~\ref{app:compute_cost}).

For span-interaction relation labeling, we instantiate $L_{\text{rel}}$ with GPT-4o~\citep{openai2024gpt4ocard}
to reduce relation-label noise (see App.~\ref{app:error_propagation} for error propagation analysis).
This step is optional: replacing $L_{\text{rel}}$ with the open-weight Qwen2.5-14B-Instruct yields comparable
downstream NLE quality with a small drop (App.~\ref{app:ablation_qwen_labeler});
we discuss this modularity and future directions in Limitations.



\section{Automatic Evaluation}
\subsection{Faithfulness}
\label{sec:faithfulness:method}

To assess whether the NLEs produced by \modelname\ are faithful to the model's uncertainty,
we adapt the Correlational Counterfactual Test (CCT)~\citep{siegel-etal-2024-probabilities}
and propose \emph{Entropy-CCT}. Entropy-CCT retains CCT's perturb-and-correlate design,
but replaces model's answer probability-shift (e.g., TVD) with \emph{predictive entropy change} to directly
target uncertainty, which is the object our NLEs aim to explain
 (see the App.~\ref{app:difference_from_cct} for comparison).

Following \citet{siegel-etal-2024-probabilities}, for each instance $X$ we construct $n$ perturbations $X'$ 
by inserting a random modifier token (adjective before a noun, or adverb before a verb; App.~\ref{app:perturbation_details}).
Let $u(X)$ denote the model's uncertainty score defined by Eq.~\ref{eq:entropy_uncertainty}.
We measure the perturbation's impact on uncertainty using Absolute Entropy Change (AEC):
\begin{equation}
\Delta u(X) = \left| u(X) - u(X^{\prime}) \right|.
\label{eq:aec}
\end{equation}

For each perturbation $X'$, we also record whether the inserted word appears in the generated NLE,
using its presence as a proxy for importance. This yields a binary mention flag
$m\in\{0,1\}$~\citep{siegel-etal-2024-probabilities,atanasova-etal-2023-faithfulness}.

Let $D_m$ denote the set of perturbed examples where the NLE \emph{mentions} the inserted word
and $D_{\lnot m}$ the complementary set where it does not.
We compute the point-biserial correlation $r_{\text{pb}}$~\citep{tate1954correlation}
between the continuous variable $\Delta u$ and the binary flag $m$ over all perturbations.
The Entropy-CCT statistic is:
\begin{equation}
\scriptstyle
\text{CCT}_{\text{entropy}} = r_{\text{pb}}=
\frac{\mathbb{E}_{m}[\Delta u] - \mathbb{E}_{\lnot m}[\Delta u]}{\mathrm{Std}(\Delta u)}
\cdot
\sqrt{ \frac{|D_m| \cdot |D_{\lnot m}|}{(|D_m| + |D_{\lnot m}|)^2} }.
\label{eq:entropy-cct}
\end{equation}

Here, $\mathbb{E}_m[\Delta u]$ and $\mathbb{E}_{\lnot m}[\Delta u]$ are the mean AEC values
for the two groups, and $\mathrm{Std}(\Delta u)$ is the standard deviation across all perturbations.
Higher Entropy-CCT indicates that the NLE more often mentions tokens whose insertion
induces larger changes in predictive uncertainty, and thus is more faithful to reflect the model's uncertainty.

\subsection{Span-Coverage}
\label{sec:spancoverage:method}

An uncertainty explanation should surface \emph{all} information conveyed by the selected span interactions.  We therefore compute \textbf{Span-Coverage}: the fraction of reference interactions that are
explicitly mentioned in the generated NLE.  Let $S_{\text{NLE}}$ be the set of span interactions extracted from the explanation, and let $S_R(k)$ be the reference set supplied in the prompt (See \S\ref{sec:nle_generation:method}).  Then
\begin{equation}
\text{Span-Coverage} \;=\;
\frac{\lvert S_{\text{NLE}} \cap S_R(k) \rvert}{\lvert S_R(k) \rvert}.
\end{equation}
A higher value indicates the NLE covers a higher proportion of the information supplied by the extracted span interactions.
\subsection{Span-Extraneous}
\label{sec:spanextraneous:method}

Ideally, the explanation should mention \emph{only} the provided interactions and avoid introducing extraneous information.  We measure the proportion of mentioned interactions that
\emph{do not} belong to the reference set, denoted
\textbf{Span-Extraneous}:
\begin{equation}
\text{Span-Extraneous} \;=\;
\frac{\lvert S_{\text{NLE}} \setminus S_R(k) \rvert }
     {\lvert S_{\text{NLE}} \rvert }.
\end{equation}
A lower value indicates closer alignment with the intended span interactions.

\subsection{Label-Explanation Entailment}
\label{sec:lee:method}
We evaluate the extent to which the uncertainty explanation agrees with the model’s predicted label by formulating the task as a natural-language inference (NLI) problem. 
First, we convert the predicted label into a hypothesis using the template 
\textit{``The claim is supported by / refuted by / neutral to the evidence.''} 
The generated explanation serves as the premise. 
The resulting premise--hypothesis pair is fed to a widely used off-the-shelf language-inference model, DeBERTa-v3~\citep{he2023debertav3improvingdebertausing}. 
The Label-Explanation Entailment (LEE) score is the proportion of examples for which the NLI model predicts \textsc{entailment}.


\subsection{Results}
\label{sec:automatic_results}

\begin{table*}[!htbp]
\centering
\begin{fitwidthtable}[\scriptsize]
\begin{tabular}{llcccccccc}
\toprule
\multirow{2}{*}{\textbf{Model}} & \multirow{2}{*}{\textbf{Method}}
& \multicolumn{4}{c}{\textbf{HealthVer}} & \multicolumn{4}{c}{\textbf{DRUID}} \\
\cmidrule(lr){3-6}\cmidrule(lr){7-10}
& & \textbf{Faith} $\uparrow$ & \textbf{Cov.} $\uparrow$ & \textbf{Ext.} $\downarrow$ & \textbf{LEE.} $\uparrow$
  & \textbf{Faith.} $\uparrow$ & \textbf{Cov.} $\uparrow$ & \textbf{Ext.} $\downarrow$ & \textbf{LEE.} $\uparrow$ \\
\midrule

\multirow{3}{*}{Qwen2.5-14B}
& \textbf{\texttt{\baselinemodelname}}
& -0.028 & \multicolumn{1}{c}{\textit{n/a}} & \multicolumn{1}{c}{\textit{n/a}} & 0.74
& -0.080 & \multicolumn{1}{c}{\textit{n/a}} & \multicolumn{1}{c}{\textit{n/a}} & 0.60 \\
& \textbf{\texttt{\modelvariantone}}
& {0.006} & {0.33} & {0.68} & {0.75}
& {0.089} & {0.20} & {0.38} & \textbf{0.78} \\
& \textbf{\texttt{\modelvarianttwo}}
& \textbf{0.033} & \textbf{0.44} & \textbf{0.53} & \textbf{0.80}
& \textbf{0.102} & \textbf{0.28} & \textbf{0.20} & {0.77} \\
\midrule
\multirow{3}{*}{OLMo-2-1124-13B}
& \textbf{\texttt{\baselinemodelname}}
& -0.100 & \multicolumn{1}{c}{\textit{n/a}} & \multicolumn{1}{c}{\textit{n/a}} & 0.55
& -0.130 & \multicolumn{1}{c}{\textit{n/a}} & \multicolumn{1}{c}{\textit{n/a}} & 0.53 \\
& \textbf{\texttt{\modelvariantone}}
& {0.005} & {0.10} & {0.83} & {0.61}
& {0.014} & {0.08} & {0.79} & {0.65} \\
& \textbf{\texttt{\modelvarianttwo}}
& \textbf{0.020} & \textbf{0.23} & \textbf{0.77} & \textbf{0.68}
& \textbf{0.099} & \textbf{0.15} & \textbf{0.70} & \textbf{0.69} \\
\midrule
\multirow{3}{*}{Gemma-2-9B-IT}
& \textbf{\texttt{\baselinemodelname}}
& -0.105 & \multicolumn{1}{c}{\textit{n/a}} & \multicolumn{1}{c}{\textit{n/a}} & 0.66
& -0.120 & \multicolumn{1}{c}{\textit{n/a}} & \multicolumn{1}{c}{\textit{n/a}} & 0.57 \\
& \textbf{\texttt{\modelvariantone}}
& {0.007} & {0.34} & {0.59} & {0.82}
& {0.043} & {0.23} & \textbf{0.43} & {0.76} \\
& \textbf{\texttt{\modelvarianttwo}}
& \textbf{0.021} & \textbf{0.39} & \textbf{0.50} & \textbf{0.85}
& \textbf{0.098} & \textbf{0.30} & {0.47} & \textbf{0.81} \\

\bottomrule
\end{tabular}
\end{fitwidthtable}
\caption{Automatic uncertainty-NLE evaluation. Higher is better for Faith.(\S\ref{sec:faithfulness:method})/Cov. (\S\ref{sec:spancoverage:method})/LEE. (\S\ref{sec:lee:method}) and lower is better for Ext. (\S\ref{sec:spanextraneous:method}).
\textit{n/a} indicates Cov./Ext. are undefined for \textbf{\texttt{\baselinemodelname}} because no reference spans are provided. Best results per metric for each dataset-model pair are in bold.}
\label{tab:healthver_druid_results}
\end{table*}

Here, we present the results of our automatic NLE evaluation for the scenario where the input consists of one \textit{Claim} and two pieces of \textit{Evidence}\footnote{See experiments with three pieces of evidence in App.~\ref{app:3_evidence}}. For brevity, we refer to Qwen2.5-14B-Instruct, OLMo-2-1124-13B-Instruct, and Gemma-2-9B-it simply as Qwen, OLMo, and Gemma, respectively\footnote{See experiments on the reasoning model DeepSeek‑R1‑Distill‑Qwen‑14B in App.~\ref{app:reasoning_model} }. 

\paragraph{Faithfulness.}
We use Entropy-CCT, a point--biserial correlation $r_\text{pb}$ (Eq.~\ref{eq:entropy-cct}) bounded by $[-1,1]$
to measure the faithfulness of NLEs to the model's uncertainty (\S\ref{sec:faithfulness:method}).  
When $r_{\text{pb}}=0$, the explanation mentions high- and low-impact
perturbation words equally often; every $+0.01$ adds roughly
\emph{one percentage point (pp)} to the chance that the explanation
names a token that is \emph{truly influential for the model’s predictive uncertainty} (App.~\ref{app:faithfulness_stats}).

Table~\ref{tab:healthver_druid_results} shows that
\textbf{\textbf{\texttt{\baselinemodelname}} is \emph{non-faithful} in all six settings} 
with
all $r_{\text{pb}}$ values negative ranging from $-0.03$ to $-0.13$.  
Thus its NLEs mention truly influential tokens 3--13\,pp
\emph{less} often than uninfluential ones-the opposite of faithful
behaviour. \textbf{Both variants of our \modelname\ reverse this trend.}
Presenting span interactions in the prompt
(\textbf{\texttt{\modelvariantone}}) raises every correlation to
non-negative values and peaks at $r_{\text{pb}} = 0.089$ on the
\textsc{DRUID}--Qwen setting.  This means the explanation now mentions
about 17\,pp more often than  \textbf{\texttt{\baselinemodelname}} ($r_{\text{pb}} = -0.080$). Adding attention steering
(\textbf{\texttt{\modelvarianttwo}}) lifts the $r_{\text{pb}}$ scores to
$0.033$ on \textsc{HealthVer} and $0.102$ on
\textsc{DRUID} with Qwen model, i.e., 
net gains of +6\,pp and +18\,pp over
\textbf{\texttt{\baselinemodelname}}.  Moreover, four of the six positive
correlations produced by \textbf{\texttt{\modelvarianttwo}} are
significant at $p<0.01$ (Table~\ref{tab:faith_sig_7200} in App.~\ref{app:faithfulness_stats}), confirming that
the improvements are both substantial and statistically reliable.
\textbf{Particularly large jumps of OLMo on Druid dataset (up to $\Delta r_{\text{pb}} = +0.23
\approx +23$\,pp)} suggest that span-interaction guidance from our \modelname\ framework is most
beneficial for models that initially struggle to align explanations with
its uncertainty.

\paragraph{Other Properties} We evaluate three further properties of the generated NLEs: (i) \textbf{Span-Coverage} of extracted conflict-/agreement- span interactions (\S\ref{sec:spancoverage:method}), (ii) \textbf{Span-Extraneous}: mention of non-extracted spans (\S\ref{sec:spanextraneous:method}), and (iii) \textbf{Label-Explanation Entailment} with the generated fact-checking label (\S\ref{sec:lee:method}). As Table \ref{tab:healthver_druid_results} shows, \textbf{\textbf{\texttt{\modelvarianttwo}} outperforms \textbf{\texttt{\modelvariantone}} in both Span-Coverage and Span-Extraneous}, consistent with the attention steering method's effectiveness in directing the model to focus on provided spans during generation \citep{ICLR2024_b99d6cc4}. Absolute numbers, however, remain modest (peak Span-Coverage: 0.44, Span-Extraneous: 0.20 with Qwen). A Span-Coverage of 1 means the NLE cites every extracted interaction, while a Span-Extraneous score of 0 means it adds none beyond them.
This gap highlights considerable headroom for better integrating critical span interactions into the explanations. Among the three backbones, \textbf{Qwen attains the highest Span-Coverage and the lowest Span-Extraneous scores}, a trend that likely reflects its stronger instruction-following ability (see benchmark scores in App.~\ref{app:llm_perf}), 
and thus larger or more capable models might further narrow the gap. \textbf{Both variants of our framework achieve stronger label-explanation entailment scores than the baseline}, yielding explanations logically consistent with the predicted labels while remaining faithful to the model’s uncertainty.

\section{Human Evaluation}

\subsection{Method}
We recruited N=12 participants from Prolific (\url{https://www.prolific.com/}) to evaluate 120 explanations generated by \textbf{\texttt{\baselinemodelname}}, \textbf{\texttt{\modelvariantone}}, \textbf{\texttt{\modelvarianttwo}} for 40 unique claims (20 from DRUID, 20 from HealthVer) (see details of participants and setup in App. \ref{app:participants_and_materials}).
Adapting \citet{atanasova-etal-2020-generating-fact}, participants ranked explanations in descending order (1$^{st}$, 2$^{nd}$, 3$^{rd}$) according to five criteria, complementary to our automatic evaluation metrics:
\begin{itemize}[noitemsep,leftmargin=*]
    \item \textbf{Helpfulness.} The explanation offers information that aids readers to fact-check the claim. 

    \item \textbf{Coverage.} The explanation captures \emph{all} salient information in the input that matters for the fact check, distinct from Span-Coverage (\S\ref{sec:spancoverage:method}), which counts overlap with pre-extracted spans.
    
    \item \textbf{Non-redundancy.} The explanation does not offer irrelevant or repetitive information to the input, distinct from Span-Extraneous (\S\ref{sec:spanextraneous:method}) which counts mentions outside the extracted spans.
    
    \item \textbf{Consistency.} The explanation contains logically consistent statements to the input, distinct from Label-Explanation Entailment (\S\ref{sec:lee:method}), which measures label-explanation alignment.
    
    \item \textbf{Overall Quality.} Ranking of explanations by considering all criteria above.

\end{itemize}

\subsection{Results}
\label{sec:human_eval_results}
Our evaluation shows that explanations generated by \modelname\ are consistently preferred to \texttt{\baselinemodelname}
(Table \ref{tab:human_eval}). Similar to prior work that has evaluated NLEs~\citep{atanasova-etal-2020-generating-fact,huang-etal-2024-chatgpt,solano-etal-2024-sparsefit}, annotator agreement was moderate to low (App. \ref{app:IAA}), which we attribute to the 
complexity of the task.

\begin{table}[t]
    \centering
    \begin{tabular}{lcc}
    \toprule
     & \textbf{\texttt{\baselinemodelname}} & \textbf{\texttt{\modelname}} \\
     \midrule
    \textbf{Helpfulness}     &  & \\
    Overall     & 0.281  & \textbf{0.73} \\
    DRUID     &  0.312 & \textbf{0.688 }\\
    HealthVer     & 0.25  & \textbf{0.772} \\
    \addlinespace[2pt]
    \textbf{Consistency}     &  & \\
    Overall     & 0.290 & \textbf{0.721}\\
    DRUID     & 0.309 & \textbf{0.691}\\
    HealthVer     & 0.27  & \textbf{0.751} \\
    \addlinespace[2pt]
    \textbf{Non-redundancy}     &  & \\
    Overall     & 0.252  & \textbf{0.762} \\
    DRUID     & 0.261 & \textbf{0.739} \\
    HealthVer     & 0.242  & \textbf{0.784}\\
    \addlinespace[2pt]
    \textbf{Coverage}     &  & \\
    Overall     & 0.275 & \textbf{0.722}\\
    DRUID     & 0.283 & \textbf{0.717}\\
    HealthVer     & 0.266 & \textbf{0.727} \\
    \addlinespace[2pt]
    \textbf{Overall Quality}     &  & \\
    Overall     & 0.325 & \textbf{0.678}\\
    DRUID     & 0.313 & \textbf{0.688}\\
    HealthVer     & 0.336 & \textbf{0.667} \\
    \bottomrule
    \end{tabular}
    \caption{Proportion of times explanations were ranked as participants' first preference for each evaluation metric. Explanations generated by \textbf{Qwen2.5-14B-Instruct} 
    (chosen for their high faithfulness; see §\ref{sec:automatic_results})}
    \label{tab:human_eval}
\end{table}

The explanations generated by our \texttt{\modelname} framework are ranked as participants' first preference 67-78\% of the time, compared to 24-34\% of the time for \texttt{\baselinemodelname}. The \textbf{explanations generated by \textbf{\texttt{\modelname}} are rated as most helpful, containing the least amount of redundant information, highest coverage, consistency, and overall quality approximately twice as often as those generated using \texttt{\baselinemodelname}}.


Although both \texttt{\modelvariantone} and \texttt{\modelvarianttwo} outperform \texttt{\baselinemodelname} in automatic metrics (\S\ref{sec:automatic_results}) and are ranked higher than \texttt{\baselinemodelname} in the human study (Table~\ref{tab:human_eval_ranks} in App.~\ref{app:CLUE_variants}),
we observe a mild \emph{faithfulness-plausibility} trade-off when comparing the two CLUE variants. \textbf{\texttt{\modelvarianttwo}} achieves the highest automatic faithfulness, whereas \textbf{\texttt{\modelvariantone}} is ranked slightly higher in human-perceived overall quality. A possible reason is that \textbf{\texttt{\modelvarianttwo}} adheres closely to the top-$k{=}3$ span interactions (higher Span-Coverage and lower Span-Extraneous), which improves faithfulness but can reduce fluency/naturalness when some extracted spans are fragmentary. In contrast, \textbf{\texttt{\modelvariantone}} may capture additional points that participants deemed important, but not captured by the extracted span interactions~\citep{choudhury-etal-2023-explaining}. This pattern aligns with the well-documented trade-off between faithfulness and plausibility~\citep{agarwal2024faithfulnessvsplausibilityunreliability,atanasova-etal-2023-faithfulness,lu2024does}, and calls for future work to enhance both aspects for improved NLE generation, e.g., by improving the quality of the extracted span interactions.


\section{Related Work}
\subsection{Uncertainty Quantification in LLMs}
Recent work estimates LLM uncertainty primarily with white-box, logit-derived measures: predictive entropy of answer distributions ~\citep{kadavath2022language,yang-etal-2025-maqa}, aggregation across generations ~\citep{malinin-gales-2021-uncertainty}, and response consistency via semantic similarity ~\citep{duan2024shifting,kuhn2023semantic,nikitin2024kernel}. For closed-source models, uncertainty is typically elicited through verbalised confidence \citep{lin2022tmlr_uncertainty,mielke-etal-2022-reducing}, which is overconfident and unreliable~\citep{yona-etal-2024-large,pmlr-v238-harsha-tanneru24a}, or approximated by diversity under paraphrased prompts ~\citep{zhang-etal-2024-luq,chen2024quantifying}, which is compute-intensive and conflates prompt noise with model uncertainty. Accordingly, our method focuses on open-source models and adopts \emph{predictive entropy}, a simple, interpretable, efficient white-box metric computed from answer logits, avoiding prompt-induced noise.

\subsection{Linguistic Expressions of Uncertainty}
Numerical uncertainty estimates are difficult for end-users to act upon as they do not address sources of uncertainty  \citep{warren2025explainablefactchecking}. Linguistic expressions of uncertainty, e.g., phrases such as ``I'm sure''~\citep{mielke-etal-2022-reducing,tian2023just,xiong2023can,ji2025calibrating,farquhar2024detecting,kim2024LLMuncertainty}, 
may be more intuitive to understand than numerical ones \citep{zimmer1983verbal,wallsten1993preferences,windschitl1996measuring}.
However, they are not necessarily faithful reflections of the model's uncertainty \citep{yona-etal-2024-large,pmlr-v238-harsha-tanneru24a}
risking misleading users \citep{steyvers2024calibrationgapmodelhuman}. Moreover, they do not explain \emph{why} the model is uncertain, as our CLUE method does.

\subsection{Generating Natural Language Explanations for Fact-Checking}
Natural language explanations (NLEs) justify model predictions for lay readers in fact-checking \citep{wei-jie-etal-2024-interpretable}, complementing summarization and social-media work that provides graphs, manipulated spans as rationales \citep{ribeiro-etal-2022-factgraph,chan-etal-2023-interpretable,huang-etal-2025-manitweet}.
In fact-checking, early systems extracted key sentences from supplied articles as explanations~\citep{atanasova-etal-2020-generating-fact}; later work improved NLE fluency via post-editing \citep{jolly2022generating} and trained models on journalist-written fact checks collected from fact-checking websites~\citep{feher2025learning,althabiti2023factex}. More recent methods use highlight-based explanations as signals to enhance NLE quality, employing techniques such as graph-based modeling~\citep{yuan-etal-2025-graph} and self-refinement~\citep{wang-atanasova-2025-self}. Another line of work focuses on few-shot approaches that avoid model supervision, including prompting GPT-3 to produce evidence summaries \citep{brown2020language,stammbach2020fever}, inserting a planning step~\citep{zhao-etal-2024-pacar}, and leveraging retrieval-augmented language models~\citep{zeng-gao-2024-justilm}. However, existing methods are often either unfaithful to model processes~\citep{atanasova-etal-2023-faithfulness,siegel-etal-2024-probabilities,siegel-etal-2025-verbosity} or fail to address model uncertainty~\citep{warren2025explainablefactchecking}, limiting their practical utility in fact-checking~\citep{schmitt2024explhuman}. Recent work in a fact-checking context has shown that explanations of model uncertainty were judged to be more helpful than providing the model verdict and numerical uncertainty alone~\citep{warren2026show}, suggesting that generating such explanations may assist people to reason about the reliability of automated fact-checking predictions.
Our framework addresses these gaps by explicitly explaining sources of uncertainty in automatic fact-checking, yielding explanations that are both faithful to model uncertainty and provide actionable insights for fact-checkers.

\section{Conclusion}
We present the first framework, \modelname, for generating NLEs of model uncertainty by referring to the conflicts and agreements between claims and multiple pieces of evidence in a fact-checking task. Our method, evaluated across three language models and two datasets, demonstrates significant improvements in both faithfulness to model uncertainty and label consistency compared to standard prompting. 
Evaluations by human participants further demonstrate that the explanations generated by \modelname\ are more helpful, more informative, less redundant, and more logically consistent with the input.
This work establishes a foundation for explainable fact-checking systems, providing end users
with grounded, faithful explanations that reflect the model's uncertainty.
Furthermore, our approach may prove useful for future work in a wide range of information-seeking and retrieval-augmented tasks (e.g., question answering) in which explaining uncertainty given conflicting context is critical.

\section*{Limitations}
\label{sec:limitations}

Our paper proposes a novel framework for generating NLEs towards the model's uncertainty by explicitly pointing to the conflicts or agreements between the claim and multiple pieces of evidence. While our framework demonstrates improved explanation quality through rigorous evaluation across multiple language models and datasets, we acknowledge several limitations. 


Our experiments are constrained to medium-sized models (Qwen2.5-14B-Instruct, Gemma2-9B-It, OLMo‑2‑1124‑13B‑Instruct, and DeepSeek‑R1‑\allowbreak Distill‑\allowbreak Qwen‑14B), which are selected based on computational limitations. Although these models show significant improvements over baseline performance, our results suggest that larger models (e.g., 70B parameter scale) may further enhance explanation quality, particularly in terms of coverage and redundancy. Our framework's modular design readily accommodates such scaling.

\modelname{} includes an auxiliary relation labeler $L_{\mathrm{rel}}$ that assigns each extracted span interaction one of \{\textsc{agree}, \textsc{disagree}, \textsc{unrelated}\}. In the main experiments we use GPT-4o for this step to reduce relation-label noise, but this step is optional: $L_{\mathrm{rel}}$ is modular, and a fully open-weight variant (e.g., Qwen2.5-14B-Instruct) yields comparable downstream NLE quality with only a small drop
(App.~\ref{app:ablation_qwen_labeler}). 
Reducing this remaining gap
while
remaining fully open-weight
may require improving relation-label prompting (more local context, higher-quality
shots) or lightly fine-tuning an open-weight model on a compact relation-judgment set
(App.~\ref{app:error_propagation_mitigation}).


Like most multi-stage pipelines, \modelname\ can occasionally propagate upstream errors: manual inspection reveals failure modes such as fragmentary spans and occasional relation mislabels, which can in turn reduce NLE faithfulness (App.~\ref{app:error_propagation}). 
These issues could be mitigated by
encouraging more complete semantic spans during extraction (e.g., mild constraints during community detection), improving relation-label robustness (prompting or lightweight fine-tuning), and adding lightweight generation-time checks that down-weight unreliable span interactions before producing the final
explanation (App.~\ref{app:error_propagation_mitigation}).

In this study, we focus on the HealthVer and DRUID datasets, which pair claims with discrete pieces of evidence, making them ideal for studying evidence-conflict scenarios. Our experiments primarily focus on setups where one claim and two or three pieces of evidence are presented. Although we did not explore setups involving more complex claim-evidence interactions, when multiple claims and/or pieces of evidence are introduced, our framework, grounded in pairwise interactions, can seamlessly accommodate more complex scenarios by simply increasing the number of input claim/evidence pieces; no methodological changes are required.

CLUE targets settings where users need to understand \emph{why} the model is uncertain by grounding uncertainty in concrete claim-evidence and inter-evidence conflicts/agreements.
This focus is motivated by qualitative findings that professional fact-checkers want systems to ``show the work'' behind a decision, especially when evidence-based tensions drive uncertainty, so that users can decide what to verify next and when to trust the output~\citep{warren2025explainablefactchecking}. While in this paper, we conduct a human study and find that uncertainty explanations generated by our proposed \modelname{} framework are more helpful than those from the simple prompting method, we do not study whether these uncertainty-oriented explanations are preferable in every context compared to alternative methods for communicating uncertainty 
(e.g., numeric confidence~\citep{zimmer1983verbal,wallsten1993preferences,van2020interpretable,liu2020intuitive} or verbal hedges~\citep{lin2022tmlr_uncertainty,mielke-etal-2022-reducing,yona-etal-2024-large,kim2024LLMuncertainty}).
In-situ evaluation with professional fact-checkers, potentially including interface-level ablation studies (e.g., tuning the number of surfaced interactions $k$ and using compact/expandable formats), would be needed to 
study the effects of uncertainty explanations on efficiency, error detection, and calibrated trust in other contexts.

Our work is limited to the scope of explaining model uncertainty arising from evidence conflicts. While this captures a critical subset of cases, real-world uncertainty may also stem from other sources, including insufficient evidence, knowledge gaps in the model, and context-memory conflicts. We view this work as a foundational step toward broader research on model uncertainty explanation.

\section*{Ethical Considerations}
\label{sec:ethics}
This work concerns automated fact-checking, which aims to reduce the harm and spread of misinformation, but nevertheless has the potential for harm or misuse through model inaccuracy, hallucination, or deployment for censorship. Our current work aims to provide explanations that allow users to examine the outputs of these systems more critically, and so we do not see any immediate risks associated with it.

Our work is limited to examining claims, evidence, and explanations in English, and so our results may not be generalisable to other languages.
As the task involved complex reasoning about technical subjects, we screened our participants to be native English speakers to ensure that they could fully understand the material and increase the chances of high-quality responses (see \ref{app:participants_and_materials} for details). However, this criteria may also introduce or reinforce existing biases and limit the generalisability of our findings.
Participants were informed about the study and its aims before agreeing to provide informed consent. No personal data was collected from participants and they received fair payment for their work (approximately 9 GBP/hour).




\section*{Acknowledgements}
$\begin{array}{l}\includegraphics[width=1cm]{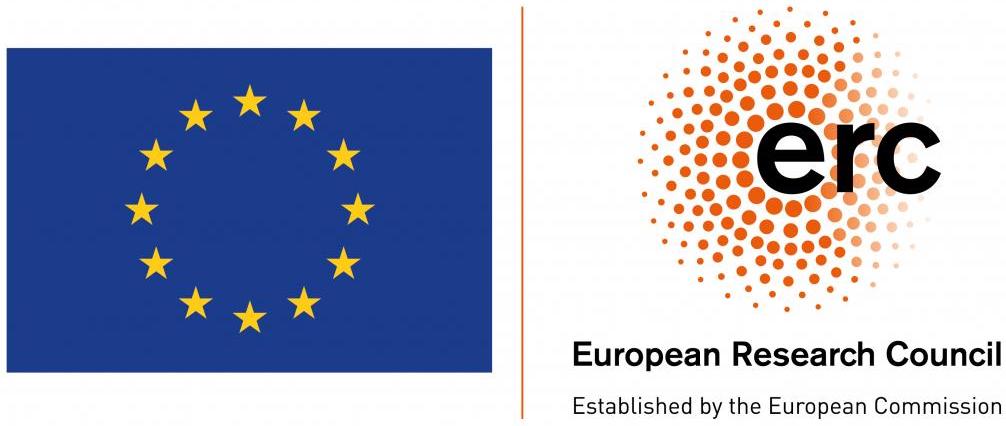} \end{array}$ 
This research was co-funded by the European Union (ERC, ExplainYourself, 101077481), by the Pioneer Centre for AI, DNRF grant number P1, as well as by The Villum Synergy Programme. Views and opinions expressed are however those of the author(s) only and do not necessarily reflect those of the European Union or the European Research Council. Neither the European Union nor the granting authority can be held responsible for them. 


\section{Appendix}
\subsection{Backbone model performance on public benchmarks}
\label{app:llm_perf}

Table~\ref{tab:llm_benchmarks} summarises the publicly reported
five-shot results on two standard reasoning benchmarks.  All figures are
taken verbatim from the official model cards or accompanying technical
reports. 


\begin{table*}[ht]
  \centering
  \small
  \setlength{\tabcolsep}{4pt}
  \renewcommand{\arraystretch}{0.95}

  \begin{tabularx}{0.98\textwidth}{
    @{}
    >{\raggedright\arraybackslash}X
    ccc
    @{}
  }
    \toprule
    \textbf{Model} & \textbf{Params} & \textbf{MMLU} & \textbf{GSM8K} \\
    \midrule
    Qwen2.5-14B-Instruct~\citep{qwen_qwen25_2025}
      & 14.7 B & 79.7 & 90.2 \\
    Gemma-2-9B-IT~\citep{gemma_arxiv_2024}
      & 9.0 B & 71.3 & 68.6 \\
    OLMo-2-1124-13B-Instruct~\citep{olmo20242olmo2furious}
      & 13 B & 67.5 & 54.2 \\
    \bottomrule
  \end{tabularx}

  \caption{Benchmark scores on MMLU~\citep{hendrycks2021measuring}
  and GSM8K~\citep{cobbe2021training} are used to characterize
  instruction-following and reasoning strength.}
  \label{tab:llm_benchmarks}
\end{table*}

These numbers corroborate our claim that Qwen2.5-14B-Instruct is the
strongest of the three for instruction-following and reasoning.

\subsection{Method: Selecting attention heads to steer}
\label{app:attention_heads_steering_selection}
Following~\citet{ICLR2024_b99d6cc4}, we steer only a selected subset of attention heads rather than all of them, because targeted steering yields larger gains in output quality.  Our selection criterion, however, differs from theirs: instead of ranking heads by their impact on task accuracy, we rank them by how strongly they affect the model’s \emph{predictive uncertainty} during fact-checking.

Concretely, for each fact-checking dataset chosen in this work (see details in \S\ref{sec:datasets}), $D$, we draw a validation subset $D_d$ with $|D_d| = 300$ examples.  For every input $X \in D_d$, we compute the model’s baseline uncertainty score $u(X)$ when it predicts the fact-checking label as stated in  \S\ref{sec:uncertainty_score_calculation}.  Then, for each attention head identified by layer $\ell$ and index $h$, we zero out that head, re-run the model, and measure the absolute change in uncertainty
$$
\Delta u(X, \ell, h) \;=\; \bigl|\,u(X)\;-\;u_{/o(l,h)}(X)\bigr|.
$$
Averaging $\Delta u(X, l, h)$ over all $X \in D_d$ yields a single importance score for head $(\ell,h)$.  We rank the heads by this score and keep the top $t$ heads for each dataset and each model. Note that we set $t = 100$ in line with the recommendation of~\citet{ICLR2024_b99d6cc4} and to balance steering effectiveness against the risk of degeneration.


\subsection{Conflict and Agreement Identification Details}
\subsubsection{Prompt Example for Assigning Relation Labels to Captured Span Interactions}
\label{app:relation_prompt}
To identify agreements and conflicts between the claim and the two evidence passages, we use the prompt in Figure \ref{fig:span-relation-prompt} to label each extracted span interaction (see \S\ref{sec:conflicts_extraction:method}).

\begin{figure}[htbp]
\centering
\begin{lstlisting}[
  basicstyle=\ttfamily\scriptsize,
  breaklines=true,
  frame=single,
  frameround=ffff
]
You are a helpful assistant. Your task:

1. Read the claim and its two evidence passages (E1, E2).
2. For each supplied span interaction, decide whether the two spans
   AGREE, DISAGREE, or are UNRELATED, taking the full context into account.
3. Output the span pairs exactly as given, followed by
   "relation: agree|disagree|unrelated".

Return format:
  1. "SPAN A" - "SPAN B"  relation: <agree|disagree|unrelated>
  2. ...
  3. ...

### SHOT 1  (annotated example)
Claim: [...]
Evidence 1: [...]
Evidence 2: [...]

Span interactions (to be labelled):
  1. "[...]" - "[...]"
  2. "[...]" - "[...]"
  3. "[...]" - "[...]"

Expected output:
  1. "[...]" - "[...]"  relation: ...
  2. "[...]" - "[...]"  relation: ...
  3. "[...]" - "[...]"  relation: ...

### SHOT 2   % omitted for brevity
### SHOT 3   % omitted for brevity

### NEW INSTANCE  (pre-filled for each new example)
Claim: {CLAIM}
Evidence 1: {E1}
Evidence 2: {E2}
Span interactions:
  1. "{SPAN1-A}" - "{SPAN1-B}"
  2. "{SPAN2-A}" - "{SPAN2-B}"
  3. "{SPAN3-A}" - "{SPAN3-B}"
\end{lstlisting}
\caption{Prompt template for span interaction relation labelling.}
\label{fig:span-relation-prompt}
\end{figure}

\subsubsection{Open-weight Alternative for Span-Interaction Relation Labeling}
\label{app:ablation_qwen_labeler}

This section studies how the choice of relation labeler affects downstream NLE quality in \ModelName{}.
While we use GPT-4o as the default relation labeler to reduce labeling noise, we show that a strong
open-weight alternative is also feasible, with only a small performance drop.

As discussed in Sec.~\ref{sec:conflicts_extraction:method}, \ModelName{} uses a relation labeler
$L_{\mathrm{rel}}$ to assign each extracted span interaction one of
$\{\textsc{agree}, \textsc{disagree}, \textsc{unrelated}\}$.
We considered two labeler options: (i) a closed-API model, which can reduce label noise
at the cost of API usage, and (ii) an open-weight model, which is fully local but may
yield slightly noisier labels and thus introduce additional noise in downstream NLE generation.

To quantify the impact of the labeler choice, we reran the pipeline with one open-weight alternative for
$L_{\mathrm{rel}}$: Qwen2.5-14B-Instruct, keeping all other components fixed.
Table~\ref{tab:labeler_ablation_qwen25} reports the results. 

\begin{table*}[!htbp]
\centering
\resizebox{\textwidth}{!}{%
\begin{tabular}{@{}l|cccc|cccc@{}}
\toprule
& \multicolumn{4}{c|}{\textbf{HealthVer}} & \multicolumn{4}{c}{\textbf{DRUID}} \\
\cmidrule(lr){2-5}\cmidrule(lr){6-9}
\textbf{Relation labeler}
& Faith. ($\uparrow$) & Span-Cov. ($\uparrow$) & Span-Ext. ($\downarrow$) & LEE ($\uparrow$)
& Faith. ($\uparrow$) & Span-Cov. ($\uparrow$) & Span-Ext. ($\downarrow$) & LEE ($\uparrow$) \\
\midrule
GPT-4o
& \textbf{0.033} & \textbf{0.44} & \textbf{0.53} & \textbf{0.80}
& \textbf{0.102} & \textbf{0.28} & \textbf{0.20} & \textbf{0.77} \\
Qwen2.5-14B-Instruct
& 0.025 & 0.41 & 0.58 & 0.78
& 0.076 & 0.25 & 0.25 & 0.75 \\
\bottomrule
\end{tabular}}
\caption{Effect of span-interaction \emph{relation labeler} choice. We rerun \ModelName\ while swapping only
$L_{\mathrm{rel}}$ (GPT-4o vs.\ Qwen2.5-14B-Instruct) and keeping all other components fixed
(backbone: Qwen2.5-14B-Instruct; variant: \modelvarianttwo{}; $k{=}3$).
An open-weight labeler is feasible with only a small drop in downstream NLE quality.
Bold indicates the better result per metric within each dataset.}

\label{tab:labeler_ablation_qwen25}
\end{table*}

Overall, Qwen2.5-14B-Instruct yields similar, though
slightly weaker, downstream NLE quality compared to GPT-4o across both datasets.
For instance, on HealthVer, Faithfulness drops from $0.033$ to $0.025$ and LEE from $0.80$ to $0.78$;
on DRUID, Faithfulness drops from $0.102$ to $0.076$ and LEE from $0.77$ to $0.75$.

\textbf{These results indicate that an open-weight labeler is feasible in \ModelName{}, and the framework is not
tied to a proprietary component.} In this work, we use GPT-4o in the main experiments because it provides
consistently stronger relation labeling and reduces the risk of error propagation from noisy relation labels
to the final explanations. We further discuss how upstream labeling errors can propagate and mitigation
strategies in App.~\ref{app:error_propagation}.

A natural concern is whether using a closed-API labeler compromises transparency.
We emphasize that \ModelName{} is \emph{white-box with respect to the backbone model}:
uncertainty scoring, span-interaction extraction, and attention steering require access to the backbone
model internals, which closed APIs do not provide.
Accordingly, running the \emph{entire} pipeline with GPT-4o would forgo these white-box signals
and reduce the method to surface-level prompting; prior work shows such verbalized uncertainty
can be unreliable, and fact-checking practice requires systems to ``show the work'' behind a decision
\citep{yona-etal-2024-large,pmlr-v238-harsha-tanneru24a,warren2025explainablefactchecking}.

GPT-4o is used only for the narrow task of pairwise relation judgment \emph{after} span interactions
have already been extracted from backbone internals; as such, it does not affect the transparency
of the core framework.





\subsubsection{Error Propagation Analysis}
\label{app:error_propagation}

We manually inspected 50 HealthVer instances (top-$3$ span pairs per instance; $N{=}150$ span pairs)
to assess (i) span extraction quality and (ii) relation labeling accuracy when using GPT-4o as $L_{\mathrm{rel}}$.

\paragraph{Span extraction quality.}
Most extracted spans are very short: 105/150 span pairs contain two-token spans (often salient nouns such as subjects/objects).
In 35/150 cases, at least one span is incomplete (e.g., ``D'' instead of ``vitamin D'') or semantically underspecified (e.g., ``is not''). Overall, the extractor tends to favor short, high-salience tokens, suggesting potential for future work to encourage more complete semantic units.

\paragraph{Relation labelling accuracy.}
On these 150 span pairs, GPT-4o achieves 86\% accuracy (129/150 correct).
Among the 21 errors, 12 are cases where an \textsc{agree} relation is mislabelled as \textsc{unrelated},
typically when one span is incomplete (e.g., ``D'' vs.\ ``vitamin D''), indicating that span completeness is a primary failure mode.

\paragraph{Propagation to downstream NLEs.}
Using Qwen2.5-14B-Instruct with CLUE-Span+Steering, 13/21 mislabeled interactions are explicitly mentioned in the generated NLEs,
showing that upstream errors can propagate into explanations.
Moreover, the subset of instances with mislabeled interactions exhibits lower faithfulness (mean Entropy-CCT 0.020)
than the overall HealthVer--Qwen CLUE-Span+Steering average (0.033; Table~1), suggesting that incorrect relation labels
can reduce explanation faithfulness.

This analysis suggests that fragmentary spans and relation-label noise can be an upstream error source and carry forward to the final NLEs, motivating mitigation strategies discussed in App.~\ref{app:error_propagation_mitigation} for future work.



\subsubsection{Mitigating Error Propagation}
\label{app:error_propagation_mitigation}

We outline several directions to mitigate error propagation in future work:
(1) \textbf{Improve span extraction:} add mild constraints during community detection so extracted spans more consistently capture complete semantic units rather than clipped tokens.
(2) \textbf{Improve relation labeling robustness:} use stronger prompting (broader local context, higher-quality example shots) or lightly fine-tune an open-weight model on a compact relation-judgment set.
(3) \textbf{Generation-time sanity checks:} add a lightweight consistency check that flags or down-weights span pairs judged internally inconsistent with the claim/evidence before producing the final NLE.

\subsection{Resources and Model Cards}
\label{app:resources}


Model cards: 
\href{https://huggingface.co/Qwen/Qwen2.5-14B-Instruct}{Qwen2.5-14B-Instruct (Hugging Face)};
\href{https://huggingface.co/google/gemma-2-9b-it}{Gemma-2-9B-IT (Hugging Face)};
\href{https://huggingface.co/allenai/OLMo-2-1124-13B-Instruct}{OLMo-2-1124-13B-Instruct (Hugging Face)};
\href{https://huggingface.co/deepseek-ai/DeepSeek-R1-Distill-Qwen-14B}{DeepSeek-R1-Distill-Qwen-14B (Hugging Face)};
\href{https://huggingface.co/MoritzLaurer/DeBERTa-v3-large-mnli-fever-anli-ling-wanli}{DeBERTa-v3 NLI (Hugging Face)};
\href{https://openai.com/index/hello-gpt-4o/}{GPT-4o (OpenAI)}.




\subsection{Computational Cost and Resource Requirements}
\label{app:compute_cost}

\paragraph{Setup.}
We report the computational overhead of \ModelName{} compared to \texttt{\baselinemodelname}.
Unless otherwise stated, measurements are collected using Qwen2.5-14B-Instruct on the HealthVer dataset
with a single NVIDIA A100 80GB GPU.
All values are reported \emph{per instance} (averaged over the evaluation set).

\subsubsection{Latency}
Table~\ref{tab:cost_latency} reports the average wall-clock time per instance for the three methods.
Both \texttt{\modelvariantone} and \texttt{\modelvarianttwo} are roughly 2 seconds slower per instance than
\texttt{\baselinemodelname}, and this overhead is entirely due to the additional span-interaction extraction and
relation labeling stage.

\begin{table*}[!htbp]
\centering
\resizebox{\textwidth}{!}{%
\begin{tabular}{@{}l|cccc@{}}
\toprule
\textbf{Approach} & \textbf{Uncertainty Scoring (s)} & \textbf{Span Extraction + Labeling (s)} & \textbf{NLE Generation (s)} & \textbf{Total (s)} \\
\midrule
\texttt{\baselinemodelname} & 0.364 & --     & 14.8969 & 15.2609 \\
\texttt{\modelvariantone}   & 0.364 & 2.0452 & 14.7730 & 17.1822 \\
\texttt{\modelvarianttwo}   & 0.364 & 2.0452 & 14.7874 & 17.1966 \\
\bottomrule
\end{tabular}}
\caption{Average time cost per instance for the three NLE generation methods on HealthVer using Qwen2.5-14B-Instruct on a single NVIDIA A100 80GB GPU. ``--'' indicates that span extraction/labeling is not used by \texttt{\baselinemodelname}.}
\label{tab:cost_latency}
\end{table*}

\subsubsection{GPU memory usage}
Table~\ref{tab:cost_vram} reports peak GPU memory usage (VRAM).
During NLE generation, VRAM usage is similar across all methods.
\texttt{\modelvarianttwo} requires about 0.32 GiB more VRAM than \texttt{\baselinemodelname}, primarily because
the prompt is longer (it includes the extracted span interactions; see App.~\ref{app:span_prompt}).
Notably, \texttt{\modelvarianttwo} does not require more VRAM than \texttt{\modelvariantone}, since during generation
KV-cache and activations dominate memory and steering does not add a separate memory footprint beyond the base forward pass.

\begin{table*}[!htbp]
\centering
\resizebox{\textwidth}{!}{%
\begin{tabular}{@{}l|ccc@{}}
\toprule
\textbf{Approach} & \textbf{Uncertainty Scoring (GiB)} & \textbf{Span Interaction Extraction (GiB)} & \textbf{NLE Generation (GiB)} \\
\midrule
\texttt{\baselinemodelname} & 29.71 & --    & 30.52 \\
\texttt{\modelvariantone}   & 29.71 & 29.72 & 30.84 \\
\texttt{\modelvarianttwo}   & 29.71 & 29.72 & 30.84 \\
\bottomrule
\end{tabular}}
\caption{Average peak VRAM per instance for the three NLE generation methods (HealthVer, Qwen2.5-14B-Instruct, NVIDIA A100 80GB).}
\label{tab:cost_vram}
\end{table*}

\subsubsection{Memory requirement for steering 100 heads}
For selecting the most important 100 attention heads (App.~\ref{app:attention_heads_steering_selection}), the peak VRAM is 29.8 GiB.
For steering these heads during NLE generation (\texttt{\modelvarianttwo}), the average peak VRAM per instance is 30.84 GiB.
Thus, a GPU with 40GB of memory satisfies the requirement for attention steering across 100 heads.

\subsubsection{GPT-4o API cost for relation labeling}
When using GPT-4o as $L_{\text{rel}}$, the total relation-labeling cost scales linearly with the number of labeled instances. In our main experiments, the cumulative cost across datasets and backbones was approximately \$30 (about \$0.003 per instance),
with an average input length of $\sim$1k tokens and an average output length of $\sim$50 tokens per call.

\subsection{Methodology Supplement: Faithfulness Evaluation Details}
\label{app:method_supp}

This supplement provides implementation and statistical details for the Entropy-CCT faithfulness
metric introduced in \S\ref{sec:faithfulness:method}. We describe (i) how counterfactual perturbations
are generated, (ii) how Entropy-CCT relates to the original CCT, and (iii) how we test statistical
significance and report results.

\subsubsection{Entropy-CCT Perturbation Generation}
\label{app:perturbation_details}
To evaluate how faithfully each NLE reflects model uncertainty, we generate multiple counterfactuals per instance, following \citet{atanasova-etal-2020-generating-fact} and \citet{siegel-etal-2024-probabilities} (see \S\ref{sec:faithfulness:method}).  For every input, comprising one claim and two evidence passages, we first tag part‐of‐speech with
spaCy~\citep{honnibal_montani_spacy2_2017} (\texttt{en\_core\_web\_sm}), then choose four random insertion sites (uniformly among tokens tagged as NOUN or VERB).  At each site we insert either (i) a random adjective before a noun or (ii) a random adverb before a verb.  The candidate modifiers are drawn uniformly from WordNet~\citep{miller_1995_wordnet} using NLTK~\citep{bird_klein_loper_2009_nltk}: we construct adjective/adverb candidate lists by collecting all lemma names from all WordNet adjective/adverb synsets, respectively.
Because we sample three random candidates for each of the four positions, this procedure yields \(4 \times 3 = 12\) perturbations per instance, providing a sufficient set for the subsequent Entropy-CCT evaluation, in which we check whether the NLE mentions the inserted word and compute the correlation between that mention and the uncertainty change induced by each perturbation.

\subsubsection{Entropy‑CCT vs. Original CCT}
\label{app:difference_from_cct}
CCT~\cite{siegel-etal-2024-probabilities} evaluates faithfulness by perturbing the input and measuring
the resulting change in the model's predicted label distribution. Specifically, it uses the Total Variation Distance (TVD) between the pre- and post-perturbation distributions $P(\cdot\mid X)$ and $P(\cdot\mid X')$:
\begin{equation}
\mathrm{TVD}(P,Q)=\frac{1}{2}\sum_i |P_i - Q_i|.
\end{equation}

Entropy-CCT is a minimal adaptation of CCT that preserves the same perturbation procedure and
the same correlational test: we correlate the perturbation impact with whether the inserted token is
mentioned in the NLE. The only change is the signal we correlate. Instead of TVD (probability-shift),
we use Absolute Entropy Change (AEC), i.e., the change in predictive entropy, to directly target
uncertainty (our object of explanation). 




\subsubsection{Statistical Testing \& Significance}
\label{app:faithfulness_stats}

We test statistical significance of $r_{\text{pb}}$ using the standard Pearson correlation $t$-test.
Each perturbation constitutes one observation: for each dataset
we evaluate 600 instances with 12 perturbations per instance, yielding $n=600\times 12=7{,}200$
observations (App.~\ref{app:perturbation_details}). Because the point-biserial correlation is
algebraically equivalent to Pearson correlation between a continuous and a binary variable, we use
the standard Pearson $t$-test for $H_0: r_{\text{pb}}=0$.



\textbf{Interpreting $r_{\text{pb}}$ and $\Delta r_{\text{pb}}$}. The Entropy-CCT score is the point-biserial correlation \citep{tate1954correlation} between the absolute entropy change $|\Delta u|$ and the binary mention flag $m$. Because it is mathematically identical to a Pearson $r$ computed between one continuous and one binary variable, it obeys $-1\le r_{\text{pb}}\le1$. When $r_{\text{pb}}=0$, it means the high- and low-impact perturbations are mentioned equally often.  If the two strata are roughly balanced, every $+0.01$ in $r_{\text{pb}}$ increases the probability that a truly uncertainty-influential token is mentioned by about one percentage point (pp).  A \emph{gain} $\Delta r_{\text{pb}}$ therefore translates to an \emph{absolute} improvement of $\approx |\Delta r_{\text{pb}}| \times 100$~pp in mention rate. For instance, moving from $-0.08$ to $+0.06$ is a swing of $0.14$, corresponding to about $14$~pp.

\textbf{Significance testing}.
Because the point-biserial is a Pearson correlation, the familiar $t$--test applies:

\begin{align}
  t &= r_{\text{pb}}\,
       \sqrt{\frac{n-2}{1-r_{\text{pb}}^{2}}}, \\[4pt]
  t &\sim t_{(n-2)}
       \qquad\text{under } H_{0}\!: r_{\text{pb}} = 0.
\end{align}

With $n=7,200$ we have $\text{df}=7,198$; the critical two-sided
values are $|t|>1.96$ for $p<0.05$ and $|t|>2.58$ for $p<0.01$.


\textbf{Faithfulness with significance results}. Table ~\ref{tab:faith_sig_7200} shows the point-biserial
coefficients $r_{\text{pb}}$, which is our faithfulness measurement for model uncertainty(See, Eq.\ref{eq:entropy-cct}), the associated $t$ statistics, and two-sided $p$ values for every model--method pair.  Values that meet the stricter
$p<0.01$ criterion are highlighted in bold.

\begin{table*}[!htbp]          
\centering
\begin{fitwidthtable}[\footnotesize]
\begin{tabular}{@{}l l r r r@{}}
\toprule
\textbf{Model} & \textbf{Method} & $r_{\text{pb}}$ & $t$ & $p$ \\
\midrule
\multicolumn{5}{c}{\textbf{HealthVer}}\\
\midrule
Qwen2.5-14B-Instruct & \textbf{\texttt{\baselinemodelname}}      & $-0.028$ & $-2.38$ & $1.7\times10^{-2}$ \\
            & \textbf{\texttt{\modelvariantone}}        & $+0.006$ & $+0.51$ & $6.1\times10^{-1}$ \\
            & \textbf{\texttt{\modelvarianttwo}}        & $+0.033$ & $+2.80$ & $\mathbf{5.1\times10^{-3}}$ \\[2pt]
OLMo-2-1124-13B-Instruct  & \textbf{\texttt{\baselinemodelname}}      & $-0.100$ & $-8.53$ & $\mathbf{<10^{-15}}$ \\
            & \textbf{\texttt{\modelvariantone}}        & $+0.005$ & $+0.42$ & $6.7\times10^{-1}$ \\
            & \textbf{\texttt{\modelvarianttwo}}        & $+0.020$ & $+1.70$ & $9.0\times10^{-2}$ \\[2pt]
Gemma-2-9B-IT  & \textbf{\texttt{\baselinemodelname}}      & $-0.105$ & $-8.96$ & $\mathbf{<10^{-15}}$ \\
            & \textbf{\texttt{\modelvariantone}}        & $+0.007$ & $+0.59$ & $5.5\times10^{-1}$ \\
            & \textbf{\texttt{\modelvarianttwo}}        & $+0.021$ & $+1.78$ & $7.5\times10^{-2}$ \\
\midrule
\multicolumn{5}{c}{\textbf{DRUID}}\\
\midrule
Qwen2.5-14B-Instruct & \textbf{\texttt{\baselinemodelname}}      & $-0.080$ & $-6.81$ & $\mathbf{9.8\times10^{-12}}$ \\
            & \textbf{\texttt{\modelvariantone}}        & $+0.089$ & $+7.58$ & $\mathbf{3.4\times10^{-14}}$ \\
            & \textbf{\texttt{\modelvarianttwo}}        & $+0.102$ & $+8.70$ & $\mathbf{<10^{-15}}$ \\[2pt]
OLMo-2-1124-13B-Instruct  & \textbf{\texttt{\baselinemodelname}}      & $-0.130$ & $-11.12$ & $\mathbf{<10^{-15}}$ \\
            & \textbf{\texttt{\modelvariantone}}        & $+0.014$ & $+1.19$ & $2.3\times10^{-1}$ \\
            & \textbf{\texttt{\modelvarianttwo}}        & $+0.099$ & $+8.44$ & $\mathbf{<10^{-15}}$ \\[2pt]
Gemma-2-9B-IT  & \textbf{\texttt{\baselinemodelname}}      & $-0.120$ & $-10.26$ & $\mathbf{<10^{-15}}$ \\
            & \textbf{\texttt{\modelvariantone}}        & $+0.043$ & $+3.65$ & $\mathbf{2.6\times10^{-4}}$ \\
            & \textbf{\texttt{\modelvarianttwo}}        & $+0.098$ & $+8.35$ & $\mathbf{<10^{-15}}$ \\
\bottomrule
\end{tabular}
\end{fitwidthtable}
\caption{Detailed faithfulness evaluation results for baseline method \textbf{\texttt{\baselinemodelname}}, and two variants of our \modelname\ framework \textbf{\texttt{\modelvariantone}} and \textbf{\texttt{\modelvarianttwo}} on Healthver and Druid dataset based on Qwen2.5-14B-Instruct(\citet{qwen_qwen25_2025}), OLMo-2-1124-13B-Instruct(\citet{olmo20242olmo2furious})and Gemma-2-9B-IT(\citet{gemma_arxiv_2024}).
Point-biserial correlation $r_{\text{pb}}$ is our Entropy-CCT measurement(\S\ref{sec:faithfulness:method}), along with $t$ statistic and two-sided $p$-value for each model--method pair ($n=7{,}200$, $df=7{,}198$). Entries with $p<0.01$ are bold.}
\label{tab:faith_sig_7200}
\end{table*}

Across both datasets and all three backbones, the \textbf{\texttt{\baselinemodelname}} exhibits negative correlations, implying a \emph{non-faithful} tendency to highlight low-impact tokens in the generated NLEs, with $\text{mean}= -0.094$.  The prompt-only variant of our \modelname\ framework \textbf{\texttt{\modelvariantone}} neutralises this bias and turns the average into $+0.027$; two of its coefficients are clear $p<0.01$, indicating a modest but significant improvement regarding faithfulness.

The full \textbf{\texttt{\modelvarianttwo}} variant pushes the mean to
$+0.062$ and achieves $p<0.01$ in four of six settings.  Interpreting
these numbers using the discussion above, the
switch from $-0.094$ to $+0.062$ yields an \emph{absolute} increase of $(0.062-(-0.094))\times100!\approx16$,pp in the probability that a truly uncertainty-influential token is named in the NLE, which is easily noticeable in
qualitative inspection.

The consistently positive, statistically significant gains therefore
substantiate the claim made in the main text: \modelname\ produces
markedly more faithful NLEs towards model uncertainty than the
\textbf{\texttt{\baselinemodelname}}, and the steer variant is particularly
beneficial for models that initially struggle with uncertainty
attribution.

\subsection{Prompt template for \textbf{\texttt{\baselinemodelname}}, \textbf{\texttt{\modelvariantone}} and \textbf{\texttt{\modelvarianttwo}} on Healthver and Druid dataset}
\label{app:explanation_prompt}
We designed two prompt templates for our experiments. The baseline prompt (Figure  \ref{fig:baseline-prompt}) gives the model no span interactions; instead, it must first identify the relevant agreements or conflicts and then discuss them in its explanation.
In contrast, the prompt used by our \modelname\ framework (Figure \ref{fig:span-prompt}) supplies the three pre-extracted span interactions (\S\ref{sec:conflicts_extraction:method}). The model is explicitly instructed to base its explanation on these spans, ensuring that the rationale remains grounded in the provided evidence.

\subsubsection{Prompt template for \textbf{\texttt{\baselinemodelname}}}
\label{app:baseline_prompt}
To generate NLEs about model uncertainty without span-interaction guidance, we craft a three-shot prompt that instructs the model to identify the interactions most likely to affect its uncertainty and to explain how these relations they represent affect it. (See Figure \ref{fig:baseline-prompt}).

\begin{figure}[htbp]  
\centering
\begin{lstlisting}[
  basicstyle=\ttfamily\scriptsize,
  breaklines=true,
  frame=single,
  frameround=ffff
]
You are a helpful assistant. Your tasks:
1. Determine the relationship between the claim and the two evidence passages.
2. Explain your prediction's uncertainty by identifying the three most
   influential span interactions from Claim-Evidence 1, Claim-Evidence 2,
   and Evidence 1-Evidence 2, and describing how each interaction's relation
   (agree, disagree, or unrelated) affects your overall confidence.
Return format: [Prediction] [Explanation]

### SHOT 1
Input
  Claim: [...]
  Evidence 1: [...]
  Evidence 2: [...]
Output
  [Prediction: ...] [Explanation: ...]

### SHOT 2   % omitted for brevity
### SHOT 3   % omitted for brevity

### NEW INSTANCE
Claim: {CLAIM}
Evidence 1: {E1}
Evidence 2: {E2}
Your answer:
\end{lstlisting}
\caption{Three-shot prompt for \textbf{\texttt{\baselinemodelname}} (Shots 2-3 omitted) on the HealthVer and DRUID datasets.}
\label{fig:baseline-prompt}
\end{figure}

\subsubsection{Prompt template for \textbf{\texttt{\modelvariantone}} and \textbf{\texttt{\modelvarianttwo}}}
\label{app:span_prompt}
To generate NLEs about model uncertainty with the span-interaction guidance, we craft a three-shot prompt that instructs the model to discuss how these interactions, along with the relations they represent, affect its uncertainty. (See Figure \ref{fig:span-prompt}).

\begin{figure}[!tbp]
\centering
\begin{lstlisting}[
  basicstyle=\ttfamily\scriptsize,
  breaklines=true,
  frame=single,
  frameround=ffff,
  numbers=none
]
You are a helpful assistant. Your tasks:
1. Determine the relationship between the claim and the two evidence passages.
2. Explain your prediction's uncertainty by referring to the three span
   interactions provided below (Claim-Evidence 1, Claim-Evidence 2,
   Evidence 1-Evidence 2) and describing how each interaction's relation
   (agree, disagree, or unrelated) affects your overall confidence.
Return format: [Prediction] [Explanation]
### SHOT 1
Input:
  Claim: [...]
  Evidence 1: [...]
  Evidence 2: [...]
  Span interactions:
    1. ``[...]'' - ``[...]''  (C-E1)  relation: [...]
    2. ``[...]'' - ``[...]''  (C-E2)  relation: [...]
    3. ``[...]'' - ``[...]''  (E1-E2) relation: [...]
Output:
  [Prediction: ...] [Explanation: ...]
### SHOT 2   % omitted for brevity
### SHOT 3   % omitted for brevity
### NEW INSTANCE
Claim: {CLAIM}
Evidence 1: {E1}
Evidence 2: {E2}
Span interactions (pre-filled):
    1. ``{SPAN1-A}'' - ``{SPAN1-B}''  (C-E1)  relation: {REL1}
    2. ``{SPAN2-A}'' - ``{SPAN2-B}''  (C-E2)  relation: {REL2}
    3. ``{SPAN3-A}'' - ``{SPAN3-B}''  (E1-E2) relation: {REL3}
Your answer:
\end{lstlisting}
\caption{Three-shot prompt for \textbf{\texttt{\modelvariantone}} and \textbf{\texttt{\modelvarianttwo}} (Shots 2-3 omitted) on the HealthVer and DRUID datasets.}
\label{fig:span-prompt}
\end{figure}

\subsection{Extended Evaluation Results}
\label{app:extended_evaluation}

\subsubsection{Extended Evaluation Results for Scenarios Involving Three Evidence pieces}
\label{app:3_evidence}

\begin{table}[t]
\centering
\scalebox{0.89}{
\begin{tabular}{lccc}
\toprule
\textbf{Metrics} 
& \textbf{\texttt{\baselinemodelnameshortened}}
& \textbf{\texttt{\modelvariantoneshortened}}
& \textbf{\texttt{\modelvarianttwoshortened}} \\
\midrule
Faith.\ ($\uparrow$)      & 0.08 & 0.10 & \textbf{0.13} \\
Span-Cov.\ ($\uparrow$)   & --   & 0.25 & \textbf{0.41} \\
Span-Ext.\ ($\downarrow$) & --   & \textbf{0.23} & 0.27 \\
LEE ($\uparrow$)          & 0.80 & \textbf{0.86} & \textbf{0.86} \\
\bottomrule
\end{tabular}}
\caption{Automatic evaluation for three types of NLE on HealthVer using \textbf{Qwen2.5-14B-Instruct} when one claim and three pieces of evidence are involved: \textbf{\texttt{\baselinemodelnameshortened}}(\textbf{\texttt{\baselinemodelname}}) is the baseline method, \textbf{\texttt{\modelvariantoneshortened}} (\textbf{\texttt{\modelvariantone}}) and \textbf{\texttt{\modelvarianttwoshortened}}(\textbf{\texttt{\modelvarianttwo}}) are the two variants of our \modelname\ framework.}
\label{tab:3_evidences}
\end{table}

To verify the generalizability of our framework in scenarios involving multiple evidence pieces, we conduct an experiment on the Healthver dataset using the Qwen2.5‑14B‑Instruct model, pairing each claim with three pieces of evidence instead of two (as shown in Tab. \ref{tab:3_evidences}). Note that we select the top‑$k=6$ span interactions for NLE generation, taking the top‑1 interaction from each pairwise combination among the $C$, $E_1$, $E_2$, and $E_3$ input parts.

Overall, we observe a similar trend with three pieces of evidence as with two (see \S\ref{sec:automatic_results}): \textbf{\modelvarianttwo} and \textbf{\modelvariantone} show better Faithfulness and Label-Explanation Entailment than \textbf{\baselinemodelname}. Interestingly, the Faithfulness of all three explanation types increases relative to the two‑evidence setting, from -0.028 to 0.08 for \textbf{\baselinemodelname}, from 0.006 to 0.10 for \textbf{\modelvariantone}, and from 0.033 to 0.13 for \textbf{\modelvarianttwo}. This may be because, when NLEs are grounded in richer evidence, they capture more nuances in both claim-evidence and inter‑evidence interactions, thereby more faithfully reflecting the model’s fact‑checking decision.

It is notable that \textbf{\modelvarianttwo} includes more targeted span interactions (higher Span‑Coverage: 0.41 > 0.27) as well as more redundant span interactions (higher Span‑Extraneous: 0.27 > 0.23) than \textbf{\modelvariantone}. This suggests that, although \textbf{\modelvarianttwo} better focuses on target interactions, it still suffers from redundancy due to the inclusion of more irrelevant interactions, which we call for future work to address.

\subsubsection{Extended Evaluation Results on Reasoning Model}
\label{app:reasoning_model}

To demonstrate that our framework generalizes to reasoning models trained primarily via large‑scale reinforcement learning, and designed to encourage chain‑of‑thought exploration, we adopt \textit{DeepSeek‑R1‑Distill‑Qwen‑14B}\footnote{\url{https://huggingface.co/deepseek-ai/DeepSeek-R1-Distill-Qwen-14B}} \citep{deepseek_r1_distill_qwen_14b} as the base model and evaluate it on the HealthVer dataset in the setting with one claim and two pieces of evidence (see Tab.~\ref{tab:reasoning_model}).

\begin{table}[!htbp]
\centering
\scalebox{0.9}{
\begin{tabular}{lccc}
\toprule
\textbf{Metrics} 
& \textbf{\texttt{\baselinemodelnameshortened}}
& \textbf{\texttt{\modelvariantoneshortened}}
& \textbf{\texttt{\modelvarianttwoshortened}} \\
\midrule
Faith.\ ($\uparrow$)      & 0.12 & 0.16 & \textbf{0.18} \\
Span-Cov.\ ($\uparrow$)   & --   & 0.58 & \textbf{0.67}\\
Span-Ext.\ ($\downarrow$) & --   & \textbf{0.59} & 0.64 \\
LEE ($\uparrow$)          & 0.88 & \textbf{0.90} & 0.89 \\
\bottomrule
\end{tabular}}
\caption{Automatic evaluation for three types of NLE on \textsc{HealthVer} using \textbf{DeepSeek-R1-Distill-Qwen-14B} when one claim and two pieces of evidence are involved: \textbf{\texttt{\baselinemodelnameshortened}}(\textbf{\texttt{\baselinemodelname}}) is the baseline method, \textbf{\texttt{\modelvariantoneshortened}} (\textbf{\texttt{\modelvariantone}}) and \textbf{\texttt{\modelvarianttwoshortened}}(\textbf{\texttt{\modelvarianttwo}}) are the two variants of our \modelname\ framework.}
\label{tab:reasoning_model}
\end{table}

Overall, both variants of our proposed CLUE model, \textbf{\texttt{\modelvariantone}} and \textbf{\texttt{\modelvarianttwo}}, show higher Faithfulness and Label‑Explanation Entailment than the baseline method \textbf{\texttt{\baselinemodelname}}, confirming the effectiveness of our framework. It is also notable that, compared with the evaluation results on other models in Tab.~\ref{tab:healthver_druid_results}, all three types of explanations generated with the DeepSeek‑R1‑Distill‑Qwen‑14B model achieve the best performance across metrics, especially the highest Faithfulness and Span‑Coverage. The former implies that reinforcement-learning-trained reasoning models may be more capable of generating highly faithful explanations with the thinking step; the latter suggests that this type of model follows our interaction-guided instructions more effectively. Consequently, both \textbf{\texttt{\modelvariantone}} and \textbf{\texttt{\modelvarianttwo}} include more target-span interactions, yielding higher Span-Coverage.

Comparing the two variants, \textbf{\texttt{\modelvarianttwo}} exhibits higher Span‑Coverage (0.67 > 0.58) than \textbf{\texttt{\modelvariantone}}, but also more Span‑Extraneous (0.64 > 0.59), suggesting that \textbf{\texttt{\modelvarianttwo}} tends to include redundant interactions while focusing on target interactions. \textbf{\texttt{\modelvarianttwo}} also achieves the highest Faithfulness (0.18 > 0.16), consistent with its higher Span‑Coverage, but yields similar Label‑Explanation Entailment to \textbf{\texttt{\modelvariantone}} (0.89 vs.\ 0.90), likely due to the increased Span‑Extraneous. Addressing this trade‑off is left for future work.

\subsection{Human Evaluation Details}
\subsubsection{Participants and Materials}
\label{app:participants_and_materials}
\paragraph{Participants}
We recruited N=12 participants from \href{https://www.prolific.com/}{Prolific}, screened to be native English speakers from Australia, Canada, Ireland, New Zealand, the United Kingdom, and the United States. The study was approved by our institution’s
Research Ethics Committee (reference number anonymised).

\begin{figure}[!t]
    \centering
    \includegraphics[width=\linewidth,height=0.44\textheight,keepaspectratio]{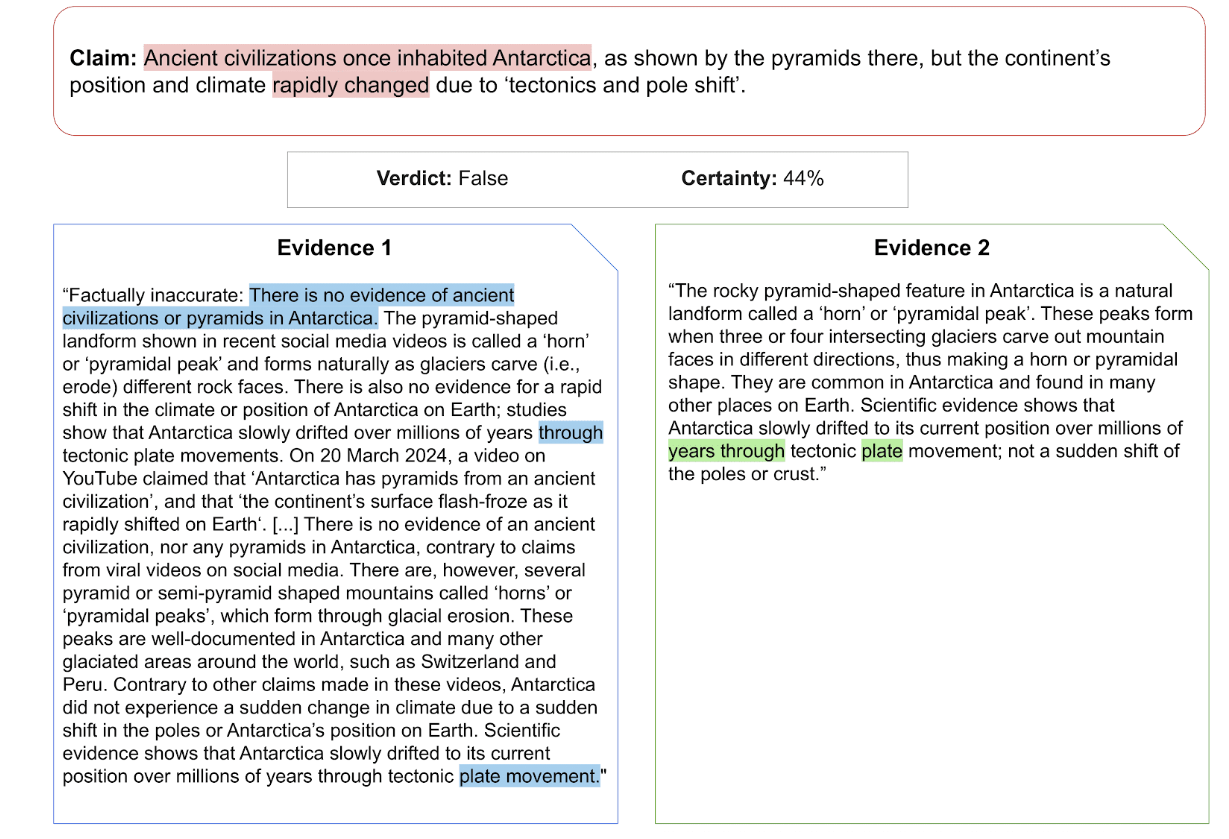}
    \medskip
    \includegraphics[width=\linewidth,height=0.44\textheight,keepaspectratio]{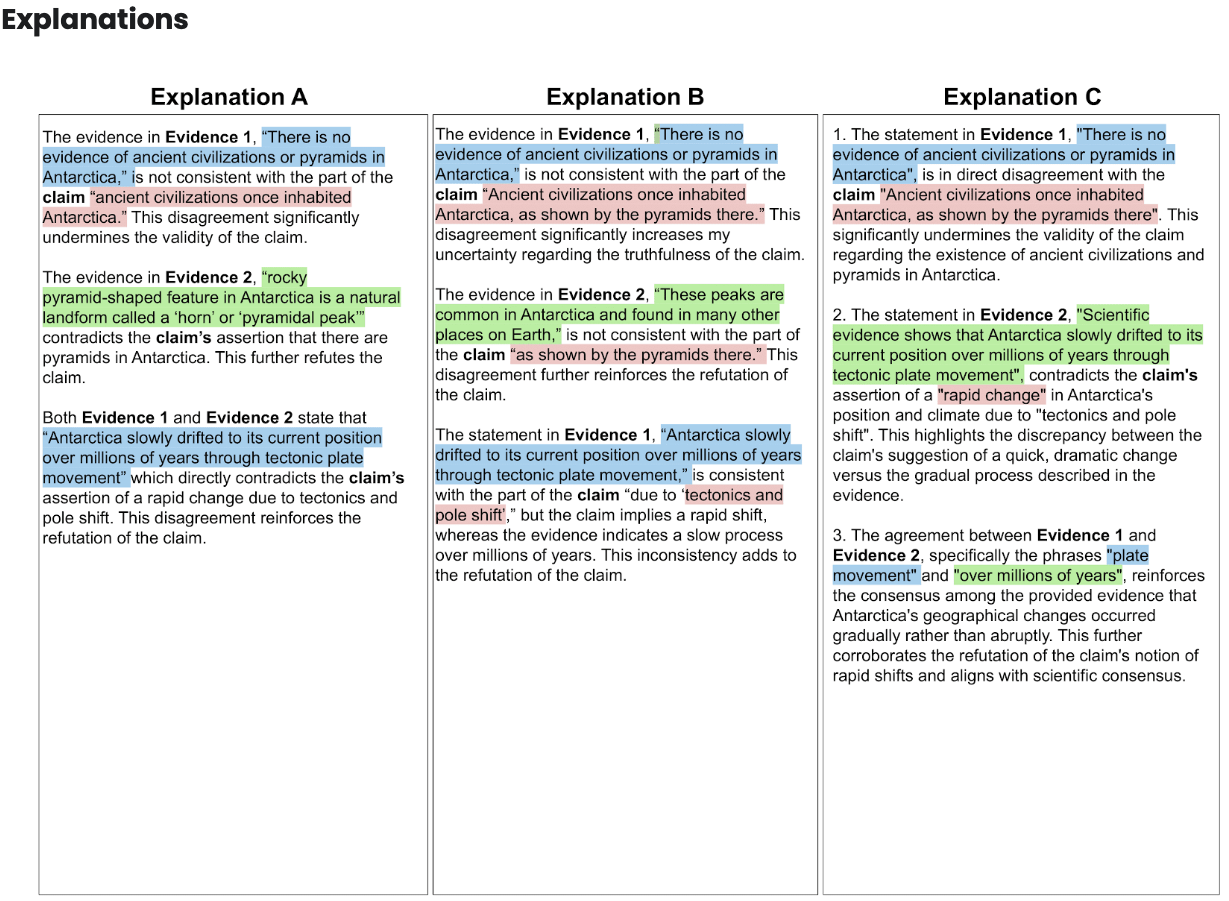}
    \caption{Example of human evaluation setup. Explanation A was generated using \baselinemodelname, Explanation B by \modelvariantone, and Explanation C by \modelvarianttwo}
    \label{fig:human_eval}
\end{figure}

\paragraph{Materials} 
Forty claims (20 from DRUID, 20 from HealthVer) were selected at random for evaluation, while 120 explanations were evaluated in total.
For each instance, participants were provided with a claim, two evidence documents, model verdict, model numerical certainty, and three alternative explanations (see Figure \ref{fig:human_eval} in \ref{app:example_annotation}).
The explanations presented to participants were those generated using Qwen2.5-14b-instruct \cite{qwen_qwen25_2025} based on its automatic evaluation performance.

Each participant evaluated explanations for 10 instances (5 labelled `True', 5 labelled `False'), in addition to two attention check instances which were used to screen responses for quality.
To mitigate the difficulty and cognitive demands of the task, we kept the number of instances per participant low, and removed participants who failed attention checks from the data.
In order to minimise the risk of bias,
each explanation presented to participants was assigned a neutral label: ‘Explanation A’, ‘Explanation B’, ‘Explanation C’, so that participants could not infer anything about how the explanations were generated or indicate that one should be favoured over another.
Each explanation was also presented in a different, random order for each instance (e.g., ABC for Claim 1, CAB for Claim 2, etc.) to mitigate order effects such as primacy bias or recency bias. The order in which the response options ‘Explanation A’, ‘Explanation B’, ‘Explanation C’ also appeared in the same random order.

\paragraph{Procedure} Participants read information about the study (see \ref{app:instructionseval}) and provided informed consent (see \ref{app:consentform}) before reading detailed task instructions and completing a practice example of the task (see \ref{app:clue_instructs}). 
Participants then progressed through the study at their own pace.
The task took approximately 20 minutes, and participants were paid £3 for their work.

\subsubsection{Human Evaluation Results} \label{app:eval_results}
\paragraph{Interrater agreement}
\label{app:IAA}

In line with similar NLE evaluations carried out by previous studies (e.g., \cite{atanasova-etal-2020-generating-fact}), interrater agreement (Kendall's W \cite{kendall1939problem}) was moderate to low (see Table \ref{tab:IAA}). We attribute this to the relative complexity of the task and individual differences in how the information was perceived.

\paragraph{CLUE Variant preferences}
\label{app:CLUE_variants}
Table \ref{tab:human_eval_ranks} shows the mean rank assigned by participants to each explanation variant.
The explanations generated by \modelname\ were preferred by our participants to those generated using \textbf{\texttt{\baselinemodelname}}: the \textbf{explanations generated by \textbf{\texttt{\modelvarianttwo}} were rated as most helpful, highest coverage, and containing the least amount of redundant information}, while \textbf{those from \texttt{\modelvariantone} were judged to have the highest consistency and overall quality}.
We observed slight variation between datasets: \textbf{\texttt{\modelvarianttwo}} tended to be rated higher than \textbf{\texttt{\modelvariantone}} for DRUID, and vice versa for HealthVer.
This may arise from differences in length and complexity of the input: DRUID evidence documents, retrieved from heterogeneous online sources and often consisting of longer form new articles, may have benefited from attention steering more than HealthVer evidence documents which consist of focused, shorter extracts from scientific abstracts.

\begin{table}[!htbp]
\centering
\scalebox{0.98}{
\begin{tabular}{lrrrr}
\hline
                & \multicolumn{2}{c}{\textbf{DRUID}}                    & \multicolumn{2}{c}{\textbf{HealthVer}}                \\
\textbf{}       & \multicolumn{1}{c}{Set A} & \multicolumn{1}{c}{Set B} & \multicolumn{1}{c}{Set A} & \multicolumn{1}{c}{Set B} \\ \hline
Helpfulness     & .016                     & .079                     & .003                     & .013                     \\
Consistency     & .44                      & .058                     & .017                    & .016                     \\
Non-redundancy  & .005                     & .084                     & .005                     & .019                     \\
Coverage        & .494                     & .113                     & .018                     & .027                     \\
Overall Quality & .005                     & .158                     & .01                      & .002                     \\ \hline
\end{tabular}
}
\caption{Interrater agreement (Kendall's W) for human evaluation}
\label{tab:IAA}
\end{table}

\begin{table}[!htbp]
\centering
\scalebox{0.9}{
\begin{tabular}{lccc}
\toprule
& \textbf{\texttt{\baselinemodelnameshortened}} 
& \textbf{\texttt{\modelvariantoneshortened}} 
& \textbf{\texttt{\modelvarianttwoshortened}} \\
\midrule
\multicolumn{4}{l}{\textbf{Helpfulness}} \\
Overall   & 2.025                                 & 1.892                             & \textbf{1.867}                      \\
DRUID     & 1.9                                   & 1.917                             & \textbf{1.767}                      \\
HealthVer & 2.15                                  & \textbf{1.867}                    & 1.967                               \\
\modelblocksep
\multicolumn{4}{l}{\textbf{Consistency}} \\
Overall   & 1.875                                 & \textbf{1.783}                    & 1.817                               \\
DRUID     & 1.717                                 & 1.75                              & \textbf{1.617}                      \\
HealthVer & 2.033                                 & \textbf{1.817}                    & 2.017                               \\ 
\addlinespace
\multicolumn{4}{l}{\textbf{Non-redundancy}} \\
Overall   & 2.05                                  & 1.908                             & \textbf{1.833}                      \\
DRUID     & 1.983                                 & 1.983                             & \textbf{1.683}                      \\
HealthVer & 2.117                                 & \textbf{1.833}                    & 1.983                               \\
\addlinespace
\multicolumn{4}{l}{\textbf{Coverage}} \\
Overall   & 1.967                                 & 1.775                             & \textbf{1.758}                      \\
DRUID     & 1.767                                 & 1.75                              & \textbf{1.617}                      \\
HealthVer & 2.167                                 & \textbf{1.8}                      & 1.9                                 \\
\addlinespace
\multicolumn{4}{l}{\textbf{Overall Quality}} \\
Overall   & 1.967                                 & \textbf{1.908}                    & 1.925                               \\
DRUID     & 1.9                                   & 1.9                               & \textbf{1.817}                      \\
HealthVer & 2.033                                 & \textbf{1.917}                    & 2.033                               \\
\bottomrule
\end{tabular}}
\caption{Mean Average Rank (MAR) for the five human-evaluation criteria applied to explanations from \textbf{Qwen2.5-14B-Instruct} on the \textsc{HealthVer} and \textsc{DRUID} datasets (chosen for its high faithfulness; see §\ref{sec:automatic_results}).  
\textbf{\texttt{\baselinemodelname}}, \textbf{\texttt{\modelvariantone}} (\textbf{\texttt{\modelvariantoneshortened}}), and \textbf{\texttt{\modelvarianttwo}} (\textbf{\texttt{\modelvarianttwoshortened}}) are compared.  
Lower MAR means a better (higher) average rank; the best score in each row is boldfaced.}
\label{tab:human_eval_ranks}
\end{table}

\subsubsection{Human Evaluation Information Screen}
\label{app:instructionseval}

Thank you for volunteering to participate in this study! Before you decide whether you wish to take part, please read this information screen carefully. 

\noindent \textbf{1. What is the project about?}

\noindent Our goal is to make sure that AI fact-checking systems can explain the decisions they produce in ways that are understandable and useful to people. This survey is part of a project to help us understand 
[anonymised]

\noindent \textbf{2. What does participation entail?}

\noindent You are invited to help us explore what kinds of explanations work better in fact-checking. In this task you will see claims, an AI system’s prediction about whether this claim is true or false and corresponding evidence used to make the prediction. You will also see an explanation for why the AI system is certain or uncertain about its prediction to help you decide how to interpret the true/false prediction. We ask you to evaluate the explanations along 5 different dimensions (the detailed explanation of the task is on the next page).
All participants who complete the survey will receive a payment of £3. There is no cost to you for participating. You may refuse to participate or discontinue your involvement at any time without penalty. 

\noindent \textbf{3. Source of funding} \\
\noindent This project has received funding from [anonymised]

\noindent \textbf{4. Consenting to participate in the project and withdrawing from the research} \\
\noindent You can consent to participating in this study by ticking the box on the next page of the study.  Participation in the study is completely voluntary. Your decision not to consent will have no adverse consequences. Should you wish to withdraw during the experiment you can simply quit the webpage. All incomplete responses will be deleted. After you have completed the study and submitted your responses, it will no longer be possible to withdraw from the study, as your data will not be identifiable and able to linked to you.

\noindent \textbf{5. Possible benefits and risks to participants}

\noindent By participating in this study you will be contributing to research related to understanding what kinds of explanations are useful to people who use or who are impacted by automated fact checking systems. This is a long-term research project, so the benefits of the research may not be seen for several years. It is not expected that taking part will cause any risk, inconvenience or discomfort to you or others.

\noindent \textbf{6. What personal data does the project process?}

\noindent The project does not process any personal data.

\noindent \textbf{7. Participants’ rights under the General Data Protection Regulation (GDPR)}

\noindent As a participant in a research project, you have a number of rights under the GDPR. Your rights are specified in 
[anonymised]

\noindent \textbf{8. Person responsible for storing and processing of data}

\noindent[anonymised]



\noindent Please click 'Next' to read more about consenting to participate in the study.

\subsubsection{Human Evaluation Consent Form}
\label{app:consentform}

We hereby request your consent for processing your data. We do so in compliance with the General Data Protection Regulation (GDPR).
See the information sheet on the previous screen for more details about the project and the processing of your data.

\begin{itemize}[leftmargin=*]
    \item I confirm that I have read the information sheet and that this forms the basis on which I consent to the processing of my data by the project.

    \item I hereby give my consent that  [anonymised]
    may register and process my data 

    \item I understand that any data I provide will be anonymous and not identifiable to me.

    \item I understand that my anonymous response data will be retained by the study team.

    \item I understand that after I submit my responses at the end of the study, they cannot be destroyed, withdrawn, or recalled, because they cannot be linked with me.

    \item I understand that there are no direct benefits to me from participating in this study

    \item I understand that anonymous data shared through publications or presentations will be accessible to researchers and members of the public anywhere in the world, not just the EU.

    \item I give my consent that the anonymous data I provided may be stored in a database for new research projects after the end of this project.

    \item I give permission for my anonymous data to be stored for possible future research related to the current study without further consent being required.

    \item I understand I will not be paid for any future use of my data or products derived from it.
\end{itemize}

\noindent By checking this box, I confirm that I agree to the above and consent to take part in this study.

$\Box$ I consent

\subsubsection{Evaluation Task Instructions}
\label{app:clue_instructs}

\textbf{What do I have to do?}\\
In this study you will see claims, an AI system’s prediction about whether this claim is true or false, how certain the system is about its label, and the corresponding evidence used to make the prediction. You will also see three different explanations for why the AI system is certain or uncertain about its prediction. These explanations are intended to help you decide how to interpret the true/false prediction. \\
Your task is to \textbf{evaluate the quality of the explanations} provided, \textbf{not} the credibility of the claims and evidence.
\\

\noindent \textbf{What information will I be shown?}\\
You will be shown examples of claims, evidence document, verdicts and explanations.

\begin{itemize}[leftmargin=*]
    \item A claim is some statement about the world. It may be true, false, or somewhere in between.
    \item Additional information is typically necessary to verify the truthfulness of a claim - this is referred to as evidence or evidence document. An evidence document consists of one or several sentences extracted from an external source for the particular claim. In this study, you will see two evidence documents that have been retrieved for a claim. These evidence documents may or may not agree with each other.
    \item Based on the available evidence, a verdict is reached regarding whether a claim is true or false.
    \item Uncertainty often arises when evaluating the claim and evidence to reach a verdict. Each verdict is accompanied by a numerical uncertainty score which represents the AI system's confidence that its predicted verdict is correct.
    \item You will see 3 alternative explanations for where uncertainty arises with regard to the verdict. Note that these explanations focus on the AI system's uncertainty, not the verdict itself.

    \item You are asked to evaluate the explanations according to 5 different properties. The properties are as follows: 

    \noindent\textbf{Helpfulness.} The explanation contains information that is helpful for evaluating the claim and the fact check.

    \noindent\textbf{Coverage. }The explanation contains important, salient information and does not miss any important points that contribute to the fact check. 

    \noindent\textbf{Non-redundancy. }The explanation does not contain any information that is redundant/repeated/not relevant to the claim and the fact check. 
    
    \noindent\textbf{Consistency.} The explanation does not contain any pieces of information that are contradictory to the claim and the fact check. 
    
    \noindent\textbf{Overall Quality. }Rank the explanations by their overall quality. 
    
    \noindent\item Please rank the explanations in descending order. For example, you should rank the explanation that you think is most helpful as ‘1’, and the explanation that you think is least helpful as ‘3’. If two explanations appear almost identical, you can assign them the same ranking, but as a general rule, you should try rank them in hierarchical order.

    \noindent\item The three explanations, Explanation A, Explanation B, and Explanation C, will appear in a different order throughout the study, so you may need to pay some attention to which is which.
\end{itemize}

\textbf{Important:}
Please only consider the provided information (claim, evidence documents, and explanations) when evaluating explanations. Sometimes you will be familiar with the claim, but we ask you to approach each claim as new, whether or not you have seen it before. It doesn’t matter whether you personally agree or disagree with the claim or evidence - we are asking you to evaluate what the AI produces: if you were to see this claim for the first time, would you find the explanation provided by the AI useful?
On the next page, you will see an example of the task.

\subsubsection{Example of human evaluation setup}
\label{app:example_annotation}

Here is an example of what you will see during the study. First, you will see a \textbf{Claim}, and two pieces of \textbf{Evidence}, along with an AI system's predicted \textbf{Verdict} and the system's \textbf{Certainty} that its prediction is correct.

The \textbf{parts of the claim and evidence that are most important to the AI system's certainty are highlighted.} Parts of the Claim are Red, parts of Evidence 1 are Blue, and parts of Evidence 2 are Green.

Underneath, you will see \textbf{three alternative explanations for the AI system's certainty}, Explanation A, Explanation B, and Explanation C. The parts of each explanation that refer to the claim and evidence are colour coded in the same way (Claim = Red, Evidence 1 = Blue, Evidence 2 = Green).

Your task is to read the claim, evidence, and explanations, and rank each explanation based on five properties.

Now, you can try this example below!


\part{Evaluating Faithfulness and Utility of Free-Text Explanations}

\chapter[Interplay between Contextual and Parametric CoT Faithfulness]
        {Investigating the Interplay between Contextual and Parametric CoT Faithfulness under Optimisation}
\chaptermark{Interplay between Contextual and Parametric CoT Faithfulness}
\label{chap:CoT_FaithMate}

\renewcommand{\thefootnote}{\fnsymbol{footnote}}
\renewcommand*{\thefootnote}{\arabic{footnote}}

This chapter consists of the following paper, which is archived as a preprint and currently under review\footnote{No changes have been made to the scientific content of the original paper. The paper is presented in a single-column
format used throughout the thesis, and its references have been merged into the bibliography at the end of the thesis.}


\begin{enumerate}
    \item \bibentry{sun2026investigatinginterplaycontextualparametric}.
\end{enumerate}

\clearpage

\begin{abstract}
    Chain-of-Thought (CoT) faithfulness, i.e., whether CoTs genuinely reflect large language models' (LLM) underlying behavior, is typically evaluated with metrics under two disjoint paradigms: \textit{contextual faithfulness}, measured by perturbing the input or CoT trace, and \textit{parametric faithfulness}, assessed by intervening on a model's parametric knowledge. Yet prior work compares them only descriptively. 
    We fill this gap by proposing \our, a unified preference-alignment interface for optimizing models towards either faithfulness paradigm.
It enables us to investigate the interplay between the two paradigms, examining whether and to what extent faithfulness gains generalize within and across paradigms. Across three models, two datasets, and six faithfulness metrics, we find that the two paradigms are positively coupled, yet asymmetric: 
    optimizing towards parametric faithfulness yields consistent gains across both paradigms, whereas the contextual counterpart delivers more variable gains. Within the contextual paradigm, faithfulness gains on one metric do not consistently transfer to others, 
    implying that existing contextual metrics capture disjoint facets of faithfulness and exposing inherent trade-offs. 
    These findings imply that CoT faithfulness is not a monolithic objective and therefore requires multifaceted optimization and evaluation.
\end{abstract}
\section{Introduction}
CoT elicits intermediate reasoning steps that have proven effective for improving LLMs' performance on various reasoning tasks \cite{kojima2022large, wei2022chain}. 
Yet, questions have been raised about CoT faithfulness, i.e., whether CoTs accurately reflect the model's actual reasoning process.
In practice, CoTs can appear plausible to users while being weakly coupled with a model's internal computations \cite{ agarwal2024faithfulnessvsplausibilityunreliability,arcuschin2025chainofthought, zaman-srivastava-2026-chain}. \looseness=-1

\begin{figure*}[t!]
\centering
\resizebox{0.88\textwidth}{!}{
\begin{minipage}{\textwidth}
\includegraphics[width=\textwidth]{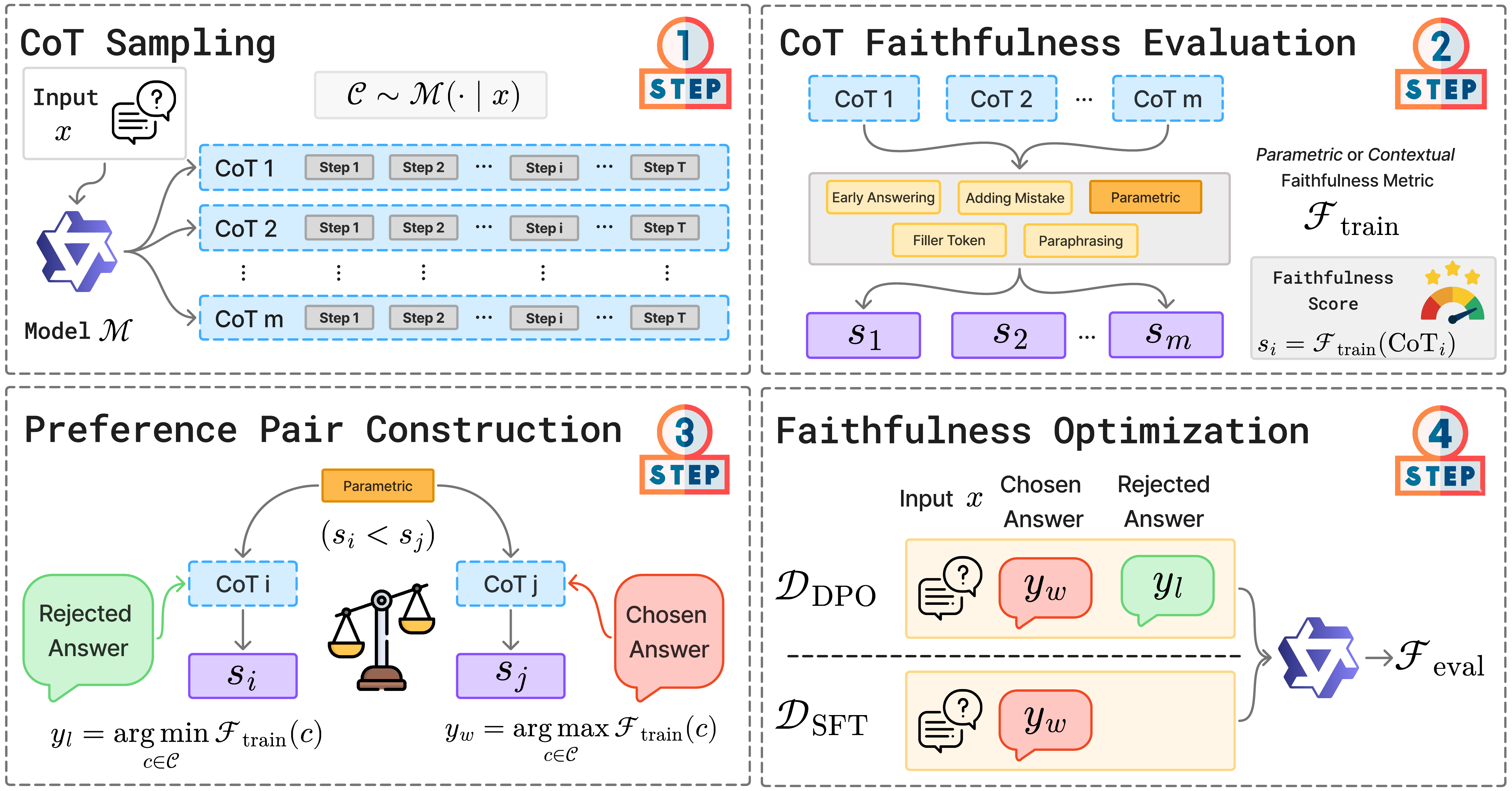}
\end{minipage}
}
\caption{Overview of our optimization-based meta-evaluation framework. We sample CoTs from a base model, score them using contextual or parametric faithfulness metrics, construct preference pairs, and optimize the model towards selected faithfulness objectives. We then evaluate cross-metric transfer across both faithfulness dimensions.}
\label{fig:pipeline}
\end{figure*}

Existing work evaluates CoT faithfulness primarily through two paradigms. \textit{Contextual faithfulness} intervenes on the input and reasoning trace in various ways, e.g., by inserting an error \cite{lanham2023measuring}, or injecting answer hints into the context \cite{NEURIPS2023_ed3fea90, chen2025reasoning}, and deems a CoT faithful if the intervention is truly reflected. 
However, such measures only consider contextual consistency instead of model internals. 
\textit{Parametric faithfulness}, in contrast, intervenes on a model's parameters rather than its context: it deems a CoT step faithful if unlearning the knowledge underlying that step changes the answer~\citep{tutek-etal-2025-measuring}. While this perspective targets model internals directly, it is limited to the knowledge explicitly verbalized by the CoT, neglecting information that shapes CoT generation but remains unexpressed.


Although the two paradigms have largely been studied in isolation,
they can yield divergent faithfulness conclusions for the same CoTs \cite{zaman-srivastava-2026-chain}, raising the question of whether they capture a shared notion of faithfulness or reward different properties of CoTs. 
As methods for improving CoT faithfulness proliferate, observational comparisons on fixed CoTs cannot reveal what each faithfulness paradigm rewards under optimization or whether and to what extent optimizing one transfers to the other.

To this end, we introduce \our (\textbf{Faith}fulness \textbf{M}etric \textbf{A}lignment and \textbf{T}ransfer \textbf{E}valuation), a unified framework for investigating CoT faithfulness under optimization. 
It constructs faithfulness-targeted preference pairs for optimizing contextual or parametric CoT faithfulness (Figure~\ref{fig:pipeline}). 
This unified optimization interface enables investigation of the optimization-based interplay between the two paradigms, while also allowing us to compare what each 
rewards. 
Experiments on three models, two datasets, and six faithfulness metrics reveal the following findings. 

First, across faithfulness paradigms, contextual and parametric faithfulness exhibit a positively coupled, yet asymmetric 
relationship (\S\ref{subsec:transfer_across_paradigm}): optimizing for parametric faithfulness delivers more consistent improvements across both paradigms, while contextual optimization, although occasionally larger in magnitude, is less consistent. \textit{This positions parametric faithfulness as the more reliable target for broad generalization.}


Second, 
within the contextual faithfulness paradigm, faithfulness gains from optimizing one contextual metric do not consistently transfer to others, e.g., between Adding Mistake and Paraphrasing (\S\ref{subsec:transfer_contextual}). This suggests that contextual faithfulness metrics are better understood as a family of related but distinct properties, rather than interchangeable measurements of a single latent quantity. \textit{Thus, we call attention to potential tensions among contextual faithfulness metrics.}
Additional experiments show that task-arithmetic merging of models, each optimized for a specific faithfulness target, 
can partially mitigate this inconsistency and amplify the faithfulness improvements. 


Third, we qualitatively characterize what different metrics actually reward under optimization (\S\ref{subsec:qualitative}). We find that contextual metric-shared gains are typically driven by improved alignment between reasoning and final answer, whereas contextual metric-specific gains are mostly prompted by the mitigation of post-hoc rationalization.\footnote{{A model decides on the answer and then generates a plausible justification that explains the answer rather than reflects how it actually arrived at it~\cite{NEURIPS2023_ed3fea90}.}} By contrast, parametric gains 
arise from more explicit grounding in task-relevant facts and knowledge. \textit{This suggests that evaluating and reporting gains on a single faithfulness metric can be misleading, as improvements may reflect distinct underlying behaviors that are not interchangeable.}





\section{Related Work}
\label{sec:related_work}
\paragraph{CoT Faithfulness Metrics.} 
Prior work has mainly studied \textit{contextual faithfulness}, which 
intervenes on the CoT or input and checks whether the intervention is reflected in the answer.
Corruption-based methods perturb the CoT, 
and measure whether the model's answer changes~\citep{lanham2023measuring, zaman-srivastava-2025-causal}. Attribution-based methods measure the 
alignment between input contributions to CoT generation and those to answer prediction
~\citep{parcalabescu-frank-2024-measuring, admoni-etal-2026-aligning}.
Another line of work uses Biasing Features, which inject answer hints and check whether the CoT verbalizes them~\citep{NEURIPS2023_ed3fea90,chen2025reasoning,chua2025deepseekr1reasoningmodels,young2026measuring}. 
Recent work further proposes \textit{parametric faithfulness}, which intervenes on a model's parametric knowledge rather than on the CoT.
\citet{tutek-etal-2025-measuring} unlearn the knowledge involved in individual CoT steps from the model parameters and measure whether the answer changes after unlearning. 
Given that contextual and parametric metrics target different aspects of CoT faithfulness, we evaluate both families within our unified \our framework under optimization.\looseness=-1

\paragraph{Meta-Evaluation of CoT Faithfulness.} 
\citet{parcalabescu-frank-2024-measuring} compare contextual faithfulness metrics and find that they often assign divergent faithfulness labels to the same CoTs, suggesting that they are better understood as measures of output-level self-consistency. \citet{zaman-srivastava-2025-causal} use model editing to construct faithful--unfaithful CoT pairs and evaluate whether the target faithfulness metrics rank the faithful CoTs higher. 
More recent work extends such comparisons to parametric faithfulness. 
\citet{zaman-srivastava-2026-chain} show that contextual and parametric faithfulness metrics can yield different conclusions on the same CoTs. 
Such observational comparisons, however, leave open how the two paradigms interact when models are explicitly optimized for faithfulness. 
\our{} fills this gap with a unified interface for optimizing either paradigm, enabling controlled study of faithfulness improvement transfers across metrics and paradigms.








\paragraph{CoT Faithfulness Improvement.} 
Several methods have been proposed to improve CoT faithfulness. Nevertheless, they are designed to \textit{optimize only one specific type of faithfulness}. FRODO trains the model to prefer correct reasoning chains over counterfactual ones \cite{paul-etal-2024-making}. \citet{li-etal-2025-towards-better} use integrated-gradient attribution to recall question information missing from the CoT, and weight sampled CoTs by their information gain when voting for the final answer. FRIT optimizes towards faithful CoTs identified via causal importance -- whether interventions on reasoning steps change the answer \cite{swaroop2025frit}. \citet{hoth2026atmanrl} leverage differentiable attention manipulation to learn faithful reasoning through reinforcement learning. In comparison, \our does not focus on better results in individual faithfulness metrics; rather, it serves as a universal preference optimization framework applicable to both \textit{contextual} and \textit{parametric} faithfulness, facilitating the study of how the two paradigms interact. 



\section{\our Framework}
\our comprises four key steps: CoT sampling, faithfulness evaluation, preference pair construction, and faithfulness optimization (Figure~\ref{fig:pipeline}).

\subsection{CoT Sampling}
In the first step, for each data point $x \in \mathcal{D}$ from a given dataset $\mathcal{D}$, we sample a set of $m$ CoTs, denoted $\mathcal{C}=\{c_{1}, ..., c_{m}\}$, potentially representing varying levels of CoT faithfulness. A specific CoT $c_{k}$ comprising $T$ steps $c_{k}={(s_1,...,s_T)}$, is sampled from a model $\mathcal{M}$ (Table~\ref{tab:cot-config}): 
\begin{align}
    c_{k} \sim \mathcal{M}(\cdot \mid x)
\end{align}

\subsection{CoT Faithfulness Evaluation}
\label{subsec:cot_faithfulness_eval}

For each CoT $c_k$, we assess faithfulness using a suite of metrics $\mathcal{F}$ (\S\ref{subsec:cot_metrics}). For each optimization run, one specific metric $f_i \in \mathcal{F}$ is selected as the \textbf{training-time metric}, i.e., $\mathcal{F}_{\text{train}}=\{f_i\}$, which is then used to rank preference pairs (\S\ref{subsec:preference_pair_ranking}). The remaining metrics are treated as \textbf{evaluation-only metrics}, $\mathcal{F}_{\text{eval}}=\mathcal{F}\setminus\mathcal{F}_{\text{train}}$.



\subsection{Preference Pair Construction}
\label{subsec:preference_pair_ranking}
For each input $x$ and sampled CoT set $\mathcal{C}=\{c_{1},...,c_{m}\}$, CoT faithfulness preferences are ranked according to a specific training-time faithfulness metric $\mathcal{F}_{\text{train}}$ (\S\ref{subsec:cot_faithfulness_eval}). The most faithful CoT is selected as the preferred response $y_{w}$, while the least faithful CoT is selected as the rejected response $y_l$
\footnote{We use the strongest contrastive signal to facilitate effective DPO training.}:
\begin{align}
\label{eq:argmax}
    y_w &= \mathop{\arg\max}_{c \in \mathcal{C}} \mathcal{F}_{\text{train}}(c) \\
    \label{eq:argmin}
    y_l &= \mathop{\arg\min}_{c \in \mathcal{C}} \mathcal{F}_{\text{train}}(c)
\end{align}

\subsection{CoT Faithfulness Optimization}
\label{subsec:model_training}
We employ two \textit{offline} learning strategies due to the high computational cost of faithfulness metrics during inference: supervised fine-tuning (SFT) and direct preference optimization (DPO) \cite{rafailov2023direct}.\footnote{Detailed information about training strategies and training details are provided in Appendix~\ref{app:training_paradigm} and~\ref{app:training}, respectively.} From Section~\ref{subsec:preference_pair_ranking}, we obtain training sets $\mathcal{D}_{\text{SFT}}=\{(x_i,y_w^i)\}_{i=1}^{N}$ and $\mathcal{D}_{\text{DPO}}=\{(x_i,y_w^i, y_l^i)\}_{i=1}^{N}$ for SFT and DPO, respectively, which are then used to optimize CoT faithfulness.  

\subsection{Faithfulness Gain Transfer}
\label{subsec:cot_faithfulness_final_eval}
After optimizing the model towards $f_i$, we evaluate it on $\mathcal{F}_{\text{train}}\cup\mathcal{F}_{\text{eval}}$ to measure direct gains on the target metric $f_i$ and transfer to non-target metrics\looseness=-1.
\footnote{We repeat this procedure for each feasible metric in $\mathcal{F}$ (\S\ref{subsec:cot_metrics}), 
varying $\mathcal{F}_{\text{eval}}$ with optimization target to study the interplay among faithfulness metrics under optimization.}

Specifically, to assess how model optimization towards $f_i \in \mathcal{F}_\text{train}$ is beneficial for improving faithfulness on the evaluation-only metrics $f_j \in \mathcal{F}_\text{eval}$, we define \textbf{faithfulness gain transfer}:\looseness=-1
\begin{equation}
\label{eq:faithfulness_transfer}
    \delta(f_i, f_j) = f_j(c^{'}) - f_j(c)
\end{equation}
where $c \sim \mathcal{M}(\cdot \mid x)$ and $c^{'}\sim \mathcal{M}^{'}_{f_i}(\cdot \mid x)$ are CoTs from the base model $\mathcal{M}$ and the $\mathcal{M}^{'}_{f_{i}}$ model optimized towards $f_i$.
$\delta$ ranges from -2 to +2, and a positive score means optimizing $f_i$ can enhance a model's faithfulness on $f_j$. Thus, a high $\delta$ indicates strong gain transfer between metric $f_i$ and $f_j$.

\section{Experimental Setup}

\subsection{Datasets}
Consistent with prior CoT faithfulness evaluation studies \cite{paul-etal-2024-making, tutek-etal-2025-measuring, zaman-srivastava-2025-causal}, we select two widely used reasoning datasets (see examples in App.~\ref{app:dataset}):

\paragraph{\data{OpenBookQA}} \cite{mihaylov-etal-2018-suit} is a question answering dataset designed to test scientific reasoning. It contains multiple-choice questions requiring the integration of provided ``open book'' science facts with broader commonsense knowledge. \looseness=-1 

\paragraph{\data{LogiQA}} \cite{liu2020logiqa} is a dataset designed to evaluate logical reasoning in natural language understanding. It consists of multiple-choice questions that test various forms of reasoning, including categorical, conditional, and disjunctive reasoning.


\subsection{Models}
In our experiments, we employ three widely used open-source instruction-tuned models from distinct families with varying parameter sizes (Appendix~\ref{app:model}): \lm{Gemma3-4B} \cite{gemmateam2025gemma3technicalreport}, \lm{Qwen2.5-7B} \cite{qwen_qwen25_2025} and \lm{Llama3.1-8B} \cite{grattafiori2024llama3herdmodels}.

\subsection{CoT Faithfulness Evaluation Metrics}
\label{subsec:cot_metrics}
We employ different metrics for contextual (\S\ref{subsubsec:contextual}) and parametric (\S\ref{subsubsec:parametric}) faithfulness, respectively, following \citet{tutek-etal-2025-measuring}.

\subsubsection{Contextual Faithfulness}
\label{subsubsec:contextual}
Contextual faithfulness is assessed in the literature by examining reasoning--answer consistency (\S\ref{sec:related_work}). 
Commonly used \textit{Corruption-based} and \textit{Attribution-based} contextual faithfulness metrics are included in this study.\looseness=-1\footnote{We intentionally exclude Biasing Features~\citep{NEURIPS2023_ed3fea90}, as they are sensitive to how hint acknowledgment is identified~\citep{young2026measuring}.}

\paragraph{Corruption-based.} \citet{lanham2023measuring,zaman-srivastava-2025-causal} propose four approaches to corrupt CoTs: (i) \textit{Early Answering}, which truncates the CoT before the original answer is reached to elicit an early response; (ii) \textit{Adding Mistake}, which injects a mistake into one CoT step  
and regenerates the remainder; (iii) \textit{Paraphrasing}, which 
rephrases the first part of the CoT while preserving the original meaning and regenerating the rest; (iv) \textit{Filler Token}, which replaces the CoT with meaningless symbols such as ellipses. Faithful CoTs should lead to answer changes after corruption, except for (iii) \textit{Paraphrasing}, where the model is expected to derive the same answer. To measure this change with greater granularity rather than treating faithfulness as a binary property of the final answer \cite{lanham2023measuring}, following \citet{zaman-srivastava-2025-causal}, we employ the \textit{change in prediction score} for the top-predicted answer \textit{before} and \textit{after} corruption, denoted as $\hat{z}_i$ and $\hat{z}_i^{'}$, respectively: 
\begin{align}
    \hat{z}_i = \frac{\exp(p_\theta(L_i \mid x))}{\sum_{L_j \in \mathcal{L}} \exp(p_\theta(L_j \mid x))}
\end{align}
where $L$ denotes the set of possible labels. The faithfulness score is then calculated as $(\hat{z}_i -\hat{z}_i^{'}) \in [-1,1]$.\footnote{For Paraphrasing, we use 
$(\hat{z}_i -\hat{z}_i^{'})$.
Appendix~\ref{app:prompt_templates} provides details on the implementation of CoT corruptions.}
Larger positive margins indicate a bigger drop in the original answer's probability after corruption, and, thus, a more faithful CoT. \looseness=-1


\paragraph{Attribution-based.} An intuition behind this line of approach 
is that a truly consistent CoT should depend on the same key input tokens that drive the prediction \cite{parcalabescu-frank-2024-measuring, admoni-etal-2026-aligning}. A representative instantiation of this idea is CC-SHAP \cite{parcalabescu-frank-2024-measuring}\footnote{Due to exceptionally high computational cost, we only employ CC-SHAP as $\mathcal{F}_\text{eval}$ instead of $\mathcal{F}_\text{train}$. See computational cost for all used metrics in App.~\ref{app:metric_cost}.}, which uses Shapley values to measure each input token's contribution to the predicted answer and to the generated CoT separately, and then computes the cosine similarity between the two contribution distributions. This assesses how well the importance of inputs for the answer aligns with their importance in the CoT, \textit{without requiring any modification or perturbation of the model's input}. Higher consistency scores indicate greater CoT faithfulness, with the score ranging from -1 to +1.

\subsubsection{Parametric Faithfulness}
\label{subsubsec:parametric}
In contrast, parametric faithfulness evaluates CoT faithfulness through interventions on models' parametric knowledge.
Specifically, given an input $x$, the target CoT $c$ is first decomposed into $T$ reasoning steps $c=(s_1,...,s_T)$. $\mathrm{ff}_{\mathrm{hard}}$ provides a binary answer as to whether a CoT is faithful by measuring whether unlearning the knowledge used by models in any CoT step causes models to change their output prediction \cite{tutek-etal-2025-measuring} (App.~\ref{app:parametric_metric_details}).
\begin{align}
    \mathrm{ff}_{\mathrm{hard}} = \mathbb{I}[\exists \, s_i \text{ such that } y \neq y^{(i)*}]
\end{align}
where $s_i$ is the $i$-th reasoning step, $y^{(i)*}$ denotes the prediction made by $\mathcal{M}^{(i)*}$ after unlearning knowledge involved in the $i$-th reasoning step, and $\mathbb{I}(\cdot)$ represents the indicator function. 

To enable more fine-grained evaluation, we further propose $\mathrm{ff}_{\mathrm{cont}}$, which extends $\mathrm{ff}_{\mathrm{hard}}$ and calculates the proportion of reasoning steps for which unlearning leads the model to change its prediction: 
\begin{align}
    \mathrm{ff}_{\mathrm{cont}} = \frac{\sum_{i=1}^{T} \mathbb{I}[s_i \text{ such that } y \neq y^{(i)*}]}{T}
\end{align}
$\mathrm{ff}_{\mathrm{cont}}$ is therefore a continuous signal ranging from 0 to 1, as opposed to $\mathrm{ff}_{\mathrm{hard}}$. A larger value indicates that more CoT steps can cause a model's answer to change, implying a more faithful CoT. 




\section{Results}

\begin{figure*}[t]
    \centering
    \begin{subfigure}{\textwidth}
        \centering
        \includegraphics[width=0.95\textwidth]{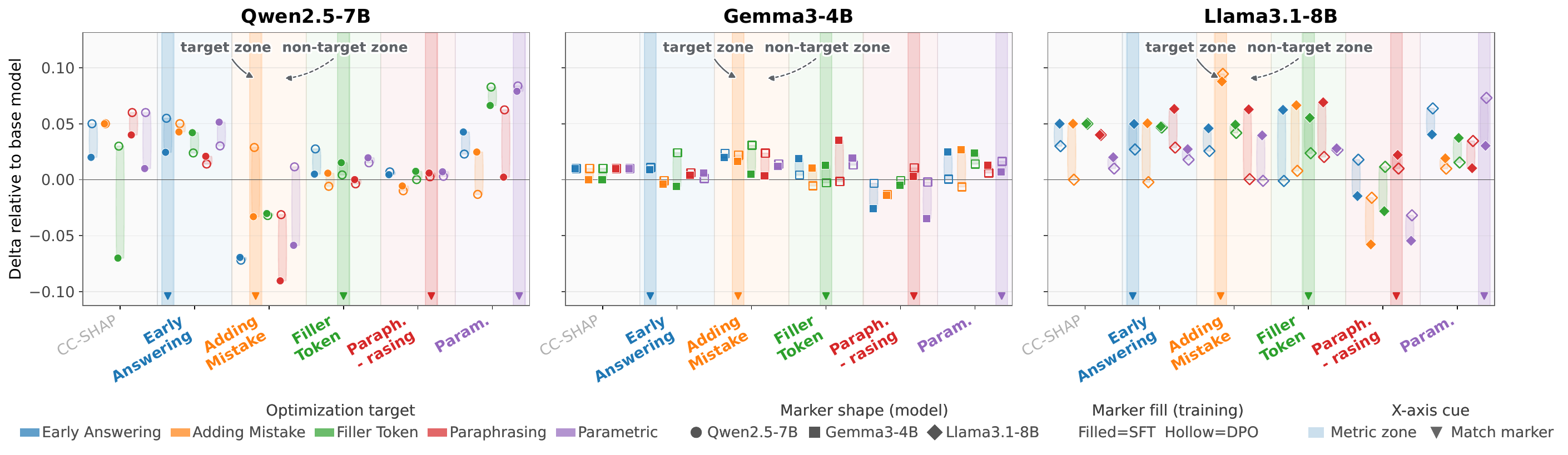}
        \caption{\data{OpenBookQA}}
        \label{fig:openbookqa-meta-faith}
    \end{subfigure}


    \begin{subfigure}{0.95\textwidth}
        \centering
        \includegraphics[width=\textwidth]{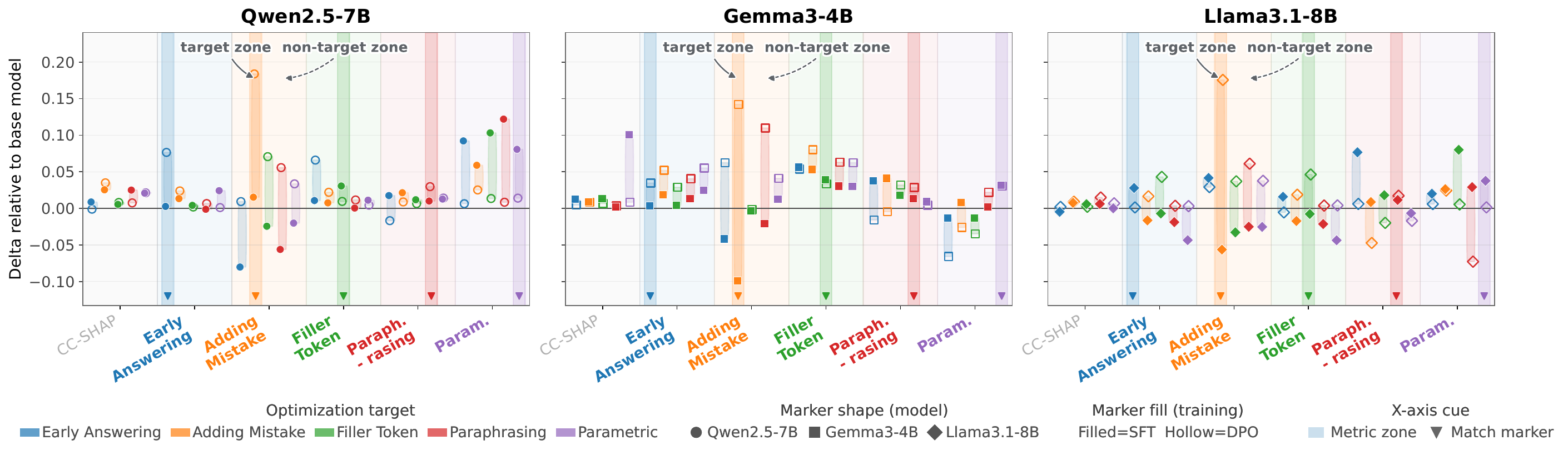}
        \caption{\data{LogiQA}}
        \label{fig:logiqa-meta-faith}
    \end{subfigure}

    
    \caption{
Cross-metric faithfulness-transfer deltas on \data{OpenBookQA} (top) and \data{LogiQA} (bottom) (raw values in Tables~\ref{tab:automatic_evaluation_openbookqa}
and~\ref{tab:automatic_evaluation_logiqa}). The $x$-axis shows evaluation metrics. Colors indicate optimization targets. Points denote deltas from the corresponding base model; filled and hollow markers indicate SFT and DPO. Translucent violins show the spread of deltas for each target--evaluation pair, and downward triangles mark matched target--evaluation settings. Bootstrap confidence interval results are presented in Appendix~\ref{app:bootstrap_ci}.
}
    \vspace{-1em}
    \label{fig:meta-faith-combined}
\end{figure*}
\label{sec:results}


We analyze two key questions: 
(1) whether a model's faithfulness on one specific metric is improved by optimizing towards another, i.e., faithfulness gain transfer $\delta$ (\S\ref{subsec:cot_faithfulness_final_eval}), especially when they are from distinct faithfulness paradigms (\S\ref{subsec:transfer_across_paradigm}); and (2) whether this faithfulness gain transfer $\delta$ remains positive across metrics within the contextual paradigm (\S\ref{subsec:transfer_contextual}). We also examine how these trends vary across base models and training paradigms, and how task performance is affected under faithfulness optimization 
\footnote{Direct target optimization can generally improve faithfulness on the matched metric, 
supporting our framework as a reliable basis for cross-metric comparison (App.~\ref{app:subsec_targeted_optimization}).}(\S\ref{subsec:other_factors}).
Finally, we conduct a qualitative analysis to examine what each faithfulness paradigm enhances (\S\ref{subsec:qualitative}).

\subsection{Faithfulness Gain Transfer Across Paradigms: Parametric $\Leftrightarrow$ Contextual}
\label{subsec:transfer_across_paradigm}

We first investigate the faithfulness gain transfer $\delta$ between contextual and parametric faithfulness (Figure~\ref{fig:meta-faith-combined}; Table~\ref{tab:ctx-eval-transfer}). Recall that $\delta>0$ means positive faithfulness improvement on one metric after optimizing towards another (\S\ref{subsec:cot_faithfulness_final_eval}).  



\paragraph{Contextual $\Rightarrow$ Parametric.}  

Contextual faithfulness gains $\delta$ can positively transfer to parametric faithfulness in 40/48 <model, contextual target, training paradigm, dataset> cases, indicating that optimizing contextual metrics can generally improve parametric faithfulness. Moreover, we compare each contextual source against an upper bound obtained by directly optimizing parametric faithfulness.
On \data{OpenBookQA}, Filler Token and Early Answering are the strongest ($\delta = +0.0399$ and $+0.0325$), followed by Paraphrasing ($+0.0214$) and Adding Mistake ($+0.0101$). Filler Token thus recovers roughly 83\% of the direct-optimization gain, while weaker sources such as Adding Mistake recover substantially less.
This pattern holds on \data{LogiQA}: Filler Token again emerges as the strongest source ($\delta = +0.0257$, recovering about 79\% of the direct-optimization gain), whereas Early Answering produces the weakest transfer ($\delta = +0.0076$, recovering only about 23\%). \looseness=-1
\textit{These results indicate that the magnitude of contextual-to-parametric gain transfer exhibits high variance, and that certain contextual metrics can achieve parametric faithfulness improvements close to those of direct parametric optimization.}

\paragraph{Parametric $\Rightarrow$ Contextual.} The reverse direction yields more consistent transfer gains. 
Optimizing for parametric faithfulness improves contextual faithfulness in 24/30 <model, training paradigm, contextual metric> cases on \data{OpenBookQA} and 23/25 cases on \data{LogiQA}.\footnote{The \lm{Llama3.1-8B} SFT configuration is excluded from \data{LogiQA} aggregation (see discussion in \S\ref{subsec:other_factors}).}
Parametric optimization transfers most consistently to Early Answering and Filler Token, each improving in all 11 <model, training paradigm, dataset> cases. Relative to direct optimization, their cross-dataset average gains $\delta$ recover 82.9\% and 85.7\% of the corresponding direct gains, respectively (Table~\ref{tab:ctx-eval-transfer}). A plausible explanation is that parametrically faithful CoTs are likely contextually faithful as well (in roughly 95\% of cases, App.~\ref{app:why}). \textit{This suggests that parametric optimization is a more reliable target for broad faithfulness generalization.
}

\paragraph{Parametric $\Leftrightarrow$ Contextual.} 
Overall, \textit{contextual and parametric faithfulness are positively coupled, yet asymmetric}: (parametric $\Rightarrow$ contextual) 
yields more consistent gains in contextual faithfulness, 
whereas (contextual $\Rightarrow$ parametric) transfer is stronger per-metric but with greater variance across metrics. 
From a practitioner's standpoint, this asymmetry suggests \textit{parametric faithfulness as the more generalizable optimization target.}

\subsection{Faithfulness Gain Transfer Within the Contextual Paradigm}
\label{subsec:transfer_contextual}

\subsubsection{Among Corruption-based Metrics}
We first ask whether contextual faithfulness assessed by one metric can be improved after optimization towards another
within the corruption-based contextual faithfulness paradigm.
Figure~\ref{fig:meta-faith-combined} illustrates that 
optimizing one of these metrics improves a single non-target metric in 45/72 and 51/72
<4 target metrics, 3 non-target metrics, 3 models, 2 training paradigms> configurations on \data{OpenBookQA} and \data{LogiQA}, respectively
, \textit{indicating non-trivial within-paradigm transfer}. 

\paragraph{Asymmetry as training target ($\mathcal{F}_{\text{train}}$).}
\textit{The four contextual metrics differ in how well they propagate gains to others}. On \data{OpenBookQA}, when employed as $\mathcal{F}_{\text{train}}$, Paraphrasing and Filler Token are the most effective training targets, with highest positive-transfer count of 14/18 and 12/18 <model, training paradigm, non-target contextual metrics> and mean $\delta$ of +0.0125 and +0.0126, respectively (Table~\ref{tab:ctx-eval-transfer}). 
Adding Mistake is the weakest, transferring in only 7/18 cases. On \data{LogiQA}, Early Answering achieves the largest mean transfer gain (13/18; mean $\delta=+0.0179$), while Adding Mistake has the highest positive-transfer
count (14/18; mean $\delta=+0.0167$). Filler Token and Paraphrasing follow, each yielding a positive transfer gain in 12/18 cases, with mean $\delta$ around $+0.01$. \textit{These distinct rankings suggest that the extent of transfer within the contextual faithfulness paradigm largely depends on the dataset.}\footnote{We hypothesize that 
for Adding Mistake, \data{OpenBookQA}'s reliance on associative factual recall allows the model to bypass corrupted reasoning steps, limiting transfer to non-target metrics.} 


\paragraph{Asymmetry as evaluation target ($\mathcal{F}_{\text{eval}}$).} 

The asymmetry also appears for  $\mathcal{F}_{\text{eval}}$: Early Answering and Filler Token are the most stable receivers, each improving in 28/36 <model, training paradigm, dataset, optimization-target> cases when other contextual metrics are optimized. Paraphrasing is least stable, improving in only 19/36 cases.
The most asymmetric effect appears between Adding Mistake and Paraphrasing: Paraphrasing decreases in 
8/12
(model, training paradigm, dataset) cases when Adding Mistake is the optimization target, 
suggesting that the two metrics may be in tension, not orthogonal (Figure~\ref{fig:pca_llama}). 
One plausible explanation is that Paraphrasing rewards \textit{invariance} under meaning-preserving reformulation, whereas the other three all reward \textit{sensitivity} to perturbations.\footnote{Paraphrased CoTs largely maintain the semantic meaning of the original ones, so that this is not an artifact of the Paraphrasing generator (App.~\ref{app:paraphrasing_quality}).}
Critically, these capabilities may be \textit{asymmetric}: invariance is foundational and supports sensitivity, but training for sensitivity can actively harm invariance. \textit{Thus, contextual faithfulness metrics share signal, but may not be interchangeable proxies for a single underlying notion of faithfulness.}


\subsubsection{Between Corruption-based and Attribution-based Metrics} 
\label{subsubsec:ccshap}
We next ask whether transfer holds across fundamentally different operationalizations of contextual faithfulness, by including CC-SHAP (\S\ref{subsubsec:contextual}), 
in contrast to the corruption-based metrics studied above.
Figure~\ref{fig:meta-faith-combined} reveals that CC-SHAP can generally be improved by \our when optimizing either contextual or parametric faithfulness. 
Prior work reports moderate correlation between CC-SHAP and intervention-based contextual metrics \cite{parcalabescu-frank-2024-measuring}; our results extend this correspondence from measurement-time correlation to optimization-time transfer. 
This provides preliminary evidence that \textit{contextual faithfulness gains may transfer beyond a single operationalization paradigm, at least for CC-SHAP.}

\subsection{Modulating Factors in Faithfulness Optimization and Task Performance}
\label{subsec:other_factors}
\paragraph{Variation across base models.}
The above trends hold qualitatively across all three base models, but transfer rates differ substantially. Within-paradigm transfer among the four corruption-based contextual faithfulness metrics succeeds in 18/24 cases for \lm{Llama3.1-8B} (75\%), compared with 14/24 (58\%) for \lm{Qwen2.5-7B} and 13/24 (54\%) for \lm{Gemma3-4B}.  
The contextual-to-parametric transfer is strongest for \lm{Llama3.1-8B} (8/8 cases), followed by \lm{Qwen2.5-7B} and \lm{Gemma3-4B} (7/8 cases each).
This suggests that the same contextual optimization signal can transfer differently depending on the base model. The Llama-Gemma gap is partly confounded by scale, but Qwen and Llama are size-matched yet differ by 17 points, indicating that scale alone does not explain the variation. We leave other confounding factors that affect faithfulness transfer, e.g., post-training recipe, to future work.

\paragraph{Training paradigm also matters.} Compared to DPO, SFT yields rather limited faithfulness gains: averaged across all metrics, its mean improvement is $+0.013$, about two-thirds of DPO's $+0.020$, and occasionally harms faithfulness. The gap between SFT and DPO is most pronounced on \data{LogiQA} with \lm{Llama3.1-8B}. These results demonstrate that simply maximizing the likelihood of highly faithful CoTs is insufficient to effectively improve faithfulness; the contrastive signal provided by DPO, which distinguishes higher- from lower-faithfulness CoTs, better guides the model towards faithful reasoning \cite{admoni-etal-2026-aligning}.



\paragraph{Faithfulness optimization largely preserves accuracy, and often helps.} 

Table~\ref{tab:automatic_evaluation_openbookqa} shows that on \data{OpenBookQA}, task accuracy improves over the base model in all configurations for both SFT and DPO, with SFT producing larger average gain ($+7.83\%$ vs. $+6.30\%$). On \data{LogiQA}, the picture is more mixed (Table~\ref{tab:automatic_evaluation_logiqa}): accuracy improves in 17/30 configurations, 
with average gains of +0.06\% for SFT and +0.07\% for DPO.
A plausible explanation is that \data{LogiQA} 
is more challenging,
given the substantially lower base-model accuracy. Overall, \textit{improving faithfulness tends to maintain or even boost overall task performance.}

\begin{table}[t]
\centering
\renewcommand{\arraystretch}{1}
\begin{tabularx}{0.85\columnwidth}{
    @{}
    >{\bfseries\raggedright\arraybackslash}p{0.22\columnwidth}
    >{\raggedright\arraybackslash}X
    @{}
}
\toprule

Question
& To improve health, what is a good strategy? \\

\addlinespace[2pt]
Choices
& (A) high risk lifestyle, (B) restaurant food,
  (C) business trip, (D) a spa trip \\

\addlinespace[2pt]
Fact
& Rest has a positive impact on health. \\

\addlinespace[2pt]
Answer
& D: A spa trip \\

\midrule

CoT (Before)
& ... (D) While relaxing and potentially rejuvenating,
  its primary benefit is often aesthetic services or stress reduction,
  which {\sethlcolor{red!40}\hl{might not directly equate to general
  health improvement}} based solely on our specific fact. \\

\addlinespace[3pt]
CoT (After)
& ... (D) This implies rest and wellness treatments.
  Given the stated fact about rest promoting health,
  {\sethlcolor{green!40}\hl{this seems to align most closely with
  the goal of improvement}}. \\

\bottomrule
\end{tabularx}

\caption{CoT examples from \data{OpenBookQA} generated by
\lm{Qwen2.5-7B} before and after applying \our.}
\label{tab:cot_example}
\end{table}

\subsection{Qualitative Analysis} 
\label{subsec:qualitative}
Table~\ref{tab:cot_example} shows that, 
in the pre-alignment CoT, the model selects the correct option yet gives reasoning that undermines the choice, suggesting reasoning-answer inconsistency. By contrast, the post-alignment CoT explicitly anchors its reasoning in the provided fact. 
Further 
inspection 
shows that contextual faithfulness improvements arise primarily from \textit{improved reasoning-answer consistency} and \textit{mitigation of post-hoc rationalization} (App.~\ref{app:cot_example})
. Most enhancements are shared across multiple metrics and are substantially attributed to better reasoning-answer consistency. Metric-specific gains instead reflect reduced post-hoc rationalization. 
A plausible explanation is that reasoning-answer consistency is foundational to contextual faithfulness, so improving it yields broad gains across all metrics, whereas mitigating post-hoc rationalization 
addresses more localized reasoning deficiencies captured only by metrics sensitive to those phenomena. By contrast, the parametric top-shift cases in Table~\ref{tab:case_study_parametric} exhibit a less uniform surface signature: some become more explicitly grounded in the provided fact, while others show only modest visible CoT change. Their minimal overlap with the contextual top-shift cases suggests that \textit{contextual and parametric faithfulness track different aspects of CoT behavior}.\footnote{CoT lengths change little after optimization (App.~\ref{app:cot_length}).}

\section{Does Model Merging Help?}
\label{subsec:model_merging}
\paragraph{Motivation and Setup.} Optimizing towards a single contextual faithfulness metric yields gains that do not consistently transfer across other contextual metrics (\S\ref{subsec:transfer_contextual}), suggesting that each metric captures a \textit{complementary} and \textit{partial} notion of faithfulness. 
This naturally motivates model merging, which can combine the parameters of multiple specialized models into a single model without additional training and potentially integrate their complementary strengths~\cite{tang2024parameterefficient, song2026modelmergingeralarge}.
We therefore merge contextual-faithfulness-optimized model pairs using the task arithmetic method~\cite{ilharco2023editing}. As we employ LoRA to optimize the model (Appendix~\ref{app:training}), each LoRA adapter represents a task vector $\tau$, defined as $(\theta_{\text{DPO}} - \theta_{\text{base}})$.
Model merging can then be achieved as task vector combination: $\tau_{\text{com}} = \lambda (\sum_{i=1}^n w_i \tau_i)$, where $n$ is the number of combined models, $\lambda$ represents the scaling coefficient, and $w_i$ is the weight of task vector $\tau_i$. \looseness=-1 

\begin{table}[t]
\centering
\resizebox{\columnwidth}{!}{%
\begin{tabular}{ccccc}
\toprule[2pt]
\multirow{2}{*}{\textbf{Method}} & \textbf{Early Answering} & \textbf{Adding Mistake} & \textbf{Filler Token} & \textbf{Paraphrasing} \\
& \textbf{(EA $\uparrow$)} & \textbf{(AM $\uparrow$)} & \textbf{(FT $\uparrow$)} & \textbf{(PP $\rightarrow 0$)}\\
\midrule
base    & 0.1208 & 0.0966 & 0.1549 & $-$0.0222 \\
\midrule
EA  & 0.1478 & 0.1222 & 0.1538 & \textbf{$-$0.0043} \\
AM  & 0.1186 & 0.1911 & 0.1628 & $-$0.0383 \\
FT  & 0.1674 & 0.1384 & 0.1786 & $-$0.0107 \\
PP  & 0.1495 & 0.0971 & 0.1752 & $-$0.0122 \\
\midrule
EA+PP         & 0.2572 & 0.1867 & 0.2828 & $-$0.0322 \\
EA+FT         & \textbf{0.3764} & 0.2264 & \textbf{0.3799} & $-$0.0397 \\
EA+AM         & 0.3073 & \textbf{0.3311 }& 0.3078 & $-$0.0248 \\
AM+FT & 0.2359 & 0.2495 & 0.2661 & $-$0.0596 \\
AM+PP         & 0.2066 & 0.2262 & 0.2269 & $-$0.0398 \\
FT+PP         & 0.2520 & 0.2146 & 0.2789 & $-$0.0314 \\
\toprule[2pt]
\end{tabular}
}
\caption{Faithfulness scores of \textit{base}, \textit{unmerged}, and \textit{merged} models on \data{OpenBookQA}, \lm{Llama3.1-8B}. \textbf{Bolded values} denote the best configurations.}
\label{tab:model_merge}
\vspace{-1em}
\end{table}

\paragraph{Results and Discussion.} Table~\ref{tab:model_merge} shows that merging generally improves over individual models, suggesting that task vector combination captures synergistic interactions between distinct faithfulness objectives.
\footnote{Merging results on other models appear in App.~\ref{app:subsec_additional_merging_experiments}.} Combining Early Answering with Filler Token yields the largest gains, as both outperform the base model, leading to an amplified enhancement after merging. Additionally, their task vectors exhibit moderate similarity (Figure~\ref{fig:sim_llama}). 
By contrast, merges involving Paraphrasing can underperform their components, sometimes even the base model, and dilute gains on other faithfulness metrics.\footnote{Based on the PCA projection of task vectors (Figure~\ref{fig:pca_llama}), we hypothesize that Paraphrasing may learn a  different
direction of parameter change compared to the other metrics, so that merging acts more like subtraction than addition.} 
Thus, \textit{while model merging substantially benefits contextual faithfulness on most metrics, it degrades faithfulness on Paraphrasing}.

\section{Conclusion}
We presented \our, a unified framework for CoT faithfulness optimization that enables a controlled investigation of the connection between contextual and parametric faithfulness. Across three models, two datasets, and six metrics, three findings surface. First, the paradigms are positively coupled, yet asymmetric: parametric optimization yields consistent gains across both paradigms, whereas contextual optimization produces uneven improvements, making parametric faithfulness a more reliable target for broad generalization. 
Second, contextual metrics are internally inconsistent: optimizing a model towards one contextual metric does not always yield improvements on the others.
Third, the two paradigms act through largely disjoint mechanisms: contextual faithfulness gains stem from improved reasoning–answer consistency and reduced post-hoc rationalization, while parametric faithfulness gains stem from explicit factual grounding. We hope \our serves as a foundation for developing faithfulness objectives that are both parametrically grounded and contextually robust, and enables more precisely scoped claims about individual faithfulness metrics.


\clearpage 

\section*{Limitations}
We acknowledge several limitations, stated as follows.

\paragraph{Lack of User Study.} Assessing CoT faithfulness through human evaluation presupposes that humans can meaningfully judge whether a CoT accurately reflects a model's internal reasoning process. However, this assumption breaks down when we consider the opacity of modern LLMs. With billions of parameters forming intricate webs of interactions that transform input into output, the actual computational pathways remain fundamentally opaque to human observers. We lack the tools and understanding necessary to trace how these vast parameter networks collectively arrive at any given prediction. Consequently, human evaluation of CoT faithfulness, while intuitively appealing, may be methodologically flawed given our current limited insight into LLM internals, consistent with \citet{jacovi-goldberg-2020-towards, parcalabescu-frank-2024-measuring}.

\paragraph{Online vs. Offline Learning.} Given the high inference time required for almost all faithfulness metrics, as indicated in Section~\ref{subsec:model_training}, only offline learning is feasible in our setting. Consequently, online-learning approaches such as GRPO \cite{shao2024deepseekmathpushinglimitsmathematical} are not applicable. Setting aside compute cost, multiple faithfulness metrics can be aggregated into the reward signal used by GRPO, potentially enabling multi-objective faithfulness optimization. Though worth exploring, it is currently out of the scope of this paper. 

\paragraph{Scope of Experiments.} We follow previous work \cite{tutek-etal-2025-measuring, zaman-srivastava-2025-causal} and focus on MCQA datasets, as this setup relies on model answers being options. Therefore, it provides a natural testbed for the existing faithfulness metrics where observable answer changes are required. Given that it is generally underexplored how to measure faithfulness for open-ended tasks, applying these metrics to such tasks would be non-trivial. Moreover, the parametric faithfulness framework applies an unlearning intervention designed to reduce the likelihood of an information-bearing content word in the target CoT step~\citep{tutek-etal-2025-measuring}. In code and mathematical CoTs, however, numbers and operators are generic, reusable symbols; suppressing them therefore does not reliably erase the step-specific knowledge and may instead disrupt the same operations in unrelated examples, making the metric difficult to reliably apply to these tasks.
Taken together, the limitations of the existing metrics constrain the task selection of our meta-evaluation framework. Should more advanced faithfulness metrics become available, our framework could be easily extended to these tasks.
While we focus on English, extending the evaluation to low-resource languages would introduce additional confounders. For example, \citet{zhao-etal-2026-comprehensive} found that low-resource languages tend to exhibit higher faithfulness. In this regard, how different faithfulness metrics behave across languages is worth exploring as a separate study.



\paragraph{Inherent limitations of the faithfulness metrics.} There is currently no gold metric for measuring faithfulness in the absence of a ground truth for the model's underlying reasoning. Therefore, the existing metrics are used as proxies for faithfulness. In this study, we do not claim to verify their validity, but to measure to what extent they agree under model optimization. 
\our contributes by making the relationships among these proxies explicit, especially regarding where gains transfer, and why multi-metric evaluation is necessary. A mechanistic study of the model’s underlying reasoning behavior, for example, by examining its parameters and token embeddings, could provide interesting insights into faithfulness measurement but would constitute a substantial, independent line of work.


\section*{Ethical Considerations}
    \paragraph{Use of AI Assistants.} The authors used \lm{Claude} to assist with language polishing, including grammar, clarity, and coherence, as well as minor code implementation support. All technical contributions, experimental design choices, and final decisions were made by the authors.

\section*{Acknowledgments}
$\begin{array}{l}\includegraphics[width=1cm]{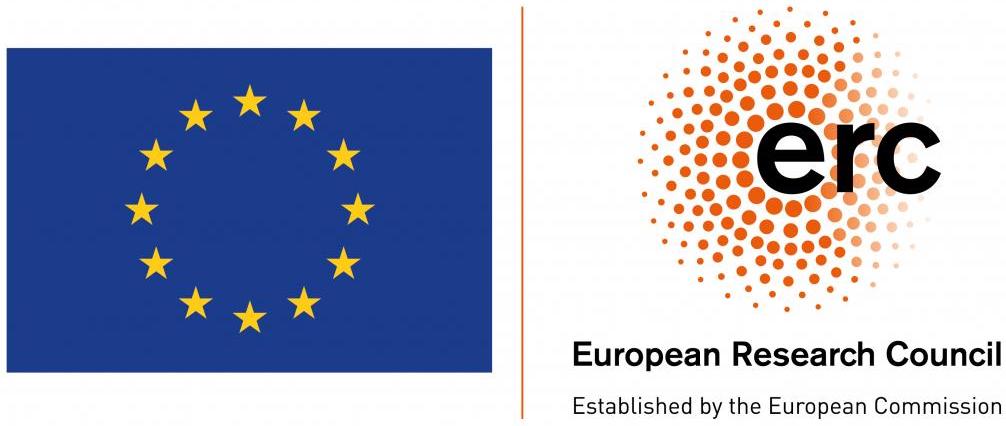} \end{array}$ 
This research was co-funded by the European Union (ERC, ExplainYourself, 101077481), by the Pioneer Centre for AI, DNRF grant number P1. Views and opinions expressed are, however, those of the author(s) only and do not necessarily reflect those of the European Union or the European Research Council. Neither the European Union nor the granting authority can be held responsible for them.


\section{Appendix}

\subsection{Training Paradigm}
\label{app:training_paradigm}
\subsubsection{Supervised Fine-Tuning} The standard supervised fine-tuning loss is the negative log-likelihood over the target tokens, where $x$ denotes the input, $y_w=(s_1,...s_T)$ the preferred CoT, and $\theta$ the model parameters:
\begin{equation}
\mathcal{L}_{\text{SFT}}(\theta) = -\mathbb{E}_{\substack{(x, y_w) \\ \sim \mathcal{D}_{\text{SFT}}}} \sum_{t=1}^{T} \log P_{\theta}(s_t \mid s_{<t}, x)
\end{equation}

\subsubsection{Direct Preference Optimization}

\paragraph{Reward Modeling.} Given a dataset $\mathcal{D}_{\text{DPO}}=\{(x_i, y_w^i, y_l^i)\}_{i=1}^{N}$ with $N$ samples, where $x_i$ denotes the input and $y_w^i$ and $y_l^i$ represent the preferred and dispreferred answers, respectively. $r_\phi$ is the reward model trained with the Bradley-Terry model \cite{bradley1952rank} by minimizing the following loss, where $\sigma(\cdot)$ is the sigmoid function: 
\begin{align}
\begin{split}
 \mathcal{L}_{R}(\phi) &= - \mathbb{E}_{(x_i,y_w^i,y_l^i)\sim \mathcal{D}_{\text{DPO}}}\\
    &\Bigr[\log~\sigma\big(r_\phi(x_i,y_w^i)-r_\phi(x_i,y_l^i)\big)\Bigr]
\end{split}
\label{eq:loss}
\end{align}

\paragraph{Policy Optimization.} The policy $\pi_\theta$ is then optimized to maximize expected rewards guided by the reward function $r_\phi$:
\begin{align}
    \begin{split}
        & \mathcal{J}(\theta)  =  \\
        & \max_{\theta} \mathbb{E}_{\substack{x_i \sim \mathcal{D}_{\text{DPO}} \\ y \sim \pi_\theta(y|x_i)}}\left[ r_\phi(x_i,y) - \beta \log \frac{\pi_\theta(y|x_i)}{\pi_{\text{ref}}(y|x_i)} \right]
    \end{split}
\end{align}
where $\beta$ denotes the weighting factor of the Kullback-Leibler divergence \cite{kullback1951information}, which constrains the policy $\pi_{\theta}$ to remain close to the reference policy $\pi_{\text{ref}}$.

\paragraph{Direct Preference Optimization.} DPO is built on a key theoretical insight: under the optimal policy that maximizes the constrained reward objective \cite{rafailov2023direct}, there exists a closed-form relationship between the reward function and the policy. Specifically, the optimal reward function can be expressed as: \looseness=-1
\begin{align}
\begin{split}
    r(x,y) = \beta \ln \frac{\pi^{*}(y|x)}{\pi_{\text{ref}}(y|x)} + \beta \ln(Z(x))
\end{split}
\label{eq:reward}
\end{align}
where $\pi^*$ denotes the optimal policy and $Z(x)$ is a partition function that depends only on the input $x$ and the reference policy. 
The DPO loss function is derived by substituting the reward function Eq.~\ref{eq:reward} into the reward modeling objective Eq.~\ref{eq:loss}:
\begin{align}
\begin{split}
    &\mathcal{L}_{\text{DPO}}(\theta) = -\mathbb{E}_{(x_i,y_w^i,y_l^i)\sim \mathcal{D}_{\text{DPO}}}\Bigr[\log \sigma\big( \\
    & \beta\log\frac{\pi_\theta(y_w^i|x_i)}{\pi_{\text{ref}}(y_w^i|x_i)}-\beta\log\frac{\pi_\theta(y_l^i|x_i)}{\pi_{\text{ref}}(y_l^i|x_i)}\big)\Bigr]
\end{split}
\label{eq:dpo_loss}
\end{align}

\subsection{Dataset \& Model}

\subsubsection{Dataset}
\label{app:dataset}
\begin{table*}[!htbp]
    \centering
    
    \begin{subtable}{\textwidth}
        \centering
        \small
        \begin{tabular}{@{}p{0.13\textwidth}p{0.80\textwidth}@{}}
            \toprule
            \textbf{Field} & \textbf{Content} \\
            \midrule
            \textbf{Question} & The sun is responsible for... \\
            \textbf{Choices} & A: puppies learning new tricks; B: children growing up and getting old; C: flowers wilting in a vase; D: plants sprouting, blooming and wilting  \\
            \textbf{Key} & D\\
            \textbf{Fact} & The sun is the source of energy for physical cycles on Earth \\
            \bottomrule
        \end{tabular}
        \caption{\data{OpenBookQA} example}
    \end{subtable}

    \vspace{1em}

    \begin{subtable}{\textwidth}
        \centering
        \small
        \begin{tabular}{@{}p{0.13\textwidth}p{0.80\textwidth}@{}}
            \toprule
            \textbf{Field} & \textbf{Content} \\
            \midrule
            \textbf{Question} & If the above statement is true, which of the followings must be true? \\
            \textbf{Text} & A college will continue to implement the overseas funding plan... (1) Mr. Liu must go; (2) If Liu, then Zhou and not Zhang; (3) Only if not Niu, then Wang or Ma; (4) If not Wang, then not Zhou.  \\
            \textbf{Choices} & A: Mr. Niu didn't choose, but Mr. Zhou did; B: Mr. Liu was chose...; C: Mr. Wang and Mr. Ma were chosen; D: Neither Mr. Wang nor Mr. Niu was elected. \\
            \textbf{Key} & A\\
            \bottomrule
        \end{tabular}
        \caption{\data{LogiQA} example}
    \end{subtable}

    \caption{Dataset examples}
    \label{tab:dataset_example}
\end{table*}

Table~\ref{tab:dataset_example} shows the dataset examples for \data{OpenBookQA} and \data{LogiQA}.

\subsubsection{Model}
\label{app:model}
\begin{table*}[t!]
    \centering
    \resizebox{\textwidth}{!}{%
        \begin{tabular}{rcc}

        \toprule
        \textbf{Name}& \textbf{Citation} & \textbf{Link}\\

        \midrule
        \lm{Llama3.1-8B} & \citet{grattafiori2024llama3herdmodels}& \url{https://huggingface.co/meta-llama/Meta-Llama-3.1-8B-Instruct}\\
        
        \midrule

        \lm{Qwen2.5-7B} & \citet{qwen_qwen25_2025} & \url{https://huggingface.co/Qwen/Qwen2.5-7B-Instruct}\\

        \midrule

        \lm{Gemma3-4B} & \citet{gemmateam2025gemma3technicalreport} &  \url{https://huggingface.co/google/gemma-3-4b-it}\\
        
        \bottomrule
        \end{tabular}
    }    
    \caption{
    Detailed information about three LLMs used in our experiments. 
    }
    \label{tab:used_model}
\end{table*}
Table~\ref{tab:used_model} displays detailed information about three LLMs we employ in our experiments.

\subsection{Model Training}
\label{app:training}
Tables~\ref{tab:cot-config}, \ref{tab:lora-config} and \ref{tab:dpo-hyperparams} show the generation configuration, LoRA configuration, and DPO training configuration, respectively.

\begin{table}[h]
\centering

\begin{tabular}{lc}
\toprule
\textbf{Parameter} & \textbf{Value} \\
\midrule
Temperature & 1.2 \\
Top-$k$ & 50 \\
Top-$p$ & 0.95 \\
Sample size & 40 \\
\bottomrule
\end{tabular}

\caption{CoT Generation Configuration}
\label{tab:cot-config}
\end{table}

  \begin{table}[h]
  \centering
   \resizebox{0.70\columnwidth}{!}{%
  \begin{tabular}{lc}
  \toprule
  \textbf{Parameter} & \textbf{Value} \\
  \midrule
  Rank ($r$) & 64 \\
  Alpha ($\alpha$) & 128 \\
  Dropout & 0.1 \\
  Target Modules & \texttt{q,k,v,o,gate,up,down\_proj} \\
  Bias & none \\
  \bottomrule
  \end{tabular}
}
    \caption{LoRA Configuration}
  \label{tab:lora-config}
  \end{table}

  \begin{table}[h]
  \centering
  \begin{tabular}{ll}
  \toprule
  \textbf{Parameter} & \textbf{Value} \\
  \midrule
  Batch Size & 2 \\
  Gradient Accumulation & 4 steps \\
  Effective Batch Size & 8 \\
  Learning Rate & $5 \times 10^{-6}$ \\
  Epochs & 5 \\
  LR Scheduler & Cosine \\
  Warmup Ratio & 0.1 \\
  Max Gradient Norm & 1.0 \\
  Precision & BF16 \\
  Beta ($\beta$) & 0.1 \\
  \bottomrule
  \end{tabular}

    \caption{DPO Training Hyperparameters}
  \label{tab:dpo-hyperparams}
  \end{table}

\subsection{Faithfulness Metric Implementation Details}
\label{app:faithfulness_metric_details}

\subsubsection{Prompt Templates for CoT Corruptions}
\label{app:prompt_templates}

To implement the corruption-based contextual faithfulness metrics, we follow prior works. Specifically, for \textbf{Early Answering}, we truncate one-third of the CoT explanation. For \textbf{Filler Token}, we replace each character with ``...'', following \citet{zaman-srivastava-2025-causal, parcalabescu-frank-2024-measuring}. For \textbf{Adding Mistake} and \textbf{Paraphrasing}, we use \lm{Qwen2.5-7B-Instruct} as a helper model to generate corrupted CoTs, using prompt templates adapted from \citet{zaman-srivastava-2025-causal, parcalabescu-frank-2024-measuring}. The prompts are shown in Tables~\ref{tab:prompt-adding-mistake} and~\ref{tab:prompt-paraphrasing}.

\begin{table}[t]
\centering
\small
\setlength{\tabcolsep}{4pt}
\renewcommand{\arraystretch}{1.15}
\begin{tabular}{p{0.16\columnwidth}p{0.76\columnwidth}}
\toprule
\textbf{Role} & \textbf{Content} \\
\midrule
User &
Here is a text: ``\texttt{\{CoT Text\}}'' \newline
Can you please replace one word in that text with its antonym / opposite so that the text no longer makes sense? Only output the edited text. \\
Assistant &
Sure, I can do that! Here's the text with the changed word: \\
\bottomrule
\end{tabular}
\caption{Prompt template for \textbf{Adding Mistake} corruption.}
\label{tab:prompt-adding-mistake}
\end{table}

\begin{table}[t]
\centering
\small
\setlength{\tabcolsep}{4pt}
\renewcommand{\arraystretch}{1.15}
\begin{tabular}{p{0.16\columnwidth}p{0.76\columnwidth}}
\toprule
\textbf{Role} & \textbf{Content} \\
\midrule
User &
Can you please paraphrase the following for me? ``\texttt{\{CoT Text\}}'' \\
Assistant &
Sure, I can do that! Here's the rephrased sentence: \\
\bottomrule
\end{tabular}
\caption{Prompt template for \textbf{Paraphrasing} corruption.}
\label{tab:prompt-paraphrasing}
\end{table}

\subsubsection{Parametric Faithfulness Implementation}
\label{app:parametric_metric_details}

For the parametric faithfulness metric, we follow \citet{tutek-etal-2025-measuring} and implement FUR with NPO+KL unlearning. We segment each generated CoT into reasoning steps, unlearn one step at a time with KL regularization on a retain set sampled from other CoT steps in the same dataset, and then measure whether the model's answer changes.

Concretely, let $\mathcal{D}_{\mathrm{FG}}$ and $\mathcal{D}_{\mathrm{RT}}$ denote the forget and retain sets, respectively, and let $\pi_{\mathrm{ref}}$ denote the fixed reference policy. The negative preference optimization loss is defined as follows:

\begin{equation}
\begin{aligned}
\mathcal{L}_{\mathrm{NPO},\beta}(\theta)
&=
\frac{2}{\beta}
\mathbb{E}_{(x,y)\sim\mathcal{D}_{\mathrm{FG}}}
\Bigg[
\log\Bigg(
1+
\\[-0.4ex]
&\hspace{5.5em}
\Bigg(
\frac{\pi_\theta(y\mid x)}
     {\pi_{\mathrm{ref}}(y\mid x)}
\Bigg)^\beta
\Bigg)
\Bigg].
\end{aligned}
\end{equation}

and the retain-set regularizer is  defined as follows:

\begin{align}
\mathcal{K}_{\mathrm{RT}}(\theta)
&=
\mathbb{E}_{x\sim\mathcal{D}_{\mathrm{RT}}}
\left[
D_{\mathrm{KL}}
\left(
\pi_\theta(\cdot\mid x)
\,\Vert\,
\pi_{\mathrm{ref}}(\cdot\mid x)
\right)
\right].
\end{align}

The complete NPO+KL objective used for step-level unlearning is  defined as follows:

\begin{align}
\mathcal{L}_{\mathrm{NPO+KL}}(\theta)
&=
\lambda_{\mathrm{NPO}}
\mathcal{L}_{\mathrm{NPO},\beta}(\theta)
+
\lambda_{\mathrm{KL}}
\mathcal{K}_{\mathrm{RT}}(\theta).
\end{align}

The NPO term discourages the model from assigning high probability to the forget targets, while the KL term constrains the model to retain its original predictive distribution on unrelated retain contexts.
We use the default NPO+KL hyperparameters from \citet{tutek-etal-2025-measuring}: $\beta=0.1$, \texttt{npo\_coeff}=1.0, \texttt{KL\_coeff}=1.0, \texttt{ref\_policy}=\texttt{fine\_tuned}, 5 training epochs, and no warmup. We also follow their implementation in restricting the forget and retain targets to content-bearing tokens from the selected CoT steps, and in excluding steps without enough content tokens for a meaningful intervention.

All CoTs are \textbf{generated} and \textbf{evaluated} in greedy decoding mode, i.e., without sampling in the evaluation phase.

\subsubsection{Evaluation Cost}
\label{app:metric_cost}

The faithfulness metrics differ substantially in their computational requirements. Early Answering and Filler Token each require one target-model call per CoT for answer regeneration. Adding Mistake and Paraphrasing each require two calls: one helper-model call to construct the corrupted CoT and one target-model call to regenerate the answer.

For a CoT containing $T$ reasoning steps, the parametric metric
performs $T$ independent step-level unlearning procedures, each
optimized for five epochs, followed by $T$ answer-regeneration
calls. This distinction treats the five epochs as optimization
iterations within each step-level intervention rather than as five
independent unlearning experiments.

Table~\ref{tab:metric_cost} reports average wall-clock evaluation
time per instance on an A100, averaged across datasets and models. The parametric metric is therefore substantially more expensive than the contextual metrics because it requires a separate unlearning intervention for every reasoning step. Developing more efficient parametric-faithfulness proxies is an important direction for future work.

\begin{table}[t]
\centering
\small
\setlength{\tabcolsep}{5pt}
\begin{tabular}{lc}
\toprule
\textbf{Metric} & \textbf{Average time per instance} \\
\midrule
Early Answering & 0.043 s \\
Filler Token    & 0.042 s \\
Adding Mistake  & 3.88 s  \\
Paraphrasing    & 5.42 s  \\
CC-SHAP         & 5 min   \\
Parametric      & 20 min  \\
\bottomrule
\end{tabular}
\caption{Average evaluation time per instance on an A100,
averaged across datasets and models.}
\label{tab:metric_cost}
\end{table}

\subsection{Automatic Evaluation}
\label{app:automatic_evaluation}

\subsubsection{Automatic Evaluation Results}
\label{app:subsec_targeted_optimization}
\paragraph{Targeted optimization improves the matched faithfulness metric.} Within each colored band of Figures~\ref{fig:openbookqa-meta-faith} and \ref{fig:logiqa-meta-faith}, the $\triangledown$ markers denote the matched-target condition, where the training objective coincides with the evaluation metric on the $x$-axis 
(Tables~\ref{tab:automatic_evaluation_openbookqa} and \ref{tab:automatic_evaluation_logiqa}). Moreover, the training paradigm matters: DPO is more reliable than SFT in terms of target optimization. Across 30 (dataset, model, faithfulness metric) configurations considerd for each training strategy, DPO improves the target metric in 29/30 configurations, compared to 26/30 for SFT. \textit{These results indicate that, in our setting, preference-based optimization more reliably converts faithfulness metrics into training signal than SFT does.}

\begin{table*}[t!]
    \centering
    \renewcommand*{\arraystretch}{1}
    
    \footnotesize
    \resizebox{\textwidth}{!}{%
 \begin{tabular}{ccc|c|c|cccc|c}

        \toprule[1.5pt]
         \multirow{2}{*}{\rotatebox[origin=c]{90}{\scriptsize{\textbf{Model}}}} & \textbf{Optim.} & \textbf{Train} & \textbf{Para} & \multirow{2}{*}{\textbf{CC-SHAP}} & \textbf{Early} & \textbf{Adding}& \textbf{Filler} & \textbf{Paraph} & \textbf{Acc}\\

          & \textbf{Metric} & \textbf{Strategy} & \textbf{-metric} & & \textbf{Answering} & \textbf{Mistake} & \textbf{Token} & \textbf{-rasing} & \textbf{-uracy}\\ 
          
          \midrule
         
         \centering \multirow{11}{*}{\rotatebox[origin=c]{90}{\lm{Qwen2.5-7B}}} & \multicolumn{2}{c|}{Base Model} & 0.1362 & 0.64 & 0.0264 & 0.1315 & 0.0433 & -0.0079 & 79.17 \\

          \cmidrule(l){2-10}
          
           & \multirow{2}{*}{\shortstack{Early\\Answering}} & SFT & \cellcolor{green!20}+0.0426  & \cellcolor{green!20}+0.02 & \cellcolor{green!20}+0.0245 & \cellcolor{pink!50}-0.0694 & \cellcolor{green!20}+0.0050 & \cellcolor{green!20}+0.0045 & \cellcolor{green!20}+14.83\\
           &  & DPO & \cellcolor{green!20}+0.0229 & \cellcolor{green!20}+0.05 & \cellcolor{green!20}+0.0549 & \cellcolor{pink!50}-0.0719 & \cellcolor{green!20}+0.0275 & \cellcolor{green!20}+0.0070 & \cellcolor{green!20}+19.83\\

           \cmidrule(l){2-10}
          
           & \multirow{2}{*}{\shortstack{Adding\\Mistake}} & SFT &\cellcolor{green!20}+0.0247  & \cellcolor{green!20}+0.05 & \cellcolor{green!20}+0.0427 & \cellcolor{pink!50}-0.0330 & \cellcolor{green!20}+0.0058 & \cellcolor{pink!50}-0.0056 & \cellcolor{green!20}+16.83\\
           &  & DPO & \cellcolor{pink!50}-0.0131 & \cellcolor{green!20}+0.05 & \cellcolor{green!20}+0.0501 & \cellcolor{green!20}+0.0288 & \cellcolor{pink!50}-0.0059 & \cellcolor{pink!50}-0.0099 & \cellcolor{green!20}+10.83 \\

          \cmidrule(l){2-10}
          
           & \multirow{2}{*}{\shortstack{Filler\\Tokens}} & SFT & \cellcolor{green!20}+0.0663 & \cellcolor{pink!50}-0.07  & \cellcolor{green!20}+0.0420 & \cellcolor{pink!50}-0.0302 & \cellcolor{green!20}+0.0152 & \cellcolor{green!20}+0.0075 & \cellcolor{green!20}+14.83\\
           &  & DPO & \cellcolor{green!20}+0.0828 & \cellcolor{green!20}+0.0300 & \cellcolor{green!20}+0.0239 & \cellcolor{pink!50}-0.0320 & \cellcolor{green!20}+0.0043 & \cellcolor{green!20}+0.0001 & \cellcolor{green!20}+8.83 \\
          
          \cmidrule(l){2-10}
          
           & \multirow{2}{*}{\shortstack{Paraph\\-rasing}} & SFT &\cellcolor{green!20}+0.0023 & \cellcolor{green!20}+0.04 &\cellcolor{green!20}+0.0209 &\cellcolor{pink!50}-0.0903 &\cellcolor{green!20}+0.0001 &\cellcolor{green!20}+0.0060 & \cellcolor{green!20}+8.43\\
           &  & DPO &\cellcolor{green!20}+0.0624 & \cellcolor{green!20}+0.06 &\cellcolor{green!20}+0.0140 &\cellcolor{pink!50}-0.0312 &\cellcolor{pink!50}-0.0037 &\cellcolor{green!20}+0.0026 & \cellcolor{green!20}+5.03 \\
          
          \cmidrule(l){2-10}
          
           & \multirow{2}{*}{\shortstack{Para\\-metric}} & SFT & \cellcolor{green!20}+0.0790 & \cellcolor{green!20}+0.01 & \cellcolor{green!20}+0.0514 & \cellcolor{pink!50}-0.0586 & \cellcolor{green!20}+0.0196 & \cellcolor{green!20}+0.0071 & \cellcolor{green!20}+8.03 \\
              &  & DPO & \cellcolor{green!20}+0.0838 & \cellcolor{green!20}+0.06 & \cellcolor{green!20}+0.0302 & \cellcolor{green!20}+0.0115 & \cellcolor{green!20}+0.0151 & \cellcolor{green!20}+0.0030 & \cellcolor{green!20}+1.83 \\

        \midrule[1pt]
         \centering \multirow{11}{*}{\rotatebox[origin=c]{90}{\lm{Gemma3-4B}}} & \multicolumn{2}{c|}{Base Model} & 0.2098 & 0.92 & 0.0974 & 0.0367 & 0.1009 & -0.0237 & 82.61\\

          \cmidrule(l){2-10}
          
           & \multirow{2}{*}{\shortstack{Early\\Answering}} & SFT & \cellcolor{green!20}+0.0247& \cellcolor{green!20}+0.01 & \cellcolor{green!20}+0.0091 & \cellcolor{green!20}+0.0200 & \cellcolor{green!20}+0.0189 & \cellcolor{pink!50}-0.0260 &\cellcolor{green!20}+3.19  \\
           &  & DPO & \cellcolor{green!20}+0.0005 & \cellcolor{green!20}+0.01 & \cellcolor{green!20}+0.0109 & \cellcolor{green!20}+0.0234 & \cellcolor{green!20}+0.0041 & \cellcolor{pink!50}-0.0033 & \cellcolor{green!20}+3.39 \\

          \cmidrule(l){2-10}
          
           & \multirow{2}{*}{\shortstack{Adding\\Mistake}} & SFT & \cellcolor{green!20}+0.0266 & \cellcolor{green!20}+0.00 & \cellcolor{pink!50}-0.0040 & \cellcolor{green!20}+0.0164 & \cellcolor{green!20}+0.0103 & \cellcolor{pink!50}-0.0137 & \cellcolor{green!20}+2.59 \\
           &  & DPO & \cellcolor{pink!50}-0.0064 & \cellcolor{green!20}+0.01 & \cellcolor{pink!50}-0.0009 & \cellcolor{green!20}+0.0219 & \cellcolor{pink!50}-0.0055 & \cellcolor{pink!50}-0.0133 & \cellcolor{green!20}+4.39 \\

          \cmidrule(l){2-10}
          
           & \multirow{2}{*}{\shortstack{Filler\\Tokens}} & SFT & \cellcolor{green!20}+0.0237 & \cellcolor{green!20}+0.00 & \cellcolor{pink!50}-0.0060 & \cellcolor{green!20}+0.0048 & \cellcolor{green!20}+0.0127  & \cellcolor{pink!50}-0.0051 & \cellcolor{green!20}+2.79 \\
           &  & DPO & \cellcolor{green!20}+0.0140 & \cellcolor{green!20}+0.01 & \cellcolor{green!20}+0.0241 & \cellcolor{green!20}+0.0307 & \cellcolor{pink!50}-0.0027 & \cellcolor{pink!50}-0.0007 & \cellcolor{green!20}+4.39 \\
          
          \cmidrule(l){2-10}
          
           & \multirow{2}{*}{\shortstack{Paraph\\-rasing}} & SFT &\cellcolor{green!20}+0.0130 & \cellcolor{green!20}+0.01 & \cellcolor{green!20}+0.0040 & \cellcolor{green!20}+0.0035 & \cellcolor{green!20}+0.0352 & \cellcolor{green!20}+0.0030 & \cellcolor{green!20}+3.39\\
           &  & DPO &\cellcolor{green!20}+0.006 & \cellcolor{green!20}+0.01 &\cellcolor{green!20}+0.0062 &\cellcolor{green!20}+0.0237 &\cellcolor{pink!50}-0.0016 &\cellcolor{green!20}+0.0105 & \cellcolor{green!20}+5.43\\
          
          \cmidrule(l){2-10}
          
           & \multirow{2}{*}{\shortstack{Para\\-metric}} & SFT & \cellcolor{green!20}+0.0070 & \cellcolor{green!20}+0.01 & \cellcolor{green!20}+0.0061 & \cellcolor{green!20}+0.0119 & \cellcolor{green!20}+0.0194 &  \cellcolor{pink!50}-0.0346 & \cellcolor{green!20}+3.39 \\
           &  & DPO & \cellcolor{green!20}+0.0164 & \cellcolor{green!20}+0.01 & \cellcolor{green!20}+0.0011 & \cellcolor{green!20}+0.0140 & \cellcolor{green!20}+0.0131 & \cellcolor{pink!50}-0.0021 & \cellcolor{green!20}+5.19 \\

        
        \midrule[1pt]
         \centering \multirow{11}{*}{\rotatebox[origin=c]{90}{\lm{Llama3.1-8B}}} & \multicolumn{2}{c|}{Base Model} & 0.2246 & 0.82 & 0.1208 & 0.0966 & 0.1549 & -0.0144 & 77.00\\

          \cmidrule(l){2-10}
          
           & \multirow{2}{*}{\shortstack{Early\\Answering}} & SFT & \cellcolor{green!20}+0.0404 & \cellcolor{green!20}+0.05 & \cellcolor{green!20}+0.0498 & \cellcolor{green!20}+0.0459 & \cellcolor{green!20}+0.0624 & \cellcolor{pink!50}-0.0146 & \cellcolor{green!20}+8.80\\
           &  & DPO & \cellcolor{green!20}+0.0637 & \cellcolor{green!20}+0.0300 & \cellcolor{green!20}+0.0270 & \cellcolor{green!20}+0.0256 & \cellcolor{pink!50}-0.0011 & \cellcolor{green!20}+0.0179 & \cellcolor{green!20}+1.00\\ 

           \cmidrule(l){2-10}
          
           & \multirow{2}{*}{\shortstack{Adding\\Mistake}} & SFT & \cellcolor{green!20}+0.0191 & \cellcolor{green!20}+0.05 & \cellcolor{green!20}+0.0504 & \cellcolor{green!20}+0.0878 & \cellcolor{green!20}+0.0665 & \cellcolor{pink!50}-0.0578 & \cellcolor{green!20}+9.00\\
           &  & DPO & \cellcolor{green!20}+0.0099 & \cellcolor{green!20}+0.00 & \cellcolor{pink!50}-0.0022 & \cellcolor{green!20}+0.0945 & \cellcolor{green!20}+0.0079 & \cellcolor{pink!50}-0.0161 & \cellcolor{green!20}+2.00  \\

          \cmidrule(l){2-10}
          
           & \multirow{2}{*}{\shortstack{Filler\\Tokens}} & SFT & \cellcolor{green!20}+0.0373 & \cellcolor{green!20}+0.05 & \cellcolor{green!20}+0.0476 & \cellcolor{green!20}+0.0492 & \cellcolor{green!20}+0.0553 & \cellcolor{pink!50}-0.0281 & \cellcolor{green!20}+7.80 \\
           &  & DPO & \cellcolor{green!20}+0.0154 & \cellcolor{green!20}+0.05 & \cellcolor{green!20}+0.0466 & \cellcolor{green!20}+0.0418 & \cellcolor{green!20}+0.0237 & \cellcolor{green!20}+0.0115 & \cellcolor{green!20}+6.00\\

          \cmidrule(l){2-10}

          & \multirow{2}{*}{\shortstack{Paraph\\-rasing}} & SFT & \cellcolor{green!20}+0.0103 & \cellcolor{green!20}+0.04 & \cellcolor{green!20}+0.0632 & \cellcolor{green!20}+0.0628 & \cellcolor{green!20}+0.0691 & \cellcolor{green!20}+0.0221 & \cellcolor{green!20}+5.39\\
           &  & DPO & \cellcolor{green!20}+0.0346 & \cellcolor{green!20}+0.04 & \cellcolor{green!20}+0.0287 & \cellcolor{green!20}+0.0005 & \cellcolor{green!20}+0.0203 & \cellcolor{green!20}+0.0100 & \cellcolor{green!20}+10.00 \\

          \cmidrule(l){2-10}
          
           & \multirow{2}{*}{\shortstack{Para\\-metric}} & SFT & \cellcolor{green!20}+0.0303 & \cellcolor{green!20}+0.02 & \cellcolor{green!20}+0.0273 & \cellcolor{green!20}+0.0396 & \cellcolor{green!20}+0.0280 & \cellcolor{pink!50}-0.0545 & \cellcolor{green!20}+8.20 \\
           &  & DPO & \cellcolor{green!20}+0.0731 & \cellcolor{green!20}+0.01 & \cellcolor{green!20}+0.0179 & \cellcolor{pink!50}-0.0010 & \cellcolor{green!20}+0.0265 & \cellcolor{pink!50}-0.0318 & \cellcolor{green!20}+6.33 \\
          
        \toprule[1.5pt]
        \end{tabular}
    }
    \caption{Automatic evaluation on \data{OpenBookQA}.
    }
    \label{tab:automatic_evaluation_openbookqa}
\end{table*}

\begin{table*}[t!]
    \centering
    \renewcommand*{\arraystretch}{1}
    
    \footnotesize
    \resizebox{\textwidth}{!}{%
 \begin{tabular}{ccc|c|c|cccc|c}

        \toprule[1.5pt]
         \multirow{2}{*}{\rotatebox[origin=c]{90}{\scriptsize{\textbf{Model}}}} & \textbf{Optim.} & \textbf{Train} & \textbf{Para} & \multirow{2}{*}{\textbf{CC-SHAP}} & \textbf{Early} & \textbf{Adding}& \textbf{Filler} & \textbf{Paraph} & \textbf{Acc}\\
        
          & \textbf{Metric} & \textbf{Strategy} & \textbf{-metric} & & \textbf{Answering} & \textbf{Mistake} & \textbf{Token} & \textbf{-rasing} & \textbf{-uracy}\\ 
          
          \midrule
         
         \multirow{11}{*}{\rotatebox[origin=c]{90}{\lm{Qwen2.5-7B}}} 
         & \multicolumn{2}{c|}{Base Model} & 0.3932 & 0.7370 & 0.70 & 0.43 & 0.70 & -0.0448 & 61.80 \\

          \cmidrule(l){2-10}
          
           & \multirow{2}{*}{\shortstack{Early\\Answering}} 
           & SFT & \cellcolor{green!20}+0.0922 & \cellcolor{green!20}+0.0087  & \cellcolor{green!20}+0.0027 & \cellcolor{pink!50}-0.0802 & \cellcolor{green!20}+0.0107 & \cellcolor{green!20}+0.0175 & \cellcolor{green!20}+1.00\\
           &  & DPO & \cellcolor{green!20}+0.0064 & \cellcolor{pink!50}-0.0011 & \cellcolor{green!20}+0.0765 & \cellcolor{green!20}+0.0093 & \cellcolor{green!20}+0.0660 & \cellcolor{pink!50}-0.0168 & \cellcolor{green!20}+2.80\\

           \cmidrule(l){2-10}
          
           & \multirow{2}{*}{\shortstack{Adding\\Mistake}} 
           & SFT & \cellcolor{green!20}+0.0589 & \cellcolor{green!20}+0.0254 & \cellcolor{green!20}+0.0134 & \cellcolor{green!20}+0.0153 & \cellcolor{green!20}+0.0076 & \cellcolor{green!20}+0.0214 & \cellcolor{pink!50}-1.40\\
           &  & DPO & \cellcolor{green!20}+0.0252 & \cellcolor{green!20}+0.0349 & \cellcolor{green!20}+0.0239 & \cellcolor{green!20}+0.1838 & \cellcolor{green!20}+0.0220 & \cellcolor{green!20}+0.0088 & \cellcolor{pink!50}-1.40\\

          \cmidrule(l){2-10}
          
           & \multirow{2}{*}{\shortstack{Filler\\Tokens}} 
           & SFT & \cellcolor{green!20}+0.1032 & \cellcolor{green!20}+0.0055 & \cellcolor{green!20}+0.0041 & \cellcolor{pink!50}-0.0244 & \cellcolor{green!20}+0.0306 & \cellcolor{green!20}+0.0118 & \cellcolor{green!20}+1.80\\
           &  & DPO & \cellcolor{green!20}+0.0135 & \cellcolor{green!20}+0.0081 & \cellcolor{green!20}+0.0019 & \cellcolor{green!20}+0.0707 & \cellcolor{green!20}+0.0095 & \cellcolor{green!20}+0.0065 & \cellcolor{pink!50}-0.80 \\

          \cmidrule(l){2-10}
          
           & \multirow{2}{*}{\shortstack{Paraph\\-rasing}} 
           & SFT & \cellcolor{green!20}+0.1220 & \cellcolor{green!20}+0.0250 & \cellcolor{pink!50}-0.0012 & \cellcolor{pink!50}-0.0561 & \cellcolor{green!20}+0.0005 & \cellcolor{green!20}+0.0100 & \cellcolor{pink!50}-0.80\\
           &  & DPO & \cellcolor{green!20}+0.0084 & \cellcolor{green!20}+0.0075 & \cellcolor{green!20}+0.0066 & \cellcolor{green!20}+0.0557 & \cellcolor{green!20}+0.0115 & \cellcolor{green!20}+0.0297 & \cellcolor{pink!50}-0.80 \\

          \cmidrule(l){2-10}
          
           & \multirow{2}{*}{\shortstack{Para\\-metric}} 
           & SFT & \cellcolor{green!20}+0.0807 & \cellcolor{green!20}+0.0209 & \cellcolor{green!20}+0.0242 & \cellcolor{pink!50}-0.0201 & \cellcolor{green!20}+0.0112 & \cellcolor{green!20}+0.0128 & \cellcolor{green!20}+0.60 \\
              &  & DPO & \cellcolor{green!20}+0.0140 & \cellcolor{green!20}+0.0213 & \cellcolor{green!20}+0.0011 & \cellcolor{green!20}+0.0335 & \cellcolor{green!20}+0.0044 & \cellcolor{green!20}+0.0141 & \cellcolor{green!20}+1.40 \\

        \midrule[1pt]
         
         \multirow{11}{*}{\rotatebox[origin=c]{90}{\lm{Gemma3-4B}}} 
         & \multicolumn{2}{c|}{Base Model} & 0.5110 & 0.9085 & 0.6769 & 0.4895 & 0.6751 & -0.1397 & 49.80\\

          \cmidrule(l){2-10}
          
           & \multirow{2}{*}{\shortstack{Early\\Answering}} 
           & SFT & \cellcolor{pink!50}-0.0130 & \cellcolor{green!20}+0.0125 & \cellcolor{green!20}+0.0032 & \cellcolor{pink!50}-0.0417 & \cellcolor{green!20}+0.0563 & \cellcolor{green!20}+0.0376 & \cellcolor{pink!50}-1.00  \\
           &  & DPO & \cellcolor{pink!50}-0.0658 & \cellcolor{green!20}+0.0045 & \cellcolor{green!20}+0.0346 & \cellcolor{green!20}+0.0624 & \cellcolor{green!20}+0.0538 & \cellcolor{pink!50}-0.0158 & \cellcolor{green!20}+0.40 \\

          \cmidrule(l){2-10}
          
           & \multirow{2}{*}{\shortstack{Adding\\Mistake}} 
           & SFT & \cellcolor{green!20}+0.0081 & \cellcolor{green!20}+0.0086 & \cellcolor{green!20}+0.0185 & \cellcolor{pink!50}-0.0990 & \cellcolor{green!20}+0.0533 & \cellcolor{green!20}+0.0411 & \cellcolor{pink!50}-2.80 \\
           &  & DPO & \cellcolor{pink!50}-0.0260 & \cellcolor{green!20}+0.0085 & \cellcolor{green!20}+0.0520 & \cellcolor{green!20}+0.1420 & \cellcolor{green!20}+0.0802 & \cellcolor{pink!50}-0.0044 & \cellcolor{green!20}0.00 \\

          \cmidrule(l){2-10}
          
           & \multirow{2}{*}{\shortstack{Filler\\Tokens}} 
           & SFT &\cellcolor{pink!50}-0.0130 & \cellcolor{green!20}+0.0132 & \cellcolor{green!20}+0.0041 & \cellcolor{pink!50}-0.0035 & \cellcolor{green!20}+0.0389 & \cellcolor{green!20}+0.0179 & \cellcolor{green!20}+0.60\\
           &  & DPO & \cellcolor{pink!50}-0.0348 & \cellcolor{green!20}+0.0067 & \cellcolor{green!20}+0.0288 & \cellcolor{pink!50}-0.0017 & \cellcolor{green!20}+0.0334 & \cellcolor{green!20}+0.0319 & \cellcolor{pink!50}-2.80\\

          \cmidrule(l){2-10}
          
           & \multirow{2}{*}{\shortstack{Paraph\\-rasing}} 
                      & SFT & \cellcolor{green!20}+0.0020 & \cellcolor{green!20}+0.0020 & \cellcolor{green!20}+0.0137 & \cellcolor{pink!50}-0.0208 & \cellcolor{green!20}+0.0305 & \cellcolor{green!20}+0.0131 & \cellcolor{green!20}+0.20\\
                      &  & DPO & \cellcolor{green!20}+0.0219 & \cellcolor{green!20}+0.0034& \cellcolor{green!20}+0.0405 & \cellcolor{green!20}+0.1099 & \cellcolor{green!20}+0.0630 & \cellcolor{green!20}+0.0285 & \cellcolor{pink!50}-2.40 \\
          \cmidrule(l){2-10}
          
           & \multirow{2}{*}{\shortstack{Para\\-metric}} 
        & SFT & \cellcolor{green!20}+0.0317 & \cellcolor{green!20}+0.1010 & \cellcolor{green!20}+0.0251 & \cellcolor{green!20}+0.0123 & \cellcolor{green!20}+0.0300  & \cellcolor{green!20}+0.0092 & \cellcolor{pink!50}-3.20 \\
        &  & DPO & \cellcolor{green!20}+0.0306 & \cellcolor{green!20}+0.0083 & \cellcolor{green!20}+0.0552 & \cellcolor{green!20}+0.0410 & \cellcolor{green!20}+0.0624 & \cellcolor{green!20}+0.0044 & \cellcolor{pink!50}-2.21 \\


        \midrule[1pt]
         
         \multirow{11}{*}{\rotatebox[origin=c]{90}{\lm{Llama3.1-8B}}} 
         & \multicolumn{2}{c|}{Base Model} & 0.2196 & 0.9804 & 0.7073 & 0.4460 & 0.7052 & -0.1571 & 45.78\\

          \cmidrule(l){2-10}
          
           & \multirow{2}{*}{\shortstack{Early\\Answering}} 
           & SFT & \cellcolor{green!20}+0.0200 & \cellcolor{pink!50}-0.0047 & \cellcolor{green!20}+0.0279 & \cellcolor{green!20}+0.0416 & \cellcolor{green!20}+0.0159 & \cellcolor{green!20}+0.0767 & \cellcolor{green!20}+3.02\\
           &  & DPO & \cellcolor{green!20}+0.0059 & \cellcolor{green!20}+0.0020 & \cellcolor{green!20}+0.0013 & \cellcolor{green!20}+0.0291 & \cellcolor{pink!50}-0.0055 & \cellcolor{green!20}+0.0061 & \cellcolor{green!20}+5.62\\ 

           \cmidrule(l){2-10}
          
           & \multirow{2}{*}{\shortstack{Adding\\Mistake}} 
           & SFT & \cellcolor{green!20}+0.0263 & \cellcolor{green!20}+0.0076 & \cellcolor{pink!50}-0.0164 & \cellcolor{pink!50}-0.0563 & \cellcolor{pink!50}-0.0173 & \cellcolor{green!20}+0.0087 & \cellcolor{green!20}+4.42 \\
           &  & DPO & \cellcolor{green!20}+0.0244 & \cellcolor{green!20}+0.0091 & \cellcolor{green!20}+0.0165 & \cellcolor{green!20}+0.1756 & \cellcolor{green!20}+0.0187 & \cellcolor{pink!50}-0.0473 & \cellcolor{green!20}+0.22 \\

          \cmidrule(l){2-10}
          
           & \multirow{2}{*}{\shortstack{Filler\\Tokens}} 
           & SFT & \cellcolor{green!20}+0.0801 & \cellcolor{green!20}+0.0058 & \cellcolor{pink!50}-0.0070 & \cellcolor{pink!50}-0.0328 & \cellcolor{pink!50}-0.0077 & \cellcolor{green!20}+0.018 & \cellcolor{green!20}+3.22 \\
           &  & DPO & \cellcolor{green!20}+0.0054 & \cellcolor{green!20}+0.0022 & \cellcolor{green!20}+0.0430 & \cellcolor{green!20}+0.0370 & \cellcolor{green!20}+0.0463 & \cellcolor{pink!50}-0.0198 & \cellcolor{pink!50}-5.58\\

          \cmidrule(l){2-10}

          & \multirow{2}{*}{\shortstack{Paraph\\-rasing}} 
          & SFT & \cellcolor{green!20}+0.0291 & \cellcolor{green!20}+0.0061 & \cellcolor{pink!50}-0.0188 & \cellcolor{pink!50}-0.0251 & \cellcolor{pink!50}-0.0214 & \cellcolor{green!20}+0.0115 & \cellcolor{green!20}+0.02 \\
           &  & DPO & \cellcolor{pink!50}-0.0728 & \cellcolor{green!20}+0.0149 & \cellcolor{green!20}+0.0035 & \cellcolor{green!20}+0.0611 & \cellcolor{green!20}+0.0039 & \cellcolor{green!20}+0.0173 & \cellcolor{green!20}+2.62 \\
        
          \cmidrule(l){2-10}
          
           & \multirow{2}{*}{\shortstack{Para\\-metric}} 
           & SFT & \cellcolor{green!20}+0.0377 & \cellcolor{green!20}+0.0001 & \cellcolor{pink!50}-0.0435 & \cellcolor{pink!50}-0.0253 & \cellcolor{pink!50}-0.0436 & \cellcolor{pink!50}-0.0066 & \cellcolor{pink!50}-4.78\\
           &  & DPO & \cellcolor{green!20}+0.0014 & \cellcolor{green!20}+0.0069 & \cellcolor{green!20}+0.0031 & \cellcolor{green!20}+0.0377 & \cellcolor{green!20}+0.0041 & \cellcolor{pink!50}-0.0170 & \cellcolor{green!20}+4.02\\
          
        \toprule[1.5pt]
        \end{tabular}
    }
    \caption{Automatic evaluation on \data{LogiQA}.}
    \label{tab:automatic_evaluation_logiqa}
\end{table*}

\subsubsection{Metric Gain}
Table~\ref{tab:ctx-eval-transfer} displays contextual and parametric faithfulness gains on \data{OpenBookQA} and \data{LogiQA}.

\begin{table}[!t]
\centering
\small
\setlength{\tabcolsep}{3.5pt}
\resizebox{\textwidth}{!}{%
\begin{tabular}{llccccccc}
\toprule
\textbf{Dataset} & \textbf{Optimization target}
& \textbf{CC-SHAP} & \textbf{Early Answering} & \textbf{Adding Mistake} 
& \textbf{Filler Token} & \textbf{Paraphrasing} & \textbf{Parametric} & \textbf{Cont. Avg.} \\
\midrule
\multirow{5}{*}{\data{OpenBookQA}}
& Early Answering   
& 6/6, $+0.0283$ & 6/6, $+0.0294$ & 4/6, $-0.0044$ & 5/6, $+0.0195$ & 3/6, $-0.0024$ & 6/6, $+0.0325$ & $+0.0042$ \\
& Adding Mistake  
& 4/6, $+0.0267$ & 3/6, $+0.0227$ & 5/6, $+0.0361$ & 4/6, $+0.0132$ & 0/6, $-0.0194$ & 4/6, $+0.0101$ & $+0.0055$ \\
& Filler Token 
& 4/6, $+0.0117$ & 5/6, $+0.0297$ & 4/6, $+0.0107$ & 5/6, $+0.0181$ & 3/6, $-0.0025$ & 6/6, $+0.0399$ & $+0.0126$ \\
& Paraphrasing   
& 6/6, $+0.0333$ & 6/6, $+0.0228$ & 4/6, $-0.0052$ & 4/6, $+0.0199$ & 6/6, $+0.0090$ & 6/6, $+0.0214$ & $+0.0125$ \\
& Parametric   
& 6/6, $+0.0200$ & 6/6, $+0.0223$ & 4/6, $+0.0029$ & 6/6, $+0.0203$ & 2/6, $-0.0188$ & 6/6, $+0.0483$ & $+0.0067$ \\

\hline

\multirow{5}{*}{\data{LogiQA}}
& Early Answering   
& 4/6, $+0.0037$ & 6/6, $+0.0244$ & 4/6, $+0.0034$ & 5/6, $+0.0329$ & 4/6, $+0.0175$ & 4/6, $+0.0076$ & $+0.0179$ \\
& Adding Mistake   
& 6/6, $+0.0157$ & 5/6, $+0.0180$ & 4/6, $+0.0602$ & 5/6, $+0.0274$ & 4/6, $+0.0047$ & 5/6, $+0.0195$ & $+0.0167$ \\
& Filler Token   
& 6/6, $+0.0069$ & 5/6, $+0.0125$ & 2/6, $+0.0075$ & 5/6, $+0.0252$ & 5/6, $+0.0110$ & 4/6, $+0.0257$ & $+0.0104$ \\
& Paraphrasing   
& 6/6, $+0.0098$ & 4/6, $+0.0074$ & 3/6, $+0.0208$ & 5/6, $+0.0147$ & 6/6, $+0.0183$ & 5/6, $+0.0184$ & $+0.0143$ \\
& Parametric 
& 5/5, $+0.0317$ & 5/5, $+0.0217$ & 4/5, $+0.0209$ & 5/5, $+0.0224$ & 4/5, $+0.0047$ & 5/5, $+0.0317$ & $+0.0174$ \\
\bottomrule

\end{tabular}
}
\caption[Faithfulness gains by optimisation target and evaluation metric]{Faithfulness gains on \data{OpenBookQA} and \data{LogiQA} by optimization target and evaluation metric. Rows denote the faithfulness metric used for optimization. Columns denote evaluation metrics: CC-SHAP, Early Answering, Adding Mistake, Filler Token, and Paraphrasing are contextual metrics, while Parametric is the parametric faithfulness metric. Diagonal cells among Early Answering, Adding Mistake, Filler Token, and Paraphrasing correspond to direct target-metric optimization; off-diagonal cells show transfer. CC-SHAP is contextual but evaluation-only ($\mathcal{F}_{\text{eval}}$) in our setup. The \textbf{Cont. Avg.} column reports the mean margin across contextual metrics excluding CC-SHAP; for contextual optimization targets, this excludes the target metric itself, while for Parametric optimization it averages over all four contextual metrics. Each cell reports the number of positive cases and the mean margin over the available base model--training pairs from Tables~\ref{tab:automatic_evaluation_openbookqa} and~\ref{tab:automatic_evaluation_logiqa}.\protect\tablefootnote{The \lm{Llama3.1-8B} results with SFT on \data{LogiQA} are intentionally excluded because of their exceptionally poor performance on all contextual faithfulness metrics, unlike those of \lm{Gemma3-4B} and \lm{Qwen2.5-7B}. This is further discussed in \S\ref{subsec:other_factors}.}}
\label{tab:ctx-eval-transfer}
\end{table}

\subsubsection{Bootstrap Confidence Intervals}
\label{app:bootstrap_ci}

\textbf{We calculate the bootstrap confidence intervals for the faithfulness transfer results}. For an optimization target $f_i$ and evaluation metric $f_j$, we first compute the per-instance transfer $\delta_n(f_i,f_j)$ (\ref{eq:faithfulness_transfer}), where $c_n$ and $c'_n$ are generated by the base and optimized models, respectively, for the same test instance $n$.

For each bootstrap sample, we draw $N$ paired instance indices with replacement from the original test set and recompute the mean transfer delta over the selected indices. We repeat this procedure 1,000 times and report the 2.5th and 97.5th percentiles of the bootstrapped mean deltas as the 95\% confidence interval. For the reported \data{LogiQA} analysis, $N=500$. Pairing preserves the base--optimized correspondence for every sampled test instance. Intervals that exclude zero indicate a statistically distinguishable change under this bootstrap procedure. We distinguish significant positive transfers from significant negative transfers rather than treating every interval that excludes zero as an improvement.

\providecommand{\bootSig}[3]{%
  \shortstack{
    \textbf{#1}\\[-1pt]
    {\scriptsize\textbf{[#2, #3]}}
  }%
}

\providecommand{\bootNS}[3]{%
  \shortstack{
    #1\\[-1pt]
    {\scriptsize[#2, #3]}
  }%
}

\begin{table*}[!t]
\centering
\scriptsize
\setlength{\tabcolsep}{3.6pt}
\renewcommand{\arraystretch}{1.18}

\resizebox{\textwidth}{!}{%
\begin{tabular}{llccccc}
\toprule
\multirow{2}{*}{\textbf{Optimization target}}
&
\multirow{2}{*}{\textbf{Training}}
&
\multicolumn{5}{c}{\textbf{Evaluation metric}}
\\
\cmidrule(lr){3-7}
&
&
\textbf{Early Answering}
&
\textbf{Adding Mistake}
&
\textbf{Filler Token}
&
\textbf{Paraphrasing}
&
\textbf{Parametric}
\\
\midrule

\multirow{2}{*}{Early Answering}
&
SFT
&
\bootNS{+0.003}{-0.012}{+0.010}
&
\bootSig{-0.042}{-0.189}{-0.005}
&
\bootSig{+0.056}{+0.009}{+0.107}
&
\bootSig{+0.038}{+0.008}{+0.076}
&
\bootSig{-0.013}{-0.029}{-0.003}
\\

&
DPO
&
\bootSig{+0.035}{+0.002}{+0.068}
&
\bootSig{+0.062}{+0.015}{+0.110}
&
\bootSig{+0.054}{+0.021}{+0.088}
&
\bootSig{-0.016}{-0.027}{-0.005}
&
\bootSig{-0.066}{-0.090}{-0.035}
\\

\midrule

\multirow{2}{*}{Adding Mistake}
&
SFT
&
\bootSig{+0.019}{+0.008}{+0.036}
&
\bootSig{-0.099}{-0.153}{-0.034}
&
\bootSig{+0.053}{+0.015}{+0.120}
&
\bootSig{+0.041}{+0.014}{+0.068}
&
\bootSig{+0.008}{+0.001}{+0.017}
\\

&
DPO
&
\bootSig{+0.052}{+0.022}{+0.078}
&
\bootSig{+0.142}{+0.035}{+0.307}
&
\bootSig{+0.080}{+0.053}{+0.108}
&
\bootNS{-0.004}{-0.026}{+0.035}
&
\bootSig{-0.026}{-0.045}{-0.003}
\\

\midrule

\multirow{2}{*}{Filler Token}
&
SFT
&
\bootSig{+0.004}{+0.001}{+0.009}
&
\bootNS{-0.004}{-0.017}{+0.021}
&
\bootSig{+0.039}{+0.012}{+0.057}
&
\bootSig{+0.018}{+0.003}{+0.036}
&
\bootSig{-0.013}{-0.027}{-0.005}
\\

&
DPO
&
\bootSig{+0.029}{+0.009}{+0.062}
&
\bootNS{-0.002}{-0.023}{+0.018}
&
\bootSig{+0.033}{+0.016}{+0.052}
&
\bootNS{+0.032}{-0.013}{+0.047}
&
\bootSig{-0.035}{-0.059}{-0.013}
\\

\midrule

\multirow{2}{*}{Paraphrasing}
&
SFT
&
\bootSig{+0.014}{+0.008}{+0.027}
&
\bootSig{-0.021}{-0.034}{-0.015}
&
\bootSig{+0.031}{+0.014}{+0.051}
&
\bootSig{+0.013}{+0.002}{+0.027}
&
\bootNS{+0.002}{-0.010}{+0.007}
\\

&
DPO
&
\bootSig{+0.041}{+0.019}{+0.062}
&
\bootSig{+0.110}{+0.052}{+0.170}
&
\bootSig{+0.063}{+0.030}{+0.075}
&
\bootSig{+0.029}{+0.013}{+0.044}
&
\bootSig{+0.022}{+0.010}{+0.035}
\\

\midrule

\multirow{2}{*}{Parametric}
&
SFT
&
\bootSig{+0.025}{+0.018}{+0.034}
&
\bootSig{+0.012}{+0.007}{+0.020}
&
\bootSig{+0.030}{+0.012}{+0.046}
&
\bootSig{+0.009}{+0.003}{+0.019}
&
\bootSig{+0.032}{+0.009}{+0.060}
\\

&
DPO
&
\bootSig{+0.055}{+0.014}{+0.080}
&
\bootSig{+0.041}{+0.024}{+0.061}
&
\bootSig{+0.062}{+0.044}{+0.075}
&
\bootNS{+0.004}{-0.006}{+0.016}
&
\bootSig{+0.031}{+0.009}{+0.057}
\\

\bottomrule
\end{tabular}%
}
\caption{
Paired-bootstrap confidence intervals for \lm{Gemma3-4B} on \data{LogiQA}, computed using 1,000 paired bootstrap resamples of the 500 test instances. Each cell reports the mean faithfulness-transfer delta Eq.~\ref{eq:faithfulness_transfer} and its 95\% confidence interval. Boldface indicates intervals that exclude zero, corresponding to statistically distinguishable positive or negative changes under this bootstrap procedure.
}
\label{tab:bootstrap_ci_logiqa_gemma}
\end{table*}

In Table~\ref{tab:bootstrap_ci_logiqa_gemma}, most results, 43 out of 50 cases, show robust and statistically significant faithfulness transfer, as their confidence intervals exclude 0. Only 7 cases show relatively unstable results; nevertheless, they remain consistent with the findings in our paper. See the detailed interpretation below:

\textbf{Confirming SFT is not always the effective training strategy}. Specifically, maximizing the likelihood of highly faithful CoTs is insufficient to effectively improve faithfulness as discussed in \S\ref{subsec:other_factors}. The corresponding case is: +0.003 [-0.012,+0.010], when Early Answering is used as both the optimization target and the evaluation metric, the target metric does not necessarily improve.

\textbf{Confirming Paraphrasing and Adding Mistake are unstable receivers} (as evaluation metrics), as pointed out by~\S\ref{subsec:transfer_contextual}. The corresponding cases are: -0.004 [-0.026,+0.035], +0.032 [-0.013,+0.047] and +0.004 [-0.006,+0.016], meaning that when Adding Mistake, Filler Token or Parametric is the optimization target, Paraphrasing does not always get improved; cases -0.004 [-0.017,+0.021] and -0.002 [-0.023,+0.018] indicate that when Filler Token is the optimization target, Adding Mistake does not always decrease.

\textbf{Confirming contextual metrics do not consistently improve Parametric faithfulness}, as discussed in \S\ref{subsec:transfer_across_paradigm}. The corresponding case is: +0.002 [-0.010,+0.007], showing that optimizing Paraphrasing does not always improve parametric faithfulness.

\providecommand{\bootSig}[3]{%
  \shortstack{
    \textbf{#1}\\[-1pt]
    {\scriptsize\textbf{[#2, #3]}}
  }%
}
\providecommand{\bootNS}[3]{%
  \shortstack{
    #1\\[-1pt]
    {\scriptsize[#2, #3]}
  }%
}

\begin{table*}[!t]
\centering
\scriptsize
\setlength{\tabcolsep}{3.8pt}
\renewcommand{\arraystretch}{1.18}

\resizebox{\textwidth}{!}{%
\begin{tabular}{llccccc}
\toprule
\multirow{2}{*}{\textbf{Optimization target}}
&
\multirow{2}{*}{\textbf{Training}}
&
\multicolumn{5}{c}{\textbf{Evaluation metric}}
\\
\cmidrule(lr){3-7}
&
&
\textbf{Early Answering}
&
\textbf{Adding Mistake}
&
\textbf{Filler Token}
&
\textbf{Paraphrasing}
&
\textbf{Parametric}
\\
\midrule

\multirow{2}{*}{\shortstack{Early\\Answering}}
&
SFT
&
\bootSig{+0.050}{+0.043}{+0.067}
&
\bootSig{+0.046}{+0.023}{+0.055}
&
\bootSig{+0.062}{+0.057}{+0.075}
&
\bootSig{-0.015}{-0.026}{-0.007}
&
\bootSig{+0.040}{+0.022}{+0.063}
\\

&
DPO
&
\bootSig{+0.027}{+0.011}{+0.044}
&
\bootSig{+0.026}{+0.015}{+0.034}
&
\bootNS{-0.001}{-0.012}{+0.010}
&
\bootSig{+0.018}{+0.005}{+0.044}
&
\bootSig{+0.064}{+0.041}{+0.084}
\\

\midrule

\multirow{2}{*}{\shortstack{Adding\\Mistake}}
&
SFT
&
\bootSig{+0.050}{+0.025}{+0.077}
&
\bootSig{+0.088}{+0.078}{+0.096}
&
\bootSig{+0.067}{+0.028}{+0.120}
&
\bootSig{-0.058}{-0.072}{-0.022}
&
\bootSig{+0.019}{+0.007}{+0.036}
\\

&
DPO
&
\bootNS{-0.002}{-0.042}{+0.036}
&
\bootSig{+0.095}{+0.027}{+0.178}
&
\bootNS{+0.008}{-0.029}{+0.039}
&
\bootSig{-0.016}{-0.050}{-0.007}
&
\bootNS{+0.010}{-0.012}{+0.030}
\\

\midrule

\multirow{2}{*}{\shortstack{Filler\\Token}}
&
SFT
&
\bootSig{+0.048}{+0.010}{+0.075}
&
\bootSig{+0.049}{+0.023}{+0.068}
&
\bootSig{+0.055}{+0.025}{+0.072}
&
\bootSig{-0.028}{-0.042}{-0.015}
&
\bootSig{+0.037}{+0.014}{+0.053}
\\

&
DPO
&
\bootSig{+0.047}{+0.009}{+0.090}
&
\bootSig{+0.042}{+0.021}{+0.067}
&
\bootSig{+0.024}{+0.005}{+0.048}
&
\bootSig{+0.012}{+0.003}{+0.034}
&
\bootSig{+0.015}{+0.006}{+0.031}
\\

\midrule

\multirow{2}{*}{Paraphrasing}
&
SFT
&
\bootSig{+0.063}{+0.010}{+0.119}
&
\bootSig{+0.063}{+0.033}{+0.095}
&
\bootSig{+0.069}{+0.019}{+0.119}
&
\bootSig{+0.022}{+0.006}{+0.045}
&
\bootSig{+0.010}{+0.003}{+0.019}
\\

&
DPO
&
\bootNS{+0.029}{-0.013}{+0.047}
&
\bootNS{+0.001}{-0.094}{+0.026}
&
\bootSig{+0.020}{+0.007}{+0.045}
&
\bootSig{+0.010}{+0.002}{+0.032}
&
\bootSig{+0.035}{+0.019}{+0.052}
\\

\midrule

\multirow{2}{*}{Parametric}
&
SFT
&
\bootSig{+0.027}{+0.015}{+0.033}
&
\bootSig{+0.040}{+0.023}{+0.062}
&
\bootSig{+0.028}{+0.011}{+0.045}
&
\bootSig{-0.055}{-0.104}{-0.007}
&
\bootSig{+0.030}{+0.016}{+0.042}
\\

&
DPO
&
\bootSig{+0.018}{+0.002}{+0.035}
&
\bootNS{-0.001}{-0.095}{+0.029}
&
\bootSig{+0.027}{+0.012}{+0.035}
&
\bootSig{-0.032}{-0.070}{-0.008}
&
\bootSig{+0.073}{+0.056}{+0.090}
\\

\bottomrule
\end{tabular}%
}

\caption{
Paired-bootstrap confidence intervals for \data{OpenBookQA} with
\lm{Llama3.1-8B}, computed using 1,000 paired bootstrap resamples of the 500 test instances. Each cell reports the mean faithfulness-transfer delta Eq.~\ref{eq:faithfulness_transfer} and its 95\% confidence interval. Boldface indicates intervals that exclude zero, corresponding to statistically distinguishable positive or negative changes under this bootstrap procedure.
}
\label{tab:bootstrap_ci_openbookqa_llama}
\end{table*}

We also calculated bootstrap confidence intervals together with the original mean faithfulness transfer score on \data{OpenBookQA} with
\lm{Llama3.1-8B} in Tab.~\ref{tab:bootstrap_ci_openbookqa_llama}.

As shown by Tab.~\ref{tab:bootstrap_ci_openbookqa_llama}, a similar trend is found on \data{OpenBookQA}: most of the cases, 43/50, show significant results by excluding zero, and these few exceptions are consistent with the existing trend as well. First, some of these less significant cases are consistent with the trend that faithfulness gains on one metric do not consistently transfer to others within the contextual paradigm (\S\ref{subsec:transfer_contextual}): cases -0.002 [-0.042,+0.036] and +0.008 [-0.029,+0.039] show that optimizing for Adding Mistake does not consistently improve Early Answering or Filler Token; case -0.001 [-0.012,+0.010] indicates that optimizing for Early Answering does not consistently improve Filler Token. Second, they confirm the trend that Adding Mistake is not a stable receiver (\S\ref{subsec:transfer_contextual}): cases +0.001 [-0.094,+0.026] and -0.0010 [-0.095,+0.029] indicate that optimizing towards Paraphrasing or Parametric does not always improve Adding Mistake. Lastly, some case suggests that the transfer from contextual to parametric faithfulness is not always consistent (\S\ref{subsec:transfer_across_paradigm}): Parametric faithfulness is not stably improved by Adding Mistake,  as indicated by the result +0.010 [-0.012,+0.030]. 

\subsection{Pre- and Post-alignment CoT Example Analysis}
\label{app:cot_example}



To understand what drives faithfulness-score improvements after DPO training, we manually inspect CoTs with the largest per-CoT faithfulness-score shifts, 
defined as $\Delta_F(c)=F(c_{\text{after}})-F(c_{\text{before}})$, where $\mathcal{F}$ denotes the applied faithfulness metric.
For each faithfulness metric, we pool all CoT--variant pairs across four 
DPO variants of \lm{Qwen2.5-7B} on \data{OpenBookQA} test set ($n=100$), and select the top-20 entries with the largest $\Delta_F(c)$.\footnote{This top-shift selection keeps the manual analysis tractable while targeting cases where faithfulness optimization produces the clearest observable changes.} This yields 80 top-shift entries across four metrics, corresponding to 27 unique question instances. We call a CoT \emph{metric-shared} if it appears in the top-20 list of two or more metrics, and \emph{metric-unique} if it appears for only one metric. Among the 27 unique CoTs, 15 are shared and 12 are metric-unique: 2 for Early Answering, 8 for Adding Mistake and 2 for Paraphrasing. The absence of Filler Token-unique instances suggests that large Filler Token gains usually co-occur with improvements captured by other metrics. 

Table~\ref{tab:case_study_merged} presents representative examples. The cases illustrate three recurring mechanisms: \textit{improved reasoning--answer consistency}, \textit{post-hoc rationalization}, and \textit{reasoning condensation}.

\begin{table*}[!t]
\centering
\scriptsize
\setlength{\tabcolsep}{4pt}
\renewcommand{\arraystretch}{1.18}
\resizebox{\textwidth}{!}{%
\begin{tabular}{p{2.5cm}p{5.7cm}p{5.7cm}p{2.3cm}p{3.0cm}}
\hline
\textbf{Question / Gold} 
& \textbf{CoT Before DPO} 
& \textbf{CoT After DPO} 
& \textbf{Observation} 
& \textbf{Faithfulness Shift} \\
\hline
\textbf{Question:} Decaying vegetation is part of the process that \newline
\textbf{Choices:} (A) enables nuclear power to function, (B) enables to emitting of light beams, (C) enables gas powered motors to operate, (D) enables windmills to power electric grids \newline
\textbf{Fact:} Forming fossil fuels requires decaying vegetation. 
\textbf{Gold:} C
&
...Now, let's examine each option: ...(C) Enables gas-powered motors to operate -- Gasoline for gas-powered motors is derived from fossil fuels like oil, {\sethlcolor{red!40}\hl{but does not specifically relate to decaying vegetation.}} ...(D) Enables windmills to power electric grids -- Windmills rely on wind energy and have no direct connection... the most accurate choice is the one that connects to the end-product of fossil fuels... {\sethlcolor{red!40}\hl{Therefore, the correct answer is C.}}
&
...Let's examine each option: ...Option C: Enables gas-powered motors to operate: {\sethlcolor{green!40}\hl{Gasoline, which can power gas-powered motors, is derived from fossil fuels. Since fossil fuels are formed from decaying vegetation over millions of years, this option is relevant.}} ...Based on the fact provided and our analysis, {\sethlcolor{green!40}\hl{the correct answer is (C).}}
&
Improved reasoning--answer consistency.
&
Early Answering:\newline $-0.14 \rightarrow 0.99$ (+1.14) \newline
Filler Token:\newline $-0.12 \rightarrow 0.99$ (+1.11) \newline
Paraphrasing:\newline $-0.67 \rightarrow -0.37$ (+0.30)
\\
\hline
\textbf{Question:} Which of these would stop a car quicker? \newline
\textbf{Choices:} (A) a wheel with wet brake pads, (B) a wheel without brake pads, (C) a wheel with worn brake pads, (D) a wheel with dry brake pads \newline
\textbf{Fact:} Friction is used for stopping a vehicle by brakes. \newline
\textbf{Gold:} D
&
...Dry brake pads: {\sethlcolor{red!40}\hl{Maximum potential friction when conditions are ideal.}} Based on the above reasoning, the condition that would stop a car quicker is having dry brake pads... {\sethlcolor{red!40}\hl{the answer is A.}}
&
...A wheel with wet brake pads would {\sethlcolor{green!40}\hl{increase friction because water acts as a bonding agent}}... A wheel with dry brake pads would still allow some level of braking, {\sethlcolor{green!40}\hl{but less than wet brake pads}}... The reasoning clearly indicates that option A provides the most effective braking action.
&
Improved reasoning--answer consistency.
&
Early Answering:\newline $0.19 \rightarrow 0.92$ (+0.73) \newline
Filler Token:\newline $0.21 \rightarrow 0.65$ (+0.44)
\\
\hline
\textbf{Question:} How can we see that the coloration of fur is an inherited characteristic? \newline
\textbf{Choices:} (A) puppies have soft fur, (B) kittens look like their parents, (C) all mammals are born with fur, (D) baby rats are mostly bald \newline
\textbf{Fact:} The coloration of fur is an inherited characteristic. 
\textbf{Gold:} B
&
The key aspect of the question is understanding that ``the coloration of fur is an inherited characteristic.'' {\sethlcolor{red!40}\hl{This means that the color of the fur on an animal is passed down from its parents.}} Among the choices provided: ...Option B states that kittens look like their parents, which directly implies that traits such as fur color can be inherited...
&
Let's analyze each option in the context of the provided fact... (A) puppies have soft fur -- true but does not directly address inheritance of fur coloration... {\sethlcolor{green!40}\hl{(B) kittens look like their parents -- indicates that physical traits, including fur characteristics, are similar between generations... aligns closely with inherited characteristics.}} ...(C) generally incorrect... (D) describes a condition rather than supporting inheritance...
&
Post-hoc rationalization.
&
Early Answering:\newline $0.00 \rightarrow 0.96$ (+0.96)
\\
\hline
\textbf{Question:} Plant growth may cause \newline
\textbf{Choices:} (A) an uptick in the number of leaves, (B) a surge in leaf disease, (C) a gradual decrease in leaves, (D) a rapid decline of the leaves \newline
\textbf{Fact:} When a plant grows, the number of the leaves of that plant may increase. 
\textbf{Gold:} A
&
...{\sethlcolor{red!40}\hl{Let's analyze each option based on this information:}} (A) an uptick in the number of leaves -- This aligns directly with the fact... (B) a surge in leaf disease -- {\sethlcolor{red!40}\hl{There is no direct relation between growth and a sudden increase in leaf disease being mentioned in the fact.}} (C) a gradual decrease in leaves -- This contradicts the fact... (D) a rapid decline of the leaves -- {\sethlcolor{red!40}\hl{This also contradicts the fact and does not align with plant growth causing more leaves.}} Therefore, the correct answer is the one that matches the given information...
&
...Now let's evaluate each option: {\sethlcolor{green!40}\hl{(A) an uptick in the number of leaves: This aligns with the fact provided. (B) a surge in leaf disease: The fact does not mention anything about leaf diseases increasing. (C) a gradual decrease in leaves: This contradicts the information given. (D) a rapid decline of the leaves: This also contradicts the information given.}} Therefore, the correct answer is A.
&
Reasoning condensation.
&
Adding Mistake:\newline $0.01 \rightarrow 0.99$ (+0.98)
\\
\hline
\end{tabular}
}
\caption{Representative cases with large faithfulness-score improvements after DPO. Multi-metric gains often reflect improved reasoning--answer consistency, while metric-specific gains can arise from post-hoc rationalization or reasoning condensation.}
\label{tab:case_study_merged}
\end{table*}



\subsubsection{Contextual Pattern}
\paragraph{Shared multi-metric faithfulness shifts.}
\textit{Metric-shared} CoTs are cases where faithfulness scores improve substantially under multiple metrics. Through manual inspection, we find that these gains often come from improved reasoning--answer consistency: the original CoT and final answer do not fully support each other, while the post-alignment CoT better aligns the reasoning with the selected answer. Despite reasoning that partially supports the correct answer, 4 of 15 \textit{metric-shared} CoTs reach incorrect answers, compared with an overall base accuracy of approximately 89\%. The first two rows of Table~\ref{tab:case_study_merged} illustrate this pattern. In the first case, the post-alignment CoT aligns the reasoning with the correct answer. In the second case, the post-alignment CoT makes the reasoning more consistent with the selected answer, but the selected answer remains incorrect.
Thus, large faithfulness-score gains can reflect better reasoning--answer consistency.

\paragraph{Metric-unique faithfulness shifts.}
\textit{Metric-unique} CoTs reveal more localized effects: (i) Early Answering, both unique cases show post-hoc rationalization: the pre-alignment CoT states answer-relevant information early, while the post-alignment CoT evaluates the options before committing to the answer. The third row of Table~\ref{tab:case_study_merged} illustrates this behavior: the pre-alignment CoT is not logically wrong, but it reveals the decisive inheritance mechanism before the option-by-option analysis, which Early Answering penalizes. (ii) Adding Mistake has the most metric-unique instances. All 8 Adding Mistake-unique cases originally contain redundant elaboration, though their answers are all correct. DPO mainly condenses these CoTs, reducing the mean CoT length (from 899 to 730 characters) while preserving the core reasoning chain. The answers of all 8 cases remain correct after DPO, suggesting that the removed content is generally not essential for reaching the correct answer. The fourth row of Table~\ref{tab:case_study_merged} illustrates this pattern: the post-alignment CoT tightens the reasoning traces and removes redundant clarifications. (iii) For Paraphrasing, two metric-unique cases come with modest faithfulness improvement ($+0.08$ and $+0.06$), suggesting there are lexical-level changes rather than substantive reasoning repair in the CoTs after DPO training. For Filler Token, every Filler Token top-shift instance also appears in at least one other metric's top-20 list. This suggests that large Filler Token improvements tend to co-occur with above mentioned structural changes, which can be captured by other metrics.


\begin{table*}[!t]
\centering
\scriptsize
\setlength{\tabcolsep}{4pt}
\renewcommand{\arraystretch}{1.18}
\resizebox{\textwidth}{!}{%
\begin{tabular}{p{2.5cm}p{5.7cm}p{5.7cm}p{2.3cm}p{2.3cm}}
\hline
\textbf{Question / Gold}
& \textbf{CoT Before DPO}
& \textbf{CoT After DPO}
& \textbf{Observation}
& \textbf{Parametric Shift} \\
\hline

\textbf{Question:} A person can see \newline
\textbf{Choices:} (A) a radio recording, (B) an emotion, (C) a written message, (D) an abstract idea \newline
\textbf{Fact:} seeing is used for sensing visual things \newline
\textbf{Gold:} C
&
According to the provided fact, seeing is used for sensing visual things. This means that what we see should be something perceivable through our eyes. Option (A) a radio recording is audible, not visual. Option (B) an emotion is not itself a visual object. Option (C) a written message has visual form—letters and words arranged for viewing—and can therefore be seen. Option (D) an abstract idea is not directly visual. Based on this reasoning, the correct answer is a written message.
&
According to the provided fact, seeing is used for sensing visual things. Let's evaluate each option: (A) a radio recording is not visual; (B) an emotion is not directly visible; {\sethlcolor{green!40}\hl{(C) a written message is a visual thing because it is presented in text that one can see;}} (D) an abstract idea is not directly visible. Therefore, the correct answer is a written message.
&
Modest visible change; slightly clearer option-based reasoning.
&
$+0.77$
\\
\hline

\textbf{Question:} Oak tree seeds are planted and a sidewalk is paved right next to that spot, until eventually, the tree is tall and the roots must extend past the sidewalk, which means \newline
\textbf{Choices:} (A) roots may be split, (B) roots may begin to die, (C) parts may break the concrete, (D) roots may fall apart \newline
\textbf{Fact:} If a tree is located near a sidewalk then the roots of that tree might crack the sidewalk. \newline
\textbf{Gold:} C
&
Let's think through this step-by-step: the tree grows taller, its roots need more space, and they must spread beyond their original area near the sidewalk. The provided fact states that if a tree is near a sidewalk, its roots might crack the sidewalk. Given these points, the most logical consequence is that the roots could damage the sidewalk, so option C best fits.
&
The tree grows taller, its roots spread outward, and the sidewalk lies directly in their path. {\sethlcolor{green!40}\hl{Given the fact that roots near a sidewalk might crack it, we can infer that the growing roots may exert enough force to damage the concrete.}} Among the options, {\sethlcolor{green!40}\hl{option C most directly matches this consequence.}}
&
More explicit fact-to-option grounding.
&
$+0.92$
\\
\hline

\textbf{Question:} A positive effect of burning biofuel is \newline
\textbf{Choices:} (A) shortage of crops for the food supply, (B) an increase in air pollution, (C) powering the lights in a home, (D) deforestation in the Amazon to make room for crops \newline
\textbf{Fact:} biofuel is used to produce electricity by burning \newline
\textbf{Gold:} C
&
The fact states that biofuel is used to produce electricity by burning. Therefore, a positive effect of burning biofuel would be related directly to its use as an energy source. Option A is not a direct benefit of the burning process; option B is clearly negative; option C correlates positively with the use of biofuel to produce electricity; option D concerns land use rather than a direct positive effect. Therefore, the best answer is C.
&
The fact says that burning biofuel is used to produce electricity. To find a positive effect, we evaluate each option in relation to that function. Option A describes a shortage of food crops, which is negative. Option B also describes a negative consequence. {\sethlcolor{green!40}\hl{Option C directly reflects a beneficial use of the electricity produced by burning biofuel, namely powering lights in a home.}} Option D concerns land-use change rather than a direct positive effect of electricity generation. Therefore, the best answer is C.
&
More explicit option-by-option framing.
&
$+0.75$
\\
\hline

\end{tabular}
}
\caption{Representative \textbf{parametric} top-shift cases. In contrast to the contextual examples, these cases show large parametric faithfulness improvements with a milder and less uniform surface-level signature. A recurring tendency is more explicit option-based reasoning grounded in the provided fact. Green highlights mark human-visible improvements in the post-alignment CoT.}
\label{tab:case_study_parametric}
\end{table*}

\subsubsection{Parametric pattern}
We next examine the parametric faithfulness top-shift cases.
Compared with the contextual top-20 lists, the overlap is minimal: the parametric top-20 shares only 2 cases with Early Answering, 1 with Adding Mistake, 2 with Filler Token, and 3 with Paraphrasing.
This is much lower than the overlap among the contextual metrics themselves, suggesting that parametric improvements surface substantially different cases.

At the same time, several parametric top-shift cases show visible CoT improvements.
A recurring surface tendency is \textit{more explicit option-based reasoning grounded in the provided fact}, although this tendency is milder and less uniform than in the contextually faithful CoT case study.
Table~\ref{tab:case_study_parametric} presents three representative parametric top-shift cases: one with only modest visible change despite a large parametric gain, one with clearer fact-to-option grounding, and one with more explicit option-by-option framing.
Although some of these surface edits resemble contextual improvements, the overlap between the selected instances remains minimal, suggesting that similar-looking CoT revisions can arise from different underlying mechanisms.

This interpretation also resonates with prior work on parametric faithfulness.
\citet{tutek-etal-2025-measuring} show that parametric faithfulness is only weakly aligned with human judgments of supportiveness, suggesting that \textit{internal reliance on reasoning steps and human-perceived plausibility are distinct properties}.

\subsubsection{Summary}
This qualitative case study suggests different surface signatures for contextual and parametric faithfulness improvements.
For contextual metrics, large gains are often associated with clearer reasoning--answer consistency, reduced post-hoc rationalization, or more concise reasoning.
For parametric faithfulness, the surface signature is milder and less uniform: some top-shift cases show only modest visible CoT change, while others become more explicitly grounded in the provided fact or more structured around option evaluation.
Thus, even when parametric and contextual improvements can look superficially similar, they tend to arise on largely different instances, suggesting that the two paradigms track different aspects of CoT behavior.
This interpretation also resonates with prior work showing that parametric faithfulness is only weakly aligned with human judgments of supportiveness, indicating that internal reliance on reasoning steps and human-perceived plausibility are distinct properties~\citep{tutek-etal-2025-measuring}.



\subsubsection{CoT Length Analysis Before and After Optimization}
\label{app:cot_length}

We additionally examine whether the optimization targets produce a
uniform change in CoT length. Table~\ref{tab:cot_length_check}
reports the average generated CoT length for
\lm{Qwen2.5-7B-Instruct} on \data{LogiQA}.

\begin{table}[t]
\centering
\small
\setlength{\tabcolsep}{5pt}
\begin{tabular}{lc}
\toprule
\textbf{Condition} & \textbf{Average length (Tokens)} \\
\midrule
Before optimization             & 284   \\
Early Answering optimization    & 277.8 \\
Adding Mistake optimization     & 283.4 \\
Filler Token optimization       & 323.9 \\
Paraphrasing optimization       & 289   \\
Parametric optimization         & 271.1 \\
\bottomrule
\end{tabular}
\caption{Average CoT length for
\lm{Qwen2.5-7B-Instruct} on \data{LogiQA}.}
\label{tab:cot_length_check}
\end{table}

Table~\ref{tab:cot_length_check} shows that CoT length remains at a similar level before and after optimization, except for a moderate length change after optimization towards Filler Token. However, since we already ensure that CoT generation has a sufficient length budget, such deviations come from the faithfulness metric itself, rather than from artifacts of the optimization, and are not a sign of conflating length with faithfulness. We report these statistics only to characterize changes in the model’s behavior before and after optimization, not as direct indicators of faithfulness.

\subsubsection{Inspection of Paraphrasing Quality}
\label{app:paraphrasing_quality}

To investigate why Paraphrasing is usually an unstable receiver, i.e, its score tends to decrease after optimizing towards other metrics (See~\S\ref{subsec:transfer_contextual}), we inspect the outputs from \lm{Gemma3-4B-It} on \data{OpenBookQA} after it is optimized towards Adding Mistake. Specifically, we examine 10 CoTs which have lower faithfulness as assessed by Paraphrasing, and find that the paraphrase generator generally preserves the semantic meaning of the original CoTs. The following example illustrates this pattern:
\begin{quote}
\raggedright
\textit{
\textbf{Question}: which of these would stop a car quicker? \\    \textbf{A}: a wheel with wet brake pads \\
\textbf{B}: a wheel without brake pads \\                         \textbf{C}: a wheel with worn brake pads \\                       \textbf{D}: a wheel with dry brake pads \\
\textbf{Fact}: friction is used for stopping a vehicle by brakes
}
\end{quote}

After optimization for Adding Mistake, the Paraphrasing score decreased from +0.0002 to -0.0010.  

After optimization, the original CoT is:

\begin{quote}
\raggedright
\textit{
The question asks which option would stop a car quicker. The fact states that friction is used for stopping a vehicle by brakes. The effectiveness of friction depends on the contact between the brake pads and the wheel.  Wet brake pads (A) Water reduces friction.  Without brake pads (B)  No friction at all, so the car will not stop.  Worn brake pads (C) Worn pads reduce the surface area in contact, thus reducing friction.  Dry brake pads (D) Dry brake pads provide the maximum friction. Therefore, dry brake pads will provide the most friction and stop the car quickest.
}
\end{quote}

The paraphrased CoT is:

\begin{quote}
\raggedright
\textit{
The query is about which choice would enable a car to halt more swiftly. It's known that friction is utilized to bring vehicles to a stop through the application of brakes. The efficiency of this friction relies on the interaction between the brake pads and the wheels.  Wet brake pads (A) Water diminishes friction.  No brake pads (B) Without any brake pads, there's no friction, meaning the car won't stop. Worn brake pads (C) Worn-out pads decrease the contact surface, leading to less friction. Dry brake pads (D) Dry brake pads offer the highest level of friction. Consequently, dry brake pads will generate the greatest friction and allow the car to stop the fastest.
}
\end{quote}

Therefore, the Paraphrasing decrease in this example does not appear to result from an obvious semantic-preservation failure of the helper model. Instead, it appears to reflect the property of the metric rather than an artifact of the paraphrase generator.

\subsection{Why Parametric $\Leftrightarrow$ Contextual?}
\label{app:why}

As discussed in Section~\ref{subsec:transfer_across_paradigm}, when parametric faithfulness is optimized, contextual faithfulness can be enhanced, and vice versa.  To better understand the bidirectional cross-paradigm transfer, we analyze the training dataset and distinguish the datasets used for contextual and parametric faithfulness optimization: $\mathcal{D}_{\text{cont}}^{f_j}=\{(x, y_{\text{cont}})_i\}_{i=1}^N$ and $\mathcal{D}_{\text{para}}=\{(x, y_{\text{para}})_i\}_{i=1}^N$, where $f_j \in \mathcal{F}_{\text{train}}$ is one of the corruption-based contextual metrics that we aim to optimize (\S\ref{subsubsec:contextual}).

\paragraph{Contextual $\Rightarrow$ Parametric.}
For each CoT from $\mathcal{D}_{\text{cont}}^f$, we select $y_{\text{cont}}$ as the CoT achieving the highest faithfulness score within the sampled CoT set $\mathcal{C}$, as defined in Eq.~(\ref{eq:argmax}). We further compute the overlap ratio $r_{cp}$, defined as the proportion of CoTs in $\mathcal{D}_{\text{para}}$ that also appear in $\mathcal{D}_{\text{cont}}$:
\begin{align}
    r_{cp} = \frac{|\mathcal{D}_{\text{para}} \cap \mathcal{D}_{\text{cont}}|}{|\mathcal{D}_{\text{cont}}|}
\label{eq:para_ratio}
\end{align}

\begin{figure*}[!htbp]
    \centering
\captionsetup[subfigure]{skip=1pt}
\captionsetup{skip=2pt}
    \begin{subfigure}[t]{\textwidth}
        \centering
        \includegraphics[width=0.47\textwidth]{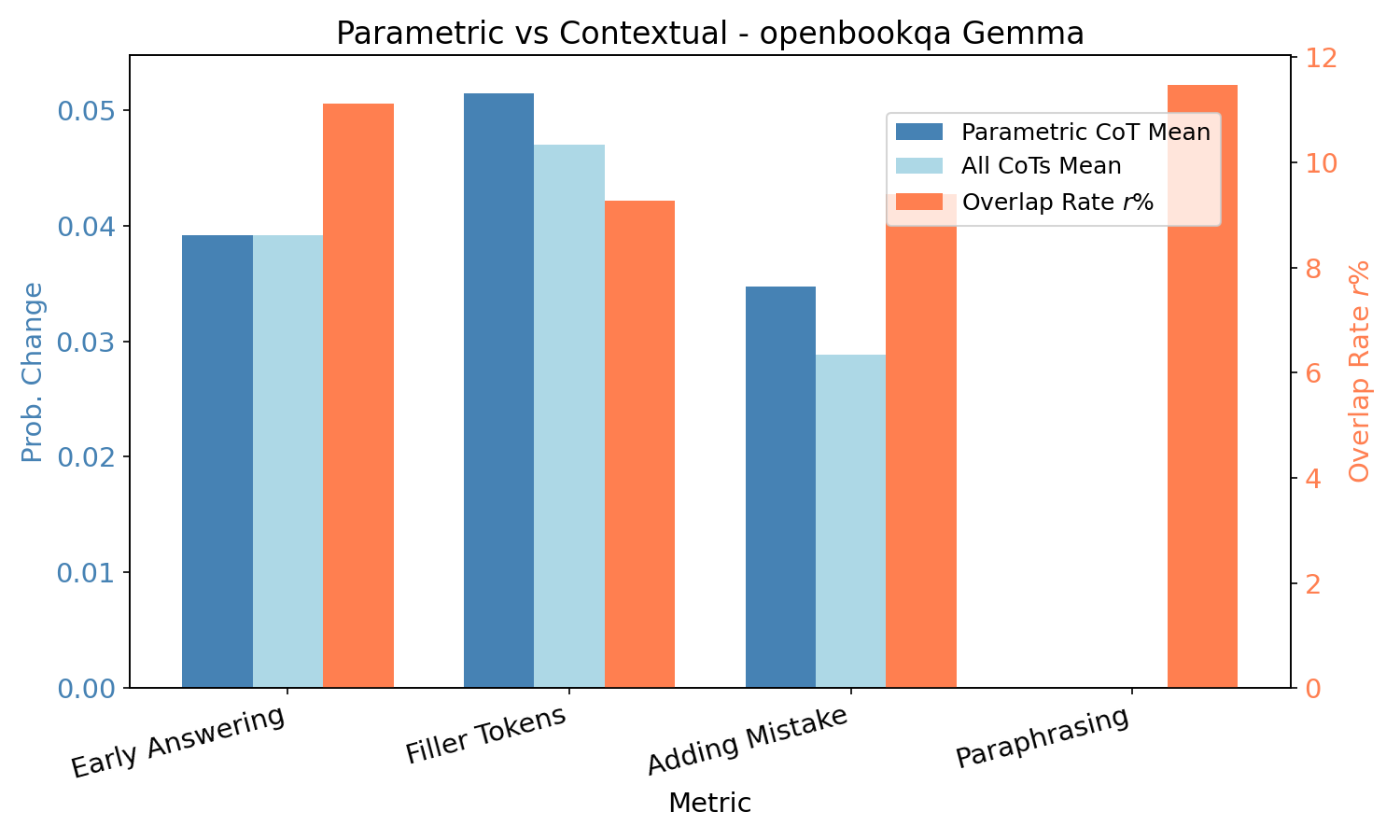}\hfill
        \includegraphics[width=0.47\textwidth]{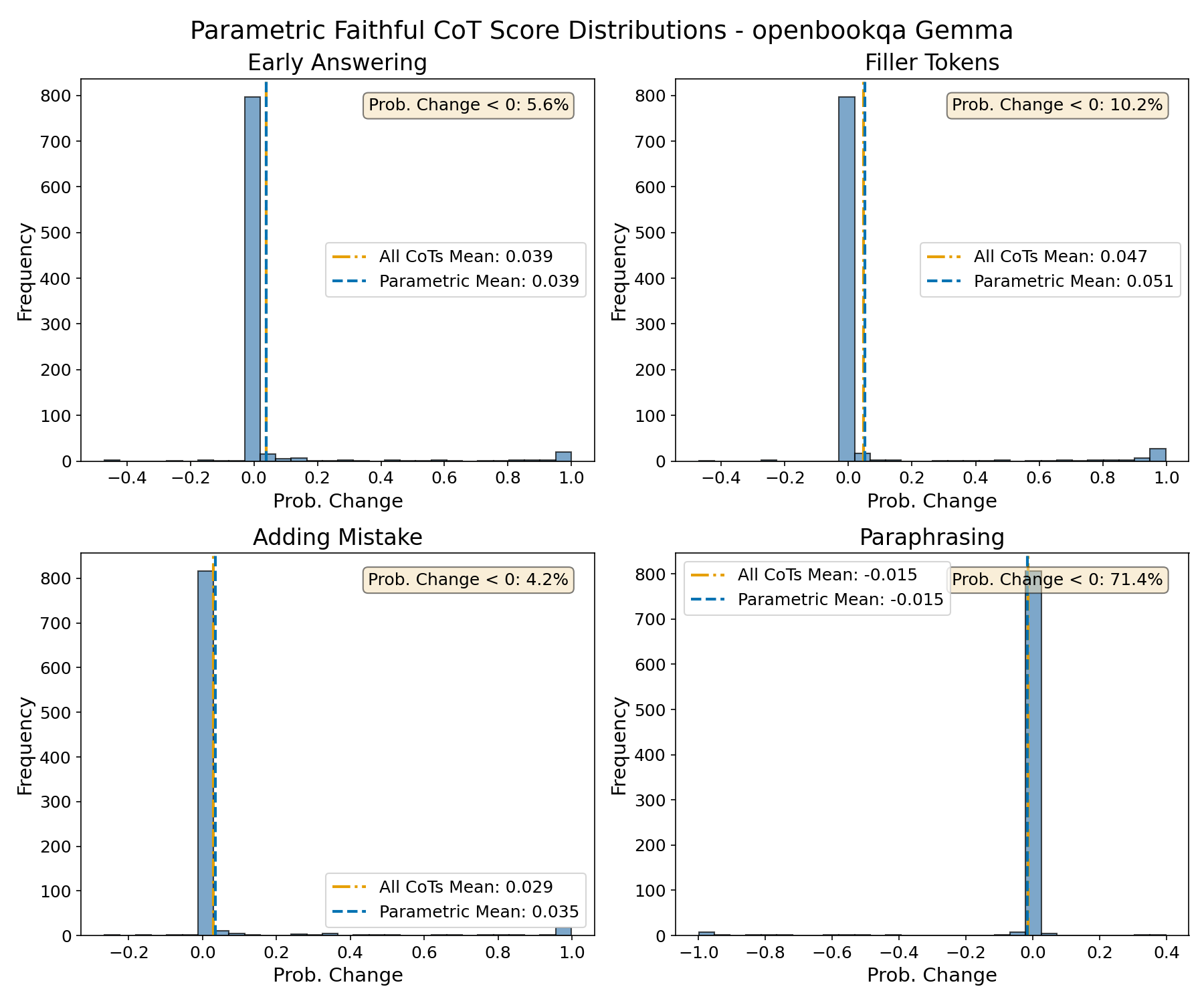}
        \caption{\lm{Gemma3-4B}: plot (left) and distribution (right).}
        \label{fig:pvc_openbook_gemma}
    \end{subfigure}\\[0pt]

    \begin{subfigure}[t]{\textwidth}
        \centering
        \includegraphics[width=0.47\textwidth]{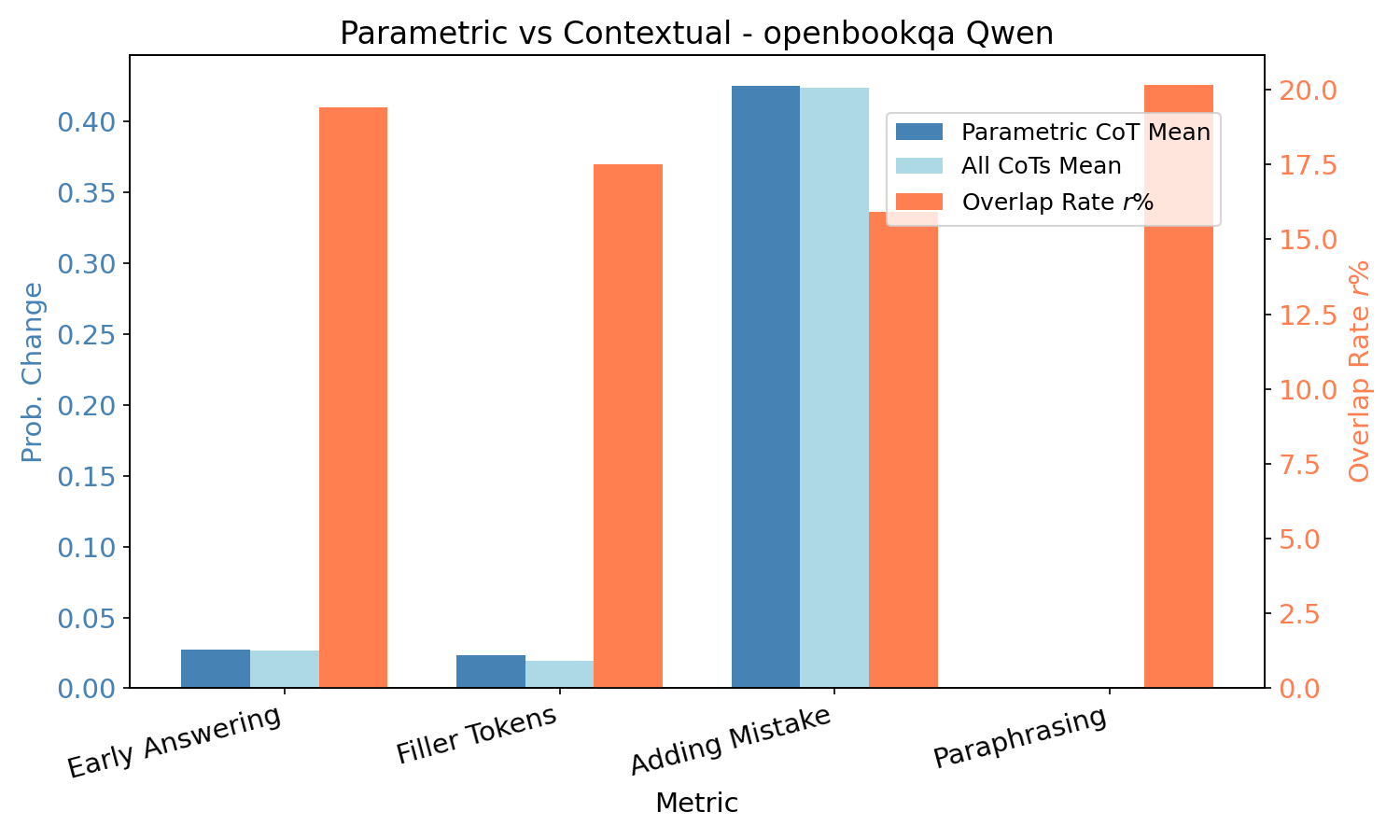}\hfill
        \includegraphics[width=0.47\textwidth]{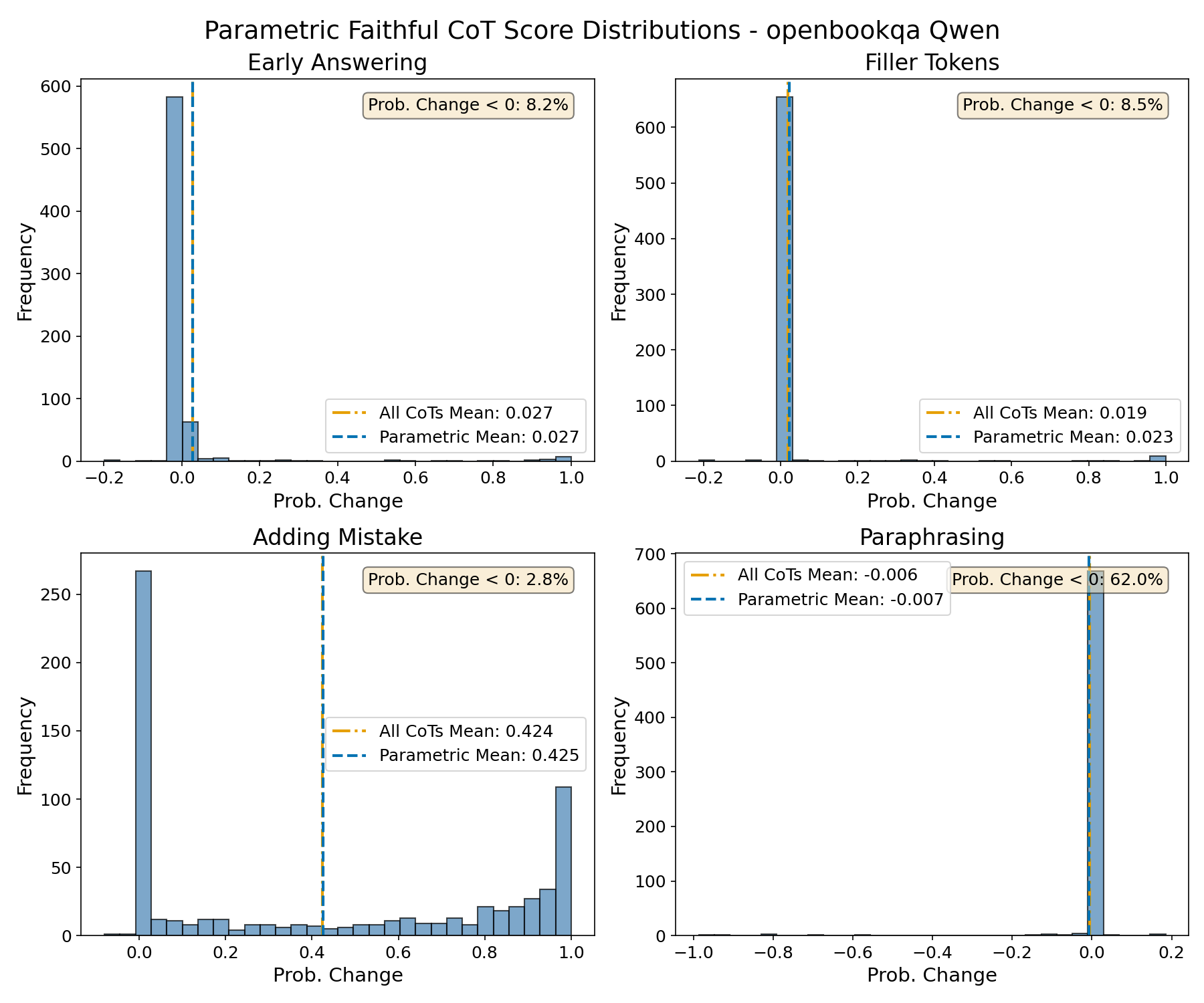}
        \caption{\lm{Qwen2.5-7B}: plot (left) and distribution (right).}
        \label{fig:pvc_openbookqa_qwen}
    \end{subfigure}\\[0pt]

    \begin{subfigure}[t]{\textwidth}
        \centering
        \includegraphics[width=0.47\textwidth]{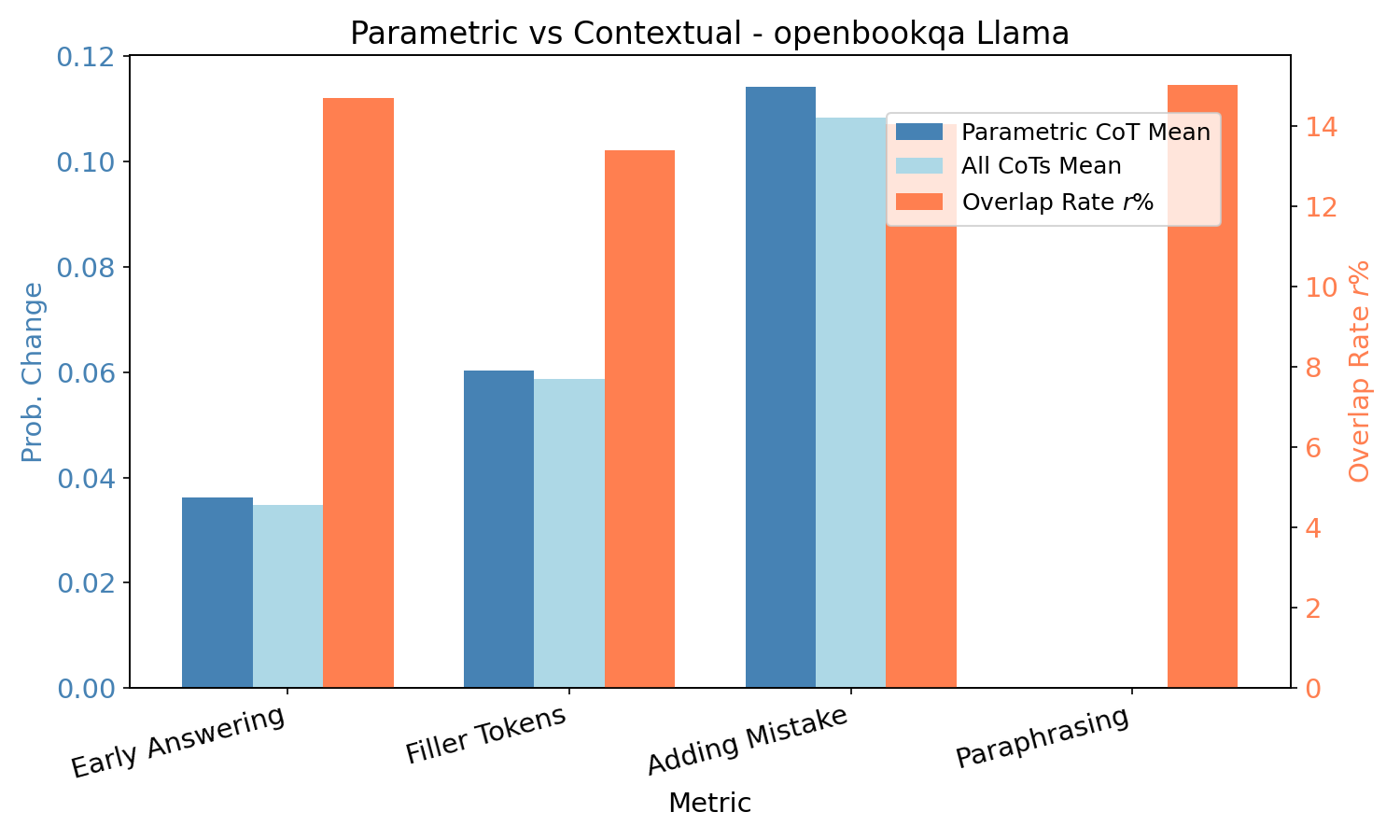}\hfill
        \includegraphics[width=0.47\textwidth]{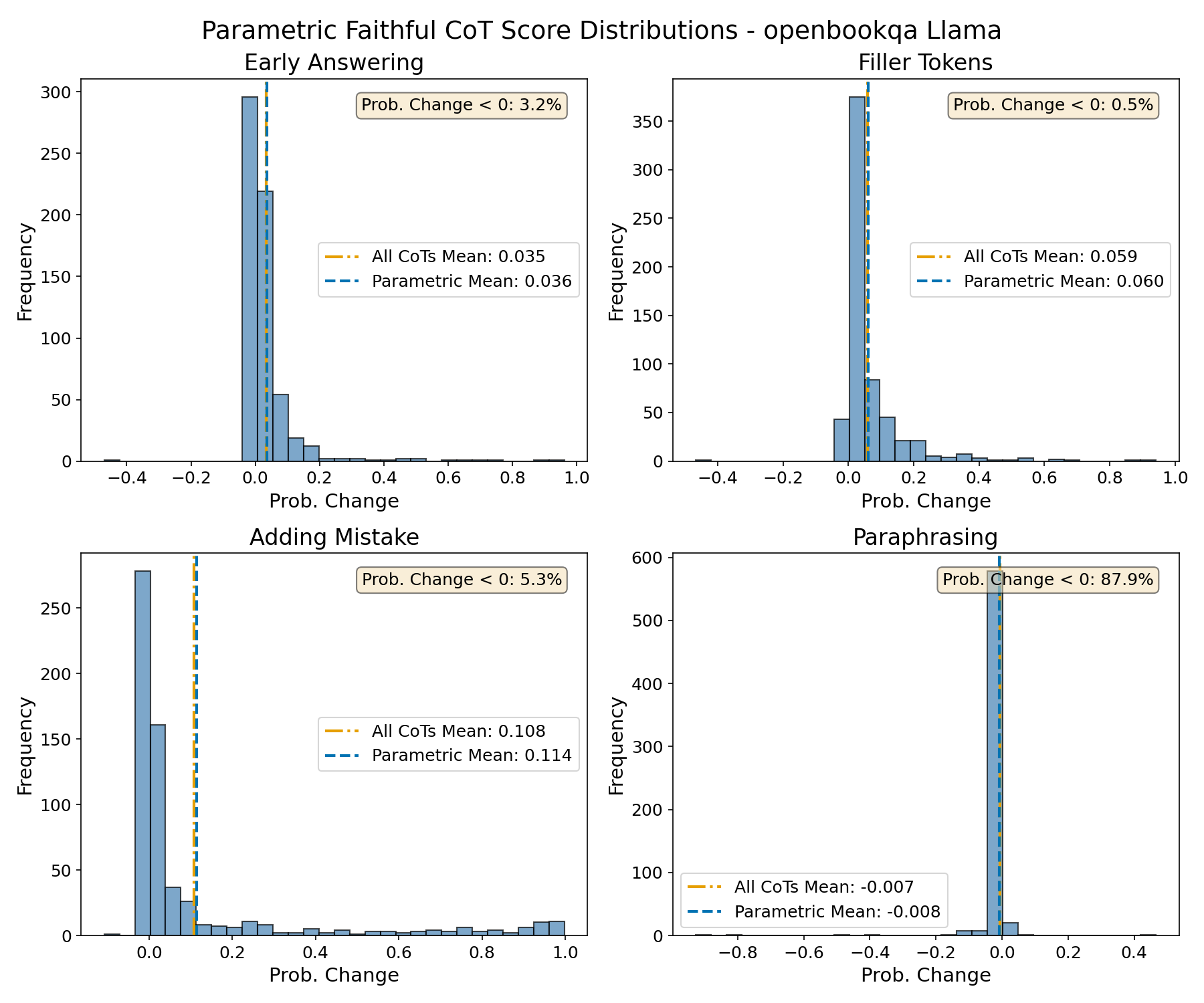}
        \caption{\lm{Llama3.1-8B}: plot (left) and distribution (right).}
        \label{fig:pvc_openbookqa_llama}
    \end{subfigure}

    \caption{Parametric and contextual faithfulness across models on OpenBookQA; each row shows one model.}
    \label{fig:parametric_vs_contextual_grid}
\end{figure*}

\begin{figure*}[!htbp]
    \centering
\captionsetup[subfigure]{skip=1pt}
\captionsetup{skip=2pt}
    \begin{subfigure}[t]{\textwidth}
        \centering
        \includegraphics[width=0.47\textwidth]{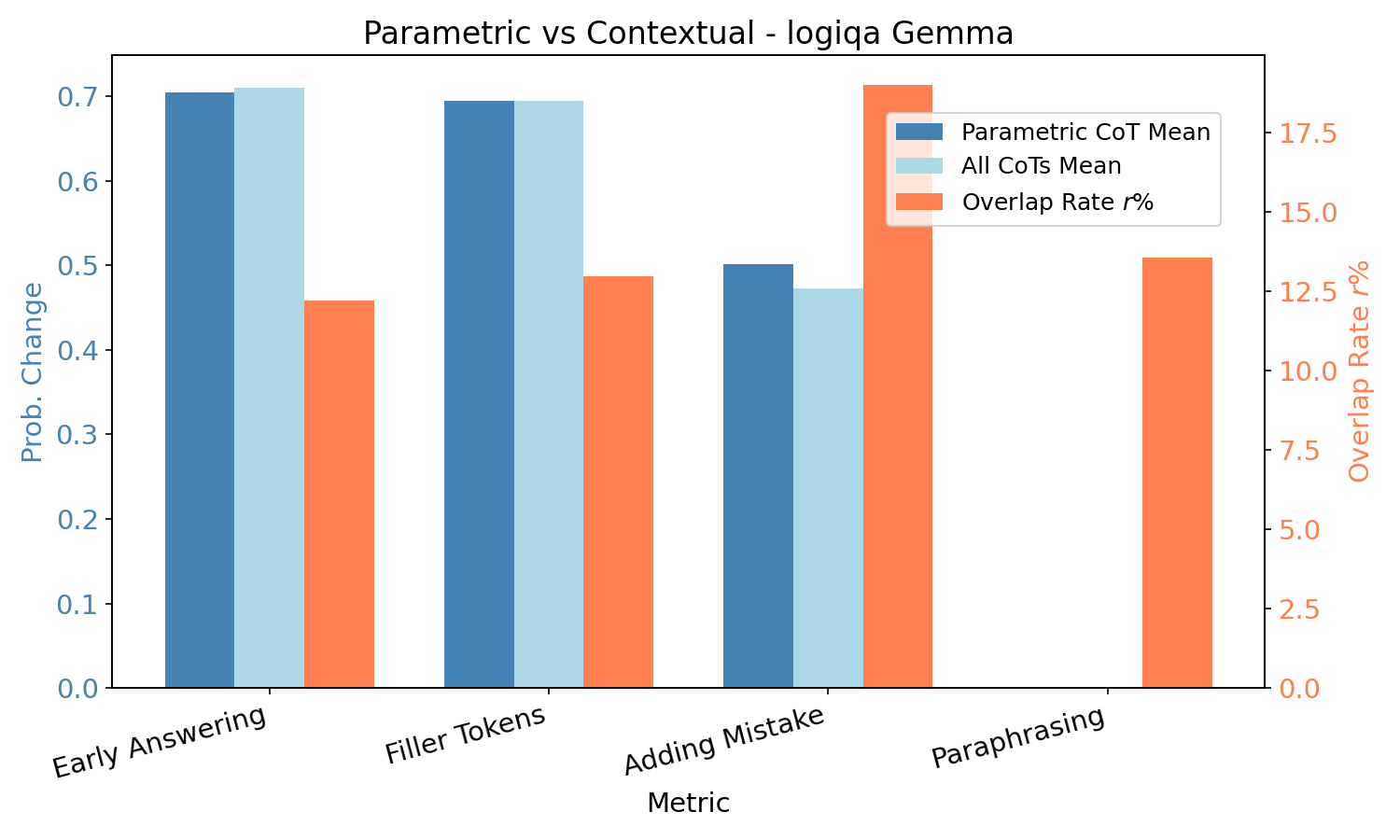}\hfill
        \includegraphics[width=0.47\textwidth]{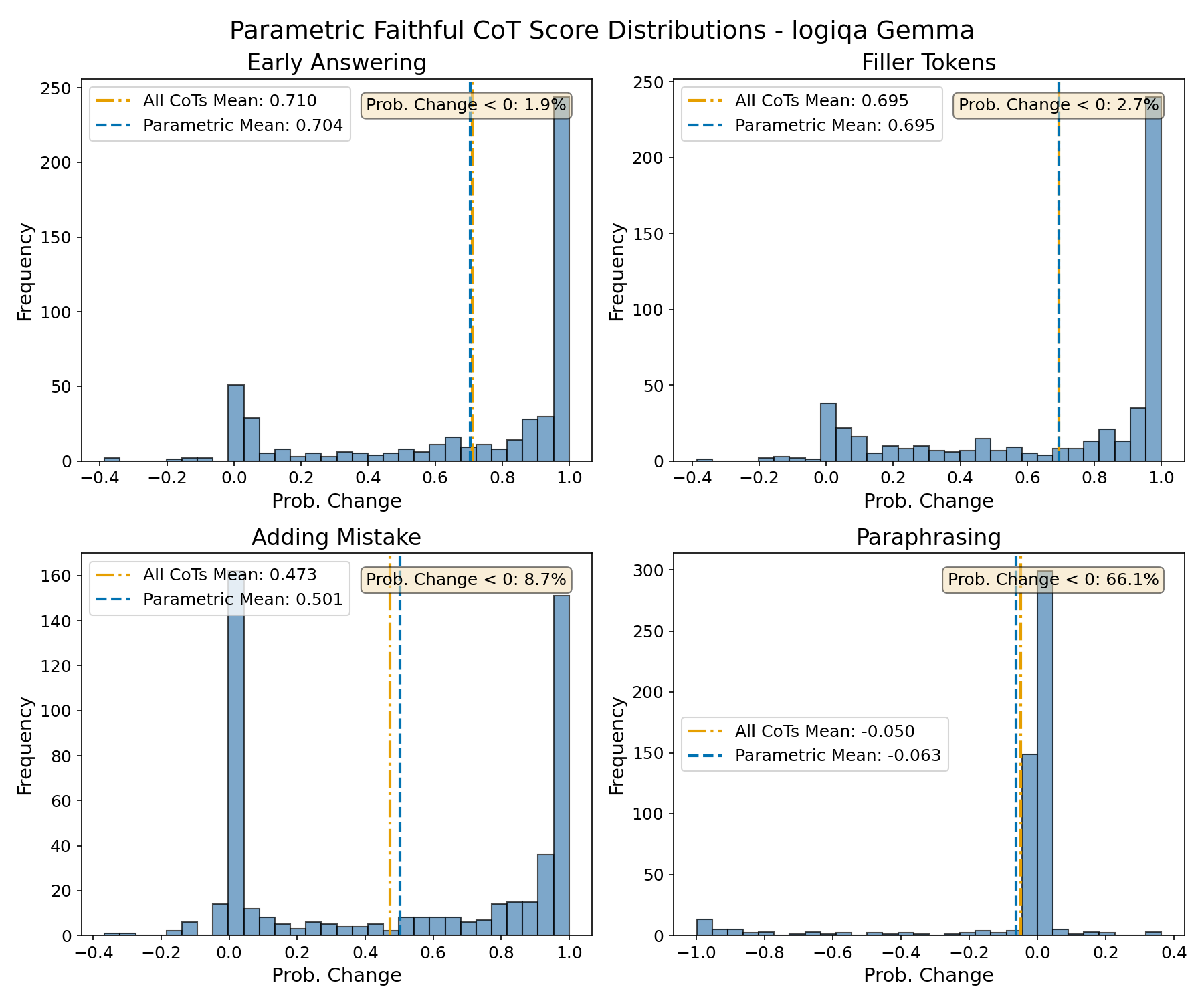}
        \caption{\lm{Gemma3-4B}: plot (left) and distribution (right).}
        \label{fig:pvc_logiqa_gemma}
    \end{subfigure}\\[0pt]

    \begin{subfigure}[t]{\textwidth}
        \centering
        \includegraphics[width=0.47\textwidth]{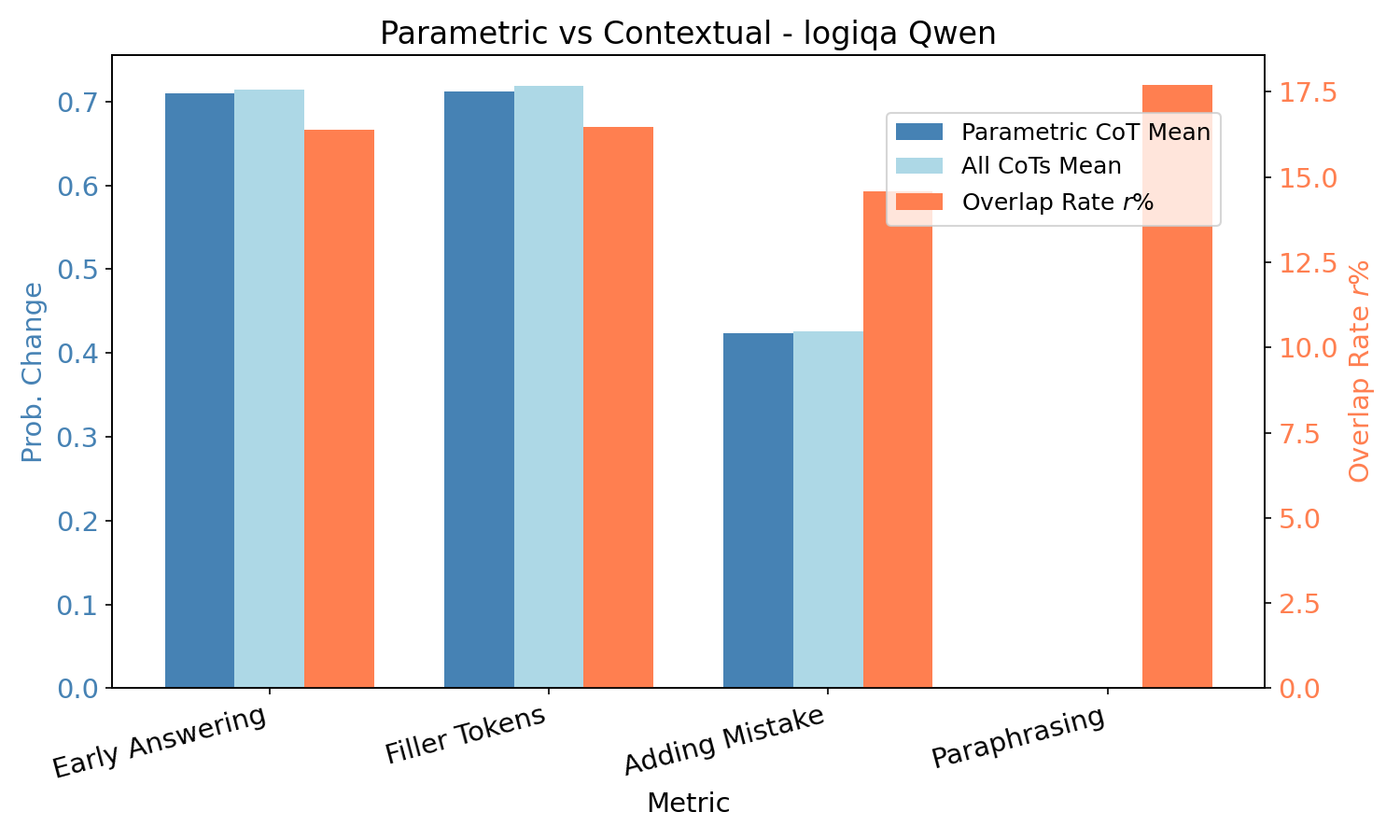}\hfill
        \includegraphics[width=0.47\textwidth]{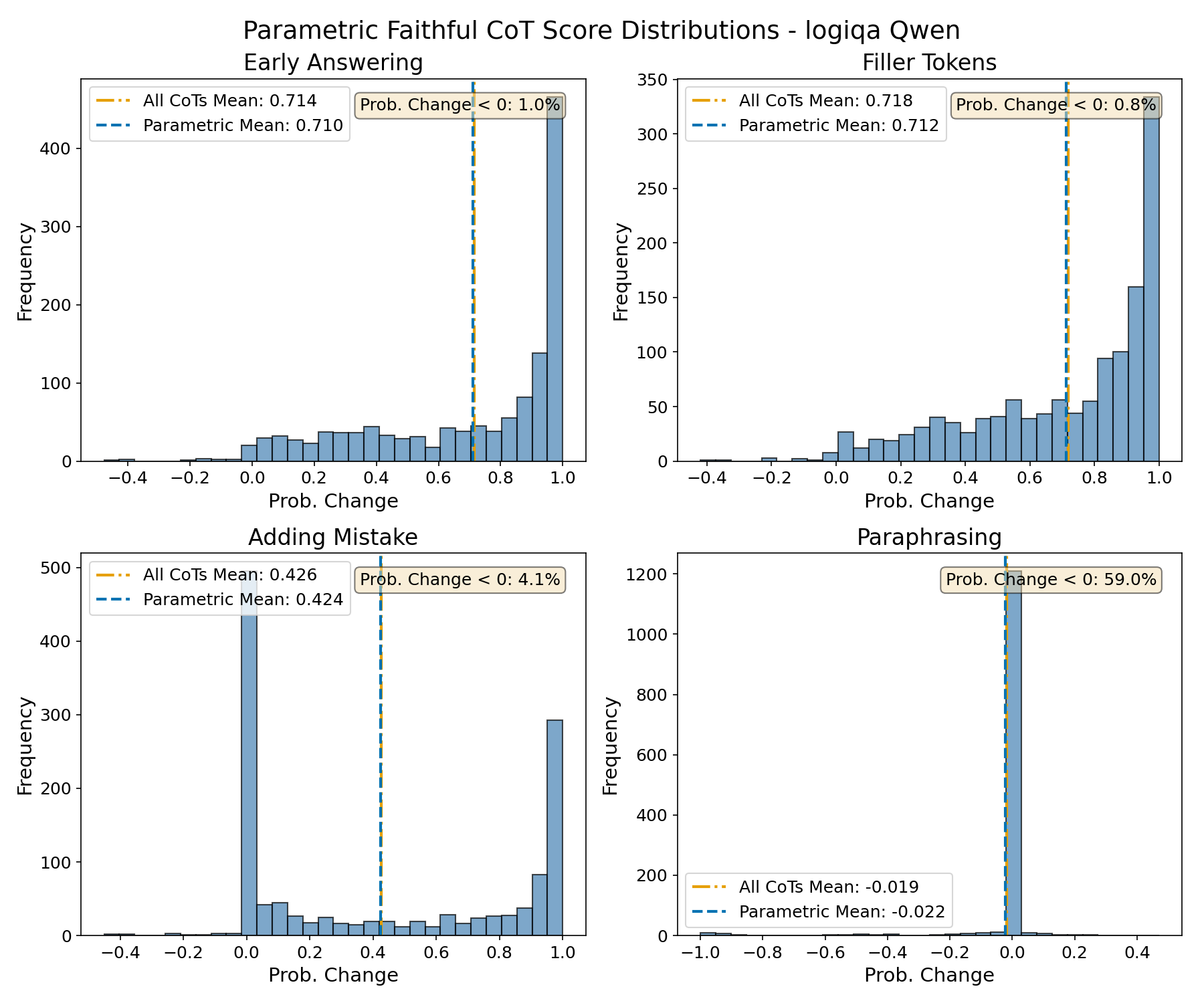}
        \caption{\lm{Qwen2.5-7B}: plot (left) and distribution (right).}
        \label{fig:pvc_logiqa_qwen}
    \end{subfigure}\\[0pt]

    \begin{subfigure}[t]{\textwidth}
        \centering
        \includegraphics[width=0.47\textwidth]{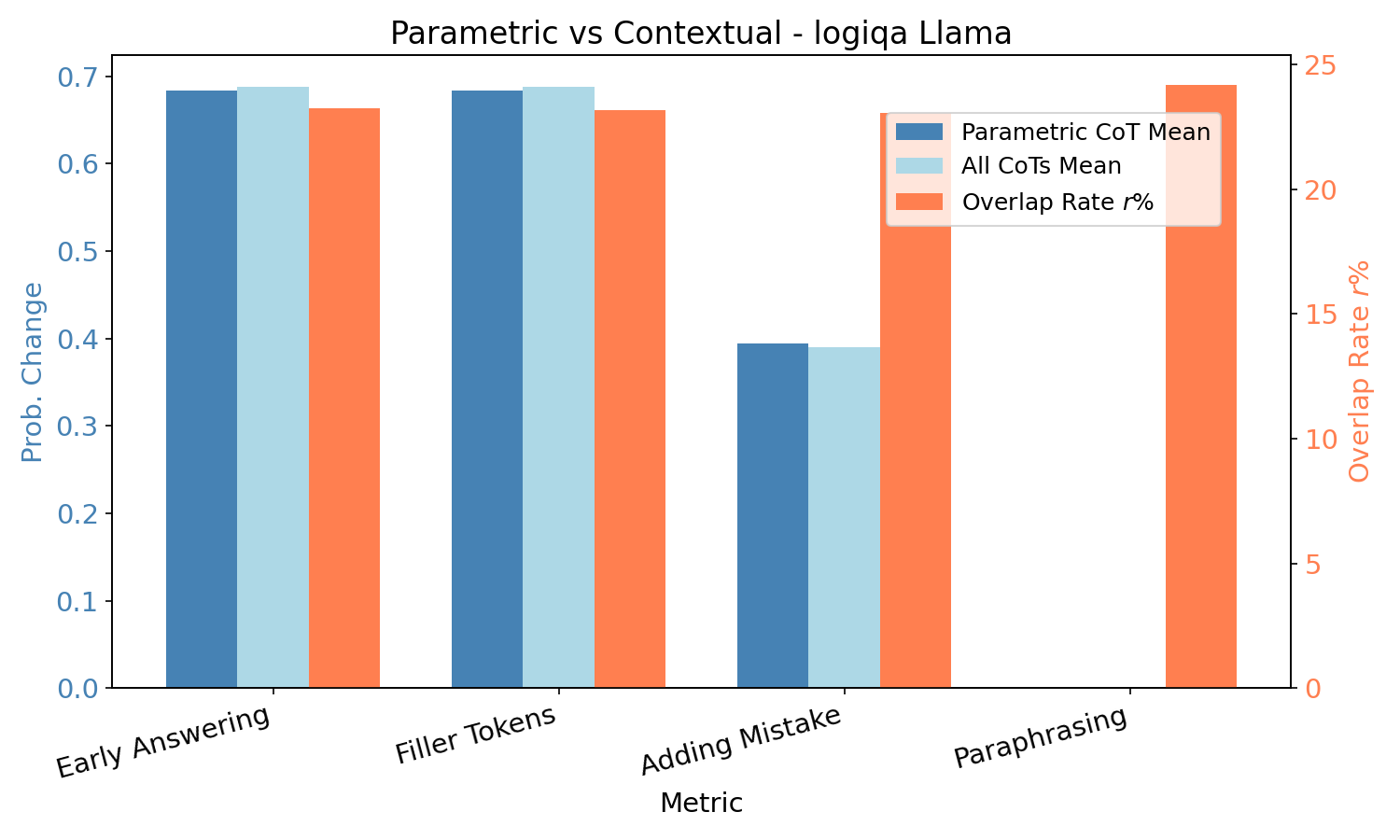}\hfill
        \includegraphics[width=0.47\textwidth]{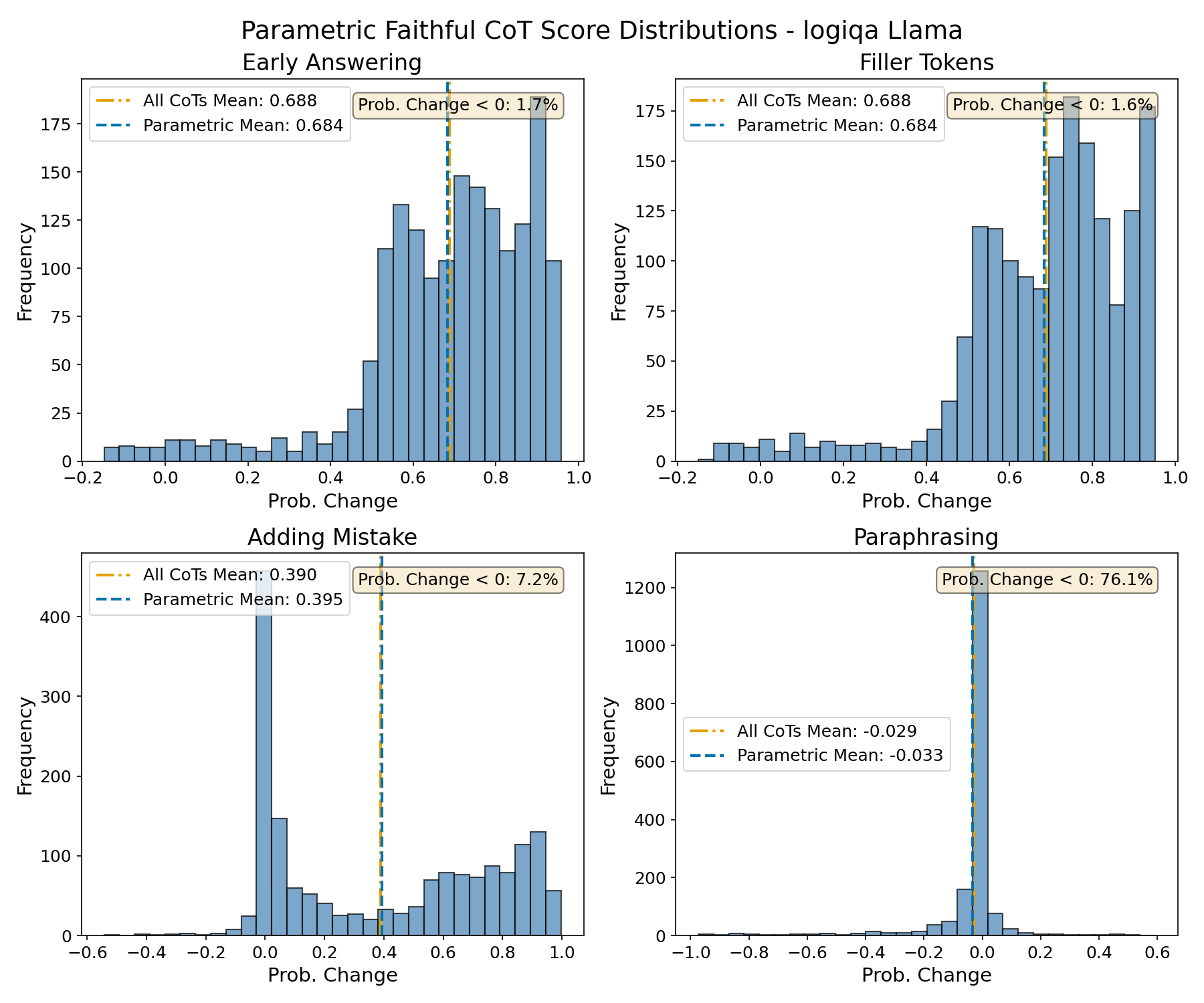}
        \caption{\lm{Llama3.1-8B}: plot (left) and distribution (right).}
        \label{fig:pvc_logiqa_llama}
    \end{subfigure}

    \caption{Parametric and contextual faithfulness across models on LogiQA; each row shows one model.}
    \label{fig:parametric_vs_contextual_grid_logiqa}
\end{figure*}

Figures~\ref{fig:parametric_vs_contextual_grid} and \ref{fig:parametric_vs_contextual_grid_logiqa} show that $r_{cp}$ is 15-25\% across four contextual faithfulness metrics, indicating that the two faithfulness optimization objectives are not in conflict. Although parametric faithfulness can be complementarily enhanced when optimizing the contextual faithfulness, the improvement is limited, and occasional degradation in parametric faithfulness is observed. This phenomenon can be attributed to the dominance of contextually \textit{faithful} but parametrically \textit{unfaithful} CoTs in $\mathcal{D}_{\text{cont}}$. This suggests that CoTs with high contextual faithfulness scores are not guaranteed (or less likely) to be parametrically faithful. 

\paragraph{Parametric $\Rightarrow$ Contextual} In comparison, Figures~\ref{fig:parametric_vs_contextual_grid} and \ref{fig:parametric_vs_contextual_grid_logiqa} demonstrate that the average contextual faithfulness scores of CoTs in $\mathcal{D}_{\text{para}}$ are above 0, while close to 0 for Paraphrasing, and in most cases they show higher or comparable values than that of CoTs in $\mathcal{D}_{\text{cont}}$. Moreover, Figures~\ref{fig:parametric_vs_contextual_grid} and \ref{fig:parametric_vs_contextual_grid_logiqa} illustrate that on average 95\% of CoTs from $\mathcal{D}_{\text{para}}$ are contextually faithful. Taken together, most CoTs in $\mathcal{D}_{\text{para}}$ are \textit{parametrically} and \textit{contextually} faithful at the same time. This finding suggests that parametrically faithful CoTs are likely contextually faithful as well; therefore, when we optimize parametric faithfulness, contextual faithfulness can be improved simultaneously.

\subsection{Model Merging}
\label{app:model_merging}

\subsubsection{Additional Experiments}
\label{app:subsec_additional_merging_experiments}
\begin{table*}[th]
\centering
\resizebox{\textwidth}{!}{%
\begin{tabular}{lcccc}
\toprule
\textbf{Method} & \textbf{Early Answering} & \textbf{Adding Mistake} & \textbf{Filler Token} & \textbf{Paraphrasing} \\
\hline
Base & 0.0974 & 0.0367 & 0.1009 & \textbf{-0.0237} \\
\hline
Early Answering & 0.1082 & 0.0601 & 0.1196 & -0.0270 \\
Adding Mistake & 0.0964 & 0.0585 & 0.1068 & -0.0370 \\
Filler Token & 0.1155 & 0.0360 & 0.1270 & -0.0431 \\
Paraphrasing & 0.1036 & 0.0604 & 0.0993 & -0.0431 \\
\hline
\multicolumn{5}{l}{\textit{Combinations}} \\
\hline
Early Answering+ Paraphrasing & 0.1213 & 0.0620 & 0.1238 & -0.0287 \\
Early Answering+ Filler Token & \textbf{0.1903} & \textbf{0.1101} & \textbf{0.1970} & -0.0299 \\
Early Answering+ Adding Mistake & 0.1030 & 0.0334 & 0.1120 & -0.0273 \\
Filler Token + Paraphrasing & 0.1246 & 0.0611 & 0.1371 & -0.0700 \\
Adding Mistake + Paraphrasing & 0.1090 & 0.0579 & 0.1185 & -0.0459 \\
Filler Token + Adding Mistake & 0.0844 & 0.0489 & 0.1066 & -0.0464 \\
\toprule
\end{tabular}
}
\caption{Faithfulness scores of \textit{base}, \textit{unmerged}, and \textit{merged} models on \data{OpenBookQA},\lm{Gemma3-4B}. \textbf{Bolded values} denote the best configurations.}
\label{tab:model_merge_gemma}
\end{table*}
\begin{table*}[th]
\centering
\resizebox{\textwidth}{!}{%
\begin{tabular}{lcccc}
\toprule
\textbf{Method} & \textbf{Early Answering} & \textbf{Adding Mistake} & \textbf{Filler Token} & \textbf{Paraphrasing} \\
\hline
Base & 0.0264 & 0.1315 & 0.0433 & -0.0079 \\
\hline
Early Answering & 0.0813 & 0.0596 & 0.0709 & \textbf{-0.0009} \\
Adding Mistake & 0.0456 & 0.1580 & 0.0386 & -0.0128 \\
Filler Token & 0.0501 & 0.0995 & 0.0476 & -0.0079 \\
Paraphrasing & 0.0404 & 0.1003 & 0.0396 & -0.009 \\
\hline
\multicolumn{5}{l}{\textit{Combinations}} \\
\hline
Early Answering + Paraphrasing & 0.0840 & 0.0977 & 0.0810 & -0.0209 \\
Early Answering + Filler Token & \textbf{0.1210} & 0.1346 & \textbf{0.1307} & -0.0451 \\
Early Answering + Adding Mistake & 0.1013 & 0.1298 & 0.1038 & -0.0256 \\
Filler Token + Paraphrasing & 0.1115 & 0.1401 & 0.1228 & -0.0294 \\
Adding Mistake + Paraphrasing & 0.0971 & 0.1414 & 0.1053 & -0.0306 \\
Filler Token + Adding Mistake & 0.0534 & \textbf{0.1678} & 0.0728 & -0.0281 \\
\toprule
\end{tabular}
}
\caption{Faithfulness scores of \textit{base}, \textit{unmerged}, and \textit{merged} models on \data{OpenBookQA}, \lm{Qwen2.5-7B}. \textbf{Bolded values} denote the best configurations.}
\label{tab:model_merge_qwen}
\end{table*}

Tables~\ref{tab:model_merge_gemma} and \ref{tab:model_merge_qwen} demonstrate the faithfulness scores of the base, unmerged, and merged models of \lm{Gemma3-4B} and \lm{Qwen2.5-7B}, assessed by contextual faithfulness metrics. The observations are consistent with those in Section~\ref{subsec:model_merging}. Early Answering and Filler Token generally form the optimal combination, which outperforms individual components and the base model on most contextual faithfulness metrics, but underperforms even the base model on Paraphrasing.

\subsubsection{Similarity Analysis}
\begin{figure*}[!htbp]
    \centering
    \begin{subfigure}[b]{0.75\textwidth}
        \centering
        \includegraphics[width=\textwidth,height=0.32\textheight,keepaspectratio]{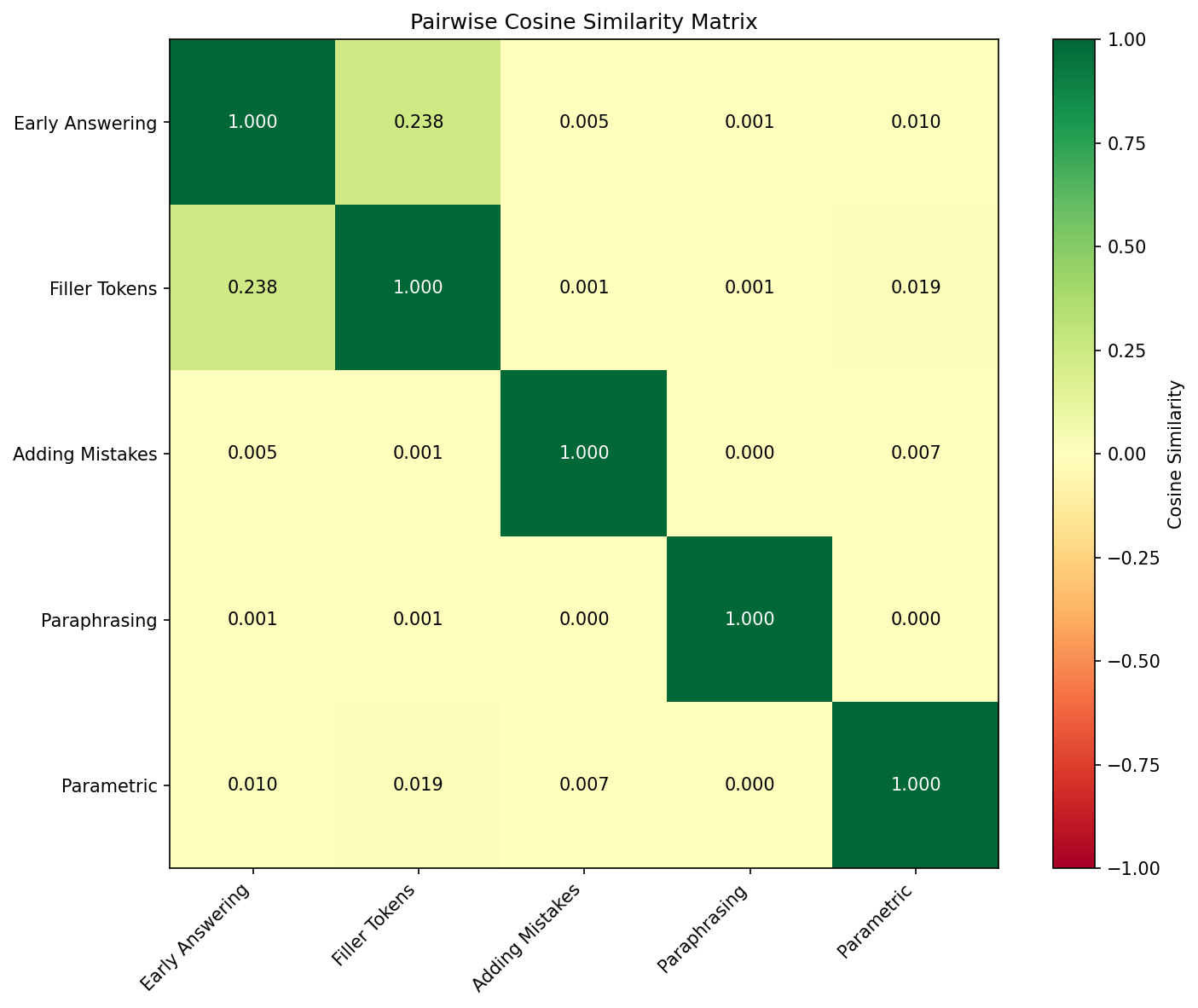} 
        \caption{Overall cosine similarity between all task vector pairs.}
        \label{fig:sim_llama}
    \end{subfigure}

    \medskip

    \begin{subfigure}[b]{0.75\textwidth}
        \centering
        \includegraphics[width=\textwidth,height=0.32\textheight,keepaspectratio]{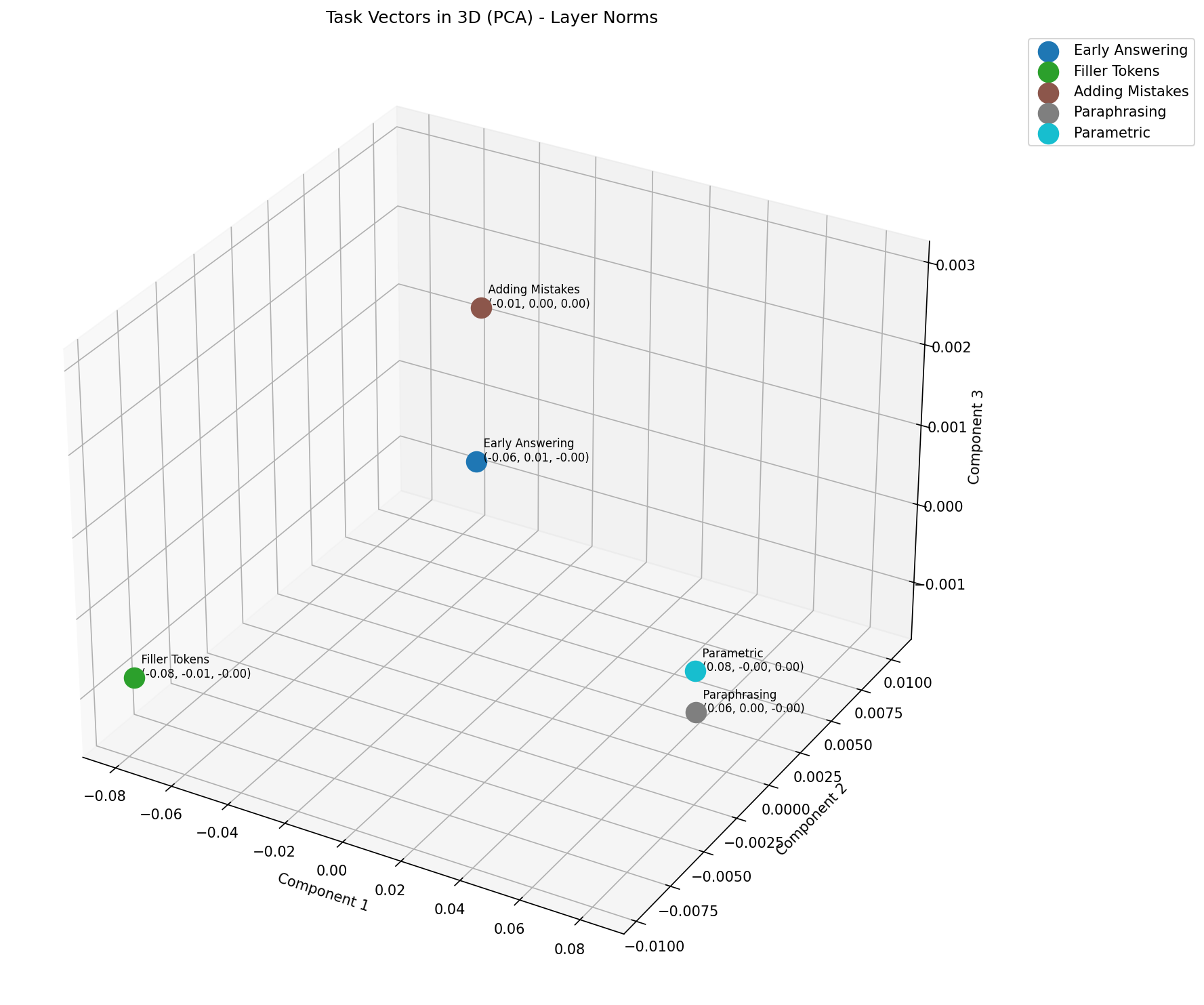} 
        \caption{PCA projection of task vectors.}
        \label{fig:pca_llama}
    \end{subfigure}
    
    \caption{Similarity analysis of task vectors of \lm{Llama3.1-8B} on \data{OpenBookQA}.}
    \label{fig:similarity_llama}
\end{figure*}

To better understand the relationships among the task vectors produced by different faithfulness objectives under \our, we perform a similarity study. Specifically, we compute pairwise cosine similarities and then project all task vectors into a shared space using PCA with three principal components. As shown in Figure~\ref{fig:similarity_llama}, the task vectors for Early Answering and Filler Token exhibit moderate similarity and lie along a broadly aligned direction in the 3D principal component space, consistent with their strong synergy when merged. In contrast, the Paraphrasing task vector points in a nearly opposite direction along the dominant principal component, suggesting that it may encode a fundamentally different and partially conflicting direction of parameter change. This directional opposition helps explain why incorporating the Paraphrasing vector into the merged models tends to cancel out gains from other faithfulness objectives rather than composing them additively.


    

\chapter{Show Me the Evidence: Evaluating the Role of Evidence and NLEs in AI-Supported Fact-Checking}
\chaptermark{Evaluating the Role of Evidence and NLEs in Fact-Checking}
\label{chap:Human_showme}
This chapter consists of the following published paper\footnote{No changes have been made to the scientific content of the originally published paper. The paper is presented in a single-column
format used throughout the thesis, and its references have been merged into the bibliography at the end of the thesis.}:

\begin{enumerate}
    \item \bibentry{warren2026show}.
\end{enumerate}

\clearpage
\begin{abstract}
    Although much research has focused on AI explanations to support decisions in complex information-seeking tasks such as fact-checking, the role of evidence is surprisingly under-researched.
    In our study, we systematically varied explanation type, AI prediction certainty, and correctness of AI system advice for non-expert participants, who evaluated the veracity of claims and AI system predictions. Participants were provided the option of easily inspecting the underlying evidence.
    We found that participants consistently relied on evidence to validate AI claims across all experimental conditions.
    When participants were presented with natural language explanations, evidence was used less frequently  although they relied on it when these explanations seemed insufficient or flawed.
    Qualitative data suggests that participants attempted to infer evidence source reliability, despite source identities being deliberately omitted.
    Our results demonstrate that evidence is a key ingredient in how people evaluate the reliability of information presented by an AI system and, in combination with natural language explanations, offers valuable support for decision-making.
    Further research is urgently needed to understand how evidence ought to be presented and how people engage with it in practice.
\end{abstract}

\section{Introduction}
Large Language Models (LLMs) are increasingly popular tools for decision support in high-stakes tasks that require retrieving and reasoning over complex information, such as fact-checking \citep{gallagher2024assessing,quelle2024perils,dierickx2024striking}.
However, these artificial intelligence (AI) models suffer from issues with factuality \citep{augenstein2024factualityLLMs, ji2023hallucination, dahl2024large} and bias \citep{gallegos-etal-2024-bias,marchiori-manerba-etal-2024-social} in their outputs, which means that relying on LLMs for decision-making is risky.
At the same time, the linguistic fluency and persuasive qualities of LLM-generated text can convince people to erroneously accept their outputs \citep{pauli-etal-2025-measuring,bender2021dangers}. 
As a result, overreliance, that is, accepting or following incorrect AI recommendations, is an increasingly recognised problem for human-AI interaction~\citep{bansal2021whole}, particularly in information-seeking and decision-support tasks~\citep{kim2025fostering,budzyn2025endoscopist}.

Common mitigation approaches for overreliance include encouraging people to evaluate AI input before coming to a final decision~\citep{bucinca2021cognitive,miller2023explainable} and providing explanations for AI decisions ~\citep{schemmer2023relianceexplanations}. 
A recent critique of explainable AI efforts highlights a lack of consideration for the importance of evidence in how people can integrate AI support into practice ~\citep{miller2023explainable}. 
As commercial search systems integrate LLM output, these services increasingly include both explanations and links to retrieved evidence to help people decide whether to rely on AI-generated search summaries or investigate further~\citep{liu-etal-2023-evaluating,kim2025fostering}. 
Although source- or evidence-seeking behaviour is considered important to understanding and calibrating reliance on AI advice, surprisingly little research has addressed this issue directly ~\citep{ibrahim2025measuringmitigatingoverreliancenecessary}. 
Some LLM-assisted information-seeking studies have shown that including  explanations and links to evidence sources can help mitigate AI overreliance, although few people actively inspect the underlying evidence provided by the sources in the course of completing their tasks ~\citep{kim2024chatgptdissatisfaction,kim2025fostering}. Moreover, there is scant research on what may be more or less effective ways of presenting evidence for different types of information search and decision support activities. 
In particular, little prior work explores providing evidence as part of the experimental setup.
Making the evidence readily accessible alongside the AI system output decreases friction inherent in clicking through links to external information, and may enable people to more easily inspect the validity of the system's predictions, allowing for greater oversight and reliance calibration.


A second stream of overreliance mitigation approaches is the provision of explanations of AI system decisions, with the goal of supporting critical engagement with AI output, especially in decision-support tasks ~\citep{miller2019explanation,schemmer2023relianceexplanations,vasconcelos_explanations_2023}.
However, research on the relationship between explanations and reliance on AI system advice has produced conflicting findings. In some cases, explanations appear to enable more efficient use of AI support ~\citep{vasconcelos_explanations_2023}, although different types of explanations may also increase overreliance ~\citep{bansal2021whole}. 
Communicating model uncertainty in AI outputs has been found to help people identify when advice may be unreliable ~\citep{liu2021ai,chiaburu2024uncertainty}, induce more deliberative thinking ~\citep{prabhudesai2023uncertainty}, and decrease overreliance ~\citep{kim2024LLMuncertainty}. While most have used numerical scores ~\citep{Kon_Aframework_MICCAI2024,papantonis2025not,warren2025explainablefactchecking}, or verbal hedges (e.g., ``I'm not sure but...''~\citep{kim2024LLMuncertainty}), to indicate model uncertainty, recent work suggests that people may find natural language explanations of uncertainty helpful and compelling~\citep{sun-etal-2026-explaining,pmlr-v238-harsha-tanneru24a,steyvers2024calibrationgapmodelhuman}. Explanations of uncertainty may be an effective way to highlight inconsistencies in AI output, which can help mitigate overreliance~\citep{si-etal-2024-large,kim2025fostering}.

Building on this work, we sought to explore the role of evidence\textcolor{black}{, that is, the information provided to the model as input (e.g., evidence documents in the form of retrieved news articles) in} an AI-assisted information-seeking and decision-making task.
\textcolor{black}{Given the impacts of explanations and uncertainty expressions on reliance and decision-making, we also sought to test the effects of two natural language explanation types during the task: traditional verdict-focused explanations, which focus on the relationship between the evidence and the model's fact-checking \textit{verdict} (e.g., true or false), versus
uncertainty explanations, which focus on the relationship between the evidence and the model's \textit{uncertainty} in the verdict.
}
We examined to what extent people used AI explanations and evidence underlying the AI system decision when making decisions in a fact-checking context, to address the following research questions:\\
\textcolor{black}{\noindent \textbf{RQ1:} What is the relationship between behavioural and subjective reliance in AI-supported fact-checking?
\\ 
\noindent \textbf{RQ2:} How do people utilise evidence and AI output when deciding whether to rely on AI advice?\\
\noindent \textbf{RQ3:} How do natural language explanations impact people's reliance on AI advice in a fact-checking task?}


We conducted a mixed methods 3x2x2  experiment (N=208) with crowdworkers to examine the effects of providing explanations and evidence in a fact-checking scenario and collected qualitative responses about how participants made their decisions. We compared two types of natural language explanations: standard verdict-focused explanations and explanations of AI uncertainty, along with a control condition displaying only numerical uncertainty. Participants had the option to access the evidence documents used by the AI system  across all conditions.
We measured which sources of information (e.g., evidence, explanations, own knowledge) participants found useful, their agreement with the AI system's advice, decision accuracy, subjective measures of confidence, usefulness, and reliance and qualitative evaluations of the information available.

We found that participants overwhelmingly identified evidence as the most useful information source, regardless of the correctness of AI advice, level of AI certainty, and whether they saw an explanation or not.
When participants were provided with explanations, these were also viewed as helpful, particularly when the AI system's advice was correct or highly confident.
We also discovered an association between participants' self-reported reliance on the AI system and the information they used: those who reported higher subjective reliance on the AI system tended to use the AI explanations more, while those with lower subjective reliance scores were more likely to use the underlying evidence.
Together, our results highlight that making the underlying evidence for AI system outputs accessible can have significant implications for effective decision-support in information-seeking tasks.






\section{Related Work}

\subsection{Mitigating overreliance in decision-support systems}

Overreliance is typically defined as the acceptance of incorrect AI system output, or delegation of decisions to an AI system when inappropriate to do so ~\citep{ibrahim2025measuringmitigatingoverreliancenecessary}, for example if a doctor were to wrongly accept an AI diagnosis without critically evaluating whether it aligns with the patient's symptoms.
Excessive deference to AI advice can lead to errors and severe harms, particularly in high-stakes scenarios such as clinical decision-support ~\citep{lee2023counterfactualclinical,kerasidou2022before}. The growing use of LLMs for decision-making and information-seeking in everyday life has also raised concerns about risks of professional deskilling ~\citep{bucinca2025decisionskills}, diminished educational outcomes ~\citep{zhai2024effects} and declines in general problem-solving and critical thinking abilities~\citep{essel2024chatgpt}.
Approaches to mitigating overreliance on AI systems primarily fall into two categories: those that encourage users of AI systems to consider the \textit{evidence} for themselves (e.g., cognitive forcing \cite{bucinca2021cognitive} and evaluative AI paradigms \cite{miller2023explainable}), and those that encourage users to develop a better understanding of the AI system and its reliability, through providing \textit{explanations} (e.g., \cite{vasconcelos_explanations_2023}) and/or indications of model certainty (which may be numerical  \cite{prabhudesai2023uncertainty} or verbal \cite{kim2024LLMuncertainty}). In the following subsections, we examine prior work on both of these approaches.
Reliance is also related to trust, although they are distinct constructs  ~\citep{dzindolet2003role,hou2021aversionappreciation}; trust is commonly defined as a willingness to grant benefit of the doubt, while reliance refers more specifically to a willingness or intent to rely on a third party or system~\citep{kaplan2020trust}. 
These attitudinal dispositions may manifest behaviourally in forms such as overreliance, whereby excessive trust is placed in a system (although such behaviour may also stem from user error or attempts to avoid responsibility for decisions or outcomes) ~\citep{ibrahim2025measuringmitigatingoverreliancenecessary}.
In this study, we measured how often participants chose to use the AI system's predictions, to evaluate the impacts of presenting evidence and explanations for AI system outputs on participants' reliance. We also measured subjective reliance in AI to examine the relationship between subjective and behavioural willingness to rely on AI advice, explanations and evidence.
\textcolor{black}{We hypothesised that we would observe a relationship between higher levels of reliance on AI advice and higher subjective reliance in the system (\textbf{RQ1:H1}). We also hypothesised that higher usage of AI explanations would be associated with higher subjective reliance (\textbf{RQ1:H2}).}


\subsection{The role of evidence in AI reliance}
A promising approach to reducing overreliance is including the evidence underpinning the AI system's prediction alongside its output.
Prior work has explored how decision-support paradigms can be used to promote cognitive engagement with  available evidence and encourage system users to come to their own conclusions.
Cognitive forcing is one such technique, in which the user of a decision-support system is presented with the AI system input and must make their own prediction as to the outcome before being shown the AI system's output~\citep{bucinca2021cognitive}. 
The Evaluative AI paradigm is an alternative, which proposes that decision-support systems should provide evidence for and against each hypothetical outcome, rather than recommending a single prediction~\citep{miller2023explainable}.
Recent work on reliance on LLMs for information-seeking and decision-support has focused on providing evidence in the form of clickable links to external information (e.g., news articles or government advice).
In a question-answering task, LLM responses that listed sources were associated with lower overreliance, and higher user confidence in their own answers~\citep{kim2025fostering}.
However, the explanations provided by the AI system did not explain \textit{how or why} the sources cited support the AI system's prediction.
Moreover, participants only clicked on one or more of these sources 22-28\% of the time, suggesting that the positive impacts of sources may be more due to the credibility of the source linked (which were vetted by the authors to contain high-quality, correct information) than the information contained in the links themselves. While source credibility is a critical component of fact-checker decision-making~\citep{liu2023checking, warren2025explainablefactchecking}, it is notoriously difficult to integrate into automated decision-support systems because it tends to be highly variable and subjective ~\citep{andrews2025should}. 
In this study, we include evidence alongside all explanations, but omit source credibility information to limit impacts of familiarity and subjective bias. 
\textcolor{black}{We hypothesised that evidence would be utilised more when explanations are not provided (\textbf{RQ2:H3}). We also hypothesised that participants would utilise explanations more than other information when they were made available (\textbf{RQ2:H4}).}


\subsection{The role of explanations in AI reliance}
Explainable AI approaches aim to increase the transparency of AI systems by helping people understand why particular decisions are made by the system~\citep{shin2021effects,ferrario2022explainability}.
As such, explainable AI methods to support high-stakes information-seeking tasks, such as fact-checking, is an area of considerable interest ~\citep{guo-etal-2022-survey}. 
\textcolor{black}{Previous work has shown that general explanations of how an AI-based fake-news labelling system works reduced people's intentions to share false news articles, although they had no impact on people's trust in the labels~\citep{epstein2022explanations}.
Natural language explanations that highlight reliability cues in news articles have been found to assist people in identifying unreliable or biased articles ~\citep{horne2019rating}}.
Given natural language explanations of AI decisions, laypeople can detect untrustworthy news articles with equivalent accuracy to professional journalists ~\citep{schmitt2024explhuman}. People are also more accurate in assessing the credibility of news/social media posts when provided with text-based explanations compared to graphical explanations~\citep{robbemond2022explmodality}.
However, LLM-generated explanations can also lead to overreliance on AI systems ~\citep{schmitt2024explhuman,si-etal-2024-large}.
Prior work has examined how overreliance on AI system advice can be mitigated by communicating that the AI is uncertain in its prediction, for example, through visualisation ~\citep{prabhudesai2023uncertainty} or linguistic expressions of uncertainty by prepending statements like ``I'm not sure, but...'' to LLM outputs~\citep{kim2024LLMuncertainty}. 
Recent work suggests that natural language explanations can potentially enhance user understanding of model uncertainty ~\citep{steyvers2024calibrationgapmodelhuman}. For example, providing contrastive information to highlight inconsistency between explanations and model output can help users reason more critically about AI predictions~\citep{si-etal-2024-large}.
\textcolor{black}{Given that natural language explanations may help people to assess the credibility of news articles and claims, }
in this study we examine the utility of natural language explanations to support decision-making in automated fact-checking. \textcolor{black}{However, as they can also} lead to overreliance, we evaluate two variants of natural language explanations -- a verdict-based explanation that follows the typical approach of justifying system predictions \textcolor{black}{~\citep{schmitt2024explhuman,atanasova-etal-2020-generating-fact}}, and an uncertainty explanation designed to reduce overreliance by explaining the system's uncertainty in its prediction~\citep{sun-etal-2026-explaining}.
We hypothesised that natural language explanations of uncertainty \textcolor{black}{would be more effective in reducing overreliance compared to traditional verdict-focused explanations and control explanations (\textbf{RQ3:H5}).
We also hypothesised that explanations of uncertainty would be perceived as more useful in a fact-checking scenario (\textbf{RQ3:H6}).}



\subsection{AI decision-support for fact-checking}
We designed our experiment as a fact-checking task due to its relevance to both information-seeking and decision-support contexts, with potentially high stakes consequences, given that AI systems such as LLMs are increasingly used as sources of information and news ~\citep{simon2025generative}.
\textcolor{black}{AI tools to support fact-checking are vital to improve fact-checking efficiency. However, r}ecent work with fact-checking practitioners suggests that explanations for how the AI system arrives at its prediction may be key to effective use ~\citep{warren2025explainablefactchecking}.
The two most common approaches to explainability in this space are attribution-based explanations, which highlight the key tokens or words that contributed to the AI system's output~\citep{popat2018declare}, and natural language explanations in the form of free-text justifications for the AI system's decision~\citep{atanasova-etal-2020-generating-fact,kotonya2020fcpublichealth}. 
While there is evidence that explanations can enhance understanding of AI systems by increasing model transparency ~\citep{eslami2019user}, research shows that they also can lead to problematic overreliance on AI advice~\citep{bansal2021whole,poursabzi-sangdeh-etal-2021-interpretability,zhang2020confidence,pafla2024LLMerrors}.
For instance, example-based and attribution-based explanations seem to have little effect on people's ability to correctly assess the veracity of claims~\citep{linder2021level}, but can lead to reliance on AI system advice even where it can be demonstrably wrong ~\citep{lim2023xai}.
In this study, we designed a task in which participants are presented with the input (claim and evidence) and outputs of an AI system (predicted verdict, model uncertainty, AI explanation) and asked to decide whether to use the AI system's prediction or not, and to identify which information they used to make their decision.

\section{Method}
\begin{figure*}[!t]
    \centering
    \includegraphics[width=1\linewidth]{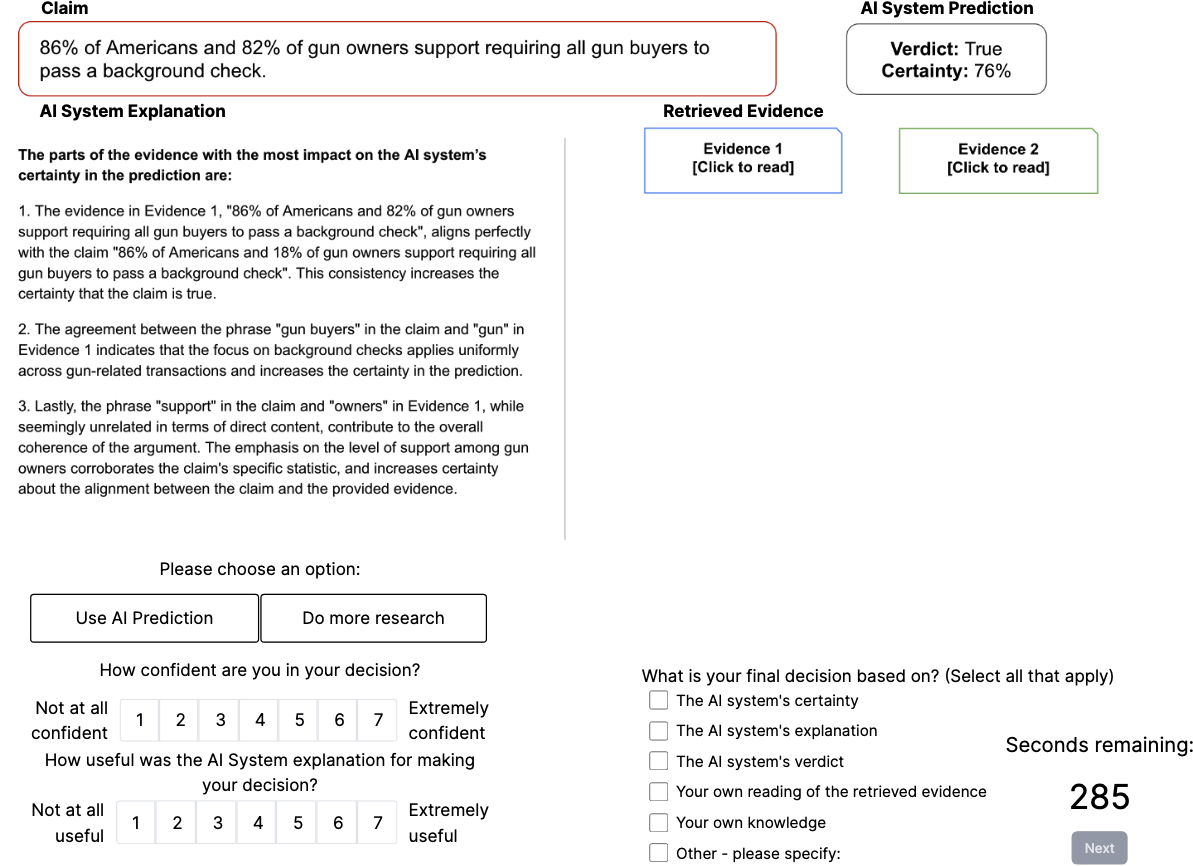}
    \caption{Example of task interface in Uncertainty Explanation condition}
    \label{fig:task_interface} 
\end{figure*}

We designed a controlled experiment to understand the role of evidence and  explanations in how people make decisions in an AI-supported fact-checking task.
Participants were presented with the outputs of an AI system designed to support fact-checking by predicting the veracity of claims (see example interface in Fig. \ref{fig:task_interface}). The system provided a numerical estimate of model certainty about the prediction, an explanation, and two evidence documents on which the prediction and explanation were based. 
The study was designed to assess the impact of explanation type on the fact-checking decision process, where participants could accept or decline the AI prediction. Each participant evaluated eight carefully selected combinations of claims and relevant evidence documents retrieved from online sources using one of three types of explanations (Uncertainty explanations, Verdict explanations, and No explanation) to support the process. We varied model certainty and model correctness counterbalanced within subjects. The study was approved by our institution's Research Ethics Committee.

\subsection{Experimental setup and design}


\subsubsection{\textcolor{black}{Claims and Evidence}}
The claims and evidence documents that formed the basis for the materials were drawn from the DRUID dataset~\citep{hagstrom-etal-2025-reality}.
\textcolor{black}{
We selected this dataset for its diversity, comprising claims from seven professional fact-checking websites, such as Politifact (\url{https://www.politifact.com/}), and FactCheckNI (\url{https://factcheckni.org/}) and quality-checked evidence documents retrieved from online sources such as news articles.
We selected claims covering science, health and politics (see Table \ref{tab:claims}) that had been fact-checked by organisations based in the United States, United Kingdom, Sri Lanka and France. The evidence documents (see App. \ref{app:materials}) consisted of online articles (see \cite{hagstrom-etal-2025-reality} for full dataset).
}

%


\subsubsection{AI Output: Verdicts, Certainty and Explanations}
For each claim and pair of evidence documents, we used Qwen2.5-14B-Instruct\footnote{\url{https://huggingface.co/Qwen/Qwen2.5-14B-Instruct}}, an open-weights, instruction-tuned language model with 14B parameters, to generate a  verdict and explanation for the given claim. We chose this model as a representative mid-sized open model that can be run locally (supporting reproducibility without relying on proprietary APIs), and that provides access to token-level logits required for our entropy-based uncertainty estimate. 
The model’s uncertainty was quantified by the entropy of its logits for generating the verdict, i.e., True or False (see App. \ref{app:uncertainty_est} for  estimation method).

We compared three types of explanations (see Table \ref{tab:example_expls} for examples). \textcolor{black}{Given the role of explanations and expressions of uncertainty in reliance on AI decisions, our aim was to compare the effects of \textit{Uncertainty explanations}, \textit{Verdict-focused explanations}, and a \textit{No explanation} control.}
The \textit{Uncertainty explanations} provided reasons for the model certainty score. For example, conflicting evidence provided to the model may decrease the model's certainty that the predicted verdict is correct, while corroborating evidence may increase the model's certainty.
To generate the uncertainty explanations, we followed the CLUE method~\citep{sun-etal-2026-explaining}. Given a claim and some evidence documents as input, this method first identifies key conflicting and concordant spans of text between the claim and evidence pieces that influence model certainty, and then instructs the model to describe how these conflicting or concordant span interactions, such as disagreements between two pieces of evidence, affect the certainty of the verdict prediction (see App. \ref{app:uncertainty_explanation_generation_prompt} for the instruction prompt).
The \textit{Verdict-focused explanations} were generated using a few-shot prompting method in which the model was instructed to explain why the verdict is predicted by referencing or summarising the provided evidence, and provided with three complete examples within the prompt (see App. \ref{app:generic_explanation_generation_prompt}). 
The `explanations' in the \textit{No explanation} condition were designed as a control condition and merely restated the AI system's verdict prediction and certainty score in text.

We assessed all generated explanations for quality, making minor edits to 
ensure consistency in structure (e.g., line-breaks and 2-3 numbered paragraphs for readability) and equivalent length in the explanations for the same claim. Where explanations did not explicitly explain the extracted text spans and the resulting AI system certainty percentage as instructed, we made minor edits to include this information to ensure that all explanations met our intended aims. We ensured that all Uncertainty explanations began with ``The parts of the evidence with the most impact on the AI system’s certainty in the prediction are:'' and that all Verdict explanations began with ``The parts of the evidence with the most impact on the AI system’s predicted verdict are:'' to ensure the target of each explanation type was clear.

\subsubsection{Interface design}
Following previous studies examining AI-supported fact-checking (e.g., \cite{schmitt2024explhuman,miranda2019automated, linder2021level}), we designed the interface to mimic a news dashboard.
Participants were presented with a fact-checking claim, the AI system's predicted verdict, certainty, and an explanation for the AI system's output. 
Participants could access the evidence by clicking on the evidence boxes in the interface to reveal them.
We chose to present the evidence within the experiment, rather than as links to external sources (as in e.g., \cite{kim2024llmsproducefaithfulexplanations,kim2025fostering}) for several reasons: 
(i) to mitigate the risk that evidence could be edited or removed, given that the AI outputs were pre-generated,
(ii) to maintain participants' focus on the experimental task, rather than becoming distracted by side-research, and hence (iii) to control the content, relevance, and length of information participants were provided, 
(iv) to allow us to measure the extent to which participants chose to view or hide the evidence,
and (v) to study the approach of providing evidence documents `in-window' used by a number of LLM-based chatbots designed specifically for knowledge intensive information-seeking tasks such as Consensus AI (\url{https://consensus.app/}) and Google Gemini's `Double-check response' feature (\url{https://gemini.google.com/app}).



\begin{table}[!htbp]
\centering
\scriptsize
\linespread{1.05}\selectfont
\setlength{\tabcolsep}{4pt}
\renewcommand{\arraystretch}{1.08}
\setlength{\extrarowheight}{1pt}

\begin{tabular}{|p{0.18\textwidth}|p{0.74\textwidth}|}

\hline
\multicolumn{2}{|p{0.94\textwidth}|}{\textbf{Claim:} A windmill could spin until it falls apart and never generate as much energy as was invested in building it.} \\ \hline
\multicolumn{2}{|p{0.94\textwidth}|}{\textcolor{black}{\textbf{Evidence 1:} No, a professor didn’t say a windmill will ‘never generate’ the energy invested in building it [...] Now, we have a meme roaming social media that claims Canadian author and professor Thomas Homer-Dixon said that
windmills will “never” generate as much energy as was expended to build the structures. [...] The meme does get a part
of the quote right, but it strategically leaves out large swaths of information in order to give people the impression that
all windmills never expend the amount of energy that was invested in it, which is not accurate. The meme’s omitted
section explains that, while poorly placed windmills may never generate enough energy payback, a good wind site
could generate it in three years or less.}} \\ \hline
\multicolumn{2}{|p{0.94\textwidth}|}{\textcolor{black}{\textbf{Evidence 2:} Factually Inaccurate: Analyzing the energy used in the production of wind turbines does not show that it’s
greater than the electricity produced over the turbine’s working lifetime.}} \\ \hline

\multicolumn{2}{|p{0.94\textwidth}|}{\textbf{Predicted Verdict:} False \quad \textbf{Certainty:} 31\%} \\ \hline
\multicolumn{2}{|p{0.94\textwidth}|}{\textbf{Uncertainty Explanation:} 
The parts of the evidence with the most impact on the AI system’s certainty in the prediction are:
\newline
1. The evidence in Evidence 1, ``No, a professor didn’t say a windmill will `never generate' the energy invested in building it'', directly contradicts the claim’s assertion that windmills will never generate sufficient energy. This contradiction increases the AI system’s certainty that the claim is False.
\newline
2. The evidence in Evidence 1, ``will ‘never generate’ the energy invested'', aligns with the part of the claim ``spin until it falls apart and never generate as much energy as was invested'', reducing the AI system’s certainty that the claim is False.
\newline
3. The evidence in Evidence 1, ``expended to'', aligns with the phrase ``invested in building'' in the claim, indicating agreement on the energy expenditure aspect, reducing the certainty that the claim is False.} \\ \hline
\multicolumn{2}{|p{0.94\textwidth}|}{\textbf{Verdict Explanation:} 
The parts of the evidence with the most impact on the AI system’s predicted verdict are:
\newline
1. Evidence 1 explains that the claim comes from a misrepresented quote. The meme states that windmills ``will never generate as much energy as was expended to build the structures,'' but the original quote clarifies that ``while poorly placed windmills may never generate enough energy payback, a good wind site could generate it in three years or less.'' This shows the claim exaggerates and omits important context.
\newline
2. The same evidence highlights that well-sited windmills can generate enough energy to offset their construction cost quickly, making the claim that all windmills ``never generate'' enough energy inaccurate. The phrase ``a good wind site could generate it in three years or less'' directly contradicts the claim.
\newline
3. Evidence 2 reinforces this, stating that ``analyzing the energy used in the production of wind turbines does not show that it's greater than the electricity produced over the turbine's working lifetime.'' This confirms that, over time, turbines produce more energy than was invested in building them.} \\ \hline

\multicolumn{2}{|p{0.94\textwidth}|}{\textbf{No Explanation:} 
The AI system has judged this claim to be False, with 31\% certainty.} \\ \hline
\end{tabular}
\caption{Example of claim, evidence,  corresponding AI prediction, certainty, and explanations}
\label{tab:example_expls}
\end{table}

\subsection{Pre-testing experimental setup through think-aloud study}
We pre-tested our initial experimental setup with a think-aloud study to refine our design and materials and to observe in-situ how people reacted to the different types of explanations. We recruited five participants with diverse expertise
from our institution (T1-T3) and wider community (T4-T5) 
(see 
App. \ref{app:thinkaloud_ppts} for demographics). 
Sessions lasted approximately 1 hour, and were audio recorded and transcribed with participants' permission (see App. \ref{app:thinkaloud_protocol}). Each participant saw six claims, two with each type of explanation, in random order. We conducted semi-structured interviews upon task completion. 
We made several key design changes based on the observations and insights gathered during these sessions.


\subsubsection{Task Design.} Our initial study design asked participants to make a decision about whether the claim was true or false and gave no time limit.
After three think-aloud sessions, we observed that people often focused on reading the evidence, to the exclusion of the AI system decisions and explanations. Through participant debriefing, we realised this happened because the decision about whether the claim was true or false did not necessarily require people to engage with the AI system, as the evidence was readily available and people could spend as much time as desired reading the evidence documents to form their decision.
We made two changes to the task to encourage participants to engage with the AI system prediction, certainty and explanations.
First, we altered the main decision participants made, from ``True'' vs ``False'' to ``Use AI Prediction'' vs ``Do more research''. In the task instructions (App. \ref{app:instructs}), participants were told to imagine that their goal was to release assessments about whether a claim is true or false, and that their task was to decide whether to rely on the AI prediction, or conduct more research before releasing the assessment. The purpose of this change was to encourage participants to evaluate the AI output, rather than merely evaluating the claim and the evidence.
Second, we introduced a time limit of 5 minutes per instance, to create a more realistic situation for decision-making, engender a greater sense of urgency, and discourage any side-research. When the time limit elapsed, evidence and explanations disappeared from the screen, and participants were prompted to make a final decision and move on to the next instance.
After introducing these changes, we conducted think-aloud sessions with two additional participants, and observed that these changes shifted participants' focus more towards the AI system. 

\subsubsection{Claims.} The think-aloud pre-testing also helped to identify potential issues with the claims presented to participants. After each think-aloud session, we reviewed the comments participants made about the claims, and replaced claims that were too vague, too familiar, or too difficult to understand (e.g., ``I don't like the claim [...] it's very vague and I feel like it should just throw it out'' -- T1).

\subsubsection{Evidence sources.} Participants frequently commented on the importance of knowing the identity of evidence sources: ``including the source of the evidence... It will allow me to know how trustworthy the evidence is, who made the claim'' (T2). While we wanted to avoid including specific names of sources to control for potential biases of participants ~\citep{andrews2025should,liao2014sourceexpertise}, we modified the task instructions to advise that all evidence provided should be considered reasonably reliable.

\begin{table}[!t]
\centering
\small
\linespread{0.94}\selectfont
\renewcommand{\arraystretch}{0.88}

\setlength{\aboverulesep}{1.5pt}
\setlength{\belowrulesep}{1.5pt}

\begin{tabularx}{\textwidth}{
    @{}
    >{\raggedright\arraybackslash}p{0.13\textwidth}
    >{\raggedright\arraybackslash}p{0.27\textwidth}
    >{\raggedright\arraybackslash}X
    @{}
}
\toprule
\textbf{Task} & \textbf{Measurement} & \textbf{Response} \\
\midrule
& AI advice use 
& Use AI prediction vs. Do more research \\

& Information usage 
& Multiple choice: The AI system's verdict / AI certainty / AI explanation / Evidence / Own knowledge / Other \\

& Decision confidence 
& 7-point scale: 1=``Not at all'' to 7=``Extremely'' \\

& Explanation usefulness 
& 7-point scale: 1=``Not at all'' to 7=``Extremely'' \\

\midrule

\textbf{Post-task} 
& Information usage (qualitative) 
& How did you use the information provided by the AI system to make your decisions? \\

& Explanation utility (qualitative)  
& Please explain why the AI system explanations were/weren't useful for determining how reliable the system was. \\

& Explanation helpfulness (App. \ref{app:helpfulness}) 
& 5-point scale: 1=``Not at all'' to 5=``Extremely'' \\

& Subjective reliance (Trust Belief \& Intention~\citep{mcknight2002impact}) (App. \ref{app:trust_scales}) 
& 5-point scale: 1=``Strongly disagree'' to 5=``Strongly agree'' \\

& Claim familiarity (App. \ref{app:claimfam}) 
& 4 options: No knowledge / Limited knowledge / Some knowledge / Full knowledge \\

\bottomrule
\end{tabularx}

\caption{Overview of tasks, measurements, and response formats.}
\label{tab:task_measurements}
\end{table}

\subsection{Controlled Experiment}

\subsubsection{Design and Measurements}
\begin{figure}[!t]
    \centering
    \includegraphics[width=0.9\linewidth]{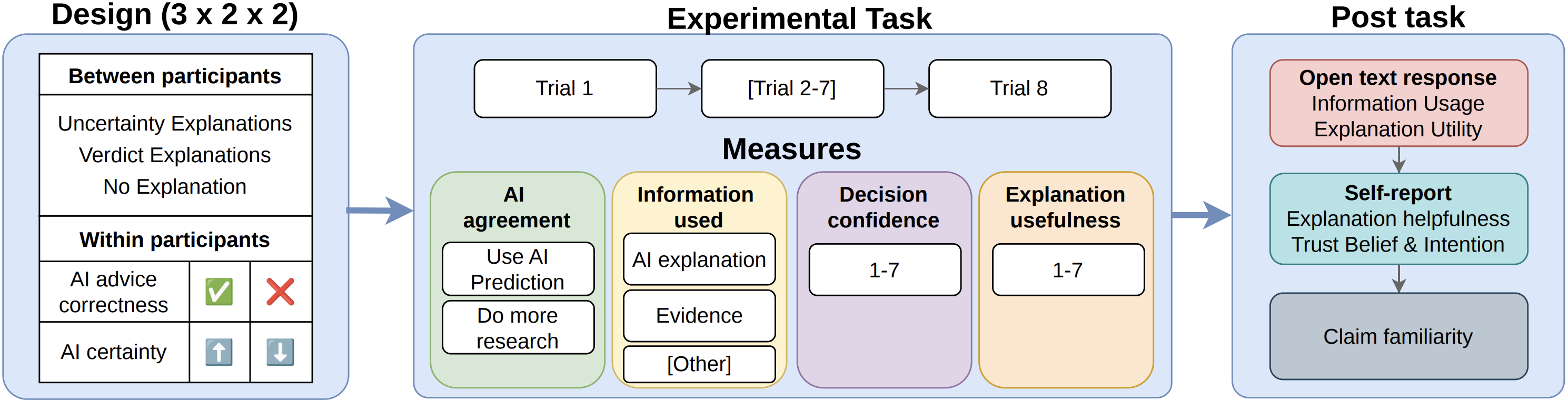}
    \caption{Overview of study design and flow}
    \label{fig:expt_flow}
\end{figure}

The final experiment had a 3 (Explanation type: Uncertainty explanation vs Verdict explanation vs No explanation) x 2 (AI advice: correct vs incorrect) x 2 (AI certainty: high vs low) design to investigate the effects of explanations and evidence across different levels of AI correctness and AI certainty \textcolor{black}{(see Fig. \ref{fig:expt_flow} for an overview)}.
Explanation type was a between-participants factor, while AI correctness and AI uncertainty were within-participant factors.
In the main task (see Figure \ref{fig:task_interface}), participants were presented with a fact-checking claim, the AI system's predicted verdict, certainty, and an explanation for the AI system's output. Participants could access the evidence by clicking on the evidence boxes in the interface to reveal them. For each claim, participants were asked to decide whether to (a) use the AI system's prediction, or (b) do more research. If a participant chose to `do more research', a follow-up question asked them to explain their decision given the options `The AI is incorrect', `More information is needed to make a decision', or `Other'.
For each decision, participants were asked to indicate what sources of information (i.e., AI verdict, AI certainty, AI explanation, evidence, their own knowledge, and other) their final decision was based on.
They were also asked to indicate (i) how confident they were in their own decision, and (ii) how useful the explanation was to their decision on a scale of 1-7. 
There was a time-limit of 5 minutes for each claim. If this elapsed, the evidence and AI explanations disappeared and participants were forced to make a decision and move on.
A post-task questionnaire collected subjective evaluations and free-text responses, see Table \ref{tab:task_measurements} and App. \ref{app:subjective_scales} for details.

\subsubsection{Participants}
A power analysis~\citep{faul2009statistical} indicated that 207 participants were required for 90\% power for a medium effect (alpha<.05) for two-tailed tests.
Participants (N=208) were recruited using Prolific (\url{https://www.prolific.com/}), and randomly assigned to one of three conditions: Uncertainty explanation (n=70), Verdict-based explanation (n=69) and No explanation (n=69). Participant demographics are reported in Table \ref{tab:education} (App. \ref{app:ppts}).
Participants were pre-screened to be native English speakers from Ireland, the United Kingdom, the United States, Australia, Canada, and New Zealand, who had not participated in previous related studies. 
Twenty-two participants who failed more than one attention or memory check were excluded prior to data analysis.

\subsubsection{Materials}
Eight unique claims (and corresponding evidence documents) were selected from the DRUID ~\citep{hagstrom-etal-2025-reality} dataset to balance (i) true and false verdicts, (ii) correct and incorrect AI predictions, (iii) high (>50\%) and low (<50\%) AI certainty (see Table \ref{tab:claims}). Explanations were consistent with the AI prediction, regardless of whether it was correct or incorrect.
For all claims, evidence and explanations see App. \ref{app:materials}.

\begin{table}[!t]
\centering
\small
\linespread{1.03}\selectfont
\setlength{\tabcolsep}{4pt}
\renewcommand{\arraystretch}{0.9}

\begin{tabularx}{\linewidth}{
    @{}
    >{\raggedright\arraybackslash}p{0.55\linewidth}
    >{\centering\arraybackslash}X
    >{\centering\arraybackslash}X
    >{\centering\arraybackslash}X
    @{}
}
\toprule
\textbf{Claim}
& \shortstack{\textbf{AI}\\\textbf{Verdict}}
& \shortstack{\textbf{AI}\\\textbf{Advice}}
& \shortstack{\textbf{AI}\\\textbf{Certainty}} \\
\midrule

        1. 86\% of Americans and 82\% of gun owners support requiring & True & Correct & High \\
        all gun buyers to pass a background check. & & & \\
        2. A satanic-themed hotel is opening in Texas. & False & Correct & High \\
        3. A scone can equal a third of one's recommended daily calories. & True & Correct & Low \\
        4. A windmill could spin until it falls apart and never generate as & False & Correct & Low \\
        much energy as was invested in building it. & & & \\
        5. All tea bags contain harmful microplastics. & True & Incorrect & High \\
        6. Lagan Valley is one of the least wooded areas in Northern Ireland,  & False & Incorrect & High \\
        particularly with regards to ancient woodland & & & \\
        7. Cristiano Ronaldo was the only European soccer captain who did not  & True & Incorrect & Low \\
        wear a rainbow armband at Euro 2020. & & & \\
        8. The top issue for college students is the economy. & False & Incorrect & Low\\

\bottomrule
\end{tabularx}

\caption{Claims presented to participants during the experiment.}
\label{tab:claims}
\end{table}

\Needspace{8\baselineskip}
\subsubsection{Procedure}
Participants received a link to the study hosted on Gorilla (\url{https://gorilla.sc/}).
Participants read detailed instructions  and completed a practice example (see App. \ref{app:instructs}).
They then progressed through the main task, providing responses for the eight claims in random order.
They then responded to a post-task questionnaire and optionally provided demographic information.
Participants were paid \pounds 4.50 for taking part. The average completion time was 29m 54s, corresponding to a payment rate of \pounds 9 per hour.

\Needspace{8\baselineskip}
\FloatBarrier
\subsubsection{Data analysis}
Quantitative analysis was performed using R ~\citep{R2021}.
Alongside traditional quantitative analysis, we analysed themes in our qualitative data from think-aloud interviews and free-text response questions. Two authors open-coded all of the data independently, periodically comparing codes and reconciling differences until achieving full agreement. Open codes were then combined into thematic categories to aid analysis~\citep{braun2006using}. 

\section{Results}
The importance of evidence in our study was striking across quantitative and qualitative data. In contrast to prior research ~\citep{kim2025fostering}, 64\% of participants opened both evidence documents for every claim they assessed and just three (of 208) did not access either of the evidence documents. Participants reported finding evidence useful significantly more often than any other available information, regardless of condition. Evidence was also the primary topic of commentary in free-text responses, where participants explained which information they used to make decisions. Explanations were judged to be helpful tools, particularly in identifying when the AI output was less reliable. In this section, we first discuss the original focus of the study, to assess the role of different types of explanations in claim assessment in a fact-checking scenario. We then discuss the role evidence played in this process and the implications of this finding. 

\subsection{The role of explanations in claim assessment}
Our initial hypothesis was that uncertainty explanations would be more effective in supporting participants to evaluate whether to agree or disagree with the AI system \textcolor{black}{(\textbf{RQ3:H5})}. We analysed the extent to which participants used the AI system's prediction, depending on the explanation condition, correctness of the advice, and level of certainty of the system (see Table~\ref{tab:evaluation_results}).
Predictably, participants were more likely to follow the AI system's advice when it was correct vs. incorrect, F(1, 207)=208.779, p <.001, $\eta_p^2$=.50, and when AI certainty was high vs. low, F(1, 207)=55.55, p <.001, $\eta_p^2$=.21.
Yet there was no difference in \textcolor{black}{overreliance, F(2, 205)=1.186, p=.307,} or \textcolor{black}{overall} reliance based on the explanation condition, F(2, 205)=1.825, p=.164, nor did this factor interact with advice correctness, F(2, 205)=0.1914, p=.9087, or certainty, F(2, 205)=0.5771, p=.749.
We did not observe any effect of explanation condition on confidence or trust judgments, nor did prior familiarity with claims affect results (see App. \ref{app:additional}).

\begin{table}[!t]
\renewcommand{\arraystretch}{1}
\centering
\resizebox{\textwidth}{!}{%
\begin{tabular}{lcccc}
\hline
\textbf{Measure} & 
\makecell{\textbf{All} \\ \textbf{conditions}} & 
\makecell{\textbf{Uncertainty} \\ \textbf{Explanation}} & 
\makecell{\textbf{Verdict} \\ \textbf{Explanation}} & 
\makecell{\textbf{No} \\ \textbf{Explanation}} \\
\hline
\multicolumn{5}{l}{\textbf{Use of AI prediction (\%)}} \\
Overall             & 54.56 & 51.79 & 57.79 & 54.17 \\
Correct AI advice   & 73.44 & 71.43 & 76.09 & 72.83 \\
Incorrect AI advice & 35.70 & 32.14 & 39.49 & 35.51 \\
High AI certainty   & 62.62 & 57.86 & 65.58 & 64.49 \\
Low AI certainty    & 46.51 & 45.71 & 50.00 & 43.84 \\

\multicolumn{5}{l}{\textbf{Explanation usefulness (1--7)}} \\
Overall             & 4.93 & 4.98 & 5.12 & 4.70 \\
Correct AI advice   & 5.30 & 5.28 & 5.47 & 5.16 \\
Incorrect AI advice & 4.56 & 4.68 & 4.76 & 4.24 \\
High AI certainty   & 5.22 & 5.15 & 5.43 & 5.07 \\
Low AI certainty    & 4.65 & 4.81 & 4.80 & 4.33 \\
\multicolumn{5}{l}{\textbf{Explanation helpfulness (1--5)}} \\
Overall             & 3.67 & 3.80 & 3.88 & 3.32 \\
\hline
\end{tabular}
}
\caption{Mean use of AI predictions, and explanation usefulness and helpfulness judgments by participants in each condition. \textcolor{black}{The `Incorrect AI advice' row corresponds to overreliance rates.}
}
\label{tab:evaluation_results}
\end{table}

\noindent We did observe a main effect of explanation condition on how often people used the explanations \textcolor{black}{(\textbf{RQ3:H6})} (see Table \ref{tab:information_use}), F(2, 205)=24.668, p<.001, $\eta_p^2$=.19. Tukey HSD tests showed that participants in the two natural language explanation groups reported using the explanations in their decisions significantly more than the No Explanation group (Uncertainty: p<.001, d=.99, Verdict: p<.001, d=1.07). 
People judged that both Uncertainty (p=.003, d=.5) and Verdict (p=.013, d=.46) Explanations were more \textit{helpful} than No Explanation and that the Verdict Explanations were more \textit{useful} than No Explanation, p=.01, d=.34. Further, participants reported using the AI explanations more frequently when the AI system was correct (M=55.5\%) than when it was incorrect (M=48.1\%), F(1, 207)=12.125, p<.001, $\eta_p^2$=.055. No differences emerged between the two explanation groups (p>.05 for all comparisons).
There was a main effect of explanation condition on how useful people judged the explanations, F(2, 205)=4.412, p=.013, $\eta_p^2$ =.02.
Bonferroni-corrected post hoc tests indicated that the Verdict Explanation group rated the explanations as more useful than No Explanation, p=.01, d=.34. 

\textcolor{black}{\textbf{Cross-referencing explanations and evidence  to identify AI errors.}} In the text responses, participants in the No Explanation condition complained about  lack of information, although several mentioned using numerical uncertainty as a cue to be more sceptical. Participants in the explanation conditions often found explanations useful as summaries of evidence when they aligned with evidence. When explanations and evidence were misaligned, however, participants commented that this helped them understand the system's reasoning, aligning with prior research~\citep{si-etal-2024-large}: ``Sometimes by reading the explanations I could spot errors in the AI's logic.''  Qualitative data also suggested reasons for why the natural language explanations did not differ along expected dimensions. Almost all responses in the No Explanation condition mentioned using evidence to assess the claim, suggesting that having access to evidence documents compensated for  differences in explanation types. Participants in the Uncertainty condition had mixed opinions on the utility of these explanations. Some found them difficult to understand: ``on some of the choices the AI was very helpful but in others I found it very contradictory and confused,''suggesting that more research on explaining uncertainty in natural language is necessary.  


\begin{table}[h]
\centering
\small
\setlength{\tabcolsep}{3pt}
\renewcommand{\arraystretch}{1.0}

\begin{tabularx}{\linewidth}{
    @{}
    >{\raggedright\arraybackslash}p{0.21\linewidth}
    *{4}{>{\centering\arraybackslash}X}
    @{}
}
\toprule
\textbf{Measure}
& \makecell{\textbf{All}\\\textbf{conditions}}
& \makecell{\textbf{Uncertainty}\\\textbf{Explanation}}
& \makecell{\textbf{Verdict}\\\textbf{Explanation}}
& \makecell{\textbf{No}\\\textbf{Explanation}} \\
\midrule
AI Verdict     & 41.10 & 40.00 & 45.83 & 37.50 \\
AI Explanation & 51.80 & 59.46 & 61.41 & 34.42 \\
AI Certainty   & 31.00 & 27.14 & 30.98 & 34.78 \\
Evidence       & 67.80 & 64.82 & 71.01 & 67.75 \\
Own knowledge  & 20.30 & 19.64 & 15.76 & 25.54 \\
Other          &  2.88 &  3.21 &  3.80 &  1.63 \\
\bottomrule
\end{tabularx}

\caption{Mean percentages of each source of information used.}
\label{tab:information_use}
\end{table}

\subsection{The role of evidence in claim assessment}
Participants reported using the evidence most frequently (67.8\% overall), regardless of AI explanation, correctness, or certainty.
A one-way repeated measures ANOVA indicated differences in how often participants used each information source, F(5, 1035)=200.6, p<.001, $\eta_p^2$=.49.
People in the No Explanation group were more likely to open all evidence documents (83\%), compared to the explanation groups (55\% Uncertainty, p=.001, d=.62; 57\% Verdict, p=.003, d=.58)  \textcolor{black}{(\textbf{RQ2:H3})}.
This suggests that natural language explanations may have offered sufficient information to make a judgment based on the explanation alone or a single evidence document.  
Mirroring the overall trend, participants in the Verdict condition relied more on the evidence than all other information sources, and relied on the explanations second most often, more so than the remaining information sources, p<.001 for all comparisons.
Participants in the Uncertainty  condition, however, reported using the evidence and explanations equally as often, p=.241 \textcolor{black}{(\textbf{RQ2:H4})}. They relied on these two information sources significantly more than the others, p<.001 for all comparisons.
Participants in the No Explanation condition used evidence significantly more than any other information source, p <.001 for all comparisons. The `explanations' (restatements of AI verdict and certainty) were no more useful here than the AI verdict (p=.408), AI certainty (p =.911), or their own knowledge (p=.028; not significant on the Bonferroni-corrected alpha).
\textcolor{black}{We also conducted exploratory analysis of participants' task responses when they reported using evidence versus when they did not (see App. \ref{app:no_evidence}). 
}

\textcolor{black}{\textbf{People inferred evidence quality even in the absence of source identities.}} While evidence generally dominated as the most important information for evaluating claims in the qualitative data, a common criticism was that without source information people ``found it difficult to fully trust'' the evidence. In some cases, people tried to reverse engineer sources by paying attention to format, language, and, most often, the use of statistics. In fact, the use of statistics was at times perceived as more credible: ``if the evidence had actual numbers and data from respectable sources then usually I would side with the AI's verdict.'' Given that there was no source information in the evidence documents on whether the sources were ``respectable'' the credibility was clearly gained through the presence of numbers alone. This has implications for what kinds of evidence AI systems might present and suggests caution in how statistical evidence might be interpreted. 

\subsection{The role of subjective reliance in behavioural reliance on AI advice, explanations, and evidence}

We investigated the relationships between participants' responses to the trust and belief intention scale and their propensity to rely on (i) AI advice and use (ii) AI explanations and (iii) the underlying evidence in their decisions. We interpret the trust belief and intention scales~\citep{mcknight2002impact} as essentially measuring subjective reliance on AI output. 
We found a moderate correlation between people's subjective reliance on the AI system and their tendency to rely on the AI's recommendation (r=.425, p<.001), and subjective reliance predicted higher agreement with AI advice, $\beta$=0.91, SE=.135, p<.001, $R^2$=.181 \textcolor{black}{(\textbf{RQ1:H1})}. 
Participants with higher subjective reliance scores were more likely to follow AI advice.
We found that all explanation conditions exhibited a negative correlation between subjective reliance and AI agreement (Uncertainty: r=-0.35, p=0.003; Verdict: r=-0.17, p=0.176; No explanation: r=-0.35, p=0.003). 
Participants' subjective reliance on the AI system was also correlated with  tendency to use the AI explanations and underlying evidence in their decisions, although the effects were relatively small \textcolor{black}{(\textbf{RQ1:H2)}}.
Our analysis showed a small positive correlation between participant's subjective reliance judgments and the frequency with which they used the AI explanation to make their final decision (r=.26, p<.001), and reliance predicted higher usage of AI explanations, $\beta$=0.838, SE=0.214, p<.001, $R^2$=.069.
On the other hand, subjective reliance in the AI system was negatively correlated with usage of evidence (r=-.29, p<.001), with reliance predicting lower evidence use, $\beta$=-1.015, SE=0.234, p<.001, $R^2$=.084. 
In other words, use of evidence was associated with lower subjective reliance in the AI system, while use of explanations was associated with higher reliance.

\textcolor{black}{\textbf{Using evidence and explanations to assess the reliability of AI advice.}}
Qualitative findings corroborated the association between low subjective reliance in the AI system and use of evidence: ``I didn't trust it so had to check what it claimed''.
Participants also explained how trust in the AI was shaped by how accurately the outcome reflected the evidence: ``I considered whether the AI's decision was based off the evidence, if their decision contradicted the evidence, this makes me feel that the AI was less trustworthy as it had not acknowledged the evidence properly''. 
People in the Explanation conditions also discussed using logical fallacies in explanations as cues to how reliable the system's decision was. Some participants in the Uncertainty condition reported using the sources of uncertainty to ``judge when to trust the system less''. 
This suggests that explanations can play an important role in helping to determine whether the system has interpreted available evidence correctly.

\section{Discussion}
Our results show that evidence is a key component of how people make decisions when assessing AI system output, prompting important considerations for developing AI decision-support tools. We observed that having access to evidence enabled participants to evaluate AI content successfully regardless of the  explanation provided, although explanations were appreciated. We also show that when people have access to evidence, explanations do not necessarily lead to overreliance (using incorrect AI advice) as previous studies have found \cite{he2025conversationalXAI,kim2025fostering}. Qualitative data suggest that access to evidence can enable more critical assessment of AI output. 


\subsection{Availability is key for evidence engagement}
Previous work has found that when AI systems provide people with links to external sources, relatively few (10-28\%) check them ~\citep{kim2024LLMuncertainty,kim2025fostering}, and it is unclear to what extent people meaningfully engage with the content of sources. 
In contrast, approximately two-thirds of participants in this study opened all of the available evidence documents and almost all opened at least one for each claim.
Several factors may account for this difference.
First, the evidence was embedded directly within the task interface. Participants could open the evidence alongside the claim, AI prediction, and explanation, instead of following a link to an external webpage (e.g., \cite{kim2025fostering}), which likely reduced friction and effort associated with accessing evidence.
Second, the explanations explicitly referenced the evidence, which may have encouraged participants to use it to verify the AI system's interpretation, or made it easier to identify key parts of the evidence more efficiently. However, we observed that people reported using the evidence in their decisions regardless of whether explanations were provided.
It may also be the case that the fact-checking context may have been perceived as more high-stakes than the question-answering setups in previous studies ~\citep{kim2024LLMuncertainty,kim2025fostering}, motivating participants to consider the information more carefully, although we note that in prior studies click rates were extremely low ($\sim$7\%) even for potentially high-stakes health stimuli ~\citep{kim2024LLMuncertainty}.
These differences indicate that more research is needed to understand what kind of evidence is needed, in what context, and what are the best approaches to make it available to people using AI systems. 

Participants were sensitive to inconsistencies between the verdict and the explanations, but even where the explanation and the verdict aligned, inconsistencies between explanation and evidence alerted them to critically consider AI system output. \textcolor{black}{We also found that people relied less on the AI system's advice when they used evidence than when they did not, and were more confident in their decisions when using evidence than when they did not.} As suggested by~\citet{kim2025fostering}, the provision of sources themselves may lead to more deliberative reasoning, but attention must be paid to how sources are presented.

\subsection{Evidence-focused explanations may mitigate overreliance}
On the one hand, our findings suggest a more optimistic state of affairs about people's reliance on and critical engagement with AI  explanations.
Prior studies in AI-supported fact-checking have highlighted that natural language explanations in the form of high-level abstractive summaries, designed to simulate how humans reason about claim veracity, can persuade people to incorrect conclusions~\citep{schmitt2024explhuman,si-etal-2024-large}.
However, providing natural language explanations, which extracted relevant parts of the evidence and explained how they contributed to the AI system's predicted verdict or certainty, did not lead participants to over-rely on incorrect advice in comparison to the No Explanation group.
This may indicate that explanations that focus on evidence facilitate people to identify inconsistencies, flawed argumentation or other cues that the explanation may be misleading. 
However, these explanations also have  potential for misuse: cherry-picking evidence or otherwise using evidence out of context to falsely push a certain narrative could be highly persuasive. Although our qualitative data suggested that our participants were relatively alert to when this occurred, this may be less feasible when evidence documents are longer or more complex.
\textcolor{black}{Furthermore, we only tested explanations generated using one type of model. Although the Verdict Explanations were prototypical justifications like those produced by many approaches (e.g.~\citet{zeng-gao-2024-justilm,deverna_fact-checking_2024}), future, more powerful models may have capability to produce more persuasive, misleading explanations. Our findings suggest that providing evidence appears to mitigate this risk to some degree, as users can verify if information is taken out of context, however} future work in HCI is needed to improve systems and explanations that allow people to detect when plausible-sounding explanations provide misleading reasoning.
Further research should examine evidence and explanation evaluation in more naturalistic, less structured task contexts and with domain experts who may interact differently with explanations than crowdworkers ~\citep{schmitt2024explhuman,szymanski2021visual}.


\subsection{Design implications for information-seeking and decision-support systems}
The role of evidence in AI-assisted decision-making has received relatively little attention in the literature to date. There is some theoretical research on the use of evidence in explanations, such as the evaluative AI framework ~\citep{miller2023explainable}, which builds on an earlier interpretability approach that utilised the information theory concept of weight of evidence~\citep{melis2021human}. However, few studies have considered the role of evidence empirically or developed methods that use it along with AI predictions and advice. This indicates a worrying research gap, especially as many commercial LLM chatbots already provide people with sources and evidence documents alongside AI output; some automatically, while others requiring explicit inquiry. As a result, there is little consensus on best practices. When current LLM implementations do give sources, LLMs struggle with reliably producing factual information ~\citep{augenstein2024factualityLLMs}, and often hallucinate references to external sources ~\citep{alkaissi2023artificial}, highlighting key gaps for further development of NLP and information retrieval tools for AI-assisted fact-checking and other complex decision-making tasks.  

Research in natural language explanations has largely focused on the production of fluent explanations~\citep{atanasova-etal-2020-generating-fact,kotonya2020fcpublichealth} and how users evaluate them~\citep{schlichtkrull-etal-2023-intended}.
Our findings call for NLP research to consider the role of evidence in explaining AI output and how to make it available effectively, alongside efforts to produce better explanations. Concurrently, HCI research is needed to gain fine-grained understanding of how evidence should be presented, and how evidence is evaluated and used by people in different AI-supported decision-making contexts.

\color{black}
\subsection{Limitations and future work}

The scope and findings of this study are subject to several additional limitations.
Firstly, with the aim of carefully controlling the information available to participants to minimise potential compounds, we designed a fact-checking task using stimuli from a single (though diverse) dataset, consisting of static output and requiring a binary decision, which did not allow participants to engage in open-ended or multi-turn interactions.
We selected this approach following similar studies that examine information-seeking and fact-checking (e.g., \cite{schmitt2024explhuman,kim2025fostering}). However, studying how people utilise explanations and evidence in a more naturalistic task setting is an important direction for future work in this area.
Secondly, we defined `high' and `low' certainty according to a 50\% threshold. In practice, what people consider as high or low certainty is subjective and dependent on individual, cultural, and  context-specific factors such as the task or stakes associated with a decision ~\citep{wallsten1988understanding,phillips1977cultural}.
Nevertheless, in our study participants relied on evidence more than any other information source, irrespective of model certainty, suggesting that the findings apply to a broad range of uncertainty scores.
Finally, our experiment only examined the effects of providing natural language explanations and evidence in the context of a fact-checking task, an area which has been underexplored in prior work. Although fact-checking is a prototypical information-seeking task, similar to other question-answering tasks in which the user must make a binary judgment (e.g., \cite{kim2024LLMuncertainty}), future research should also examine how people use evidence in a broad range of tasks and diverse formats, especially where evidence and explanations may look quite different, such as tabular data features or image pixels.

\color{black}

\section{Conclusion}
We conducted a mixed methods experiment examining how evidence and explanations shape people's use of an LLM-based AI system for fact-checking.
We found that the evidence underpinning the AI output was the most important consideration for participants, regardless of whether explanatory information was provided.
Quantitative and qualitative data indicated that people found explanations to be useful indicators of where the AI system may be less reliable.
Our results highlight the central importance of evidence in AI-assisted information-seeking and decision support.
They also show how presenting evidence alongside AI outputs can substantially increase engagement with sources compared to current conventions of linking to external information.
Together, these findings suggest promising directions for developing best practices for mitigating overreliance through accessible evidence and evidence-focused explanations.

\section{Acknowledgements}
\noindent $\begin{array}{l}\includegraphics[width=1cm]{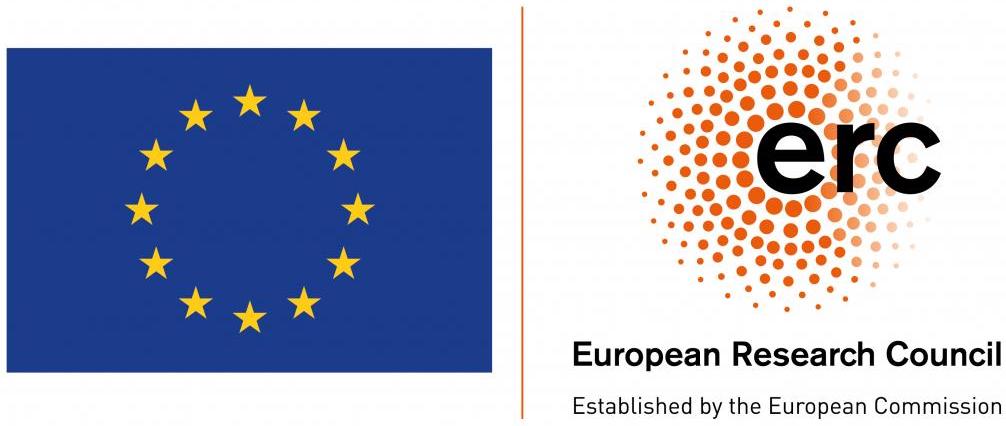} \end{array}$ This research was co-funded by the European Union (ERC, ExplainYourself, 101077481), and supported by the Pioneer Centre for AI, DNRF grant number P1. Views and opinions expressed are those of the authors only and do not necessarily reflect those of the European Union or the European Research Council. Neither the European Union nor the granting authority can be held responsible for them. The authors wish to thank the participants who generously volunteered to participate in pilot studies, without whom this research would not have been possible.


\section{Appendix}

\subsection{Generative AI usage statement}
\label{app:genAI_statement}
Large Language Models (ChatGPT-5, Claude Sonnet 4) were used to assist with programming the quantitative statistical analyses, as well as with formatting the tables that appear in the paper.

\subsection{Think-Aloud Study Participants}
\label{app:thinkaloud_ppts}

\begin{table}[!htbp]
    \centering
    \begin{tabular}{llll}
    \hline
      \textbf{ID}  &  \textbf{Age} & \textbf{Gender} & \textbf{Background} \\ \hline
      TI  &  24 & Female & Physics \\
      T2  &  58 & Female & International Relations \\
      T3  &  33 & Male & Computer Science \\
      T4  &  66 & Female & Translation \\
      T5  &  28 & Male & Sustainability \\
      \hline
    \end{tabular}
    \caption{Demographics of participants in think-aloud study}
    \label{tab:thinkaloudppts}
\end{table}

\subsection{Think-Aloud Study Protocol}
\label{app:thinkaloud_protocol}

\subsubsection{Background knowledge}
\begin{itemize}
    \item How familiar are you with artificial intelligence or AI? Have you used any AI systems or LLMs like ChatGPT?
    \item How familiar are you with fact-checking?
\end{itemize}

\subsubsection{Introduction to think-aloud study}

\begin{itemize}
    \item In this experiment, you’ll be testing an AI system designed to help with fact-checking. You’ll see some examples of claims that the AI system has been given, and its prediction about whether the claim is true or false. 
    \item You’ll see some more detailed instructions once you begin the study, and we’d like you to describe what you’re thinking as you read through the information you see, for example if something is unclear or confusing.
    \item Then, during the main study, as you’re evaluating the AI system and making decisions about the claims, we’d like you to also describe your impressions and thought processes, and talk through your decisions as you make them. At some points, I may ask you to explain a little more about why you did something.
    \item Then, at the end of the study I’ll ask you some questions about how you found the study overall.
    \end{itemize}

\subsubsection{Think-aloud prompts (to be asked if participant doesn’t verbalise their thoughts)
}

\begin{itemize}
    \item What was your first impression when you saw this screen?
    \item Why did you check [the evidence/explanation/AI certainty]?
    \item How did you decide whether the claim was true or false? What did you base your decision on?
    \item Why were you confident (or not) in your decision?
    \item Why did you find the explanation (not) useful? What was (not) useful about it?
    \item Did the AI’s certainty score impact your decision in any way?
\end{itemize}

\subsubsection{Post-study interview}

\begin{itemize}
    \item How did you find the task overall? Was anything unclear or confusing about what you were asked to do?
    \item Did you find anything difficult about the task?
    \item You saw different styles of explanation in the study. Were they helpful? Why/why not?
\end{itemize}

\subsection{Subjective Evaluation Scales} \label{app:subjective_scales}

\subsubsection{Explanation Helpfulness} \label{app:helpfulness}
To what extent were the AI system explanations helpful in determining how reliable the system was?

\noindent(1 = Not at all helpful,
2 = Somewhat helpful, 
3 = Neither, 
4 = Somewhat helpful, 
5 = Extremely helpful)

\subsubsection{Subjective reliance: Trust Belief and Intention Scales~\cite{mcknight2002impact}} \label{app:trust_scales}
On this page you are asked to provide your opinion on the AI system you used in the experiment.
Please select one answer for each statement
(1 = Strongly disagree,
2 = Somewhat disagree, 
3 = Neither, 
4 = Somewhat agree, 
5 = Strongly agree)

\noindent*Reverse-scored items

\begin{enumerate}
    \item The AI system is competent and effective in fact-checking claims
    \item Overall, the AI system is a capable and proficient information provider
    \item I would characterise the AI system as honest
    \item The AI system is NOT truthful in providing information to me*
    \item I believe that the AI system was developed to act in my (or the users') best interest
    \item The AI system was developed with good intentions to do its best to help me (or the users)
    \item When an important claim arises, I would NOT feel comfortable depending on the information provided by the AI system*
    \item I can always rely on the AI system to find information about claims
    \item I would feel comfortable acting on the information given to me by the AI system
    \item I would not hesitate to use the information the AI system supplied to me
\end{enumerate}

\subsubsection{Claim familiarity}\label{app:claimfam}

In the study, you saw 8 different real life claims, some of which you may have come across some of these claims before in your day-to-day life.

\noindent Please indicate whether you had read or come across any of these claims **before this experiment**, and if so, how much you knew about them.

\noindent\textbf{No knowledge} = I had never come across this claim before\\
\textbf{Limited knowledge} = I had not come across this specific claim before but I had previously come across something similar \\
\textbf{Some knowledge} = I had come across this claim before but I could not be sure that I knew whether it was true/false \\
\textbf{Full knowledge} = I had come across this claim and was confident that I knew it was true/false

\subsection{Main Study Participants} \label{app:ppts}

\begin{table}[h!]
    \centering
    \small
    \setlength{\tabcolsep}{4pt}
    \renewcommand{\arraystretch}{1.05}

    \begin{tabularx}{0.95\linewidth}{
        @{}
        l|
        >{\raggedright\arraybackslash}X c|
        >{\raggedright\arraybackslash}X c
        @{}
    }
    \hline
    \textbf{Age} & \textbf{Gender} & \textbf{\%} & \textbf{Education level} & \textbf{\%} \\ \hline
    M = 41.44  & Woman & 45.67 &   Doctorate degree & 5.28  \\
    SD = 13.34 & Man & 51.44  & Master's degree & 13.94  \\
     & Non-binary & 1.92 & Bachelor's degree & 35.58 \\
     & Prefer not to answer & .96 & Some college/university & 24.52  \\
    & & &    High school graduate  & 17.79 \\
    & & &    Less than high school & 1.92 \\
    & & &    Prefer not to answer  & 0.96 \\ \hline
    \end{tabularx}

    \caption{Participant demographic details.}
    \label{tab:education}
\end{table}

\subsection{Additional analyses}\label{app:additional}
\textcolor{black}{\subsubsection{Participants who did not use evidence} \label{app:no_evidence}
We analysed participants’ task responses when they reported using evidence versus when they did not.
For trials in which people used evidence, agreement with AI advice was significantly lower (50\%) than when they did not (64\%), $\beta$ = -0.12, SE = 0.02, t(1130.84) = -4.93, p<.001. People reported that the AI explanation was less useful when they chose to use evidence (5.36) than when they did not (4.73), $\beta$ = -0.36, SE = 0.09, t(1648.93) = -4.04, p<.001. When people used the evidence, they also self-reported higher confidence in their decision (5.82) than when they did not (5.62), ($\beta$ = 0.14, SE = 0.07, t(1623.57) = 2.12, p=.034).
Finally, for trials in which participants did not report using evidence, they tended to rely most frequently on the AI explanation on average (59\%), followed by AI verdict (48\%), AI certainty (40\%), their own knowledge (27\%) and other (3\%). This suggests that in these cases where people did not use evidence, many people viewed the AI output on its own as sufficient to help to make a decision.}

\subsubsection{Claim familiarity}
We collected data on how familiar participants were with the claims prior to the experiment, and on participants' confidence and subjective reliance judgements during the experiment. These results did not appear to impact the main findings of the study, but we report them here for completeness and transparency.

On average, people were unfamiliar with the claims prior to the study 58\% of the time, had some knowledge 14\% of the time, had limited knowledge 24.94\% of the time, and had full knowledge of the claims 2.5\% of the time. 
People relied more on their own knowledge when they had some knowledge of the claim than when they did not, t(384)=6.87, p<.001, otherwise, no differences emerged in how people used the information available to them (p>.05 for all comparisons).
Verdict explanations were rated as more useful than no explanations when people had no familiarity with the claims, $\chi^2$(2)= 8.22, p=.016.
Otherwise, claim familiarity did not affect how people judged the explanations, or how much they agreed with the AI system (p>.05 for all comparisons, see Table \ref{tab:familiarity_agreement_conf_use}).

\begin{table}[!htbp]
\centering
\small
\setlength{\tabcolsep}{4pt}
\renewcommand{\arraystretch}{1.05}

\begin{adjustbox}{max width=\linewidth}
\begin{tabular}{@{}llcccc@{}}
\toprule
\textbf{Condition}
& \textbf{Familiarity}
& \makecell[c]{\textbf{Number of}\\[-1pt]\textbf{responses}}
& \makecell[c]{\textbf{Agreement}\\[-1pt]\textbf{(\%)}}
& \makecell[c]{\textbf{Confidence}\\[-1pt]\textbf{(1--7)}}
& \makecell[c]{\textbf{Usefulness}\\[-1pt]\textbf{(1--7)}} \\
\midrule

No Explanation     & Any knowledge & 230 & 53.5 & 5.78 & 4.83 \\
     & No knowledge & 322 & 55.7 & 5.79 & 4.60 \\
Verdict     & Any knowledge & 206 & 55.3 & 5.62 & 5.04 \\
     & No knowledge & 338 & 59.2 & 5.73 & 5.24 \\
Uncertainty & Any knowledge & 258 & 50.8 & 5.86 & 4.96 \\
 & No knowledge & 302  & 52.6 & 5.71 & 4.99 \\
\bottomrule
\end{tabular}
\end{adjustbox}

\caption{Participants' agreement with the AI, confidence, and usefulness
ratings for claims about which they had either (a) some prior knowledge
or (b) no prior knowledge, across explanation conditions.}
\label{tab:familiarity_agreement_conf_use}
\end{table}



\subsubsection{Confidence \& Subjective reliance}

\paragraph{Confidence}
There was no impact of explanation condition on how confident people were in their decisions, F(2, 205) = 0.428, p = .652 (see Table \ref{tab:confidence_trust}).
Participants were more confident in their decisions when the AI's advice was correct than when it was incorrect, F(1, 207) = 6.122, p = .014, $\eta_p^2$ = .03, and when the AI system's certainty was high rather than low, F(1, 207) = 27.886, p < .001, $\eta_p^2$ = .12.

\begin{table}[h]
\centering
\small
\setlength{\tabcolsep}{4pt}
\renewcommand{\arraystretch}{0.95}

\begin{tabular}{@{}lccc@{}}
\hline
\textbf{Measure} & 
\makecell{\textbf{Uncertainty} \\ \textbf{Explanation}} & 
\makecell{\textbf{Verdict} \\ \textbf{Explanation}} & 
\makecell{\textbf{No} \\ \textbf{Explanation}} \\
\hline
\multicolumn{4}{l}{\textbf{Confidence (1--7)}} \\
Overall           & 5.78 & 5.70  & 5.78 \\
Correct AI advice    & 5.80 & 5.81 & 5.86 \\
Incorrect AI advice  & 5.76 & 5.59 & 5.70 \\
High AI certainty   & 5.87 & 5.88 & 5.88 \\
Low AI certainty    & 5.68 & 5.52 & 5.68  \\
\multicolumn{4}{l}{\textbf{Subjective reliance (10--50)}} \\
Overall              & 33.4 & 33.9 & 32.0 \\
\hline
\end{tabular}

\caption{Mean confidence and subjective reliance judgments by participants
in each condition.}
\label{tab:confidence_trust}
\end{table}

\paragraph{Subjective reliance.}
The explanation groups did not differ in their judgments for the subjective reliance scales, F(2, 205) = .091, p = 0.40.

\subsection{Uncertainty Estimation and Explanation Generation Method}
\label{app:method_uncertainty_and_explanation}
\subsubsection{Uncertainty Estimation} \label{app:uncertainty_est}
By curating the instruction, we guided the model to generate the verdict along with the explanations, see App. \ref{app:generic_explanation_generation_prompt} and \ref{app:uncertainty_explanation_generation_prompt} for the prompts. Based on the model's verdict prediction, we derived its uncertainty as follows.
\label{app:uncertainty_score_generation_method}
To quantify model uncertainty for a verdict label on input $X$ (one claim plus two evidence passages), we compute the predictive entropy~\citep{shannon1948,ulmer2022predictive} of the softmax distribution over logits, which requires only a single forward pass and is widely used. For candidate labels $\mathcal{Y}=\{\mathrm{True},\mathrm{False},\mathrm{NEI}\}$ with logits $\ell(y_i)$, the probability of label $y_i$ is
\begin{equation}
P(y_i \mid X) \;=\; \frac{\exp\!\big(\ell(y_i)\big)}{\sum_{j=1}^{|\mathcal{Y}|}\exp\!\big(\ell(y_j)\big)}.
\end{equation}
The uncertainty score for $X$ is the entropy of this distribution,
\begin{equation}
u(X) \;=\; -\sum_{y_i\in\mathcal{Y}} P(y_i \mid X)\,\log P(y_i \mid X),
\end{equation}

\subsubsection{Few-shot Prompt for Verdict Explanation Generation}
\label{app:generic_explanation_generation_prompt}
For the verdict explanation generation, we leveraged a three-shot prompt for instructing the model to generate the verdict along with the reasons towards this verdict, see Table \ref{tab:prompt_generic_explanation}.

\subsubsection{Few-shot Prompt for Uncertainty Explanation Generation}
\label{app:uncertainty_explanation_generation_prompt}
For the uncertainty explanation generation, we used a three-shot prompt to instruct the model to generate the verdict, along with the reasons to forward its verdict prediction certainty by referring to the key span interactions extracted in the previous step, following \cite{sun-etal-2026-explaining}, see the prompts in Table \ref{tab:prompt_uncertainty_explanation}.

\begin{table}[p]
\centering

\captionsetup{
    position=top,
    aboveskip=0pt,
    belowskip=1pt,
    skip=0pt
}

\caption{
Verdict explanation generation prompt. Adapted from Table~10 of \citet{warren2026show} for space.}
\label{tab:prompt_generic_explanation}

\begingroup
\fontsize{8.3pt}{9.2pt}\selectfont
\setlength{\parskip}{0.5pt}
\setlength{\tabcolsep}{0pt}
\setlength{\parindent}{0pt}
\setlength{\emergencystretch}{1em}
\renewcommand{\arraystretch}{1.0}

\begin{tabularx}{0.98\linewidth}{
    @{}
    >{\raggedright\arraybackslash}X
    @{}
}
\toprule

\textbf{Task Description:}
You are a helpful assistant. Here are your tasks:
1. Supply the verdict for the claim based on the relationship between
the claim and the two evidence passages.
2. Explain your reasoning for the verdict prediction by identifying
the three most influential span interactions from Claim--Evidence 1,
Claim--Evidence 2, and Evidence 1--Evidence 2, and describing how each
interaction (agree, disagree, or unrelated) affects your overall
prediction.
\par
\textbf{Return format:}
[Verdict Prediction] [Explanation]
\tabularnewline

\midrule

\textbf{SHOT 1 Input}\par
\textbf{Claim:}
Winston Churchill once stated, ``A pessimist sees the difficulty in
every opportunity; an optimist sees the opportunity in every
difficulty.''\par
\textbf{Evidence 1:}
``A pessimist,'' says Rev.\ F. W. Cole, president of the
Cardiff Free Church Council, \ldots{}\par
\textbf{Evidence 2:}
Winston Churchill's quote, ``A pessimist sees the difficulty in every
opportunity; an optimist sees the opportunity in every difficulty,''
\ldots{}
\tabularnewline

\addlinespace[1pt]

\textbf{SHOT 1 Output}\par
\textbf{Verdict Prediction:} False
\quad
\textbf{Explanation:}
The parts of the evidence with the most impact on my predicted verdict
are: The evidence in Evidence 1, ``says the Rev.\ F. W. Cole,''
conflicts with the part of the claim ``Winston Churchill once stated'';
the evidence in Evidence 2 that ``Winston Churchill's quote,
`A pessimist sees the difficulty in every opportunity \ldots{}' ''
aligns with the claim's attribution; taken together, the attribution
in Evidence 1 (``Rev.\ F. W. Cole'') versus Evidence 2
(``Winston Churchill'') is inconsistent, which weighs against the claim.
\tabularnewline

\midrule

\textbf{SHOT 2 Input}\par
\textbf{Claim:}
During the last hundred years, the temperature increased by about
0.1$^\circ$C because of carbon dioxide. The human contribution was
about 0.01$^\circ$C.\par
\textbf{Evidence 1:}
``\ldots human-caused global surface temperature
increase \ldots{} is 0.8$^\circ$C to 1.3$^\circ$C
(best estimate 1.07$^\circ$C [2.01$^\circ$F]). \ldots greenhouse gases (GHGs) contributed a warming of
1.0$^\circ$C to 2.0$^\circ$C, and other human drivers
(principally aerosols) contributed a cooling of 0.0$^\circ$C to
0.8$^\circ$C; natural (solar and volcanic) drivers changed global
surface temperature by $-0.1^\circ$C to $+0.1^\circ$C, and internal
variability changed it by $-0.2^\circ$C to $+0.2^\circ$C.''\par
\textbf{Evidence 2:}
``After using a bunch of nonsense calculations to say that
0.1$^\circ$C warming is from CO2, they say that 90\% of the change in
CO2 is caused by the oceans. \ldots{} The oceans are actually absorbing
CO2 \ldots{} their conclusions cannot be supported.''
\tabularnewline

\addlinespace[1pt]

\textbf{SHOT 2 Output}\par
\textbf{Verdict Prediction:} False
\quad
\textbf{Explanation:}
\ldots Evidence 1 reports human-caused warming near 1$^\circ$C overall
and 1.0--2.0$^\circ$C from well-mixed greenhouse gases, contradicting
the claim's ``about 0.1$^\circ$C because of carbon dioxide'' and
``about 0.01$^\circ$C'' human contribution; Evidence 2 challenges the
assertion that ``0.1$^\circ$C warming is from CO2,'' which also goes
against the claim; the segment in Evidence 1 about aerosol cooling and
natural drivers ($-0.1^\circ$C to $+0.1^\circ$C) is unrelated to
Evidence 2's point about oceans absorbing CO2.
\tabularnewline

\midrule

\textbf{SHOT 3 Input}\par
\textbf{Claim:}
Vitamin D appears to increase COVID-19 mortality rates.\par
\textbf{Evidence 1:}
If vitamin D does in fact reduce the severity of COVID-19 \ldots{} supplements would offer a relatively easy option
to decrease the impact of the pandemic.\par
\textbf{Evidence 2:}
A vitamin D deficiency that is not sufficiently treated is related to COVID-19 risk.
\tabularnewline

\addlinespace[1pt]

\textbf{SHOT 3 Output}\par
\textbf{Verdict Prediction:} False
\quad
\textbf{Explanation:}
The parts of the evidence with the most impact on my predicted verdict
are: Evidence 1 indicates that vitamin D reduces COVID-19 severity,
which opposes the claim that vitamin D increases mortality; Evidence 2
links untreated vitamin D deficiency with higher COVID-19 risk, which
also runs against the claim; Evidence 1's indication of benefits from
adequate vitamin D is consistent with Evidence 2's association between
deficiency and risk, jointly weighing against the claim.
\tabularnewline

\midrule

\textbf{NEW INSTANCE}\par
\textbf{Claim:} \{CLAIM\}
\textbf{Evidence 1:} \{E1\}
\textbf{Evidence 2:} \{E2\}
\textbf{Your answer:}
\tabularnewline

\bottomrule
\end{tabularx}

\endgroup
\end{table}

\begin{table}[p]
\centering
\vspace*{-2mm}
\captionsetup{
    position=top,
    aboveskip=0pt,
    belowskip=0pt
}
\caption{
Uncertainty explanation generation prompt. Adapted from Table~11 of \citet{warren2026show} for space.}
\label{tab:prompt_uncertainty_explanation}
\begingroup
\fontsize{8.3pt}{9.2pt}\selectfont
\setlength{\parskip}{0.5pt}
\setlength{\tabcolsep}{0pt}
\setlength{\parindent}{0pt}
\setlength{\emergencystretch}{1em}
\setlength{\abovetopsep}{0pt}
\renewcommand{\arraystretch}{1.0}

\begin{tabularx}{0.98\linewidth}{
    @{}
    >{\raggedright\arraybackslash}X
    @{}
}
\toprule

\textbf{Task Description:}
You are a helpful assistant. Here are your tasks:
1. Supply the verdict for the claim based on the relationship between
the claim and the two evidence passages.
2. Explain your certainty in the verdict prediction by referring to
the three span interactions provided below---Claim--Evidence 1,
Claim--Evidence 2, and Evidence 1--Evidence 2---and describing how
each interaction's relation (agree, disagree, or unrelated) affects
your overall certainty.
\par
\textbf{Return format:}
[Verdict Prediction] [Explanation]
\tabularnewline

\midrule

\textbf{SHOT 1 Input}\par
\textbf{Claim:}
Winston Churchill once stated, ``A pessimist sees the difficulty in
every opportunity; an optimist sees the opportunity in every
difficulty.''\par
\textbf{Evidence 1:}
``A pessimist,'' says Rev.\ F. W. Cole, president of the Cardiff
Free Church Council, \ldots{}\par
\textbf{Evidence 2:}
Winston Churchill's quote, ``A pessimist sees the difficulty in every
opportunity; an optimist sees the opportunity in every difficulty,''
\ldots{}\par
\textbf{Span interactions (pre-filled):}\par
(1) ``Winston Churchill once stated'' -- ``says the Rev.\ F. W. Cole''
(C--E1), relation: [disagree];\par
(2) ``Winston Churchill once stated'' -- ``Winston Churchill's quote,
`A pessimist sees the difficulty in every opportunity' ''
(C--E2), relation: [agree];\par
(3) ```A pessimist,' says Rev.\ F. W. Cole, president of the
Cardiff Free Church Council'' -- ``Winston Churchill's quote,
`A pessimist sees the difficulty in every opportunity' ''
(E1--E2), [disagree].
\tabularnewline

\addlinespace[1pt]

\textbf{SHOT 1 Output}\par
\textbf{Verdict Prediction:} False
\quad
\textbf{Uncertainty Explanation:}
The parts of the evidence with the most impact on my certainty in the
prediction are: the claim's phrase ``Winston Churchill once stated''
conflicts with Evidence 1's attribution to ``the Rev.\ F. W. Cole,''
which decreases my certainty in the claim; the same claim wording
matches Evidence 2, which introduces the line as ``Winston Churchill's
quote,'' increasing my certainty; since one passage attributes the
quote to Rev.\ Cole and the other to Churchill, the evidence passages
disagree with each other, and that inconsistency lowers my overall
certainty and supports predicting False.
\tabularnewline

\midrule

\textbf{SHOT 2 Input}\par
\textbf{Claim:}
During the last hundred years, the temperature increased by about
0.1$^\circ$C because of carbon dioxide. The human contribution was
about 0.01$^\circ$C.\par
\textbf{Evidence 1:}
\ldots{} well-mixed greenhouse gases (GHGs) contributed a warming of
1.0$^\circ$C to 2.0$^\circ$C \ldots{}\par
\textbf{Evidence 2:}
\ldots{} ``0.1$^\circ$C warming is from CO2 \ldots{} cannot be
supported.'' \ldots{}\par
\textbf{Span interactions (pre-filled):}\par
(1) ``temperature increased by about 0.1$^\circ$C because of carbon
dioxide'' -- ``well-mixed greenhouse gases (GHGs) contributed a
warming of 1.0$^\circ$C to 2.0$^\circ$C''
(C--E1), relation: [disagree];\par
(2) ``0.1$^\circ$C warming is from CO2'' -- ``0.1$^\circ$C warming is
from CO2 \ldots{} cannot be supported''
(C--E2), relation: [disagree];\par
(3) ``well-mixed greenhouse gases \ldots{} 1.0$^\circ$C to
2.0$^\circ$C'' -- ``0.1$^\circ$C warming is from CO2 \ldots{} cannot
be supported''
(E1--E2), relation: [agree].
\tabularnewline

\addlinespace[1pt]

\textbf{SHOT 2 Output}\par
\textbf{Verdict Prediction:} False
\quad
\textbf{Uncertainty Explanation:}
The parts of the evidence with the most impact on my certainty in the
prediction are: the claim's ``about 0.1$^\circ$C because of carbon
dioxide'' contrasts with Evidence 1's ``1.0$^\circ$C to
2.0$^\circ$C'' contribution from well-mixed greenhouse gases, which
pushes my certainty toward False; Evidence 2 explicitly states that
``0.1$^\circ$C warming is from CO2 \ldots{} cannot be supported,''
further moving my certainty toward False; together, the larger range
in Evidence 1 and the rejection of the 0.1$^\circ$C figure in
Evidence 2 reinforce each other, increasing my certainty in the False
prediction.
\tabularnewline

\midrule

\emph{SHOT 3 omitted for space; its structure is identical to that of
SHOT 1 and SHOT 2.}
\tabularnewline

\midrule

\textbf{NEW INSTANCE}\par
\textbf{Claim:} \{CLAIM\} 
\textbf{Evidence 1:} \{E1\}
\textbf{Evidence 2:} \{E2\}
\textbf{Span interactions (pre-filled):}\par
(1) ``\{SPAN1-A\}'' -- ``\{SPAN1-B\}''
(C--E1), relation: \{REL1\};\par
(2) ``\{SPAN2-A\}'' -- ``\{SPAN2-B\}''
(C--E2), relation: \{REL2\};\par
(3) ``\{SPAN3-A\}'' -- ``\{SPAN3-B\}''
(E1--E2), relation: \{REL3\}.
\textbf{Your answer:}
\tabularnewline

\bottomrule
\end{tabularx}

\endgroup

\end{table}

\subsection{Task Instructions}
\label{app:instructs}

\subsection*{Screen 1}
\textbf{Welcome to the experiment!}\\
\textbf{What is this study about?}\\
This study is about AI systems and fact-checking.
Fact-checking is the process of verifying the truthfulness of claims, statements, or stories by comparing them against credible evidence and sources.

\noindent In this study, you will see examples of an AI system designed to help people with fact-checking. Your task will be to carefully read and evaluate the information provided by the AI system.

\noindent You will see and evaluate 8 different claims in total. 
After you have evaluated these claims, you will answer some questions about your experiences of and opinions on the AI system and demographics.

\noindent You will have limited time to evaluate each claim. The main study will take 25-30 minutes to complete, while the questionnaire will take 5-10 minutes.

\noindent Before starting, make sure your browser window is set to full screen. Please make sure you are sitting somewhere comfortable and quiet so that you can complete the experiment in one sitting!

\noindent Click `Next' to read the instructions for the task.

\subsection*{Screen 2}
\textbf{What do I have to do?} \\
Imagine your job is to release assessments about whether a claim is true or false. You will see claims, an AI system’s prediction about whether this claim is true or false, how certain the system is about its label, and an explanation of the AI system’s prediction. These explanations are intended to help you decide how to interpret the AI prediction. You have the option to see the evidence that the AI system used to make its prediction. Your task is to decide whether you can rely on the AI system prediction or need to do more research before releasing the assessment of the claim. \\

\noindent\textbf{Important: }
Please only consider the provided information — claim, verdict, uncertainty, explanations \& evidence documents (if applicable) — when evaluating explanations. Sometimes you will be familiar with the claim, but we ask you to approach each claim as new, whether or not you have seen it before. It doesn’t matter whether you personally agree or disagree with the claim or evidence. We are asking you to evaluate what the AI produces: if you were to see this claim for the first time, would you find the explanation provided by the AI useful? Would you be willing to rely on the AI verdict to make a statement about the claim? 

\subsection*{Screen 3}
\textbf{About the AI System}\\
The AI System you will evaluate in this study is based on a Large Language Model (LLM).
This AI System has been designed to help people with fact-checking.
It has been trained to predict whether the claim is true or false based on the evidence it has been provided.
The AI System also provides explanations for its outputs, which you will evaluate in this study.

\noindent \textbf{What information will I see? (Definitions)}

\begin{itemize}
    \item A \textbf{claim} is some statement about the world. It may be true, false, or somewhere in between.
    \item \textbf{Evidence} is additional information is typically necessary to verify the truthfulness of a claim. An evidence document consists of one or several sentences extracted from an external source for the particular claim. In this study, you will have the option of reading two evidence document excerpts that have been retrieved for a claim. These evidence document excerpts may or may not agree with each other. Please assume that all evidence documents are reasonably reliable. 
    \item A \textbf{verdict} is reached based on the available evidence, regarding whether a claim is true or false.
    \item A \textbf{certainty score} indicates how confident an AI system is about the correctness of its prediction. For example, a Certainty score of 99\% indicates very high confidence that the predicted verdict is correct, but a score of 1\% indicates very low confidence that the predicted verdict is correct. 
    \item An \textbf{explanation} for the AI system’s prediction explains the sources of uncertainty regarding the verdict. In other words, the explanation highlights parts of the evidence that increase or reduce its certainty that its verdict is correct.
    \item Your \textbf{TASK} is to decide whether you can rely on the verdict about the claim produced by the AI system, based on the information available, and how confident you are in your judgment, and decide how helpful the explanation is in making your decision. 
\end{itemize}

\noindent On the next page, you will see an example of the task.

\subsection*{Screen 4}

\begin{figure}[ht]
    \centering
    \includegraphics[width=.9\linewidth]{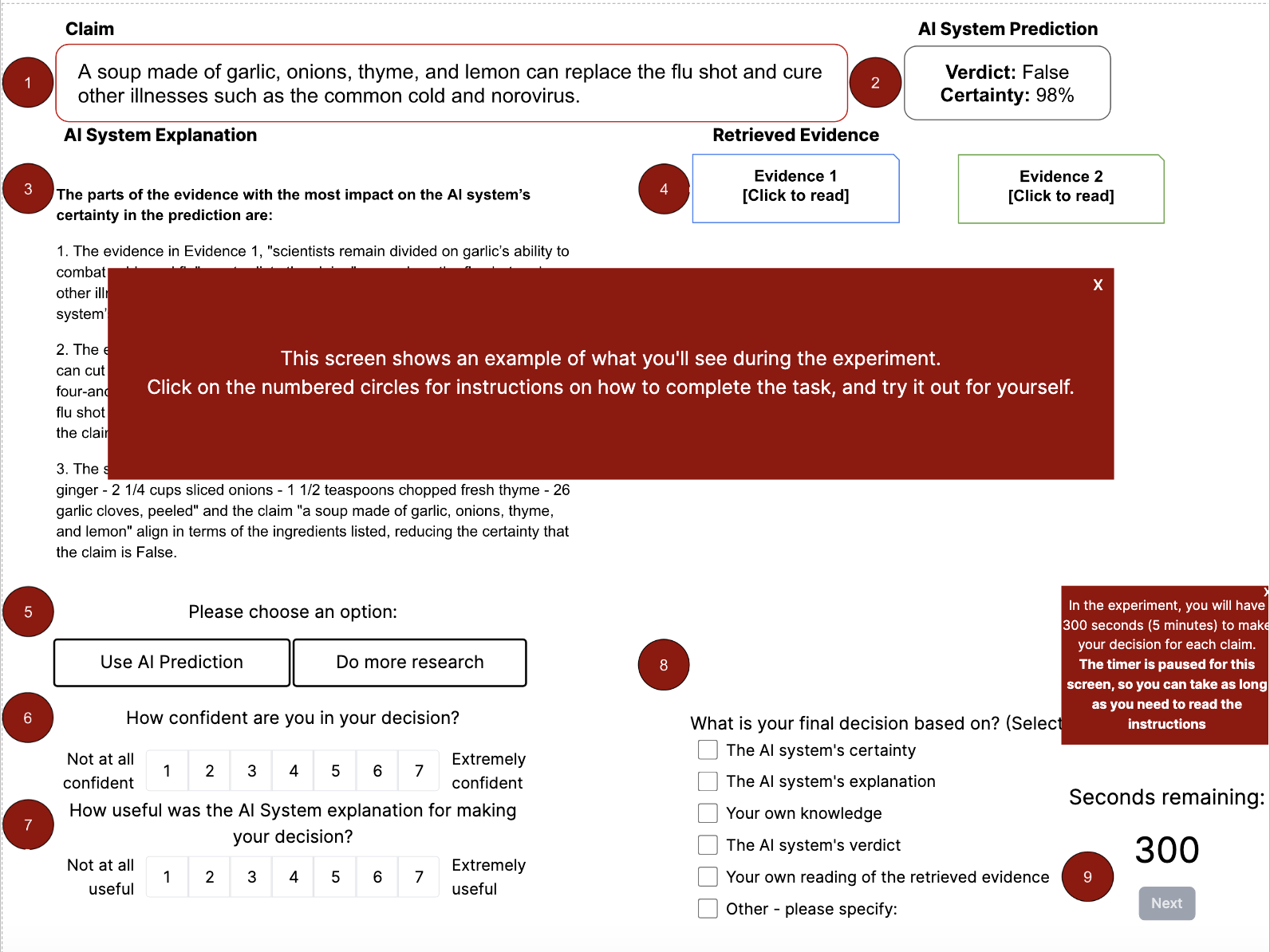}
    \label{fig:app:instruct_example}
\end{figure}

\begin{enumerate}
    \item First, read the claim that the AI system has been given.
    \item Next, read the verdict output by the AI system, along with its certainty that the output is correct.
    \item Next, read the explanation for the AI system's output. Here,  the explanation highlights parts of the evidence that increase or reduce its certainty that its verdict is correct.
    \item If you want to, you can check the evidence that the AI system used to make its prediction, by clicking on `Evidence 1' and/or `Evidence 2'.
    \item Based on the information provided, decide whether you can rely on the verdict produced by the AI system. Try to disregard any prior knowledge or opinions you may have and focus only on the information presented here. Based on the information provided, decide whether you can rely on the verdict produced by the AI system. Try to disregard any prior knowledge or opinions you may have and focus only on the information presented here.
    \item Indicate how confident you are about your decision whether to rely on the AI system verdict or not.
    \item Rate how useful the AI system explanation above was for making your decision. 
    \item Select the sources of information that you based your final decision on. There is no limit to how many you can select. If you used other sources not listed here, you can type them in `Other'.
    \item When you have completed all these steps, press `Next' to move on to the next claim.
You have 300 seconds (5 minutes) to complete your decision for each claim. If the timer runs out before you finish, your response will not be counted.
\end{enumerate}

\subsection*{Screen 5}
Thank you for completing the practice example!\\
\noindent Now, you're ready to start the study. You will see 8 different claims in total.\\
\noindent When you're ready, click `Next' to begin the study.

\subsection{Study Materials}
\label{app:materials}

\subsubsection{Claim 1 (True, Correct, High Certainty)}
\textbf{Claim:} 86\% of Americans and 82\% of gun owners support requiring all gun buyers to pass a background check.

\noindent \textbf{AI Prediction:} True

\noindent \textbf{AI Certainty:} 76\%

\subsubsection*{Evidence 1}
Statistic: 86\% of Americans and 82\% of gun owners support requiring all gun buyers to pass a background check, no matter where they buy the gun and no matter who they buy it from''. [...] One of the survey’s questions was: ``Do you support or oppose requiring a background check on all gun buyers?'' 86 percent of respondents said support, 7 percent said oppose, and 8 percent said unsure, according to the Giffords report on the poll. [...] A March 2019 poll conducted by Quinnipiac University found that 87 percent of national gun owners support ``requiring background checks for all gun owners''. The poll surveyed 1,120 national voters and had a margin of error of plus or minus 3.4 percentage points.

\subsubsection*{Evidence 2}
Iowa voters will decide in 2022 whether to add a state Constitution amendment affirming the right to bear arms. Support for the amendment has fallen along party lines. Opposing Democrats refer to a poll that says an overwhelming number of even gun owners support background checks for gun buyers.

\subsubsection*{Uncertainty Explanation}
The parts of the evidence with the most impact on the AI system’s certainty in the prediction are: \\
1. The evidence in Evidence 1, ``86\% of Americans and 82\% of gun owners support requiring all gun buyers to pass a background check'', aligns perfectly with the claim ``86\% of Americans and 18\% of gun owners support requiring all gun buyers to pass a background check''. This consistency increases the certainty that the claim is true. \\
2. The agreement between the phrase ``gun buyers'' in the claim and ``gun'' in Evidence 1 indicates that the focus on background checks applies uniformly across gun-related transactions and increases the certainty in the prediction. \\
3. Lastly, the phrase ``support'' in the claim and ``owners'' in Evidence 1, while seemingly unrelated in terms of direct content, contribute to the overall coherence of the argument. The emphasis on the level of support among gun owners corroborates the claim's specific statistic, and increases certainty about the alignment between the claim and the provided evidence.

\subsubsection*{Verdict Explanation}
The parts of the evidence with the most impact on the AI system’s predicted verdict are: \\
1. Evidence 1 directly supports the claim, stating that ``86\% of Americans and 82\% of gun owners support requiring all gun buyers to pass a background check, no matter where they buy the gun and no matter who they buy it from''. The poll details show that when asked, ``Do you support or oppose requiring a background check on all gun buyers?'', 86\% responded ``support''. This aligns exactly with the percentages in the claim. \\
2. Evidence 1 cites a Quinnipiac University poll reporting that ``87 percent of national gun owners support requiring background checks for all gun owners'', which reinforces the claim’s assertion of broad support among both the general public and gun owners. \\
3. Evidence 2 does not give exact percentages but echoes the same conclusion. It states that ``an overwhelming number of even gun owners support background checks for gun buyers'', which is consistent with the data presented in Evidence 1. While it is less specific, it corroborates the general finding of widespread support.

\subsubsection*{No Explanation}
The AI system has judged this claim to be True, with 76\% certainty.

\subsubsection{Claim 2 (False, Correct, High Certainty)}
\textbf{Claim:} A satanic-themed hotel is opening in Texas.

\noindent \textbf{AI Prediction:} False

\noindent \textbf{AI Certainty:} 67\%

\subsubsection*{Evidence 1}
Baphomet bedside buddies, upside down crosses, spewing Satan sinks and a devilish welcome, one Facebook user claimed a satanic hotel would soon be opening in Plano, Texas. But don't pull out the rosaries and holy water yet, the hotel is not actually real. On Feb. 23, a Facebook user claimed that a satanic hotel was supposedly opening in an abandoned office building in the heart of the Downtown Plano Art District. [...] But the Baphomet-themed hotel won’t be coming anywhere near Plano. The images were created using an AI art platform. According to The Buzz, an Instagram account posted photos showing how AI art can be made using a specific algorithm. In this case, a satanically-spooky hotel.

\subsubsection*{Evidence 2}
A video shared on Facebook allegedly shows images of a satanic-themed hotel that is set to open in Plano, Texas. [...] A Facebook video allegedly shows photos of a Satanic-themed hotel that is set to open in Plano on June 6. The post shares a photo of a red bedroom with a goat-like figure. ``This is a satanic hotel located in Plano, Texas'' the caption reads. ``This brand new hotel will open June 6 at 6pm (666).'' The claim is fabricated. Check Your Fact found no credible news reports about a satanic-themed hotel opening in Plano, Texas. The photo originates from Ink Poisoning, an apparel brand. The company posted the images on Facebook with a disclaimer stating the images are ``AI concept art created by us''.

\subsubsection*{Uncertainty Explanation}
The parts of the evidence with the most impact on the AI system’s certainty in the prediction are: \\
1. The evidence in Evidence 1, ``Baphomet bedside buddies, upside down crosses, spewing Satan sinks and a devilish welcome, one Facebook user claimed a satanic hotel would soon be opening in Plano, Texas'', is somewhat related to the claim ``A satanic-themed hotel is opening in Texas''. This reduces the AI system’s certainty that the claim is false.\\
2. The evidence in Evidence 1, ``the hotel is not actually real'', is in direct contradiction to the claim ``A satanic-themed hotel is opening in Texas''. This statement increases the certainty that the claim is false by clarifying that the images and descriptions of the hotel are merely AI-generated concepts and not an actual establishment.\\
3. The agreement between the evidence in Evidence 1, that  ``the images were created using an AI art platform'', and the phrase ``supposedly opening'' increase AI system uncertainty that the claim is false.

\subsubsection*{Verdict Explanation}
The parts of the evidence with the most impact on the AI system’s predicted verdict are: \\
1. Evidence 1 makes clear that the idea of a satanic-themed hotel opening in Plano, Texas, comes from a social media post and is not real. It explains that a Facebook user claimed ``a satanic hotel would soon be opening in Plano, Texas'', but the article quickly clarifies that ``the hotel is not actually real''. The supposed images of the ``Baphomet bedside buddies, upside down crosses, spewing Satan sinks and a devilish welcome'' were not photos of a real hotel but instead ``created using an AI art platform''. \\
2. The Buzz further explained that the Instagram account that posted them did so to show ``how AI art can be made using a specific algorithm''. This confirms the images were fictional concept art, not evidence of an actual hotel. \\
3. Evidence 2 supports this conclusion. It describes how a Facebook video claimed ``a satanic-themed hotel… is set to open in Plano on June 6'' and even added details like ``open June 6 at 6pm (666)''. However, the evidence directly states that ``the claim is fabricated'' and that ``no credible news reports'' support the idea of such a hotel opening. It also identifies the true origin of the images: they came from Ink Poisoning, an apparel brand, which shared them with a disclaimer noting they were ``AI concept art created by us''.

\subsubsection*{No Explanation}
The AI system has judged this claim to be False, with 67\% certainty.

\subsubsection{Claim 3 (True, Correct, Low Certainty)}

\textbf{Claim:} A scone can equal a third of one’s recommended daily calories.

\noindent \textbf{AI Prediction:} True

\noindent \textbf{AI Certainty:} 35\%

\subsubsection*{Evidence 1}
An average scone equals a fifth of recommended daily calories for females, and a sixth for males. The number of calories tends to be proportionate to the size of a scone rather than the luxuriousness of its filling.

\subsubsection*{Evidence 2}
On 11 February 2019, a headline of the Irish News and the Belfast Telegraph stated that ``One large scone can equal a third of recommended daily calories''. Both articles used a Press Association report that referred to ``a survey''. [...] The report states that ``the highest calorie scone, without the addition of spread or jam, provides over a third (38\%) of the recommended daily calorie intake''. [...] For people aged 19 to 64, the Government Dietary Recommendations suggest a daily intake of 2,000kcal for females and 2,500kcal for males. The largest scone of the survey, then, indeed equals a third of recommended calories; taking up 37.8\% of a daily female diet, and 30.2\% of a daily male diet.

\subsubsection*{Uncertainty Explanation}
The parts of the evidence with the most impact on the AI system’s certainty in the prediction are:
1. The evidence in Evidence 1, ``An average scone equals a fifth of recommended daily calories for females, and a sixth for males'', aligns with the claim ``A scone can equal a third of one’s recommended daily calories''. Although the specific fractions differ, the core message that a scone can be a substantial contributor to daily calorie consumption is consistent, increasing the certainty that the claim is true.\\
2. The evidence in Evidence 2, ``without the addition of spread or jam, provides over a third (38\%) of the recommended daily calorie intake'', corroborates the claim by providing a concrete example where a larger scone meets or exceeds the claim's assertion. This specific instance strengthens the certainty that the claim is true.\\
3. The evidence in Evidence 2, ``the largest scone of the survey, then, indeed equals a third of recommended calories; taking up 37.8\% of a daily female diet, and 30.2\% of a daily male diet'', further reinforces the certainty that the claim is true by offering precise figures that closely match the claim's statement.

\subsubsection*{Verdict Explanation}
The parts of the evidence with the most impact on the AI system’s predicted verdict are:
1. Evidence 1 states that an average scone provides ``a fifth of recommended daily calories for females, and a sixth for males''. This shows that even a regular-sized scone contains a significant share of daily energy needs.\\
2. Evidence 1 also emphasizes that calorie content is more closely tied to the size of the scone than to its filling, suggesting that larger scones will contribute an even higher proportion of daily calories.\\
3. Evidence 2 confirms that in some cases a single scone can equal about a third of recommended calories. It explains that ``the highest calorie scone, without the addition of spread or jam, provides over a third (38\%) of the recommended daily calorie intake''. This is further broken down to ``37.8\% of a daily female diet, and 30.2\% of a daily male diet'', based on standard intake levels of 2,000 kcal for women and 2,500 kcal for men.

\subsubsection*{No Explanation}
The AI system has judged this claim to be True, with 35\% certainty.

\subsubsection{Claim 4 (False, Correct, Low Certainty)}

\textbf{Claim:} A windmill could spin until it falls apart and never generate as much energy as was invested in building it.

\noindent \textbf{AI Prediction:} False

\noindent \textbf{AI Certainty:} 31\%

\subsubsection*{Evidence 1}
No, a professor didn't say a windmill will `never generate' the energy invested in building it [...] Now, we have a meme roaming social media that claims Canadian author and professor Thomas Homer-Dixon said that windmills will ``never'' generate as much energy as was expended to build the structures. [...] The meme does get a part of the quote right, but it strategically leaves out large swaths of information in order to give people the impression that all windmills never expend the amount of energy that was invested in it, which is not accurate. The meme's omitted section explains that, while poorly placed windmills may never generate enough energy payback, a good wind site could generate it in three years or less.

\subsubsection*{Evidence 2}
Factually Inaccurate: Analyzing the energy used in the production of wind turbines does not show that it's greater than the electricity produced over the turbine's working lifetime.

\subsubsection*{Uncertainty Explanation}
The parts of the evidence with the most impact on the AI system’s certainty in the prediction are:\\
1. The evidence in Evidence 1, ``No, a professor didn't say a windmill will `never generate' the energy invested in building it'', directly contradicts the claim's assertion that windmills will never generate sufficient energy. This contradiction increases the AI system’s certainty that the claim is False.\\
2. The evidence in Evidence 1, ``will `never generate' the energy invested'', aligns with the part of the claim ``spin until it falls apart and never generate as much energy as was invested'', reducing the AI system’s certainty that the claim is False. \\
3. The evidence in Evidence 1, ``expended to'', aligns with the phrase ``invested in building'' in the claim, indicating agreement on the energy expenditure aspect, reducing the certainty that the claim is False.

\subsubsection*{Verdict Explanation}
The parts of the evidence with the most impact on the AI system’s predicted verdict are:\\
1. Evidence 1 explains that the claim comes from a misrepresented quote. The meme states that windmills ``will never generate as much energy as was expended to build the structures'', but the original quote clarifies that ``while poorly placed windmills may never generate enough energy payback, a good wind site could generate it in three years or less''. This shows the claim exaggerates and omits important context.\\
2. The same evidence highlights that well-sited windmills can generate enough energy to offset their construction cost quickly, making the claim that all windmills `never generate' enough energy inaccurate. The phrase ``a good wind site could generate it in three years or less'' directly contradicts the claim.\\
3. Evidence 2 reinforces this, stating that ``analyzing the energy used in the production of wind turbines does not show that it’s greater than the electricity produced over the turbine's working lifetime''. This confirms that, over time, turbines produce more energy than was invested in building them.

\subsubsection*{No Explanation}
The AI system has judged this claim to be False, with 31\% certainty.

\subsubsection{Claim 5 (True, Incorrect, High Certainty)}

\textbf{Claim:} All tea bags contain harmful microplastics.

\noindent \textbf{AI Prediction:} True

\noindent \textbf{AI Certainty:} 73\%

\subsubsection*{Evidence 1}
At ORGANIC INDIA, we believe in plastic-free tea bags to protect your body and the earth; and preserve the pure, organic tea experience that you deserve. Unfortunately, the industry norm is to include plastic in the mesh tea bag, which releases billions of microplastics and nanoplastics into your infusion. Consuming microplastics, as you can imagine, can have negative impacts on your health and on the earth. [...] Yes, but not all tea bags. The vast majority of brands on the shelves have mesh tea bags that are composed of 20-30\% plastic. According to studies, a single standard tea bag releases 11.6 billion microplastics and 3.1 billion nanoplastics into every cup of tea. Consumers are increasingly becoming aware of this fact and opting either for loose teas, or selecting tea brands that have no plastic in the teabag. ORGANIC INDIA tea bags do not contain plastic, and are made instead of unbleached filter paper that is verified home compostable, biodegradable, Non-GMO, allergen-free and chemical-free. [...] In short, yes. Microplastics in tea bags can be harmful not only for the environment, but for human health as well. Microplastics may enter the human system through ingestion, inhalation or dermal absorption. Studies have shown that microplastics have potential to cause damage to the cells in the quantities currently found in food and beverages like tea.

\subsubsection*{Evidence 2}
Therefore, the use of plastic in tea bags has come under scrutiny due to the potential release of microplastics when exposed to heat. As we mentioned earlier, many tea bag manufacturers use nylon or polyethylene terephthalate (PET) in their tea bags, which are not biodegradable and can release harmful microplastics when exposed to heat. These microplastics can end up in your drink, and studies have shown that they may pose a risk to human health. [...] Another study published in the journal Environmental Science \& Technology in 2020 found that plastic tea bags could release up to 16 micrograms of microplastics per liter of tea when brewed at a temperature of 95°C. These studies highlight the importance of choosing plastic-free tea bags to minimize exposure to potentially harmful plastics. [...] Yes, some tea bags can contain microplastics, particularly those made with plastic materials like nylon or polypropylene. However, The Tea Spot's sachets are made from sugar cane fibers, which are biodegradable, compostable, and completely free from petroleum-based plastic. Therefore, you can rest assured that when you brew our tea, you're not consuming any harmful plastics or microplastics.

\subsubsection*{Uncertainty Explanation}
The parts of the evidence with the most impact on the AI system’s certainty in the prediction are:\\
1. The statement in Evidence 2, ``Therefore, the use of plastic in tea bags has come under scrutiny due to the potential release of microplastics when exposed to heat'', does not support the blanket statement that ``All tea Bags contain harmful microplastics''. This disagreement reduces the AI system certainty that the claim is True.\\
2. The statement in Evidence 1, ``Yes, but not all tea bags. The vast majority of brands on the shelves have mesh tea bags that are composed of 10-30\% plastic'', contradicts the claim by indicating that not all tea bags contain harmful microplastics, reducing certainty that the claim is True.\\
3. The statement in Evidence 1, ``ORGANIC INDIA tea bags do not contain plastic, and are made instead of unbleached filter paper that is verified home compostable, biodegradable, Non-GMO, allergen-free and chemical-free'', undermines the claim by providing a specific example of a brand that produces tea bags without harmful microplastics, further reducing certainty that the claim is True.

\subsubsection*{Verdict Explanation}
The parts of the evidence with the most impact on the AI system’s predicted verdict are:\\
1. Evidence 1 states that ``the vast majority of brands on the shelves have mesh tea bags that are composed of 20–30\% plastic''. This shows that most commercially available tea bags contain plastic, making the exposure to microplastics nearly universal for consumers using standard tea bags.\\
2. Evidence 1 notes that ``a single standard tea bag releases 11.6 billion microplastics and 3.1 billion nanoplastics into every cup of tea'', and Evidence 2 confirms that materials like nylon or PET ``can release harmful microplastics when exposed to heat''. Both sources directly link the presence of plastic in tea bags to microplastic contamination in brewed tea.\\
3. Evidence 1 explains that microplastics ``can be harmful not only for the environment, but for human health as well'' and ``may cause damage to the cells in the quantities currently found in food and beverages like tea''. Evidence 2 echoes this, stating that microplastics ``may pose a risk to human health''.

\subsubsection*{No Explanation}
The AI system has judged this claim to be True, with 73\% certainty.

\subsubsection{Claim 6 (False, Incorrect, High Certainty)}

\textbf{Claim:} Lagan Valley is one of the least wooded areas in Northern Ireland, particularly with regards to ancient woodland.

\noindent \textbf{AI Prediction:} False

\noindent \textbf{AI Certainty:} 78\%

\subsubsection*{Evidence 1}
The Lagan Valley (Irish: Cluain an Lagáin, Ulster Scots: Glen Lagan) is an area of Northern Ireland between Belfast and Lisburn. The Lagan is a famous river that flows into Belfast Lough. For a section, the river forms part of the border between the counties of Antrim and Down. It has a number of interesting features including a towpath which runs alongside the River Lagan. The towpath is popular with walkers, runners, cyclists, dog owners etc. It is a very scenic and peaceful area and is ideal for walking, cycling etc. The towpath begins in the Stranmillis area of south Belfast and runs all the way to Lisburn. The cycle route forms part of National Cycle Route 9. There are a number of ``off route'' mountain bike trails along the towpath. Lagan Valley is an Area of Outstanding Natural Beauty (AONB). The AONB was established in 1965 and the greater part of it lies within the Greater Belfast area. [...] Belvoir Park Forest Education Centre 

\subsubsection*{Evidence 2}
Therefore, it is fair to say that Lagan Valley is one of the least wooded areas in NI. [...] Lisburn \& Castlereagh is the least densely wooded council (3.7\%) in Northern Ireland, followed by Armagh, Banbridge \& Craigavon (4.0\%) and Antrim \& Newtownabbey (4.4\%). [...] Based on all this, Ms Eastwood’s claim that ``Lagan Valley is one of the least wooded areas in NI'' is backed by evidence – although the precise figures are available broken down by council areas, rather than Westminster/Assembly constituencies.

\subsubsection*{Uncertainty Explanation}
The parts of the evidence with the most impact on the AI system’s certainty in the prediction are:\\
1. The evidence in Evidence 1, ``Lagan Valley is an Area of Outstanding Natural Beauty (AONB)'', does not directly address the claim about the level of woodlands in the area, reducing the certainty of the AI system’s prediction.\\
2. The evidence in Evidence 2, ``Therefore, it is fair to say that Lagan Valley is one of the least wooded areas in NI'', directly supports the claim that Lagan Valley is one of the least wooded areas in Northern Ireland, particularly with regards to ancient woodland. This agreement reduces the certainty that the claim is False.\\
3. The evidence in Evidence 2 that ``Lisburn \& Castlereagh is the least densely wooded council (3.7\%) in Northern Ireland'' aligns with the evidence in Evidence 1 that the Lagan Valley ``is an area of Northern Ireland between Belfast and Lisburn''. This reduces the certainty that the claim is False.

\subsubsection*{Verdict Explanation}
The parts of the evidence with the most impact on the AI system’s predicted verdict are:\\
1. Evidence 1 describes Lagan Valley as ``an Area of Outstanding Natural Beauty (AONB)'' and mentions that it ``has a number of interesting features including a towpath which runs alongside the River Lagan''. It also emphasizes that the area is ``very scenic and peaceful'' and ``ideal for walking, cycling etc.'', suggesting the presence of natural green spaces rather than a sparsely wooded landscape.\\
2. Evidence 2 notes that ``Lisburn \& Castlereagh is the least densely wooded council (3.7\%) in Northern Ireland, followed by Armagh, Banbridge \& Craigavon (4.0\%) and Antrim \& Newtownabbey (4.4\%)''. It adds that Ms Eastwood’s claim that ``Lagan Valley is one of the least wooded areas in NI'' is ``backed by evidence – although the precise figures are available broken down by council areas, rather than Westminster/Assembly constituencies'', showing that the claim is not directly supported for Lagan Valley itself.
3. Neither piece of evidence provides specific data about ancient woodland in Lagan Valley. The lack of precise figures for the area means the claim about ancient woodland cannot be verified.

\subsubsection*{No Explanation}
The AI system has judged this claim to be False, with 78\% certainty.

\subsubsection{Claim 7 (True, Incorrect, Low Certainty)}

\textbf{Claim:} Cristiano Ronaldo was the only European soccer captain who did not wear a rainbow armband at Euro 2020.

\noindent \textbf{AI Prediction:} True

\noindent \textbf{AI Certainty:} 29\%

\subsubsection*{Evidence 1}
Cristiano Ronaldo is refusing to wear the ``One Love'' armband, which advocates for the LGBTQ+ community, during the 2022 World Cup. The world-renowned soccer player, Cristiano Ronaldo, revealed in a press conference that he will not be participating in wearing the ``One Love'' armband. The armband consists of a heart with a rainbow inside of it and the number one. This ``One Love'' band is being used to advocate inclusivity, being that the 2022 World Cup is being held in Qatar, a country where same-sex relationships are illegal. [...] Players who wear the armband are threatened with receiving a yellow card, according to FIFA’s regulations. This yellow card can result in a suspension from the games if the player were to receive another yellow card during the game. Obviously, any mention of social issues is discouraged on the field. Due to the consequences, none of the captains from any of the other national teams have worn the armband, but none have faced backlash like Ronaldo has.

\subsubsection*{Evidence 2}
There is evidence that three team captains — from Germany, England, and Denmark — wore rainbow or pro-LGBTQ+ armbands during this tournament, which included 24 teams. Germany’s captain, Manuel Neuer, was briefly investigated by the Union of European Football Associations for his rainbow armband. Cristiano Ronaldo was among many other captains who wore normal armbands.

\subsubsection*{Uncertainty Explanation}
The parts of the evidence with the most impact on the AI system’s certainty in the prediction are:\\
1. The evidence in Evidence 1, ``team captains — from Germany, England, and Denmark — wore rainbow or pro-LGBTQ+ armbands during this tournament'', is not consistent with the part of the claim ``Cristiano Ronaldo was the only European soccer captain who did not wear a rainbow armband at Euro 2020''. This indicates that there were multiple captains who did not wear the armband, not just Ronaldo and reduces AI system certainty that the claim is True.\\
2. The evidence in Evidence 1, ``Cristiano Ronaldo was among many other captains who wore normal armbands'', is not consistent with the part of the claim ``Cristiano Ronaldo was the only European soccer captain.'', implying he was not the only captain without a rainbow armband, reducing certainty that the claim is True.\\
3. The evidence in Evidence 1, ``during this tournament'', is not consistent with the part of the claim ``Euro 2020'', as it refers to a different event. This discrepancy highlights that the evidence discusses a different tournament, possibly the 2022 World Cup, rather than Euro 2020 and reduces certainty that the claim is True.

\subsubsection*{Verdict Explanation}
The parts of the evidence with the most impact on the AI system’s predicted verdict are:\\
1. Evidence 1 notes that Ronaldo ``is refusing to wear the `One Love' armband, which advocates for the LGBTQ+ community,'' and describes how he revealed in a press conference that he ``will not be participating in wearing the `One Love' armband.'' The evidence emphasizes that while ``players who wear the armband are threatened with receiving a yellow card'', Ronaldo’s decision not to wear it drew notable attention, suggesting he stood out among team captains. The passage also clarifies that ``none of the captains from any of the other national teams have worn the armband'' at other tournaments, highlighting Ronaldo’s high-profile refusal.\\
2. Evidence 2 provides additional support, stating that ``three team captains — from Germany, England, and Denmark — wore rainbow or pro-LGBTQ+ armbands during this tournament''. It also specifically mentions that ``Cristiano Ronaldo was among many other captains who wore normal armbands'', directly confirming that he did not wear the rainbow armband while other European captains did.

\subsubsection*{No Explanation}
The AI system has judged this claim to be True, with 29\% certainty.

\subsubsection{Claim 8 (False, Incorrect, Low Certainty)}

\textbf{Claim:} The top issue for college students is the economy.

\noindent \textbf{AI Prediction:} False

\noindent \textbf{AI Certainty:} 49\%

\subsubsection*{Evidence 1}
``The everyday concerns of Americans are the economy — even young people that we work with at Young America's Foundation, our polling shows nationwide, their top issue for college students is the economy'', Walker said. [...] Is Walker right that ``The top issue for college students is the economy''? [...] The economy was the top issue for college students who participated in the group's poll. Wisconsin pollster Charles Franklin said the polling company is reputable and the questions were overall evenhanded.

\subsubsection*{Evidence 2}
The midterm election is right around the corner and the economy and abortion rights are the top issues of concern for college students according to a new survey by BestColleges.com. These issues are also ranked as the most important for Americans overall. BestColleges.com surveyed over 1000 students about their political beliefs and plans to vote. Thirty-seven percent (37\%) of those surveyed noted the economy, inflation, and employment as the most important issues in this election. Students are worried about finding jobs, their future finances, and the growing cost of everything from housing to peanut butter. White college men, in particular, are overwhelmingly concerned about the economy, employment, and inflation, with almost half (49\%) selecting these three issues as their top political issues — a significantly higher percentage than other demographic groups surveyed. Abortion rights, which have been the subject of heated national and local political campaigns, were ranked as a top concern by college students with 33\% of those surveyed ranking these rights as a top issue. College women were particularly concerned about legislation having an impact on their bodies. While college men chose the economy as their top issue (44\%), abortion rights were the most important issue for college women (43\%), with women of color expressing the deepest concern. Only 21\% of college men considered abortion rights an important issue. For college men, the economy, gun policy, healthcare, racial and ethnic inequality, and climate change all came out ahead of abortion rights as leading issues.

\subsubsection*{Uncertainty Explanation}
The parts of the evidence with the most impact on the AI system’s certainty in the prediction are:\\
1. The evidence in Evidence 1, ``The economy was the top issue for college students who participated in the group’s poll'', is consistent with the claim ``The top issue for college students is the economy'', reducing AI system certainty that the claim is False.
2. The evidence in Evidence 2, ``abortion rights were the most important issue for college women (43\%)'' contrasts with the claim, increasing certainty that the claim is False.\\
3. The statement in Evidence 1, ``The everyday concerns of Americans are the economy'', is consistent with the part of Evidence 2, ``Students are worried about finding jobs, their future finances, and the growing cost of everything'',  reducing AI system certainty that the claim is False.

\subsubsection*{Verdict Explanation}
The parts of the evidence with the most impact on the AI system’s predicted verdict are:\\
1. Evidence 1 reports that, according to polling by Young America’s Foundation, ``the economy was the top issue for college students who participated in the group’s poll''. While this appears to support the claim, the evidence also raises the question: ``Is Walker right that `The top issue for college students is the economy'?'' suggesting uncertainty about whether this finding is representative of all college students.\\
2. Evidence 2 provides more detailed data, showing that college students are concerned about multiple issues. It states that ``the economy and abortion rights are the top issues of concern for college students'', with 37\% ranking the economy and 33\% ranking abortion rights as their primary concern. The evidence further breaks this down by gender: ``college men chose the economy as their top issue (44\%)'', whereas ``abortion rights were the most important issue for college women (43\%), with women of color expressing the deepest concern''. This demonstrates that the economy is not universally the top issue across all college students.

\subsubsection*{No Explanation}
The AI system has judged this claim to be False, with 49\% certainty.





\addtocontents{toc}{\vspace{1\baselineskip}}
\mymiscpagestyle{}

\clearpage
\addcontentsline{toc}{section}{Bibliography}
{\normalsize\bibliography{references_all}}

\end{document}